\documentclass{article}

\usepackage[english]{babel}

\usepackage[letterpaper,top=2cm,bottom=2cm,left=3cm,right=3cm,marginparwidth=1.75cm]{geometry}

\usepackage{graphicx}
\usepackage{nicefrac}
\usepackage{booktabs,longtable,array,tabularx,multirow,hhline}
\usepackage{algorithm,algpseudocode,enumitem,xcolor,xurl}
\usepackage{colortbl,pifont}

\usepackage[normalem]{ulem}
\definecolor{secThreePurple}{RGB}{112,48,160}

\usepackage{amsmath,amsfonts,amssymb,amsthm,mathtools,bm}
\usepackage{aliascnt}
\usepackage{bbm}
\usepackage{dsfont}
\usepackage{isomath}
\DeclareMathSymbol{\Gamma}{\mathalpha}{operators}{"00}
\DeclareMathSymbol{\Delta}{\mathalpha}{operators}{"01}
\DeclareMathSymbol{\Theta}{\mathalpha}{operators}{"02}
\DeclareMathSymbol{\Lambda}{\mathalpha}{operators}{"03}
\DeclareMathSymbol{\Xi}{\mathalpha}{operators}{"04}
\DeclareMathSymbol{\Pi}{\mathalpha}{operators}{"05}
\DeclareMathSymbol{\Sigma}{\mathalpha}{operators}{"06}
\DeclareMathSymbol{\Upsilon}{\mathalpha}{operators}{"07}
\DeclareMathSymbol{\Phi}{\mathalpha}{operators}{"08}
\DeclareMathSymbol{\Psi}{\mathalpha}{operators}{"09}
\DeclareMathSymbol{\Omega}{\mathalpha}{operators}{"0A}
\allowdisplaybreaks

\def\eqref#1{(\ref{#1})}

\def\1{\bm{1}}

\DeclareMathAlphabet{\mathsfit}{\encodingdefault}{\sfdefault}{m}{sl}
\SetMathAlphabet{\mathsfit}{bold}{\encodingdefault}{\sfdefault}{bx}{n}

\def\gA{{\mathcal{A}}}
\def\gB{{\mathcal{B}}}
\def\gC{{\mathcal{C}}}

\def\gF{{\mathcal{F}}}
\def\gG{{\mathcal{G}}}

\def\gK{{\mathcal{K}}}

\def\gO{{\mathcal{O}}}
\def\gP{{\mathcal{P}}}

\def\gW{{\mathcal{W}}}

\def\sP{{\mathbb{P}}}
\def\sQ{{\mathbb{Q}}}
\def\sR{{\mathbb{R}}}

\newcommand{\E}{\mathbb{E}}

\newcommand{\R}{\mathbb{R}}

\newcommand{\KL}{D_{\mathrm{KL}}}

\DeclareMathOperator*{\argmin}{arg\,min}

\DeclareMathOperator{\Tr}{Tr}

\newcommand{\Reg}[1][T]{\operatorname{Reg}_{#1}}

\newcommand{\cA}{\mathcal A}
\newcommand{\cF}{\mathcal F}

\newcommand{\ind}{\mathbbm 1}
\newcommand{\Conv}{\operatorname{Conv}}
\newcommand{\norm}[1]{\left\lVert #1\right\rVert}
\newcommand{\ip}[2]{\left\langle #1,#2\right\rangle}

\newcommand{\logp}{\log_+}

\newcommand{\cstar}{c^{*}}
\newcommand{\cctr}{c_{\mathrm{ctr}}^{*}}
\newcommand{\diag}{\operatorname{diag}}
\newcommand{\TV}{\operatorname{TV}}
\newcommand{\kl}{\operatorname{kl}}

\newcommand{\lo}{\mathcal L}
\newcommand{\up}{\mathcal U}
\newcommand{\BB}{\mathbb B^d}

\newcommand{\namedeqref}[2]{\hyperref[#2]{#1~\textup{(\ref*{#2})}}}
\DeclareRobustCommand{\IAFTRL}{\hyperref[alg:G-coherent]{\mbox{\textup{IA-FTRL}}}}
\DeclareRobustCommand{\GIAFTRL}{\hyperref[alg:sharp]{\mbox{\textup{GIA-FTRL}}}}

\newcolumntype{Y}{>{\raggedright\arraybackslash}X}
\newcolumntype{P}[1]{>{\raggedright\arraybackslash}p{#1}}

\definecolor{crossrefcolor}{RGB}{200,55,85}
\usepackage[colorlinks=true, allcolors=blue, linkcolor=crossrefcolor]{hyperref}
\usepackage[nameinlink]{cleveref}

\numberwithin{equation}{section}
\newtheorem{theorem}{Theorem}[section]
\newaliascnt{lemma}{theorem}
\newtheorem{lemma}[lemma]{Lemma}
\aliascntresetthe{lemma}
\newtheorem{proposition}{Proposition}[section]
\newtheorem{corollary}{Corollary}[section]
\theoremstyle{definition}
\newaliascnt{definition}{theorem}
\newtheorem{definition}[definition]{Definition}
\aliascntresetthe{definition}

\title{Best-of-Both-Worlds FTRL for Linear Bandits: The Interplay of Information Acquisition, Corruption, and Geometry}
\author{%
  Canzhe Zhao\textsuperscript{1}\qquad Vincent Y. F. Tan\textsuperscript{1,2}\\[0.7em]
  \small\textsuperscript{1}Department of Mathematics, National University of Singapore\\
  \small\textsuperscript{2}Department of Electrical and Computer Engineering, National University of Singapore\\[0.4em]
  \small\href{mailto:canzhezhao@nus.edu.sg,vtan@nus.edu.sg}{\texttt{\{canzhezhao, vtan\}@nus.edu.sg}}%
}
\date{}
\begin{document}
\maketitle

\begin{abstract}

We study \textit{best-of-both-worlds} (BOBW) algorithms for linear bandits,
which aim to simultaneously guarantee $\gO(\sqrt T)$ regret in the
adversarial regime and $\gO(\log T)$ regret in the stochastic regime over
$T$ rounds. For finite action sets, we introduce
\emph{information-acquisition follow-the-regularized-leader} (\IAFTRL{}), the first BOBW algorithm
based purely on the follow-the-regularized-leader (FTRL) framework that attains an
$\gO(\cstar\log T)$-type regret in the stochastic regime, where $\cstar$ is the
information-acquisition complexity that characterizes instance-dependent
optimality in linear bandits, while retaining the $\gO(\sqrt{T})$ regret guarantee in the adversarial regime. This resolves the open problem posed
in~\cite{LeeEtAl2021,KongZ023,ItoT23,ZhaoItoLi2025}.

Motivated by the existing
$\gO(d\log T/\Delta_{\min}+\sqrt{Cd\log T/\Delta_{\min}})$ regret bound
for BOBW linear bandit algorithms in the corrupted stochastic
regime~\cite{KongZ023,ItoT23,ItoTsuchiyaHonda2024}, where $C$ is the
corruption budget and $\Delta_{\min}$ is the minimum suboptimality gap,
we investigate how corruption affects the $\gO(\cstar\log T)$-type regret in the corrupted stochastic regime.
In particular, 
we ask whether the coefficient in the corruption regret term can be bounded solely in terms of $\cstar$, 
as in the main term $\gO(\cstar\log T)$, yielding an
$\gO(\sqrt{\cstar C})$ corruption-dependent regret term. 
On the positive side, we show that
\IAFTRL{} achieves $ \gO ( \cstar\log K\cdot\log T
       +\sqrt{Cd\log K}\,\Delta_{\min}^{-129/256} )$ regret in the corrupted stochastic regime (ignoring remainder terms independent of $T$ and $C$),
where $K$ is the number of actions and $d$ is the ambient dimension.
On the negative side, for sufficiently large corruption budgets
and time horizons, we establish an
$\Omega(\min\{\sqrt T,\sqrt{C/\Delta_{\min}}\})$ regret lower bound for instances with $\cstar=\Theta(1)$ yet arbitrarily small
$\Delta_{\min}$.
This shows that the multiplicative dependence of the corruption term on $\Delta_{\min}^{-1}$ cannot, in general, be replaced by dependence on $\cstar$ alone.

Finally, we investigate whether analogous guarantees can be extended to infinite
action sets.
On the positive side, 
we identify a subclass of linear bandit instances with infinite action sets, which we call \textit{sharp} linear bandit instances. 
On these instances, the distance from a suboptimal action $a$ to the optimal action $a^*$ decreases faster than the gap of action $a$.
In particular, we propose the \textit{centered} information-acquisition complexity $\cctr$, which extends the original information-acquisition complexity $c^*$ of \cite{LS17} to infinite action sets, and show that sharpness is necessary and sufficient for $\cctr<\infty$.
We further devise a second algorithm, \emph{geometric information-acquisition
FTRL} (\GIAFTRL{}), which achieves
$\gO(d\cctr\log T + J_\theta\sqrt C\logp(C))$ regret
on sharp instances in the corrupted stochastic regime,
up to an additive remainder independent of $T$ and $C$,
while retaining an $\gO(d\sqrt T\log T)$ adversarial guarantee.
Here $J_\theta$ is an instance-dependent coefficient independent
of $T$ and $C$.
On the negative side, we demonstrate the role of action-set geometry
by exhibiting a fixed nonsharp linear bandit instance with an infinite
compact action set and an $\Omega(C^{2/3})$ regret lower bound
when $C\asymp\sqrt T$.
This rules out extending the preceding $\gO(\sqrt{C})$ corruption guarantee
to all instances with infinite action sets.

\end{abstract}

\clearpage
\setcounter{tocdepth}{2}
\tableofcontents
\clearpage

\section{Introduction}\label{sec:introduction}

Linear bandits provide a fundamental model for sequential decision making
with large, or even infinite, action sets.
In this problem, at each round $t$, a learner
selects an action $A_t\in\cA\subset\R^d$ and observes a scalar loss
whose conditional mean is $A_t^\top\theta_t$, where $\theta_t$ is an unknown
loss vector. The linear structure allows an observation from one
action to reveal information about other actions, making learning possible
without exploring every action separately~\cite{AbbasiYadkoriPalSzepesvari2011}.
In this problem, the learner seeks to minimize regret against the best fixed action.
As in other online learning problems, two standard regimes in linear bandits capture how the loss of each action is generated:
in the \emph{stochastic regime}, $\theta_t=\theta$ for all $t\in[T]$ for a fixed unknown
parameter $\theta$, whereas in the \emph{adversarial regime}, $\theta_t$
may vary from round to round and depend on the past history. For adversarial linear bandits with bounded
losses, \cite{BubeckCesaBianchiKakade2012} establish a regret upper bound of
$\gO(\sqrt{dT\log K})$ for a finite set of $K$ actions and of 
$\gO(d\sqrt{T\log T})$ for general compact action sets; the latter
is minimax optimal up to a logarithmic factor without further geometric
assumptions~\cite{BubeckCesaBianchiKakade2012}.

In contrast to the adversarial regime, the stochastic regime can result in logarithmic regret.
Let $a^*$ be a unique optimal action, define
$\Delta(a)=(a-a^*)^\top\theta$, and, for a finite action set, set
$\Delta_{\min}=\min_{a\ne a^*}\Delta(a)$. Canonical multi-armed bandit (MAB)
guarantees have the form
$\gO(\sum_{a\ne a^*}\log T/\Delta(a))$, which is 
$\gO(K\log T/\Delta_{\min})$~\cite{AuerCesaBianchiFischer2002}.
For linear bandits, however, the statistical difficulty depends on the
regret cost of collecting information across the entire action set, 
and thus an appropriate metric must account for information shared across actions.
\cite{LS17} characterize this cost through $\cstar=\cstar(\gA,\theta)$: 
in linear bandits with finite actions,
every consistent learner satisfies
$\liminf_{T\to\infty}\Reg/\log T\ge\cstar$, and an
asymptotically optimal learner attains
$\limsup_{T\to\infty}\Reg/\log T\le\cstar$.
In this work, we refer to $\cstar$ as the \emph{information-acquisition complexity}, which is formally defined in
\eqref{eq:overview-cstar}.
Intuitively, $\cstar$ measures the minimal regret cost of acquiring enough information to distinguish each suboptimal action from the optimal action $a^*$.
On linear bandit instances with informative actions, $\cstar$ can be arbitrarily smaller than the inverse of the minimum gap $1/\Delta_{\min}$.

Since the seminal work of \cite{BubeckSlivkins2012}, there has been growing interest over the past decade in developing algorithms with the \emph{best-of-both-worlds} (BOBW) guarantee in the online learning community, which requires simultaneously achieving
a (nearly) minimax-optimal $\gO(\sqrt{T})$-type regret in the adversarial regime and a (nearly) optimal instance-dependent $\gO(\log T)$-type regret in the stochastic regime, \textit{without} knowing the nature of the regime in advance.
Early BOBW algorithms for MABs
follow a detect-and-switch approach: the learner first conducts statistical tests to identify the regime
and then invokes subroutines specialized to different
regimes~\cite{BubeckSlivkins2012,SeldinSlivkins2014,AuerChiang2016,SeldinLugosi2017}.
A second line of research obtains better adaptation using only the follow-the-regularized-leader
(FTRL) framework, 
\textit{without} explicit regime
detection~\cite{WeiLuo2018,ZimmertSeldin2021,Ito21,JinLiuLuo2023}.
This purely FTRL-based approach
also yields guarantees between the two endpoints, notably
for the \textit{corrupted stochastic regime}~\cite{lykouris2018stochastic,ZimmertSeldin2021}.

For linear bandits, 
\cite{LeeEtAl2021} 
give the first detect-and-switch-based BOBW
algorithm with nearly instance-optimal regret of $\gO(\cstar\log T\log(KT))$
in the stochastic regime and nearly optimal regret of $\gO(\sqrt{dT}\log(KT))$ in the adversarial regime. 
Subsequent works develop purely FTRL-based BOBW algorithms
for linear bandits~\cite{ItoT23,KongZ023,dann2023blackbox,ItoTsuchiyaHonda2024}.
However, the regret guarantee of these approaches in the stochastic regime typically scales as
$\gO(\operatorname{poly}(\log T)/\Delta_{\min})$,
rather than reflecting the finer information-acquisition complexity $\cstar$.
Thus, obtaining BOBW guarantees with information-acquisition regret using a purely FTRL-based algorithm remains a central open problem in the literature, as mentioned in \cite{LeeEtAl2021,KongZ023,ItoT23,ZhaoItoLi2025}:
\begin{quote}
\hypertarget{intro-question-information}{}%
\raggedright
\textbf{Motivating Question 1 (Information acquisition).}
Can a purely FTRL-based algorithm (\textit{i.e.}, the algorithm only runs FTRL without using the detect-and-switch scheme) achieve 
$\gO(\cstar\log T)$-type stochastic regret, while retaining an $\gO(\sqrt{T})$ adversarial regret?
\end{quote}
Further, existing purely FTRL-based algorithms for linear bandits can obtain an $ \gO (\operatorname{poly}(\log T)/ \Delta_{\min}+\sqrt{ C\operatorname{poly}(\log T) /\Delta_{\min}} )$ regret in the corrupted stochastic regime
\cite{ItoT23,KongZ023,ItoT23EXO,ItoTsuchiyaHonda2024},
where $C\ge \sum_{t\le T}\sup_{a\in\cA}|a^\top c_t^{\rm vec}|$ is the total amount of corruption, and $c_t^{\rm vec}$ is the corruption at round $t$ so that $\theta_t=\theta+c_t^{\rm vec}$.
If the main term in the regret in the stochastic regime can be improved to $\gO(\cstar\log T)$, it is then
natural to ask whether the multiplicative coefficient of $\sqrt C$ in the
corruption term can be refined to $\gO(\sqrt{\cstar})$.
This leads to the following question.
\begin{quote}
\hypertarget{intro-question-corruption}{}%
\raggedright
\textbf{Motivating Question 2 (Corruption).}
Can the same algorithm preserve the $\gO(\cstar\log T)$-type main term in the corrupted stochastic regime, paying only an additional $\gO(\sqrt{C})$ corruption term with a horizon-independent coefficient?
In particular, can the corruption term be refined to
$\gO(\sqrt{\cstar C})$, with no horizon-dependent multiplier?
\end{quote}

In many practical applications, linear bandit problems may have infinite action sets, which introduce a further obstacle.
Some existing BOBW algorithms for linear bandits are intrinsically unable to handle infinite action sets (see, \textit{e.g.}, \cite{KongZ023,ItoTsuchiyaHonda2024}).
While the remaining algorithms can work with infinite action sets in the adversarial regime (say, \cite{ItoT23,ItoT23EXO}), the $\gO(\log T/\Delta_{\min})$-type regret upper bounds still become vacuous in the stochastic regime when $\inf_{a\neq a^*}\Delta(a)=0$.
Indeed, analyzing BOBW linear bandits with infinite action sets is
nontrivial: 
even for algorithms tailored to the stochastic regime, 
the instance-optimal upper bound of~\cite{LS17} is restricted to finite action sets.
Extending this guarantee to infinite action sets in the stochastic regime already requires
further advances, let alone obtaining a BOBW guarantee.
When corruption arises, the difficulty
also depends on how the loss gap $\Delta(a)$ relates to the distance
$\|a-a^*\|_{\mathcal S}$ between a suboptimal action $a$ and the optimal action $a^*$:
nearby actions with very small gaps may respond differently to the same
perturbation caused by corruption. 
Thus the coefficient in the corruption regret term may depend on geometric structures beyond the cost of acquiring information.
Taken together, these considerations lead us to the following question concerning the geometry of the action set:
\begin{quote}
\hypertarget{intro-question-geometry}{}%
\raggedright
\textbf{Motivating Question 3 (Geometry).}
Is there an infinite-action analogue of information-acquisition
complexity $c^*$ that supports instance-dependent logarithmic regret
within a BOBW algorithm? Under what geometric conditions can such
an algorithm also achieve an additional $\gO(\sqrt C)$ corruption
term, up to logarithmic factors, and can this guarantee extend to
all compact instances?
\end{quote}

\subsection{Overview of the Main Results}\label{sec:intro-results}

We answer \hyperlink{intro-question-information}{Question~1}
affirmatively, obtain both positive and negative answers to the two
parts of \hyperlink{intro-question-corruption}{Question~2}, and identify
the role of geometry in
\hyperlink{intro-question-geometry}{Question~3}.
In \Cref{tab:intro-regret-comparison}, we compare the regret upper bounds of our algorithms with those of existing BOBW algorithms for linear bandits.

\begin{table}[!t]
\caption{Comparisons with existing works for BOBW linear bandits.
Additive remainders independent of $T$ are omitted from upper bounds.
}
\label{tab:intro-regret-comparison}
\begin{minipage}{\textwidth}
\centering
\footnotesize
\renewcommand{\arraystretch}{1.5}
\setlength{\tabcolsep}{3pt}
\setlength{\minrowclearance}{3pt}
\renewcommand{\tabularxcolumn}[1]{m{#1}}
\newcommand{\tableboundmark}[2][]{\makebox[1.6em][l]{\ding{#2}\ensuremath{^{#1}}}}
\begin{tabularx}{\textwidth}{|
>{\centering\arraybackslash}m{0.018\textwidth}|
>{\centering\arraybackslash}m{0.17\textwidth}|
>{\centering\arraybackslash}X|
>{\centering\arraybackslash}X|
>{\centering\arraybackslash}m{0.135\textwidth}|
>{\centering\arraybackslash}m{0.115\textwidth}|}\hline
\multicolumn{1}{|c|}{} &
& \textbf{Stochastic Regime}
& \textbf{Adversarial Regime}
& \shortstack{\textbf{Purely}\\\textbf{FTRL-based?}}
& \shortstack{\textbf{Infinite}\\\textbf{Action Set?}} \\\hline
& \cite{LeeEtAl2021}$^{\S}$
& $\gO\!\left(\begin{gathered}
\cstar\log T\log(KT)\\[-2pt]
{}+\frac{d\log^{3/2}(KT)}{\Delta_{\min}}
\end{gathered}\right)$
& $\gO\!\left(\begin{gathered}
\sqrt{dT\log T\log(KT)}\\[-2pt]
{}+d\log^{3/2}(KT)
\end{gathered}\right)$
& \tableboundmark{55} & \tableboundmark{55} \\\hhline{|~|-|-|-|-|-|}
& \cite{KongZ023}
& $\gO\!\left(\frac{d\log T\log(KT)}{\Delta_{\min}}\right)$
& $\gO\!\left(\sqrt{dT\log T\log(KT)}\right)$
& \tableboundmark{51} & \tableboundmark{55} \\\hhline{|~|-|-|-|-|-|}
& \cite{ItoT23}
& $\gO\!\left(\frac{d^3\log T}{\Delta_{\min}}\right)$
& $\gO\!\left(d^{3/2}\sqrt{T\log T}\right)$
& \tableboundmark{51} & \tableboundmark[\dagger]{55} \\\hhline{|~|-|-|-|-|-|}
& \cite[Thm.~3]{DannWZ23}
& $\gO\!\left(\frac{d^3\log T}{\Delta_{\min}}\right)$
& $\gO\!\left(d^{3/2}\sqrt{T\log T}\right)$
& \tableboundmark{51} & \tableboundmark[\dagger]{55} \\\hhline{|~|-|-|-|-|-|}
& \cite[Cor.~7]{DannWZ23}
& $\gO\!\left(\frac{d^2\log T}{\Delta_{\min}}\right)$
& $\gO\!\left(d\sqrt{T\log T}\right)$
& \tableboundmark{51} & \tableboundmark[\dagger]{55} \\\hhline{|~|-|-|-|-|-|}
& \cite[Cor.~12]{DannWZ23}
& $\gO\!\left(\frac{d\log K\log T}{\Delta_{\min}}\right)$
& $\gO\!\left(\sqrt{dT\log K}\right)$
& \tableboundmark{51} & \tableboundmark{55} \\\hhline{|~|-|-|-|-|-|}
& \cite{ItoT23EXO}
& $\gO\!\left(\frac{d^2\log T}{\Delta_{\min}}\right)$
& $\gO\!\left(d\sqrt{T\log T}\right)$
& \tableboundmark{51} & \tableboundmark[\dagger]{55} \\\hhline{|~|-|-|-|-|-|}
& \cite{ItoTsuchiyaHonda2024}
& $\gO\!\left(\frac{d\log K\log T}{\Delta_{\min}}\right)$
& $\gO\!\left(\sqrt{dT\log K}\right)$
& \tableboundmark{51} & \tableboundmark{55} \\\hhline{|~|-|-|-|-|-|}
 & \cellcolor{lightgray!30} \Cref{thm:G-main}
& \cellcolor{lightgray!30} $\gO\!\left(\cstar\log K\log T\right)$
& \cellcolor{lightgray!30} $\gO\!\left(\sqrt{dT\log K}\right)$
& \cellcolor{lightgray!30} \tableboundmark{51} & \cellcolor{lightgray!30} \tableboundmark{55} \\\hhline{|~|-|-|-|-|-|}
\multirow{-13.3}{*}[3.5pt]{\rotatebox{90}{\textbf{Upper Bounds}}} & \cellcolor{lightgray!30} \Cref{thm:main-nonseparated-upper}
& \cellcolor{lightgray!30} $\gO\!\left(d\cctr\log T\right)$
& \cellcolor{lightgray!30} $\gO\!\left(d\sqrt T\log T\right)$
& \cellcolor{lightgray!30} \tableboundmark{51} & \cellcolor{lightgray!30} \tableboundmark[\ddagger]{51} \\\hline
& \cite[Cor.~2]{LS17}
& $\Omega\!\left(\cstar\log T\right)$
& -- & -- & \tableboundmark{55} \\\hhline{|~|-|-|-|-|-|}
 & \cellcolor{lightgray!30} \Cref{thm:main-sharp-lower}\textup{(i)}
& \cellcolor{lightgray!30} $\Omega\!\left(\cctr\log T\right)$
& \cellcolor{lightgray!30} -- & \cellcolor{lightgray!30} -- & \cellcolor{lightgray!30} \tableboundmark{51} \\\hhline{|~|-|-|-|-|-|}
& \cite[Thm.~4.3]{DaniHayesKakade2008Price}
& -- & $\Omega\!\left(\sqrt{dT\log K}\right)$
& -- & \tableboundmark{55} \\\hhline{|~|-|-|-|-|-|}
\multirow{-5.6}{*}[-1.95pt]{\rotatebox{90}{\textbf{Lower Bounds}}} & \cite[Thm.~3]{DaniHayesKakade2008Stochastic}
& -- & $\Omega\!\left(d\sqrt T\right)$
& -- & \tableboundmark{51} \\\hline
\end{tabularx}
\par\smallskip
\begin{minipage}{\textwidth}
\footnotesize
Here $K=|\cA|$ for finite action sets.
``\textbf{Infinite Action Set?}'' refers to whether the corresponding regret guarantee holds for infinite action sets with
$\inf_{a\ne a^*}\Delta(a)=0$ in the stochastic regime, and whether it holds for infinite action sets in the adversarial regime.
In this paper, ``\textbf{Purely FTRL-based}'' refers to FTRL-based algorithms without using a detect-and-switch scheme.
The self-concordance parameter in Theorem~3 of \cite{DannWZ23} is taken to be $\gO(d)$.
$^{\dagger}$The adversarial bound applies to infinite action sets, whereas the displayed stochastic bound requires a positive minimum gap and gives no finite guarantee when $\inf_{a\ne a^*}\Delta(a)=0$.
$^{\ddagger}$The sharpness condition is assumed to obtain our regret guarantee in the stochastic regime.
The $\Omega(\sqrt{dT\log K})$ lower bound of
\cite{DaniHayesKakade2008Price} holds on a finite hypercube with $K=2^d$.
\end{minipage}
\end{minipage}
\end{table}

\paragraph{Questions 1 and 2: Information Acquisition and Corruption}
For finite actions, we introduce \emph{information-acquisition FTRL}
(\IAFTRL{}; \Cref*{alg:G-coherent}). 
It is a purely FTRL-based algorithm without using the detect-and-switch scheme.
In particular, \IAFTRL{} minimizes empirical
exploration cost subject to gap-calibrated moment constraints and an
exponential stability constraint, which
 enable comparison with
low-cost sampling allocations once the gap estimates are
sufficiently controlled.
Its regret guarantee is as follows.
\begin{theorem}[Finite-action upper bound; informal version of
\Cref{thm:G-main}]
For linear bandits with finite action sets,
\IAFTRL{} simultaneously guarantees $\Reg=\gO (\sqrt{dT\log K} )$ in the adversarial regime, and 
\begin{align*}
\Reg
=
 \gO \left(\cstar\log K\cdot \log T 
       +M_{\cA,\theta}
       +J_{d,K,\Delta_{\min}}\sqrt C 
 \right)
\end{align*}
in corrupted stochastic regimes,
where
$M_{\cA,\theta}$ is a finite instance-dependent
remainder term independent of $T$ and $C$,
and $ J_{d,K,\Delta_{\min}}=\gO(
 \sqrt{d\log K}\Delta_{\min}^{-129/256})$.
\end{theorem}
This establishes a BOBW guarantee for a purely FTRL-based algorithm with 
an $\gO(c^*\log T)$-type regret in the stochastic regime,
resolving \hyperlink{intro-question-information}{Question~1}.
The additional $J_{d,K,\Delta_{\min}}\sqrt C$ term
answers the first part of
\hyperlink{intro-question-corruption}{Question~2} in the affirmative.
The following lower bound shows why its dependence on $\Delta_{\min}$
cannot in general be replaced by $\cstar$.

\begin{theorem}[Finite-action lower bounds; informal version of
\Cref{thm:main-finite-lower}]
There is a family of linear bandit instances on a fixed finite action set
with arbitrarily small minimum gap $\Delta_{\min}>0$ and
$\cstar=\Theta(1)$.
 In particular,
$\cstar=o(\Delta_{\min}^{-1})$ as $\Delta_{\min}\downarrow0$.
For sufficiently large $C$ and $T$,
there exists a corrupted stochastic environment based on an instance from this family, with total corruption at most $C$, in which every policy incurs regret $\Reg
 =
 \Omega (
       \min\{\sqrt T,\sqrt{C/\Delta_{\min}}\} )$.
\end{theorem}
Consequently, the corruption regret term cannot in general be upper bounded by
$\gO(\sqrt{\cstar C})$ on these instances.
On the other hand, our upper bound has an inverse-gap exponent of
$129/256$,
leaving a multiplicative gap of
$\Delta_{\min}^{-1/256}$ between the upper and lower bounds in their
dependence on $\Delta_{\min}$.

\paragraph{Question 3, Part (i): Information Acquisition beyond Positive Gaps}

We first identify a subclass of linear bandit instances with infinite action sets called \textit{sharp} instances, on which the gap $\Delta(a)$ of any suboptimal action $a$ is lower bounded by the distance from $a$ to the optimal action $a^*$,
even though $\Delta(a)$ may be arbitrarily small (see \Cref{def:sharp} for details).
Further, we propose the \emph{centered information-acquisition
complexity} $\cctr$, which extends the original \emph{information-acquisition
complexity} considered by \cite{LS17} to infinite action sets. In particular, the centered information-acquisition
complexity $\cctr$ measures the least
regret cost of a feasible sampling allocation that controls all loss differences from the optimal action $a^*$ (see \eqref{eq:SH-centered-program}). On finite action sets, we have $\cctr=c^*$.
We then introduce \emph{geometric information-acquisition FTRL}
(\GIAFTRL{}; \Cref*{alg:sharp}), which combines geometrically scaled
moment constraints with a mixed discrete and continuous reference measure
in FTRL with relative-entropy regularization.

\begin{theorem}[Sharp compact upper bound; informal version of
\Cref{thm:main-nonseparated-upper}]\label{thm:simp:sharp_upper}
For linear bandits with sharp action sets,
\GIAFTRL{} simultaneously guarantees $ \Reg=\gO(d\sqrt T\log T)$ in the adversarial regime and
\begin{align*}
 \Reg
 =
 \gO\left( \cctr d\log T+M_\theta
                         +J_\theta\sqrt C\logp(C)\right)
\end{align*}
in the corrupted stochastic regime, where
$M_\theta$ and $J_\theta$ are finite instance-dependent
terms independent of $T$ and $C$.
\end{theorem}
This gives an affirmative answer to the algorithmic part of
\hyperlink{intro-question-geometry}{Question~3} under sharpness.
The next result gives separate lower bounds for the $\gO(\log T)$ term and the $\gO(\sqrt C)$ corruption term within this class.
The latter shows that the multiplicative coefficient in the $\gO(\sqrt C)$ corruption term must also account for the geometry of the instance.

\begin{theorem}[Sharp compact lower bounds; informal version of
\Cref{thm:main-sharp-lower}]
\leavevmode\par
\begin{itemize}[labelindent=0pt,leftmargin=*,align=left,itemsep=4pt,topsep=4pt]
\item
There is a class of sharp stochastic linear bandit instances
on a fixed compact action set such that every consistent policy
incurs $\Omega(\cctr\log T)$ regret on each instance in the class.
\item
There is another class of sharp stochastic linear bandit instances on a
fixed compact action set, with sharpness modulus $\varsigma=\Theta(\phi)$
and $\cctr=\Theta(1)$ for any $0<\phi\le1/8$.
For every such $\phi$, every $C\ge\phi^{-1}$, and every integer
$T\ge\lceil\phi^{-2}\rceil$, every policy incurs
$\Reg=\Omega(\min\{\sqrt T,\sqrt{C/\phi}\})$ in some corrupted stochastic environment based on an instance in this class,
with total corruption budget at most $C$.
\end{itemize}
\end{theorem}

Both lower bounds hold on sharp stochastic instances with arbitrarily small
positive gaps. The second part shows that, even on a fixed infinite action set
with bounded $\cctr$, the multiplicative coefficient in the $\sqrt C$
corruption term can grow as the sharpness modulus decreases.
Intuitively, weaker sharpness allows a smaller loss gap between a suboptimal
action $a$ and the optimal action $a^*$ for a given distance between them.
Corruption can then shrink this gap at lower cost, making it harder for
the learner to identify the optimal action.
On the other hand, unlike in the finite-action case, we remark that a multiplicative gap
of $\logp(C)$ remains between the upper and lower bounds in 
the corruption term $C$.

\paragraph{Question 3, Part (ii): Beyond Sharpness}
When sharpness fails, $\cctr=\infty$, so the preceding allocation-based
guarantee no longer provides a finite bound. 
In this case, we construct a fixed nonsharp instance
on which the $\gO(\sqrt{C})$ corruption guarantee fails.

\begin{theorem}[Negative result on nonsharp instances; informal version of
\Cref{thm:main-critical}]\label{thm:simp:sharp_lower_2}
There is a stochastic linear bandit instance with a fixed compact action set in $\R^2$ 
that does not satisfy the sharpness condition
and $\cctr=\infty$ such that, for every
integer $T\ge4$ and $C\ge0$, every policy satisfies
$ \Reg=\Omega (
       \min\left\{\frac{C^{4/3}}{T^{1/3}},T^{1/3}\right\})$
under deterministic corruption of budget at most $C$.
At $C\asymp\sqrt T$, the lower bound is $\Omega(C^{2/3})$.
\end{theorem}
On this fixed instance, no policy can guarantee an
$\gO(\log T+\sqrt C)$ regret for any horizon $T$
and corruption budgets $C$ in the corrupted stochastic regime.
Consequently, the $\gO(\log T+\sqrt C)$ regret guarantee in the sharp case
cannot extend to all linear bandit instances with infinite action sets. 

\subsection{Main Techniques}\label{sec:intro-techniques}

Our general idea is to incorporate the optimal sampling allocation $n^*$, whose information-acquisition cost is $c^*$,
into exploration, with an exploration strength of $1/t$, so that an
$\gO(\cstar\log T)$ regret can be achieved in the stochastic regime.
The difficulty is, of course, that neither the optimal sampling allocation $n^*$ nor any other feasible sampling allocation is known because the gaps are unknown. In addition, we also need to control the regret contributed by running FTRL, so that the regret of FTRL will not dominate the desired $\gO(\cstar\log T)$ exploration regret cost in the stochastic regime.
Meanwhile, we also need to ensure 
that the algorithm conducts sufficient exploration so that an $\gO(\sqrt{T})$ regret is also guaranteed in the adversarial regime. 
These two challenges become even more pronounced in the infinite action set case, since there will be infinitely many suboptimal actions with arbitrarily small gaps.
We discuss these challenges in detail below.

\paragraph{Challenge 1: Unknown Gaps and Optimal Sampling Allocation}
To bound the exploration regret by the optimal information-acquisition
cost $\cstar=\cstar(\cA,\theta)$, the first bottleneck is that the
optimal sampling allocation, and even the gaps themselves, are unknown.
To cope with this obstacle, we control the empirical gaps and use them to
guide exploration toward a low-cost sampling allocation.

\begin{itemize}[leftmargin=*,itemsep=4pt,topsep=4pt]
\item \textbf{Technique 1: empirical gap ratio control.}
We maintain empirical gaps from the cumulative loss estimates.
Let $\widehat\ell_{t,a}$ be an unbiased estimate of the loss of
action $a$ in round $t$ and
$\widehat L_{t-1,a}=\sum_{s<t}\widehat\ell_{s,a}$. Define
$r_t\in\argmin_a\widehat L_{t-1,a}$ as the empirical best action and
$\widehat\Delta_t(a)=\min\{2,
(\widehat L_{t-1,a}-\widehat L_{t-1,r_t})/t\}$ as the empirical gap.
We construct an optimization problem \eqref{eq:G-program} with two carefully chosen
constraints \eqref{eq:G-program-mag} and \eqref{eq:G-program-var}
that control the
magnitude and second moment of each loss-difference estimate $\widehat\ell_{t,a}-\widehat\ell_{t,r_t}$ relative
to its current cumulative difference $Z_{t,a}\coloneq \widehat L_{t-1,a}-\widehat L_{t-1,r_t}$. 
Intuitively, an action with a
small empirical gap $\widehat\Delta_t(a)$ requires more accurate estimation to separate it from the optimal action $a^*$, whereas a
well-separated action permits larger estimation errors and less exploration.

In the analysis, these two constraints limit abrupt changes in the cumulative loss estimates by controlling the increments of the cumulative
loss differences $X_{s,a}\coloneq \widehat L_{s-1,a}-\widehat L_{s-1,a^*}$ (with increment $X_{s+1,a}-X_{s,a}=\widehat\ell_{s,a}-\widehat\ell_{s,a^*}$) relative to their current scale,
 $Y_{s,a}^{\dagger}\coloneq X_{s,a}+2\tau_s$ (where $s$ ranges over a fixed analysis window and $\tau_s>0$ is a deterministic buffer ensuring $Y_{s,a}^{\dagger}>0$ on the histories retained in the moment recursion).
Consequently, 
the second-order Taylor remainders for $(Y_{s+1,a}^{\dagger})^P$ and $(Y_{s+1,a}^{\dagger})^{-P}$ are controlled by the second-moment bound on $\xi_{s,a}\coloneq Y_{s+1,a}^{\dagger}-Y_{s,a}^{\dagger}$, where $P\ge2$ is a fixed even moment order.
Thus,
we can derive recursive bounds for both positive and inverse moments of the cumulative loss differences $Y_{s,a}^{\dagger}$.
These respectively control overestimation and underestimation
of the empirical gaps, 
giving high-probability ratio bounds for $\widehat\Delta_t(a)/\Delta(a)$
and $\Delta(a)/\widehat\Delta_t(a)$ for suboptimal actions (see \Cref{sec:G-algorithm,ps:prop3-1}).

\item \textbf{Technique 2: mixtures of FTRL updates and any feasible sampling allocations.}
Although the optimal sampling allocation is unknown, the constraints in \eqref{eq:G-program}
are designed 
so that they can be satisfied by a mixture $p_t^n$ of the FTRL update $q_t$
and any feasible sampling allocation $n$.
The allocation is scaled
at order $1/t$, with an adjustment for empirical-gap errors.
Once the empirical gaps are sufficiently controlled, this mixture
satisfies the two gap-calibrated constraints, as well as the stability
constraint discussed below. Since the sampling distribution $p_t$
of our algorithm minimizes the empirical exploration objective in \eqref{eq:G-program},
$\sum_a p_{t,a}\widehat\Delta_t(a)
 \le\sum_a p^n_{t,a}\widehat\Delta_t(a)$.
\textbf{Technique~1} then translates this comparison into one in terms of the true gaps, \textit{i.e.},
$\sum_a p_{t,a} \Delta(a)
 \le\sum_a p^n_{t,a} \Delta(a)$.
Consequently, the exploration cost of $p_t-q_t$ can be
controlled by the regret cost $c^n=\sum_{a\ne a^*}n_a\Delta(a)$ of
any feasible allocation $n$, accumulated at the $1/t$ scale.
Comparing with the best feasible allocation $n^*$ gives the desired
$\gO(\cstar\log T)$-type exploration bound (see
\Cref{ps:prop3-1,app:B-exploration}).
\end{itemize}

\paragraph{Challenge 2: FTRL Regret}
Bounding the exploration cost by $\gO(\cstar\log T)$ alone is not sufficient: we also need to
control the regret of running FTRL on the loss estimates. For FTRL with negative Shannon entropy and (inverse) learning rate $\beta_t$,
standard analysis shows that
\begin{align*}
       \Reg\le
 \underbrace{\beta_1\log K+
 \sum_{t=2}^T(\beta_t-\beta_{t-1})\E[H(q_t)]}_{\text{penalty term}}
 +\underbrace{\sum_{t=1}^T\E[\mathsf D_t]}_{\text{stability term}}
 +\underbrace{\sum_{t=1}^T\E \left[
      \left\langle p_{t}-q_{t},  \widehat{\ell}_t \right\rangle
       \right]}_{\text{exploration term}}\,,
\end{align*}
where $H(q)=-\sum_aq_a\log q_a$ is Shannon entropy and
$\mathsf D_t=\sum_aq_{t,a}\widehat\ell_{t,a}
 +\beta_t\log\sum_aq_{t,a}e^{-\widehat\ell_{t,a}/\beta_t}$.
We need to ensure both the penalty term and stability term are small enough so that they will not dominate the desired $\gO(\cstar\log T)$ leading term.

\begin{itemize}[leftmargin=*,itemsep=4pt,topsep=4pt]
\item \textbf{Technique 3: constant regret control of the penalty term.}
In standard FTRL analyses, the common learning rate schedule $\beta_t\asymp\sqrt t$ combined with direct use of
$H(q_t)\le\log K$ only gives an $\gO(\sqrt T)$ penalty bound.
The data-dependent learning rate $\beta_t\asymp \sum_{\tau=1}^{t-1} 1 / \sqrt{1+\sum_{s=1}^\tau H\left(q_s\right)}$ of~\cite{Ito2022nearly,KongZ023} improves this
by relating cumulative entropy to the probability of selecting
suboptimal actions. However, converting the cumulative entropy into regret
uses 
$\langle q_t,\Delta\rangle\ge
\Delta_{\min}(1-q_{t,a^*})$, with
$\Delta_{\min}=\min_{a\ne a^*}\Delta(a)$,
eventually leading to an $\gO(\log T/\Delta_{\min})$ penalty bound, which loses the finer information-acquisition cost.

Based on \textbf{Technique~1}, we instead show that the probability of each
suboptimal action decays exponentially fast. 
On good histories where the
gap estimates are well controlled, we demonstrate that the update $q_{t}$ under FTRL with Shannon entropy and
 $\beta_t\asymp \sqrt{t}$ satisfies
$q_{t,a}\lesssim e^{-X_{t,a}/\beta_t}
 \lesssim e^{-4t^{1/4}}$ 
 for $a\ne a^*$.
Thus the entropy decreases fast enough to make the cumulative penalty
bounded by a $T$-dependent constant. After accounting for
estimation failures, this constant remains independent of $T$ in the stochastic regime.

\item \textbf{Technique 4: constant regret control of the stability term with large loss estimates.}
Controlling the stability term to obtain the overall $\gO(c^*\log T)$ regret is particularly challenging, since 
standard analyses for FTRL with Shannon entropy in adversarial and BOBW linear bandits use a $G$-optimal design distribution for exploration to prevent excessively negative
loss estimates~\cite{BubeckCesaBianchiKakade2012,KongZ023}. 
In particular, mixing with a $G$-optimal design can ensure $\widehat\ell_{t,a}/\beta_t\ge-1$, allowing the stability term to be bounded by a second-order term of $\gO(\E[\sum_t\beta_t^{-1}\sum_aq_{t,a}\widehat\ell_{t,a}^2])$ and ultimately by an $\gO(\log T/\Delta_{\min})$ term for a data-dependent learning rate $\beta_t$.
However, our exploration distribution $p_t-(1-\gamma_t)q_t$ 
in 
the stochastic regime may not use a $G$-optimal design distribution. Instead, the exploration distribution $p_t-(1-\gamma_t)q_t$ in our algorithm can be any feasible sampling allocation in the stochastic regime, as elaborated in \textbf{Technique~2}.
To this end,
we need to permit larger loss estimates and thus cannot rely on the uniform lower bound $\widehat\ell_{t,a}/\beta_t\ge-1$. In other words, we need to achieve a tighter upper bound on the stability term using less amount of exploration.

To deal with the above difficulty, we instead bound the one-round stability term $\mathsf D_t$ directly by a proxy function
\[
 \mathcal C_t(p)=\beta_t\E_{b\sim p}\!\left[
 \E_{a\sim q_t}\!\left[
 \chi\!\left(\frac{(a-\bar a_t)^\top v_b(p)}{\beta_t}\right)
 \right]\right]\,,
\]
where $\bar a_t=\sum_aq_{t,a}a$, $\chi(x)=e^x+e^{-x}-2$, and
$v_b(p)=M(p)^{-1}b$ for the positive definite design matrix
$M(p)=\sum_ap_aaa^\top$.
On the same well-controlled histories as above,
due to the magnitude constraint \eqref{eq:G-program-mag},
we show that
$\gC_t(p_t) \le \gO(\beta_t\sum_{a\ne a^*}q_{t,a}
e^{2X_{t,a}/(B\beta_t)})$, where $B\ge4$ is a fixed parameter in
the gap-calibrated constraints.
The key here is again the intuition from \textbf{Technique~1}: for a suboptimal action $a$ with a sufficiently large empirical gap $\widehat\Delta_t(a)$, it is already well separated from $a^*$, and we can significantly reduce exploration on $a$, even if this makes its loss estimate no longer satisfy $\widehat\ell_{t,a}\ge -\beta_t$.
Indeed, on good events, the decay $q_{t,a}\lesssim e^{-X_{t,a}/\beta_t}$ can absorb the exponentially large factor
caused by a large loss estimate as
$q_{t,a}e^{2X_{t,a}/(B\beta_t)}
 \le e^{-(1-2/B)X_{t,a}/\beta_t}$.
This makes the overall stability term a $T$-independent constant as well.

Algorithmically, we also impose the third constraint
$\mathcal C_t(p)\le2d/\beta_t$ as
\eqref{eq:G-program-cert} in \eqref{eq:G-program} to 
control the stability term on every history. In particular, this 
ensures that mixing $q_t$ with a $G$-optimal design satisfies all the
constraints in  \eqref{eq:G-program}, so the same algorithm retains an $\gO(\sqrt{T})$
regret guarantee in the adversarial regime. 
\end{itemize}

\paragraph{Challenge 3: Infinite Action Sets}
The finite-action analysis relies critically on $\Delta_{\min}>0$, for example, to ensure that the difference in cumulative estimated losses $\widehat L_{t,a}-\widehat L_{t,a^*}$ of every suboptimal action $a$ is large enough,
and obtain uniform decay $q_{t,a}\lesssim e^{-4t^{1/4}}$ for the FTRL update.
These arguments no longer
apply directly as an infinite action set may have $\inf_{a\ne a^*}\Delta(a)=0$. 

\begin{itemize}[leftmargin=*,itemsep=4pt,topsep=4pt]
\item \textbf{Technique 5: relative entropy with a nonuniform mixture reference measure.}
For infinite action sets, FTRL with relative-entropy regularization $\KL(\cdot,\mu)$,
equivalently continuous exponential weights, typically uses a uniform
reference measure $\mu$ (see, \textit{e.g.}, \cite{ItoT23EXO}).
Although it allows the learner to compete with arbitrary actions to obtain an $\gO(\sqrt{T})$ adversarial guarantee, a uniform reference measure $\mu$ is insufficient for obtaining the desired $\gO(c^*\log T)$ regret, which requires $q_t$ to concentrate rapidly at $a^*$ in the stochastic regime.
To meet both requirements,
we construct a reference measure $\mu=\mu_{d}+\mu_c$ with a discrete component $\mu_{d}$ and a continuous
component $\mu_{c}$. Since the finite-action uniform decay bound $q_{t,a}\lesssim e^{-4t^{1/4}}$ is unavailable,
we instead control the reference mass
$\mu\{a:0<\Delta(a)\le\varepsilon\}$ of suboptimal actions with sufficiently small gaps.

\begin{itemize}
      \item For the discrete component $\mu_{d}$, we enumerate rational directions in
$\mathbb Q^d$ and record their minimizing actions in order, removing
duplicates to obtain $(b_j)_{j\in J}$, with $J\subseteq\mathbb N$.
Sharpness ensures that $a^*$ remains optimal in a neighborhood of the
unknown $\theta$. 
Since $\mathbb Q^d$ is dense in $\mathbb R^d$, the list must
therefore contain $a^*=b_{j^*}$.
We then penalize the weight $\pi_j$ as $\pi_j=e^{-h_j}$ for each recorded action $b_j$, according to its
distances $h_j\asymp \sum_{i\in J,\,i<j}
 \exp(\|b_j-b_i\|_{\mathcal S}^{-8})$ from earlier actions, where $\|x\|_{\mathcal S}\coloneq \inf\{r>0:x\in r\mathcal S\}$ is the norm induced by the symmetric convex hull $\mathcal S\coloneq \Conv(\cA\cup(-\cA))$.
The finitely many actions preceding
$a^*$ in $(b_j)_{j\in J}$, if any, have gaps bounded below by the positive prefix gap $\Delta_{\rm pre}=\min(\{1\}\cup\{\Delta(b_j):j\in J,\ j<j^*\})$ due to uniqueness of the optimal action,
so they
contribute no mass for sufficiently small $\varepsilon$.
For actions succeeding
$a^*$ in $(b_j)_{j\in J}$, 
sharpness guarantees that the distance $\|a-a^*\|_{\mathcal S}$ of action $a$ to $a^*$ needs to be small enough for the gap $\Delta(a)$ to be sufficiently small. 
The distance penalty then makes their
weights sufficiently small. Thus, in both cases, the total reference mass
on suboptimal actions with gaps close to zero is small.
The resulting discrete measure is defined as $\mu_c=\sum_{j\in J}\pi_j\delta_{b_j}$, where $\delta_{b_j}$ is the Dirac measure at $b_j$.

\item For the continuous component $\mu_{c}$, the key observation is that $a^*$ lies
on the boundary of $\mathcal K=\Conv(\cA)$.
Therefore, we take the law of
$\bar{U}_{m}=m^{-1}\sum_{i=1}^{m}U_i$, scaled by $\omega_\mu=1-\sum_j\pi_j>0$, as $\mu_{c}$, where $U_1,\ldots,U_{m}\overset{\rm i.i.d.}{\sim}\lambda_{\mathcal K}$ and $\lambda_{\mathcal K}$ is the uniform volume measure on $\mathcal K$.
Then, a small gap of $\bar{U}_{m}$ implies that all samples must lie in a
small-gap region near the boundary. 
Independence of $\{U_i\}_{i}$ ensures that this probability is sufficiently small:
$\sP(\Delta(\bar{U}_{m})\le\varepsilon)
 \lesssim \varepsilon^{m}$,
while $\bar{U}_{m}$ still has support throughout $\mathcal K$ for the adversarial regret guarantee. In our algorithm, we set $m=16$.
\end{itemize}
Together, we can show that the mixture
$\mu=\mu_d+\mu_c= \sum_{j\in J}\pi_j\delta_{b_j}+\omega_\mu\mu_m$,
satisfies 
$\mu\{a:0<\Delta(a)\le\varepsilon\}
 \lesssim \varepsilon^{16}$.
This sufficiently small probability mass of suboptimal actions with sufficiently small gaps
allows $q_t$ to concentrate fast enough 
to ensure the penalty and
stability term to be
$T$-independent constants in the stochastic regime (see \Cref{sec:SH-algorithm}).

In addition, 
the finite-action exploration constraints in \eqref{eq:G-program} also do not apply due to potentially arbitrarily small gaps. 
We incorporate 
the distance of action $a$ to the empirical best action 
into their magnitude and second-moment bounds.
In this way, the exploration can still be well calibrated even for a suboptimal action $a$ with arbitrarily small $\Delta(a)$, 
by leveraging the sharpness condition.
\end{itemize}

\section{Preliminaries}\label{ch:overview}\label{sec:G-cone-def}

In linear bandits, there is a \textit{known} compact action set $\cA\subseteq\BB$, where
$\BB\coloneq \{x\in\R^d:\|x\|_2\le1\}$ is the Euclidean unit ball.
We assume without loss of generality that
$\operatorname{span}(\gA)=\sR^d$ and denote by $K\coloneq |\gA|$ the size of $\gA$, when $\gA$ is a finite set.
At round $t$,
the environment selects a loss vector $\theta_t\in\R^d$, unknown to the learner.
Meanwhile, the learner chooses $A_t\in\cA$, and observes random feedback $Y_t$ such that $ \E[Y_t\mid\cF_{t-1},A_t]=A_t^\top\theta_t$ and $ \|\theta_t\|_\cA\coloneq \sup_{a\in\cA}|a^\top\theta_t|\le1$.
Throughout this work, we assume $Y_t$ satisfies $ |Y_t|\le 1$.
The learning objective is to minimize the \textit{pseudo-regret} (\textit{regret} for short in what follows)
\[
\Reg\coloneq \sup_{u\in\cA}\Reg(u)\,,\quad\text{where}\quad
 \Reg(u)=\E \left[\sum_{t=1}^T(A_t-u)^\top\theta_t\right]\,,
\]
and the expectation is taken over the potential randomness of both the learner and the environment.
\paragraph{Regime of the Environment}
We denote by $\gF_t$ the $\sigma$-algebra generated by $\{A_1,Y_1,\ldots,A_t,Y_t\}$.
In the \textit{stochastic regime}, $\theta_t=\theta$ for all $t\in[T]$, where $\theta$ is an unknown fixed loss vector with $\|\theta\|_\cA\le1$.
Throughout this work, we assume there exists a unique optimal action $a^*$ in the stochastic regime. This is required in all existing studies of BOBW linear bandits 
\cite{LeeEtAl2021,KongZ023,ItoT23,DannWZ23,ItoTsuchiyaHonda2024,ZhaoItoLi2025} (see \Cref{sec:scope} for further discussion).
For $a\in\Conv(\cA)$, let $\Delta(a)\coloneq(a-a^*)^\top\theta$ denote the loss gap.
When $\cA$ is finite, let $\Delta_{\min}=\min_{a\in\cA\setminus\{a^*\}}\Delta(a)$ be the minimum gap, which satisfies $\Delta_{\min}>0$ by uniqueness of $a^*$.
We further assume $\Delta_{\min}\in (0,1]$ without loss of generality.
In the \textit{adversarial regime}, the environment may choose $\theta_t$ depending on $\gF_{t-1}$ and its internal randomness.
In addition, we also consider the \textit{corrupted stochastic regime}, which interpolates between the \textit{stochastic regime} and the \textit{adversarial regime}.
In this regime, there exists some fixed yet unknown $\theta$ so that the actual loss vector $\theta_t$ is $\theta_t=\theta+c_t^{\rm vec}$, where $\kappa_t=\|c_t^{\rm vec}\|_\cA$.
Let $C\ge0$ be a deterministic corruption level, unknown to the
learner, such that $\sum_{t\le T}\kappa_t\le C$ almost surely.

\paragraph{Sampling Allocations and Information Acquisition Complexity}
Canonical worst-case minimax lower and upper bounds for linear bandits over general compact action sets are $\Omega(d\sqrt{T})$ and $\gO(d\sqrt{T\log T})$, respectively; for a finite action set of size $K$, the regret admits the upper bound $\gO(\sqrt{dT\log K})$~\cite{BubeckCesaBianchiKakade2012}.
On the other hand, the \textit{instance-optimal regret} for linear bandits is more subtle. In particular,
 \cite{LS17} show that 
for a stochastic linear bandit instance $(\gA,\theta)$ with a finite action set $\gA$,
the regret lower bound\footnote{Though \cite{LS17} consider Gaussian noise in the feedback while we consider bounded feedback, it is straightforward to obtain similar results in our setting by considering Bernoulli feedback, with only the constants changed (see \cite{LeeEtAl2021} for detailed discussions on this).}
\begin{align}\label{eq:linear_bandit_ins_opt}
    \liminf_{T\to \infty}
    \frac{\Reg}{\log T}\ge \Omega(c^*(\gA,\theta))
\end{align}
holds
for any \textit{consistent} algorithm. Here, an algorithm is called \textit{consistent} if it guarantees a regret $\Reg=o(T^{\alpha})$ for any $\gA$, $\theta$, and $\alpha>0$. Specifically, $c^*(\gA,\theta)$ in \eqref{eq:linear_bandit_ins_opt} is defined as
\begin{equation}\label{eq:overview-cstar}
 \cstar(\gA,\theta)
 =
 \inf_{n\in \sR_{\ge 0}^{\gA}}\left\{c^{n}(\gA,\theta):
 H_n=\sum_a n_a a a^\top\succ0\,,
 \quad
 \|a\|_{H_n^{-1}}^2\le\Delta(a)^2/2\,,\quad \forall a\ne a^* \right\} \,,
\end{equation}
where $c^{n}(\gA,\theta)\coloneq \sum_{a\in\gA} n_a\Delta(a)$ and $\|x\|_M\coloneq \|M^{1/2}x\|_2$ for a positive definite matrix $M$. In this work, we call $n\in \sR_{\ge 0}^{\gA}$ a {sampling allocation} if $n$ satisfies the feasibility condition in \eqref{eq:overview-cstar}, and call $c^{n}(\gA,\theta)$ the {information-acquisition cost} of $n$, which measures its regret cost. 
Also, we refer to $\cstar(\gA,\theta)$ as the \textit{information-acquisition complexity} of the instance $(\gA,\theta)$ since it measures the regret cost for acquiring sufficient yet minimal information to distinguish all the suboptimal actions from $a^*$.

\paragraph{Infinite Action Sets with Sharpness}
As illustrated in \hyperlink{intro-question-geometry}{Motivating Question~3}, we are further interested in whether an analogue of $\gO(\cstar(\gA,\theta)\log T)$-type regret can be achieved in the infinite action set case. In this work, we propose and identify the following \textit{sharpness} condition, under which we show $\gO(\cstar(\gA,\theta)\log T)$-type regret and even BOBW regret can be achieved.

Specifically, for a compact infinite action set $\gA$, we say $\gB$ is a \textit{generator} for $\gA$ if $\mathcal B\subseteq\cA\subseteq\Conv(\mathcal B)$.
Note that $\mathcal B$ may still be uncountable.
For a generator $\mathcal B\subseteq\BB$ of $\cA$, let $\mathcal K=\Conv(\mathcal B)$, $k=\dim\operatorname{aff}(\mathcal K)$, $\mathsf D=\operatorname{span}(\mathcal K-\mathcal K)$, and $\mathcal S=\Conv(\mathcal B\cup(-\mathcal B))$.
Let $\|x\|_{\mathcal S}=\inf\{r>0:x\in r\mathcal S\}$ be the Minkowski function, which is a norm as $\mathcal B$ spans $\R^d$. The sharpness condition is defined as follows.
\begin{definition}[Sharpness]
    \label{def:sharp}
A stochastic linear bandit instance $(\gA,\theta)$ with a unique minimizer $a^*$ is called \textit{geometrically sharp} on
 $\mathcal K$ if
\begin{equation}\label{eq:SH-overview-sharp}
 \varsigma\coloneq \inf_{a\in\mathcal K\setminus\{a^*\}}
 \frac{(a-a^*)^\top\theta}{\|a-a^*\|_{\mathcal S}}>0\,.
\end{equation}
\end{definition}
We call $\varsigma$ the sharpness modulus. Intuitively, sharpness requires the loss gap to be at least proportional to the distance from the optimal action $a^*$, measured in $\|\cdot\|_{\mathcal S}$; thus the gap cannot vanish faster than this distance as an action approaches $a^*$.

For a nonnegative measure $n$ on $\cA$, we say it is finite if its total mass $ N_n=n(\gA)<\infty$. Define the design matrix as $ H_n=\int_{\cA}aa^\top n(da)$ for a nonnegative finite measure $n$ on $\cA$.
Let $ c^{n}(\gA,\theta)
 =\int_{\gA}(a-a^*(\gA,\theta))^\top\theta\,n(da)
 =\int_{\gA}\Delta(a)n(da)$ and $c_{\mathrm{ctr}}^{n}(\gA,\theta)\coloneq c^{n}(\gA,\theta)$.
We then propose the following \textit{centered} information-acquisition complexity, which extends the raw information-acquisition complexity $\cstar(\gA,\theta)$ of \cite{LS17} 
in \eqref{eq:overview-cstar}.

\begin{definition}[Centered information-acquisition complexity]\label{def:D-physical-design}
For a stochastic linear bandit instance $(\gA,\theta)$, its \textit{centered} information-acquisition complexity is defined as
\begin{equation}\label{eq:SH-centered-program}
 \cctr(\gA,\theta)
 =
 \inf_{n\in \sR_{\ge 0}^{\gA}}
 \left\{c^{n}_{\mathrm{ctr}}(\gA,\theta):
 H_n
 \succ0\,,
 \quad
\|u-a^*\|_{H_n^{-1}}^2\le\Delta(u)^2/2
 \,,\quad \forall u\in\Conv(\cA) \right\} \,,
\end{equation}
\end{definition}
Importantly, we show that the sharpness of the instance $(\cA,\theta)$ 
is necessary and sufficient for $\cctr(\cA,\theta)< \infty$.
Equivalently, sharpness holds exactly when a finite centered feasible
sampling allocation exists. 
Besides, for a finite action set $\gA$ with a unique optimal action $a^*$, the positive minimum gap $\Delta_{\min}$ implies sharpness, and we can show that $\cstar(\cA,\theta)=\cctr(\cA,\theta)$. 
However, we note that
this equality concerns only the infimum values;
 the optimal sampling allocation that attains the infimum, if any, might still be different. 

\paragraph{Additional Notation}
Throughout this work, we omit the dependence on $(\gA,\theta)$ in $\cstar(\gA,\theta)$ and $ \cctr(\gA,\theta)$ when the context is clear.
We use $\gP(\gA)$ to denote the set of all probability measures on $\gA$.
For two probability measures $p,q$,
the relative entropy $\KL(q,p)$ is defined as $\KL(q,p)=\int\log(dq/dp)\,dq$ when $q\ll p$, and is $+\infty$ otherwise.
Besides, for $x\ge0$, we define
$\logp(x)\coloneq\log(\max\{e,x\})$. 
The positive part of a real number $z$ is defined by $(z)_+\coloneq\max\{0,z\}$.

\section{Finite Action Sets}\label{sec:G-research}

In this section, we first introduce the proposed information-acquisition FTRL algorithm (\IAFTRL{};
\Cref*{alg:G-coherent}) in \Cref{sec:G-algorithm}. Then we present its regret guarantees in \Cref{sec:finite-upper} and compare them with lower bounds in \Cref{sec:finite-lower-main}.

\subsection{Algorithm}\label{sec:G-algorithm}

At a high level, \IAFTRL{} uses FTRL with the (negative) Shannon entropy regularizer to update the distribution $q_t$ over $\cA$. Based on $q_t$, it then computes a sampling distribution $p_t\in\gP(\cA)$ by
minimizing the empirical exploration cost subject to gap-calibrated moment constraints and an exponential stability
constraint on its parameter estimator.

\paragraph{Loss Estimates and FTRL Updates}
For any $p\in\gP(\cA)$, let $ M(p)=\sum_{a\in\cA}p_a aa^\top$ be the
      design matrix induced by $p$.
To construct the loss estimator, 
      for any $p\in\gP(\cA)$ with $M(p)\succ0$, we define $v_a(p)=M(p)^{-1}a$ for all $ a\in\cA$.
At round $t$, let $v_{t,a}=v_a(p_t)$.
As is common in the literature on adversarial linear bandits, we then construct the
      least-squares estimator
as $\widehat\theta_t=v_{t,A_t} Y_t$, where
      $A_t\sim p_t$ is the action played by the learner
 and $Y_t$ is the noisy feedback received by the learner in round $t$. 
By definition, we have $\sum_a p_a v_a(p)a^\top=M(p)^{-1}M(p)=I_d$, which guarantees $\E[\widehat\theta_t\mid\cF_{t-1}]=\theta_t$.
The learner's sampling distribution $p_t$ will be specified in \eqref{eq:G-program} below.

Denote by
$\widehat\ell_{t,a}=a^\top\widehat\theta_t$ the estimated loss of action $a$ in round $t$ and by $\widehat L_{t,a}=\sum_{s\le t}\widehat\ell_{s,a}$ the sum of the estimated losses of action $a$ up to round $t$. Define $\widehat L_{0,a}=0$.
At the beginning of round $t$, \IAFTRL{} solves the following optimization problem to obtain the FTRL update $q_t$:
\begin{align}
       q_t=\argmin_{q\in\gP(\cA)}
 \left\{
            \left\langle q, \widehat L_{t-1}\right\rangle
                 +\beta_t
            \psi(q) \right\} 
            \label{eq:finite:ftrl-qt}     
            \,,
\end{align}
where $\beta_t=16d+\sqrt{dt/ (1+\log K)}$ is the (inverse) learning rate and $\psi(q)=\sum_aq_a\log q_a$ is the (negative) Shannon entropy.

\paragraph{Gap-Calibrated Exploration}
The above procedures are standard in the bandit literature, whereas the main technical innovation of our \IAFTRL{} algorithm lies in how we conduct exploration based on the FTRL update $q_t$.
      Given 
      $(\widehat L_{t-1,a})_{a\in\cA}$, we define the \textit{reference action} $r_t$
      and the \textit{empirical gap} $\widehat\Delta_t(a)$ respectively as
\begin{align}\label{eq:finite:reference-action}
       r_t\in\argmin_{a\in\cA} \widehat L_{t-1,a} \,,\qquad
  \widehat\Delta_t(a)=\min\{2,Z_{t,a}/t\}
 \,,\qquad
  Z_{t,a}=\widehat L_{t-1,a}-\widehat L_{t-1,r_t} 
 \,,
\end{align}
where we use a fixed ordering to break ties. 
The nonnegative $Z_{t,a}$ is the estimated cumulative loss difference, while the empirical gap
$\widehat\Delta_t(a)$ is its clipped, normalized value.
For any $u,v\in\R^d$ and $y\in[-1,1]$, we
define the candidate loss-difference estimate for action $u$ relative to $r_t$
as $d_t(u;v,y)=(u-r_t)^\top v y$.
Let
$\overline d_t(v,y)=\sum_aq_{t,a} d_t(a;v,y)$ and $ \bar a_t=\sum_aq_{t,a}a$. Then $ d_t(u;v,y)-\overline d_t(v,y)=(u-\bar a_t)^\top vy$.
In \eqref{eq:G-program},
we set $ \Gamma_t=\frac{4d}{\beta_t}\le\frac14$,
 $B=2^{25}\lceil256+\log(2K)\rceil$, and $\chi(x)=e^x+e^{-x}-2$.
 Let $\mathcal Q$ contain all families $Q=(Q_b)_{b\in\cA}$ of feedback distributions supported on $[-1,1]$.

We now introduce the proposed \textit{gap-calibrated exploration} (GCE) mechanism, in which the final sampling distribution $p_t$ is obtained by solving the optimization problem in \eqref{eq:G-program}.
Intuitively, this optimization problem simultaneously requires that (a) 
we can obtain reliable comparisons of empirical and true gaps and sufficient cumulative-loss separation of the optimal action $a^*$ in the stochastic regime; and (b) the algorithm's update is sufficiently stable that an $\gO(\sqrt{T})$-type worst-case regret can still be guaranteed in the adversarial regime.
The objective
\eqref{eq:G-program-objective} in \eqref{eq:G-program} 
is to minimize $\langle p, \widehat\Delta_t\rangle$, subject to the constraints below.
The resulting solution $p_t$ may not be unique; in this case, we simply choose $p_t$ according to a fixed $\gF_{t-1}$-measurable rule (\textit{e.g.}, the lexicographically smallest $p_t$).
Note that \eqref{eq:G-program} optimizes only $p$: its mixture constraints require the
existence of auxiliary $w\ge0$ and $\gamma=\sum_bw_b\le\Gamma_t$.
Equivalently, they require $p_a\ge(1-\Gamma_t)q_{t,a}$ for every $a$.
Since $q_{t,a}>0$ and the actions span $\R^d$, this domination ensures
$M(p)\succ0$.
We explain each constraint of \eqref{eq:G-program} below. 

\begin{subequations}\label{eq:G-program}
\allowdisplaybreaks[0]
\predisplaypenalty=0
\begin{align}
 &\underset{p\in\gP(\cA)}
       {\operatorname{minimize}}
       &&\sum_a p_a\widehat\Delta_t(a)\,,
       \label{eq:G-program-objective}\\
 &\text{subject to}
       &&\begin{gathered}
            p=(1-\gamma)q_t+w \,,
            \quad
            \gamma=\sum_{b\in\cA} w_b \,,
         \end{gathered}
       \label{eq:G-program-mix}\\
 &&&w_b\ge0\quad (\forall b\in\cA)\,,\quad \gamma\le\Gamma_t \,,
       \label{eq:G-program-dom}\\
 &&&\sup_{b\in\cA\,,\,Q\in\mathcal Q}
       \E_{Y_b\sim Q_b}
       \left[
       \bigl|d_t(a;v_b(p),Y_b)\bigr|\right]
           \le\beta_t+\frac{Z_{t,a}}B\quad(\forall a\in\cA) \,,
       \label{eq:G-program-mag}\\
 &&&\sup_{Q\in\mathcal Q}
       \E_{\substack{A\sim p\\Y\sim Q_A}}\!\left[
       d_t(a;v_A(p),Y)^2
       \mid\cF_{t-1}\right]
           \le\beta_t+\frac{Z_{t,a}^2}{Bt}\quad(\forall a\in\cA) \,,
       \label{eq:G-program-var}\\
 &&&\begin{aligned}
       \beta_t\sup_{Q\in\mathcal Q}
         \E_{\substack{A\sim p\\Y\sim Q_A}}\!
         \left[\left.
            \sum_aq_{t,a}
         \chi\!\left(\frac{d_t(a;v_A(p),Y)
                            -\overline d_t(v_A(p),Y)}{\beta_t}\right)
         \right|
         \cF_{t-1}\right]
       \le\frac{2d}{\beta_t} \,.
     \end{aligned}
       \label{eq:G-program-cert}
\end{align}
\end{subequations}

\begin{algorithm}[!t]
\caption{Information-Acquisition FTRL (IA-FTRL)}\label{alg:G-coherent}
\begin{algorithmic}[1]
\Require A finite action set $\cA$.
\State Set $\widehat L_{0,a}=0$ for all $a\in\cA$.
\For{$t=1,2,\ldots$}
 \State 
Solve \eqref{eq:finite:ftrl-qt} to obtain the FTRL update $q_t$.
 \State Compute the reference action $r_t$, cumulative loss differences $Z_t$, and empirical gaps $\widehat\Delta_t$
 by \eqref{eq:finite:reference-action}.
 \State 
 Solve \eqref{eq:G-program} 
 to obtain the sampling distribution $p_t$.
 \State Draw $A_t\sim p_t$, play $A_t$, and observe $Y_t\in[-1,1]$.
 \State Set $\widehat\theta_t=v_{t,A_t}Y_t$ and $\widehat\ell_{t,a}=a^\top\widehat\theta_t$ for $a\in\cA$.
 \State Update $\widehat L_{t,a}=\widehat L_{t-1,a}+\widehat\ell_{t,a}$ for $a\in\cA$.
\EndFor
\end{algorithmic}
\end{algorithm}

\begin{itemize}[leftmargin=*,itemsep=4pt,topsep=4pt]

 \item \textbf{Sampling distribution \eqref{eq:G-program-mix} and exploration \eqref{eq:G-program-dom}.}
      In FTRL, the constructed loss estimator $\hat\ell_{t}$ is usually constrained to ensure a stable update. 
      In linear bandits, this in turn requires the sampling distribution $p_t$ to be ``diverse'' enough so that each direction in the space can be explored sufficiently often. 
      Therefore, in constraint \eqref{eq:G-program-mix}, we mix the FTRL update $q_t$ with an additional distribution $\{w_a/\gamma\}_{a\in\cA}$ to conduct exploration. The total amount of exploration is also limited in \eqref{eq:G-program-dom} so that 
there would not be unnecessary exploration which prevents us from obtaining
      an $\gO(\sqrt{T})$ regret guarantee in the adversarial regime. 
      However, a key distinction between our work and previous works studying FTRL with Shannon entropy for adversarial linear bandits \cite{BubeckCesaBianchiKakade2012} and BOBW linear bandits \cite{KongZ023,ItoTsuchiyaHonda2024} is that these works explicitly require the exploration distribution $w/\gamma$ to be a G-optimal design distribution, while we replace this requirement with more adaptive and flexible constraints, detailed below.

 \item \textbf{Update magnitude constraint \eqref{eq:G-program-mag}.}
 Since $\mathcal Q$ includes every Dirac feedback distribution, \eqref{eq:G-program-mag}
is equivalent to controlling every possible update magnitude.
 Thus its equivalent
      computable form is $|(a-r_t)^\top v_b(p)|\le\beta_t+Z_{t,a}/B$ for all $a,b\in\cA$.

 \item \textbf{Second-moment control \eqref{eq:G-program-var}.}
      The bound in \eqref{eq:G-program-var} controls the conditional second moment
      of the loss-difference estimate.
 A smaller estimated cumulative difference $ Z_{t,a}$ on the RHS of \eqref{eq:G-program-var} imposes a tighter budget, requiring more information to distinguish plausibly optimal actions.
Moreover, since the feedback satisfies $|Y|\le1$,
      the constraint \eqref{eq:G-program-var} is equivalent to
\[
 \sum_bp_b\bigl[(a-r_t)^\top v_b(p)\bigr]^2
 =(a-r_t)^\top M(p)^{-1}(a-r_t)
 \le\beta_t+\frac{Z_{t,a}^2}{Bt} \,.
\]
Constraints \eqref{eq:G-program-mag} and \eqref{eq:G-program-var}
are used to bound the Taylor remainder
in our moment recursions for (buffered shifted) $X_{t,a}=\widehat L_{s-1,a}-\widehat L_{s-1,a^*}$, the cumulative difference between action $a$ and the optimal action $a^*$:
the magnitude constraint \eqref{eq:G-program-mag} controls the curvature
along the update, while the second-moment constraint \eqref{eq:G-program-var} controls the conditional expectation of the quadratic remainder.
This, in turn, guarantees that the constructed empirical gaps $\widehat\Delta_t(a)$ are sufficiently accurate to be bounded both above and below by the true gap $ \Delta(a)$.

 \item \textbf{FTRL stability \eqref{eq:G-program-cert}.}
The purpose of constraint
\eqref{eq:G-program-cert} is to control the \textit{stability term} of FTRL.
Typically, the standard analysis of FTRL with Shannon entropy first enforces the hard threshold $\widehat\ell_{t,a}/\beta_t\ge-1$ and then bounds the one-round stability
term 
of
$\beta_t^{-1}\E[\sum_a p_{t,a}\widehat\ell_{t,a}^2\mid\cF_{t-1}]$ (see \textit{e.g.}, Theorem 27.1 in \mbox{\cite{lattimore2019bandit}}).
      The obstruction in our case is that satisfying
$\widehat\ell_{t,a}/\beta_t\ge-1$ usually requires mixing the FTRL update $q_t$ with the G-optimal design distribution.
In this way, using a carefully designed data-dependent learning rate $\beta_t$, one can balance the stability term against the regret contributed by the G-optimal design distribution \mbox{\cite{KongZ023,ItoTsuchiyaHonda2024}}, but this still yields only a coarse $\gO(\log T/\Delta_{\min})$ regret bound, whereas our goal is a finer $\gO(c^*\log T)$ regret.
  In contrast, our constraint \eqref{eq:G-program-cert} directly bounds the stability term via 
an exponential proxy function 
  for the centered candidate difference
$d_t(a;v_A(p),Y)-\overline d_t(v_A(p),Y)$, 
whose realized value is $(a-\bar a_t)^\top\widehat\theta_t$. 
To see this, first notice that because $\chi$ is even and increasing in absolute value, the computable form of constraint \eqref{eq:G-program-cert} 
      can be written as
 \[
  \gC_t(p)\coloneq \beta_t\sum_bp_b\sum_aq_{t,a}
   \chi \left(\frac{(a-\bar a_t)^\top v_b(p)}{\beta_t}\right)
     \le\frac{2d}{\beta_t} \,.
 \]
Meanwhile, the one-round stability term of FTRL with Shannon entropy is
$ D_t=\beta_t\log \E_{a\sim q_{t}}[
       \exp(-(a-\bar a_t)^\top\widehat\theta_t/\beta_t)]$, which can be shown to satisfy $\E[D_t\mid\cF_{t-1}]\le\gC_t(p_t)$.

In the stochastic regime, after FTRL converges to the optimal action $a^*$, the estimated cumulative loss difference $Z_{t,a}$ may be as large as $\gO(t\Delta(a))$. In this case, one might wonder whether
our stability constraint \eqref{eq:G-program-cert} can be satisfied, given that the magnitude constraint \eqref{eq:G-program-mag} allows $d_t$ to be as large as $\gO(Z_{t,a})$ and thus
$\gC_t(p)\asymp \sum_a \chi (Z_{t,a})$ seems to be exponentially large in $t\Delta(a)$. However, we remark that on appropriate good events with $r_t=a^*$, we can show that $q_{t,a}$ satisfies an exponential decay bound of $q_{t,a}\le \exp(-Z_{t,a}/\beta_t)$ and thus we can prove that $\gC_t(p_t)\lesssim \beta_t\sum_{a\ne a^*}q_{t,a}\exp(2Z_{t,a}/(B\beta_t))\lesssim  \beta_t\sum_{a\ne a^*}\exp(-(1-2/B)Z_{t,a}/\beta_t)$ on good events
(see \Cref{lem:F-good-stability}). Therefore, for sufficiently large $B$, with high probability, the stability term will not violate constraint \eqref{eq:G-program-cert}; instead, it will decay exponentially fast. In other words, a suboptimal action with a large cumulative loss
difference $Z_{t,a}$ can allow updates much larger than $\beta_t$ while making a very small contribution to the stability term. This explains exactly why we can significantly reduce the amount of exploration on suboptimal actions, without the need to satisfy the usual $\widehat\ell_{t,a}/\beta_t\ge-1$ condition. 
This refinement from hard-threshold exploration to gap-dependent exploration is
essential: summing only the worst-case bound $2d/\beta_t$ would only give
an $\gO(\sqrt{T})$ rate.

\end{itemize}
      In the stochastic regime,
      we can show that by mixing the FTRL update $q_t$ with any feasible sampling allocation $n$ defined in \eqref{eq:overview-cstar} in an appropriate manner, all the constraints in \eqref{eq:G-program} can be satisfied on good events.
      This is the crucial reason why we can obtain an $\gO(c^*\log T)$-type regret instead of only an $\gO(\log T/\Delta_{\min})$-type regret in the stochastic regime.      
      Meanwhile, in the adversarial regime, 
      we show that mixing the FTRL update $q_t$ with a G-optimal design distribution still satisfies all the constraints in \eqref{eq:G-program}. This thus guarantees a minimax-optimal regret upper bound in the adversarial regime.

\subsection{Regret Upper Bound}\label{sec:finite-upper}

\begin{proposition}
      \label{prop:G-compatibility}
For linear bandits with finite action sets, \IAFTRL{} guarantees
\begin{equation}\label{eq:G-adv}
\Reg = \gO\left(d\log K+\sqrt{dT\log K}\right) \,,
\end{equation}
in the adversarial regime. Simultaneously, for every feasible sampling allocation $n$ in
\eqref{eq:overview-cstar}, \IAFTRL{} guarantees
\begin{equation}
 \Reg
 =
 \gO
 \left(
c^{n}(\gA,\theta)\log K\cdot \log T+M_n+J_{d,K,\Delta_{\min}}\sqrt{C}
 \right) \label{eq:G-compat-cor}
 \,,
\end{equation}
in the corrupted stochastic regime, 
where 
$J_{d,K,\Delta_{\min}}=\gO ( \sqrt{d\log K} \Delta_{\min}^{-129/256} )$, and $M_{n}$ is a finite constant depending on the instance $(\gA,\theta)$ and the sampling allocation \(n\), but independent of \(T\) and \(C\)
(formally defined in \eqref{eq:F-cor-constants}).
\end{proposition}
Note that the remainder $M_n$ in \Cref{prop:G-compatibility} depends on each chosen sampling allocation $n$, so we cannot directly take the infimum of \eqref{eq:G-compat-cor} to obtain the desired $\gO(c^*\log T)$-type regret. 
Instead, we choose one fixed feasible sampling allocation $n^{\circ}$, independent of $T,C$, with $c^{n^{\circ}}(\gA,\theta)\le2\cstar(\gA,\theta)$ 
(see \Cref{lem:F-design-mass}), 
leading to the following result.

\begin{theorem}
      \label{thm:G-main}
For linear bandits with finite action sets, \IAFTRL{} simultaneously guarantees \eqref{eq:G-adv}
in the adversarial regime, and
\begin{equation}\label{eq:G-cor-main}
 \Reg
 =
 \gO
 \left(
 \cstar(\gA,\theta)\log K\cdot \log T
 +
 M_{\cA,\theta}
               +J_{d,K,\Delta_{\min}}\sqrt C 
 \right) \,,
\end{equation}
in the corrupted stochastic regime, where $ M_{\cA,\theta}=M_{n^{\circ}}$ and $J_{d,K,\Delta_{\min}}=\gO ( \sqrt{d\log K} \Delta_{\min}^{-129/256} )$ are instance-dependent constants independent of $T,C$ (formally defined in \eqref{eq:F-cor-constants}).
\end{theorem}

Proofs of \Cref{prop:G-compatibility} and \Cref{thm:G-main} are deferred to
\Cref{app:finite-compatibility-proof} and \Cref{app:finite-results}, respectively.
\subsubsection{Proof Sketch for Proposition~\ref*{prop:G-compatibility}}\label{ps:prop3-1}

We define the parameter-dependent constants
\begin{equation}\label{eq:G-analysis-constants}
 c_{\beta}=\sqrt{d/\log(eK)}\,,\qquad C_{\beta}=16d+c_{\beta}\,,\qquad c_{R}=\sqrt{d\log(eK)}\,.
\end{equation}
They satisfy
$\beta_t=16d+c_{\beta}\sqrt t\le C_{\beta}\sqrt t$ and
$d/\beta_t\le c_{R}/\sqrt t$ for $t\ge1$.

\begingroup
\interdisplaylinepenalty=10000
Fix arbitrary feasible sampling allocation $n$ satisfying \eqref{eq:overview-cstar},
with $c^{n}=c^{n}(\gA,\theta)=\sum_{a\ne a^*}n_a\Delta(a)$ and $N_n=\sum_a n_a<\infty$,
independently of $T,C$. Let
$e_{t,a}=(a-a^*)^\top c_t^{\rm vec}$ and
$\kappa_t=\|c_t^{\rm vec}\|_\cA$.
Thus $|e_{t,a}-e_{t,b}|\le2\kappa_t$, $0\le\kappa_t\le2$, and
$\sum_t\kappa_t\le C$. Define $H(q)=-\sum_aq_a\log q_a$ and recall $ \mathsf D_t=\langle q_t,\widehat\ell_t^{\cA}\rangle
       +\beta_t\log\sum_aq_{t,a}e^{-\widehat\ell_{t,a}/\beta_t}$.
Via the unbiasedness of $\hat\ell_t$ and the canonical analysis for FTRL with Shannon entropy (see, \textit{e.g.}, Exercise~28.12 of \cite{lattimore2019bandit}), for every fixed comparator $u$, it holds that
\begin{align}\label{eq:G-sketch-decomposition}
 \Reg(u)
 \le
 \underbrace{\beta_1\log K+
  \sum_{t=2}^T(\beta_t-\beta_{t-1})\E[H(q_t)]}_{\mathsf{Penalty\text{-}Term}}
 +\underbrace{\sum_{t=1}^T\E[\mathsf D_t]}_{\mathsf{Stability\text{-}Term}}
 +\underbrace{\sum_{t=1}^T\E[\langle p_t-q_t,\Delta\rangle]}_{\mathsf{Exploration\text{-}Term}}
 +\underbrace{\sum_{t=1}^T\E[\langle p_t-q_t,e_t\rangle]}_{\mathsf{Corruption\text{-}Term}}\,.
\end{align}

\paragraph{Step 1: Bounding $\mathsf{Penalty\text{-}Term}$}
Via the definition of $\beta_t$ and $H(q)$, it is straightforward to see $\mathsf{Penalty\text{-}Term}\le\beta_T\log K=\gO( d\log K+\sqrt{dT\log K})$.

Meanwhile, to obtain an ultimate $\gO(\cstar\log T)$-type regret in the stochastic regime, we instead exploit that
little entropy remains once the cumulative loss estimates separate $a^*$ from the suboptimal actions.
Specifically, let $X_{t,a}=\widehat L_{t-1,a}-\widehat L_{t-1,a^*}$ be the difference between the
cumulative loss estimates of action $a$ and the optimal action $a^*$.
On good event $\mathcal G_t$ (formally defined in Step~3(a) below), 
for all $t\ge T_\circ$ ($T_\circ$ is defined in \eqref{eq:F-thresholds-step-4}), 
\Cref{lem:B-good-event-tail} guarantees: 
\begin{align}
      \text{(a)\quad} r_t=a^* \,, \text{\quad(b)\quad} X_{t,a}\ge4t^\alpha\ge4t^{3/4}\ (\forall a\ne a^*)\,, \text{\quad and (c)\quad} q_{t,a}\le e^{-X_{t,a}/\beta_t} \,,
      \label{eq:prop3.1:psketch:X_t_floor}
\end{align}
where $\alpha=\frac34+\frac{13}{10P}$ and $P=2\lceil256+\log(2K)\rceil$.
These further imply 
          \begin{align}
            q_{t,a}\le e^{-X_{t,a}/\beta_t}
          \le e^{-4t^{1/4}/C_{\beta}}\,,\label{eq:prop3.1:psketch:q_t_decay}
          \end{align}
          which means that 
           the probability of each suboptimal action decays rapidly on $\gG_t$, and $q_t$ concentrates on $a^*$.
Therefore, we have
\begin{align}
 \mathsf{Penalty\text{-}Term}
 &=\beta_1\log K
   +\sum_{t=2}^T(\beta_t-\beta_{t-1})\E[\ind_{\mathcal G_t}H(q_t)]
   +\sum_{t=2}^T(\beta_t-\beta_{t-1})\E[\ind_{\mathcal G_t^c}H(q_t)]
       \notag\\
 &\le\beta_1\log K+2c_{\beta}K\sum_{t=2}^Tt^{-1/2}e^{-2t^{1/4}/C_{\beta}}
                    +c_{R}\sum_{t=2}^Tt^{-1/2}\sP(\mathcal G_t^c)
       \label{eq:G-sketch-pen-estimate}\\
 &\le\beta_1\log K+S_{\rm pen}+c_{R}\mathcal E_T
       \label{eq:G-sketch-pen-final}
       \,,
\end{align}
where \eqref{eq:G-sketch-pen-estimate} is because $\ind_{\mathcal G_t}H(q_t)\le2K e^{-2t^{\alpha-1/2}/C_{\beta}}$ using $ q_{t,a}\le e^{-4t^{1/4}/C_{\beta}}$ (see \Cref{lem:F-good-stability}) and $\beta_t-\beta_{t-1}\le c_{\beta}/\sqrt t$; and 
\eqref{eq:G-sketch-pen-final} uses
$\sum_{t\ge1}e^{-t^{1/4}/C_{\beta}}\le24C_{\beta}^4$ (see \Cref{lem:AB-good-sums}), and we define $S_{\rm pen}\coloneq 48c_{\beta}KC_{\beta}^4$ as well as $\mathcal E_T\coloneq \sum_{t=1}^Tt^{-1/2}\sP(\mathcal G_t^c)$.
Thus the good histories contribute the
explicit constant $S_{\rm pen}$, while every failed history costs at most
$c_{R}/\sqrt t$. 

\paragraph{Step 2: Bounding $\mathsf{Stability\text{-}Term}$}
The proxy function constraint \eqref{eq:G-program-cert} immediately gives $\E_{t-1}[\mathsf D_t]\le2c_{R}/\sqrt t$,
and hence a total at most $4c_{R}\sqrt T$. 

As aforementioned, it is nontrivial to control the contribution of $\mathsf{Stability\text{-}Term}$ in the stochastic regime so that an $\gO(\cstar\log T)$ regret is guaranteed, as the update of suboptimal action can be exponentially large in \eqref{eq:G-program-cert}.

We must use the update magnitude constraint and the FTRL exponential weight
$q_{t,a}$ together to control the total stability cost on $\gG_t$.
First notice that
\begin{align}
 \mathsf{Stability\text{-}Term}
 =\sum_{t=1}^T\E[\E_{t-1}[\mathsf D_t]]
 \le\sum_{t=1}^T\E[\gC_t(p_t)]
 =\sum_{t=1}^T\E[\ind_{\mathcal G_t}\gC_t(p_t)]
   +\sum_{t=1}^T\E[\ind_{\mathcal G_t^c}\gC_t(p_t)]\,,
       \label{eq:G-sketch-stab-split}
\end{align}
where the inequality uses $\mathsf D_t \le \gC_t(p_t)$
(see \Cref{lem:B-scalar-stability}).
To bound the good-history sum
$\sum_{t=1}^T\E[\ind_{\mathcal G_t}\gC_t(p_t)]$, let
$d_{a,b}=(a-a^*)^\top v_{t,b}$ and
$\bar d_b=\sum_aq_{t,a}d_{a,b}$ on $\mathcal G_t$.
The update magnitude constraint in \eqref{eq:G-program-mag} is precisely
$|d_{a,b}|\le\beta_t+X_{t,a}/B$, and $d_{a^*,b}=0$.
Therefore, we have
\begin{align}
 \gC_t(p_t)
 &=\beta_t\sum_bp_{t,b}\sum_aq_{t,a}
          \chi((d_{a,b}-\bar d_b)/\beta_t)
 \le\beta_t\sum_bp_{t,b}\sum_{a\ne a^*}q_{t,a}\chi(2d_{a,b}/\beta_t)
       \notag\\
 &\le e^2\beta_t\sum_{a\ne a^*}q_{t,a}e^{2X_{t,a}/(B\beta_t)}
 \le e^2\beta_t\sum_{a\ne a^*}e^{-X_{t,a}/(2\beta_t)}
 \le e^2C_{\beta}K\sqrt t\,e^{-2t^{1/4}/C_{\beta}} 
 \,,
       \label{eq:G-sketch-stab-good}
\end{align}
where the first inequality applies \eqref{eq:chi-center} in \Cref{lem:chi} as well as $d_{a^*,b}=0$; the second inequality follows from $\chi(x)\le e^{|x|}$ and the magnitude constraint $|d_{a,b}|\le\beta_t+X_{t,a}/B$; the third inequality is by \eqref{eq:prop3.1:psketch:q_t_decay} and the fact that $B\ge 4$; and the last inequality uses (b) in \eqref{eq:prop3.1:psketch:X_t_floor}.
Intuitively, this shows that the allowed exponentially large update factor
$e^{2X_{t,a}/(B\beta_t)}$ is dominated by the decaying FTRL weight $q_t$, which eventually guarantees a constant upper bound the total expected stability cost from good histories,
even though the upper bound $\beta_t+X_{t,a}/B$ grows with $X_{t,a}$. 

Substituting \eqref{eq:G-sketch-stab-good} into \eqref{eq:G-sketch-stab-split} yields
\begin{align}
 \mathsf{Stability\text{-}Term}
 &\le e^2C_{\beta}K\sum_{t=1}^T\sqrt t\,e^{-2t^{1/4}/C_{\beta}}
             +2c_{R}\sum_{t=1}^Tt^{-1/2}\sP(\mathcal G_t^c)
       \notag\\
 &\le e^2C_{\beta}^3K\sum_{t\ge1}e^{-t^{1/4}/C_{\beta}}+2c_{R}\mathcal E_T
 \le S_{\rm stab}+2c_{R}\mathcal E_T \,,
       \label{eq:G-sketch-stab-final}
\end{align}
where the first inequality uses the global constraint
$\gC_t(p_t)\le2d/\beta_t\le2c_{R}/\sqrt t$ on $\gG_t^t$;
the second inequality follows from $\sqrt t\,e^{-2t^{1/4}/C_{\beta}}\le C_{\beta}^2e^{-t^{1/4}/C_{\beta}}$; and in the last inequality,
we define $S_{\rm stab}\coloneq 24e^2KC_{\beta}^7$ and
use the same exponential
sum as Step~1. 

Together, \eqref{eq:G-sketch-pen-final} and
\eqref{eq:G-sketch-stab-final} give
\[
 \mathsf{Penalty\text{-}Term}+\mathsf{Stability\text{-}Term}
 \le\beta_1\log K+S_{\rm pen}+S_{\rm stab}+3c_{R}\mathcal E_T\,.
\]
Step~3(a) proves $\mathcal E_T\le E_0+E_1\sqrt C$, with finite
$E_0,E_1$ independent of $T,C$. In the purely stochastic case $C=0$,
this bounds the complete penalty and stability sums uniformly over all
$T$, including the contributions from histories where the cumulative
loss estimates fail to separate $a^*$ from the suboptimal actions. These two contributions
are therefore included in the finite constant $M_n$ in
\eqref{eq:G-compat-cor}. Under corruption, their additional cost is
at most $3c_{R}E_1\sqrt C$.

\paragraph{Step 3: Bounding $\mathsf{Exploration\text{-}Term}$}
First notice that
\begin{equation}\label{eq:G-sketch-expl-def}
 \mathsf{Exploration\text{-}Term}
 =\sum_{t=1}^T\E[\langle p_t-q_t,\Delta\rangle]
 =\sum_{t=1}^T\E[\ind_{\mathcal G_t}\langle p_t-q_t,\Delta\rangle]
  +\sum_{t=1}^T\E[\ind_{\mathcal G_t^c}\langle p_t-q_t,\Delta\rangle] \,,
\end{equation}
where the second sum is at most $8c_{R}\mathcal E_T$, since
$\langle p_t-q_t,\Delta\rangle\le2\Gamma_t\le8c_{R}/\sqrt t$.
This bound alone still gives an $\gO(\sqrt T)$ exploration cost.

To obtain the $\gO(c^{n}\log T)$ regret,
the one-round $\mathsf{Exploration\text{-}Term}$ on $\gG_t$ must instead have an $\gO(c^{n}/t)$ term, the summation of which will lead to the desired $\gO(c^{n}\log T)$ regret bound. Meanwhile, $\mathsf{Exploration\text{-}Term}$ also needs to be bounded by a $T$-independent constant term on bad events.
The is the most challenging part in this proof due to two issues:
(a) the objective function in program \eqref{eq:G-program-objective} minimizes empirical gaps $\widehat\Delta_t$, whereas regret is evaluated on true gaps $\Delta$; and (b) a sampling allocation with low regret cost must yield a comparison distribution satisfying the update magnitude, second-moment, and exponential stability constraints simultaneously.
To this end, in Step~3(a), we show that the $\widehat\Delta_t$ converge to the $\Delta$, in the sense that $\widehat\Delta_t$ is both lower and upper bounded by $\Delta$. Then, in Step~3(b), we demonstrate that a mixture sampling distribution $p_t^c$ carefully constructed with $q_t$ and $n$ satisfies all the required constraints in program \eqref{eq:G-program}. Together, Step~3(a) and Step~3(b) translates the one-round exploration cost $\gO(c^{n}/t)$ of $n$ into the one-round exploration cost $\langle p_t-q_t,\Delta\rangle$.
Step~3(c) sums the total cost of $\mathsf{Exploration\text{-}Term}$.

\paragraph{Step 3(a): Gap Ratios and the Cost of Failure Histories}
The comparison in Step~3(b) needs to turn a small empirical-gap objective
$\langle p_t,\widehat\Delta_t\rangle$ into a small true cost
$\langle p_t,\Delta\rangle$. 
This requires control of both gap underestimation (\textit{i.e.}, $\Delta(a)/\widehat\Delta_t(a)$) and gap overestimation (\textit{i.e.}, $\widehat\Delta_t(a)/\Delta(a)$). 
We first consider the case of $C=0$, then explain what need to be adjusted
when $C>0$.

\begingroup
\setlist[itemize,1]{labelindent=2em}
\subparagraph{(i) Corruption Level $C=0$.}
\begin{itemize}[leftmargin=*,itemsep=4pt,topsep=4pt]
\item \textbf{Constructing the buffered window.}
Let $D_{s,a}\coloneq X_{s+1,a}-X_{s,a}$.
By definitions, we have $ \E_{s-1}[D_{s,a}]=\Delta(a)$ and $\E[X_{t,a}]=\sum_{s<t}\E[D_{s,a}]=(t-1)\Delta(a)$. Meanwhile, if $X_{t,a}>0$ for all $a\ne a^*$, it will hold that $ r_t=a^*$, $Z_{t,a}=X_{t,a}$, and $\widehat\Delta_t(a)=\min\{2,X_{t,a}/t\}$.

Therefore, one may consider bounding the underestimation ratio $\Delta(a)/\widehat\Delta_t(a)$ by $t\Delta(a)/X_{t,a}$. However, this is might be problematic since $X_{t,a}$ can be close to zero with positive probability even when
$\E[X_{t,a}]$ is large, making $t\Delta(a)/X_{t,a}$ to be upper bounded. Obviously, the other obstacle of doing this is that $ r_t\neq a^*$ may happen, in which case $\widehat\Delta_t(a)\neq \min\{2,X_{t,a}/t\}$.
Instead, in this work, we  transfer the constraints to increments of $X_{s,a}$ on a precisely defined set of histories, and derive the needed moments there.
In particular, for sufficiently large $t$,
we make inverse powers well defined throughout a recursive argument on a recent interval of integer rounds $[m_t,t]$, 
where $m_t\asymp t^{25/32}$.
This analysis window $[m_t,t]$ lets us replace the possibly nonpositive $X_{s,a}$ by the positive buffered process $Y_{s,a}^{\dagger,(t)}\coloneq X_{s,a}+2\tau_s^{(t)}$, where $\tau_s^{(t)}\coloneq s^\alpha+4z_t\quad(m_t\le s\le t)$ and $z_t\asymp t^{1/256}$.
To translate the ratio bound $\nicefrac{\Delta(a)}{(Y_{t,a}^{\dagger,(t)}/t)}$ back into $\nicefrac{\Delta(a)}{\widehat\Delta_t(a)}$, we also
require a sufficiently large gap of $X_{t,a}$ at the endpoint $t$. Specifically, we define
\begin{align}
       I_s^{(t)}&\coloneq \ind\{X_{j,a}\ge-\tau_j^{(t)}\,, \forall a\in\cA
                       \text{ and }j\in [m_t,s]\}\,,
                       \label{eq:G-3a-survival-definition}\\
 \mathcal G_t&\coloneq \{I_t^{(t)}=1\}\cap \left(\cap_{a\ne a^*}\{X_{t,a}\ge4\tau_t^{(t)}\}
 \right) \quad(t\ge T_\circ)\,,\quad
 \mathcal G_t\coloneq \varnothing \quad(t<T_\circ) \label{eq:G-3a-clean-good}\,,
\end{align}
so that
\begin{align}
       I_s^{(t)}=1&\Longrightarrow Y_{s,a}^{\dagger,(t)}\ge\tau_s^{(t)}\ge s^\alpha>0\,,
 \quad I_{s+1}^{(t)}\le I_s^{(t)}\,,\quad I_s^{(t)}\in\cF_{s-1} \,,
 \label{eq:G-3a-survival-properties}\\
  \mathcal G_t&\Longrightarrow
 X_{t,a}\le Y_{t,a}^{\dagger,(t)}=X_{t,a}+2\tau_t^{(t)}
       \le\tfrac32X_{t,a}\le2X_{t,a}\quad(a\ne a^*) \,,
 \label{eq:G-3a-clean-comparison}\\
 \mathcal G_t&\Longrightarrow
 \frac{t\Delta(a)}{X_{t,a}}\le\frac{2t\Delta(a)}{Y_{t,a}^{\dagger,(t)}} \,,
 \quad
 \frac{\widehat\Delta_t(a)}{\Delta(a)}
       \le\frac{X_{t,a}}{t\Delta(a)}
       \le\frac{Y_{t,a}^{\dagger,(t)}}{t\Delta(a)}\quad(a\ne a^*) \,,
 \label{eq:G-3a-moment-targets}
\end{align}
where the measurability $I_s^{(t)}\in\cF_{s-1}$ in \eqref{eq:G-3a-survival-properties} holds because $X_{j,a}$ for $j\le s$ is known before round $s$.
Equation \eqref{eq:G-3a-moment-targets} identifies exactly the two moments we need: an inverse moment of
$Y_{t,a}^{\dagger,(t)}=X_{t,a}+2\tau_t^{(t)}$ to control underestimation, and a
positive moment of the same process to control overestimation.
The endpoint condition also gives $r_t=a^*$ and
$X_{t,a}\ge4t^\alpha$, as required by Steps~1 and~2.
Also, the condition $I_s^{(t)}=1$ means that no action has crossed its lower
boundary at any round from $m_t$ through $s$; we call this a \textit{surviving
window}.

\definecolor{stepthreearevisionpurple}{RGB}{112,48,160}

\item \textbf{Bounding inverse and positive moments.}
Our next objective is a recursive bound on
$\E[I_s^{(t)}(Y_{s,a}^{\dagger,(t)})^{-P}]$. Since the program directly controls
only the first two increment moments and the update magnitude, we will
use Taylor's formula for $y^{-P}$ to express its one-round change in
terms of precisely those controlled quantities. 
Define $M_s=\max_b(-X_{s,b})=Z_{s,a^*}$,
$v_s\coloneq \tau_{s+1}^{(t)}-\tau_s^{(t)}=(s+1)^\alpha-s^\alpha$ and
$\xi_{s,a}\coloneq Y_{s+1,a}^{\dagger,(t)}-Y_{s,a}^{\dagger,(t)}=D_{s,a}+2v_s$.
We first verify the increment bounds, without assuming $r_s=a^*$.
Indeed, due to constraints \eqref{eq:G-program-mag} and \eqref{eq:G-program-var},
\Cref{lem:AB-card-compensation}, specifically \eqref{eq:F-local-inc-step-2} shows that
\begin{align}
 \frac{|\xi_{s,a}|}{Y_{s,a}^{\dagger,(t)}}&\le\frac3B\,,
 \qquad
 \E_{s-1}[\xi_{s,a}^2]
       \le8\beta_s+\frac{4(Y_{s,a}^{\dagger,(t)})^2}{Bs}\,,
 \label{eq:G-sketch-local-increment-budgets}\\
 Y_{s+1,a}^{\dagger,(t)}&=Y_{s,a}^{\dagger,(t)}+\xi_{s,a}
       \ge(1-3/B)Y_{s,a}^{\dagger,(t)}>0\,.
 \label{eq:G-3a-next-positive}
\end{align}
Importantly, note that \eqref{eq:G-3a-next-positive} holds on $I_s^{(t)}=1$, even if $I_{s+1}^{(t)}=0$, and thus
\eqref{eq:G-3a-next-positive} ensures the next inverse power is nonnegative and well defined for every one-step continuation of a surviving history, when we remove $I_{s+1}^{(t)}$ in the conditional expectation below.

We now turn these increment bounds into a closed recursion of one-round moment with inverse power via Taylor's expansion. For fixed $a\ne a^*$, let $y=Y_{s,a}^{\dagger,(t)}$ and $\xi=\xi_{s,a}$
on $I_s^{(t)}=1$ in the following. Then, we have
\begin{subequations}\label{eq:G-3a-inverse-chain}
\begin{align}
 \E_{s-1}[(y+\xi)^{-P}]
 &= \E_{s-1}[
      y^{-P}-Py^{-P-1}\xi
       +P(P+1)/2\cdot(y+\eta\xi)^{-P-2}\xi^2
       ]
 \label{eq:G-3a-inverse-taylor0}\\
 &\le y^{-P}-Py^{-P-1}\E_{s-1}[\xi]
                +P(P+1)y^{-P-2}\E_{s-1}[\xi^2]
 \label{eq:G-3a-inverse-taylor}\\
 &\le y^{-P}-P(\Delta(a)+2v_s)y^{-P-1}
       +P(P+1)y^{-P-2}\left(8\beta_s+\frac{4y^2}{Bs}\right)
 \label{eq:G-3a-inverse-substitute}\\
 &\le\left[1+P\left(\frac{8(P+1)\beta_s}{y^2}
                         +\frac{4(P+1)}{Bs}\right)\right]y^{-P}
                   -P\Delta(a) y^{-P-1}
\notag\\
 &\le(1+P/(2s))y^{-P}-P\Delta(a) y^{-P-1} \,,
 \label{eq:G-3a-inverse-conditional}
\end{align}
\end{subequations}
where \eqref{eq:G-3a-inverse-taylor0} uses Taylor's formula for some $\eta\in(0,1)$; \eqref{eq:G-3a-inverse-taylor} holds because (a) $y+\eta\xi\ge y-|\xi|\ge(1-3/B)y>0$ with the last inequality given by the first part in \eqref{eq:G-sketch-local-increment-budgets}; and (b) $(1-3/B)^{-P-2}\le2$;
\eqref{eq:G-3a-inverse-substitute} substitutes $\E_{s-1}[\xi_{s,a}]=\Delta(a)+2v_s$ and the second part of \eqref{eq:G-sketch-local-increment-budgets}; and \eqref{eq:G-3a-inverse-conditional} is because the two curvature coefficients can be controlled as 
(see \Cref{lem:F-local}):
\begin{align}
 \frac{8(P+1)\beta_s}{y^2}
 \le 
 8(P+1)C_{\beta} s^{1/2-2\alpha}
   =\frac1{4s}\frac{32(P+1)C_{\beta}}{s^{2\alpha-3/2}}
   \le\frac1{4s} 
   \,,\quad
 \frac{4(P+1)}{Bs}
 \le\frac1{4s} \,.
\notag
\end{align}
Denote $u_s^{(t)}\coloneq \E[I_s^{(t)}(Y_{s,a}^{\dagger,(t)})^{-P}]$. We further have
\begin{align}
       u_{s+1}^{(t)}
 &=
 \E[I_{s+1}^{(t)}(Y_{s+1,a}^{\dagger,(t)})^{-P}]
\le
 \E[I_{s}^{(t)}(Y_{s+1,a}^{\dagger,(t)})^{-P}]
       =
       \E\left[I_s^{(t)}\E_{s-1}[(Y_{s+1,a}^{\dagger,(t)})^{-P}]\right]
\notag\\
 &\le
 (1+P/(2s))u_s^{(t)}-P\Delta(a)\E[I_s^{(t)}(Y_{s,a}^{\dagger,(t)})^{-P-1}]
 \le
 (1+P/(2s))u_s^{(t)}-P\Delta(a) (u_s^{(t)})^{1+1/P} \,,
 \label{eq:G-sketch-inverse-recursion}
\end{align}
where the first inequality uses $I_{s+1}^{(t)}\le I_s^{(t)}$ and \eqref{eq:G-3a-next-positive}; the second equality is by the fact that $I_s^{(t)}\in\cF_{s-1}$; the second inequality substitutes \eqref{eq:G-3a-inverse-conditional}; and the last inequality applies Jensen's inequality to the nonnegative variable $I_s^{(t)}(Y_{s,a}^{\dagger,(t)})^{-P}$.
Solving the recursion in \eqref{eq:G-sketch-inverse-recursion} leads to (see \Cref{lem:F-local}):
\begin{align}
 \E[I_t^{(t)}(Y_{t,a}^{\dagger,(t)})^{-P}]
 =
 u_t^{(t)}
       \le\left(\frac4{t\Delta(a)}\right)^P \,.
 \label{eq:G-3a-inverse-explicit}
\end{align}

The other target in \eqref{eq:G-3a-moment-targets} is an upper bound
on $\widehat\Delta_t(a)/\Delta(a)$. We therefore need a positive
moment of $Y_{s,a}^{\dagger,(t)}$, which penalizes large estimates.
We follow a similar argument for obtaining \eqref{eq:G-3a-inverse-explicit} to bound the recursive positive moment of $Y_{s,a}^{\dagger,(t)}$. A slight difference is that we need to additionally ensure that the initial value $\E[I_{m_t}^{(t)}(Y_{m_t,a}^{\dagger,(t)})^P]$ at the starting of the recursion inherited from observations before $m_t$ is well bounded.
Overall, we can bound the positive moment as (see \eqref{eq:F-positive-local} in \Cref{lem:B-positive-local}):
\begin{align}\label{eq:G-3a-positive-explicit}
      \E[I_t^{(t)}(Y_{t,a}^{\dagger,(t)}/(t\Delta(a)))^P]
      \le
      (8\Psi_t)^P \,,
\end{align}
where 
$\Psi_t\asymp 1+\Delta_{\min}^{-1} t^{-(255/256)^2}$.

\item \textbf{Bounding gap ratios.}
To obtain the ratios required by the objective comparison in Step 3(b), define
\begin{equation}\label{eq:G-3a-ratio-definition}
 \lo_t\coloneq \max\{1,\max_{a\ne a^*}t\Delta(a)/X_{t,a}\}\,,\qquad
 \up_t\coloneq \max_{a\ne a^*}\widehat\Delta_t(a)/\Delta(a)
       \quad\text{on }\mathcal G_t \,,
\end{equation}
and set $\lo_t=\up_t=0$ on $\mathcal G_t^c$. Via the single-action moment bounds in \eqref{eq:G-3a-inverse-explicit} and \eqref{eq:G-3a-positive-explicit}, we upper bounds the maximum moment bounds over all the actions as

\begin{subequations}\label{eq:G-3a-ratio-moments}
\begin{align}
 \E[\lo_t^P]
 &\le
 1+\sum_{a\ne a^*}(t\Delta(a))^P
                   \E[\ind_{\mathcal G_t}X_{t,a}^{-P}]
 \le
 1+\sum_{a\ne a^*}(2t\Delta(a))^P \E[I_t^{(t)}(Y_{t,a}^{\dagger,(t)})^{-P}]
       \le1+K8^P \,,
 \label{eq:G-3a-ratio-inverse-substitution}\\
 \E[\up_t^P]
 &\le\sum_{a\ne a^*}\E\!\left[
        \ind_{\mathcal G_t}\left(\frac{X_{t,a}}{t\Delta(a)}\right)^P\right]
       \le\sum_{a\ne a^*}\E\!\left[
        I_t^{(t)}\left(\frac{Y_{t,a}^{\dagger,(t)}}{t\Delta(a)}\right)^P\right]
       \le K(8\Psi_t)^P \,,
 \label{eq:G-3a-ratio-positive-substitution}\\
 \|\lo_t\|_P&\le(1+K8^P)^{1/P}\le1+8k\le25 \,,
       \qquad\|\up_t\|_P\le8k\Psi_t\le16\Psi_t \,,
 \label{eq:G-3a-ratio-roots}\\
 \E[\lo_t^j\up_t]
 &\le\|\lo_t\|_P^j\|\up_t\|_P
           \sP(\mathcal G_t)^{1-(j+1)/P}
       \le16\cdot25^j\Psi_t\qquad(j=1,3,5) \,,
 \label{eq:G-3a-ratio-products}
\end{align}
\end{subequations}
where the second inequalities in \eqref{eq:G-3a-ratio-positive-substitution} and \eqref{eq:G-3a-ratio-positive-substitution} use $\ind_{\mathcal G_t}\le I_t^{(t)}$ and \eqref{eq:G-3a-moment-targets}; \eqref{eq:G-3a-ratio-roots} is by
$k=K^{1/P}<7/4<2$; \eqref{eq:G-3a-ratio-products} uses generalized H\"older's inequality
with reciprocal exponents $j/P$, $1/P$, and $1-(j+1)/P$ ($P\ge512>6$ ensures these exponents are valid). 
Collectively, on $\mathcal G_t$, \eqref{eq:G-3a-ratio-definition}, together with the fact $\Delta(a)/X_{t,a}$, gives
\begin{equation}\label{eq:G-3a-clipping}
 \lo_t\widehat\Delta_t(a)
 =\min\{2\lo_t,\lo_t X_{t,a}/t\}
 \ge\min\{2,\Delta(a)\}=\Delta(a)\,,\qquad
 \lo_t\widehat\Delta_t(a)\le\lo_t\up_t\Delta(a)\quad(a\ne a^*)\,.
\end{equation}
Note that for $a=a^*$, the comparisons hold with both gaps zero.

\item \textbf{Bounding failure probabilities and their weighted cost.}
It remains to control the probability of the failure event $\mathcal G_t^c$. By definition \eqref{eq:G-3a-clean-good}, we have $ \sP(\mathcal G_t^c)
\le\sP(I_t^{(t)}=0)+\sum_{a\ne a^*}\sP(I_t^{(t)}=1,X_{t,a}<4\tau_t^{(t)})$. For the first part, we use a recursive moment argument to bound a crossing $X_{s,a}\le -\tau^{(t)}_s$ anywhere in $[m_t,t]$ (see \Cref{lem:F-window}). This yields
\begin{align}
      \sP(I_t^{(t)}=0)&\le\mathsf A t^{-65/64}+(n_0/t)^{r_0}\,,\label{eq:G-3a-window-tail}
\end{align}
where $\mathsf A=\exp(\Theta(P\log(PC_\beta)))$ and $r_0\asymp \log K$ is specified in (see \eqref{eq:F-exponents-step-4}).
For the second part, applying Markov inequality to the
nonnegative inverse power and then using \eqref{eq:G-3a-inverse-explicit} guarantee that
\begin{align}
       \sP(I_t^{(t)}=1,X_{t,a}<4\tau_t^{(t)})
 &\le(7\tau_t^{(t)})^P\E[I_t^{(t)}(Y_{t,a}^{\dagger,(t)})^{-P}]
       \le\left(\frac{28\tau_t^{(t)}}{t\Delta(a)}\right)^P \,.
 \label{eq:G-3a-clean-markov}
\end{align}

The costs on $\mathcal G_t^c$ in Steps~1--3 are proportional to
$t^{-1/2}$.
Therefore the relevant target is the
weighted sum $\mathcal E_T=\sum_{t=1}^Tt^{-1/2}\sP(\mathcal G_t^c)$.
Substitution of $\tau_t^{(t)}$ into
\eqref{eq:G-3a-clean-markov} and summation of the resulting tails
will give $\mathcal E_T <\infty$ at $C=0$.

\end{itemize}

\subparagraph{(ii) Corruption Level $C>0$.}
\begin{itemize}[leftmargin=*,itemsep=4pt,topsep=4pt]
\item \textbf{Compensating the drift.}

The first change occurs in the inverse-power Taylor step
\eqref{eq:G-3a-inverse-substitute}: keeping the old buffer
$X_{s,a}+2\tau_s^{(t)}$ would introduce the extra linear term
$-Pe_{s,a}y^{-P-1}$, whose sign is uncontrolled, so the previous
inverse-moment recursion no longer follows. We remove this term by
subtracting $E_{s,a}\coloneq \sum_{j<s}e_{j,a}$ and add $2C$ to keep the
correction nonnegative, since $|E_{s,a}|\le2C$. Define
\begin{align}
 Y_{s,a}^{\dagger,(t)}\coloneq X_{s,a}-E_{s,a}+2C+2\tau_s^{(t)}
 \,,\quad
 \xi_{s,a}\coloneq Y_{s+1,a}^{\dagger,(t)}-Y_{s,a}^{\dagger,(t)}
 =D_{s,a}-e_{s,a}+2v_s \,.\label{eq:G-3a-corrupt-buffer}
\end{align}
Then $\E_{s-1}[\xi_{s,a}]=\Delta(a)+2v_s$, so similar uncorrupted inverse- and positive-power Taylor arguments apply.

For the probability of failure event, we also modify the
boundary-crossing argument behind \eqref{eq:G-3a-window-tail}.
Its negative-part Taylor step discards a nonpositive first-order term;
the raw drift $\Delta(a)+e_{s,a}$ no longer guarantees that sign. Let $b_s^{\rm rem}\coloneq C-\sum_{j<s}\kappa_j$ and
$W_{s,a}\coloneq X_{s,a}-2b_s^{\rm rem}$.
Since $b_{s+1}^{\rm rem}=b_s^{\rm rem}-\kappa_s$, the shift contributes
$2\kappa_s$ to the increment of $W_{s,a}$, giving
$\E_{s-1}[W_{s+1,a}-W_{s,a}]
=\Delta(a)+e_{s,a}+2\kappa_s\ge\Delta(a)$.
Thus the same moment argument applies to
$\sum_a(-W_{s,a})_+^P$
(\Cref{lem:AB-card-shift,lem:F-negative,lem:F-window}). Use the same window construction with
$n_0=\max\{N_{\rm burn},\lceil C\rceil\}$, so $m_t\ge C$ and
$\tau_s^{(t)}\ge4C$, and define $I_s^{(t)}$ by survival of every
$W_{j,a}> -\tau_j^{(t)}$ for $m_t\le j\le s$.
On survival, defining $\Lambda_{s,a}\coloneq 2b_s^{\rm rem}-E_{s,a}+2C\in[0,4C]$
gives $Y_{s,a}^{\dagger,(t)}=W_{s,a}+2\tau_s^{(t)}+\Lambda_{s,a}
\ge\tau_s^{(t)}$ and
$Z_{s,a}+M_s\le W_{s,a}+2\tau_s^{(t)}\le Y_{s,a}^{\dagger,(t)}$.
This transfers the original algorithmic constraints to the new buffer,
just as in the uncorrupted argument. Subtracting the $\gF_{s-1}$-measurable term $e_{s,a}$
leaves the conditional variance unchanged, while $|e_{s,a}|\le4$ and
$v_s\le1$ add only constants absorbed by $\beta_s\ge16$.
Consequently, the same Taylor remainder bounds hold (see \Cref{lem:AB-card-compensation}):
\begin{align}
       \frac{|\xi_{s,a}|}{Y_{s,a}^{\dagger,(t)}}\le\frac3B \,,
 \quad
 \E_{s-1}[\xi_{s,a}^2]
 \le8\beta_s+\frac{4(Y_{s,a}^{\dagger,(t)})^2}{Bs}
 \quad\text{on }I_s^{(t)}=1 \,.\label{eq:G-3a-corrupt-local-bounds}
\end{align}

\item \textbf{Recovering moments and gap ratios.}
Applying similar recursive moment arguments leads to the same inverse recursion \eqref{eq:G-sketch-inverse-recursion} and
positive-moment bound \eqref{eq:G-3a-positive-explicit}; 
the extra
$\gO(C)$ initial shift can be absorbed into $\gO(m_t)$ because $m_t\ge C$.
For $t\ge T_\circ$, define $\mathcal G_t$ by adding
$W_{t,a}\ge4\tau_t^{(t)}$ for all $a\ne a^*$ to $I_t^{(t)}=1$,
and keep $\mathcal G_t=\varnothing$ for $t<T_\circ$.
The bounds $|E_{t,a}|\le2C$ and $0\le\Lambda_{t,a}\le4C\le\tau_t^{(t)}$
then give $r_t=a^*$ and
$X_{t,a}\le Y_{t,a}^{\dagger,(t)}\le2X_{t,a}$ for $a\ne a^*$.
Hence the same gap-ratio bounds in \eqref{eq:G-3a-ratio-moments} can be recovered.
\item \textbf{Bounding failure probabilities and their weighted cost.}
Finally, substituting $X_{t,a}$ with $W_{t,a}$ in the gap event, reusing the crossing/small-endpoint-gap split and the inverse-moment bound with Markov inequality leads to a similar bound in \eqref{eq:G-3a-clean-markov}.
The enlarged $n_0$ changes the scale of the tails, while upper bounding them at
one and summing with weights $t^{-1/2}$ still gives
$\mathcal E_T\le E_0+E_1\sqrt C$, with $E_0,E_1$ independent of $T,C$
(\Cref{lem:F-sums}): the budget enters through
$\sqrt{n_0}\le\sqrt{N_{\rm burn}}+\sqrt C$.

\end{itemize}
\endgroup

\paragraph{Step 3(b): Verifying the Mixture Distribution Induced by the Sampling Allocation}
Recall $n$ is an arbitrary feasible sampling allocation in \eqref{eq:overview-cstar}, with
$H_n=\sum_a n_a a a^\top$, regret cost $c^{n}=\sum_{a\ne a^*}n_a\Delta(a)$, and mass $N_n=\sum_a n_a$.
The key idea in this part is to show that mixing FTRL update $q_t$ with this sampling allocation $n$ with an exploration strength of $\Theta(1/t)$ satisfies all the constraints required in \eqref{eq:G-program-mix}--\eqref{eq:G-program-cert}.

Specifically, set $a_n=\max\{1,4BC_{\beta}N_n/d\}$, $a_{\Delta_{\min}}=64B\Delta_{\min}^{-2}$ and
\[
 \rho_t=16B\lo_t^2/t\,,
 \quad
 \mathcal W_t=\mathcal G_t\cap\{\lo_t^2\le\sqrt t/a_n\}
                      \cap\{\lo_t^2\le t/a_{\Delta_{\min}}\}\,,
                      \quad
 w_{t,a}^c=\rho_tn_a\ind\{a\ne a^*\} \,.
\]
On $\mathcal W_t$, let $\gamma_t^c=\sum_aw_{t,a}^c$,
$p_t^c=(1-\gamma_t^c)q_t+w_t^c$, $M_c=\sum_bp_{t,b}^cbb^\top$,
and $v_b^c=v_b(p_t^c)=M_c^{-1}b$.
We show $p_t^c$ satisfies the constraints in \eqref{eq:G-program} below.
\begin{itemize}[leftmargin=*,itemsep=4pt,topsep=4pt]

\item \textbf{Verifying constraint \eqref{eq:G-program-dom}.}
The condition $\lo_t^2\le\sqrt t/a_n$ in $\mathcal W_t$ gives $\rho_tN_n\le\Gamma_t$ (see \eqref{eq:rev-B-witness-mass-step-1}).
By \Cref{sec:G-algorithm}, 
$\Gamma_t=4d/\beta_t\le1/4$.
Since $n_a\ge0$ and $\rho_t\ge0$, we have $w_{t,a}^c\ge0$ and
$\gamma_t^c=\rho_t\sum_{a\ne a^*}n_a\le\rho_tN_n\le\Gamma_t$,
which verifies the nonnegativity and exploration-mass constraints.

\item \textbf{Verifying constraint \eqref{eq:G-program-var}.}
On $\mathcal G_t$, summing the suboptimal-mass bound \eqref{eq:prop3.1:psketch:q_t_decay} over $a\ne a^*$ yields $q_{t,a^*}\ge1-Ke^{-4t^{1/4}/C_\beta}\ge3/4$ for $t\ge T_\circ$ (see \eqref{eq:rev-B-witness-domination-step-1}).
Since $w_{t,a^*}^c=0$, the exploration-mass bound
$\rho_tN_n\le\Gamma_t$ therefore yields
\begin{align}\label{eq:G-sketch-optimal-mass}
 \gamma_t^c
 &
 =\rho_t\sum_{a\ne a^*}n_a\le\rho_tN_n\le\Gamma_t\le\tfrac14
 \,,
 \quad
 p_{t,a^*}^c
 =(1-\gamma_t^c)q_{t,a^*}
   \ge(1-\Gamma_t)\tfrac34 \ge\tfrac12 \,, 
\end{align}
and $ \rho_tn_{a^*}\le\rho_tN_n\le\Gamma_t\le\tfrac14$.
Hence $p_{t,a^*}^c\ge \rho_tn_{a^*}$ and $M_c\succeq\rho_tH_n$.
Denote $\lambda_a=a^\top M_c^{-1}a$. 
Since $n$ is a feasible sampling allocation in \eqref{eq:overview-cstar}, we have $\lambda_a
\le \rho_t^{-1}a^\top H_n^{-1}a
\le\Delta(a)^2/(2\rho_t)$ for $a\ne a^*$.
Besides, for optimal action $a^*$, we have
\begin{align}\label{eq:G-sketch-optimal-leverage-inequality}
\lambda_{a^*}
 =
 (M_c^{-1}a^*)^\top M_c(M_c^{-1}a^*)
 \ge
  p_{t,a^*}^c\bigl((a^*)^{\top} M_c^{-1}a^*\bigr)^2
   =
   p_{t,a^*}^c\lambda_{a^*}^2 \,,
\end{align}
which along with \eqref{eq:G-sketch-optimal-mass} implies that $\lambda_{a^*}\le1/p_{t,a^*}^c \le 2$.
Thus, for $a\ne a^*$,
\begin{equation}\label{eq:G-sketch-design-centered-moment}
 (a-a^*)^\top M_c^{-1}(a-a^*)
 \le2\lambda_a+2\lambda_{a^*}
 \le\frac{\Delta(a)^2}{\rho_t}+4
 \le\frac{2\Delta(a)^2}{\rho_t} \,,
\end{equation}
where the last inequality uses $4\rho_t\le\Delta_{\min}^2\le\Delta(a)^2$ implied by the condition $\lo_t^2\le t/a_{\Delta_{\min}}$ on $\gW_t$ (see \eqref{eq:rev-B-witness-mass-step-2}).
\eqref{eq:G-sketch-design-centered-moment}, together with the fact that $r_t=a^*$ on $\mathcal W_t$ and $|Y|\le1$, demonstrates that
\begin{align}
 \sup_{Q\in\mathcal Q}
 \E_{\substack{A\sim p_t^c\\Y\sim Q_A}}
 \!\left[d_t(a;M_c^{-1}A,Y)^2\mid\cF_{t-1}\right]
 &=\sum_bp_{t,b}^c\bigl((a-a^*)^\top M_c^{-1}b\bigr)^2
  =(a-a^*)^\top M_c^{-1}(a-a^*)
\notag
       \\
 &\le\frac{2\Delta(a)^2}{\rho_t}
  =\frac{t\Delta(a)^2}{8B\lo_t^2}
  \le\frac{X_{t,a}^2}{8Bt}
  \le\beta_t+\frac{X_{t,a}^2}{Bt} \,,\notag
\end{align}
where the second inequality uses $t\Delta(a)\le\lo_tX_{t,a}$.

\item \textbf{Verifying constraint \eqref{eq:G-program-mag}.}
\Cref{lem:AB-physical-witness} gives $ \max_b|(a-a^*)^\top M_c^{-1}b|
 \le\frac{X_{t,a}}{8B}\le\beta_t+\frac{X_{t,a}}B$.
For $a=a^*$, the estimated
difference is zero.

\item \textbf{Verifying constraint \eqref{eq:G-program-cert}.}
For the same reconstruction, \Cref{lem:AB-physical-witness}
also proves $\gC_t(p_t^c)\le2d/\beta_t$.

\item \textbf{Objective comparison and worst-case feasibility.}
Using the feasibility and optimality of $p_t$ in \eqref{eq:G-program},
\begin{align}
 \langle p_t,\Delta\rangle
 &\le\lo_t\langle p_t,\widehat\Delta_t\rangle
 \le\lo_t\langle p_t^c,\widehat\Delta_t\rangle
=
\lo_t(1-\gamma_t^c)\langle q_t,\widehat\Delta_t\rangle
                  +\lo_t\rho_t\sum_{a\ne a^*}n_a\widehat\Delta_t(a)
 \notag\\
&\le
 \lo_t\up_t\langle q_t,\Delta\rangle
                  +\frac{16Bc^{n}}{t}\lo_t^3\up_t \,,
       \label{eq:G-sketch-objective-cost}
\end{align}
where the first inequality uses
\eqref{eq:G-3a-clipping}; the second one is by the optimality of $p_t$; and the last inequality again follows from \eqref{eq:G-3a-clipping}.
Besides, on $\mathcal G_t\setminus\mathcal W_t$, the feasible G-optimal law of
\Cref{lem:AB-fallback} gives
$\langle p_t,\Delta\rangle\le\lo_t\up_t\langle q_t,\Delta\rangle
+2\Gamma_t\lo_t\up_t$.
\end{itemize}

\paragraph{Step 3(c): Summing the Exploration Cost}
For $x>0$, \eqref{eq:G-3a-clipping} implies
\begin{equation}\label{eq:G-sketch-mass-tail}
 \E[\lo_t\up_t\ind\{\lo_t^2>x\}]
 \le\min\{\E[\lo_t\up_t],x^{-2}\E[\lo_t^5\up_t]\}
 \le\Psi_t\min\{400,16\cdot25^5x^{-2}\} \,.
\end{equation}
With
$m_t=\min\{400,16\cdot25^5a_n^2/t\}
      +\min\{400,16\cdot25^5a_{\Delta_{\min}}^2/t^2\}$,
 combining
\eqref{eq:G-sketch-objective-cost} on $\mathcal W_t$ with
$\langle p_t,\Delta\rangle\le\lo_t\up_t\langle q_t,\Delta\rangle
+2\Gamma_t\lo_t\up_t$ on $\mathcal G_t\setminus\mathcal W_t$ and substituting it in \eqref{eq:G-sketch-expl-def} gives (see \Cref{lem:F-exploration})
\begin{align}
 \mathsf{Exploration\text{-}Term}
 &\le\sum_{t=T_\circ}^T
       \left\{\frac{2^{22}Bc^{n}\Psi_t}{t}
        +800K\Psi_t e^{-4t^{1/4}/C_{\beta}}
        +8c_{R}\Psi_t t^{-1/2}m_t\right\}+8c_{R}\mathcal E_T
       \notag\\
 &\lesssim
 Bc^{n}\log(eT)
       +
       BA_0c^{n}\Delta_{\min}^{-\varepsilon_0}
       +\Psi_{\Delta_{\min}} S_b
 +c_{R}\Psi_{\Delta_{\min}}(a_n+\sqrt{a_{\Delta_{\min}}})+c_{R}\mathcal E_T
       \notag\\
 &\lesssim
 Bc^{n}\log(eT)+M_{3,n}+c_{R}\mathcal E_T\,,
       \label{eq:G-sketch-expl-final}
\end{align}
where $ S_b\asymp KC_{\beta}^4$ and $M_{3,n}\asymp BA_0c^{n}\Delta_{\min}^{-\varepsilon_0}
       +\Psi_{\Delta_{\min}} S_b+c_{R}\Psi_{\Delta_{\min}}(a_n+\sqrt{a_{\Delta_{\min}}})$.
In \eqref{eq:G-sketch-expl-final}, the first inequality is by
$\langle q_t,\Delta\rangle\le2Ke^{-4t^{1/4}/C_{\beta}}$ on $\mathcal G_t$, which follows from \eqref{eq:prop3.1:psketch:q_t_decay} and $0\le\Delta(a)\le2$;
and the second inequality uses $\Psi_t\le\Psi_{\Delta_{\min}}$ and
$\sum_{t\ge T_\circ}(\Psi_t-1)/t\le3A_0\Delta_{\min}^{-\varepsilon_0}$ from
\Cref{lem:B-positive-local}, the exponential sum from
\Cref{lem:AB-good-sums}, and the two mass sums
\eqref{eq:F-mass-sum1}--\eqref{eq:F-mass-sum2} from
\Cref{lem:B-mass-sums}. 

\paragraph{Step 4: Bounding $\mathsf{Corruption\text{-}Term}$}
Since $(a^*)^{\top} c_t^{\rm vec}\sum_a(p_{t,a}-q_{t,a})=0$,
we have
\begin{align}
 \mathsf{Corruption\text{-}Term}
 &=\sum_{t=1}^T\E[\langle p_t-q_t,e_t\rangle]
 =\sum_{t=1}^T\E[\sum_a(w_{t,a}-\gamma_tq_{t,a})a^\top c_t^{\rm vec}]
       \notag\\
 &\le\sum_{t=1}^T\E[2\gamma_t\kappa_t]
 \le8c_{R}\,\E[\sum_{t=1}^T\kappa_t/\sqrt t]
 \le16\sqrt2c_{R}\sqrt C \,,
       \label{eq:G-sketch-cor-final}
\end{align}
where the first inequality uses
$|a^\top c_t^{\rm vec}|\le\kappa_t$ and $\sum_aw_{t,a}=\gamma_t$, the second inequality is by $\gamma_t\le \Gamma_t \le4c_{R}/\sqrt t$, and the last inequality follows from
the budget bound for corruption in \eqref{eq:cor-weighted} of
\Cref{lem:capped}. 

\paragraph{Putting the Four Terms Together}
Define the finite constants
\begin{equation}\label{eq:G-sketch-final-constants}
 M_n\coloneq \beta_1\log K+S_{\rm pen}+S_{\rm stab}+M_{3,n}+11c_{R}E_0\,,
 \qquad J\coloneq c_{R}(11E_1+16\sqrt2) \,.
\end{equation}
Substituting
\eqref{eq:G-sketch-pen-final}, \eqref{eq:G-sketch-stab-final},
\eqref{eq:G-sketch-expl-final}, and \eqref{eq:G-sketch-cor-final} into \eqref{eq:G-sketch-decomposition} yields
\begin{align}
 \Reg(u)
 &\lesssim Bc^{n}\log(eT)+\beta_1\log K+S_{\rm pen}+S_{\rm stab}
          +M_{3,n}
          +
          c_{R}\mathcal E_T
          +
          c_{R}\sqrt C
       \notag\\
 &\lesssim
 Bc^{n}\log(eT)+M_n+J\sqrt C
  \,,
       \label{eq:G-sketch-total-final}
\end{align}
where $J\lesssim c_{R}\Delta_{\min}^{-129/256}$,
\eqref{eq:G-sketch-total-final} uses
$\mathcal E_T\le E_0+E_1\sqrt C$, with $E_0,E_1$ independent of $T,C$
(see \Cref{lem:F-sums})
and \eqref{eq:G-sketch-final-constants}.
Meanwhile, in the adversarial regime, the global upper bound on the stability term and exploration on every history (see \Cref{prop:AB-adversarial}) gives
\begin{equation}\label{eq:G-sketch-adversarial}
 \Reg\le\min\{2T,\beta_T\log K+10d\sum_{t\le T}\beta_t^{-1}\}
 \lesssim
 \min\{T, d\log(eK)+\sqrt{dT\log(eK)}\} \,.
\end{equation}
\endgroup

\subsection{Regret Lower Bound}\label{sec:finite-lower-main}
The
 lower bound of $\Omega(c^*\log T)$ for the leading term from \cite{LS17} applies to our problem.
We now give a lower bound
showing that the multiplicative coefficient in the $\sqrt C$ corruption term
can grow even when $c^*$ remains bounded.
In the following, 
$\Reg[T,\sigma]$ denotes the regret in the environment indexed by $\sigma$.

\begin{theorem}
      \label{thm:main-finite-lower}
Let $\cA=\{-e_1,e_1,e_2\}$ and
$\theta_\sigma=(\sigma\Delta_{\min}/2,1/2)$, with $\sigma=\pm1$ and
$0<\Delta_{\min}\le1/2$. Both instances have minimum gap $\Delta_{\min}$ and $ \cstar(\cA,\theta_\sigma)=\frac{4}{1+\Delta_{\min}}=\Theta(1)$
for all
$0<\Delta_{\min}\le1/2$.
For every corruption budget $C\ge\Delta_{\min}^{-1}$
and every integer $T\ge\lceil\Delta_{\min}^{-2}\rceil$, there are two corrupted stochastic environments indexed by $\sigma=\pm1$, each with loss vector
$\theta_\sigma$,
such that every policy satisfies 
$ \max_{\sigma=\pm1}\Reg[T,\sigma]
 \ge\frac1{16}\min\{\sqrt T,\sqrt{C/\Delta_{\min}}\}$ in the corrupted stochastic regime with budget at most $C$.
\end{theorem}

\begin{proof}[Proof sketch of \Cref{thm:main-finite-lower}]\label{ps:thm3-2}
      For every integer $n\ge\Delta_{\min}^{-2}$, we construct two
corrupted environments on $\cA$, indexed by $\sigma=\pm1$ and with
respective fixed loss vectors $\theta_\sigma$. In each environment, the
per-round corruption is $\Delta_{\min}/2-1/(8\sqrt n)\in[3\Delta_{\min}/8,\Delta_{\min}/2]$,
and the optimal action is unchanged by corruption. The total corruption
used is therefore
$C_{\Delta_{\min},n}=n(\Delta_{\min}/2-1/(8\sqrt n))\in[3\Delta_{\min} n/8,\Delta_{\min} n/2]$.
A two-environment testing argument gives, for every policy,
$\max_{\sigma=\pm1}\Reg[n,\sigma]\ge\sqrt n/16
\ge(\sqrt2/16)\sqrt{C_{\Delta_{\min},n}/\Delta_{\min}}$.

Now fix the prescribed budget cap $C\ge\Delta_{\min}^{-1}$ and horizon
$T\ge\lceil\Delta_{\min}^{-2}\rceil$.
If $C\ge\Delta_{\min} T/2$, take $n=T$ and use the construction in all $T$
rounds, with per-round corruption $\Delta_{\min}/2-1/(8\sqrt T)$ and total
corruption $C_{\Delta_{\min},T}\le\Delta_{\min} T/2\le C$.
If $\Delta_{\min}^{-1}\le C<\Delta_{\min} T/2$, take
$n=\lfloor2C/\Delta_{\min}\rfloor\ge\Delta_{\min}^{-2}$ and use the construction
in the first $n$ rounds, with per-round corruption
$\Delta_{\min}/2-1/(8\sqrt n)$ and total corruption
$C_{\Delta_{\min},n}\le n\Delta_{\min}/2\le C$; use the fixed loss vector
$\theta_\sigma$ thereafter, with no further corruption.

In both cases, $n=\min\{T,\lfloor2C/\Delta_{\min}\rfloor\}
\ge\min\{T,C/\Delta_{\min}\}$. The optimal action in each environment is
unchanged throughout, so any remaining rounds add nonnegative regret.
Hence $\max_{\sigma=\pm1}\Reg[T,\sigma]
 \ge\frac{\sqrt n}{16}
 \ge\frac1{16}\min\{\sqrt T,\sqrt{C/\Delta_{\min}}\}$.
\end{proof}

We can even construct a one-dimensional example on $\cA^{\prime}=\{-1,1\}$
with $\theta_\sigma=\sigma\Delta_{\min}/2$, for which $\cstar=0$ but the same
lower bound $\max_{\sigma=\pm1}\Reg[T,\sigma]
\ge\frac1{16}\min\{\sqrt T,\sqrt{C/\Delta_{\min}}\}$ holds for every
$C\ge\Delta_{\min}^{-1}$ and integer $T\ge\lceil\Delta_{\min}^{-2}\rceil$ (see \Cref{ch:finite-lower}).
Thus a bound on $\cstar$ alone does not bound the multiplicative coefficient
in the $\sqrt C$ corruption term.
On the other hand, we note that our upper bound's extra $\Delta_{\min}^{-1/256}$ dependence remains unmatched.

\section{Sharp Action Sets}\label{sec:overview-sharp}
\definecolor{algTwoRevisionPurple}{RGB}{112,48,160}
In this section, we first introduce geometric information-acquisition FTRL
(\GIAFTRL{}; \Cref*{alg:sharp}), which extends the
information-acquisition guarantee to sharp compact action sets.
Then, in \Cref{sec:shap-upper-main} and \Cref{sec:sharp-lower-main}, we present the regret upper and lower bounds, respectively.

\subsection{Algorithm}\label{sec:SH-algorithm}
We run \GIAFTRL{} on the known compact generator
$\mathcal B$ defined in \Cref{ch:overview}.
In particular, \GIAFTRL{} uses FTRL with relative-entropy regularization, 
equivalently the continuous exponential-weights method, on the compact domain
$\mathcal K=\Conv(\mathcal B)$ \cite{HoevenErvenKotlowski2018}.
However, a key difference between our algorithm and the continuous exponential-weights algorithm for linear bandits of \cite{HoevenErvenKotlowski2018} is that their algorithm uses the uniform
volume measure as its reference measure, whereas our \GIAFTRL{} uses a mixture of a discrete measure and a continuous
measure in its relative-entropy regularizer.
In addition, as in \IAFTRL{}, a sampling distribution $p_t$ will be computed based on the FTRL update $q_t$ by solving an optimization problem, but the optimization problem is now combined with geometrically scaled moment constraints.
In the following, recall that $\mathcal S=\Conv(\mathcal B\cup(-\mathcal B))$ is the symmetric convex hull of $\mathcal B$, and
we assume $k=\dim\operatorname{aff}(\mathcal K)\ge1$.

\paragraph{Loss Estimates and FTRL Updates}
Let $\gP(\mathcal K)$ denote the set of probability measures on
$\mathcal K$. 
As in \IAFTRL, for any candidate sampling probability measure $p\in\gP(\mathcal K)$
with $M(p)\succ0$, we define $ M(p)=\int_{\mathcal K}xx^\top p(dx)$, and $v_x(p)=M(p)^{-1}x $ for all $x\in\mathcal K$.
At round $t$, \GIAFTRL{} draws $X_t\sim p_t$, where the sampling distribution $p_t$ is
specified by \eqref{eq:SH-program} below.
If $X_t\in\mathcal B$, the learner takes $A_t=X_t$. Otherwise, we decompose $X_t$ as a convex combination $X_t=\sum_{i=1}^m\lambda_i a_i$ over $\gB$, where $m\le d+1$,
$a_i\in\mathcal B$, $\lambda_i\ge0$, and $\sum_i\lambda_i=1$, using a prescribed measurable rule. Then the learner draws an index $i$ with probability $\lambda_i$ and plays $A_t=a_i$, which implies that $\E[A_t\mid\cF_{t-1},X_t,v_{X_t}(p_t)]=X_t$.
After observing the feedback $Y_t$, we construct $\widehat\ell_{t,x}=x^\top\widehat\theta_t$ and update $\widehat L_{t,x}=\widehat L_{t-1,x}+\widehat\ell_{t,x}$.

At the beginning of round $t$, we use the learning rate $\beta_t=16d+\sqrt t$ and compute the FTRL update
\begin{equation}\label{eq:SH-ftrl}
 q_t=\argmin_{q\in\gP(\mathcal K)}
 \left\{\int_{\mathcal K}\widehat L_{t-1,x}\,q(dx)
                      +\beta_t\KL(q,\mu)\right\} \,,
\end{equation}
where $\mu$ is a fixed reference measure constructed below from the
known action geometry.  
The unique minimizer of \eqref{eq:SH-ftrl} has the exponential-weights form $q_t(dx)=\frac{\exp(-\widehat L_{t-1,x}/\beta_t)}{\mathcal Z_t} \mu(dx)$, where $\mathcal Z_t=\int_{\mathcal K}\exp(-\widehat L_{t-1,z}/\beta_t)\mu(dz)$. We define $\bar x_t=\int_{\mathcal K}xq_t(dx)$.

\paragraph{Reference Measure}
Denote by $\delta_x$ the Dirac probability measure which places all its mass at $x\in\gK$. 
In this work, we call $x$ an \textit{atom} of $\mu$ if $\mu(\{x\})>0$.
For finite action sets, the (negative) Shannon entropy regularizer in \IAFTRL{} implicitly uses the
uniform reference distribution $\mu=K^{-1}\sum_{a\in\cA}\delta_a$.
 On an infinite action set, we need to specify a reference measure to define the density and its relative
entropy. The choice also determines which comparators have finite
KL cost and how quickly the FTRL distribution $q_t$ concentrates.
In the adversarial regime, we need to compete with arbitrary comparators. This requires $\mu$ to have a continuous component with support
$\mathcal K$, which supplies quantitative coverage near every adversarial comparator. 
Meanwhile, in the stochastic regime, we need to compare with always playing the unknown optimal action $a^*$, requiring a bounded $\KL(\delta_{a^*},\mu)$. Since $\KL(\delta_{a^*},\mu)=\log(1/\mu(\{r\}))$ if $\mu(\{r\})>0$ and $\KL(\delta_{a^*},\mu)=\infty$ otherwise, this in turn requires a reference measure $\mu$ that puts positive probability mass on each possible optimal action $a$.

However, obtaining $\gO(c^*\log T)$-type regret in the stochastic regime also requires $q_t$ to converge fast to $\delta_{a^*}$, while finite $\KL(\delta_{a^*},\mu)$ alone does not provide the required rates. To see this, let $I_t=\int_{a\ne a^*}
                 \exp(-(\widehat L_{t-1,a}-\widehat L_{t-1,a^*})/\beta_t)\mu(da)$. Then we have $ q_t(\mathcal K\setminus\{a^*\})=\frac{I_t}{\pi^*+I_t}$.
                 The numerator is
controlled by the mass of suboptimal actions whose estimated loss
differences $\widehat L_{t-1,a}-\widehat L_{t-1,a^*}$ are still small enough to receive substantial exponential weight.
Consequently, both the discrete and continuous components of $\mu$ must put sufficiently little mass on suboptimal actions with very small gaps.
The construction of $\mu$ below ensures that the total reference
mass of all actions $a$ with $0<\Delta(a)\le\varepsilon$ decreases as a
high power of $\varepsilon$, for both its discrete and continuous parts.

\begin{itemize}[leftmargin=*,itemsep=4pt,topsep=4pt]
 \item \textbf{Discrete measure: candidate actions and prior weights.}
The unique optimal action $a^*$ for a fixed $\theta$ is an extreme point of
$\mathcal K=\Conv(\mathcal B)$ and belongs to $\mathcal B$.
However, a compact convex set may have uncountably many extreme
points, whereas a probability measure can assign positive mass to
only countably many points. 
Fortunately, the sharpness condition in \eqref{eq:SH-overview-sharp} allows us to construct a
countable family containing every possible optimal action, using
only the known generator $\mathcal B$.
Let $\mathbb Q^d$ be the set of vectors with $d$ rational coordinates.
It is countable, so fix an enumeration $(v^{(m)})_{m\ge1}$ containing
every such vector. 
For each $v^{(m)}$, we select a minimizer of $a\mapsto a^\top v^{(m)}$ on $\mathcal B$ (breaking ties lexicographically). 
Then, we delete repeated selected actions, keeping their first occurrence. This gives
a finite or countably infinite family of distinct actions
$(b_j)_{j\in J}\subseteq\mathcal B$, even when $\mathcal B$ is uncountable.
We retain their original enumeration indices, so $J\subseteq\mathbb N$
is the set of first-occurrence indices and $1\in J$.

As $\sQ^d$ is dense in $\sR^d$ and the sharpness condition guarantees that the optimal action $a^*$ with respect to $\theta$ is still optimal even if $\theta$ is slightly perturbed, a particular loss direction $v^{(m)}$, with respect to which the original action $a$ is still optimal, must be enumerated in the above process. This guarantees that $a^*\in (b_j)_{j\in J}$.
Formally, fix any loss vector $\theta$ with optimal action
$a^*$ and sharpness constant $\varsigma>0$. If a loss direction $v$
satisfies $\|v-\theta\|_{\mathcal B}<\varsigma$, then, for every
$a\in\mathcal K\setminus\{a^*\}$, we have $(a-a^*)^\top v
 \ge(\varsigma-\|v-\theta\|_{\mathcal B})
                          \|a-a^*\|_{\mathcal S}>0$.
Thus $a^*$ remains the unique minimizer throughout an open
neighborhood of $\theta$. 
As $\sQ^d$ is dense in $\sR^d$, that neighborhood contains a rational
vector, so the enumeration selects $a^*$ and retains it as
$b_{j^*}$ for some $j^*\in J$.

We assign positive prior mass to every candidate while ensuring that
the total prior mass of suboptimal candidates with gap at most
$\varepsilon$ decays sufficiently fast as $\varepsilon\to0$.
We therefore penalize a new candidate when it is close to
an earlier one.
For every $j\in J$, we define $\pi_j$ as the prior probability for action $b_j$ in $\mu$, where
\begin{equation}\label{eq:SH-priority}
      \pi_j=e^{-h_j}\,,
       \quad
        h_j=j+1+\sum_{i\in J\,,i<j}
       \exp \left(\norm{b_j-b_i}_{\mathcal S}^{-8}\right)
  \,.
\end{equation}
The main purpose of constructing such $(\pi_j)_{j\in J}$ is to make sure $\mu(\{j\ne j^*\,,0<\Delta(b_j)\le\varepsilon \})$ decreases sufficiently fast as $\varepsilon$ decreases. To see this, first note that the prefix set $\{b_i:i\in J,\ i<j^*\}$ is finite.
Define $\Delta_{\rm pre}\coloneq\min(\{1\}\cup\{\Delta(b_i):i\in J,\ i<j^*\})$.
Uniqueness of $a^*$ gives $0<\Delta_{\rm pre}\le1$, with $\Delta_{\rm pre}=1$ when the prefix is empty.
Its contribution to the prior mass of
$\{a:0<\Delta(a)\le\varepsilon\}$ is therefore at most
$\Delta_{\rm pre}^{-k}\varepsilon^{k}\sum_{i<j^*}\pi_i$ for any $k>0$;
an empty prefix contributes zero.
Let $d_j=\|b_j-a^*\|_{\mathcal S}$.
The suffix $\{b_j:j\in J,\ j>j^*\}$ may contain infinitely many candidates whose gaps approach zero. However, the distance penalty in $h_j$ gives
$\pi_j\le e^{-j-1}e^{-d_j^{-8}}\le e^{-j-1}d_j^{16}$ and the sharpness condition gives $d_j\le\Delta(b_j)/\varsigma$.
Collectively, we can show
 $\sum_{{j\ne j^*\,,0<\Delta(b_j)\le\varepsilon}}\pi_j
 \le C_{\rm at}\varepsilon^{16}$ for some instance-dependent constant
$C_{\rm at}<\infty$,
 guaranteeing that the total mass of all the actions with sufficiently small gaps in $(b_j)_{j\in J}$ is well controlled.
The stronger penalty
$\exp(d_j^{-8})$ in $h_j$ further makes a very close suboptimal candidate
difficult to select as the reference action after corruption.
Besides, one can see that $\omega_\mu\coloneq 1-\sum_{j\in J}\pi_j>\frac12$.

 \item \textbf{Continuous measure.}
The remaining mass $\omega_\mu$ supplies coverage of the whole $\gK$ for arbitrary adversarial comparators. 
Denote by 
$
\lambda_{\mathcal K}(A)
=\frac{\operatorname{vol}_k(A\cap\mathcal K)}
       {\operatorname{vol}_k(\mathcal K)}$ 
       the uniform volume measure on $\mathcal K$.
A single sample $U\sim\lambda_{\mathcal K}$ obeys the volume bound $ \sP(\Delta(U)\le\varepsilon)\le k\varepsilon/\Delta_{\max}^{\gK}$, where $\Delta_{\max}^{\gK}=\max_{a\in\mathcal K}\Delta(a)>0$.
To ensure that the probability mass of suboptimal actions near $a^*$ is sufficiently small, we instead take the law of $\bar U_m=\frac1m\sum_{i=1}^m U_i$, where $U_1,\ldots,U_m\overset{\rm i.i.d.}{\sim}\lambda_{\mathcal K}$, as the continuous component of $\mu$.
Denote by $\mu_m$ the law of $\bar U_m$.
As $a^*$ lies on the boundary of $\gK$, all $m$ i.i.d. samples must be close to the boundary for $\Delta(U_i)$ to be sufficiently small, which thus
guarantees an exponentially small probability for sufficiently small $\Delta(\bar U_m)$. 
Formally, $ \sP(\Delta(\bar U_m)\le\varepsilon)
 \le\prod_{i=1}^m\sP(\Delta(U_i)\le m\varepsilon)
 \le\left(km\varepsilon / \Delta_{\max}^{\gK}\right)^m$.
As $\mu_m$ is still supported on $\gK$, the averaging thus preserves the coverage required in the adversarial
analysis while significantly reducing probability mass near $a^*$.
We set $m=16$ to match the powers required in the analysis.

\end{itemize}
Finally, we define the reference measure $\mu$ as
\begin{equation}\label{eq:SH-prior}
 \mu=\sum_{j\in J}\pi_j\delta_{b_j}+\omega_\mu\mu_m \,.
\end{equation}

\paragraph{Geometric Gap-Calibrated Exploration}
As in \IAFTRL{}, the sampling measure $p_t$ minimizes the clipped empirical exploration cost subject to moment and stability constraints.  
Given the cumulative action-loss estimates $\widehat L_{t-1,\cdot}$ and the FTRL update $q_t$ in \eqref{eq:SH-ftrl}, we define the index of the
\textit{reference action} $r_t$ and the \textit{empirical gap} $\widehat\Delta_t(x)$ as
\begin{align}
 r_t=\argmin_{j\in J}\{\widehat L_{t-1,b_j}+\beta_t h_j\} \,,\quad
 Z_t(x)=[\widehat L_{t-1,x}-\widehat L_{t-1,b_{r_t}}]_+ 
 \,,\quad
 \widehat\Delta_t(x)=\min\{2,Z_t(x)/t\} \,,
                   \label{eq:SH-score}
\end{align}
where we break ties in \eqref{eq:SH-score} by the least index. 
Note that the reference action $b_{r_t}$ is now selected using both the cumulative loss estimate $\widehat L_{t-1,x}$ and the prior penalty. Thus $b_{r_t}$ need not minimize the unpenalized cumulative loss estimate $\widehat L_{t-1,x}$ over $x\in\mathcal K$. 
However, the reference action $b_{r_t}$ is still the atom with the largest probability under $q_t$, because
$q_t(\{b_j\})=\frac{\exp(-\widehat L_{t-1,x}/\beta_t)}{\mathcal Z_t} \mu(\{b_j\})=\frac{\exp\{-(\widehat L_{t-1,b_j}+\beta_t h_j)/\beta_t\}}
        {\mathcal Z_t}$ using $\mu(\{b_j\})=\pi_j$.
        Besides, we note that although $J$ may be infinite, at each round we can find the minimizer in \eqref{eq:SH-score} by enumerating only a finite subset of $J$
whose indices $j\in J$ do not exceed a finite threshold, since all candidates beyond this threshold have larger objective values than the first candidate (see \Cref{app:sec:algo2:reference-action-computing}).

We now introduce the proposed \textit{geometric gap-calibrated exploration} ($\text{G}^2$CE) mechanism.
For every $x\in\mathcal K$, we define the upper bounds in the constraints as
\begin{equation}\label{eq:SH-budgets}
 \begin{split}
 U_t(x)=\frac{\beta_t}{2}\norm{x-b_{r_t}}_{\mathcal S}+\frac{Z_t(x)}B \,,\quad
 W_t(x)=\frac{\beta_t}{4}\norm{x-b_{r_t}}_{\mathcal S}^2
                         +\frac{Z_t(x)^2}{Bt} \,.
 \end{split}
\end{equation}
$d_t(u;v,y)=(u-b_{r_t})^\top v\,y$ and $\overline d_t(v,y)=\int d_t(u;v,y)q_t(du)$ are defined as in \IAFTRL{}.  
Let $\mathcal M_+(\mathcal K)$ denote the set of finite nonnegative Borel
measures on $\mathcal K$. 
The program optimizes only $p$; its mixture
constraints require the existence of auxiliary
$w\in\mathcal M_+(\mathcal K)$ and $\gamma=w(\mathcal K)\le\Gamma_t$.
In \eqref{eq:SH-program}, we set $\Gamma_t=4d/\beta_t<1/4$, and $B=2^{36}(d+1)$.

\begin{subequations}\label{eq:SH-program}
\allowdisplaybreaks[0]
\predisplaypenalty=0
\begin{align}
 &\underset{p\in\gP(\mathcal K)}
       {\operatorname{minimize}}
       &&\int\widehat\Delta_t(x)p(dx)\,,
       \label{eq:SH-program-objective}\\
 &\text{subject to}
       &&\gamma=w(\mathcal K)\,,\qquad p=(1-\gamma)q_t+w\,,
       \label{eq:SH-program-mix}\\
 &&&w\ge0\,,\qquad \gamma\le\Gamma_t\,,
       \label{eq:SH-program-dom}\\
 &&&\sup_{x\in\mathcal K\,,\,Q\in\mathcal Q}
       \E_{Y_x\sim Q_x}\!\left[
       \bigl|d_t(u;v_x(p),Y_x)\bigr|\right]
       \le U_t(u)\quad(u\in\mathcal K)\,,
       \label{eq:SH-program-mag}\\
 &&&\sup_{Q\in\mathcal Q}\E_{\substack{X\sim p\\Y\sim Q_X}}\!\left[
       d_t(u;v_X(p),Y)^2\mid\cF_{t-1}\right]
       \le W_t(u)\quad(u\in\mathcal K)\,,
       \label{eq:SH-program-var}\\
 &&&\begin{aligned}
       &\beta_t\sup_{Q\in\mathcal Q}\E_{\substack{X\sim p\\Y\sim Q_X}}\left[
       \left.\int\chi\!\left(
       \frac{d_t(u;v_X(p),Y)-\overline d_t(v_X(p),Y)}{\beta_t}\right)
       q_t(du)\right|\cF_{t-1}\right]
       \le\frac{2d}{\beta_t} \,.
     \end{aligned}
       \label{eq:SH-program-cert}
\end{align}
\end{subequations}

\begin{algorithm}[!t]
\caption{Geometric Information-Acquisition FTRL (GIA-FTRL)}\label{alg:sharp}
\begin{algorithmic}[1]
\Require A generator $\mathcal B$ of action set $\gA$.
\State Compute the candidate family $(b_j)_{j\in J}$,
the weight in discrete reference measure $(\pi_j)_{j\in J}$ by \eqref{eq:SH-priority}, and the reference measure
$\mu$ by \eqref{eq:SH-prior}.
\State Set $\widehat L_{0,x}=0$ for all $x\in\mathcal K$.
\For{$t=1,2,\ldots$}
\State Solve \eqref{eq:SH-ftrl} to obtain the FTRL update $q_t$.
 \State Compute the reference action $b_{r_t}$, cumulative loss differences $Z_t$, and empirical gaps $\widehat\Delta_t$
 by \eqref{eq:SH-score}.
 \State Solve \eqref{eq:SH-program} to obtain the sampling distribution $p_t$.
\State Draw $X_t\sim p_t$ and then
$A_t\in\gA$ so that $\E[A_t\mid\cF_{t-1},X_t,v_{X_t}(p_t)]=X_t$.
 \State Play $A_t$ and observe $Y_t\in[-1,1]$.
 \State Set $\widehat\theta_t=v_{X_t}(p_t)Y_t$ and $\widehat\ell_{t,x}=x^\top\widehat\theta_t$ for $x\in\mathcal K$.
 \State Update $\widehat L_{t,x}=\widehat L_{t-1,x}+\widehat\ell_{t,x}$ for $x\in\mathcal K$.
\EndFor
\end{algorithmic}
\end{algorithm}

The mixture constraints
\eqref{eq:SH-program-mix}--\eqref{eq:SH-program-dom}, analogous to
\eqref{eq:G-program-mix}--\eqref{eq:G-program-dom} in
\IAFTRL{}, limit exploration and its
per-round cost to $2\Gamma_t$.
Constraint \eqref{eq:SH-program-cert} controls the FTRL stability
term by $2d/\beta_t$ on every history, similar to
\eqref{eq:G-program-cert}.
Recall  $\chi(z)=e^z+e^{-z}-2$.
Since $\chi$ is even and increasing in absolute value,
the worst feedback has $|Y|=1$, so constraint \eqref{eq:SH-program-cert} is equivalent to
\begin{equation}\label{eq:SH-program-cert-computable}
 \gC_t(p)\coloneq \beta_t\int\!\int
 \chi\!\left(\frac{(u-\bar x_t)^\top v_x(p)}{\beta_t}\right)
 q_t(du)p(dx)\le\frac{2d}{\beta_t} \,.
\end{equation}
Compared to program \eqref{eq:G-program}, the main changes are that the two moment constraints include geometric constraints to handle arbitrarily small gaps, \textit{i.e.}, \eqref{eq:SH-program-mag} and \eqref{eq:SH-program-var}.
In particular, compared with constraints \eqref{eq:G-program-mag} and \eqref{eq:G-program-var}, $\beta_t$ in the upper bound of the constraints \eqref{eq:SH-program-mag} and \eqref{eq:SH-program-var} is replaced by $(\beta_t/2)s_t(u)$ in $U_t(u)$, and $(\beta_t/4)s_t(u)^2$ in $W_t(u)$, while the additive
term remains the same. These changes make the constraint budget vanish as the
distance $s_t(u)=\|u-b_{r_t}\|_{\mathcal S}$ from $u$ to the reference action $b_{r_t}$
tends to zero. For instance, along $u=b_{r_t}+\lambda h$, $U_t(u)$ scales
linearly with $\lambda\ge0$ whenever $u\in\mathcal K$.
When $b_{r_t}=a^*$, sharpness gives
$s_t(u)\le\Delta(u)/\varsigma$. Dividing the loss-difference
estimate by $\Delta(u)$ therefore leaves a bounded geometric
control even as $\Delta(u)\downarrow0$.
Meanwhile, since $s_t(u)\le2$, the new budget still satisfies $U_t(u)\le\beta_t+Z_t(u)/B$ and
$W_t(u)\le\beta_t+Z_t(u)^2/(Bt)$ on every history.

\subsection{Regret Upper Bound}\label{sec:shap-upper-main}
\begin{proposition}
\label{prop:SH-compatibility}
For linear bandits with compact action sets, \GIAFTRL{} guarantees
\begin{equation}\label{eq:SH-adv}
 \Reg=\gO\left(d\sqrt T\log T\right)
\end{equation}
in the adversarial regime. Simultaneously, 
if 
the linear bandit instance satisfies
the sharpness condition in \eqref{eq:SH-overview-sharp}, then for
every feasible centered sampling allocation $n$ in
\eqref{eq:SH-centered-program} and an arbitrary generator $\mathcal B$ of $\gA$, 
 \GIAFTRL{} guarantees
\begin{equation}\label{eq:SH-per-design-rate}
 \Reg=\gO\left(c_{\mathrm{ctr}}^{n}(\gB,\theta) d\log T+M_n
       +J_\theta^{\rm cor}\sqrt C\logp(C)\right)
\end{equation}
in the corrupted stochastic regime, where
$J_\theta^{\rm cor}$ is a finite instance-dependent constant independent
of $T,C,n$, and $M_n$ is a finite constant depending on the instance and
the sampling allocation $n$, but independent of $T$ and $C$ (formally
defined in \eqref{eq:SC-constants}).
\end{proposition}

The above proposition, together with $\cctr(\gB,\theta)=\cctr(\gA,\theta)$ (see \Cref{lem:C-hull}), implies the following theorem by choosing a feasible centered
allocation $n^\circ$ on $\gB$, independently of $T,C$, with $ c_{\mathrm{ctr}}^{n^\circ}(\gB,\theta)
 \le 2c_{\mathrm{ctr}}^*(\gB,\theta)$.

\begin{theorem}[Regret in the sharp action case]
\label{thm:main-nonseparated-upper}
For linear bandits with
compact action sets, \GIAFTRL{} guarantees \mbox{\eqref{eq:SH-adv}}
in the adversarial regime. Simultaneously, if the linear bandit instance satisfies
the sharpness condition in \eqref{eq:SH-overview-sharp}, \GIAFTRL{} guarantees
\begin{equation}\label{eq:SH-cor-rate}
 \Reg=\gO\left(\cctr(\gA,\theta) d \log T+M_{\gA,\theta}
       +J_\theta^{\rm cor}\sqrt C\logp(C)\right)
 \,,
\end{equation}
in the corrupted stochastic regime, where $M_{\gA,\theta}=M_{n^\circ}$ and
$J_\theta^{\rm cor}$ are finite parameter-dependent constants independent
of $T,C$ (formally defined in \eqref{eq:SC-constants}).
\end{theorem}

Proofs of \Cref{prop:SH-compatibility} and
\Cref{thm:main-nonseparated-upper} are deferred to
\Cref{app:sharp-compatibility-proof} and
\Cref{app:sharp-results}, respectively.
\subsubsection{Proof Sketch for Proposition~\ref*{prop:SH-compatibility}}\label{ps:prop4-1}
\begingroup
\interdisplaylinepenalty=10000
In the proofs, we define $ P=4096(d+1)$, $C_{\mathrm{rat}}=64$, and $C_{\beta,\mathrm{sh}}=16d+1$.
Here $P$ is the even moment order, $C_{\mathrm{rat}}$ is the gap-ratio
moment constant, $B=2^{24}P$, and
$\beta_t=16d+\sqrt t\le C_{\beta,\mathrm{sh}}\sqrt t$ for $t\ge1$.

Fix an arbitrary finite feasible centered sampling allocation $n$, independently of $T,C$,
with $c_{\mathrm{ctr}}^{n}=c_{\mathrm{ctr}}^{n}(\gB,\theta)=\int_{\gB}\Delta\,dn$ and $N_n=n(\gB)$.
Let $X_t(x)=\widehat L_{t-1,x}-\widehat L_{t-1,a^*}$,
$\Delta(x)=(x-a^*)^\top\theta$, and
$\kappa_t=\|c_t^{\rm vec}\|_{\mathcal B}$.
As explained in \Cref{sec:SH-algorithm}, there exist a $j^*\in J$ such that $a^*=b_{j^*}$. Let $\pi^*\coloneq \mu(\{a^*\})>0$.
In particular, $h^*\coloneq \log(1/\pi^*)=\KL(\delta_{a^*},\mu)<\infty$.

Define the entropy difference
$F_t\coloneq \KL(\delta_{a^*},\mu)-\KL(q_t,\mu)
=h^*-\KL(q_t,\mu)$, and
$\mathsf D_t\coloneq \langle q_t,\widehat\ell_t\rangle+
\beta_t\log\langle q_t,e^{-\widehat\ell_t/\beta_t}\rangle$, where
$\langle q,f\rangle=\int f\,dq$.
Using unbiasedness of $\widehat{\theta}_t$,
the FTRL telescoping argument of
\eqref{eq:G-sketch-decomposition} gives
\begin{align}\label{eq:SH-sketch-decomposition}
 \Reg(a^*)&\le
 \underbrace{\beta_1h^*+\sum_{t=2}^T(\beta_t-\beta_{t-1})\E[F_t]}
       _{\mathsf{Penalty\text{-}Term}}
 +\underbrace{\sum_{t=1}^T\E[\mathsf D_t]}
       _{\mathsf{Stability\text{-}Term}}
 +\underbrace{\sum_{t=1}^T\E[\langle p_t-q_t,\Delta\rangle]}
       _{\mathsf{Exploration\text{-}Term}}
 +\underbrace{\sum_{t=1}^T\E[\langle p_t-q_t,x^\top c_t^{\rm vec}\rangle]}
       _{\mathsf{Corruption\text{-}Term}}\,.
\end{align}
In Step~3(a), we introduce a $\gF_{t-1}$-measurable event $\mathcal G_t$ on which $t\ge T_*$, the reference is correct (\textit{i.e.}, $b_{r_t}=a^*$), and
\begin{equation}\label{eq:SH-sketch-good}
 X_t(x)\ge4t^{7/8}\Delta(x)\qquad(\forall x\in\mathcal K) \,.
\end{equation}
Similarly to \eqref{eq:G-sketch-pen-final}, we define $\mathcal E_T=\sum_{t\le T}t^{-1/2}\sP(\mathcal G_t^c)$.

\paragraph{Step 1: Bounding $\mathsf{Penalty\text{-}Term}$}
Since $F_t=h^*-\KL(q_t,\mu)\le h^*$ by nonnegativity of relative
entropy, the good/bad-history split in
\eqref{eq:G-sketch-pen-estimate}, with
$\beta_t-\beta_{t-1}\le t^{-1/2}$, gives
\begin{equation}\label{eq:SH-sketch-pen-reduction}
 \mathsf{Penalty\text{-}Term}\le\beta_1h^*+
 \sum_{t=2}^T(\beta_t-\beta_{t-1})\E[\ind_{\mathcal G_t}F_t]
 +h^*\mathcal E_T\,.
\end{equation}
It remains to bound $F_t$ on $\mathcal G_t$.
In the finite case, \eqref{eq:prop3.1:psketch:X_t_floor} and
\eqref{eq:prop3.1:psketch:q_t_decay} use a positive minimum gap $\Delta_{\min}$ to show that $q_t$ concentrates on $a^*$; that is, $q_{t, a} \leq e^{-4 t^{1 / 4} / C_\beta}$ holds uniformly for all $a\in\gA$.
However, in infinite action set, suboptimal action $a$ can approach
$a^*$ with $\Delta(a)$ tending to zero, so the same argument no longer applies.

To overcome the above obstacle, we now also need to control the prior mass at all these small gaps.
On $\mathcal G_t$, let $\xi_t(x)=X_t(x)/\beta_t\ge0$ and
$Z_t^*=\int e^{-\xi_t}\,d\mu$. 
Intuitively, if $q_t$ assigns sufficiently small probability mass on all suboptimal actions with small loss gaps, then $q_t\approx \delta_{a^*}$ and thus $F_t\approx 0$, indicating that a sufficiently small penalty term can still be guaranteed. 
Indeed, to see this intuition formally, first notice that for $F_t$, we have
\begin{align}
 F_t
 &=h^*-\int\log\!\left(\frac{e^{-\xi_t(x)}}{Z_t^*}\right)q_t(dx)
   =\log(Z_t^*/\pi^*)+\int\xi_t(x)q_t(dx)
   \label{eq:prop4-1:penlty-start-0}\\
 &=\log\!\left(1+\frac1{\pi^*}
             \int_{x\ne a^*}e^{-\xi_t(x)}\mu(dx)\right)
   +\frac1{Z_t^*}\int_{x\ne a^*}\xi_t(x)e^{-\xi_t(x)}\mu(dx)
   \label{eq:SH-sketch-entropy-suboptimal}\\
 &\le\frac1{\pi^*}\int_{x\ne a^*}e^{-\xi_t(x)}\mu(dx)
   +\frac1{Z_t^*}\int_{x\ne a^*}\xi_t(x)e^{-\xi_t(x)}\mu(dx)
   \label{eq:SH-sketch-entropy-log}\\
 &\le\frac1{\pi^*}\int_{x\ne a^*}
              (1+\xi_t(x))e^{-\xi_t(x)}\mu(dx)\label{eq:SH-sketch-entropy-normalizer}\\
 &\le\frac2{\pi^*}\int_{x\ne a^*}e^{-\xi_t(x)/2}\mu(dx)
   \eqcolon 2J_t \,,
 \label{eq:SH-sketch-entropy-good}
\end{align}
where \eqref{eq:prop4-1:penlty-start-0} substitutes $dq_t/d\mu=e^{-\xi_t}/Z_t^*$;
\eqref{eq:SH-sketch-entropy-suboptimal} uses $\xi_t(a^*)=0$;
\eqref{eq:SH-sketch-entropy-log} applies $\log(1+z)\le z$;
\eqref{eq:SH-sketch-entropy-normalizer} uses $Z_t^*=\pi^*+\int_{x\ne a^*}e^{-\xi_t(x)}\mu(dx)\ge\pi^*$ and
$\xi_t\ge0$; and \eqref{eq:SH-sketch-entropy-good} follows from
$1+z\le2(1+z/2)\le2e^{z/2}$ for $z\ge0$.
Meanwhile, one can see that the concentration on $\delta_{a^*}$ can also be measured by $J_t$:
\begin{equation}\label{eq:SH-sketch-odds}
 \frac{q_t(\mathcal K\setminus\{a^*\})}{q_t(\{a^*\})}
 =\frac{(Z_t^*)^{-1}\int_{x\ne a^*}e^{-\xi_t(x)}\mu(dx)}
        {\pi^*/Z_t^*}
 =\frac1{\pi^*}\int_{x\ne a^*}e^{-\xi_t(x)}\mu(dx)
 \le J_t \,,
\end{equation}
where \eqref{eq:SH-sketch-odds} uses $\xi_t\ge0$. 
Consequently, small $J_t$ forces most of $q_t$'s mass onto the optimal atom, which in turn ensures a small $F_t$.
Importantly, the exponent $-\xi_t/2$ in $J_t$ upper bounds the mass ratio $q_t(\mathcal K\setminus\{a^*\})/q_t(\{a^*\})$ while leaving half of the FTRL exponential decay available to absorb additional factors.
As we will shortly see, the enables the same $J_t$ to also control the update exponential in the stability proxy function \eqref{eq:SH-sketch-stab-half-exponent}.

It remains to bound $J_t$, which is exactly why we need to construct the $\mu$ in \Cref{sec:SH-algorithm}.
Let $\mu_{\rm at}^\circ=\sum_{j\ne j^*}\pi_j\delta_{b_j}$ for the suboptimal atoms of the prior.
By \eqref{eq:SH-sketch-good},
$\xi_t(x)/2\ge\lambda_t\Delta(x)$, where
$\lambda_t\coloneq 2t^{7/8}/\beta_t\ge2t^{3/8}/C_{\beta,\mathrm{sh}}$. Since $\xi_t(a^*)=0$ and $\mu_m(\{a^*\})=0$,
\begin{align}
 J_t
=\frac1{\pi^*}\int_{x\ne a^*}e^{-\xi_t(x)/2}\mu(dx)
 \le\frac1{\pi^*}\int_{x\ne a^*}e^{-\lambda_t\Delta(x)}\mu(dx)
   =\frac1{\pi^*}\left\{
       \int e^{-\lambda_t\Delta}\,d\mu_{\rm at}^\circ
       +\omega_\mu\int e^{-\lambda_t\Delta}\,d\mu_m\right\}\,.
 \label{eq:SH-sketch-laplace-reduction}
\end{align}
We bound the atomic and continuous integrals in
\eqref{eq:SH-sketch-laplace-reduction} separately. In each case,
the key is a bound on the mass of $\{\Delta\le z\}$: integrating this
bound against $\lambda e^{-\lambda z}$ yields the desired exponential decay.

\begin{itemize}[leftmargin=*,itemsep=4pt,topsep=4pt]
\item \textbf{Discrete contribution.}
Only finitely many atoms precede $a^*$ in the priority order, so
$\Delta_{\rm pre}\coloneq\min(\{1\}\cup\{\Delta(b_j):j\in J,\ j<j^*\})\in(0,1]$,
with $\Delta_{\rm pre}=1$ when the prefix is empty.
Later atoms may accumulate at $a^*$; unlike the finite case, their weights
must compensate for arbitrarily small gaps. Set
$d_j=\|b_j-a^*\|_{\mathcal S}$. For $j>j^*$, the reinforced priority
and sharpness give
\begin{align}
 \pi_j=e^{-h_j}
 &\le e^{-j-1}\exp\{-\exp(d_j^{-8})\}\label{eq:SH-sketch-atom-priority}\\
 &\le e^{-j-1}d_j^{16}\label{eq:SH-sketch-atom-polynomial}\\
 &\le e^{-j-1}\varsigma^{-16}\Delta(b_j)^{16}\,.
 \label{eq:SH-sketch-atom-weight}
\end{align}
\eqref{eq:SH-sketch-atom-priority} follows from \eqref{eq:SH-priority};
\eqref{eq:SH-sketch-atom-polynomial} uses
$\exp\{-\exp(u)\}\le u^{-2}$ for $u>0$; and
\eqref{eq:SH-sketch-atom-weight} uses sharpness,
$d_j\le\Delta(b_j)/\varsigma$.
With $C_{\rm at}\coloneq \Delta_{\rm pre}^{-16}+\varsigma^{-16}$, this yields
\begin{align}
 \mu_{\rm at}^\circ\{\Delta\le z\}
 &
 =\sum_{j<j^*}\pi_j\ind_{\{\Delta(b_j)\le z\}}
   +\sum_{j>j^*}\pi_j\ind_{\{\Delta(b_j)\le z\}}
   \notag\\
 &
 \le(z/\Delta_{\rm pre})^{16}\sum_{j<j^*}\pi_j
   +\varsigma^{-16}z^{16}\sum_{j>j^*}e^{-j-1}
 \le C_{\rm at}z^{16} \,,
 \label{eq:SH-sketch-atom-cap}
\end{align}
where
\eqref{eq:SH-sketch-atom-cap} uses
$\ind_{\{\Delta(b_j)\le z\}}\le(z/\Delta_{\rm pre})^{16}$ for $j<j^*$; and
\eqref{eq:SH-sketch-atom-weight} for $j>j^*$, and the bounds
$\sum_{j<j^*}\pi_j\le1$ and $\sum_{j>j^*}e^{-j-1}\le1$.
We can now bound the integral regarding $\mu_{\rm at}$ itself:
\begin{align}
 \int e^{-\lambda\Delta}\,d\mu_{\rm at}^\circ
 =\int\!\left(\int_{\Delta(x)}^\infty
                    \lambda e^{-\lambda z}\,dz\right)\mu_{\rm at}^\circ(dx)
 &=\lambda\int_0^\infty e^{-\lambda z}
                 \mu_{\rm at}^\circ\{\Delta\le z\}\,dz
                 \notag\\
 &\le C_{\rm at}\lambda\int_0^\infty e^{-\lambda z}z^{16}\,dz
   \label{eq:SH-sketch-atom-laplace-cap}\\
 &= C_{\rm at}\lambda^{-16}\int_0^\infty e^{-v}v^{16}\,dv
   =16!C_{\rm at}\lambda^{-16} \,,
 \label{eq:SH-sketch-atom-laplace}
\end{align}
where \eqref{eq:SH-sketch-atom-laplace-cap} uses
\eqref{eq:SH-sketch-atom-cap}; and \eqref{eq:SH-sketch-atom-laplace}
substitutes $v=\lambda z$ and evaluates the Gamma integral.

\item \textbf{Continuous contribution.}
The key point for controlling the probability mass of small gaps is that
averaging $m=16$ samples in $\mu_m$ gives a sixteenth-power decay bound.
Specifically,
      \eqref{eq:SHr-convolution-laplace} in \Cref{lem:SH-caps}, 
      with $f=\Delta$, $h=1/\lambda$, and $m=16$,
guarantees that
\begin{align}
 \int e^{-\lambda\Delta}\,d\mu_m
 \le\left(\frac{16k}{\lambda \Delta_{\max}^{\gK}}\right)^{16} \,.
 \label{eq:SH-sketch-laplace-product}
\end{align}
\end{itemize}
Substituting the discrete contribution \eqref{eq:SH-sketch-atom-laplace}
and the continuous contribution \eqref{eq:SH-sketch-laplace-product}
into \eqref{eq:SH-sketch-laplace-reduction} gives
\begin{align}
 J_t
 \le\frac1{\pi^*}\left\{16!C_{\rm at}
       +\omega_\mu(16k/\Delta_{\max}^{\gK})^{16}\right\}\lambda_t^{-16}
 \le\frac1{\pi^*}\left\{16!C_{\rm at}+(16k/\Delta_{\max}^{\gK})^{16}\right\}
             \left(\frac{C_{\beta,\mathrm{sh}}}{2t^{3/8}}\right)^{16}
   =A_Lt^{-6} \,,
 \label{eq:SH-sketch-laplace}
\end{align}
where $A_L\coloneq (\pi^*)^{-1}(C_{\beta,\mathrm{sh}}/2)^{16}
\{16!C_{\rm at}+(16k/\Delta_{\max}^{\gK})^{16}\}$ is independent of $T,C$,
and the second inequality uses $\omega_\mu\le1$ and
$\lambda_t\ge2t^{3/8}/C_{\beta,\mathrm{sh}}$. Thus even though individual suboptimal gaps can vanish,
the prior puts sufficiently little total mass at those gaps to make
$J_t$ summable.

The final summation has the same logic as \eqref{eq:G-sketch-pen-final}.
Substituting \eqref{eq:SH-sketch-entropy-good} and
\eqref{eq:SH-sketch-laplace} into
\eqref{eq:SH-sketch-pen-reduction} gives
\begin{align}
 \mathsf{Penalty\text{-}Term}
 \le\beta_1h^*+2A_L\sum_{t=2}^Tt^{-13/2}+h^*\mathcal E_T
 \le\beta_1h^*+4A_L+h^*\mathcal E_T\,.
 \label{eq:SH-sketch-pen-final}
\end{align}

\paragraph{Step 2: Bounding $\mathsf{Stability\text{-}Term}$}
As in \eqref{eq:G-sketch-stab-split}, the conditional stability cost is
bounded by the proxy function $\gC_t(p_t)$. On bad histories the global
constraint gives $\gC_t(p_t)\le2d/\beta_t$.
On $\mathcal G_t$, the centering argument, the subsequent removal of the
zero $a^*$ term, and the absorption of the update exponential are the
same as in \eqref{eq:G-sketch-stab-good}.
The new difficulty is how to control the resulting integral, as the uniform upper bound $q_{t, a} \leq e^{-4 t^{1 / 4} / C_\beta}$ in finite action case no longer holds now. However, one can see that $\mathsf{Stability\text{-}Term}$ is also controlled by $J_t$, which has already been controlled in \eqref{eq:SH-sketch-pen-final} when dealing with $\mathsf{Penalty\text{-}Term}$.

Specifically, on $\mathcal G_t$, let $d_t(x,y)=(x-a^*)^\top v_y(p_t)$.
The magnitude constraint gives
$|d_t(x,y)|/\beta_t\le\|x-a^*\|_{\mathcal S}/2+\xi_t(x)/B
\le1+\xi_t(x)/B$ and $d_t(a^*,y)=0$. Then, we have
\begin{align}
 \gC_t(p_t)
 &\le e^2\beta_t\int_{x\ne a^*}e^{2\xi_t(x)/B}q_t(dx)\label{eq:SH-sketch-stab-centered}\\
 &=\frac{e^2\beta_t}{Z_t^*}\int_{x\ne a^*}
                          e^{-(1-2/B)\xi_t(x)}\mu(dx)\label{eq:SH-sketch-stab-density}\\
 &\le\frac{e^2\beta_t}{\pi^*}\int_{x\ne a^*}
                          e^{-\xi_t(x)/2}\mu(dx)
   =e^2\beta_tJ_t\label{eq:SH-sketch-stab-half-exponent}\\
 &\le e^2C_{\beta,\mathrm{sh}}A_Lt^{-11/2} \,,
 \label{eq:SH-sketch-stab-good}
\end{align}
where \eqref{eq:SH-sketch-stab-centered} uses the same argument in \eqref{eq:G-sketch-stab-good};
\eqref{eq:SH-sketch-stab-density} substitutes $dq_t/d\mu=e^{-\xi_t}/Z_t^*$;
\eqref{eq:SH-sketch-stab-half-exponent} uses $Z_t^*\ge\pi^*$, $B\ge4$,
and $\xi_t\ge0$; and \eqref{eq:SH-sketch-stab-good} follows from
$\beta_t\le C_{\beta,\mathrm{sh}}\sqrt t$ and the mass bound of $J_t$ in
\eqref{eq:SH-sketch-laplace}.

Combining the good-history bound \eqref{eq:SH-sketch-stab-good}
with the global proxy function \eqref{eq:SH-program-cert-computable},
and summing as in \eqref{eq:G-sketch-stab-final}, yields
\begin{align}
 \mathsf{Stability\text{-}Term}
 \le e^2C_{\beta,\mathrm{sh}}A_L\sum_{t=1}^Tt^{-11/2}
                  +2d\mathcal E_T
\le2e^2C_{\beta,\mathrm{sh}}A_L+2d\mathcal E_T\,.
 \label{eq:SH-sketch-stab-final}
\end{align}

\paragraph{Step 3: Bounding $\mathsf{Exploration\text{-}Term}$}
The split in \eqref{eq:G-sketch-expl-def} applies with
$\langle p_t-q_t,\Delta\rangle\le2\Gamma_t\le8d/\sqrt t$, yielding
\begin{equation}\label{eq:SH-sketch-expl-reduction}
 \mathsf{Exploration\text{-}Term}\le\sum_{t=1}^T
       \E[\ind_{\mathcal G_t}\langle p_t,\Delta\rangle]+8d\mathcal E_T\,.
\end{equation}
We use the same three steps in the Step 3(a) of 
the proof sketch of \Cref{prop:G-compatibility}.
The additional challenges here are controlling gap-estimation ratios uniformly over all suboptimal actions, identifying a penalized reference action, and 
maintaining second-moment bounds proportional to \(\Delta(x)^2\), even for arbitrarily small gaps.
Throughout Step~3, an asterisk (*) marks a procedure that differs from its counterpart in Step~3 of the proof sketch of \Cref{prop:G-compatibility} or has no counterpart there.

\paragraph{Step 3(a): Gap Ratios and the Cost of Failure Histories}
Use $P=4096(d+1)$, $C_{\mathrm{rat}}=64$, and $B=2^{24}P$.
As in Step~3(a) of the finite case, we first consider $C=0$ and then the case of $C>0$.

\begingroup
\setlist[enumerate,1]{labelindent=2em}
\subparagraph{(i) Corruption Level $C=0$.}
Procedure 4, 5, 7, and 8 are similar to the finite case. We detail Procedure 1, 2, 3, and 6 below.
\begin{enumerate}[leftmargin=*,itemsep=4pt,topsep=4pt]
\item 
\textbf{Raw moment bounds.*}
In the finite proof, positivity of every $X_{s,a}$ forces $r_s=a^*$.
Here the priority rule only implies
$r_s=j\Rightarrow X_s(b_j)\le\beta_s(h^*-h_j)$: an atom with
$h_j<h^*$ may still be selected despite a positive estimated gap.
Thus we first control the estimated cumulative loss gaps
$X_s(x)$ for all $x\in\mathcal K$, without assuming that the
selected reference is $a^*$. We then use these bounds to show that
$b_{r_s}=a^*$ with high probability. On this event, we can apply
the algorithm's magnitude and variance constraints with $a^*$
as the reference.

Let $D_s(x)=X_{s+1}(x)-X_s(x)$ and
$M_s=\max_{\mathcal K}(-X_s)$.
Since $X_s(a^*)=0$, we have $M_s\ge0$;
the reference-centered score satisfies
$Z_s(x)=[X_s(x)-X_s(b_{r_s})]_+\le X_s(x)+M_s$ and
$Z_s(a^*)\le M_s$.
The same subtraction of reference-centered increments as in
\eqref{eq:G-sketch-local-increment-budgets} consequently gives
$|D_s(x)|\le2\beta_s+[X_s(x)+2M_s]/B$ and
$\E_{s-1}[D_s(x)]=\Delta(x)$, together with 
the following bound from \eqref{eq:SH-global-var} in \Cref{lem:SH-feasible}:
\begin{equation}\label{eq:SH-sketch-raw-budgets}
 \E_{s-1}[D_s(x)^2]
 \le8\beta_s+\frac{2[X_s(x)+2M_s]^2}{Bs}\,.
\end{equation}

The next step requires bounds on $\E[(R_s^X)^P]$ and $\E[M_s^P]$, where
$R_s^X=\max_{x\in\mathcal K}|X_s(x)|$.
The new obstacle is that the Taylor remainder contains the
maximum $X_s$ and $M_s$, whereas our moment recursion will be for an integral.
Specifically, for a finite set of random variables $\{Y_i\}_{i=1}^n$, we have
$ \E \left[\max_{1\le i\le n}Y_i\right]
 \le\E\!\left[\sum_{i=1}^nY_i\right]
 =\sum_{i=1}^n\E[Y_i] \,.$
Thus expectations of sums give bounds on expectations of maxima.
This is the step used in the proof of \Cref{prop:G-compatibility} to
obtain \eqref{eq:G-3a-ratio-inverse-substitution} and
\eqref{eq:G-3a-ratio-positive-substitution} from single-action moment
bounds.
With infinitely many actions, however, the corresponding sum may
diverge even when the maximum is finite.
We therefore need a different way to control the maximum.

An \emph{affine contraction argument} lets us replace the finite sum
by a spatial integral. Compactness ensures that a maximum is attained;
convexity keeps a contraction toward a maximizing point inside the
domain; and affinity ensures that function values remain large on the
contracted region. For a simple example, let $h$ be a nonnegative affine
function on $[0,1]$ with maximum $m=h(1)$, such as $h(x)=x$.
On the half-interval $[1/2,1]$, affinity gives
$h(x)=(1-x)h(0)+xh(1)\ge m/2$. Hence the integral controls the maximum:
\[
 \int_0^1h(x)\,dx
 \ge\int_{1/2}^1h(x)\,dx
 \ge\frac12\cdot\frac m2=\frac m4\,,
 \qquad
 \max_{x\in[0,1]}h(x)\le4\int_0^1h(x)\,dx\,.
\]

For the Taylor remainder, \eqref{eq:taylor-positivepart} in
\Cref{lem:taylor} gives, for real $y,z$,
 \begin{align}
 (y+z)_+^P
 =y_+^P+Py_+^{P-1}z+\mathcal R_P(y,z)\,,\quad 
 0\le\mathcal R_P(y,z)
 \le\frac{P(P-1)}2(y+|z|)_+^{P-2}z^2\,.
 \end{align}
For the negative-moment calculation, take $y=-X_s(x)$ and
$z=-D_s(x)$. The conditional expectation of $z^2$ is controlled by
\eqref{eq:SH-sketch-raw-budgets}, whose bound depends on $M_s$.
After bounding the increment and integrating over $x$, we therefore
need to control terms involving a squared maximum and an integrated
$(P-2)$-nd power, including shifts caused by the increment.
We now apply the affine contraction argument on a
compact convex set $D$ of affine dimension $j\le P$, with normalized
relative volume $\sigma_D$. For affine $g$ and $M=\max_Dg_+$, the
argument together with H\"older's inequality gives 
\begin{align}\label{eq:SH-sketch-curvature-bound}
      M^2\int_Dg_+^{P-2}\,d\sigma_D
 \le16\int_Dg_+^P\,d\sigma_D\,.
\end{align}
The full derivation is in \Cref{lem:C-affine}, which also gives the
shifted version \eqref{eq:C-aff-shift} needed when the Taylor remainder
contains $(g+c)_+^{P-2}$ for $c\ge0$. These estimates control the
integrated Taylor remainder using an integral moment of the same order $P$.

Using \eqref{eq:SH-sketch-curvature-bound} and \eqref{eq:C-aff-shift}
to control the integrated Taylor remainder, we use the same Taylor expansion and
induction strategy as in the arguments for
\eqref{eq:G-3a-positive-explicit} and \eqref{eq:G-3a-window-tail}
in the proof of \Cref{prop:G-compatibility}. Setting
$G_s=\int_{\mathcal K}|X_s|^P\,d\sigma_{\mathcal K}$ and
$F_s^-=\int_{\mathcal K}(-X_s)_+^P\,d\sigma_{\mathcal K}$, we obtain
\begin{equation}\label{eq:SH-sketch-integral-moment-bounds}
 \E[G_s]\le C_G(C_{\beta,\mathrm{sh}})s^P\,,
 \qquad
 \E[F_s^-]\le C_F(C_{\beta,\mathrm{sh}})s^{3P/4}\,.
\end{equation}
The smaller exponent for the negative part uses its nonpositive
first-order Taylor term. The recurrences and their constants are given
in \Cref{lem:SH-field-interface}, specifically
\eqref{eq:SH-G-rec}--\eqref{eq:SH-F-linear}.

To obtain moment bounds for the maxima from
\eqref{eq:SH-sketch-integral-moment-bounds}, apply the affine contraction
argument once more. With
$j=\dim\operatorname{aff}(\mathcal K)$, \Cref{lem:C-affine} gives
directly
\begin{equation}\label{eq:SH-sketch-maxima-from-integrals}
 (R_s^X)^P\le2^P4^jG_s\,,
 \qquad
 M_s^{P+j}\le2^P4^j(R_s^X)^jF_s^-\,.
\end{equation}
The second inequality also involves the absolute maximum because the
negative values may occupy only a small part of the domain.
Taking expectations in the first inequality of
\eqref{eq:SH-sketch-maxima-from-integrals} and using the bound on
$\E[G_s]$ in \eqref{eq:SH-sketch-integral-moment-bounds} gives
\begin{equation}\label{eq:SH-sketch-absolute-envelope}
 \E[(R_s^X)^P]\le R^Ps^P\,.
\end{equation}
For the negative maximum, raise the second inequality of
\eqref{eq:SH-sketch-maxima-from-integrals} to the power $P/(P+j)$,
take expectations, and apply H\"older's inequality. Substituting
\eqref{eq:SH-sketch-absolute-envelope} and the bound on $\E[F_s^-]$
in \eqref{eq:SH-sketch-integral-moment-bounds} then yields
\begin{equation}\label{eq:SH-sketch-clean-global-negative}
 \E[M_s^P]\le C_Ms^{\kappa P}\,,
 \qquad
 \kappa=\frac34+\frac{d}{4(P+d)}<\frac78\,.
\end{equation}
Here $R$ and $C_M$ are defined in \eqref{eq:SH-global-constants}.
See \Cref{lem:D-global-moments} for the full details.

\item \textbf{Identifying the penalized reference.*}
Take $m_r=\lceil\sqrt t\rceil$ and let
$I_s^r=\ind\{M_j\le j^{7/8}\text{ for }m_r\le j\le s\}$.
The buffer $Y_s^r(x)=X_s(x)+2s^{7/8}$ is then positive on the event $I_s^r=1$.
The surviving-window argument in \eqref{eq:G-3a-window-tail}, applied
to \eqref{eq:SH-sketch-clean-global-negative}, gives
\begin{equation}\label{eq:SH-sketch-clean-reference-preparation-tail}
 \sP(I_t^r=0)\le A_Ht^{-\eta/2}\,,
\end{equation}
where $\eta=P(7/8-\kappa)$ and $A_H=2^{\kappa P+1}C_M$.
Likewise, the inverse-moment argument in
\eqref{eq:G-3a-inverse-chain}--\eqref{eq:G-3a-inverse-explicit},
using \eqref{eq:SH-sketch-raw-budgets}, yields
\begin{equation}\label{eq:SH-sketch-clean-reference-inverse}
 \E[I_t^rY_t^r(x)^{-P}]\le\left(\frac4{t\Delta(x)}\right)^P\,.
\end{equation}
\eqref{eq:SH-sketch-clean-reference-preparation-tail} and
\eqref{eq:SH-sketch-clean-reference-inverse} hold beyond the required threshold (see \eqref{eq:SH-anchor-constants}).
In a finite set, a positive minimum gap makes it possible to sum such
pointwise bounds. Here neither that minimum nor a finite sum is
available. 
To cope with this issue, we split at $\varepsilon_t=t^{-1/16}$, set
$\tau=t^{7/8}$ and $p=P/2$, and use a different mechanism in each range.

\begin{itemize}[leftmargin=*,itemsep=4pt,topsep=4pt]
\item \emph{Larger gaps.}
Let $D_t=\mathcal K\cap\{\Delta\ge\varepsilon_t\}$ and
$E_t^r=\{I_t^r=1,b_{r_t}\ne a^*,b_{r_t}\in D_t\}$.
To control $\sP(E_t^r)$, we use the fact that selecting a reference
index $r_t$ whose action lies in the larger-gap region forces a large
spatial inverse moment. Suppose $D_t$ is nonempty.
On $E_t^r$,
the condition $I_t^r=1$ ensures $Y_t^r\ge\tau>0$ throughout
$\mathcal K$. Since the selected action $b_{r_t}$ belongs to $D_t$,
reference optimality and $\beta_th^*\le\tau$ give the upper bound
\[
 \min_{x\in D_t}Y_t^r(x)
 \le Y_t^r(b_{r_t})
 \le\beta_th^*+2\tau
 \le3\tau \,.
\]
Affinity then ensures that $Y_t^r$ remains small on a sufficiently
contracted region around a minimizer $x_-$. Specifically, the affine
contraction argument, using $\max_{D_t}Y_t^r\le R_t^X+2\tau$, gives
$Y_t^r\le4\tau$ on $D'_t=(1-a_t)x_-+a_tD_t$, where
$a_t=\tau/(R_t^X+3\tau)$ and
$\sigma_{D_t}(D'_t)=a_t^j$ for $j=\dim\operatorname{aff}(D_t)$.

Consequently, $(Y_t^r)^{-p}\ge(4\tau)^{-p}$ on $D'_t$.
Geometrically, the region under the graph of $(Y_t^r)^{-p}$ contains
the cylinder $D'_t\times[0,(4\tau)^{-p}]$, whose base has relative
volume $a_t^j$ and whose height is $(4\tau)^{-p}$. Thus, on $E_t^r$,
\[
 \int_{D_t}Y_t^r(x)^{-p}\,d\sigma_{D_t}
 \ge(4\tau)^{-p}a_t^j
 =(4\tau)^{-p}\left(\frac{\tau}{R_t^X+3\tau}\right)^j\,.
\]
In particular, the ratio of this integral to the cylinder's volume
is at least one whenever $E_t^r$ occurs. Define the inverse integral
to be zero off $\{I_t^r=1\}$; the ratio is then a nonnegative random
variable that bounds the indicator of $E_t^r$ from above.
Taking expectations and applying Cauchy--Schwarz, using
\eqref{eq:SH-sketch-absolute-envelope} and
\eqref{eq:SH-sketch-clean-reference-inverse}, yields
\[
 \sP(E_t^r)\le16^p(R+3)^d t^{-P/32+d/8}\,.
\]
The detailed derivations are deferred to \eqref{eq:SHr-anchor-contraction}--\eqref{eq:SHr-anchor-large-gap-probability}
in \Cref{lem:SH-anchor}.

\item \emph{Smaller gaps.}
The bound on $\sP(0<\Delta(b_{r_t})<\varepsilon_t)$ comes from the
large reference penalties attached to very small gaps.
Only finitely many reference atoms precede $a^*$, and their
positive gaps are bounded away from zero. Thus, once
$\varepsilon_t\le\Delta_{\rm pre}$, every atom with
$0<\Delta(b_j)<\varepsilon_t$ must satisfy $j>j^*$.
Sharpness places such an atom close to $a^*$, and the priority rule
\eqref{eq:SH-priority} assigns it a large penalty for being close to
this earlier atom. At the threshold $\varepsilon_t=t^{-1/16}$, this gives
$h_j>\varsigma^8\sqrt t$. Since $h^*$ is fixed and
$\beta_t\ge\sqrt t$, the penalty difference
$\beta_t(h_j-h^*)$ is at least a positive constant times $t$ for
sufficiently large $t$.

Therefore, selecting such an atom  requires its estimated cumulative 
      loss to satisfy $\widehat L_{t-1,b_j}+\beta_t h_j\le\widehat L_{t-1,a^*}+\beta_t h^*$, and hence its cumulative
loss to fall below that of $a^*$ by enough to offset this penalty
difference;
      that is, $X_t(b_j)\le-\beta_t(h_j-h^*)\le-\varsigma^8t/2$ once $t\ge(2h^*/\varsigma^8)^2$.
In particular, selection forces
$M_t\ge\varsigma^8t/2$. This reduces all small-gap reference errors
to one event in which the most negative estimated gap has magnitude
of order $t$. Applying Markov's inequality with the moment bound
\eqref{eq:SH-sketch-clean-global-negative} gives, for
$t\ge T_A$ as defined in \eqref{eq:SH-anchor-constants},
\[
 \sP(0<\Delta(b_{r_t})<\varepsilon_t)
 \le\sP(M_t\ge\varsigma^8t/2)
 \le2^PC_M\varsigma^{-8P}t^{-(1-\kappa)P}\,.
\]
See \eqref{eq:SHr-anchor-small-gap-probability} in \Cref{lem:SH-anchor} for details.

\end{itemize}
Set $a_A=\min\{\eta/2,P/32-d/8,(1-\kappa)P\}>16$ and
$A_A=A_H+16^p(R+3)^d+2^PC_M\varsigma^{-8P}$.
Combining the preparation failure
\eqref{eq:SH-sketch-clean-reference-preparation-tail}, the large-gap
error \eqref{eq:SHr-anchor-large-gap-probability}, and the small-gap
error \eqref{eq:SHr-anchor-small-gap-probability} gives
\begin{align}
 \sP(b_{r_t}\ne a^*)
 \le\sP(I_t^r=0)+\sP(E_t^r)
       +\sP(0<\Delta(b_{r_t})<\varepsilon_t)
 \le A_At^{-a_A}\,.
 \label{eq:SH-sketch-clean-reference-tail}
\end{align}

\item \textbf{Normalizing vanishing gaps.*}

Uniform bounds on the estimated-gap ratios $X_s(x)/\Delta(x)$
must remain valid even as the true gaps of suboptimal actions
approach zero. 
Dividing 
the constraint $|D_s(x)|\le2\beta_s+[X_s(x)+2M_s]/B$ and the second-moment constraint \eqref{eq:SH-sketch-raw-budgets}
by $\Delta(x)$ introduces terms such as
$\beta_s/\Delta(x)$ and $M_s/\Delta(x)$, which need not be bounded.
Normalizing the vector difference $x-a^*$ by the loss gap $\Delta(x)$
avoids this problem: sharpness gives
$\|(x-a^*)/\Delta(x)\|_{\mathcal S}\le\varsigma^{-1}$ for $x\ne a^*$.
Once the reference is $a^*$, 
       the magnitude and second-moment constraints \eqref{eq:SH-program-mag}--\eqref{eq:SH-program-var}, now centered at $a^*$, with $d_s(x;v,y)=(x-a^*)^\top vy$, $U_s(x)=\frac{\beta_s}{2}\|x-a^*\|_{\mathcal S}+[X_s(x)]_+/B$, and $W_s(x)=\frac{\beta_s}{4}\|x-a^*\|_{\mathcal S}^2+[X_s(x)]_+^2/(Bs)$
       scale linearly or quadratically
under scalar multiplication of the vector difference $x-a^*$,
and normalization preserves finite increment bounds.

Specifically, define
${\mathsf Z}_s(u)=u^\top\sum_{j<s}\widehat\theta_j$ and the normalized vector difference
$u_x\coloneq(x-a^*)/\Delta(x)$ for $x\ne a^*$. Sharpness and linearity give
\[
 \|u_x\|_{\mathcal S}\le\varsigma^{-1}\,,
 \qquad \frac{X_s(x)}{\Delta(x)}={\mathsf Z}_s(u_x)\,.
\]
The closure $\mathcal U_\theta$ of these directions is compact and
convex, and $\theta^\top u=1$ on it. Thus the normalized estimates
form one linear function on a bounded convex domain, where the affine
contraction argument can subsequently control their extrema.
On $b_{r_s}=a^*$, the uniform increment bounds are precisely
\eqref{eq:SH-normalized-increments} in \Cref{lem:D-normalized-interface};
the conditional drift is one on $\mathcal U_\theta$.
The observations preceding a window starting at $m$ also satisfy
\[
 \left\|\max_{u\in\mathcal D_\theta}|{\mathsf Z}_m(u)|\right\|_P
 \le\Lambda Rm\,,
 \qquad \mathcal D_\theta=\Conv(\{0\}\cup\mathcal U_\theta)\,.
\]
This follows from \eqref{eq:SH-initial-extension} and
\eqref{eq:SH-sketch-absolute-envelope}, so the full history can be
retained. The geometric and increment details are in
\Cref{lem:D-projective,lem:D-normalized-interface}.

\item \textbf{Constructing the buffered window.}
We can now use the buffered-window construction of
\eqref{eq:G-3a-survival-definition}--\eqref{eq:G-3a-moment-targets}
on the normalized field. Set $m=\lceil\sqrt t\rceil$,
$\tau_s=s^{7/8}+4t^{5/8}$, and
$N_s^{\mathcal U}=\max_{\mathcal D_\theta}(-{\mathsf Z}_s)$.
Let $I_s=\ind\{b_{r_j}=a^*,\ N_j^{\mathcal U}\le\tau_j
\text{ for }m\le j\le s\}$ and
$Y_s^\dagger(u)={\mathsf Z}_s(u)+2\tau_s$.
The good-history event is
\begin{equation}\label{eq:SH-sketch-clean-window}
 \mathcal G_t=
 \left\{I_t=1,\ \min_{u\in\mathcal U_\theta}{\mathsf Z}_t(u)
                        \ge4\tau_t\right\}
 \quad(t\ge T_*)\,,
\end{equation}
and $\mathcal G_t=\varnothing$ before $T_*$.
\item \textbf{Bounding inverse and positive moments.}
With \eqref{eq:SH-normalized-increments} and
\eqref{eq:SH-initial-extension} together with \eqref{eq:SH-sketch-absolute-envelope}, the finite inverse recursion
\eqref{eq:G-sketch-inverse-recursion} applies with drift one and
$\beta_s'=\Lambda^2\beta_s$. It yields
\begin{equation}\label{eq:SH-sketch-clean-local-moments}
 \E[I_tY_t^\dagger(u)^{-P}]\le(4/t)^P
 \quad(u\in\mathcal U_\theta)\,.
\end{equation}
The positive-moment argument in \eqref{eq:G-3a-positive-explicit}
similarly gives
\begin{equation}\label{eq:SH-sketch-clean-local-positive}
 \E[I_t(Y_t^\dagger(u)/t)^P]\le8^P
 \quad(u\in\mathcal U_\theta)\,.
\end{equation}
The threshold and initialization are verified in
\Cref{lem:SH-local}; the scalar Taylor expansions are unchanged.

\item \textbf{Spatial control of extrema.*}
Again, the finite maximum-to-sum argument
\eqref{eq:G-3a-ratio-inverse-substitution}--\eqref{eq:G-3a-ratio-roots} does not apply in continuous case. On $\mathcal G_t$, let
$f(u)={\mathsf Z}_t(u)/t>0$ and denote its maximum and minimum on
$\mathcal U_\theta$ by $R_f$ and $m_f$.
The comparison ${\mathsf Z}_t\le Y_t^\dagger\le2{\mathsf Z}_t$ and
the positive and inverse moment bounds
\eqref{eq:SH-sketch-clean-local-positive} and
\eqref{eq:SH-sketch-clean-local-moments}, respectively, give $ J_\pm\coloneq \int_{\mathcal U_\theta}f^{\pm P}\,d\sigma_{\mathcal U_\theta}$ and $\E[\ind_{\mathcal G_t}J_\pm]\le8^P$,
where $\sigma_{\mathcal U_\theta}$ is normalized relative volume.
To control the gap ratios, we need to convert these integral bounds
into moment bounds for $R_f$ and $m_f^{-1}$. The points where $f$
attains its maximum or minimum depend on the observed history, so
moment bounds for a fixed point cannot simply be applied at these
random points.

To deal with the above obstacle, we use the affine contraction argument to relate these extrema to integrals
over regions of positive relative volume. If the maximum $R_f$ is large,
contracting $\mathcal U_\theta$ toward a maximizer
gives a region of fixed relative
volume on which $f$ remains comparable to $R_f$. The positive integral
$J_+$ must therefore be large. If the minimum $m_f$ is small,
contracting toward a minimizer gives a region on which $f\le2m_f$,
and hence $f^{-P}\ge(2m_f)^{-P}$. This forces a lower bound on the
inverse integral $J_-$. Letting
$j=\dim\operatorname{aff}(\mathcal U_\theta)$, the contraction factor
$m_f/R_f$ in the latter case gives relative volume $(m_f/R_f)^j$.
This dependence on the size of the region explains both the appearance
of $R_f$ and the loss of $j$ powers in the inverse estimate.

More precisely, the affine contraction argument in \Cref{lem:C-affine} gives directly, on
$\mathcal G_t$,
\[
 R_f^P\le2^P4^jJ_+\,,
 \qquad
 m_f^{-(P-j)}\le2^PR_f^jJ_-\,.
\]
Combining these bounds with
$\E[\ind_{\mathcal G_t}J_\pm]\le8^P$ and applying generalized
H\"older yields the required uniform moment bounds (see \eqref{eq:SHr-ratio-integrals}--\eqref{eq:SHr-inverse-ratio-holder}
in the proof of \Cref{lem:SH-ratios}):
\[
 \|\ind_{\mathcal G_t}R_f\|_P\le64\,,
 \qquad
 \|\ind_{\mathcal G_t}m_f^{-1}\|_{P/4}<18\,.
\]

\item \textbf{Bounding gap ratios.}
Define $\lo_t=\max\{1,m_f^{-1}\}$ and $\up_t=R_f$ on
$\mathcal G_t$, and both to be zero otherwise.
\eqref{eq:SHr-ratio-integrals} and
\eqref{eq:SHr-inverse-ratio-holder} imply
$\|\lo_t\|_{P/4},\|\up_t\|_{P/4}\le C_{\mathrm{rat}}$.
The same product estimate as \eqref{eq:G-3a-ratio-products} then gives
\begin{equation}\label{eq:SH-sketch-clean-ratio-products}
 \E[\lo_t^r\up_t]\le C_{\mathrm{rat}}^{r+1}
 \quad(r=1,3,5)\,.
\end{equation}
The clipping argument in \eqref{eq:G-3a-clipping} likewise yields
$\Delta(x)\le\lo_t\widehat\Delta_t(x)\le\lo_t\up_t\Delta(x)$.
Finally, \eqref{eq:SH-sketch-clean-window} gives
$X_t(x)=\Delta(x){\mathsf Z}_t(u_x)
\ge4\tau_t\Delta(x)\ge4t^{7/8}\Delta(x)$,
as required in \eqref{eq:SH-sketch-good}.

\item \textbf{Bounding failure probabilities and their weighted cost.}
The failure probability $\sP(\mathcal G_t^c)$ contributes to regret
through the weighted cost
$\mathcal E_T=\sum_{t\le T}t^{-1/2}\sP(\mathcal G_t^c)$.
Bounding it requires control over the whole observation window and
over infinitely many normalized directions at the endpoint.
We separate three causes of failure: a wrong reference, a
negative threshold crossing during the window, and an insufficient
minimum at the endpoint. Freezing the entire function at the first
crossing preserves the offending value, allowing the same argument as in \eqref{eq:G-3a-window-tail} to control the whole window (see \Cref{lem:SH-local}).

For the endpoint, use the affine contraction argument toward a
minimizer: a small minimum of the positive buffered function forces
it to remain small on a contracted region, making its inverse-power
integral large. The inverse moment bound
\eqref{eq:SH-local-moments} then controls this event.
Combining these estimates and using the same weighted summation as
after \eqref{eq:G-3a-clean-markov} gives the results (see \Cref{lem:SH-ratios})
\[
 \sP(\mathcal G_t^c)\le C_{\rm bad}t^{-2}
 \quad(t\ge T_*)\,,
 \qquad
 \mathcal E_T\le2\sqrt{T_*}+2C_{\rm bad}\,.
\]
The initial interval contributes at most $2\sqrt{T_*}$; the later failure probabilities have a summable weighted tail. 

\end{enumerate}

\subparagraph{(ii) Corruption Level $C>0$.}
\begin{itemize}[leftmargin=*,itemsep=4pt,topsep=4pt]
\item \textbf{Compensating the drift.}
Apply the same signed-drift compensation as in the $C>0$ case of
\Cref{prop:G-compatibility},
\eqref{eq:G-3a-corrupt-buffer}--\eqref{eq:G-3a-corrupt-local-bounds},
and the same remaining-budget shift as in \Cref{lem:AB-card-shift}.
With $A_s^c=\sum_{j<s}\kappa_j$,
$W_s(x)=X_s(x)-2(C-A_s^c)$, and
$Q_s=\max_{\mathcal K}(-W_s)$, the $C=0$ affine contraction and
moment arguments for \eqref{eq:SH-sketch-absolute-envelope} and
\eqref{eq:SH-sketch-clean-global-negative} give (see \Cref{lem:D-cor-interface,lem:SC-global})
\[
 \E[(R_s^X)^P]\le R^Ps^P\,,
 \qquad
 \E[Q_s^P]\le A_gs^{\kappa P}+B_gn_C^{P-q}s^q
 \quad(s\ge n_C)\,,
\]
where $n_C=\lceil C\rceil+2$ and $q=P(d+v)/(P+d)$.
The additional term accounts for the initial corrupted history. 

\item \textbf{Reference recovery over a block.*}
The two gap regimes from the $C=0$ reference argument in
\Cref{lem:SH-anchor} still apply. The new difficulty is to absorb
the inherited-history term while controlling every round of
$[u,2u]$. We use a slowly decreasing gap threshold to limit the
inverse-gap cost, and stop at the first error to avoid a factor of
$u$ from summing roundwise bounds. The resulting choices are
\[
 \epsilon_u=\zeta[4\log(eu)]^{-1/8}\,,
 \qquad
 N=\left\lceil K_0n_C\log^2(en_C)\right\rceil\,,
 \qquad \zeta=\min\{1,\Delta_{\rm pre},\varsigma\}\,.
\]
The factor $\log^2(en_C)$ absorbs the logarithmic inverse-gap cost
in the inherited-history term. The two cases below, together with
the preparation-failure bound, yield
\[
 \sP(\exists s\in[u,2u]:b_{r_s}\ne a^*)
 \le A_r\{u^{-a}+(N/u)^a\}\,,
 \qquad u\ge N\,,\quad a=P/128\,.
\]
See \eqref{eq:SC-reference-block} in \Cref{lem:SC-reference} for details.
In the above two displays, $K_0$ and $A_r$ are specified in \eqref{eq:SC-clock} and \eqref{eq:SC-reference-constant}.

\begin{itemize}[leftmargin=*,itemsep=4pt,topsep=4pt]
\item \emph{Small gaps.}
Use the same reduction from a large priority penalty to a negative
estimated gap as in the $C=0$ small-gap argument
\eqref{eq:SHr-anchor-small-gap-probability}.
At the logarithmic threshold $\epsilon_u$, however, 
the polynomial priority bound used there is not sufficient.
Retaining the full exponential
in \eqref{eq:SH-priority} makes the penalty large enough: selecting
$b_j$ with $0<\Delta(b_j)<\epsilon_u$ at any $s\in[u,2u]$ forces
\[
 h_j>(eu)^4\,,
 \qquad X_s(b_j)\le-\tfrac12u^{9/2}\,.
\]
For the event $\mathcal R_u^{\rm small}$ of any such selection,
the same 
argument as in \eqref{eq:G-3a-window-tail},
using the frozen absolute moment from \Cref{lem:SC-global}, gives (see \eqref{eq:SCr-reference-small-priority}--\eqref{eq:SCr-reference-small-probability} in \Cref{lem:SC-reference})
\[
 \sP(\mathcal R_u^{\rm small})\le(4R)^Pu^{-7P/2}\,.
\]

\item \emph{Larger gaps.}
Let $\mathcal R_u^{\rm large}$ be the event that a wrong reference
is selected from $D_u=\mathcal K\cap\{\Delta\ge\epsilon_u\}$
during $[u,2u]$ before preparation fails.
Preparation supplies the inverse-moment bound
\eqref{eq:SC-prepared-inverse} for the positive function at the
deterministic starting time $u$.
The new issue compared to the case of $C=0$ is that its first selection time is
random, so the fixed-time inverse bound cannot be substituted there.
We freeze the whole positive function, including its compensator
and buffer, at that selection, and assign the inverse process value
zero if preparation fails first. This lets the same inverse
recursion \eqref{eq:G-3a-inverse-conditional} bound the frozen
function $\widetilde Y_{2u}$ at a deterministic endpoint:
\[
 \E[\ind_{\mathcal R_u^{\rm large}}\widetilde Y_{2u}(x)^{-P}]
 \le\left(\frac8{u\Delta(x)}\right)^P\,,
 \qquad x\in D_u\,.
\]
The $C=0$ affine contraction and Cauchy--Schwarz argument
\eqref{eq:SHr-anchor-contraction}--\eqref{eq:SHr-anchor-large-gap-probability}
then applies unchanged to the frozen function. With $p=P/2$, it gives
\[
 \sP(\mathcal R_u^{\rm large})
 \le2^{12P}(2R+10)^d\zeta^{-p}
       \{u^{-a}+(N/u)^a\}\,.
\]
See \Cref{lem:SC-reference} for the stopping details in
\eqref{eq:SCr-reference-stopped-inverse} and the reduction of
logarithmic factors in \eqref{eq:SCr-reference-power-reductions}.

\end{itemize}

\item \textbf{Recovering moments and gap ratios.}
Use the compensated window of \Cref{def:D-cor-window}, with
$m=\lceil\sqrt{Nt}\rceil$ for $t\ge t_0=\lceil F_\theta N\rceil$.
As in the $C>0$ moment argument of \Cref{prop:G-compatibility},
compensation restores the scalar recursions and $m\ge\Lambda C$
absorbs the initial corruption shift. The resulting local moments
are \eqref{eq:SC-local-moments} in \Cref{lem:SC-local}.
The $C=0$ affine contraction and H\"older calculation
\eqref{eq:SHr-ratio-integrals}--\eqref{eq:SHr-inverse-ratio-holder}
therefore gives the same ratio bounds:
\[
 \|\lo_t\|_{P/4},\ \|\up_t\|_{P/4}\le C_{\mathrm{rat}}\,,
 \qquad
 \E[\lo_t^r\up_t]\le C_{\mathrm{rat}}^{r+1}
 \quad(r=1,3,5)\,.
\]
The same clipping step \eqref{eq:G-3a-clipping} and the good-event
definition give, on $\mathcal G_t$,
\[
 \Delta(x)\le\lo_t\widehat\Delta_t(x)
       \le\lo_t\up_t\Delta(x)\,,
 \qquad X_t(x)\ge4t^{7/8}\Delta(x)\,.
\]
See \Cref{lem:SC-good}, specifically
\eqref{eq:SCr-good-ratios-and-gaps}, for the details.

\item \textbf{Bounding failure probabilities and their weighted cost.}
Apply the sam reference, crossing, and endpoint split of the $C=0$ case in \eqref{eq:SHr-relative-good-probability}, 
using the block bound
\eqref{eq:SC-reference-block} and the compensated moments of
\Cref{lem:SC-local}. The weighted summation is the same as in the
$C>0$ case of \Cref{prop:G-compatibility}. With
$\mathcal E_T=\sum_{t\le T}t^{-1/2}\sP(\mathcal G_t^c)$, the results are
\[
 \sP(\mathcal G_t^c)\le D_\theta\{t^{-2}+(N/t)^2\}
 \quad(t\ge t_0)\,,
 \qquad \mathcal E_T\le W_\theta\sqrt N\,.
\]
Here the recovery scale is
$N=\lceil K_0n_C\log^2(en_C)\rceil$, so its square root contributes
the additional logarithm:
\[
 (h^*+10d)\mathcal E_T
 \le E_\theta[1+\sqrt C\log(e+C)]\,.
\]
Details are in \Cref{lem:SC-good}, especially
\eqref{eq:SC-good-tail} and \eqref{eq:SCr-good-weighted-sum}; the
constants are specified in \eqref{eq:SC-final-threshold}.

\end{itemize}

\endgroup

\paragraph{Step 3(b): Verifying the Mixture Distribution Induced by the Sampling Allocation}
As in finite Step~3(b), we compare the optimizer with a mixture of $q_t$
and the fixed sampling allocation $n$, scaled by $\rho_t=16B\lo_t^2/t$. Define $H_n=\int aa^\top n(da)$, and set
$a_n=\max\{1,4BC_{\beta,\mathrm{sh}}N_n/d\}$ and $a_\rho=16B$.
The two restrictions in
$\mathcal W_t=\mathcal G_t\cap\{\lo_t^2\le\sqrt t/a_n\}
\cap\{\lo_t^2\le t/a_\rho\}$ ensure, respectively,
$\rho_tN_n\le\Gamma_t<1/4$ and $\rho_t\le1$.
On this event, let $\gamma_t^c=\rho_t n(\mathcal B\setminus\{a^*\})$
and consider $ p_t^c=(1-\gamma_t^c)q_t+\rho_t n|_{\mathcal B\setminus\{a^*\}}$,
where $n|_{\mathcal B\setminus\{a^*\}}$ is the restriction of $n$ to the suboptimal actions: for every Borel set $A$, $n|_{\mathcal B\setminus\{a^*\}}(A)=n(A\cap(\mathcal B\setminus\{a^*\}))$. 
We verify the constraints in the following.

\begin{itemize}[leftmargin=*,itemsep=4pt,topsep=4pt]
\item \textbf{Mixture and exploration-mass constraints: \eqref{eq:SH-program-mix}--\eqref{eq:SH-program-dom}.}
With $w_t^c=\rho_t n|_{\mathcal B\setminus\{a^*\}}$, the candidate
has the required mixture form and $w_t^c(\mathcal K)=\gamma_t^c$.
Since $n$ is nonnegative and $\rho_t\ge0$, we have $w_t^c\ge0$;
on $\mathcal W_t$, its total mass satisfies
$0\le\gamma_t^c=\rho_t n(\mathcal B\setminus\{a^*\})
\le\rho_tN_n\le\Gamma_t$.
Also, $\gamma_t^c\le\Gamma_t$ gives the required marginal domination
$p_t^c\ge(1-\Gamma_t)q_t$.

\item \textbf{The second-moment constraint \eqref{eq:SH-program-var}: why centering is necessary.}
We first establish the matrix lower bound $M_c\succeq\rho_tH_n$,
which will let us apply the information constraint to the second moment.
The concentration estimates \eqref{eq:SH-sketch-odds} and
\eqref{eq:SH-sketch-laplace} play the role of the finite exponential mass bound \eqref{eq:prop3.1:psketch:q_t_decay}.
Indeed, \eqref{eq:SH-sketch-odds} gives
\begin{equation}\label{eq:SH-sketch-gibbs-mass}
 q_t(\mathcal K\setminus\{a^*\})
 \le\frac{q_t(\mathcal K\setminus\{a^*\})}{q_t(\{a^*\})}
 \le J_t\le A_Lt^{-6}
 \qquad(\mathcal G_t)\,.
\end{equation}
Thus the same argument as \eqref{eq:G-sketch-optimal-mass} gives
$p_t^c(\{a^*\})\ge(1-\Gamma_t)(1-A_Lt^{-6})\ge9/16$,
where $T_*$ ensures $A_Lt^{-6}\le1/4$.
This exceeds $\rho_tn(\{a^*\})\le\rho_tN_n<1/4$.
The mass already concentrated at $a^*$ therefore supplies the allocation's
optimal atom at zero regret cost. Since the other atoms of the allocation are
included explicitly, $p_t^c\ge\rho_t n$; hence
\begin{equation}\label{eq:SH-sketch-design-covariance}
 M_c\coloneq M(p_t^c)=\int xx^\top p_t^c(dx)
 \succeq\rho_t\int aa^\top n(da)=\rho_tH_n\succ0\,.
\end{equation}

On $\mathcal W_t$, the reference is $a^*$ and the feedback has
absolute value at most one, so \eqref{eq:SH-program-var} follows
if $V_t^c(x)\le W_t(x)$ for every $x\in\mathcal K$, with $V_t^c$
defined below. 
In the finite proof, the positive minimum gap allows
$4\rho_t\le\Delta_{\min}^2\le\Delta(a)^2$, so the constant $4$ in
\eqref{eq:G-sketch-design-centered-moment} is absorbed into
$\Delta(a)^2/\rho_t$.
Here, convexity ensures that $x_\lambda=a^*+\lambda(x-a^*)$
remains in $\mathcal K$ for arbitrarily small $\lambda>0$.
For a fixed history on $\mathcal W_t$, \eqref{eq:SH-budgets} gives
$W_t(x_\lambda)=\lambda^2W_t(x)\to0$ as $\lambda\downarrow0$.
Therefore, the argument in the finite case does not apply here.
We therefore need an upper bound
that also tends to zero as the distance $\|x-a^*\|_{\mathcal S}$ from $x$ to $a^*$ tends to zero.
For the vector difference $x-a^*$,
the covariance identity
and \eqref{eq:SH-sketch-design-covariance} give
\[
 V_t^c(x)\coloneq \int[(x-a^*)^\top M_c^{-1}y]^2p_t^c(dy)
 =\|x-a^*\|_{M_c^{-1}}^2
 \le\rho_t^{-1}\|x-a^*\|_{H_n^{-1}}^2\,.
\]
For a fixed history, this expression also scales by $\lambda^2$
when the vector difference $x-a^*$ is multiplied by $\lambda$. To bound its
size relative to the budget, we use feasibility of $n$ in
\eqref{eq:SH-centered-program}, which gives
$\|a-a^*\|_{H_n^{-1}}^2\le\Delta(a)^2/2$ for $a\in\mathcal B$.
Taking square roots, convexity of the norm and affinity of the gap
extend this bound to $\mathcal K=\Conv(\mathcal B)$ (see \eqref{eq:D-centered-convex-gap} in \Cref{lem:D-centered-value}): $ \|x-a^*\|_{H_n^{-1}}^2\le\frac{\Delta(x)^2}{2}
 \qquad(x\in\mathcal K)$.
Substituting $\rho_t=16B\lo_t^2/t$ and using
$t\Delta(x)\le\lo_tX_t(x)$ on $\mathcal W_t$ now yields $ V_t^c(x)
 \le\frac{\Delta(x)^2}{2\rho_t}
 \le\frac{X_t(x)^2}{32Bt}\le W_t(x)$
where the last inequality uses \eqref{eq:SH-budgets}.
See \eqref{eq:SHr-witness-variance} and
\eqref{eq:D-witness-variance-budget} in \Cref{lem:SH-witness} for details.
.

\item \textbf{The magnitude constraint \eqref{eq:SH-program-mag}.}
\Cref{lem:SH-witness} demonstrates that, on $\mathcal W_t$, $ \sup_{y\in\mathcal K}|(x-a^*)^\top M_c^{-1}y|
 \le\frac{X_t(x)}{8B}\le U_t(x)
 \qquad(x\in\mathcal K)$.
The estimated difference is zero for $x=a^*$.

\item \textbf{The stability proxy function \eqref{eq:SH-program-cert}.}
For the same reconstruction, \Cref{lem:SH-witness} gives
$\gC_t(p_t^c)\le2d/\beta_t$ on $\mathcal W_t$.

\item \textbf{Objective comparison and worst-case feasibility: \eqref{eq:SH-program}.}
With feasibility verified above, the objective comparison follows exactly
\eqref{eq:G-sketch-objective-cost}.
On $\mathcal W_t$ it gives
\begin{equation}\label{eq:SH-sketch-objective-cost}
 \langle p_t,\Delta\rangle
 \le\lo_t\up_t\langle q_t,\Delta\rangle
          +\frac{16Bc_{\mathrm{ctr}}^{n}}{t}\lo_t^3\up_t\,.
\end{equation}
On $\mathcal G_t\setminus\mathcal W_t$, we set
$p_t^0=(1-\Gamma_t)q_t+\Gamma_t\pi$, where $\pi$ is G-optimal.
\Cref{lem:SH-feasible} guarantees that this law satisfies all the required constraints.
Further, using $p_t^0$ in the objective comparison
\eqref{eq:G-sketch-objective-cost} on
$\mathcal G_t\setminus\mathcal W_t$ gives $ \langle p_t,\Delta\rangle
 \le\lo_t\up_t\langle q_t,\Delta\rangle+2\Gamma_t\lo_t\up_t$.

\end{itemize}

\paragraph{Step 3(c): Summing the Exploration Cost}
The fifth-moment tail argument in \eqref{eq:G-sketch-mass-tail}, now
using $\E[\lo_t^r\up_t]\le C_{\mathrm{rat}}^{r+1}$ for $r=1,3,5$, gives
\begin{equation}\label{eq:SH-sketch-mixed-ratio-tail}
 \E[\ind_{\mathcal G_t}\lo_t\up_t\ind\{\lo_t^2>x\}]
 \le\min\{C_{\mathrm{rat}}^2,C_{\mathrm{rat}}^6x^{-2}\}\,.
\end{equation}
Applying \eqref{eq:SH-sketch-mixed-ratio-tail} at $x=\sqrt t/a_n$
and $x=t/a_\rho$, the two thresholds defining $\mathcal W_t$,
bounds the expected fallback ratio by
\begin{equation}\label{eq:SH-sketch-mass-tail}
 m_t=\min\{C_{\mathrm{rat}}^2,C_{\mathrm{rat}}^6a_n^2/t\}
       +\min\{C_{\mathrm{rat}}^2,C_{\mathrm{rat}}^6a_\rho^2/t^2\}\,.
\end{equation}
Also, \eqref{eq:SH-sketch-gibbs-mass} implies
$\langle q_t,\Delta\rangle\le2A_Lt^{-6}$ on $\mathcal G_t$.
Substituting \eqref{eq:SH-sketch-objective-cost} and
$\langle p_t,\Delta\rangle\le\lo_t\up_t\langle q_t,\Delta\rangle+2\Gamma_t\lo_t\up_t$ into
\eqref{eq:SH-sketch-expl-reduction}, using the ratio moments
$\E[\lo_t^r\up_t]\le C_{\mathrm{rat}}^{r+1}$ for $r=1,3,5$, the FTRL mass bound
\eqref{eq:SH-sketch-gibbs-mass}, and the fallback tail
\eqref{eq:SH-sketch-mass-tail}, and summing as in
\eqref{eq:G-sketch-expl-final}, gives
\begin{align}
 \mathsf{Exploration\text{-}Term}
 &\le\sum_{t=1}^T\left\{
       \frac{16BC_{\mathrm{rat}}^4c_{\mathrm{ctr}}^{n}}{t}
       +2A_LC_{\mathrm{rat}}^2t^{-6}
       +8dt^{-1/2}m_t\right\}+8d\mathcal E_T\notag\\
 &\le16BC_{\mathrm{rat}}^4c_{\mathrm{ctr}}^{n}\log(eT)
                    +M_{3,\theta,n}+8d\mathcal E_T\,,
 \label{eq:SH-sketch-expl-final}
\end{align}
where 
$ M_{3,\theta,n}=4A_LC_{\mathrm{rat}}^2
       +8d\left(4C_{\mathrm{rat}}^4a_n
                        +\tfrac83C_{\mathrm{rat}}^3\sqrt{a_\rho}\right)$.

\paragraph{Step 4: Bounding $\mathsf{Corruption\text{-}Term}$}
Similar to \eqref{eq:G-sketch-cor-final}, with
$\widetilde p_t=[p_t-(1-\Gamma_t)q_t]/\Gamma_t$,
$\Gamma_t\le4d/\sqrt t$, $0\le\kappa_t\le2$, and
$\sum_t\kappa_t\le C$, we have
\begin{equation}\label{eq:SH-sketch-cor-final}
 \mathsf{Corruption\text{-}Term}\le16\sqrt2d\sqrt C\,.
\end{equation}
Thus the direct corruption term introduces no new logarithm.

\paragraph{Putting the Four Terms Together}
Substituting \eqref{eq:SH-sketch-pen-final},
\eqref{eq:SH-sketch-stab-final}, \eqref{eq:SH-sketch-expl-final}, and
\eqref{eq:SH-sketch-cor-final} into \eqref{eq:SH-sketch-decomposition}
combines the four terms as in \eqref{eq:G-sketch-total-final}.
Using $(h^*+10d)\mathcal E_T\le E_\theta[1+\sqrt C\log(e+C)]$ for the resulting failure cost
$(h^*+10d)\mathcal E_T$, we obtain
\begin{align}
 \Reg(a^*)
 \le16BC_{\mathrm{rat}}^4c_{\mathrm{ctr}}^{n}\log(eT)+M_n
                      +J_\theta^{\rm anc}\sqrt C\log(e+C)\,,
 \label{eq:SH-sketch-total-final}
\end{align}
where $J_\theta^{\rm anc}=E_\theta+16\sqrt2d$ and $ M_n=\beta_1h^*+4A_L+2e^2C_{\beta,\mathrm{sh}}A_L
                         +M_{3,\theta,n}+E_\theta$.
It remains to obtain the adversarial guarantee and change the comparator.
These require two geometric replacements for the finite-action arguments.

\paragraph{Regret in the Corrupted Stochastic Regime}

The regret bound so far uses the uncorrupted optimal action $a^*$ as its
comparator. We now compare with the best fixed action for the actual
corrupted losses. 
In a finite action set,
$\Delta_{\min}=\min_{a\ne a^*}\Delta(a)>0$, thus
every other action has
uncorrupted cumulative loss at least $T\Delta_{\min}$ larger than that of
$a^*$.
Corruption can change this loss difference by at most $2C$.
Thus, when $T\Delta_{\min}\ge2C$, no action has smaller cumulative expected
corrupted loss than $a^*$, so $a^*$ remains the best fixed action even after corruptions.

On an infinite action set, $\inf_{x\ne a^*}\Delta(x)$ may be zero,
so the above argument no longer applies.
To this end, we retain the distance from $x$ to $a^*$ in the corruption bound.
Let $\bar c_T=T^{-1}\sum_{t\le T}\E[c_t^{\rm vec}]$.
The budget gives $\|\bar c_T\|_{\mathcal B}\le C/T$.
Then, we have
\begin{align}
 \Reg-\Reg(a^*)
 =T\sup_{x\in\mathcal K}
       \{-\Delta(x)-(x-a^*)^\top\bar c_T\}
 \le T\sup_{x\in\mathcal K}
       \{-\Delta(x)+(C/T)\|x-a^*\|_{\mathcal S}\}\,.
 \label{eq:SH-sketch-comparator}
\end{align}
We consider two cases.
\begin{itemize}[leftmargin=*,itemsep=4pt,topsep=4pt]
\item \textbf{Case $T\varsigma\ge C$.}
Sharpness bounds the uncorrupted gap from below by the same distance
factor that controls corruption. For every $x\in\mathcal K$,
\begin{align}
 -\Delta(x)-(x-a^*)^\top\bar c_T
 \le-\varsigma\|x-a^*\|_{\mathcal S}
       +\|\bar c_T\|_{\mathcal B}\|x-a^*\|_{\mathcal S}
 \le(C/T-\varsigma)\|x-a^*\|_{\mathcal S}\le0\,.
 \label{eq:SH-sketch-comparator-zero}
\end{align}
Hence the supremum in \eqref{eq:SH-sketch-comparator} is zero,
and $\Reg=\Reg(a^*)$: the uncorrupted optimal action $a^*$ remains a best fixed
action without requiring a positive minimum gap. The bound
\eqref{eq:SH-sketch-total-final} therefore applies directly.

\item \textbf{Case $T\varsigma<C$.}
Since $T<C/\varsigma$, the adversarial bound
\eqref{eq:SH-sketch-adv-final} gives
\begin{align}
 \Reg&\lesssim (d+1)\sqrt T\log(eT)\notag\\
 &\lesssim
 \frac{(d+1)}{\sqrt\varsigma}\sqrt C
       [1+(\log C)_++(\log\varsigma^{-1})_+]\notag\\
 &\le K_\theta^{\rm cmp}\sqrt C\log(e+C)\,,
 \label{eq:SH-sketch-comparator-short}
\end{align}
where $K_\theta^{\rm cmp}=320(d+1)\varsigma^{-1/2}
[2+\log(1+\varsigma^{-1})]$.
\end{itemize}
Combining the two cases with \eqref{eq:SH-sketch-total-final}, and setting
$J_\theta^{\rm cor}=J_\theta^{\rm anc}+K_\theta^{\rm cmp}$, gives
\begin{equation}\label{eq:SH-sketch-cor-capped}
 \Reg\le\min\{U_{\rm SH}(T),\,
 16BC_{\mathrm{rat}}^4c_{\mathrm{ctr}}^{n}\log(eT)+M_n
                    +J_\theta^{\rm cor}\sqrt C\log(e+C)\}\,.
\end{equation}
By linearity, the comparator supremum and the corruption norm are unchanged
when $\mathcal B$ is replaced by any action set $\cA$ satisfying
$\mathcal B\subseteq\cA\subseteq\Conv(\mathcal B)$.

\paragraph{Regret in the Adversarial Regime}
For a finite action set with the uniform prior \(\mu_K\), \(\KL(\delta_u,\mu_K)=\log K\) for every action \(u\). On an infinite action set, however, \(\KL(\delta_u,\mu)=+\infty\) whenever \(\mu(\{u\})=0\).
We thus approximate $\delta_u$ by competing with a smoothed comparator. Specifically,
let $\nu_{u,T}$ be the law of $(1-1/T)u+U/T$ for
$U\sim\lambda_{\mathcal K}$, and let $q_{u,T}$ be the law of the average
of $m=16$ independent samples from $\nu_{u,T}$.
We then have
$\KL(\nu_{u,T},\lambda_{\mathcal K})=k\log T$.
The entropy of ($q_{u,T},\mu$) is bounded by
\begin{align}
 \KL(q_{u,T},\mu)
 &\le\log(1/\omega_\mu)+\KL(q_{u,T},\mu_m)
 \le\log2+\KL(\nu_{u,T}^{\otimes16},\lambda_{\mathcal K}^{\otimes16})
 \label{eq:SH-sketch-comparison-kl-processing}\\
 &=\log2+16\KL(\nu_{u,T},\lambda_{\mathcal K})
 =\log2+16k\log T \,,
 \label{eq:SH-sketch-comparison-kl}
\end{align}
where the first inequality uses
$\mu\ge\omega_\mu\mu_m$;  the second inequality is by $\omega_\mu>1/2$
and the data-processing inequality for KL divergence under the averaging map; 
and the last equality follows from $\KL(\nu_{u,T},\lambda_{\mathcal K})=k\log T$.
Meanwhile, we have
\begin{align}
 |\langle q_{u,T}-\delta_u,x^\top\theta_t\rangle|
 =\frac1T|(\E[U]-u)^\top\theta_t|
 =\frac1T\left|\E[(U-u)^\top\theta_t]\right|
 \le\frac1T\E[|(U-u)^\top\theta_t|]\le\frac2T\,.
 \label{eq:SH-sketch-comparison-error}
\end{align}
Substituting the entropy bound \eqref{eq:SH-sketch-comparison-kl}
and the approximation bound \eqref{eq:SH-sketch-comparison-error}
into the worst-case guarantee of \GIAFTRL{}
gives
\begin{align}
 \Reg
 &\le\min\{2T,\beta_T[\log2+16k\log T]+20d\sqrt T+2\}
 \le 320(d+1)\sqrt T\log(eT) \,.
 \label{eq:SH-sketch-adv-final}
\end{align}
\endgroup

\subsection{Regret Lower Bound}\label{sec:sharp-lower-main}
The next two classes of sharp linear bandit instances provide lower bounds on the logarithmic main term
and the multiplicative coefficient in the $\sqrt C$ corruption term.
A policy is \emph{consistent} on a class of stochastic instances if
$\Reg[T,\theta]=o(T^\alpha)$ for every fixed instance $\theta$ in that class and
every $\alpha>0$ \cite{LS17}.

\begin{theorem}
      \label{thm:main-sharp-lower}

\leavevmode\par
\begin{itemize}[labelindent=0pt,leftmargin=*,align=left,itemsep=4pt,topsep=4pt]
\item \textup{ Information-acquisition lower bound.}
On $ \mathcal A_{\rm curve}
 =\{(-1/2+s/4,(s+s^2)/4):0\le s\le1\}$, $ \theta_\tau=(\tau,0)$ and $ 0<\tau\le1$,
we have $\cctr=8/\tau$. Every consistent policy satisfies 
\begin{equation*}
 \liminf_{T\to\infty}\frac{\Reg[T,\theta_\tau]}{\log T}
 \ge\frac7{16\tau}=\frac7{128}\cctr\,. \notag
\end{equation*}

\item \textup{Corruption on a sharp instance with arbitrarily small gaps.}
Let $ \mathcal A_{\rm arch}=\{(x,(1-x^2)/2):-1\le x\le1\}$, $\theta_\sigma=(\sigma\phi/2,1/2)$, and $
 \sigma=\pm1$,
where $0<\phi\le1/8$. For each $\sigma=\pm1$, the sharpness modulus
$\varsigma_\sigma$ in \eqref{eq:SH-overview-sharp} and the centered
information-acquisition complexity satisfy 
$\frac{\phi}{6}\le\varsigma_\sigma\le\frac{\phi}{2}$ and $\frac{128}{25}\le\cctr(\mathcal A_{\rm arch},\theta_\sigma)\le10$.
For every corruption budget $C\ge\phi^{-1}$ and every integer
$T\ge\lceil\phi^{-2}\rceil$, there exist two corrupted stochastic
environments indexed by $\sigma=\pm1$, each with fixed loss vector
$\theta_\sigma$ and a deterministic corruption sequence of total budget
at most $C$, such that every policy satisfies
\[
 \max_{\sigma=\pm1}\Reg[T,\sigma]
 \ge\frac1{32}\min\{\sqrt T,\sqrt{C/\phi}\}\,.
\]
\end{itemize}
\end{theorem}
Part~\textup{(i)} uses a proof idea similar to that of Theorem~1 of~\cite{LS17}, adapted to the
infinite action set
$\mathcal A_{\rm curve}$.
In part~\textup{(ii)}, these bounds imply that the lower bound scales as
$\min\{\sqrt T,\sqrt{C/\varsigma_\sigma}\}$ up to universal constants,
even though the centered information-acquisition complexity remains bounded.
Part~\textup{(ii)} uses the same
budget-truncation proof idea as \Cref{thm:main-finite-lower}, applied
to $\mathcal A_{\rm arch}$. We defer the full proofs to \Cref{app:sharp-lower-proof}.

\section{Beyond Sharpness: Impossibility of an $\gO(\sqrt{C})$ Corruption Guarantee}\label{sec:lower-main}

In this section, we show that the $\gO(\sqrt{C})$ corruption guarantee in \Cref{thm:main-nonseparated-upper} cannot be extended to all nonsharp instances.

\begin{theorem}
       \label{thm:main-critical}
For every even integer $p\ge4$, let $ \cA_p=\{(1/2+s,s^p):|s|\le1/4\}$ and $\theta_0=(0,1)$.
This fixed compact set has 
a unique optimal action $a_0=(1/2,0)$, but is not sharp and has $\cctr=\infty$.
For every $C\ge0$ and every integer
$T\ge\max\{1,[4^{p-1}/(8p)]^2\}$, there exist two corrupted stochastic environments indexed by
$\sigma\in\{-1,+1\}$ that share the same fixed loss vector $\theta_0$.
Only the deterministic corruption sequence depends on $\sigma$,
and in both environments the total corruption budget is at most $C$.
For every policy,
these environments satisfy
\begin{equation}\label{eq:critical-full}
 \max_{\sigma=\pm1}\Reg[T,\sigma]
 \ge\frac{(p-1)T}{4p^{p/(p-1)}}
 \left(\min\left\{\frac{4C}{3T},\frac1{8\sqrt T}\right\}\right)^{p/(p-1)}\,.
\end{equation}
\end{theorem}
For example, if $p=4$ and $T\ge4$, then $ \max_\sigma \Reg[T,\sigma]
 \ge\min\left\{\frac{C^{4/3}}{4\,3^{1/3}T^{1/3}},
                \frac{3T^{1/3}}{4\,32^{4/3}}\right\}$.
In particular, this is $\Omega(C^{2/3})$ when $C=3\sqrt T/32$.
More generally, setting $C=3\sqrt T/32$ yields a regret lower bound of
$\Omega_p(C^{\beta_p})$, where $\beta_p=(p-2)/(p-1)$ approaches one as
$p$ grows. Hence for any
fixed $\gamma<1$ and finite $q\ge0$, no policy can satisfy $\Reg\le A_{\cA,\theta}\log T+M_{\cA,\theta}
       +J_{\cA,\theta}C^\gamma[\log T]^q$
on every instance in this family for all $T,C>0$, where
$A_{\cA,\theta}$, $M_{\cA,\theta}$, and $J_{\cA,\theta}$ are finite
constants that may depend on the instance $(\cA,\theta)$ but not on
$T$ or $C$.
We remark that our \Cref{thm:main-critical} shows that the $\gO(\sqrt{C})$ regret guarantee cannot hold on the instance considered there, and it does not rule out the $\gO(\sqrt{C})$ guarantee on other nonsharp instances.
Besides, for any corruption budget $C$, the RHS of
\eqref{eq:critical-full} is at most a constant depending only on $p$
times $T^{(p-2)/(2(p-1))}$. For every fixed $p$, this is $o(\sqrt T)$
because $(p-2)/(2(p-1))<1/2$, so the lower bound does not contradict
the $\gO(\sqrt T)$ regret upper bound in the adversarial regime.
The complete proof is deferred to \Cref{ch:critical}.

\section{Further Discussions}\label{sec:scope}
We believe our work provides a better understanding of FTRL in linear bandit problems 
and the established algorithmic and analysis framework might be of independent interest in
other bandit problems with structured feedback. On the other hand, our work also raises more interesting questions than it answers.
We conclude by discussing some of them.
\begin{itemize}[leftmargin=*,itemsep=4pt,topsep=4pt]
\item \textbf{Optimal BOBW result with information-acquisition complexity.}
The leading terms of the stochastic regret guarantees for \IAFTRL{} and \GIAFTRL{} are $c^*(\mathcal{A}, \theta) \log K \cdot \log T$ and $ c_{\text {ctr }}^*(\mathcal{A}, \theta)d \log T$, respectively, where suboptimal multiplicative factors $\log K$ and $d$ appear. \cite{LS17} have developed an algorithm with a leading term of $\Theta(c^*(\mathcal{A}, \theta) \log T)$, yet only for the purely stochastic regime.
It remains an interesting and very challenging direction to investigate whether
a BOBW result with a $\Theta(c^*(\mathcal{A}, \theta) \log T)$ leading term in the stochastic regime can be achieved, and, in particular, whether it can be achieved using FTRL.

\item \textbf{Uniqueness of the optimal action $a^*$.} For both finite and infinite action sets, the assumption of uniqueness of the optimal action $a^*$ is essential in the analyses. Indeed, even to obtain an $\gO(\log T/\Delta_{\min})$ regret, this assumption already is required by
 all existing studies of BOBW linear bandits 
\cite{KongZ023,ItoT23,DannWZ23,ItoTsuchiyaHonda2024,ZhaoItoLi2025}.
This assumption is also required by the detect-and-switch-based algorithm of \cite{LeeEtAl2021}. \cite{Ito21} has lifted the uniqueness assumption in the special case of BOBW MABs. Extending the analysis of \cite{Ito21} to the linear bandit case seems promising, but remains challenging even if we only aim to obtain a BOBW guarantee with an $\gO(\log T/\Delta_{\min})$ regret, let alone the $\gO(c^*\log T)$ regret.

\item \textbf{Dependence of the corruption term on gaps and sharpness.}
Our lower bounds in \Cref{thm:main-finite-lower,thm:main-sharp-lower} show that
the multiplicative coefficient of $\sqrt C$ in the corruption term need not
remain bounded when information-acquisition complexity remains bounded.
This coefficient can grow as the minimum gap $\Delta_{\min}$ decreases in the finite-action
case, or as the sharpness modulus $\varsigma$ decreases in the sharp
case. 
The finite upper bound's extra factor
$\Delta_{\min}^{-1/256}$ and the sharp upper bound's $\log(e+C)$ factor remain
unmatched by the lower bounds. Determining the optimal dependence of the
corruption term on the minimum gap $\Delta_{\min}$ and sharpness modulus $\varsigma$, and whether
the extra $\log(e+C)$ factor can be removed, also remains an interesting future direction.

\item 
\textbf{An open problem: uncorrupted BOBW learning beyond sharpness.}
Beyond the sharpness condition, the centered information-acquisition complexity is
infinite, so our current upper bound in \Cref{thm:main-nonseparated-upper} becomes vacuous. An open question is
whether a more general information-acquisition complexity can yield
logarithmic stochastic regret guarantees on a nontrivial class of
nonsharp instances. 
If this is true, the next question is whether a purely FTRL-based algorithm
can achieve an $\gO(\log T)$ regret guarantee in the stochastic regime
while retaining an $\gO(\sqrt{T})$ regret guarantee in the adversarial
regime on this class of instances.
Such a
complexity should reflect the information cost of learning and reduce to
the information-acquisition complexity $\cstar(\gA,\theta)$
when $\gA$ is finite and the optimal action is unique.
We leave this for future work.
\end{itemize}

\subsection*{AI Use Statement}
The authors have developed the main ideas and guided the overall direction of this work. 
GPT-6 Astra has assisted in developing some proofs from ideas provided by the authors, helped write parts of the proofs, and improved the presentation of the paper.
The authors have checked and verified all proof derivations and text generated by GPT-6 Astra.
The authors take responsibility for the final content of this work,
including text, claims or artifacts produced with the aid of LLMs.

\clearpage
\bibliographystyle{alpha}
\bibliography{ref}

\clearpage
\appendix
\crefalias{section}{appendix}
\crefalias{subsection}{subappendix}
\crefalias{subsubsection}{subsubappendix}

\section{Additional Literature Review}\label{sec:intro-literature}
In general, BOBW algorithms for bandit problems fall into two categories.
Early BOBW algorithms leverage a detect-and-switch scheme. Specifically, this framework uses statistical tests to distinguish the stochastic and adversarial regimes 
and then invokes a specific algorithm tailored to the detected regime \cite{BubeckSlivkins2012,SeldinSlivkins2014,AuerChiang2016,SeldinLugosi2017}.  
A second line of research avoids explicit regime detection and derives BOBW guarantees from a single FTRL procedure \cite{WeiLuo2018,ZimmertSeldin2021,Ito21,JinLiuLuo2023}. In particular, these methods can also obtain 
refined guarantees in intermediate regimes, including 
the corrupted stochastic regime
\cite{lykouris2018stochastic,ZimmertSeldin2021}.

More recently, the BOBW guarantee has also been extended to bandit models with structured feedback, including combinatorial semi-bandits \cite{ZimmertLuoWei2019,Ito21Hybrid,TsuchiyaIH23}, online learning to rank \cite{ChenZL22}, bandits with graph feedback~\cite{ErezK21,Ito2022nearly,Rouyer2022nearoptimal,0002ZL22,DannWZ23,ito2024adaptive,tsuchiyasimple}, and Markov decision processes \cite{JinLuo2020,JinHuangLuo2021,DannWeiZimmert2023}.
For linear bandits, \cite{LeeEtAl2021} establish the first BOBW regret guarantee using the detect-and-switch framework.
In particular, \cite{LeeEtAl2021} achieve $\gO(c^*\log^2 T)$ regret in the stochastic regime.
Subsequently,
\cite{KongZ023,ItoT23,DannWZ23,ItoT23EXO,ItoTsuchiyaHonda2024,ZhaoItoLi2025} establish the BOBW regret guarantee via data-dependent learning rate schedules, though all of these algorithms are only shown to achieve an $\gO(\operatorname{poly}(\log T)/\Delta_{\min})$ regret in the stochastic regime.

Among the above works, a closely related one is \cite{ItoT23EXO}, which shows that the exploration by optimization (EXO) framework \cite{LattimoreSzepesvari2020EXO,LattimoreGyorgy2021}  can achieve the BOBW regret guarantee for linear bandits.
Specifically, the EXO framework \cite{LattimoreSzepesvari2020EXO,LattimoreGyorgy2021,ItoT23EXO} jointly optimizes sampling and loss
estimation to minimize a worst-case one-round regret bound:
\[
\begin{aligned}
\inf_{p,g}\ \sup_{\nu\in\mathcal M}\quad &
\underbrace{\ell_\nu(p)-\ell_\nu(q_t)}_{\text{exploration cost}}
+\underbrace{\operatorname{bias}_\nu(p,g;q_t)}_{\text{estimation bias}}
+\beta_t\underbrace{\E_{\nu,p}
  [\operatorname{stab}_{q_t}(\widehat\ell_g/\beta_t)]}_{\text{stability cost}}
\\
\text{subject to}\quad &p\in\mathcal P(\mathcal X)\,,\qquad g\in\mathcal G \,,
\end{aligned}
\]
where $q_t$ is a distribution representing the current FTRL update, $\beta_t>0$ is the (inverse) learning rate, $\mathcal M$ contains
the admissible loss and feedback models, and $\ell_\nu(p)$ is the
expected loss under $p$ in model $\nu$, with $\E_{\nu,p}$ averaging over
the sampled action and feedback.
The constraints require $p$ to be a probability distribution on the
decision domain $\mathcal X$ and $g$ to produce the loss estimate
$\widehat\ell_g$ using only the sampled action and its observed feedback.
The admissible classes also encode positivity where needed for importance
weighting and, in \cite{ItoT23EXO}, linearity of the estimated loss in the
evaluated action.
The bias term bounds the error in expected loss differences for all comparators, while the stability term, or its upper bound, controls the
effect of the estimated loss on the FTRL update.
Thus, these programs trade exploration cost against estimation bias and
stability, yet unbiasedness is not imposed as a separate constraint.
\cite{ItoT23EXO} establish the BOBW result of EXO for linear bandits by
combining the log-concavity of continuous exponential weights and
covariance comparisons with data-dependent learning rates.
However, their analysis bounds
 the total probability assigned to suboptimal actions by a comparison distribution
$\widetilde p$ with the same mean as $q_t$ via
$1-\widetilde p(a^*)\le
\E_{a\sim\widetilde p}[\Delta(a)]/\Delta_{\min}$, which eventually leads to the coarser $\gO(\log T/\Delta_{\min})$ regret in the stochastic regime.

At a very high level, our algorithm is similar to the EXO framework, in the sense that
both this framework and our algorithms solve an optimization problem to determine how to explore.
However, we remark that the ideas underlying our algorithm designs and analyses differ substantially from those in these works.
For instance, algorithmically, 
our algorithms use a fixed unbiased least-squares estimator and a
deterministic learning-rate schedule $\beta_t\asymp \sqrt{t}$, and
optimize only the sampling
distribution $p_t$ to minimize the empirical exploration cost subject to carefully designed gap-calibrated magnitude and second-moment constraints
and an exponential stability constraint. 
Our analysis uses positive and inverse
moments of (buffer-shifted) cumulative loss differences to compare the optimizer
with any feasible sampling allocation, while concentration of the FTRL
distribution $q_t$ ensures that the total penalty and stability costs are $T$-independent constants
 in the stochastic regime. 
This leaves the exploration term to determine the
logarithmic coefficient, yielding the desired
$\gO(\cstar\log T)$-type regret in the stochastic regime. 
For sharp compact instances with arbitrarily small gaps, 
we further incorporate geometrically scaled constraints and a mixed discrete and
continuous reference measure to extend our analysis and obtain an
$\gO(\cctr\log T)$-type regret, without requiring a positive minimum gap.

\section{Proof of Theorem~\ref*{thm:G-main}}\label{app:finite-results}
This appendix deduces \Cref{thm:G-main} from
\Cref{prop:G-compatibility} and a finite-mass sampling allocation.
The proposition for a fixed allocation and its supporting estimates are proved in
\Cref{app:finite-compatibility-proof}.
All sampling allocations used in the analysis are fixed independently of the horizon.
In the finite-action proofs, $c^{n}=c^{n}(\gA,\theta)$ is the cost of a
raw allocation and $\cstar=\cstar(\gA,\theta)$ is its optimal value.

\subsection{Completion of the Proof}\label{app:finite-upper-proof}
\begin{proof}[Proof of \Cref{thm:G-main}]
By \Cref{lem:AB-selection}, the selected sampling law is a
measurable feasible optimizer with its prescribed reconstruction.
In particular, for every feasible covariance
reconstruction $(p^c,v(p^c))$,
\begin{align}
 \langle p_t,\widehat\Delta_t\rangle
 &=\min_{p\text{ satisfying }\eqref{eq:G-program}}
                 \langle p,\widehat\Delta_t\rangle
 \le\langle p^c,\widehat\Delta_t\rangle \,.
 \label{eq:rev-A-objective}
\end{align}
This is the objective comparison used in the proof of
\Cref{prop:G-compatibility}. All quantities in the algorithm depend only on
$\cA,t,\sum_{s=1}^{t-1}\widehat\theta_s$, with $\widehat L_{0,a}=0$, so neither the horizon nor an unknown
instance parameter is an input.

Fix the finite sampling allocation $n$ supplied by
\Cref{lem:F-design-mass}, independently of $T,C$.
Set $M_{\cA,\theta}=M_n$, where $M_n$ is the constant for the fixed allocation
in \Cref{prop:G-compatibility}. Its corrupted stochastic
bound gives
\begin{align}
 \Reg
 &\le2^{22}Bc^{n}\log(eT)+M_n+J\sqrt C
 \label{eq:rev-A-cor-design}\\
 &\le2^{23}B\cstar\log(eT)+M_{\cA,\theta}+J\sqrt C\,.
 \label{eq:rev-A-corruption}
\end{align}
The last inequality uses $c^{n}\le2\cstar$. The sampling allocation has finite mass,
including a zero-cost sampling allocation when $\cstar=0$, so all transient constants
are finite and independent of $T,C$. No limiting sequence of sampling allocations is
used to define them.

The tuning satisfies
\begin{align}
 P=2\lceil256+\log(2K)\rceil
 &\le514+2\log2+2\log K\le2048\log(eK)\,,
 \label{eq:rev-A-coefficients-P}\\
 B=2^{24}P&\le2^{35}\log(eK)\,,\qquad 2^{23}B\le2^{58}\log(eK)\,.
 \label{eq:rev-A-coefficients}
\end{align}
The first inequality uses $\lceil x\rceil\le x+1$ and $\log(eK)=1+\log K$;
the second follows by multiplication. The proposition for a fixed allocation also
gives $J\le2^{14}\sqrt{d\log(eK)}\Delta_{\min}^{-129/256}$.
Taking the minimum of \eqref{eq:rev-A-corruption} and the adversarial
bound \eqref{eq:G-adv}, both valid for the same run and every fixed
comparator, proves \eqref{eq:G-cor-main}.

A singleton action set has identically zero regret.
\end{proof}

\subsection{Finite Allocation Mass and the Least-Squares Lift}
\begin{definition}[Finite geometric interpolation constant]\label{def:AB-geometry}
For each nonzero action $b$, set $u_b=b/\norm{b}$ and let $U_b$ be an orthonormal basis matrix of $u_b^\perp$. For $a\ne b$, define
$x_{b,a}=U_b^\top a$, $z_{b,a}=u_b^\top a$. For each invertible $(d-1)$-row submatrix $\mathsf B_{b,I}$ with rows $x_{b,a}^\top$, let
$v_{b,I}=\mathsf B_{b,I}^{-1}(z_{b,a})_{a\in I}$ and define
\begin{equation}\label{eq:F-geometry}
 \mathcal G_\cA=\max\left\{1,
 \max_{b:b\ne0}\max_{I:\det\mathsf B_{b,I}\ne0}\max_{a\ne b}
 \frac{|z_{b,a}-x_{b,a}^\top v_{b,I}|}{\norm{b}}\right\}\,.
\end{equation}
In dimension one, use the empty basis and the zero prediction.
The residual $z_{b,a}-x_{b,a}^\top v_{b,I}$ measures how much of
$a$ in the direction of $b$ is not predicted by an interpolant
on $b^\perp$. Dividing this residual by $\|b\|$ expresses the
free-direction information cost in units of the action $b$.
\end{definition}

The allocation-mass bound below uses $\mathcal G_\cA$ to control how much
information must be supplied in the optimal-action direction. Before using
it as an instance constant, we check that it is finite and does not depend
on the orthonormal coordinates chosen for the projections.

\begin{lemma}[Finiteness and coordinate invariance of the interpolation constant]
\label{lem:AB-geometry-invariance}
The bases in \eqref{eq:F-geometry} in \Cref{def:AB-geometry} exist, and the number
$\mathcal G_\cA$ is finite and independent of the choice of each $U_b$.
\end{lemma}
\begin{proof}[Proof of \Cref{lem:AB-geometry-invariance}]
Fix $b\ne0$. If $h\in\R^{d-1}$ is orthogonal to every projected
action, then
\begin{align}
 h^\top x_{b,a}=0\ (a\ne b)
 &\ \Longrightarrow\ (U_bh)^\top a=0\ (a\ne b)\,,
 \qquad (U_bh)^\top b=\|b\|h^\top U_b^\top u_b=0\,,
 \label{eq:AB-geometry-span}\\
 &\ \Longrightarrow\ U_bh=0
 \ \Longrightarrow\ h=U_b^\top U_bh=0\,.
 \label{eq:AB-geometry-rank}
\end{align}
Equation \eqref{eq:AB-geometry-span} substitutes $x_{b,a}=U_b^\top a$;
\eqref{eq:AB-geometry-rank} uses spanning of $\cA$ and $U_b^\top U_b=I$.
Thus the projected rows have rank $d-1$ and contain an invertible basis.

For another orthonormal basis $\widetilde U_b=U_bO$ with $O^\top O=I$,
\begin{align}
 \widetilde x_{b,a}&=O^\top x_{b,a}\,,\qquad
 \widetilde{\mathsf B}_{b,I}=\mathsf B_{b,I}O\,,
 \label{eq:AB-geometry-change-basis}\\
 \widetilde v_{b,I}
 &=(\mathsf B_{b,I}O)^{-1}z_I
 =O^\top\mathsf B_{b,I}^{-1}z_I=O^\top v_{b,I}\,,
 \label{eq:AB-geometry-interpolant-change}\\
 z_{b,a}-\widetilde x_{b,a}^\top\widetilde v_{b,I}
 &=z_{b,a}-x_{b,a}^\top OO^\top v_{b,I}
 =z_{b,a}-x_{b,a}^\top v_{b,I}\,.
 \label{eq:AB-geometry-residual-invariance}
\end{align}
Equation \eqref{eq:AB-geometry-change-basis} applies the coordinate change
row by row; \eqref{eq:AB-geometry-interpolant-change} inverts the product;
\eqref{eq:AB-geometry-residual-invariance} uses orthogonality of $O$.
There are finitely many pairs $(b,I)$ and actions $a$, every retained
matrix is invertible, and each denominator $\|b\|$ is positive.
Their finitely many residual ratios therefore have a finite maximum.
The same argument holds in dimension one with the stated empty-basis
convention.
\end{proof}

We first optimize the information supplied away from the optimal-action
direction, where every unit of mass has positive regret cost. The next
lemma shows that this projected problem has an optimizer of controlled
mass and relates its information constraints to those of the full sampling allocation.
These are the ingredients needed to construct a finite full sampling allocation in
\Cref{lem:F-design-mass}.

\begin{lemma}[Attained projected sampling allocation and Schur comparison]\label{lem:AB-projected-design}
Suppose $a^*\ne0$ and $r=d-1\ge1$. In the orthonormal coordinates
\eqref{eq:rev-A-coordinates}, the program \eqref{eq:rev-A-quotient}
has a minimizer $w^*$ satisfying
\begin{equation*}
 V(w^*)\succ0\,,\qquad
 \sum_{a\ne a^*}w_a^*\Delta(a)=c_Q\le \cstar\,,\qquad
 c_Q\le2r/\Delta_{\min}\,,\qquad \sum_{a\ne a^*}w_a^*\le2r/\Delta_{\min}^2\,. \notag
\end{equation*}
Every feasible full sampling allocation satisfies the Schur identity
\eqref{eq:schur-general}. When $a^*=0$, the attainment conclusion
holds with $x_a=a$, $r=d$, and $c_Q=\cstar$, without a removed direction.
\end{lemma}
\begin{proof}[Proof of \Cref{lem:AB-projected-design}]
First suppose $a^*\ne0$ and $r=d-1\ge1$. Set
$u=a^*/\|a^*\|$, $\nu^*=\|a^*\|$, and let the columns of $U$ form an
orthonormal basis of $u^\perp$. Thus
\begin{align}
 a&=Ux_a+uz_a\,,\qquad
 x_a=U^\top a\,,\quad z_a=u^\top a\,,\quad
 x_{a^*}=0\,,\quad z_{a^*}=\nu^*\,. \label{eq:rev-A-coordinates}
\end{align}
Define the quotient program by
\begin{align}
 c_Q&=\inf_{w\ge0}\left\{\sum_{a\ne a^*}w_a\Delta(a):
 V(w)=\sum_{a\ne a^*}w_ax_ax_a^\top\succ0\,,\quad
 x_a^\top V(w)^{-1}x_a\le\frac{\Delta(a)^2}{2}\ (a\ne a^*)\right\}\,.
 \label{eq:rev-A-quotient}
\end{align}
For a feasible full sampling allocation, let
$V=\sum_{a\ne a^*}n_ax_ax_a^\top$,
$v=\sum_{a\ne a^*}n_ax_az_a$, and
$s_0=\sum_{a\ne a^*}n_az_a^2$. In the orthonormal coordinates $[U,u]$,
\begin{align}
 \mathsf H&\coloneq [U,u]^\top H_n[U,u]
 =\begin{pmatrix}V&v\\v^\top&n_{a^*}(\nu^*)^{2}+s_0\end{pmatrix}\,,
 \qquad V\succ0\,,
 \label{eq:AB-Schur-coordinate-matrix}\\
 h&=n_{a^*}(\nu^*)^{2}+s_0-v^\top V^{-1}v>0\,,\qquad
 T=\begin{pmatrix}I_r&0\\v^\top V^{-1}&1\end{pmatrix}\,,
 \label{eq:AB-Schur-complement}\\
 T\begin{pmatrix}V&0\\0&h\end{pmatrix}T^\top
 &=\begin{pmatrix}V&0\\v^\top&h\end{pmatrix}
      \begin{pmatrix}I_r&V^{-1}v\\0&1\end{pmatrix}
 =\begin{pmatrix}V&v\\v^\top&v^\top V^{-1}v+h\end{pmatrix}
 =\mathsf H\,,
 \label{eq:AB-Schur-factorization}\\
 T^{-1}&=\begin{pmatrix}I_r&0\\-v^\top V^{-1}&1\end{pmatrix}\,,\qquad
 T^{-1}\begin{pmatrix}x_a\\z_a\end{pmatrix}
 =\begin{pmatrix}x_a\\z_a-v^\top V^{-1}x_a\end{pmatrix}\,,
 \label{eq:AB-Schur-coordinate-transform}\\
 \mathsf H^{-1}
 &=T^{-\top}\begin{pmatrix}V^{-1}&0\\0&h^{-1}\end{pmatrix}T^{-1}
 =\begin{pmatrix}
 V^{-1}+V^{-1}vv^\top V^{-1}/h&-V^{-1}v/h\\
 -v^\top V^{-1}/h&1/h
 \end{pmatrix}\,,
 \label{eq:AB-Schur-inverse}\\
 a^\top H_n^{-1}a
 &=\begin{pmatrix}x_a\\z_a-v^\top V^{-1}x_a\end{pmatrix}^{\!\top}
   \begin{pmatrix}V^{-1}&0\\0&h^{-1}\end{pmatrix}
   \begin{pmatrix}x_a\\z_a-v^\top V^{-1}x_a\end{pmatrix}\,,
 \label{eq:AB-Schur-quadratic-substitution}\\
 &=x_a^\top V^{-1}x_a+
 \frac{(z_a-x_a^\top V^{-1}v)^2}
 {n_{a^*}(\nu^*)^{2}+s_0-v^\top V^{-1}v}\,.
 \label{eq:schur-general}
\end{align}
Equations \eqref{eq:AB-Schur-coordinate-matrix} and
\eqref{eq:AB-Schur-complement} define the projected covariance and its
positive Schur complement; their positivity follows because $H_n\succ0$.
Equation \eqref{eq:AB-Schur-factorization} multiplies the triangular
factors explicitly. Equation \eqref{eq:AB-Schur-coordinate-transform}
subtracts the least-squares prediction from $z_a$.
Equation \eqref{eq:AB-Schur-inverse} inverts the factorization,
\eqref{eq:AB-Schur-quadratic-substitution} substitutes the transformed
action, and \eqref{eq:schur-general} expands the resulting diagonal
quadratic form.
The residual square in \eqref{eq:schur-general} quantifies the extra
variance in the removed optimal-action direction. Its nonnegativity gives
\begin{align}
 x_a^\top V^{-1}x_a
 &\le x_a^\top V^{-1}x_a+
       \frac{(z_a-x_a^\top V^{-1}v)^2}{h}
 =a^\top H_n^{-1}a\le\Delta(a)^2/2\,,
 \label{eq:AB-quotient-feasibility-comparison}\\
 c_Q&\le\sum_{a\ne a^*}n_a\Delta(a)
 \quad\text{for every feasible }n\,,
 \qquad c_Q\le\inf_{n\text{ feasible}}\sum_{a\ne a^*}n_a\Delta(a)=\cstar\,.
 \label{eq:AB-quotient-cost-comparison}
\end{align}
Equation \eqref{eq:AB-quotient-feasibility-comparison} applies the raw
information constraint after the exact decomposition. Equation
\eqref{eq:AB-quotient-cost-comparison} applies the quotient infimum to
that induced sampling allocation and then takes the infimum over full sampling allocations.

For a concrete quotient feasible sampling allocation, apply
\Cref{lem:gopt} to the spanning vectors
$y_a=x_a/\sqrt{\Delta(a)}$ in $\R^r$. Its law $\pi$ has matrix
$M_\pi=\sum_{a\ne a^*}\pi_a x_ax_a^\top/\Delta(a)\succ0$ and
$y_a^\top M_\pi^{-1}y_a\le r$. With
$w_a=(2r/\Delta_{\min})\pi_a/\Delta(a)$, direct substitution gives
\begin{subequations}\label{eq:rev-A-quotient-feasible}
\begin{align}
 V(w)&=\sum_{a\ne a^*}\frac{2r\pi_a}{\Delta_{\min}\Delta(a)}x_ax_a^\top
       =\frac{2r}{\Delta_{\min}}M_\pi\,,
 \label{eq:rev-A-quotient-feasible-step-1}\\
 x_a^\top V(w)^{-1}x_a
 &=\frac{\Delta_{\min}}{2r}x_a^\top M_\pi^{-1}x_a
 =\frac{\Delta_{\min}\Delta(a)}{2r}y_a^\top M_\pi^{-1}y_a
 \le\frac{\Delta_{\min}\Delta(a)}{2}\le\frac{\Delta(a)^2}{2}\,,
 \label{eq:rev-A-quotient-feasible-step-2}\\
 \sum_{a\ne a^*}w_a\Delta(a)
 &=\frac{2r}{\Delta_{\min}}\sum_{a\ne a^*}\pi_a=\frac{2r}{\Delta_{\min}}\,,
 \qquad c_Q\le\frac{2r}{\Delta_{\min}}\,.
 \label{eq:rev-A-quotient-feasible-step-3}
\end{align}
\end{subequations}
Equation \eqref{eq:rev-A-quotient-feasible-step-1} substitutes every $w_a$ in the covariance before collecting the scalar multiple of $M_\pi$. Equation \eqref{eq:rev-A-quotient-feasible-step-2} uses $(cM)^{-1}=c^{-1}M^{-1}$, substitutes $x_a=\sqrt{\Delta(a)}y_a$, applies the G-optimal leverage bound $y_a^\top M_\pi^{-1}y_a\le r$, and then uses $\Delta_{\min}\le\Delta(a)$. Equation \eqref{eq:rev-A-quotient-feasible-step-3} substitutes $w_a\Delta(a)=(2r/\Delta_{\min})\pi_a$ and $\sum_a\pi_a=1$.

To justify attainment, take a minimizing sequence $w^{(m)}$ with bounded
cost. Since $\Delta(a)\ge\Delta_{\min}>0$,
\begin{subequations}\label{eq:rev-A-attainment-lmi}
\begin{align}
 \sum_{a\ne a^*}w_a^{(m)}
 &\le\Delta_{\min}^{-1}\sum_{a\ne a^*}w_a^{(m)}\Delta(a)\,,
 \label{eq:rev-A-attainment-lmi-step-1}\\
 x_a^\top V(w^{(m)})^{-1}x_a\le\Delta(a)^2/2
 &\iff
 \begin{pmatrix}V(w^{(m)})&x_a\\x_a^\top&\Delta(a)^2/2\end{pmatrix}
 \succeq0\,.
 \label{eq:rev-A-attainment-lmi-step-2}
\end{align}
\end{subequations}
Equation \eqref{eq:rev-A-attainment-lmi-step-1} sums $\Delta_{\min} w_a^{(m)}\le\Delta(a)w_a^{(m)}$ over all suboptimal actions, so a cost sublevel is bounded. Equation \eqref{eq:rev-A-attainment-lmi-step-2} is the Schur-complement equivalence for $V(w^{(m)})\succ0$; expressing feasibility by a positive semidefinite block matrix makes its preservation under a convergent subsequence explicit.

A subsequence converges to $w^*\ge0$, and the block inequalities in
\eqref{eq:rev-A-attainment-lmi} survive the limit. If
$h\in\ker V(w^*)$, positivity of each limiting block matrix gives
\begin{subequations}\label{eq:rev-A-attainment-kernel}
\begin{align}
 0&\le
 \begin{pmatrix}h\\\lambda\end{pmatrix}^{\!\top}
 \begin{pmatrix}V(w^*)&x_a\\x_a^\top&\Delta(a)^2/2\end{pmatrix}
 \begin{pmatrix}h\\\lambda\end{pmatrix}
 =2\lambda h^\top x_a+\lambda^2\Delta(a)^2/2
 \quad(\lambda\in\R)\,,
 \label{eq:rev-A-attainment-kernel-step-1}\\
 0&\le2\left(-\frac{2h^\top x_a}{\Delta(a)^2}\right)h^\top x_a
   +\left(-\frac{2h^\top x_a}{\Delta(a)^2}\right)^2\frac{\Delta(a)^2}{2}
 =-\frac{2(h^\top x_a)^2}{\Delta(a)^2}\,,
 \label{eq:AB-quotient-kernel-test}\\
 h^\top x_a&=0\quad(a\ne a^*)\,,\qquad h=0\,.
 \label{eq:rev-A-attainment-kernel-step-2}
\end{align}
\end{subequations}
Equation \eqref{eq:rev-A-attainment-kernel-step-1} evaluates the limiting positive semidefinite block form on $(h,\lambda)$ and uses $h^\top V(w^*)h=0$. To obtain \eqref{eq:rev-A-attainment-kernel-step-2}, set $\lambda=-2h^\top x_a/\Delta(a)^2$ in \eqref{eq:rev-A-attainment-kernel-step-1}: its right side becomes $-2(h^\top x_a)^2/\Delta(a)^2$, so $h^\top x_a=0$ for every $a$; spanning then gives $h=0$.

Equation \eqref{eq:AB-quotient-kernel-test} substitutes
$\lambda=-2h^\top x_a/\Delta(a)^2$ in the positive semidefinite quadratic
form. Its RHS is nonpositive, so it vanishes.
Equation \eqref{eq:rev-A-attainment-kernel-step-2} then uses spanning
of the $x_a$. Thus $V(w^*)\succ0$,
$w^*$ is feasible, and
\begin{align}
 \sum_{a\ne a^*}w_a^*\Delta(a)=c_Q\,,\qquad
 \sum_{a\ne a^*}w_a^*\le c_Q/\Delta_{\min}\le2r/\Delta_{\min}^2\,.
 \label{eq:rev-A-quotient-mass}
\end{align}

When $a^*=0$, set $x_a=a$ and $r=d$. Removing the zero action
leaves a spanning family, and the same G-optimal construction and
compact-sublevel argument apply to these explicitly chosen $x_a$.
The two feasible programs now coincide:
\begin{align}
 H_n&=n_{a^*}0\,0^\top+\sum_{a\ne a^*}n_aaa^\top
      =\sum_{a\ne a^*}n_ax_ax_a^\top=V(n_{-a^*})\,,
 \label{eq:AB-zero-quotient-covariance}\\
 a^\top H_n^{-1}a&=x_a^\top V(n_{-a^*})^{-1}x_a\,,\qquad
 \sum_a n_a\Delta(a)=\sum_{a\ne a^*}n_a\Delta(a)\,,
 \label{eq:AB-zero-quotient-identities}\\
 c_Q&=\cstar\,,\qquad V(w^*)\succ0\,,\qquad
 \sum_{a\ne a^*}w_a^*\Delta(a)=\cstar\le2d/\Delta_{\min}\,,\qquad
 \sum_{a\ne a^*}w_a^*\le2d/\Delta_{\min}^2\,.
 \label{eq:AB-zero-quotient-conclusion}
\end{align}
Equation \eqref{eq:AB-zero-quotient-covariance} removes the zero matrix;
\eqref{eq:AB-zero-quotient-identities} identifies both constraints and
objectives; \eqref{eq:AB-zero-quotient-conclusion} applies
\eqref{eq:rev-A-quotient-feasible} and
\eqref{eq:rev-A-quotient-mass} at $r=d$.
\end{proof}

The proposition for a fixed allocation compares the learner with a fixed sampling allocation of
finite total mass, whereas $\cstar$ is defined by an infimum that need not be
attained by a full sampling allocation. The next lemma supplies a single finite sampling allocation
whose cost is at most $2\cstar$. It allows the proof of
\Cref{thm:G-main} to use horizon-independent constants,
including when $\cstar=0$.

\begin{lemma}[Finite mass at cost at most $2\cstar$]\label{lem:F-design-mass}
There is a finite feasible sampling allocation with
\begin{equation*}
 c^{n}\le2\cstar\,,\qquad N_n\le4(d+\mathcal G_\cA^2)\Delta_{\min}^{-2}\,. \notag
\end{equation*}
Also $\cstar\le2d\Delta_{\min}^{-1}$.
\end{lemma}
\begin{proof}[Proof of \Cref{lem:F-design-mass}]
For $a^*\ne0$ and $r=d-1\ge1$, use the attained sampling allocation $w^*$,
coordinates $(x_a,z_a)$, and $\nu^*=\|a^*\|$ from
\Cref{lem:AB-projected-design}. In particular,
\begin{equation*}
 x_a^\top V(w^*)^{-1}x_a\le\Delta(a)^2/2\,,\qquad
 \sum_{a\ne a^*}w_a^*\Delta(a)=c_Q\le \cstar\,,\qquad
 \sum_{a\ne a^*}w_a^*\le2(d-1)/\Delta_{\min}^2\,. \notag
\end{equation*}
The following construction restores the removed $a^*$ direction while
controlling the extra mass assigned to the zero-gap action.

Let $V^*=V(w^*)$, $v_{\rm LS}^*=\sum_{a\ne a^*}w_a^*x_az_a$, and
$\widehat v=(V^*)^{-1}v_{\rm LS}^*$.
\Cref{lem:WLS} supplies weights $\omega_I\ge0$ summing
to one, so \eqref{eq:F-geometry} in \Cref{def:AB-geometry} gives
\begin{align}
 z_a-x_a^\top\widehat v
 &=z_a-x_a^\top\sum_I\omega_Iv_{a^*,I}
 =\sum_I\omega_I(z_a-x_a^\top v_{a^*,I})\,,
 \label{eq:AB-lift-residual-convex-substitution}\\
 |z_a-x_a^\top\widehat v|
 &\le\sum_I\omega_I|z_a-x_a^\top v_{a^*,I}|
 \le \nu^*\mathcal G_\cA\sum_I\omega_I=\nu^*\mathcal G_\cA\,,
 \label{eq:AB-lift-residual-uniform}\\
 \sum_{a\ne a^*}w_a^*(z_a-x_a^\top y)^2
 &=\sum_{a\ne a^*}w_a^*\{z_a^2-2z_ax_a^\top y+y^\top x_ax_a^\top y\}\,,
 \label{eq:AB-lift-LS-expand}\\
 &=\sum_{a\ne a^*}w_a^*z_a^2-2y^\top v_{\rm LS}^*+y^\top V^*y\,,
 \label{eq:AB-lift-LS-collect}\\
 &=\sum_{a\ne a^*}w_a^*z_a^2-\widehat v^\top V^*\widehat v
                         +(y-\widehat v)^\top V^*(y-\widehat v)\,,
 \label{eq:AB-lift-LS-complete-square}\\
 \min_y\sum_{a\ne a^*}w_a^*(z_a-x_a^\top y)^2
 &=\sum_{a\ne a^*}w_a^*z_a^2
             -(v_{\rm LS}^*)^\top (V^*)^{-1}v_{\rm LS}^*\ge0\,.
 \label{eq:rev-A-lift-residual}
\end{align}
Equation \eqref{eq:AB-lift-residual-convex-substitution} substitutes the
interpolant representation in \Cref{lem:WLS} and
$\sum_I\omega_I=1$; \eqref{eq:AB-lift-residual-uniform} applies the
residual bound defining $\mathcal G_\cA$. Equations
\eqref{eq:AB-lift-LS-expand} and \eqref{eq:AB-lift-LS-collect} first expand
each square and then use the definitions of $V^*$ and $v_{\rm LS}^*$.
Equation \eqref{eq:AB-lift-LS-complete-square} uses
$V^*\widehat v=v_{\rm LS}^*$. The minimum in
\eqref{eq:rev-A-lift-residual} is attained at $y=\widehat v$, and its
nonnegativity follows from the original sum of weighted squares.
For $\epsilon>0$, set
$n_a=(1+\epsilon)w_a^*$ for $a\ne a^*$ and
$n_{a^*}=2(1+\epsilon)\mathcal G_\cA^2\Delta_{\min}^{-2}/\epsilon$.
The suboptimal block and cross-vector are both multiplied by
$1+\epsilon$, so their least-squares coefficient remains $\widehat v$.
Substituting this sampling allocation in \eqref{eq:schur-general} in \Cref{lem:AB-projected-design}, before bounding
either summand, gives
\begin{subequations}\label{eq:rev-A-lift-substitution}
\begin{align}
 a^\top H_n^{-1}a
 &=\frac{x_a^\top (V^*)^{-1}x_a}{1+\epsilon}
 +\frac{(z_a-x_a^\top\widehat v)^2}
 {n_{a^*}(\nu^*)^{2}+(1+\epsilon)
  \{\sum_{b\ne a^*}w_b^*z_b^2-(v_{\rm LS}^*)^\top (V^*)^{-1}v_{\rm LS}^*\}}
 \label{eq:rev-A-lift-substitution-step-1}\\
 &\le\frac{\Delta(a)^2}{2(1+\epsilon)}
       +\frac{(\nu^*)^{2}\mathcal G_\cA^2}{n_{a^*}(\nu^*)^{2}}
 \label{eq:rev-A-lift-substitution-step-2}\\
 &=\frac{\Delta(a)^2}{2(1+\epsilon)}
       +\frac{\epsilon}{2(1+\epsilon)\Delta_{\min}^{-2}}
 \label{eq:rev-A-lift-substitution-step-3}\\
 &\le\frac{\Delta(a)^2}{2(1+\epsilon)}
       +\frac{\epsilon\Delta(a)^2}{2(1+\epsilon)}
 =\frac{\Delta(a)^2}{2}\,.
 \label{eq:rev-A-lift-substitution-step-4}
\end{align}
\end{subequations}
Equation \eqref{eq:rev-A-lift-substitution-step-1} substitutes $V=(1+\epsilon)V^*$, $v=(1+\epsilon)v_{\rm LS}^*$, and the stated $n_{a^*}$ into the Schur identity. Equation \eqref{eq:rev-A-lift-substitution-step-2} applies quotient feasibility, the squared residual bound, and nonnegativity of the least-squares residual sum. Equation \eqref{eq:rev-A-lift-substitution-step-3} substitutes $n_{a^*}=2(1+\epsilon)\mathcal G_\cA^2\Delta_{\min}^{-2}/\epsilon$ and cancels $(\nu^*)^{2}\mathcal G_\cA^2$; \eqref{eq:rev-A-lift-substitution-step-4} uses $\Delta_{\min}^{2}\le\Delta(a)^2$ and combines the two fractions.

Here $\Delta_{\min}^2\le\Delta(a)^2$; positivity of the
Schur complement follows from $n_{a^*}(\nu^*)^{2}>0$. Therefore
\begin{subequations}\label{eq:rev-A-lift-cost-mass}
\begin{align}
 c_Q\le \cstar\le(1+\epsilon)c_Q\quad(\epsilon>0)
 &\quad\Longrightarrow\quad \cstar=c_Q\,,
 \label{eq:rev-A-lift-cost-mass-step-1}\\
 c^{n}\big|_{\epsilon=1}&=2\cstar\,,
 \label{eq:rev-A-lift-cost-mass-step-2}\\
 N_n\big|_{\epsilon=1}
 &=2\sum_{a\ne a^*}w_a^*+4\mathcal G_\cA^2\Delta_{\min}^{-2}
 \le4r/\Delta_{\min}^2+4\mathcal G_\cA^2\Delta_{\min}^{-2}
 \label{eq:rev-A-lift-cost-mass-step-3}\\
 &\le4(d+\mathcal G_\cA^2)\Delta_{\min}^{-2}\,.
 \label{eq:rev-A-lift-cost-mass-step-4}
\end{align}
\end{subequations}
Equation \eqref{eq:rev-A-lift-cost-mass-step-1} takes $\epsilon\downarrow0$ after the feasible lift of cost $(1+\epsilon)c_Q$ has been constructed; thus $c_Q\le \cstar\le c_Q$. Equation \eqref{eq:rev-A-lift-cost-mass-step-2} then chooses the fixed value $\epsilon=1$, not a horizon-dependent limit. Equation \eqref{eq:rev-A-lift-cost-mass-step-3} substitutes both groups of weights and the quotient mass bound; \eqref{eq:rev-A-lift-cost-mass-step-4} uses $r\le d$.

When $a^*\ne0$ and $d=1$, set
$n_{a^*}=4\mathcal G_\cA^2\Delta_{\min}^{-2}$ and $n_a=0$ for $a\ne a^*$.
The empty-basis convention in \eqref{eq:F-geometry} in \Cref{def:AB-geometry} gives
$|a/a^*|\le\mathcal G_\cA$, and hence
\begin{subequations}\label{eq:rev-A-one-dimensional}
\begin{align}
 H_n&=n_{a^*}(a^*)^{2}>0\,,\qquad c^{n}=0\,,
 \label{eq:rev-A-one-dimensional-step-1}\\
 a^\top H_n^{-1}a
 &=\frac{(a/a^*)^2}{4\mathcal G_\cA^2\Delta_{\min}^{-2}}
 \le\frac1{4\Delta_{\min}^{-2}}\le\frac{\Delta(a)^2}{2}\,,\qquad \cstar=0\,.
 \label{eq:rev-A-one-dimensional-step-2}
\end{align}
\end{subequations}
Equation \eqref{eq:rev-A-one-dimensional-step-1} substitutes the sampling allocation supported only on the nonzero optimal action. Equation \eqref{eq:rev-A-one-dimensional-step-2} inverts that scalar covariance, uses $|a/a^*|\le\mathcal G_\cA$, and then $\Delta_{\min}^{2}\le\Delta(a)^2$; feasibility at zero cost implies $0\le \cstar\le c^{n}=0$.

When $a^*=0$, use the attained sampling allocation in
\eqref{eq:AB-zero-quotient-conclusion} of
\Cref{lem:AB-projected-design} and set $n_a=2w_a^*$ for
$a\ne a^*$ and $n_{a^*}=0$. Direct substitution gives
\begin{align}
 H_n&=\sum_{a\ne a^*}2w_a^*aa^\top=2V(w^*)\succ0\,,
 \label{eq:AB-zero-lift-covariance}\\
 a^\top H_n^{-1}a
 &=\tfrac12 a^\top V(w^*)^{-1}a
 \le\tfrac12(\Delta(a)^2/2)\le\Delta(a)^2/2\,,
 \label{eq:AB-zero-lift-feasibility}\\
 c^{n}&=\sum_{a\ne a^*}2w_a^*\Delta(a)=2\cstar\,,\qquad
 N_n=\sum_{a\ne a^*}2w_a^*\le4d/\Delta_{\min}^2
                  \le4(d+\mathcal G_\cA^2)\Delta_{\min}^{-2}\,,
 \label{eq:AB-zero-lift-cost-mass}\\
 \cstar&\le2d\Delta_{\min}^{-1}\,.
 \label{eq:AB-design-final-gap-bound}
\end{align}
Equation \eqref{eq:AB-zero-lift-covariance} substitutes the doubled
weights; \eqref{eq:AB-zero-lift-feasibility} inverts the scalar multiple;
\eqref{eq:AB-zero-lift-cost-mass} substitutes the attained cost and mass.
Equation \eqref{eq:AB-design-final-gap-bound} holds also in the nonzero
cases above, using $r\le d$ or $\cstar=0$, respectively.
\end{proof}

Lifting the projected sampling allocation back to the full space requires controlling
a weighted least-squares residual. The next identity expresses the fitted
coefficient as a convex combination of exact interpolants, so each residual
is bounded by the interpolation constant $\mathcal G_\cA$. This is the
step used in \Cref{lem:F-design-mass} to bound the additional
mass at the optimal action.

\begin{lemma}[Weighted least squares as a convex combination of interpolants]\label{lem:WLS}
If $V=\sum_aw_a x_ax_a^\top\succ0$, $w_a\ge0$, then
\[
 V^{-1}\sum_aw_ax_a z_a=\sum_{I:\det\mathsf B_I\ne0}\omega_I v_I\,,
 \quad\omega_I=\frac{\det(\mathsf B_I)^2\prod_{a\in I}w_a}{\det V}\,,
 \quad\omega_I\ge0\,,\quad\sum_I\omega_I=1\,.
\]
\end{lemma}
\begin{proof}[Proof of \Cref{lem:WLS}]
Let $r=\dim x_a$, let $X$ have rows $x_a^\top$, set
$W=\diag(w_a)$, and let $z=(z_a)_a$. For $|I|=r$, let
$\mathsf B_I=X_{I,:}$. The Cauchy--Binet identity gives
\begin{subequations}\label{eq:rev-A-WLS-weights}
\begin{align}
 \det V
 &=\det(X^\top WX)
 =\sum_{|I|=r}\det(X_{I,:}^\top)
                   \det(W_{I,I}X_{I,:})
 \label{eq:rev-A-WLS-weights-step-1}\\
 &=\sum_{|I|=r}\det(\mathsf B_I)^2\prod_{a\in I}w_a>0\,,
 \label{eq:rev-A-WLS-weights-step-2}\\
 \omega_I&=\frac{\det(\mathsf B_I)^2\prod_{a\in I}w_a}{\det V}
 \ge0\,,\qquad
 \sum_{I:\det\mathsf B_I\ne0}\omega_I=1\,.
 \label{eq:rev-A-WLS-weights-step-3}
\end{align}
\end{subequations}
Equation \eqref{eq:rev-A-WLS-weights-step-1} applies Cauchy--Binet to the product $X^\top(WX)$. Equation \eqref{eq:rev-A-WLS-weights-step-2} uses $\det(\mathsf B_I^\top)=\det(\mathsf B_I)$ and $\det(W_{I,I}\mathsf B_I)=(\prod_{a\in I}w_a)\det(\mathsf B_I)$; positivity follows from $V\succ0$. Equation \eqref{eq:rev-A-WLS-weights-step-3} divides the nonnegative summands by their strictly positive sum.

Fix $k\in[r]$. Replace column $k$ of $X$ by $z$ to form $X^{(k)}$;
then $X^\top WX^{(k)}$ is $V$ with column $k$ replaced by
$X^\top Wz$. For invertible $\mathsf B_I$, its corresponding
row submatrix $\mathsf B_I^{(k)}$ is $\mathsf B_I$ with column $k$
replaced by $z_I$. Cramer's rule, followed by Cauchy--Binet, now gives
\begin{subequations}\label{eq:rev-A-WLS-coordinate}
\begin{align}
 [V^{-1}X^\top Wz]_k
 &=\frac{\det(X^\top WX^{(k)})}{\det V}
 \label{eq:rev-A-WLS-coordinate-step-1}\\
 &=\frac1{\det V}\sum_{|I|=r}
       \det(\mathsf B_I)\det(\mathsf B_I^{(k)})\prod_{a\in I}w_a
 \label{eq:rev-A-WLS-coordinate-step-2}\\
 &=\sum_{I:\det\mathsf B_I\ne0}
   \frac{\det(\mathsf B_I)^2\prod_{a\in I}w_a}{\det V}
   \frac{\det(\mathsf B_I^{(k)})}{\det(\mathsf B_I)}
 \label{eq:rev-A-WLS-coordinate-step-3}\\
 &=\sum_{I:\det\mathsf B_I\ne0}\omega_I[\mathsf B_I^{-1}z_I]_k
 =\sum_{I:\det\mathsf B_I\ne0}\omega_I[v_I]_k\,.
 \label{eq:rev-A-WLS-coordinate-step-4}
\end{align}
\end{subequations}
Equation \eqref{eq:rev-A-WLS-coordinate-step-1} is Cramer's rule for coordinate $k$; \eqref{eq:rev-A-WLS-coordinate-step-2} is Cauchy--Binet. In \eqref{eq:rev-A-WLS-coordinate-step-3}, singular minors and zero-weight summands vanish before division, and each nonzero determinant is factored. Equation \eqref{eq:rev-A-WLS-coordinate-step-4} substitutes $\omega_I$ and applies Cramer's rule to $\mathsf B_Iv_I=z_I$. Since $k\in[r]$ was arbitrary, these coordinate identities give the asserted vector identity.
\end{proof}

\clearpage
\section{Proof of Proposition~\ref*{prop:G-compatibility}}
\label{app:finite-compatibility-proof}
The first subsection gives the complete proof. The next four subsections
prove the penalty, stability, exploration, and direct corruption bounds
used in that proof and in the main-text sketch. The last subsection
contains only inputs shared across the term proofs: the regret
decomposition, estimator identities, bounds on entropy and the proxy function, and
the common good-event and moment estimates.
All comparisons concern the same run of \IAFTRL{}.

\subsection{Complete Formal Proof of Proposition~\ref*{prop:G-compatibility}}\label{app:B-complete}
\begin{proof}[Proof of \Cref{prop:G-compatibility}]
Fix a feasible sampling allocation $n$, and let $N_n=\sum_a n_a$.
Let $\mathsf I_1,\ldots,\mathsf I_4$
be respectively the penalty, stability, exploration under $\theta$, and
exploration-corruption sums in \eqref{eq:F-master}, and set
$\mathcal E_T=\sum_{t\le T}t^{-1/2}\sP(\mathcal G_t^c)$.
The four term lemmas proved in the next four subsections give
\begin{align}
 \Reg(u)&\le\mathsf I_1+\mathsf I_2+\mathsf I_3+\mathsf I_4
       \label{eq:B-complete-decomposition}\\
 &\le(\beta_1\log K+S_{\rm pen}+c_{R}\mathcal E_T)
       +(S_{\rm stab}+2c_{R}\mathcal E_T)\notag\\
 &\quad+(2^{22}Bc^{n}\log(eT)+M_{3,n}+8c_{R}\mathcal E_T)
       +16\sqrt2c_{R}\sqrt C
       \label{eq:B-complete-four-terms}\\
 &=2^{22}Bc^{n}\log(eT)+\beta_1\log K+S_a+M_{3,n}
       +11c_{R}\mathcal E_T+16\sqrt2c_{R}\sqrt C\,.
       \label{eq:B-complete-collection}
\end{align}
Equation \eqref{eq:B-complete-decomposition} is
\Cref{lem:AB-card-master}.
Equation \eqref{eq:B-complete-four-terms} applies
\Cref{lem:B-penalty-term,lem:B-stability-term,lem:B-exploration-term,lem:B-corruption-term}; these
lemmas explicitly define $S_{\rm pen},S_{\rm stab},M_{3,n}$.
Equation \eqref{eq:B-complete-collection} uses
$S_a=S_{\rm pen}+S_{\rm stab}$ and $1+2+8=11$.
Substituting the full weighted failure sum from
\Cref{lem:F-sums} is exactly the bound for a fixed allocation
\eqref{eq:F-per-design}, also recorded below with every constant expanded.

The all-history calculation in
\Cref{prop:AB-adversarial} gives, for every fixed comparator,
\begin{align}
 \Reg(u)&\le\beta_T\log K+10d\sum_{t=1}^T\beta_t^{-1}
 \le16d\log(eK)+21\sqrt{dT\log(eK)}\,,\qquad \Reg(u)\le2T\,.
 \notag
\end{align}
The bound is independent of $u$, so its supremum proves
\eqref{eq:G-adv} in \Cref{thm:G-main}.

For ordinary corruption, \Cref{prop:F-per-design}
gives the complete bound for a fixed allocation \eqref{eq:F-per-design}. To expose
its dependence on the budget, use the definitions in
\eqref{eq:F-thresholds} and \Cref{def:AB-card-negative}:
\begin{equation}\label{eq:F-cor-thresholds}
 \sqrt{n_0}\le\sqrt{N_{\rm burn}}+\sqrt C\,,\qquad
 \sqrt{T_\circ}\le\sqrt{Q+1}\Delta_{\min}^{-5/2}
                  +32\Delta_{\min}^{-1/2}\sqrt{n_0}\,.
\end{equation}
Indeed, $n_0=\max\{N_{\rm burn},\lceil C\rceil\}\le N_{\rm burn}+C$
and $T_\circ\le(Q+1)\Delta_{\min}^{-5}+1024n_0\Delta_{\min}^{-1}$; taking square roots
proves both inequalities. All remaining auxiliary constants in
\eqref{eq:F-per-design} are independent of $C$.
Consequently, define
\begin{subequations}\label{eq:F-cor-constants}
\begin{align}
 M_n={}&3\cdot2^{22}BA_0c^{n}\Delta_{\min}^{-\varepsilon_0}
       +\beta_1\log K+S_a+\Psi_{\Delta_{\min}} S_b
 \label{eq:F-cor-transient-1}\\
 &+22c_{R}\sqrt{Q+1}\Delta_{\min}^{-5/2}+33c_{R}\mathsf A+22c_{R}K56^P
 \label{eq:F-cor-transient-2}\\
 &+c_{R}\{704\Delta_{\min}^{-1/2}+1430
             +33c_{\rm tail}\Delta_{\min}^{-\gamma_0}\}\sqrt{N_{\rm burn}}
 \label{eq:F-cor-transient-3}\\
 &+8c_{R}\Psi_{\Delta_{\min}}(10^6a_n+30000\sqrt{a_{\Delta_{\min}}})\,,
 \label{eq:F-cor-transient-4}\\
 J_F={}&c_{R}\{704\Delta_{\min}^{-1/2}+1430
             +33c_{\rm tail}\Delta_{\min}^{-\gamma_0}+16\sqrt2\}\,.
 \label{eq:F-cor-coefficient}
\end{align}
\end{subequations}
These formulas collect every budget-independent term into
$M_n$ and every coefficient of $\sqrt C$ into $J_F$.
Substitution into \eqref{eq:F-per-design} gives
\begin{equation}\label{eq:rev-B-completion-corruption}
 \Reg\le2^{22}Bc^{n}\log(eT)+M_n+J_F\sqrt C\,.
\end{equation}
Here $c_{\rm tail}=(224u_0k)^{128/255}$, $u_0=18k+1$, and
$k^2<11/4$, $k<7/4$. Thus $u_0k<205/4$ and $c_{\rm tail}<256$.
Since $\Delta_{\min}^{-1}\ge1$ and $1/2<\gamma_0=128/255<129/256$,
\begin{equation}\label{eq:rev-B-completion-J}
 J_F\le c_{R}\Delta_{\min}^{-129/256}(704+1430+33\cdot256+16\sqrt2)
      <2^{14}\sqrt{d\log(eK)}\Delta_{\min}^{-129/256}\,.
\end{equation}
Set $J=J_F$. The constants $M_n,J$ are finite and independent of
$T,C$, and $J$ is also independent of the sampling allocation $n$ and its mass $N_n$.
This proves \eqref{eq:G-compat-cor}; its specialization at $C=0$
uses the same constant $M_n$.

\end{proof}

\subsubsection{The Regret Bound for a Fixed Allocation}\label{app:finite-per-design}
The following intermediate estimate keeps the regret cost
$c^{n}=\sum_{a\ne a^*}n_a\Delta(a)$ and total mass $N_n=\sum_a n_a$ of a fixed
feasible sampling allocation explicit. Its constants are defined in the scalar
estimates below using \eqref{eq:F-analysis-tuning} in
\Cref{def:AB-card-tuning}. The objective comparison
\eqref{eq:F-objective-comparison} follows from the minimization in
\IAFTRL{}.

\begin{proposition}[Cardinality-dependent regret bound for a fixed allocation]\label{prop:F-per-design}
\IAFTRL{} obeys, for each fixed finite
feasible sampling allocation $n$ and every $T$,
\begin{align}\label{eq:F-per-design}
 \Reg\le{}&2^{22}Bc^{n}\log(eT)+3\cdot2^{22}BA_0c^{n}\Delta_{\min}^{-\varepsilon_0}
 +\beta_1\log K+S_a+\Psi_{\Delta_{\min}} S_b\notag\\
 &+22c_{R}\sqrt{T_\circ}+33c_{R}\mathsf A+1430c_{R}\sqrt{n_0}+22c_{R}K56^P\notag\\
 &+33c_{R} c_{\rm tail}\Delta_{\min}^{-\gamma_0}\sqrt{n_0}
 +8c_{R}\Psi_{\Delta_{\min}}(10^6a_n+30000\sqrt{a_{\Delta_{\min}}})+16\sqrt2c_{R}\sqrt C\,.
\end{align}
\end{proposition}
\begin{proof}
All events $\mathcal G_t$ are $\mathcal F_{t-1}$-measurable, so the
conditional quantities in \eqref{eq:F-master} in \Cref{lem:AB-card-master} may be split with
$\ind_{\mathcal G_t}$ and $\ind_{\mathcal G_t^c}$ before expectation.
Define the three nonnegative good-event contributions
\begin{subequations}\label{eq:rev-F-per-design-components}
\begin{align}
 A_T^{\rm good}
 &\coloneq \sum_{t=2}^T(\beta_t-\beta_{t-1})
                      \E[\ind_{\mathcal G_t}H(q_t)]
     +\sum_{t=1}^T\E[\ind_{\mathcal G_t}\gC_t(p_t)]\,,
 \label{eq:rev-F-per-design-components-step-1}\\
 E_T^{\rm good}
 &\coloneq \sum_{t=T_\circ}^T
      \left\{\frac{2^{22}Bc^{n}\Psi_t}{t}
       +800K\Psi_t e^{-4t^{\alpha-1/2}/C_{\beta}}\right\}\,,
 \label{eq:rev-F-per-design-components-step-2}\\
 M_T^{\rm good}
 &\coloneq 8c_{R}\sum_{t=T_\circ}^T\frac{\Psi_t}{\sqrt t}
   \left[\min\{400,16\cdot25^5a_n^2/t\}\right.
 \label{eq:rev-F-per-design-components-step-3}\\
 &\hspace{42mm}\left.
             +\min\{400,16\cdot25^5a_{\Delta_{\min}}^2/t^2\}\right]\,.
 \label{eq:rev-F-per-design-components-step-4}
\end{align}
\end{subequations}
Equation \eqref{eq:rev-F-per-design-components-step-1} defines the combined entropy-growth and proxy function cost on good histories. Equation \eqref{eq:rev-F-per-design-components-step-2} defines the regret-cost and exponentially small exploration contributions. Equations \eqref{eq:rev-F-per-design-components-step-3} and \eqref{eq:rev-F-per-design-components-step-4} define the separate contributions from failures of the two feasibility thresholds for the comparison distribution; the two minima will be summed with different integral exponents.

A sum whose lower endpoint exceeds $T$ is zero.
\Cref{lem:F-good-stability} and \Cref{lem:F-exploration}, followed by
\eqref{eq:F-good-sums} in \Cref{lem:AB-good-sums}, \eqref{eq:F-positive-local} in \Cref{lem:B-positive-local}, and
\eqref{eq:F-mass-sum1}--\eqref{eq:F-mass-sum2} in \Cref{lem:B-mass-sums}, give
\begin{subequations}\label{eq:rev-F-per-design-good-substitution}
\begin{align}
 A_T^{\rm good}&\le S_a\,,
 \label{eq:rev-F-per-design-good-substitution-step-1}\\
 E_T^{\rm good}
 &\le2^{22}Bc^{n}
        \left\{\sum_{t=1}^T\frac1t
                     +\sum_{t\ge T_\circ}\frac{\Psi_t-1}{t}\right\}
        +800K\Psi_{\Delta_{\min}}\sum_{t\ge1}e^{-4t^{\alpha-1/2}/C_{\beta}}
 \label{eq:rev-F-per-design-good-substitution-step-2}\\
 &\le2^{22}Bc^{n}\log(eT)
      +3\cdot2^{22}BA_0c^{n}\Delta_{\min}^{-\varepsilon_0}
      +\Psi_{\Delta_{\min}} S_b\,,
 \label{eq:rev-F-per-design-good-substitution-step-3}\\
 M_T^{\rm good}
 &\le8c_{R}\Psi_{\Delta_{\min}}
       \left\{10^6a_n+30000\sqrt{a_{\Delta_{\min}}}\right\}\,.
 \label{eq:rev-F-per-design-good-substitution-step-4}
\end{align}
\end{subequations}
Equation \eqref{eq:rev-F-per-design-good-substitution-step-1} uses the full good-history sum in \Cref{lem:AB-good-sums}. Equation \eqref{eq:rev-F-per-design-good-substitution-step-2} substitutes $\Psi_t=1+(\Psi_t-1)$ before bounding the harmonic sum. Equation \eqref{eq:rev-F-per-design-good-substitution-step-3} uses $\sum_{t\le T}t^{-1}\le\log(eT)$, the bound on $\sum_{t\ge T_\circ}(\Psi_t-1)/t$, independent of $T$, in \Cref{lem:B-positive-local}, and the exponential sum. Equation \eqref{eq:rev-F-per-design-good-substitution-step-4} substitutes both capped mass integrals and $\Psi_t\le\Psi_{\Delta_{\min}}$.

On $\mathcal G_t^c$, entropy growth, the proxy function, and exploration under $\theta$ are respectively at most $c_{R}/\sqrt t$, $2c_{R}/\sqrt t$,
and $8c_{R}/\sqrt t$. Thus their total, denoted $B_T^{\rm bad}$, obeys
\begin{subequations}\label{eq:rev-F-per-design-bad-substitution}
\begin{align}
 B_T^{\rm bad}
 &\le\sum_{t=1}^T\frac{(1+2+8)c_{R}}{\sqrt t}
                         \sP(\mathcal G_t^c)
 \label{eq:rev-F-per-design-bad-substitution-step-1}\\
 &\le11c_{R}\left\{2\sqrt{T_\circ}+3\mathsf A+130\sqrt{n_0}
           +2K56^P+3c_{\rm tail}\Delta_{\min}^{-\gamma_0}\sqrt{n_0}\right\}
 \label{eq:rev-F-per-design-bad-substitution-step-2}\\
 &=22c_{R}\sqrt{T_\circ}+33c_{R}\mathsf A+1430c_{R}\sqrt{n_0}
           +22c_{R}K56^P+33c_{R}c_{\rm tail}\Delta_{\min}^{-\gamma_0}\sqrt{n_0}\,.
 \label{eq:rev-F-per-design-bad-substitution-step-3}
\end{align}
\end{subequations}
Equation \eqref{eq:rev-F-per-design-bad-substitution-step-1} adds the pointwise entropy, proxy function, and exploration coefficients $1,2,8$ and takes expectations of their bad-history indicators. Equation \eqref{eq:rev-F-per-design-bad-substitution-step-2} substitutes the complete five-component failure sum, including the prefix before $T_\circ$. Equation \eqref{eq:rev-F-per-design-bad-substitution-step-3} multiplies every component by $11c_{R}$ explicitly.

Substituting every term of
\eqref{eq:rev-F-per-design-good-substitution} and
\eqref{eq:rev-F-per-design-bad-substitution}, and the actual
corruption bound \eqref{eq:F-direct-cor} in \Cref{lem:AB-card-master}, into \eqref{eq:F-master} in \Cref{lem:AB-card-master}
now yields
\begin{subequations}\label{eq:rev-F-per-design-final}
\begin{align}
 \Reg(u)
 &\le\beta_1\log K+A_T^{\rm good}+E_T^{\rm good}
                     +M_T^{\rm good}+B_T^{\rm bad}
                     +16\sqrt2c_{R}\sqrt C
 \label{eq:rev-F-per-design-final-step-1}\\
 &\le2^{22}Bc^{n}\log(eT)+3\cdot2^{22}BA_0c^{n}\Delta_{\min}^{-\varepsilon_0}
       +\beta_1\log K+S_a+\Psi_{\Delta_{\min}} S_b
 \label{eq:rev-F-per-design-final-step-2}\\
 &\quad+22c_{R}\sqrt{T_\circ}+33c_{R}\mathsf A+1430c_{R}\sqrt{n_0}
       +22c_{R}K56^P+33c_{R}c_{\rm tail}\Delta_{\min}^{-\gamma_0}\sqrt{n_0}
 \label{eq:rev-F-per-design-final-step-3}\\
 &\quad+8c_{R}\Psi_{\Delta_{\min}}(10^6a_n+30000\sqrt{a_{\Delta_{\min}}})
       +16\sqrt2c_{R}\sqrt C\,.
 \label{eq:rev-F-per-design-final-step-4}
\end{align}
\end{subequations}
Equation \eqref{eq:rev-F-per-design-final-step-1} is the comparator inequality with the good, bad, and actual-corruption contributions separated. Equations \eqref{eq:rev-F-per-design-final-step-2}--\eqref{eq:rev-F-per-design-final-step-4} substitute all the component estimates, including both mass constraints and $16\sqrt2c_{R}\sqrt C$. The RHS contains no comparator-dependent quantity.

The RHS of \eqref{eq:rev-F-per-design-final} does not
depend on $u$. Taking $\max_u \Reg(u)$, with this maximum outside
expectation, proves \eqref{eq:F-per-design} for all $T$.
\end{proof}

\subsubsection{Attainment and Measurable Selection}
For every realized history, \IAFTRL{} needs an
optimal solution to its one-round program; the prescribed selection must
choose a unique solution measurably from that history, and the resulting
estimates must be integrable. The next lemma verifies these requirements.
It makes the optimization step a well-defined decision rule and justifies
the subsequent expectation calculations. In particular, it supplies the
measurable optimizer used in the proof of
\Cref{prop:G-compatibility}.

\begin{lemma}[Attainment and measurable sampling-law selection]\label{lem:AB-selection}
On every finite history, the objective in \eqref{eq:G-program}, with
$v=v(p)$ prescribed by $M(p)=\sum_{a\in\cA}p_a aa^\top$ and $v_a(p)=M(p)^{-1}a$, attains its minimum over
$p$. The lexicographically least minimizing $p_t$, in the fixed action
ordering, is uniquely defined and Borel measurable in the history.
For every fixed finite horizon, its covariance estimates and
action-loss differences are integrable.
\end{lemma}
\begin{proof}[Proof of \Cref{lem:AB-selection}]
\emph{Attainment and a fixed tie rule.}
Fix $t$ and the cumulative sum $\sum_{s=1}^{t-1}\widehat\theta_s$. Eliminating the existential mixture variables
leaves the compact set
\[
 \mathcal P_t=\{p\in\gP(\cA):p_b\ge(1-\Gamma_t)q_{t,b}
                                      \text{ for every }b\}\,.
\]
Indeed, each such $p$ admits the fixed representation
$\gamma=\Gamma_t$, $w=p-(1-\Gamma_t)q_t$; conversely, every allowed
mixture has this domination. Since $q_{t,b}>0$ and the actions span,
\[
 M(p)\succeq(1-\Gamma_t)M(q_t)\succ0\qquad(p\in\mathcal P_t)\,.
\]
Thus $p\mapsto M(p)^{-1}$ is continuous and uniformly bounded on
$\mathcal P_t$, as are the prescribed coefficients
$v_b(p)=M(p)^{-1}b$. After using the explicit definition
$v_b(p)=M(p)^{-1}b$ with $M(p)=\sum_{a\in\cA}p_a aa^\top$, the remaining estimator constraints
\eqref{eq:G-program-mag}--\eqref{eq:G-program-cert} are continuous
inequalities in $p$. Their intersection $\mathcal F_t\subseteq\mathcal P_t$
is compact and nonempty by \Cref{lem:AB-fallback}.
The continuous linear objective therefore attains its minimum.

Let $\mathcal F_t^{(0)}$ be its compact set of minimizers, and define
successively along the fixed action ordering $a^{(1)},\ldots,a^{(K)}$, for $j=1,\ldots,K$,
\begin{equation}\label{eq:rev-B-measurable-selection}
 m_t^{(j)}=\min_{p\in\mathcal F_t^{(j-1)}}p_{a^{(j)}}\,,\qquad
 \mathcal F_t^{(j)}=\{p\in\mathcal F_t^{(j-1)}:p_{a^{(j)}}=m_t^{(j)}\}\,.
\end{equation}
Every set is nonempty and compact, and $\mathcal F_t^{(K)}$ contains
exactly one vector. This defines the fixed lexicographic tie rule.
No convexity or minimum-norm uniqueness assertion is needed.

\emph{Measurability.}
The fixed-order reference rule is Borel in $\sum_{s=1}^{t-1}\widehat\theta_s$, and the Gibbs
weights and all prescribed estimator coefficients are Borel in the
state and continuous in $p$ on their domain. Hence the feasible graph
is Borel with compact sections. For each real $c$, its sections with
objective strictly below $c$ are relatively open subsets of compact
metric spaces, hence $\sigma$-compact. The Borel projection theorem
for $\sigma$-compact sections \cite{Larman72} makes the optimal
value Borel. Its minimizer graph is consequently Borel with compact
sections. Repeating the same argument with the continuous coordinate
objectives $p_{a^{(j)}}$ proves that every $m_t^{(j)}$ and every graph in
\eqref{eq:rev-B-measurable-selection} is Borel. The final singleton
graph defines a Borel map $p_t$. The formulas for $v(p_t)$,
$w_t=p_t-(1-\Gamma_t)q_t$, and $\gamma_t=\Gamma_t$ are Borel as well.
Starting from the empty cumulative sum, the update gives Borel dependence on the observed
history by induction.

\emph{Integrability.}
Let $\mu_K=K^{-1}\sum_b\delta_{b}$ and
$\lambda_{\cA}=\lambda_{\min}(M(\mu_K))>0$.
If $\|\sum_{s=1}^{t-1}\widehat\theta_s\|\le R$, bounded actions give
$q_{t,b}\ge K^{-1}e^{-2R/\beta_t}$. Consequently, uniformly over
$p\in\mathcal P_t$,
\[
 M(p)\succeq\tfrac34e^{-2R/\beta_t}M(\mu_K)\,,
\]
and therefore
\begin{align}
 \max_b\|v_b(p)\|
 &=\max_b\|M(p)^{-1}b\|
 \le\frac{4}{3\lambda_{\cA}}e^{2R/\beta_t}=:Q_t(R)\,,
 \label{eq:AB-selection-selected-bound}\\
 \|\widehat\theta_t\|&\le Q_t(R)\,,
 \label{eq:AB-selection-update-bound}
\end{align}
since $\|b\|\le1$ and $|Y_t|\le1$.
Define $R_0^{\rm det}=0$ and
\[
 R_t^{\rm det}=R_{t-1}^{\rm det}+Q_t(R_{t-1}^{\rm det})\,.
\]
Induction gives $\|\sum_{s=1}^{t}\widehat\theta_s\|\le R_t^{\rm det}<\infty$ and a deterministic
bound on every fixed-time covariance estimate. These bounds establish
integrability only; they do not enter the regret constants.

For an explicit bound on the algorithm's action-loss differences, fix any action $c$ and set
$X_{t,a}=\widehat L_{t-1,a}-\widehat L_{t-1,c}$,
$R_t^{\mathrm{abs}}=\max_a|X_{t,a}|$, and $M_t=-\min_aX_{t,a}$.
Since $X_{t,c}=0$, we have $0\le M_t\le R_t^{\mathrm{abs}}$ and
$Z_{t,a}+Z_{t,c}=X_{t,a}+2M_t\le3R_t^{\mathrm{abs}}$. The two update magnitude constraints
therefore bound each difference update by $2\beta_t+3R_t^{\mathrm{abs}}/B$, giving
\begin{equation}\label{eq:AB-selection-difference-integrability}
 R_{t+1}^{\mathrm{abs}}\le(1+3/B)R_t^{\mathrm{abs}}+2\beta_t\,,\qquad
 R_t^{\mathrm{abs}}\le2\sum_{s=1}^{t-1}\beta_s(1+3/B)^{t-1-s}<\infty\,.
\end{equation}
This also bounds every fixed-time action-loss difference and its update,
directly from the update magnitude constraints.
\end{proof}

\subsubsection{The All-History Adversarial Guarantee}
The all-history stability bound and the cap on added exploration already
give the adversarial guarantee, without requiring accurate gap estimates.
The next proposition records this part of the proposition for a fixed allocation.
It ensures that the later instance-dependent analysis applies to a learner
that remains protected on arbitrary loss sequences.

\begin{proposition}[The all-history adversarial guarantee]\label{prop:AB-adversarial}
\IAFTRL{} satisfies, for every $T\ge1$,
\begin{equation*}
 \sup_{u\in\cA}\Reg(u)
 \le\min\{2T,16d\log(eK)+21\sqrt{dT\log(eK)}\}\,. \notag
\end{equation*}
No moment event or objective comparison is required for this bound.
\end{proposition}
\begin{proof}[Proof of \Cref{prop:AB-adversarial}]
Fix a deterministic comparator action $u$ and define
$m_{t,a}=a^\top\theta_t\in[-1,1]$. The change from the FTRL law to
the sampling law costs at most
\begin{align}
 p_{t,a}-q_{t,a}&=(1-\gamma_t)q_{t,a}+w_{t,a}-q_{t,a}
                         =w_{t,a}-\gamma_tq_{t,a}\,,
 \label{eq:AB-adv-mixture-difference}\\
 \langle p_t-q_t,m_t\rangle
 &=\sum_aw_{t,a}m_{t,a}-\gamma_t\sum_aq_{t,a}m_{t,a}\,,
 \label{eq:AB-adv-exploration-substitute}\\
 &\le\left(\sum_aw_{t,a}\right)\max_a m_{t,a}
             -\gamma_t\left(\sum_aq_{t,a}\right)\min_a m_{t,a}
 =\gamma_t(\max_a m_{t,a}-\min_a m_{t,a})\,,
 \label{eq:AB-adv-exploration-oscillation}\\
 &\le2\gamma_t\le2\Gamma_t=8d/\beta_t\,.
 \label{eq:AB-adv-exploration-bound}
\end{align}
Equation \eqref{eq:AB-adv-mixture-difference} substitutes the sampling law;
\eqref{eq:AB-adv-exploration-substitute} expands its loss.
Equation \eqref{eq:AB-adv-exploration-oscillation} uses $w_{t,a}\ge0$,
$\sum_aw_{t,a}=\gamma_t$, and $\sum_aq_{t,a}=1$.
Equation \eqref{eq:AB-adv-exploration-bound} uses $|m_{t,a}|\le1$ and
the exploration cap. No division by $\gamma_t$ is made, so the argument
also covers $\gamma_t=0$.

Unbiasedness of the differences, with the predictable weights
$q_{t,a}-(e_u)_a$, gives
\begin{align}
 \E[\langle q_t-e_u,g_t\rangle]
 &=\E\left[\sum_a(q_{t,a}-(e_u)_a)\E_{t-1}[g_{t,a}]\right]\,,
 \label{eq:AB-adv-tower-predictable}\\
 &=\E\left[\sum_a(q_{t,a}-(e_u)_a)(m_{t,a}-m_{t,r_t})\right]
 =\E[\langle q_t-e_u,m_t\rangle]\,,
 \label{eq:AB-adv-reference-cancel}\\
 \Reg(u)&=\sum_{t=1}^T\E[\langle p_t-e_u,m_t\rangle]
 =\E\left[\sum_{t=1}^T\langle q_t-e_u,g_t\rangle\right]
       +\sum_{t=1}^T\E[\langle p_t-q_t,m_t\rangle]\,,
 \label{eq:AB-adv-regret-decomposition}\\
 &\le\beta_T\log K+\sum_{t=1}^T\E [D_t]
                    +2\sum_{t=1}^T\Gamma_t\,,
 \label{eq:AB-adv-ftrl}\\
 &\le\beta_T\log K+\sum_{t=1}^T\frac{2d}{\beta_t}
                       +2\sum_{t=1}^T\frac{4d}{\beta_t}
 =\beta_T\log K+10d\sum_{t=1}^T\beta_t^{-1}\,.
 \label{eq:AB-adv-budgets}
\end{align}
Equation \eqref{eq:AB-adv-tower-predictable} is the tower property with the
specified $\cF_{t-1}$-measurable weights.
Equation \eqref{eq:AB-adv-reference-cancel} uses
\eqref{eq:AB-est-mean-reconstruct} in \Cref{lem:AB-estimator}
and $\sum_a(q_{t,a}-(e_u)_a)=0$.
Equation \eqref{eq:AB-adv-regret-decomposition} inserts $q_t$ between $p_t$
and $e_u$. Equation \eqref{eq:AB-adv-ftrl} applies
\eqref{eq:entropy-coarse} in \Cref{lem:entropy-FTRL} and
\eqref{eq:AB-adv-exploration-bound}; \eqref{eq:AB-adv-budgets} substitutes
\eqref{eq:rev-B-actual-stability} in \Cref{lem:B-centered-stability}
and $\Gamma_t=4d/\beta_t$.

The tuning and the scalar sum used in \eqref{eq:AB-adv-budgets} satisfy
\begin{align}
 \beta_T\log K&=(16d+c_{\beta}\sqrt T)\log K
 \le(16d+\sqrt{d/\log(eK)}\sqrt T)\log(eK)
 =16d\log(eK)+\sqrt{dT\log(eK)}\,,
 \label{eq:AB-adv-temperature}\\
 \beta_t^{-1}&=(16d+c_{\beta}\sqrt t)^{-1}
 \le(c_{\beta}\sqrt t)^{-1}=\sqrt{\log(eK)/d}\,t^{-1/2}\,,
 \label{eq:AB-adv-temperature-inverse}\\
 \sum_{t=1}^Tt^{-1/2}
 &\le\sum_{t=1}^T\int_{t-1}^t x^{-1/2}\,dx
 =\int_0^T x^{-1/2}\,dx=2\sqrt T\,,
 \label{eq:AB-adv-root-sum}\\
 \Reg(u)&\le(16d\log(eK)+\sqrt{dT\log(eK)})
                +10d\sum_{t=1}^T\sqrt{\log(eK)/d}\,t^{-1/2}\,,
 \label{eq:AB-adv-final-substitution}\\
 &\le16d\log(eK)+\sqrt{dT\log(eK)}+10\sqrt{d\log(eK)}\,(2\sqrt T)
 =16d\log(eK)+21\sqrt{dT\log(eK)}\,.
 \label{eq:rev-B-adversarial-calculation}
\end{align}
Equation \eqref{eq:AB-adv-temperature} uses $c_{\beta}=\sqrt{d/\log(eK)}$ and
$\log K\le \log(eK)$; \eqref{eq:AB-adv-temperature-inverse} drops $16d>0$ from
the denominator. Equation \eqref{eq:AB-adv-root-sum} integrates the
decreasing function on each interval $[t-1,t]$.
Equation \eqref{eq:AB-adv-final-substitution} substitutes both bounds on the regularization coefficient into \eqref{eq:AB-adv-budgets};
\eqref{eq:rev-B-adversarial-calculation} then substitutes the complete sum.
Finally,
\begin{align}
 \Reg(u)&=\sum_{t=1}^T\E[(A_t-u)^\top\theta_t]
 \le\sum_{t=1}^T\E[|A_t^\top\theta_t|+|u^\top\theta_t|]
 \le\sum_{t=1}^T2=2T\,,
 \label{eq:AB-adv-linear-cap}\\
 \sup_{u\in\cA}\Reg(u)
 &\le\sup_{u\in\cA}\min\{2T,16d\log(eK)+21\sqrt{dT\log(eK)}\}
 \notag\\
 &=\min\{2T,16d\log(eK)+21\sqrt{dT\log(eK)}\}\,.
 \label{eq:AB-adv-supremum}
\end{align}
Equation \eqref{eq:AB-adv-linear-cap} uses the bounded means of both
physical actions; \eqref{eq:AB-adv-supremum} takes the supremum over the
same fixed comparators in the two bounds.
\end{proof}

\subsection{Proof of Penalty Term}\label{app:B-penalty}
\begin{lemma}[Penalty contribution]\label{lem:B-penalty-term}
Let $\mathcal E_T=\sum_{t=1}^Tt^{-1/2}\sP(\mathcal G_t^c)$ and
$S_{\rm pen}=48c_{\beta}KC_{\beta}^4$. The penalty sum satisfies
\[
 \mathsf I_1\coloneq \beta_1\log K+
 \sum_{t=2}^T(\beta_t-\beta_{t-1})\E[H(q_t)]
 \le\beta_1\log K+S_{\rm pen}+c_{R}\mathcal E_T\,.
\]
\end{lemma}
\begin{proof}
\interdisplaylinepenalty=10000
The event in \Cref{def:AB-card-good} is predictable and
empty before $T_\circ$. \Cref{lem:F-good-stability} proves
$H(q_t)\le2Ke^{-2t^{\alpha-1/2}/C_{\beta}}$ on this event and
$(\beta_t-\beta_{t-1})H(q_t)\le c_{R}/\sqrt t$ on its complement.
Therefore
\begin{align}
 \mathsf I_1
 &=\beta_1\log K+\sum_{t=2}^T(\beta_t-\beta_{t-1})\E[H(q_t)]
       \label{eq:B-penalty-definition}\\
 &=\beta_1\log K
   +\sum_{t=2}^T(\beta_t-\beta_{t-1})\E[\ind_{\mathcal G_t}H(q_t)]
   +\sum_{t=2}^T(\beta_t-\beta_{t-1})\E[\ind_{\mathcal G_t^c}H(q_t)]
       \label{eq:B-penalty-split}\\
 &\le\beta_1\log K+2c_{\beta}K\sum_{t=2}^Tt^{-1/2}e^{-2t^{\alpha-1/2}/C_{\beta}}
       +c_{R}\sum_{t=2}^Tt^{-1/2}\sP(\mathcal G_t^c)
       \label{eq:B-penalty-envelope}\\
 &\le\beta_1\log K+2c_{\beta}K\sum_{t\ge1}e^{-t^{1/4}/C_{\beta}}+c_{R}\mathcal E_T
 \le\beta_1\log K+48c_{\beta}KC_{\beta}^4+c_{R}\mathcal E_T\,.
       \label{eq:B-penalty-final}
\end{align}
Equation \eqref{eq:B-penalty-split} partitions each expectation.
Equation \eqref{eq:B-penalty-envelope} uses the two bounds just stated and
$\beta_t-\beta_{t-1}\le c_{\beta}/\sqrt t$.
Equation \eqref{eq:B-penalty-final} uses $\alpha\ge3/4$, $t^{-1/2}\le1$,
and the evaluated integral \eqref{eq:rev-F-exponential-sums-step-1}.
This includes all early histories through $\mathcal E_T$.
\end{proof}

\subsection{Proof of Stability Term}\label{app:B-stability}
\begin{lemma}[Stability contribution]\label{lem:B-stability-term}
For the one-step FTRL stability $\mathsf D_t$ in
\Cref{lem:B-scalar-stability}, set $S_{\rm stab}=24e^2KC_{\beta}^7$.
Then
\[
 \mathsf I_2\coloneq \sum_{t=1}^T\E[\mathsf D_t]
 \le S_{\rm stab}+2c_{R}\mathcal E_T\,.
\]
\end{lemma}
\begin{proof}
\interdisplaylinepenalty=10000
The conditional proxy function bound in \Cref{lem:B-scalar-stability}
applies to the selected reconstruction on every history.
\Cref{lem:F-good-stability} proves its good-event envelope
by explicitly applying the centering inequality to the reconstruction
rows in \eqref{eq:rev-F-good-actual-certificate}. Substitution gives
\begin{align}
 \mathsf I_2
 &=\sum_{t=1}^T\E[\E_{t-1}[\mathsf D_t]]
 \le\sum_{t=1}^T\E[\gC_t(p_t)]
       \label{eq:B-stability-conditional}\\
 &=\sum_{t=1}^T\E[\ind_{\mathcal G_t}\gC_t(p_t)]
   +\sum_{t=1}^T\E[\ind_{\mathcal G_t^c}\gC_t(p_t)]
       \label{eq:B-stability-split}\\
 &\le e^2C_{\beta}K\sum_{t=1}^T\sqrt t\,e^{-2t^{\alpha-1/2}/C_{\beta}}
       +2c_{R}\sum_{t=1}^Tt^{-1/2}\sP(\mathcal G_t^c)
       \label{eq:B-stability-envelope}\\
 &\le e^2C_{\beta}^3K\sum_{t\ge1}e^{-t^{1/4}/C_{\beta}}+2c_{R}\mathcal E_T
 \le24e^2KC_{\beta}^7+2c_{R}\mathcal E_T\,.
       \label{eq:B-stability-final}
\end{align}
Equation \eqref{eq:B-stability-conditional} uses the tower property and
$\E_{t-1}[\mathsf D_t]\le\gC_t(p_t)$.
Equation \eqref{eq:B-stability-split} partitions the proxy function expectation;
\eqref{eq:B-stability-envelope} uses its good-event bound and
$\gC_t(p_t)\le2d/\beta_t\le2c_{R}/\sqrt t$ elsewhere.
Equation \eqref{eq:B-stability-final} applies both evaluated bounds in
\eqref{eq:rev-F-exponential-sums}.
\end{proof}

\subsubsection{Conditional Stability of the Executed Reconstruction}
The following two lemmas convert the program's centered proxy function into
an upper bound on the actual FTRL mixability gap. The first uses physical
reconstruction vectors; the second gives the row form used in the scalar
analysis. Both estimates belong to the stability contribution.

\begin{lemma}[Executed reconstruction and centered stability]\label{lem:B-centered-stability}
For \IAFTRL{}, set
$g_{t,a}=(a-r_t)^\top v_{t,A_t}Y_t$ and let $D_t$ be the entropy-FTRL
mixability gap. Then
\[
 \E_{t-1}[D_t]\le\gC_t(p_t)\le2d/\beta_t\,.
\]
\end{lemma}
\begin{proof}
All of $q_t,p_t,r_t,v_t$ are $\mathcal F_{t-1}$-measurable.
To identify the quantity used by entropy stability, define
\begin{align}
 \bar a_t&=\sum_aq_{t,a}a\,,\qquad
 u_{t,a,b}=(a-\bar a_t)^\top v_{t,b}\,,
 \label{eq:AB-est-centered-response}\\
 \sum_aq_{t,a}u_{t,a,b}
 &=(\sum_aq_{t,a}a-\bar a_t\sum_aq_{t,a})^\top v_{t,b}=0\,,
 \label{eq:AB-est-centered-zero}\\
 g_{t,a}-\langle q_t,g_t\rangle
 &=Y_t(a-r_t-\sum_cq_{t,c}(c-r_t))^\top v_{t,b}
 =Y_tu_{t,a,b}\quad(A_t=b)\,.
 \label{eq:AB-est-centered-update}
\end{align}
Equation \eqref{eq:AB-est-centered-response} defines the response to feedback
for the executed reconstruction; its covariance special case is
\eqref{eq:AB-fallback-response} in \Cref{lem:AB-fallback}.
Equations \eqref{eq:AB-est-centered-zero} and
\eqref{eq:AB-est-centered-update} use $\sum_aq_{t,a}=1$ to remove the
common reference. This centering changes neither the estimated cumulative
loss differences nor the executed FTRL distribution.

Let $D_t$ have the definition in \Cref{lem:entropy-FTRL}.
Conditional on $A_t=b$, its complete comparison with the proxy function is
\begin{align}
 D_t&=\langle q_t,g_t\rangle
           +\beta_t\log\sum_aq_{t,a}e^{-g_{t,a}/\beta_t}
 =\beta_t\log\sum_aq_{t,a}e^{-Y_tu_{t,a,b}/\beta_t}\,,
 \label{eq:AB-stability-center}\\
 &\le\beta_t\left(\sum_aq_{t,a}e^{-Y_tu_{t,a,b}/\beta_t}-1\right)
 =\beta_t\sum_aq_{t,a}
       \{e^{-Y_tu_{t,a,b}/\beta_t}+Y_tu_{t,a,b}/\beta_t-1\}\,,
 \label{eq:AB-stability-log}\\
 &\le\beta_t\sum_aq_{t,a}\chi(-Y_tu_{t,a,b}/\beta_t)
 =\beta_t\sum_aq_{t,a}\chi(Y_tu_{t,a,b}/\beta_t)\,,
 \label{eq:AB-stability-chi}\\
 &\le\beta_t\sum_aq_{t,a}\chi(u_{t,a,b}/\beta_t)\,,
 \label{eq:AB-stability-amplitude}\\
 \E_{t-1}[D_t]
 &=\sum_bp_{t,b}\E[D_t\mid\cF_{t-1},A_t=b]
 \le\sum_bp_{t,b}\beta_t\sum_aq_{t,a}\chi(u_{t,a,b}/\beta_t)\,,
 \label{eq:AB-stability-tower}\\
 &=\beta_t\sum_bp_{t,b}\sum_aq_{t,a}
       \chi\!\left(\frac{(a-\bar a_t)^\top v_{t,b}}{\beta_t}\right)
 =\gC_t(p_t)\le2d/\beta_t\,.
 \label{eq:rev-B-actual-stability}
\end{align}
Equation \eqref{eq:AB-stability-center} substitutes
\eqref{eq:AB-est-centered-update}; \eqref{eq:AB-stability-log} uses
$\log x\le x-1$ and \eqref{eq:AB-est-centered-zero}.
Equation \eqref{eq:AB-stability-chi} applies
$e^x-x-1\le\chi(x)$ and evenness from \Cref{lem:chi}.
Equation \eqref{eq:AB-stability-amplitude} uses $|Y_t|\le1$ and monotonicity
of $\chi$ in absolute value. Equation \eqref{eq:AB-stability-tower}
averages with the executed probabilities $p_{t,b}$, and
\eqref{eq:rev-B-actual-stability} substitutes $u_{t,a,b}$ and applies
\eqref{eq:G-program-cert}.
\end{proof}

\begin{lemma}[Conditional stability in scalar row notation]\label{lem:B-scalar-stability}
Consider the scalar updates \eqref{eq:F-update} of
\IAFTRL{}. They satisfy \eqref{eq:F-program-dom}--\eqref{eq:F-program-cert}.
If $D_t$ is the FTRL mixability gap in \Cref{lem:entropy-FTRL}, then
\begin{equation}\label{eq:F-stability}
 \E_{t-1}[D_t]\le\gC_t(p_t)\le2d/\beta_t\,.
\end{equation}
\end{lemma}
\begin{proof}
Condition on $\mathcal F_{t-1}$ and abbreviate
$p_b=p_{t,b}$, $h_{a,b}=h_{t,a,b}$, $q_a=q_{t,a}$, and $\beta=\beta_t$.
For each observed column $b$, define
$\bar h_b=\sum_aq_a h_{a,b}$ and $z_{a,b}=(h_{a,b}-\bar h_b)/p_b$, so that
$\sum_aq_a z_{a,b}=0$. Centering the mixability gap, then using
$\log x\le x-1$ and \Cref{lem:chi}, yields
\begin{subequations}\label{eq:rev-F-stability-calculation}
\begin{align}
 D_t\bigm|_{A_t=b}
 &=\beta\log\sum_aq_a
       \exp\!\left(-\frac{Y_tz_{a,b}}\beta\right)
 \label{eq:rev-F-stability-calculation-step-1}\\
 &\le\beta\sum_aq_a
       \left\{\exp\!\left(-\frac{Y_tz_{a,b}}\beta\right)
                         +\frac{Y_tz_{a,b}}\beta-1\right\}
 \label{eq:rev-F-stability-calculation-step-2}\\
 &\le\beta\sum_aq_a\chi(z_{a,b}/\beta)\,,
 \label{eq:rev-F-stability-calculation-step-3}\\
 \E_{t-1}[D_t]
 &=\sum_bp_b\E[D_t\mid\mathcal F_{t-1},A_t=b]
 \label{eq:rev-F-stability-calculation-step-4}\\
 &\le\beta\sum_bp_b\sum_aq_a
          \chi\!\left(\frac{h_{a,b}-\bar h_b}{\beta p_b}\right)
 =\gC_t(p_t)\le\frac{2d}{\beta_t}\,.
 \label{eq:rev-F-stability-calculation-step-5}
\end{align}
\end{subequations}
Equation \eqref{eq:rev-F-stability-calculation-step-1} subtracts the $q_t$-mean of the realized estimate inside the logarithm. Equation \eqref{eq:rev-F-stability-calculation-step-2} uses $\log x\le x-1$ and $\sum_aq_a z_{a,b}=0$. Equation \eqref{eq:rev-F-stability-calculation-step-3} uses $e^{-x}+x-1\le\chi(x)$, $|Y_t|\le1$, and monotonicity of $\chi$ in $|x|$. Equation \eqref{eq:rev-F-stability-calculation-step-4} is the tower property with the actual probabilities $p_b$. Equation \eqref{eq:rev-F-stability-calculation-step-5} substitutes $z_{a,b}=(h_{a,b}-\bar h_b)/p_b$, identifies the proxy function, and uses its program constraint.

Thus the averaging over the observed column in
\eqref{eq:rev-F-stability-calculation} is with its actual sampling
probability $p_b$; no unweighted bound on $|g_{t,a}|/\beta_t$ is used.
\end{proof}

\subsection{Proof of Exploration Term}\label{app:B-exploration}
\begin{lemma}[Exploration contribution]\label{lem:B-exploration-term}
Fix a finite feasible sampling allocation $n$. Define
\[
 a_n=\max\{1,4BC_{\beta}N_n/d\}\,,\quad a_{\Delta_{\min}}=64B\Delta_{\min}^{-2}\,,\quad
 S_b=19200KC_{\beta}^4\,,
\]
\[
 M_{3,n}=3\cdot2^{22}BA_0c^{n}\Delta_{\min}^{-\varepsilon_0}
       +\Psi_{\Delta_{\min}} S_b+8c_{R}\Psi_{\Delta_{\min}}(10^6a_n+30000\sqrt{a_{\Delta_{\min}}})\,.
\]
Then
\[
 \mathsf I_3\coloneq \sum_{t=1}^T\E[\langle p_t-q_t,\Delta\rangle]
 \le2^{22}Bc^{n}\log(eT)+M_{3,n}+8c_{R}\mathcal E_T\,.
\]
\end{lemma}
\begin{proof}
\interdisplaylinepenalty=10000
Feasibility of the comparison distribution constructed from the sampling allocation $n$
and the resulting objective comparison are proved below in
\Cref{lem:AB-physical-witness,lem:F-exploration}.
Set $L_5=16\cdot25^5$ and
$m_t=\min\{400,L_5a_n^2/t\}+\min\{400,L_5a_{\Delta_{\min}}^2/t^2\}$.
Since $\mathcal G_t=\varnothing$ for $t<T_\circ$, their bounds imply
\begin{align}
 \mathsf I_3
 &=\sum_{t=1}^T\E[\ind_{\mathcal G_t}\langle p_t-q_t,\Delta\rangle]
   +\sum_{t=1}^T\E[\ind_{\mathcal G_t^c}\langle p_t-q_t,\Delta\rangle]
       \label{eq:B-exploration-split}\\
 &\le\sum_{t=T_\circ}^T
   \left\{2^{22}Bc^{n}\Psi_t/t+800K\Psi_t e^{-4t^{\alpha-1/2}/C_{\beta}}
                        +8c_{R}\Psi_t t^{-1/2}m_t\right\}
        +8c_{R}\mathcal E_T
       \label{eq:B-exploration-envelope}\\
 &\le2^{22}Bc^{n}\left\{\log(eT)
               +\sum_{t\ge T_\circ}(\Psi_t-1)/t\right\}
       +800K\Psi_{\Delta_{\min}}\sum_{t\ge1}e^{-t^{1/4}/C_{\beta}}
       +8c_{R}\Psi_{\Delta_{\min}}\sum_{t\ge1}t^{-1/2}m_t+8c_{R}\mathcal E_T
       \label{eq:B-exploration-sums}\\
 &\le2^{22}Bc^{n}\log(eT)+3\cdot2^{22}BA_0c^{n}\Delta_{\min}^{-\varepsilon_0}
       +\Psi_{\Delta_{\min}} S_b
       \notag\\*
 &\quad+8c_{R}\Psi_{\Delta_{\min}}(10^6a_n+30000\sqrt{a_{\Delta_{\min}}})+8c_{R}\mathcal E_T
       \label{eq:B-exploration-evaluated}\\
 &=2^{22}Bc^{n}\log(eT)+M_{3,n}+8c_{R}\mathcal E_T\,.
       \label{eq:B-exploration-final}
\end{align}
Equation \eqref{eq:B-exploration-split} partitions the defining sum.
Equation \eqref{eq:B-exploration-envelope} uses
\Cref{lem:F-exploration} on both events.
Equation \eqref{eq:B-exploration-sums} writes $\Psi_t=1+(\Psi_t-1)$,
uses $\sum_{t\le T}t^{-1}\le\log(eT)$, $\Psi_t\le\Psi_{\Delta_{\min}}$, and
$\alpha-1/2\ge1/4$. Equation \eqref{eq:B-exploration-evaluated} applies
\eqref{eq:F-positive-local} in \Cref{lem:B-positive-local},
\eqref{eq:rev-F-exponential-sums-step-1} in
\Cref{lem:AB-good-sums}, and
\eqref{eq:F-mass-sum1}--\eqref{eq:F-mass-sum2} in
\Cref{lem:B-mass-sums}.
Equation \eqref{eq:B-exploration-final} substitutes the displayed
constant, retaining both failed-mass contributions.
\end{proof}

\subsubsection{The Comparison Distribution Constructed from the Sampling Allocation \texorpdfstring{$n$}{n} and Its Objective Cost}
\label{sec:G-witness}
A low-regret exploration distribution is useful for comparison only if
its prescribed covariance estimator satisfies every constraint of the
learner's program. The next lemma verifies this for the scaled sampling allocation,
including the exponential stability constraint. Its feasibility justifies
the objective comparison used in the exploration bound.

\begin{lemma}[One low-cost physical reconstruction satisfies all budgets]\label{lem:AB-physical-witness}
Let $n$ be a finite feasible sampling allocation. Define $a_n,a_{\Delta_{\min}},\rho_t$,
$\mathcal W_t$, and $(p_t^c,w_t^c)$ by
\eqref{eq:rev-B-witness-definition}. On $\mathcal W_t$ with
$t\ge T_\circ$, the sampling law $p_t^c$ with its prescribed
$v_b^c=v_b(p_t^c)=\Sigma(p_t^c)^{-1}b$ satisfies every constraint
of \eqref{eq:G-program} simultaneously.
Its reference-difference rows satisfy \eqref{eq:G-witness-rows}, and
its proxy function satisfies \eqref{eq:rev-B-witness-certificate-final}.
\end{lemma}
\begin{proof}[Proof of \Cref{lem:AB-physical-witness}]
Fix a raw feasible sampling allocation $n$, independently of observations
and time. Set
\begin{subequations}\label{eq:rev-B-witness-definition}
\begin{align}
 c^{n}&=\sum_{a\ne a^*}n_a\Delta(a)\,,\qquad N_n=\sum_a n_a\,,
 \label{eq:rev-B-witness-definition-step-1}\\
 a_n&=\max\{1,4BC_{\beta}N_n/d\}\,,\qquad a_{\Delta_{\min}}=64B\Delta_{\min}^{-2}\,,
 \label{eq:rev-B-witness-definition-step-2}\\
 \rho_t&=16B\lo_t^2/t\,,\qquad
 \mathcal W_t=\mathcal G_t\cap\{\lo_t^2\le\sqrt t/a_n\}
                               \cap\{\lo_t^2\le t/a_{\Delta_{\min}}\}\,,
 \label{eq:rev-B-witness-definition-step-3}\\
 w_{t,a}^c&=\rho_tn_a\ind\{a\ne a^*\}\,,\quad
 \gamma_t^c=\sum_aw_{t,a}^c\,,\quad
 p_t^c=(1-\gamma_t^c)q_t+w_t^c\,.
 \label{eq:rev-B-witness-definition-step-4}
\end{align}
\end{subequations}
Equation \eqref{eq:rev-B-witness-definition-step-1} fixes the regret cost and total mass of the same raw sampling allocation. Equation \eqref{eq:rev-B-witness-definition-step-2} sets the total-mass and leverage thresholds. Equation \eqref{eq:rev-B-witness-definition-step-3} defines the scale $\rho_t$ and the event on which both restrictions hold. Equation \eqref{eq:rev-B-witness-definition-step-4} allocates exploration only to suboptimal actions; the FTRL mass on the optimal action will supply its otherwise free information. None of these analysis choices is executed as a separate policy.

The candidate is used only on $\mathcal W_t$; this event is analytical
and is not tested by the learner. On that event, direct substitution gives
\begin{subequations}\label{eq:rev-B-witness-mass}
\begin{align}
 \rho_tN_n
 &=\frac{16B\lo_t^2N_n}{t}
 \le\frac{16BN_n}{a_n\sqrt t}
 \le\frac{4d}{C_{\beta}\sqrt t}\le\frac{4d}{\beta_t}=\Gamma_t<\frac14\,,
 \label{eq:rev-B-witness-mass-step-1}\\
 \rho_t&=\frac{16B\lo_t^2}{t}
 \le\frac{16B}{a_{\Delta_{\min}}}
 =\frac{\Delta_{\min}^2}{4}\,,\qquad\rho_t\le1\,.
 \label{eq:rev-B-witness-mass-step-2}
\end{align}
\end{subequations}
Equation \eqref{eq:rev-B-witness-mass-step-1} substitutes $\rho_t=16B\lo_t^2/t$, then uses the event restriction $\lo_t^2\le\sqrt t/a_n$, $a_n\ge4BC_{\beta}N_n/d$, and $\beta_t\le C_{\beta}\sqrt t$. Equation \eqref{eq:rev-B-witness-mass-step-2} instead uses $\lo_t^2\le t/a_{\Delta_{\min}}$, substitutes $a_{\Delta_{\min}}=64B\Delta_{\min}^{-2}$, and uses $\Delta_{\min}\le1$.

Here $\alpha=3/4+13/(10P)\ge3/4$ and $\tau_t\ge t^\alpha$.
The time threshold $T_\circ$ includes
$t\ge(32C_{\beta})^8$ and $t\ge(4C_{\beta}\log(eK))^4$. Setting $x=t^{1/8}\ge32C_{\beta}$ gives
\begin{align}
 8\log x&\le x/8\quad(x\ge512)\,,\qquad
 \log t=8\log x\le x/8\le x^2/(4C_{\beta})=t^{1/4}/(4C_{\beta})\,,
 \label{eq:AB-witness-log-threshold}\\
 \log K&\le \log(eK)\le t^{1/4}/(4C_{\beta})\,,\qquad
 \log(Kt)\le t^{1/4}/(2C_{\beta})\,,
 \label{eq:AB-witness-cardinality-threshold}\\
 K e^{-t^{1/4}/(2C_{\beta})}&\le K e^{-\log(Kt)}=1/t\,.
 \label{eq:AB-witness-exponential-threshold}
\end{align}
For \eqref{eq:AB-witness-log-threshold}, the derivative of
$x/8-8\log x$ is $1/8-8/x\ge0$ for $x\ge512$ and its value at
$512$ is positive; also $C_{\beta}\ge16$. Equation
\eqref{eq:AB-witness-cardinality-threshold} uses the second threshold,
and \eqref{eq:AB-witness-exponential-threshold} exponentiates the result.
Since $X_{t,a}\ge4\tau_t$ and $\beta_t\le C_{\beta}\sqrt t$, the exponential-weights
distribution therefore satisfies
\begin{subequations}\label{eq:rev-B-witness-domination}
\begin{align}
 \sum_{a\ne a^*}q_{t,a}
 &\le K e^{-4t^{\alpha-1/2}/C_{\beta}}
 \le K e^{-t^{1/4}/(2C_{\beta})}\le1/t\le1/4\,,
 \label{eq:rev-B-witness-domination-step-1}\\
 q_{t,a^*}&\ge3/4\,,\qquad
 p_{t,a^*}^c\ge(1-\Gamma_t)q_{t,a^*}\ge9/16\ge1/2
                                      \ge\rho_tn_{a^*}\,,
 \label{eq:rev-B-witness-domination-step-2}\\
 p_{t,a}^c&\ge\rho_tn_a\quad(a\ne a^*)\,,\qquad
 M_c\coloneq \sum_bp_{t,b}^cbb^\top\succeq\rho_tH_n\succ0\,.
 \label{eq:rev-B-witness-domination-step-3}
\end{align}
\end{subequations}
Equation \eqref{eq:rev-B-witness-domination-step-1} uses $q_{t,a}\le e^{-X_{t,a}/\beta_t}$, $X_{t,a}\ge4t^\alpha$, and the explicit exponential threshold. Equation \eqref{eq:rev-B-witness-domination-step-2} converts this tail bound into optimal-action mass and uses $\rho_tn_{a^*}\le\rho_tN_n<1/4$. Equation \eqref{eq:rev-B-witness-domination-step-3} uses the nonnegative exploration coordinates and sums $p_{t,b}^cbb^\top\succeq\rho_tn_bbb^\top$ over $b$.

The inequality $\rho_tn_{a^*}\le1/2$ follows already from
\eqref{eq:rev-B-witness-mass}. The prescribed coherent reconstruction is
$v_b^c=v_b(p_t^c)=M_c^{-1}b$. It satisfies
$\sum_bp_{t,b}^cv_b^cb^\top=M_c^{-1}M_c=I_d$.

Let $\lambda_a=a^\top M_c^{-1}a$. For $a\ne a^*$, raw sampling allocation
feasibility gives $\lambda_a\le\Delta(a)^2/(2\rho_t)$.
For $a^*$, $M_c\succeq p_{t,a^*}^ca^*(a^*)^{\top}$ gives
\begin{subequations}\label{eq:rev-B-witness-optimum-leverage}
\begin{align}
 \lambda_{a^*}&=(M_c^{-1}a^*)^\top M_c(M_c^{-1}a^*)
 \ge p_{t,a^*}^c((a^*)^{\top} M_c^{-1}a^*)^2=p_{t,a^*}^c\lambda_{a^*}^2\,,
 \label{eq:rev-B-witness-optimum-leverage-step-1}\\
 \lambda_{a^*}&\le1/p_{t,a^*}^c\le2\,,\qquad
 \max_b\lambda_b\le2/\rho_t\,.
 \label{eq:rev-B-witness-optimum-leverage-step-2}
\end{align}
\end{subequations}
Equation \eqref{eq:rev-B-witness-optimum-leverage-step-1} evaluates $M_c\succeq p_{t,a^*}^ca^*(a^*)^{\top}$ on $M_c^{-1}a^*$. Equation \eqref{eq:rev-B-witness-optimum-leverage-step-2} divides by $p_{t,a^*}^c\lambda_{a^*}$ when $\lambda_{a^*}>0$; the case $\lambda_{a^*}=0$ is immediate. The maximum also uses $\lambda_a\le\Delta(a)^2/(2\rho_t)\le2/\rho_t$ for $a\ne a^*$, and $\rho_t\le1$.

For $\lambda_{a^*}=0$, the bound in \eqref{eq:rev-B-witness-optimum-leverage-step-2}
holds directly; for $\lambda_{a^*}>0$, it follows by dividing
\eqref{eq:rev-B-witness-optimum-leverage-step-1} by $p_{t,a^*}^c\lambda_{a^*}$. Let $b_{a,b}^c=(a-a^*)^\top M_c^{-1}b$.
Then
\begin{subequations}\label{eq:rev-B-witness-raw-rows}
\begin{align}
 \sum_bp_{t,b}^c[(a-a^*)^\top v_b^c]^2
 &=\sum_bp_{t,b}^c(b_{a,b}^c)^2
 =(a-a^*)^\top M_c^{-1}(a-a^*)
 \label{eq:rev-B-witness-raw-rows-step-1}\\
 &\le2\lambda_a+2\lambda_{a^*}
 \le\Delta(a)^2/\rho_t+4
 \le2\Delta(a)^2/\rho_t\,,
 \label{eq:rev-B-witness-raw-rows-step-2}\\
 \max_b|(a-a^*)^\top v_b^c|
 &=\max_b|b_{a,b}^c|
 \le\sqrt{(a-a^*)^\top M_c^{-1}(a-a^*)\max_b\lambda_b}
 \label{eq:rev-B-witness-raw-rows-step-3}\\
 &\le\sqrt{(2\Delta(a)^2/\rho_t)(2/\rho_t)}
 =2\Delta(a)/\rho_t\,.
 \label{eq:rev-B-witness-raw-rows-step-4}
\end{align}
\end{subequations}
Equation \eqref{eq:rev-B-witness-raw-rows-step-1} substitutes $v_b^c=M_c^{-1}b$ and contracts the covariance sum $\sum_bp_{t,b}^cbb^\top=M_c$. Equation \eqref{eq:rev-B-witness-raw-rows-step-2} uses $\|a-a^*\|_{M_c^{-1}}^2\le2\lambda_a+2\lambda_{a^*}$ and $4\le\Delta(a)^2/\rho_t$. Equation \eqref{eq:rev-B-witness-raw-rows-step-3} applies Cauchy--Schwarz in the $M_c^{-1}$ norm; \eqref{eq:rev-B-witness-raw-rows-step-4} inserts the two leverage bounds before taking the square root.

Here $4\le\Delta(a)^2/\rho_t$ follows from
$\rho_t\le\Delta_{\min}^2/4$.
Since $\lo_t\ge t\Delta(a)/X_{t,a}$ and $\lo_t\ge1$, substituting
$\rho_t=16B\lo_t^2/t$ gives the required row bounds
\begin{subequations}\label{eq:G-witness-rows}
\begin{align}
 \sum_bp_{t,b}^c[(a-a^*)^\top v_b^c]^2
 &\le\frac{2\Delta(a)^2}{16B\lo_t^2/t}
 =\frac{t\Delta(a)^2}{8B\lo_t^2}
 \le\frac{X_{t,a}^2}{8Bt}\,,
 \label{eq:G-witness-rows-step-1}\\
 \max_b|(a-a^*)^\top v_b^c|
 &\le\frac{2\Delta(a)}{16B\lo_t^2/t}
 =\frac{t\Delta(a)}{8B\lo_t^2}
 \le\frac{X_{t,a}}{8B\lo_t}\le\frac{X_{t,a}}{8B}\,.
 \label{eq:G-witness-rows-step-2}
\end{align}
\end{subequations}
Equation \eqref{eq:G-witness-rows-step-1} substitutes $\rho_t$ into the raw quadratic bound and then uses $t\Delta(a)\le\lo_tX_{t,a}$. Equation \eqref{eq:G-witness-rows-step-2} substitutes $\rho_t$ into the upper bound $2\Delta(a)/\rho_t$, applies the same ratio inequality, and finally uses $\lo_t\ge1$.

On $\mathcal G_t$, $r_t=a^*$ and $Z_{t,a}=X_{t,a}$, so these
bounds satisfy the variance and update magnitude constraints, whose
upper bounds are $\beta_t+X_{t,a}^2/(Bt)$ and $\beta_t+X_{t,a}/B$,
respectively. The row corresponding to $a^*$
is exactly zero.

For the same reconstruction, apply \eqref{eq:chi-center} in \Cref{lem:chi} for each
fixed column $b$, with $f(a)=b_{a,b}^c/\beta_t$. Then use
$\chi(x)\le x^2e^{|x|}$ and \eqref{eq:rev-B-witness-raw-rows}:
\begin{subequations}\label{eq:rev-B-witness-certificate-substitution}
\begin{align}
 \gC_t(p_t^c)
 &=\beta_t\sum_bp_{t,b}^c\sum_aq_{t,a}
       \chi\!\left(\frac{b_{a,b}^c-\sum_cq_{t,c}b_{c,b}^c}{\beta_t}\right)
 \label{eq:rev-B-witness-certificate-substitution-step-1}\\
 &\le\beta_t\sum_bp_{t,b}^c\sum_{a\ne a^*}q_{t,a}
                          \chi(2b_{a,b}^c/\beta_t)
 \label{eq:rev-B-witness-certificate-substitution-step-2}\\
 &\le\frac4{\beta_t}\sum_{a\ne a^*}q_{t,a}
       e^{2\max_b|b_{a,b}^c|/\beta_t}\sum_bp_{t,b}^c(b_{a,b}^c)^2
 \label{eq:rev-B-witness-certificate-substitution-step-3}\\
 &\le\frac8{\beta_t\rho_t}\sum_{a\ne a^*}q_{t,a}\Delta(a)^2
                          e^{4\Delta(a)/(\beta_t\rho_t)}\,.
 \label{eq:rev-B-witness-certificate-substitution-step-4}
\end{align}
\end{subequations}
Equation \eqref{eq:rev-B-witness-certificate-substitution-step-1} substitutes the centered covariance reconstruction into the proxy function. Equation \eqref{eq:rev-B-witness-certificate-substitution-step-2} applies \eqref{eq:chi-center} in \Cref{lem:chi} for each fixed column; its optimal row is zero. Equation \eqref{eq:rev-B-witness-certificate-substitution-step-3} applies $\chi(x)\le x^2e^{|x|}$ and takes a column maximum only in the exponential. Equation \eqref{eq:rev-B-witness-certificate-substitution-step-4} substitutes both raw row bounds, giving factors $2\Delta(a)/\rho_t$ and $2\Delta(a)^2/\rho_t$.

The exponent can be absorbed in its own exponential weight because
\begin{subequations}\label{eq:rev-B-witness-certificate-final}
\begin{align}
 \frac{4\Delta(a)}{\rho_t}
 &=\frac{t\Delta(a)}{4B\lo_t^2}
 \le\frac{X_{t,a}}{4B\lo_t}\le\frac{X_{t,a}}{4B}\,,
 \label{eq:rev-B-witness-certificate-final-step-1}\\
 q_{t,a}e^{4\Delta(a)/(\beta_t\rho_t)}
 &\le e^{-X_{t,a}/\beta_t+X_{t,a}/(4B\beta_t)}
 \le e^{-X_{t,a}/(2\beta_t)}\,,
 \label{eq:rev-B-witness-certificate-final-step-2}\\
 \gC_t(p_t^c)
 &\le\frac{32K}{\beta_t\rho_t}e^{-\tau_t/(2\beta_t)}
 \le\frac{2Kt}{B\beta_t}e^{-t^{1/4}/(2C_{\beta})}
 \le\frac2{B\beta_t}\le\frac{2d}{\beta_t}\,.
 \label{eq:rev-B-witness-certificate-final-step-3}
\end{align}
\end{subequations}
Equation \eqref{eq:rev-B-witness-certificate-final-step-1} substitutes $\rho_t$ and uses the inverse gap ratio. Equation \eqref{eq:rev-B-witness-certificate-final-step-2} combines this exponent with $q_{t,a}\le e^{-X_{t,a}/\beta_t}$ and $1-1/(4B)\ge1/2$. Equation \eqref{eq:rev-B-witness-certificate-final-step-3} then uses $\Delta(a)^2\le4$, $\rho_t^{-1}\le t/(16B)$, $\tau_t/\beta_t\ge t^{1/4}/C_{\beta}$, and the explicit exponential threshold, ending with $1/B\le d$.

This uses $\Delta(a)\le2$, $\rho_t\ge16B/t$, and
$\tau_t/\beta_t\ge t^{\alpha-1/2}/C_{\beta}\ge t^{1/4}/C_{\beta}$.
Equations \eqref{eq:rev-B-witness-definition}--\eqref{eq:rev-B-witness-certificate-final}
check every constraint for the comparison distribution constructed from
the sampling allocation $n$, together with its prescribed covariance estimator.
\end{proof}

\subsubsection{A Feasible Comparison Distribution Constructed from the Sampling Allocation \texorpdfstring{$n$}{n}}
\begin{definition}[Comparison distribution constructed from the sampling allocation $n$ and its feasibility thresholds]\label{def:AB-card-witness}
Fix any finite feasible sampling allocation $n$ in \eqref{eq:overview-cstar}; let
$c^{n}=\sum_{a\ne a^*}n_a\Delta(a)$ and $N_n=\sum_a n_a$. Set
\begin{equation*}
 a_n=\max\{1,4BC_{\beta}N_n/d\}\,,\qquad a_{\Delta_{\min}}=64B\Delta_{\min}^{-2}\,,
 \qquad\rho_t=16B\lo_t^2/t\,. \notag
\end{equation*}
On $\mathcal G_t$, consider
\[
 w^c_{t,a}=\rho_t n_a\ (a\ne a^*)\,,\quad w^c_{t,a^*}=0\,,\quad
 p_t^c=(1-\gamma_t^c)q_t+w_t^c\,,
\]
and the analysis event
\[
 \mathcal W_t=\mathcal G_t\cap\{\lo_t^2\le\sqrt t/a_n\}
                   \cap\{\lo_t^2\le t/a_{\Delta_{\min}}\}\,.
\]
\end{definition}

The program minimizes estimated regret, but the analysis needs its cost
in true gaps. The next lemma uses the gap-ratio estimates to transfer the
scaled allocation's cost to the actual optimizer. It separates the leading
$Bc^{n}/t$ term from the exponential-weights tail and the two allocation-feasibility failures
that will be summed later.

\begin{lemma}[Expected exploration cost]\label{lem:F-exploration}
For $t\ge T_\circ$,
\begin{align}\label{eq:F-exploration}
 \E[\ind_{\mathcal G_t}\ip{p_t-q_t}{\Delta}]
 \le{}&\frac{2^{22}Bc^{n}\Psi_t}{t}+800K\Psi_t e^{-4t^{\alpha-1/2}/C_{\beta}}\notag\\
 &+\frac{8c_{R}\Psi_t}{\sqrt t}
 \left[\min\{400,16\cdot25^5a_n^2/t\}
      +\min\{400,16\cdot25^5a_{\Delta_{\min}}^2/t^2\}\right]\,.
\end{align}
On every history, $\ip{p_t-q_t}{\Delta}\le8c_{R}/\sqrt t$.
\end{lemma}
\begin{proof}[Proof of \Cref{lem:F-exploration}]
On $\mathcal W_t$, \Cref{lem:F-witness} makes
$(p_t^c,v^c)$ a feasible comparison for the actual optimizer.
The two gap-ratio inequalities and then the definition of $w_t^c$
give
\begin{subequations}\label{eq:rev-F-exploration-witness}
\begin{align}
 \langle p_t,\Delta\rangle
 &\le\lo_t\langle p_t,\widehat\Delta_t\rangle
 \le\lo_t\langle p_t^c,\widehat\Delta_t\rangle
 \label{eq:rev-F-exploration-witness-step-1}\\
 &=\lo_t(1-\gamma_t^c)\langle q_t,\widehat\Delta_t\rangle
     +\lo_t\rho_t\sum_{a\ne a^*}n_a\widehat\Delta_t(a)
 \label{eq:rev-F-exploration-witness-step-2}\\
 &\le\lo_t\up_t\langle q_t,\Delta\rangle
                   +\lo_t\rho_t\up_t\sum_{a\ne a^*}n_a\Delta(a)
 \label{eq:rev-F-exploration-witness-step-3}\\
 &=\lo_t\up_t\langle q_t,\Delta\rangle
                   +\frac{16Bc^{n}}{t}\lo_t^3\up_t\,.
 \label{eq:rev-F-exploration-witness-step-4}
\end{align}
\end{subequations}
Equation \eqref{eq:rev-F-exploration-witness-step-1} uses $\Delta(a)\le\lo_t\widehat\Delta_t(a)$ and objective comparison with the feasible comparison distribution constructed from $n$. Equation \eqref{eq:rev-F-exploration-witness-step-2} substitutes its mixture probabilities exactly. Equation \eqref{eq:rev-F-exploration-witness-step-3} uses $\widehat\Delta_t(a)\le\up_t\Delta(a)$ and $1-\gamma_t^c\le1$. Equation \eqref{eq:rev-F-exploration-witness-step-4} substitutes $c^{n}=\sum_{a\ne a^*}n_a\Delta(a)$ and $\rho_t=16B\lo_t^2/t$.

On $\mathcal G_t\setminus\mathcal W_t$, use the feasible fallback
$p^0=(1-\Gamma_t)q_t+\Gamma_t\pi$ only as a comparison. Its objective
gives instead
\begin{subequations}\label{eq:rev-F-exploration-fallback}
\begin{align}
 \langle p_t,\Delta\rangle
 &\le\lo_t\langle p^0,\widehat\Delta_t\rangle
 \label{eq:rev-F-exploration-fallback-step-1}\\
 &\le\lo_t\up_t
       \{(1-\Gamma_t)\langle q_t,\Delta\rangle
                               +\Gamma_t\langle\pi,\Delta\rangle\}
 \label{eq:rev-F-exploration-fallback-step-2}\\
 &\le\lo_t\up_t\langle q_t,\Delta\rangle
                             +2\Gamma_t\lo_t\up_t\,.
 \label{eq:rev-F-exploration-fallback-step-3}
\end{align}
\end{subequations}
Equation \eqref{eq:rev-F-exploration-fallback-step-1} compares the selected objective with the feasible fallback. Equation \eqref{eq:rev-F-exploration-fallback-step-2} substitutes its mixture and the upper estimated-gap ratio. Equation \eqref{eq:rev-F-exploration-fallback-step-3} uses $\langle\pi,\Delta\rangle\le2$ and $1-\Gamma_t\le1$.

Set $x_n=\sqrt t/a_n$ and $x_{\Delta_{\min}}=t/a_{\Delta_{\min}}$.
Subtracting $\langle q_t,\Delta\rangle\ge0$ from
\eqref{eq:rev-F-exploration-witness} and
\eqref{eq:rev-F-exploration-fallback}, and using
$\mathcal G_t\setminus\mathcal W_t
 \subseteq\{\lo_t^2>x_n\}\cup\{\lo_t^2>x_{\Delta_{\min}}\}$, yields
\begin{subequations}\label{eq:rev-F-exploration-pathwise}
\begin{align}
 \ind_{\mathcal G_t}\langle p_t-q_t,\Delta\rangle
 &\le\ind_{\mathcal G_t}\lo_t\up_t\langle q_t,\Delta\rangle
       +\frac{16Bc^{n}}{t}\ind_{\mathcal G_t}\lo_t^3\up_t
 \label{eq:rev-F-exploration-pathwise-step-1}\\
 &\quad+2\Gamma_t\ind_{\mathcal G_t}\lo_t\up_t
       \{\ind\{\lo_t^2>x_n\}+\ind\{\lo_t^2>x_{\Delta_{\min}}\}\}\,.
 \label{eq:rev-F-exploration-pathwise-step-2}
\end{align}
\end{subequations}
Equations \eqref{eq:rev-F-exploration-pathwise-step-1} and \eqref{eq:rev-F-exploration-pathwise-step-2} combine the comparisons with the distribution constructed from $n$ and with the fallback distribution after subtracting $\langle q_t,\Delta\rangle\ge0$. The union bound $\ind_{\mathcal G_t\setminus\mathcal W_t}\le\ind_{\mathcal G_t}(\ind\{\lo_t^2>x_n\}+\ind\{\lo_t^2>x_{\Delta_{\min}}\})$ accounts for failures of both distinct feasibility thresholds for the comparison distribution.

For each $x>0$, the two alternatives from
\eqref{eq:F-ratio-bounds} in \Cref{lem:F-ratios} are explicitly
\begin{subequations}\label{eq:rev-F-weighted-mass-tail}
\begin{align}
 \E[\ind_{\mathcal G_t}\lo_t\up_t\ind\{\lo_t^2>x\}]
 &\le\E[\ind_{\mathcal G_t}\lo_t\up_t]\le400\Psi_t\,,
 \label{eq:rev-F-weighted-mass-tail-step-1}\\
 \E[\ind_{\mathcal G_t}\lo_t\up_t\ind\{\lo_t^2>x\}]
 &\le x^{-2}\E[\ind_{\mathcal G_t}\lo_t^5\up_t]
 \le16\cdot25^5\Psi_t x^{-2}\,.
 \label{eq:rev-F-weighted-mass-tail-step-2}
\end{align}
\end{subequations}
Equation \eqref{eq:rev-F-weighted-mass-tail-step-1} drops the extra indicator and applies the mixed first-moment ratio bound. Equation \eqref{eq:rev-F-weighted-mass-tail-step-2} uses $\ind\{\lo_t^2>x\}\le\lo_t^4/x^2$ and then the mixed fifth-moment ratio bound. Taking the smaller of these two bounds retains the probability cap.

Taking expectation in \eqref{eq:rev-F-exploration-pathwise},
substituting \eqref{eq:rev-F-weighted-mass-tail} twice, and then
substituting $x_n,x_{\Delta_{\min}}$ gives
\begin{subequations}\label{eq:rev-F-exploration-expectation}
\begin{align}
 \E[\ind_{\mathcal G_t}\langle p_t-q_t,\Delta\rangle]
 &\le2Ke^{-4t^{\alpha-1/2}/C_{\beta}}(400\Psi_t)
       +\frac{16Bc^{n}}{t}(250000\Psi_t)
 \label{eq:rev-F-exploration-expectation-step-1}\\
 &\quad+\frac{8c_{R}\Psi_t}{\sqrt t}
       \{\min(400,16\cdot25^5x_n^{-2})
                  +\min(400,16\cdot25^5x_{\Delta_{\min}}^{-2})\}
 \label{eq:rev-F-exploration-expectation-step-2}\\
 &=800K\Psi_t e^{-4t^{\alpha-1/2}/C_{\beta}}
       +\frac{4000000Bc^{n}\Psi_t}{t}
 \label{eq:rev-F-exploration-expectation-step-3}\\
 &\quad+\frac{8c_{R}\Psi_t}{\sqrt t}
       \{\min(400,16\cdot25^5a_n^2/t)
                  +\min(400,16\cdot25^5a_{\Delta_{\min}}^2/t^2)\}\,.
 \label{eq:rev-F-exploration-expectation-step-4}
\end{align}
\end{subequations}
Equations \eqref{eq:rev-F-exploration-expectation-step-1} and \eqref{eq:rev-F-exploration-expectation-step-2} take expectations of every term of the pathwise exploration bound and insert the two capped ratio estimates at $x_n$ and $x_{\Delta_{\min}}$. Equations \eqref{eq:rev-F-exploration-expectation-step-3} and \eqref{eq:rev-F-exploration-expectation-step-4} multiply $2\cdot400$ and $16\cdot250000$, and substitute $x_n^{-2}=a_n^2/t$ and $x_{\Delta_{\min}}^{-2}=a_{\Delta_{\min}}^2/t^2$ before relaxing the coefficient $4000000$ to $2^{22}$.

Since $4000000<2^{22}$, \eqref{eq:rev-F-exploration-expectation}
proves \eqref{eq:F-exploration}. Finally, on every history,
\begin{align}
 \langle p_t-q_t,\Delta\rangle
 &=\langle w_t,\Delta\rangle-\gamma_t\langle q_t,\Delta\rangle
 \le\sum_a2w_{t,a}=2\gamma_t
 \le2\Gamma_t\le8c_{R}/\sqrt t\,.
 \notag
\end{align}
\end{proof}

The preceding objective comparison requires the proposed distribution
to be feasible for the sampling-law program. We now verify that its
prescribed covariance reconstruction satisfies the mixture, reconstruction,
moment, and stability constraints simultaneously. This supplies the
feasibility step in \Cref{lem:F-exploration}.

\begin{lemma}[Feasibility of the comparison distribution constructed from the sampling allocation $n$]\label{lem:F-witness}
On $\mathcal W_t$ with $t\ge T_\circ$, the above distribution admits a
covariance reconstruction satisfying every constraint of
\eqref{eq:G-program}.
\end{lemma}
\begin{proof}[Proof of \Cref{lem:F-witness}]
Fix a history in $\mathcal W_t$ with $t\ge T_\circ$. Directly
substituting $\rho_t=16B\lo_t^2/t$ and the two additional conditions
in $\mathcal W_t$ gives
\begin{subequations}\label{eq:rev-F-witness-mass}
\begin{align}
 \gamma_t^c&=\rho_t\sum_{a\ne a^*}n_a\le\rho_tN_n
 =\frac{16BN_n\lo_t^2}{t}
 \le\frac{16BN_n}{a_n\sqrt t}
 \le\frac{4d}{C_{\beta}\sqrt t}\,,
 \label{eq:rev-F-witness-mass-step-1}\\
 \rho_t&=\frac{16B\lo_t^2}{t}
 \le\frac{16B}{a_{\Delta_{\min}}}
 =\frac{\Delta_{\min}^2}4\,,
 \qquad \rho_t\le\frac14\le1\,.
 \label{eq:rev-F-witness-mass-step-2}
\end{align}
\end{subequations}
Equation \eqref{eq:rev-F-witness-mass-step-1} substitutes the exploration scale and $\lo_t^2\le\sqrt t/a_n$, then uses $a_n\ge4BC_{\beta}N_n/d$. Equation \eqref{eq:rev-F-witness-mass-step-2} instead uses $\lo_t^2\le t/a_{\Delta_{\min}}$ and the definition $a_{\Delta_{\min}}=64B\Delta_{\min}^{-2}$. These are different constraints and neither is inferred from the other.

Since $t\ge(32C_{\beta})^8$ and $\beta_t\le C_{\beta}\sqrt t$,
$4d/(C_{\beta}\sqrt t)\le4d/\beta_t=\Gamma_t<1/4$.
In particular the mixture constraint holds.

To verify the exponential threshold, let $x=t^{1/8}$. The two
conditions $t\ge(32C_{\beta})^8$ and $t\ge(4C_{\beta}\log(eK))^4$ imply
\begin{subequations}\label{eq:F-exp-threshold}
\begin{align}
 \frac{t^{1/4}}{2C_{\beta}}
 &=\frac{x^2}{4C_{\beta}}+\frac{x^2}{4C_{\beta}}
 \ge8x+\log(eK)\ge8\log x+\log K
 =\log t+\log K\,,
 \label{eq:F-exp-threshold-step-1}\\
 e^{-t^{1/4}/(2C_{\beta})}&\le e^{-\log t-\log K}=\frac1{Kt}\,.
 \label{eq:F-exp-threshold-step-2}
\end{align}
\end{subequations}
Equation \eqref{eq:F-exp-threshold-step-1} writes $t^{1/4}/(2C_{\beta})$ as two equal pieces. The inequalities use $x=t^{1/8}\ge32C_{\beta}$, $t^{1/4}\ge4C_{\beta}\log(eK)$, $\log x\le x$, and $\log K\le \log(eK)$. Equation \eqref{eq:F-exp-threshold-step-2} exponentiates the negative of the complete bound.

On $\mathcal G_t$, $X_{t,a}\ge4\tau_t\ge4t^\alpha$, so
\begin{subequations}\label{eq:rev-F-witness-coordinate-mass}
\begin{align}
 q_{t,a}&\le e^{-X_{t,a}/\beta_t}
 \le e^{-4t^{\alpha-1/2}/C_{\beta}}
 \le e^{-t^{1/4}/C_{\beta}}\quad(a\ne a^*)\,,
 \label{eq:rev-F-witness-coordinate-mass-step-1}\\
 q_{t,a^*}&=1-\sum_{a\ne a^*}q_{t,a}
 \ge1-Ke^{-t^{1/4}/C_{\beta}}\ge1-\frac1{Kt^2}\ge\frac34\,,
 \label{eq:rev-F-witness-coordinate-mass-step-2}\\
 p_{t,a^*}^c&=(1-\gamma_t^c)q_{t,a^*}
 \ge\frac34\frac34=\frac9{16}\ge\frac12
 >\rho_tN_n\ge\rho_t n_{a^*}\,,
 \label{eq:rev-F-witness-coordinate-mass-step-3}\\
 p_{t,a}^c&=(1-\gamma_t^c)q_{t,a}+\rho_tn_a
                                  \ge\rho_tn_a\quad(a\ne a^*)\,.
 \label{eq:rev-F-witness-coordinate-mass-step-4}
\end{align}
\end{subequations}
Equation \eqref{eq:rev-F-witness-coordinate-mass-step-1} uses that the normalizing denominator of the exponential weights is at least one and the good-history lower gap threshold. Equation \eqref{eq:rev-F-witness-coordinate-mass-step-2} sums over the suboptimal actions, then squares the exponential threshold to obtain $Ke^{-t^{1/4}/C_{\beta}}\le1/(Kt^2)$. Equation \eqref{eq:rev-F-witness-coordinate-mass-step-3} uses $\gamma_t^c\le1/4$, $q_{t,a^*}\ge3/4$, and $\rho_tN_n\le1/4$. Equation \eqref{eq:rev-F-witness-coordinate-mass-step-4} retains the nonnegative added allocation mass of each suboptimal action.

The coordinate mass inequalities imply the matrix comparison directly:
\begin{align}
 M-\rho_tH_n
 &=\sum_b(p_{t,b}^c-\rho_tn_b)bb^\top\succeq0\,,\qquad
 M\succeq\rho_tH_n\succ0\,,
 \label{eq:AB-card-witness-matrix-difference}\\
 \sum_bp_{t,b}^c\{d_a^\top M^{-1}b\}^2
 &=\sum_bd_a^\top M^{-1}(p_{t,b}^cbb^\top)M^{-1}d_a\,,
 \label{eq:AB-card-witness-square-expanded}\\
 &=d_a^\top M^{-1}\left(\sum_bp_{t,b}^cbb^\top\right)M^{-1}d_a
 =d_a^\top M^{-1}MM^{-1}d_a=d_a^\top M^{-1}d_a\,,
 \label{eq:AB-card-witness-square-contracted}
\end{align}
where $d_a=a-a^*$. Equation
\eqref{eq:AB-card-witness-matrix-difference} sums positive semidefinite
rank-one matrices with nonnegative coefficients. Equation
\eqref{eq:AB-card-witness-square-expanded} rewrites each scalar square
as a quadratic form; \eqref{eq:AB-card-witness-square-contracted}
uses the covariance definition only after that expansion.
Consequently $M=\Sigma(p_t^c)\succeq\rho_tH_n\succ0$. For
$v_b^c=M^{-1}b$, $b_{a,b}^c=(a-a^*)^\top v_b^c$,
and $h_{a,b}^c=p_{t,b}^cb_{a,b}^c$, the reconstruction and leverage
identities are
\begin{subequations}\label{eq:rev-F-witness-leverage}
\begin{align}
 \sum_bp_{t,b}^cv_b^cb^\top
 &=M^{-1}\sum_bp_{t,b}^cbb^\top=M^{-1}M=I_d\,,
 \label{eq:rev-F-witness-leverage-step-1}\\
 \lambda_a\coloneq a^\top M^{-1}a
 &\le\rho_t^{-1}a^\top H_n^{-1}a
 \le\frac{\Delta(a)^2}{2\rho_t}\quad(a\ne a^*)\,,
 \label{eq:rev-F-witness-leverage-step-2}\\
 \lambda_{a^*}&=(a^*)^{\top} M^{-1}MM^{-1}a^*
 \ge p_{t,a^*}^c((a^*)^{\top} M^{-1}a^*)^2=p_{t,a^*}^c\lambda_{a^*}^2\,,
 \label{eq:rev-F-witness-leverage-step-3}\\
 \lambda_{a^*}&\le(p_{t,a^*}^c)^{-1}\le2\,,
 \qquad \max_b\lambda_b\le2/\rho_t\,.
 \label{eq:rev-F-witness-leverage-step-4}
\end{align}
\end{subequations}
Equation \eqref{eq:rev-F-witness-leverage-step-1} substitutes $v_b^c=M^{-1}b$ and $M=\sum_bp_{t,b}^cbb^\top$. Equation \eqref{eq:rev-F-witness-leverage-step-2} uses $M\succeq\rho_tH_n$ and the information constraint for the fixed allocation. Equation \eqref{eq:rev-F-witness-leverage-step-3} expands $M$ and retains only its nonnegative $b=a^*$ contribution. Equation \eqref{eq:rev-F-witness-leverage-step-4} divides by $\lambda_{a^*}$ when positive, treats $\lambda_{a^*}=0$ separately, and uses $\Delta(a)\le2$ and $\rho_t\le1$ for the maximum leverage.

The bound on $\lambda_{a^*}$ includes $\lambda_{a^*}=0$; for $\lambda_{a^*}>0$ divide by $\lambda_{a^*}$.
Now, for $a\ne a^*$, the squared triangle inequality and
Cauchy--Schwarz in the $M^{-1}$ norm yield
\begin{subequations}\label{eq:rev-F-witness-budget-substitution}
\begin{align}
 \sum_b\frac{(h_{a,b}^c)^2}{p_{t,b}^c}
 &=\sum_bp_{t,b}^c\{(a-a^*)^\top M^{-1}b\}^2
 \label{eq:rev-F-witness-budget-substitution-step-1}\\
 &=(a-a^*)^\top M^{-1}MM^{-1}(a-a^*)=:V_a^c
 \label{eq:rev-F-witness-budget-substitution-step-2}\\
 &\le2\lambda_a+2\lambda_{a^*}\le\frac{\Delta(a)^2}{\rho_t}+4
 \le\frac{2\Delta(a)^2}{\rho_t}\,,
 \label{eq:rev-F-witness-budget-substitution-step-3}\\
 \max_b\frac{|h_{a,b}^c|}{p_{t,b}^c}
 &=\max_b|b_{a,b}^c|
 \le\sqrt{V_a^c\max_b\lambda_b}
 \le\frac{2\Delta(a)}{\rho_t}=:U_a^c\,,
 \label{eq:rev-F-witness-budget-substitution-step-4}\\
 \frac{2\Delta(a)^2}{\rho_t}
 &=\frac{t\Delta(a)^2}{8B\lo_t^2}
 \le\frac{X_{t,a}^2}{8Bt}
 \le\beta_t+\frac{Z_{t,a}^2}{Bt}\,,
 \label{eq:rev-F-witness-budget-substitution-step-5}\\
 \frac{2\Delta(a)}{\rho_t}
 &=\frac{t\Delta(a)}{8B\lo_t^2}
 \le\frac{X_{t,a}}{8B\lo_t}
 \le\frac{X_{t,a}}{8B}
 \le\beta_t+\frac{Z_{t,a}}B\,.
 \label{eq:rev-F-witness-budget-substitution-step-6}
\end{align}
\end{subequations}
Equation \eqref{eq:rev-F-witness-budget-substitution-step-1} directly substitutes $h_{a,b}^c=p_{t,b}^cb_{a,b}^c$ and cancels one power of $p_{t,b}^c$. Equation \eqref{eq:rev-F-witness-budget-substitution-step-2} sums the resulting quadratic forms using $M=\sum_bp_{t,b}^cbb^\top$. Equation \eqref{eq:rev-F-witness-budget-substitution-step-3} uses the squared triangle inequality in the $M^{-1}$ norm and $4\rho_t\le\Delta_{\min}^2\le\Delta(a)^2$. Equation \eqref{eq:rev-F-witness-budget-substitution-step-4} applies Cauchy--Schwarz in that norm and the maximum leverage. Equation \eqref{eq:rev-F-witness-budget-substitution-step-5} substitutes $\rho_t=16B\lo_t^2/t$, then $t\Delta(a)\le\lo_tX_{t,a}$ and $Z_{t,a}=X_{t,a}$. Equation \eqref{eq:rev-F-witness-budget-substitution-step-6} repeats that substitution for the absolute coefficient and uses $\lo_t\ge1$.

Here $4\rho_t\le\Delta_{\min}^2\le\Delta(a)^2$,
$t\Delta(a)\le\lo_tX_{t,a}$, $\lo_t\ge1$, and $Z_{t,a}=X_{t,a}$
on $\mathcal G_t$. The row for $a=a^*$ is identically zero.

It remains to check the proxy function constraint, rather than just the row budgets.
Columnwise use of \eqref{eq:chi-center} in \Cref{lem:chi}, followed by
$\chi(x)\le x^2e^{|x|}$, gives
\begin{subequations}\label{eq:rev-F-witness-certificate-start}
\begin{align}
 \gC_t(p_t^c)
 &\le\beta_t\sum_bp_{t,b}^c\sum_{a\ne a^*}q_{t,a}
                         \chi(2b_{a,b}^c/\beta_t)
 \label{eq:rev-F-witness-certificate-start-step-1}\\
 &\le\frac4{\beta_t}\sum_{a\ne a^*}q_{t,a}
                \sum_bp_{t,b}^c(b_{a,b}^c)^2
                            e^{2U_a^c/\beta_t}
 \label{eq:rev-F-witness-certificate-start-step-2}\\
 &\le\frac8{\beta_t\rho_t}\sum_{a\ne a^*}
                 q_{t,a}\Delta(a)^2e^{4\Delta(a)/(\beta_t\rho_t)}\,.
 \label{eq:rev-F-witness-certificate-start-step-3}
\end{align}
\end{subequations}
Equation \eqref{eq:rev-F-witness-certificate-start-step-1} is the columnwise centering inequality of \Cref{lem:chi}. Equation \eqref{eq:rev-F-witness-certificate-start-step-2} substitutes $\chi(x)\le x^2e^{|x|}$ at $x=2b_{a,b}^c/\beta_t$ and uses $|b_{a,b}^c|\le U_a^c$. Equation \eqref{eq:rev-F-witness-certificate-start-step-3} substitutes both $\sum_bp_{t,b}^c(b_{a,b}^c)^2\le2\Delta(a)^2/\rho_t$ and $U_a^c=2\Delta(a)/\rho_t$.

The exponent in \eqref{eq:rev-F-witness-certificate-start} satisfies
\begin{subequations}\label{eq:rev-F-witness-certificate-exponent}
\begin{align}
 \frac{4\Delta(a)}{\rho_t}
 &=\frac{t\Delta(a)}{4B\lo_t^2}
 \le\frac{X_{t,a}}{4B\lo_t}\le\frac{X_{t,a}}{4B}\,,
 \label{eq:rev-F-witness-certificate-exponent-step-1}\\
 q_{t,a}e^{4\Delta(a)/(\beta_t\rho_t)}
 &\le e^{-X_{t,a}/\beta_t+X_{t,a}/(4B\beta_t)}
 \le e^{-X_{t,a}/(2\beta_t)}
 \le e^{-\tau_t/(2\beta_t)}\,.
 \label{eq:rev-F-witness-certificate-exponent-step-2}
\end{align}
\end{subequations}
Equation \eqref{eq:rev-F-witness-certificate-exponent-step-1} substitutes the exploration scale before using $t\Delta(a)\le\lo_tX_{t,a}$ and $\lo_t\ge1$. Equation \eqref{eq:rev-F-witness-certificate-exponent-step-2} inserts the resulting exponent in $q_{t,a}\le e^{-X_{t,a}/\beta_t}$, uses $1-1/(4B)\ge1/2$, and then $X_{t,a}\ge\tau_t$.

Substituting \eqref{eq:rev-F-witness-certificate-exponent},
$\Delta(a)^2\le4$, $\rho_t\ge16B/t$, and
\eqref{eq:F-exp-threshold} now gives
\begin{subequations}\label{eq:rev-F-witness-certificate-end}
\begin{align}
 \gC_t(p_t^c)
 &\le\frac{32K}{\beta_t\rho_t}e^{-\tau_t/(2\beta_t)}
 \le\frac{2Kt}{B\beta_t}e^{-t^{1/4}/(2C_{\beta})}
 \label{eq:rev-F-witness-certificate-end-step-1}\\
 &\le\frac{2Kt}{B\beta_t}\frac1{Kt}
 =\frac2{B\beta_t}\le\frac{2d}{\beta_t}\,.
 \label{eq:rev-F-witness-certificate-end-step-2}
\end{align}
\end{subequations}
Equation \eqref{eq:rev-F-witness-certificate-end-step-1} substitutes the exponent bound, $\Delta(a)^2\le4$, $\rho_t\ge16B/t$, and $\tau_t/\beta_t\ge t^{1/4}/C_{\beta}$. Equation \eqref{eq:rev-F-witness-certificate-end-step-2} substitutes the explicit bound $e^{-t^{1/4}/(2C_{\beta})}\le1/(Kt)$, cancels $Kt$, and uses $Bd\ge1$.

Equations \eqref{eq:rev-F-witness-mass},
\eqref{eq:rev-F-witness-leverage},
\eqref{eq:rev-F-witness-budget-substitution}, and
\eqref{eq:rev-F-witness-certificate-end} verify respectively the
mixture, reconstruction, row, and proxy function constraints.
\end{proof}

\subsubsection{A Feasible Exploration Fallback on Every History}
\label{sec:G-feasibility}
The sampling-law optimization must remain feasible even when the estimated gaps
are inaccurate. The next lemma constructs a feasible exploration law and
reconstruction from a G-optimal design on every history. This guarantees
a nonempty feasible set and provides a comparison whenever a cheaper
gap-dependent sampling allocation does not fit the exploration constraints.

\begin{lemma}[A feasible physical covariance reconstruction]\label{lem:AB-fallback}
For every finite pre-action history, let $\pi$ be a G-optimal law and set
$p^0=(1-\Gamma_t)q_t+\Gamma_t\pi$, $M=\sum_bp_b^0bb^\top$,
$w_b^0=\Gamma_t\pi_b$, and $v_b^0=v_b(p^0)=M^{-1}b$.
Then $p^0$, with this prescribed reconstruction and mixture, satisfies every constraint in
\eqref{eq:G-program}. In particular,
\begin{equation*}
 \sum_bp_b^0v_b^0b^\top=I_d\,,\qquad
 \sum_bp_b^0[(a-r_t)^\top v_b^0]^2\le\beta_t\,,
 \qquad \gC_t(p^0)<2d/\beta_t\,. \notag
\end{equation*}
This reconstruction is a comparison point, not a replacement for the executed optimizer.
\end{lemma}
\begin{proof}[Proof of \Cref{lem:AB-fallback}]
Fix $t$ and $\cF_{t-1}$ and suppress $t$ in this proof. Define
$\Sigma(q)=\sum_aq_a a a^\top$ and
$\Sigma(\pi)=\sum_a\pi_a a a^\top$. The proposed exploration and
reconstruction are
\begin{align}
 \gamma^0&=\sum_b\Gamma\pi_b=\Gamma\,,\qquad w_b^0=\Gamma\pi_b\ge0\,,
 \label{eq:AB-fallback-mixture-mass}\\
 p_b^0&=(1-\Gamma)q_b+\Gamma\pi_b
       \ge(1-\Gamma)q_b>\tfrac34q_b>0\,,
 \label{eq:AB-fallback-positive}\\
 \sum_bp_b^0&=(1-\Gamma)\sum_bq_b+\Gamma\sum_b\pi_b
             =(1-\Gamma)+\Gamma=1\,,
 \label{eq:AB-fallback-normalization}\\
 M&=\sum_b\{(1-\Gamma)q_b+\Gamma\pi_b\}bb^\top
    =(1-\Gamma)\Sigma(q)+\Gamma\Sigma(\pi)\,,\qquad
 v_b^0=M^{-1}b\,.
 \label{eq:rev-B-fallback-definition}
\end{align}
Here \eqref{eq:AB-fallback-mixture-mass} substitutes $w_b^0=\Gamma\pi_b$;
\eqref{eq:AB-fallback-positive} uses
$\Gamma=4d/(16d+c_{\beta}\sqrt t)<1/4$ and the strictly positive exponential weights;
\eqref{eq:AB-fallback-normalization} uses that $q,\pi$ are probability laws.
The matrix $M$ in \eqref{eq:rev-B-fallback-definition} is the covariance
of the physical action sampled from $p^0$. Its inverse transforms a
single observed action measurement into a parameter-vector estimate.

Spanning and $q_b>0$ give $\Sigma(q)\succ0$. For any $x\in\R^d$,
\begin{align}
 M&\succeq\Gamma\Sigma(\pi)\succ0\,,
 \qquad M\succeq(1-\Gamma)\Sigma(q)\,,
 \label{eq:AB-fallback-PSD}\\
 x^\top M^{-1}x&\le\Gamma^{-1}x^\top\Sigma(\pi)^{-1}x\,,
 \label{eq:AB-fallback-inverse-order}\\
 b^\top M^{-1}b
 &\le\Gamma^{-1}b^\top\Sigma(\pi)^{-1}b
 \le d/\Gamma=d/(4d/\beta)=\beta/4\,,
 \label{eq:AB-fallback-leverage-bound}\\
 \sum_bp_b^0v_b^0b^\top
 &=\sum_b p_b^0M^{-1}bb^\top
 =M^{-1}\!\left(\sum_bp_b^0bb^\top\right)
 =M^{-1}M=I_d\,.
 \label{eq:rev-B-fallback-leverage}
\end{align}
Equation \eqref{eq:AB-fallback-PSD} follows from the two nonnegative
summands of $M$; \eqref{eq:AB-fallback-inverse-order} reverses positive
definite matrix order. Equation \eqref{eq:AB-fallback-leverage-bound}
uses the G-optimal inequality in \Cref{lem:gopt}, and
\eqref{eq:rev-B-fallback-leverage} substitutes $v_b^0$ before collecting
the covariance sum.

Set $d_a=a-r$ and $\|x\|_{M^{-1}}^2=x^\top M^{-1}x$.
The row variance and update magnitude calculations are
\begin{align}
 \sum_bp_b^0(d_a^\top v_b^0)^2
 &=\sum_bp_b^0(d_a^\top M^{-1}b)^2
 =\sum_bp_b^0d_a^\top M^{-1}bb^\top M^{-1}d_a\,,
 \label{eq:AB-fallback-row-substitute}\\
 &=d_a^\top M^{-1}\!\left(\sum_bp_b^0bb^\top\right)M^{-1}d_a
 =d_a^\top M^{-1}d_a\,,
 \label{eq:AB-fallback-row-covariance}\\
 \|a-r\|_{M^{-1}}^2
 &=2\|a\|_{M^{-1}}^2+2\|r\|_{M^{-1}}^2
                          -\|a+r\|_{M^{-1}}^2\,,
 \label{eq:AB-fallback-square-identity}\\
 &\le2(\beta/4)+2(\beta/4)=\beta
 \le\beta+Z_a^2/(Bt)\,,
 \label{eq:AB-fallback-row-variance}\\
 |d_a^\top v_b^0|
 &=|d_a^\top M^{-1}b|
 \le\|d_a\|_{M^{-1}}\|b\|_{M^{-1}}\,,
 \label{eq:AB-fallback-row-amplitude-CS}\\
 &\le\sqrt\beta\sqrt{\beta/4}=\beta/2
 \le\beta+Z_a/B\,.
 \label{eq:rev-B-fallback-rows}
\end{align}
Equation \eqref{eq:AB-fallback-row-substitute} substitutes the reconstruction;
\eqref{eq:AB-fallback-row-covariance} uses the definition of $M$.
Equation \eqref{eq:AB-fallback-square-identity} expands both quadratic forms;
\eqref{eq:AB-fallback-row-variance} drops a nonnegative square and uses
\eqref{eq:AB-fallback-leverage-bound}. Equation
\eqref{eq:AB-fallback-row-amplitude-CS} applies Cauchy--Schwarz to
$M^{-1/2}d_a$ and $M^{-1/2}b$; \eqref{eq:rev-B-fallback-rows}
substitutes their squared-norm bounds and $Z_a\ge0$.

For the centered proxy function define
\begin{align}
 \bar a&=\sum_aq_a a\,,\qquad
 u_{a,b}=(a-\bar a)^\top M^{-1}b\,,
 \label{eq:AB-fallback-response}\\
 d_t(a;v_b^0,y)
   -\overline d_t(v_b^0,y)
 &=y\left\{(a-r)^\top M^{-1}b
               -\sum_cq_c(c-r)^\top M^{-1}b\right\}
 =yu_{a,b}\,.
 \label{eq:AB-fallback-response-role}
\end{align}
The coefficient $u_{a,b}$ in \eqref{eq:AB-fallback-response} is the response
of action $a$'s estimated loss, relative to the current FTRL-average loss,
to one unit of feedback from physical action $b$.
Equation \eqref{eq:AB-fallback-response-role} explains why the proxy function
uses this coefficient: entropy stability depends on the centered update,
not on its common reference offset.

The absolute response values and their average squared size satisfy
\begin{align}
 \|\bar a\|_{M^{-1}}&=\left\|\sum_aq_a a\right\|_{M^{-1}}
 \le\sum_aq_a\|a\|_{M^{-1}}
 \le\sum_aq_a\sqrt{\beta/4}=\sqrt{\beta/4}\,,
 \label{eq:AB-fallback-mean-norm}\\
 |u_{a,b}|&\le\|a-\bar a\|_{M^{-1}}\|b\|_{M^{-1}}
 \le(\sqrt{\beta/4}+\sqrt{\beta/4})\sqrt{\beta/4}=\beta/2\,,
 \label{eq:AB-fallback-response-amplitude}\\
 \sum_bp_b^0\sum_aq_a u_{a,b}^2
 &=\sum_aq_a(a-\bar a)^\top M^{-1}
          \left(\sum_bp_b^0bb^\top\right)M^{-1}(a-\bar a)\,,
 \label{eq:AB-fallback-response-second-substitute}\\
 &=\sum_aq_a\{a^\top M^{-1}a
               -2a^\top M^{-1}\bar a+\bar a^\top M^{-1}\bar a\}\,,
 \label{eq:AB-fallback-response-expand}\\
 &=\operatorname{Tr}(M^{-1}\Sigma(q))
       -2\bar a^\top M^{-1}\bar a+\bar a^\top M^{-1}\bar a
 \le\operatorname{Tr}(M^{-1}\Sigma(q))\,,
 \label{eq:AB-fallback-response-trace}\\
 M^{-1/2}\Sigma(q)M^{-1/2}&\preceq(1-\Gamma)^{-1}I_d\,,
 \label{eq:AB-fallback-trace-order}\\
 \operatorname{Tr}(M^{-1}\Sigma(q))
 &=\operatorname{Tr}(M^{-1/2}\Sigma(q)M^{-1/2})
 \le d/(1-\Gamma)\,.
 \label{eq:AB-fallback-trace-bound}
\end{align}
Equation \eqref{eq:AB-fallback-mean-norm} is the norm triangle inequality;
\eqref{eq:AB-fallback-response-amplitude} uses Cauchy--Schwarz and that bound.
Equation \eqref{eq:AB-fallback-response-second-substitute} substitutes $u_{a,b}$;
\eqref{eq:AB-fallback-response-expand} uses $M^{-1}MM^{-1}=M^{-1}$ and
expands the centered quadratic form. Equation
\eqref{eq:AB-fallback-response-trace} uses $\sum_aq_a a=\bar a$ and
$\sum_aq_a=1$. Equation \eqref{eq:AB-fallback-trace-order} conjugates
$M\succeq(1-\Gamma)\Sigma(q)$ by $M^{-1/2}$, and
\eqref{eq:AB-fallback-trace-bound} takes traces.

Since $|u_{a,b}/\beta|\le1/2$, the small-argument inequality in
\Cref{lem:chi} applies at every pair $(a,b)$. Its substitution is
\begin{align}
 \chi(u_{a,b}/\beta)&\le\tfrac32(u_{a,b}/\beta)^2
                         =3u_{a,b}^2/(2\beta^2)\,,
 \label{eq:AB-fallback-chi-pointwise}\\
 \gC_t(p^0)
 &=\beta\sum_bp_b^0\sum_aq_a
       \chi\!\left(\frac{(a-\bar a)^\top M^{-1}b}{\beta}\right)
 =\beta\sum_bp_b^0\sum_aq_a\chi(u_{a,b}/\beta)\,,
 \label{eq:AB-fallback-cert-substitute}\\
 &\le\beta\sum_bp_b^0\sum_aq_a\frac{3u_{a,b}^2}{2\beta^2}
 =\frac3{2\beta}\sum_bp_b^0\sum_aq_a u_{a,b}^2\,,
 \label{eq:AB-fallback-cert-quadratic}\\
 &\le\frac3{2\beta}\frac d{1-\Gamma}
 <\frac{3d}{2\beta(3/4)}=\frac{2d}\beta\,.
 \label{eq:rev-B-fallback-certificate}
\end{align}
Equation \eqref{eq:AB-fallback-chi-pointwise} records the exact scalar
inequality used; \eqref{eq:AB-fallback-cert-substitute} substitutes the
covariance reconstruction into the algorithm's proxy function.
Equation \eqref{eq:AB-fallback-cert-quadratic} sums the pointwise bounds,
and \eqref{eq:rev-B-fallback-certificate} uses
\eqref{eq:AB-fallback-trace-bound} and $\Gamma<1/4$.
Together with \eqref{eq:AB-fallback-mixture-mass},
\eqref{eq:rev-B-fallback-leverage}, \eqref{eq:AB-fallback-row-variance},
and \eqref{eq:rev-B-fallback-rows}, this verifies all constraints
for the same physical reconstruction.
\end{proof}

\subsubsection{Empirical Gap Ratios for the Exploration Comparison}
\begin{definition}[Empirical gap ratios]\label{def:B-gap-ratios}
Use the common event in \Cref{def:AB-card-good}.
On $\mathcal G_t$ define
\[
 \lo_t=\max\{1,\max_{a\ne a^*}t\Delta(a)/X_{t,a}\}\,,\qquad
 \up_t=\max_{a\ne a^*}\widehat\Delta_t(a)/\Delta(a)\,,
\]
and define them as zero off the event.
\end{definition}

\begin{lemma}[Finite empirical gap-ratio moments]\label{lem:F-ratios}
For $t\ge T_\circ$,
\begin{align}\label{eq:F-ratio-bounds}
 \|\ind_{\mathcal G_t}\lo_t\|_P&\le25\,,&
 \|\ind_{\mathcal G_t}\up_t\|_P&\le16\Psi_t\,,\notag\\
 \E[\ind_{\mathcal G_t}\lo_t\up_t]&\le400\Psi_t\,,&
 \E[\ind_{\mathcal G_t}\lo_t^3\up_t]&\le250000\Psi_t\,,\notag\\
 \E[\ind_{\mathcal G_t}\lo_t^5\up_t]&\le16\cdot25^5\Psi_t\,.
\end{align}
On $\mathcal G_t$, $\Delta(a)\le\lo_t\widehat\Delta_t(a)\le\lo_t\up_t\Delta(a)$ for every arm.
\end{lemma}
\begin{proof}
The separation of the cumulative loss estimates in \Cref{lem:B-good-event-tail} and the
compensator identity in \Cref{lem:AB-card-compensation} give
the following two-sided comparison on $\mathcal G_t$:
\begin{subequations}\label{eq:rev-F-score-compensated-comparison}
\begin{align}
 X_{t,a}
 &\le X_{t,a}-E_{t,a}+2C+2\tau_t=Y_{t,a}^\dagger
 \label{eq:rev-F-score-compensated-comparison-step-1}\\
 &=W_{t,a}+2\tau_t+\Lambda_{t,a}
 \le W_{t,a}+3\tau_t
 \le\tfrac74W_{t,a}\le2X_{t,a}\,.
 \label{eq:rev-F-score-compensated-comparison-step-2}
\end{align}
\end{subequations}
Equation \eqref{eq:rev-F-score-compensated-comparison-step-1} uses $E_{t,a}\le2C$. Equation \eqref{eq:rev-F-score-compensated-comparison-step-2} substitutes the identity for the compensated cumulative loss estimate, then $\Lambda_{t,a}\le\tau_t$, $W_{t,a}\ge4\tau_t$, and $W_{t,a}\le X_{t,a}$. The two directions of this comparison control the lower and upper
empirical-gap ratios, respectively.

The inverse and positive local moments used below are supplied by
\Cref{lem:F-local,lem:B-positive-local}.
Substitute \eqref{eq:rev-F-score-compensated-comparison} into the
definitions of both ratios, before applying the local moment bounds:
\begin{subequations}\label{eq:rev-F-ratio-moment-substitution}
\begin{align}
 \E[\ind_{\mathcal G_t}\lo_t^P]
 &\le1+\sum_{a\ne a^*}(t\Delta(a))^P
                           \E[\ind_{\mathcal G_t}X_{t,a}^{-P}]
 \label{eq:rev-F-ratio-moment-substitution-step-1}\\
 &\le1+\sum_{a\ne a^*}(2t\Delta(a))^P
                           \E[I_t(Y_{t,a}^\dagger)^{-P}]
 \label{eq:rev-F-ratio-moment-substitution-step-2}\\
 &\le1+\sum_{a\ne a^*}(2t\Delta(a))^P
                                \left(\frac4{t\Delta(a)}\right)^P
 \le1+K8^P\,,
 \label{eq:rev-F-ratio-moment-substitution-step-3}\\
 \E[\ind_{\mathcal G_t}\up_t^P]
 &\le\sum_{a\ne a^*}\E\!\left[
      \ind_{\mathcal G_t}\left(\frac{X_{t,a}}{t\Delta(a)}\right)^P\right]
 \label{eq:rev-F-ratio-moment-substitution-step-4}\\
 &\le\sum_{a\ne a^*}\E\!\left[
         I_t\left(\frac{Y_{t,a}^\dagger}{t\Delta(a)}\right)^P\right]
 \le K(8\Psi_t)^P\,.
 \label{eq:rev-F-ratio-moment-substitution-step-5}
\end{align}
\end{subequations}
Equation \eqref{eq:rev-F-ratio-moment-substitution-step-1} bounds a maximum of nonnegative inverse-gap powers by their sum, retaining the floor at one. Equation \eqref{eq:rev-F-ratio-moment-substitution-step-2} uses $X_{t,a}^{-1}\le2(Y_{t,a}^\dagger)^{-1}$ on $\mathcal G_t$ and $\ind_{\mathcal G_t}\le I_t$. Equation \eqref{eq:rev-F-ratio-moment-substitution-step-3} substitutes the inverse moment before canceling $t\Delta(a)$. Equation \eqref{eq:rev-F-ratio-moment-substitution-step-4} bounds the upper-ratio maximum by a sum. Equation \eqref{eq:rev-F-ratio-moment-substitution-step-5} uses $X_{t,a}\le Y_{t,a}^\dagger$ and substitutes the positive moment at every action.

Hence the $P$th roots obey
\begin{subequations}\label{eq:rev-F-ratio-roots}
\begin{align}
 \|\ind_{\mathcal G_t}\lo_t\|_P
 &\le(1+K8^P)^{1/P}\le1+8K^{1/P}=1+8k\le17\le25\,,
 \label{eq:rev-F-ratio-roots-step-1}\\
 \|\ind_{\mathcal G_t}\up_t\|_P
 &\le8k\Psi_t\le16\Psi_t\,.
 \label{eq:rev-F-ratio-roots-step-2}
\end{align}
\end{subequations}
Equation \eqref{eq:rev-F-ratio-roots-step-1} takes the $P$th root, uses $(x+y)^{1/P}\le x^{1/P}+y^{1/P}$, and $k<2$. Equation \eqref{eq:rev-F-ratio-roots-step-2} takes the root of the upper-ratio sum and uses the same bound on $k$.

For $j\in\{1,3,5\}$, generalized H\"older with exponents
$P/j$, $P$, and $P/(P-j-1)$ yields
\begin{subequations}\label{eq:rev-F-three-holder-products}
\begin{align}
 \E[\ind_{\mathcal G_t}\lo_t^j\up_t]
 &\le\{\E[\ind_{\mathcal G_t}\lo_t^P]\}^{j/P}
       \{\E[\ind_{\mathcal G_t}\up_t^P]\}^{1/P}
       \sP(\mathcal G_t)^{1-(j+1)/P}
 \label{eq:rev-F-three-holder-products-step-1}\\
 &\le25^j(16\Psi_t)\,,
 \label{eq:rev-F-three-holder-products-step-2}\\
 \E[\ind_{\mathcal G_t}\lo_t\up_t]&\le400\Psi_t\,,\qquad
 \E[\ind_{\mathcal G_t}\lo_t^3\up_t]\le250000\Psi_t\,,
 \label{eq:rev-F-three-holder-products-step-3}\\
 \E[\ind_{\mathcal G_t}\lo_t^5\up_t]&\le16\cdot25^5\Psi_t\,.
 \label{eq:rev-F-three-holder-products-step-4}
\end{align}
\end{subequations}
Equation \eqref{eq:rev-F-three-holder-products-step-1} is generalized H\"older applied to $\ind_{\mathcal G_t}\lo_t^j$, $\ind_{\mathcal G_t}\up_t$, and $\ind_{\mathcal G_t}$ with reciprocal exponents $j/P$, $1/P$, and $1-(j+1)/P$. Equation \eqref{eq:rev-F-three-holder-products-step-2} substitutes both ratio roots and $\sP(\mathcal G_t)\le1$. Equations \eqref{eq:rev-F-three-holder-products-step-3} and \eqref{eq:rev-F-three-holder-products-step-4} evaluate $25^j\cdot16$ for $j=1,3,5$.

All exponents are finite because $P\ge512>6$.
Finally, for each $a\ne a^*$, the two clipping cases give
\begin{subequations}\label{eq:rev-F-clipped-gap-comparison}
\begin{align}
 X_{t,a}/t\le2
 &\Longrightarrow\widehat\Delta_t(a)=X_{t,a}/t\,,
 \quad\lo_t\widehat\Delta_t(a)
       \ge(t\Delta(a)/X_{t,a})(X_{t,a}/t)=\Delta(a)\,,
 \label{eq:rev-F-clipped-gap-comparison-step-1}\\
 X_{t,a}/t>2
 &\Longrightarrow\widehat\Delta_t(a)=2\,,
 \quad\lo_t\widehat\Delta_t(a)\ge2\ge\Delta(a)\,,
 \label{eq:rev-F-clipped-gap-comparison-step-2}\\
 \widehat\Delta_t(a)/\Delta(a)\le\up_t
 &\Longrightarrow\lo_t\widehat\Delta_t(a)
                           \le\lo_t\up_t\Delta(a)\,.
 \label{eq:rev-F-clipped-gap-comparison-step-3}
\end{align}
\end{subequations}
Equation \eqref{eq:rev-F-clipped-gap-comparison-step-1} treats the untruncated estimate $\widehat\Delta_t(a)=X_{t,a}/t$. Equation \eqref{eq:rev-F-clipped-gap-comparison-step-2} separately treats truncation at two and uses $\Delta(a)\le2$. Equation \eqref{eq:rev-F-clipped-gap-comparison-step-3} uses the definition of $\up_t$ and multiplies by the nonnegative $\lo_t$.

For $a=a^*$, $\Delta(a^*) =\widehat\Delta_t(a^*)=0$, so the gap comparisons
in \eqref{eq:rev-F-clipped-gap-comparison} hold as equalities.
\end{proof}

\subsubsection{Positive Local Moments and Accumulated Ratio Distortion}
The upper empirical-gap ratio and its distortion factor enter only the
exploration comparison. Their proof uses the same compensated process as
the common inverse-moment estimate, but requires its positive moments.

\begin{lemma}[Positive local moments and bounded accumulated distortion]\label{lem:B-positive-local}
For $a\ne a^*$ and $t\ge T_\circ$, use the window of
\Cref{lem:AB-card-window-setup} and the compensated process of
\Cref{lem:AB-card-compensation}.
Set
\begin{equation}\label{eq:F-Psi}
 \Psi_t=1+A_0u_0^{1-\kappa_0}\Delta_{\min}^{-1}(n_0/t)^{\sigma_0}\,,\qquad
 \Psi_{\Delta_{\min}}=1+A_0\Delta_{\min}^{-\varepsilon_0}\,.
\end{equation}
Then
\begin{equation}\label{eq:F-positive-local}
 \left\{\E\left[I_t\left(\frac{Y_{t,a}^{\dagger}}{t\Delta(a)}\right)^P\right]\right\}^{1/P}
 \le8\Psi_t\,,
 \quad\Psi_t\le\Psi_{\Delta_{\min}}\,,
 \quad\sum_{t\ge T_\circ}\frac{\Psi_t-1}{t}\le3A_0\Delta_{\min}^{-\varepsilon_0}\,.
\end{equation}
\end{lemma}
\begin{proof}
Fix $a\ne a^*$ and abbreviate $Y_{s,a}^{\dagger}$ by $Y_s$.
On $I_s=1$, let $y=Y_s$ and $\xi=\xi_{s,a}$.
The increment bounds in \Cref{lem:AB-card-compensation}
give $|\xi|/y\le3/B$ and the conditional moments used below.
Taylor's formula and
$(1+3/B)^{P-2}\le2$ give
\begin{subequations}\label{eq:rev-F-positive-taylor}
\begin{align}
 \E_{s-1}[(y+\xi)^P]
 &\le y^P+Py^{P-1}(\Delta(a)+2v_s)
                     +P(P-1)y^{P-2}\{8\beta_s+4y^2/(Bs)\}
 \label{eq:rev-F-positive-taylor-step-1}\\
 &=\left(1+\frac{4P(P-1)}{Bs}\right)y^P
       +P\left(\Delta(a)+2v_s+\frac{8(P-1)\beta_s}{y}\right)y^{P-1}
 \label{eq:rev-F-positive-taylor-step-2}\\
 &\le(1+P\kappa_0/s)y^P
       +P\{\Delta(a)+2\alpha s^{\alpha-1}
                           +8(P-1)C_{\beta} s^{1/2-\alpha}\}y^{P-1}\,.
 \label{eq:rev-F-positive-taylor-step-3}
\end{align}
\end{subequations}
Equation \eqref{eq:rev-F-positive-taylor-step-1} is positive-power Taylor after both conditional increment moments are substituted. Equation \eqref{eq:rev-F-positive-taylor-step-2} distributes the product and groups the $y^P$ and $y^{P-1}$ coefficients. Equation \eqref{eq:rev-F-positive-taylor-step-3} uses $4(P-1)/B\le\kappa_0$, $v_s\le\alpha s^{\alpha-1}$, $y\ge s^\alpha$, and $\beta_s\le C_{\beta}\sqrt s$.

Here $4(P-1)/B\le\kappa_0$, $y\ge s^\alpha$, and
$\beta_s\le C_{\beta}\sqrt s$. Set
$w_s=(\E[I_sY_s^P])^{1/P}$ and
$a_s=\Delta(a)+2\alpha s^{\alpha-1}+8(P-1)C_{\beta} s^{1/2-\alpha}$.
H\"older and the binomial tangent bound imply
\begin{subequations}\label{eq:rev-F-positive-root-recursion}
\begin{align}
 w_{s+1}^P
 &\le(1+P\kappa_0/s)w_s^P
                  +Pa_s\E[I_sY_s^{P-1}]
 \label{eq:rev-F-positive-root-recursion-step-1}\\
 &\le(1+P\kappa_0/s)w_s^P+Pa_sw_s^{P-1}
 \label{eq:rev-F-positive-root-recursion-step-2}\\
 &\le\{(1+\kappa_0/s)w_s+a_s\}^P\,,
 \label{eq:rev-F-positive-root-recursion-step-3}\\
 w_{s+1}&\le(1+\kappa_0/s)w_s+a_s\,.
 \label{eq:rev-F-positive-root-recursion-step-4}
\end{align}
\end{subequations}
Equation \eqref{eq:rev-F-positive-root-recursion-step-1} takes expectations with the predictable surviving-history indicator. Equation \eqref{eq:rev-F-positive-root-recursion-step-2} uses $\E[I_sY_s^{P-1}]\le w_s^{P-1}$. Equation \eqref{eq:rev-F-positive-root-recursion-step-3} expands $\{w_s+(\kappa_0/s)w_s+a_s\}^P$ and retains its nonnegative constant and linear terms. Equation \eqref{eq:rev-F-positive-root-recursion-step-4} takes the monotone $P$th root.

For $j\ge m\ge2$, the product is bounded by
\begin{align}
 \prod_{u=j}^{t-1}(1+\kappa_0/u)
 &\le\exp\!\left(\kappa_0\sum_{u=j}^{t-1}\frac1u\right)
 \le\left(\frac{2t}j\right)^{\kappa_0}
 \le2(t/j)^{\kappa_0}\,.
 \label{eq:rev-F-local-product}
\end{align}
Unroll \eqref{eq:rev-F-positive-root-recursion}, apply
\eqref{eq:rev-F-local-product}, and integrate each decreasing power:
\begin{subequations}\label{eq:rev-F-positive-source-sums}
\begin{align}
 w_t
 &\le2(t/m)^{\kappa_0}w_m
        +2t^{\kappa_0}\sum_{s=m}^{t-1}s^{-\kappa_0}
              \{\Delta(a)+2\alpha s^{\alpha-1}
                               +8(P-1)C_{\beta} s^{1/2-\alpha}\}
 \label{eq:rev-F-positive-source-sums-step-1}\\
 &\le2(t/m)^{\kappa_0}w_m
       +2\Delta(a) t^{\kappa_0}\int_0^t x^{-\kappa_0}dx
 \label{eq:rev-F-positive-source-sums-step-2}\\
 &\quad+4\alpha t^{\kappa_0}\int_0^t x^{\alpha-1-\kappa_0}dx
       +16(P-1)C_{\beta} t^{\kappa_0}
                         \int_0^t x^{1/2-\alpha-\kappa_0}dx
 \label{eq:rev-F-positive-source-sums-step-3}\\
 &=2(t/m)^{\kappa_0}w_m
       +\frac2{1-\kappa_0}\Delta(a) t
       +\frac{4\alpha}{\alpha-\kappa_0}t^\alpha
       +\frac{16(P-1)C_{\beta}}{3/2-\alpha-\kappa_0}t^{3/2-\alpha}
 \label{eq:rev-F-positive-source-sums-step-4}\\
 &\le2(t/m)^{\kappa_0}w_m
       +\frac{512}{255}\Delta(a) t+5t^\alpha
       +32(P-1)C_{\beta} t^{3/2-\alpha}\,.
 \label{eq:rev-F-positive-source-sums-step-5}
\end{align}
\end{subequations}
Equation \eqref{eq:rev-F-positive-source-sums-step-1} unrolls the root recurrence and applies the product bound to the initial value and every source. Equations \eqref{eq:rev-F-positive-source-sums-step-2} and \eqref{eq:rev-F-positive-source-sums-step-3} replace each decreasing power sum by its integral, keeping the three sources distinct. Equation \eqref{eq:rev-F-positive-source-sums-step-4} evaluates all three integrals. Equation \eqref{eq:rev-F-positive-source-sums-step-5} uses $2/(1-\kappa_0)=512/255$, $4\alpha/(\alpha-\kappa_0)\le5$, and $16/(3/2-\alpha-\kappa_0)\le32$.

All three integral exponents belong to $(-1,0)$.
The initial moment is controlled by the full-history moment, not by
restarting the learner. Specifically,
\begin{subequations}\label{eq:rev-F-local-initial-moment}
\begin{align}
 w_m
 &=\|I_m(X_{m,a}-E_{m,a}+2C+2\tau_m)\|_P
 \label{eq:rev-F-local-initial-moment-step-1}\\
 &\le\|X_{m,a}\|_P+\|E_{m,a}\|_P+2C+2\tau_m
 \label{eq:rev-F-local-initial-moment-step-2}\\
 &\le(16k+1)m+4C+2\tau_m
 \le(16k+1)m+4m+10m=(16k+15)m\,.
 \label{eq:rev-F-local-initial-moment-step-3}
\end{align}
\end{subequations}
Equation \eqref{eq:rev-F-local-initial-moment-step-1} substitutes the definition of $Y_{m,a}^\dagger$ inside its norm. Equation \eqref{eq:rev-F-local-initial-moment-step-2} applies Minkowski and $I_m\le1$. Equation \eqref{eq:rev-F-local-initial-moment-step-3} uses the full-history absolute moment, $|E_{m,a}|\le2C$, $m\ge C$, and $\tau_m\le5m$, without resetting the learner's cumulative loss estimates at $m$.

This uses $m\ge N_{\rm burn},C$, $|E_{m,a}|\le2C$, and
$\tau_m\le5m$.
The inequality $m\le2\max(t^\lambda,z_t)$ gives
\begin{subequations}\label{eq:rev-F-local-initial-substitution}
\begin{align}
 \frac{2(t/m)^{\kappa_0}w_m}{t\Delta(a)}
 &\le2(16k+15)\Delta_{\min}^{-1}(m/t)^{1-\kappa_0}
 \label{eq:rev-F-local-initial-substitution-step-1}\\
 &\le A_0\Delta_{\min}^{-1}\{t^{-(1-\lambda)(1-\kappa_0)}
                          +(z_t/t)^{1-\kappa_0}\}
 \label{eq:rev-F-local-initial-substitution-step-2}\\
 &=A_0\Delta_{\min}^{-1} t^{-a_0}
        +A_0u_0^{1-\kappa_0}\Delta_{\min}^{-1}(n_0/t)^{\sigma_0}\,.
 \label{eq:rev-F-local-initial-substitution-step-3}
\end{align}
\end{subequations}
Equation \eqref{eq:rev-F-local-initial-substitution-step-1} substitutes the initial-moment bound and $\Delta(a)^{-1}\le \Delta_{\min}^{-1}$. Equation \eqref{eq:rev-F-local-initial-substitution-step-2} uses $m\le2\max(t^\lambda,z_t)$ and $\max(x,y)^{1-\kappa_0}\le x^{1-\kappa_0}+y^{1-\kappa_0}$. Equation \eqref{eq:rev-F-local-initial-substitution-step-3} substitutes $z_t$, $a_0$, and $\sigma_0$ directly.

The coefficient uses $2^{1-\kappa_0}\le2$ and
$A_0=4(16k+15)$. Substituting this into
\eqref{eq:rev-F-positive-source-sums}, and dividing every remaining
term by $t\Delta(a)$, yields
\begin{subequations}\label{eq:rev-F-local-final-before-threshold}
\begin{align}
 \frac{w_t}{t\Delta(a)}
 &\le\frac{512}{255}+A_0\Delta_{\min}^{-1} t^{-a_0}
       +5\Delta_{\min}^{-1} t^{-(1-\alpha)}
       +32(P-1)C_{\beta}\Delta_{\min}^{-1} t^{-(\alpha-1/2)}
 \label{eq:rev-F-local-final-before-threshold-step-1}\\
 &\quad+A_0u_0^{1-\kappa_0}\Delta_{\min}^{-1}(n_0/t)^{\sigma_0}\,.
 \label{eq:rev-F-local-final-before-threshold-step-2}
\end{align}
\end{subequations}
Equations \eqref{eq:rev-F-local-final-before-threshold-step-1} and \eqref{eq:rev-F-local-final-before-threshold-step-2} substitute the full initial contribution into the solved root recurrence and divide every source term by $t\Delta(a)$. The retained homogeneous term depends on $n_0/t$ and must not be absorbed into the coefficient of $\log T$.

Each pure-time error in \eqref{eq:rev-F-local-final-before-threshold}
is at most one, as the substitutions explicitly show:
\begin{subequations}\label{eq:rev-F-local-threshold-substitution}
\begin{align}
 A_0\Delta_{\min}^{-1} t^{-a_0}
 &\le A_0Q^{-a_0}\Delta_{\min}^{5a_0-1}\le1\,,
 \label{eq:rev-F-local-threshold-substitution-step-1}\\
 5\Delta_{\min}^{-1} t^{-(1-\alpha)}
 &\le5Q^{-(1-\alpha)}\Delta_{\min}^{5(1-\alpha)-1}\le1\,,
 \label{eq:rev-F-local-threshold-substitution-step-2}\\
 32(P-1)C_{\beta}\Delta_{\min}^{-1} t^{-(\alpha-1/2)}
 &\le32(P-1)C_{\beta}Q^{-(\alpha-1/2)}
                         \Delta_{\min}^{5(\alpha-1/2)-1}\le1\,,
 \label{eq:rev-F-local-threshold-substitution-step-3}\\
 \frac{w_t}{t\Delta(a)}
 &\le\frac{512}{255}+3+(\Psi_t-1)
 \le8\Psi_t\,.
 \label{eq:rev-F-local-threshold-substitution-step-4}
\end{align}
\end{subequations}
Equations \eqref{eq:rev-F-local-threshold-substitution-step-1}, \eqref{eq:rev-F-local-threshold-substitution-step-2}, and \eqref{eq:rev-F-local-threshold-substitution-step-3} each first substitute $t\ge Q\Delta_{\min}^{-5}$ and then use the corresponding threshold in $Q$, together with $5a_0>1$, $5(1-\alpha)>1$, and $5(\alpha-1/2)>1$. Equation \eqref{eq:rev-F-local-threshold-substitution-step-4} inserts all three resulting bounds and the definition of $\Psi_t$, then uses $\Psi_t\ge1$.

Finally, substitute $t\ge1024n_0\Delta_{\min}^{-1}$ into the distortion and sum
that distortion, rather than multiplying a gap-dependent constant
by the logarithmic leading term:
\begin{subequations}\label{eq:rev-F-distortion-sum}
\begin{align}
 \Psi_t-1
 &=A_0u_0^{1-\kappa_0}\Delta_{\min}^{-1}(n_0/t)^{\sigma_0}
 \label{eq:rev-F-distortion-sum-step-1}\\
 &\le A_0\left(\frac{u_0}{1024^\omega}\right)^{1-\kappa_0}
                    \Delta_{\min}^{\sigma_0-1}
 \le A_0\Delta_{\min}^{-\varepsilon_0}=\Psi_{\Delta_{\min}}-1\,,
 \label{eq:rev-F-distortion-sum-step-2}\\
 \sum_{t\ge T_\circ}\frac{\Psi_t-1}{t}
 &=A_0u_0^{1-\kappa_0}\Delta_{\min}^{-1} n_0^{\sigma_0}
                           \sum_{t\ge T_\circ}t^{-1-\sigma_0}
 \label{eq:rev-F-distortion-sum-step-3}\\
 &\le A_0u_0^{1-\kappa_0}\Delta_{\min}^{-1} n_0^{\sigma_0}
       (1+1/\sigma_0)T_\circ^{-\sigma_0}
 \label{eq:rev-F-distortion-sum-step-4}\\
 &\le3A_0\left(\frac{u_0}{1024^\omega}\right)^{1-\kappa_0}
                       \Delta_{\min}^{-\varepsilon_0}
 \le3A_0\Delta_{\min}^{-\varepsilon_0}\,.
 \label{eq:rev-F-distortion-sum-step-5}
\end{align}
\end{subequations}
Equation \eqref{eq:rev-F-distortion-sum-step-1} is the distortion definition. Equation \eqref{eq:rev-F-distortion-sum-step-2} substitutes $t\ge1024n_0\Delta_{\min}^{-1}$ and $\sigma_0=\omega(1-\kappa_0)$. Equation \eqref{eq:rev-F-distortion-sum-step-3} substitutes that definition before summing. Equation \eqref{eq:rev-F-distortion-sum-step-4} compares the power tail with its first term plus integral. Equation \eqref{eq:rev-F-distortion-sum-step-5} uses $T_\circ\ge1024n_0\Delta_{\min}^{-1}$, $1+1/\sigma_0<3$, and $u_0<1024^\omega$.

We used $u_0/1024^\omega<1$ and $1+1/\sigma_0<3$.
\end{proof}

\subsubsection{Weighted Allocation-Mass Failure Sums}
Only the exploration comparison uses these two mass-failure terms.
The common weighted good-event failure bound remains in
\Cref{lem:F-sums}.

\begin{lemma}[Weighted allocation-mass failure sums]\label{lem:B-mass-sums}
For the nonnegative thresholds $a_n,a_{\Delta_{\min}}$ in
\Cref{def:AB-card-witness},
\begin{align}
 \sum_{t\ge1}t^{-1/2}\min\{400,16\cdot25^5a_n^2/t\}&\le10^6a_n\,,\label{eq:F-mass-sum1}\\
 \sum_{t\ge1}t^{-1/2}\min\{400,16\cdot25^5a_{\Delta_{\min}}^2/t^2\}&\le30000\sqrt{a_{\Delta_{\min}}}\,.\label{eq:F-mass-sum2}
\end{align}
\end{lemma}
\begin{proof}
For the mass failures, the summands are decreasing. Set
$L_5=16\cdot25^5=400\cdot625^2$ and first split the integral at
$x_n=625^2a_n^2$. Both integrals are explicit:
\begin{subequations}\label{eq:rev-F-first-mass-integral}
\begin{align}
 \sum_{t\ge1}t^{-1/2}\min\{400,L_5a_n^2/t\}
 &\le400\int_0^{x_n}x^{-1/2}dx
             +L_5a_n^2\int_{x_n}^\infty x^{-3/2}dx
 \label{eq:rev-F-first-mass-integral-step-1}\\
 &=800\sqrt{x_n}+2L_5a_n^2x_n^{-1/2}
 \label{eq:rev-F-first-mass-integral-step-2}\\
 &=800(625a_n)
       +\frac{2(400\cdot625^2)a_n^2}{625a_n}
 =10^6a_n\,.
 \label{eq:rev-F-first-mass-integral-step-3}
\end{align}
\end{subequations}
Equation \eqref{eq:rev-F-first-mass-integral-step-1} uses monotonicity and splits at the exact crossing $x_n=L_5a_n^2/400=625^2a_n^2$. Equation \eqref{eq:rev-F-first-mass-integral-step-2} evaluates both antiderivatives. Equation \eqref{eq:rev-F-first-mass-integral-step-3} substitutes $\sqrt{x_n}=625a_n$ and $L_5=400\cdot625^2$, so both contributions equal $500000a_n$.

For the second integral split at $x_{\Delta_{\min}}=625a_{\Delta_{\min}}$:
\begin{subequations}\label{eq:rev-F-second-mass-integral}
\begin{align}
 \sum_{t\ge1}t^{-1/2}\min\{400,L_5a_{\Delta_{\min}}^2/t^2\}
 &\le400\int_0^{x_{\Delta_{\min}}}x^{-1/2}dx
             +L_5a_{\Delta_{\min}}^2\int_{x_{\Delta_{\min}}}^\infty x^{-5/2}dx
 \label{eq:rev-F-second-mass-integral-step-1}\\
 &=800\sqrt{x_{\Delta_{\min}}}
          +\tfrac23L_5a_{\Delta_{\min}}^2x_{\Delta_{\min}}^{-3/2}
 \label{eq:rev-F-second-mass-integral-step-2}\\
 &=20000\sqrt{a_{\Delta_{\min}}}
       +\tfrac{20000}3\sqrt{a_{\Delta_{\min}}}
 \le30000\sqrt{a_{\Delta_{\min}}}\,.
 \label{eq:rev-F-second-mass-integral-step-3}
\end{align}
\end{subequations}
Equation \eqref{eq:rev-F-second-mass-integral-step-1} splits at $x_{\Delta_{\min}}=\sqrt{L_5/400}\,a_{\Delta_{\min}}=625a_{\Delta_{\min}}$. Equation \eqref{eq:rev-F-second-mass-integral-step-2} evaluates the $x^{-1/2}$ and $x^{-5/2}$ integrals. Equation \eqref{eq:rev-F-second-mass-integral-step-3} substitutes $\sqrt{x_{\Delta_{\min}}}=25\sqrt{a_{\Delta_{\min}}}$ and $L_5=400\cdot625^2$, obtaining $20000(1+1/3)\sqrt{a_{\Delta_{\min}}}\le30000\sqrt{a_{\Delta_{\min}}}$.

These are \eqref{eq:F-mass-sum1} and \eqref{eq:F-mass-sum2}.
\end{proof}

\subsection{Proof of Direct Corruption Term}\label{app:B-corruption}
\begin{lemma}[Direct corruption contribution]\label{lem:B-corruption-term}
For $e_{t,a}=(a-a^*)^\top c_t^{\rm vec}$,
\[
 \mathsf I_4\coloneq \sum_{t=1}^T\E[\langle p_t-q_t,e_t\rangle]
 \le\sum_{t=1}^T\E\left[\left|
       \sum_a(p_{t,a}-q_{t,a})a^\top c_t^{\rm vec}\right|\right]
 \le16\sqrt2c_{R}\sqrt C\,.
\]
\end{lemma}
\begin{proof}
\interdisplaylinepenalty=10000
Let $\gamma_t=\sum_aw_{t,a}\le\Gamma_t$ and
$\kappa_t=\|c_t^{\rm vec}\|_\cA$. The two probability distributions have
the same total mass, so their signed difference removes the common
$(a^*)^{\top} c_t^{\rm vec}$ term. Hence
\begin{align}
 \mathsf I_4
 &=\sum_{t=1}^T\E[\sum_a(w_{t,a}-\gamma_tq_{t,a})a^\top c_t^{\rm vec}]
       \label{eq:B-corruption-shift}\\
 &\le\E[\sum_{t=1}^T\kappa_t
                  \left(\sum_aw_{t,a}+\gamma_t\sum_aq_{t,a}\right)]
 =\E[\sum_{t=1}^T2\gamma_t\kappa_t]
       \label{eq:B-corruption-mass}\\
 &\le8c_{R}\,\E[\sum_{t=1}^T\kappa_t/\sqrt t]
 \le8c_{R}\int_0^{C/2}2x^{-1/2}\,dx
 =16\sqrt2c_{R}\sqrt C\,.
       \label{eq:B-corruption-final}
\end{align}
Equation \eqref{eq:B-corruption-shift} uses
$p_t-q_t=w_t-\gamma_tq_t$.
Equation \eqref{eq:B-corruption-mass} uses
$|a^\top c_t^{\rm vec}|\le\kappa_t$ and the two mass identities.
Equation \eqref{eq:B-corruption-final} uses
$\Gamma_t\le4c_{R}/\sqrt t$ and the filled-prefix inequality of
\Cref{lem:capped} (expanded below), valid pathwise for
$0\le\kappa_t\le2$ and $\sum_t\kappa_t\le C$.

The residual term
$\sum_t\E\left[\sum_a(p_{t,a}-q_{t,a})a^\top c_t^{\rm vec}\right]$ in
\eqref{eq:F-master-step-6} charges only the exploration perturbation
$p_t-q_t=w_t-\gamma_tq_t$. Its per-round absolute value and weighted
sum satisfy
\begin{subequations}\label{eq:F-direct-cor}
\begin{align}
 \left|\sum_a(p_{t,a}-q_{t,a})a^\top c_t^{\rm vec}\right|
 &\le\kappa_t\sum_a|w_{t,a}-\gamma_tq_{t,a}|
 \label{eq:F-direct-cor-step-1}\\
 &\le\kappa_t\left(\sum_aw_{t,a}
                      +\gamma_t\sum_aq_{t,a}\right)
 =2\gamma_t\kappa_t\le2\Gamma_t\kappa_t\,,
 \label{eq:F-direct-cor-step-2}\\
 \sum_{t=1}^T2\Gamma_t\kappa_t
 &\le8c_{R}\sum_{t=1}^T\frac{\kappa_t}{\sqrt t}
 \le8c_{R}\int_0^{C/2}\frac{2\,dx}{\sqrt x}
 =16\sqrt2\,c_{R}\sqrt C\,.
 \label{eq:F-direct-cor-step-3}
\end{align}
\end{subequations}
Equation \eqref{eq:F-direct-cor-step-1} substitutes $p_t-q_t=w_t-\gamma_tq_t$ and $|a^\top c_t^{\rm vec}|\le\kappa_t$. Equation \eqref{eq:F-direct-cor-step-2} applies the triangle inequality, $\sum_aw_{t,a}=\gamma_t$, and $\sum_aq_{t,a}=1$. Equation \eqref{eq:F-direct-cor-step-3} uses $\Gamma_t\le4c_{R}/\sqrt t$ and the filled-prefix calculation below; the integral equals $4\sqrt{C/2}=2\sqrt{2C}$.

For the integral bound in \eqref{eq:F-direct-cor}, the decreasing
weights $t^{-1/2}$ and constraints $0\le\kappa_t\le2$,
$\sum_t\kappa_t\le C$ make a filled prefix maximal.
More explicitly, for $m=\lfloor C/2\rfloor$ and
$r=C-2m\in[0,2)$,
\begin{subequations}\label{eq:rev-F-budget-prefix}
\begin{align}
 \sum_{t=1}^T\frac{\kappa_t}{\sqrt t}
 &\le2\sum_{j=1}^{m}j^{-1/2}+\frac r{\sqrt{m+1}}
 \label{eq:rev-F-budget-prefix-step-1}\\
 &\le2\int_0^m x^{-1/2}dx
       +2\int_m^{m+r/2}x^{-1/2}dx
 =2\sqrt{2C}\,.
 \label{eq:rev-F-budget-prefix-step-2}
\end{align}
\end{subequations}
Equation \eqref{eq:rev-F-budget-prefix-step-1} maximizes a decreasing weighted sum subject to $0\le\kappa_t\le2$ and total mass at most $C=2m+r$. Equation \eqref{eq:rev-F-budget-prefix-step-2} bounds each filled weight by the integral over its unit interval and the fractional weight by the integral over $[m,m+r/2]$, then evaluates $2\int_0^{C/2}x^{-1/2}dx$.

This remains valid for $m=0$, and all the inequalities are pathwise
for predictable adaptive corruption.
\end{proof}

\subsection{Auxiliary Lemmas}\label{app:G-proof}\label{app:finite-moment-tools}
This subsection collects inputs shared by the term proofs. In particular,
the penalty, stability, and exploration bounds use the same failure charge
$\mathcal E_T=\sum_{t\le T}t^{-1/2}\sP(\mathcal G_t^c)$. The window,
compensated inverse, negative, and full-history absolute moment estimates
below are all needed to control this common charge, so they remain here.
The conditional stability calculation is in
\Cref{app:B-stability}; empirical gap ratios, positive local
moments, allocation-mass sums, and the fallback design are in
\Cref{app:B-exploration}. The direct corruption calculation
is in \Cref{app:B-corruption}.

\subsubsection{Shared Cardinality Parameters}\label{app:finite-cardinality}
The shared finite-action identities and moment bounds below are used in
\Cref{prop:G-compatibility}. All cumulative loss estimates and reconstruction
rows below are derived from \IAFTRL{}. The scalar
moment estimates require only the displayed row conditions.

\begin{definition}[Cardinality analysis parameters]\label{def:AB-card-tuning}
Use the tuning specified in \Cref{sec:G-algorithm}, with
$c_{\beta},C_{\beta},c_{R}$ as in \eqref{eq:G-analysis-constants}.
For the corrupted-stochastic analysis, assume a bounded fixed loss vector $\theta$
with a unique optimal action $a^*$, minimum gap $\Delta_{\min}>0$, and ordinary pathwise
corruption budget $C$. Use
\begin{equation}\label{eq:F-analysis-tuning}
 P=2\lceil256+\log(2K)\rceil\,,\qquad
 \omega=\frac{255}{256}\,,\qquad k=K^{1/P}\,.
\end{equation}
Here $P$ is the even moment order used in the analysis;
$P\ge512$, $k^2<e<11/4$, and $B=2^{24}P\ge1024P$.
Also $\beta_t\le C_{\beta}\sqrt t$ and $d/\beta_t\le c_{R}/\sqrt t$.
\end{definition}

An analysis horizon can be extended by setting future corruption to zero
and future conditional means to $a^\top\theta$ at each action $a$. This leaves the budget
unchanged and permits full dyadic blocks in the proof.

\subsubsection{Shared Conditional Estimator Identities}
The program constrains candidate reconstructions before an action is
sampled. We next translate those constraints into properties of the
estimates actually produced by the learner: correct conditional means,
controlled second moments, and controlled update magnitudes. These
identities enter the regret decomposition and the shared loss-estimate moment
analysis. The calculation relating the proxy function to stability is proved separately
in \Cref{lem:B-centered-stability}.

\begin{lemma}[Executed reconstruction and conditional estimator identities]\label{lem:AB-estimator}
For \IAFTRL{}, define
$g_{t,a}=(a-r_t)^\top v_{t,A_t}Y_t$. Then
\begin{align}
 \E_{t-1}[g_{t,a}]&=(a-r_t)^\top\theta_t\,,
 &\E_{t-1}[(g_{t,a}-g_{t,c})]&=(a-c)^\top\theta_t\,,
 \notag\\
 \E_{t-1}[g_{t,a}^2]&\le\beta_t+Z_{t,a}^2/(Bt)\,,
 &|g_{t,a}|&\le\beta_t+Z_{t,a}/B\,.
 \notag
\end{align}
\end{lemma}
\begin{proof}[Proof of \Cref{lem:AB-estimator}]
All of $q_t,p_t,r_t,Z_t,v_t$ are $\cF_{t-1}$-measurable. Let
$\xi_{t,a}=a-r_t$ and
$m_{t,b}^{(r)}=\E[Y_t^r\mid\cF_{t-1},A_t=b]$ for $r\in\{1,2\}$.
The feedback assumptions give
\begin{equation}
 \label{eq:AB-feedback-conditional-moments}
 m_{t,b}^{(1)}=b^\top\theta_t\,,\qquad
 0\le m_{t,b}^{(2)}\le1\,,\qquad |Y_t|\le1\,.
\end{equation}
Because the sampling law is $\sP(A_t=b\mid\cF_{t-1})=p_{t,b}$,
conditioning on each realized action and then using reconstruction gives
\begin{align}
 \E_{t-1}[g_{t,a}]
 &=\sum_bp_{t,b}\E\!\left[
     \left.\xi_{t,a}^\top v_{t,b}Y_t
     \right|\cF_{t-1},A_t=b\right]\,,
 \label{eq:AB-est-mean-tower}\\
 &=\sum_bp_{t,b}(\xi_{t,a}^\top v_{t,b})
                      m_{t,b}^{(1)}
 =\sum_bp_{t,b}(\xi_{t,a}^\top v_{t,b})(b^\top\theta_t)\,,
 \label{eq:AB-est-mean-cancel}\\
 &=\xi_{t,a}^\top\!\left(\sum_bp_{t,b}v_{t,b}b^\top\right)\theta_t
 =\xi_{t,a}^\top I_d\theta_t=(a-r_t)^\top\theta_t\,,
 \label{eq:AB-est-mean-reconstruct}\\
 \E_{t-1}[(g_{t,a}-g_{t,c})]
 &=(a-r_t)^\top\theta_t-(c-r_t)^\top\theta_t
 =(a-c)^\top\theta_t\,.
 \label{eq:AB-est-mean-difference}
\end{align}
Equation \eqref{eq:AB-est-mean-tower} uses the conditional action law;
\eqref{eq:AB-est-mean-cancel} pulls the history-measurable reconstruction
outside the inner expectation and substitutes
\eqref{eq:AB-feedback-conditional-moments}. Equation
\eqref{eq:AB-est-mean-reconstruct} uses
$\sum_bp_{t,b}v_{t,b}b^\top=M(p_t)^{-1}M(p_t)=I_d$, which follows
from the prescribed formula $v_{t,b}=M(p_t)^{-1}b$ with $M(p_t)=\sum_{a\in\cA}p_{t,a}aa^\top$, and
\eqref{eq:AB-est-mean-difference} subtracts the two explicit means.

The corresponding second-moment calculation does not discard the
conditional feedback factor before it has been displayed:
\begin{align}
 \E_{t-1}[g_{t,a}^2]
 &=\sum_bp_{t,b}\E\!\left[
   \left.\left((\xi_{t,a}^\top v_{t,b})Y_t\right)^2
   \right|\cF_{t-1},A_t=b\right]\,,
 \label{eq:AB-est-second-tower}\\
 &=\sum_bp_{t,b}(\xi_{t,a}^\top v_{t,b})^2m_{t,b}^{(2)}\,,
 \label{eq:AB-est-second-substitute}\\
 &\le\sum_bp_{t,b}(\xi_{t,a}^\top v_{t,b})^2\cdot1\,,
 \label{eq:AB-est-second-bounded-feedback}\\
 &=\sup_{Q\in\mathcal Q}\E_{A\sim p_t,Y\sim Q_A}
     \left[\left.\left((\xi_{t,a}^\top v_{t,A})Y\right)^2
       \right|\cF_{t-1}\right]
 \le\beta_t+\frac{Z_{t,a}^2}{Bt}\,,
 \label{eq:AB-est-second-budget}\\
 |g_{t,a}|&=|\xi_{t,a}^\top v_{t,A_t}|\,|Y_t|
 \le|\xi_{t,a}^\top v_{t,A_t}|\,,
 \label{eq:AB-est-amplitude-observed}\\
 &\le\max_b|\xi_{t,a}^\top v_{t,b}|
 \le\beta_t+Z_{t,a}/B\,.
 \label{eq:rev-B-estimator-identities}
\end{align}
Equation \eqref{eq:AB-est-second-tower} conditions the squared update;
\eqref{eq:AB-est-second-substitute} squares its coefficient and retains
the sampling weight $p_{t,b}$. Equation \eqref{eq:AB-est-second-bounded-feedback}
uses $m_{t,b}^{(2)}\le1$. In \eqref{eq:AB-est-second-budget}, laws
supported on $\{-1,1\}$ attain the displayed supremum, and
\eqref{eq:G-program-var} bounds it. Equation
\eqref{eq:AB-est-amplitude-observed} uses $|Y_t|\le1$;
\eqref{eq:rev-B-estimator-identities} applies
\eqref{eq:G-program-mag} with the Dirac feedback laws $Q_b=\delta_1$
and the supremum over $b$.

\end{proof}

\subsubsection{Cumulative Loss-Estimate Identities and Row Conditions}\label{app:finite-score-interface}
The moment analysis uses scalar cumulative loss estimates relative to
the reference sequence, while the algorithm stores one vector $\sum_{s=1}^{t}\widehat\theta_s$.
The next lemma identifies these estimates and the reconstruction rows
obtained from that vector, and verifies the conditions
used below. Thus the scalar analysis concerns exactly the executed
learner.

\begin{lemma}[Cumulative loss estimates and row constraints]\label{lem:AB-score-interface}
The cumulative loss estimates and reconstruction rows defined in \eqref{eq:F-coherent-scores} reproduce
exactly the learner's exponential-weights distribution, reference, and updates in
\eqref{eq:F-scores} and \eqref{eq:F-update}. They satisfy
\eqref{eq:F-program-dom}--\eqref{eq:F-program-cert} and the objective
comparison \eqref{eq:F-objective-comparison}. No new cumulative loss-estimate process is run.
\end{lemma}
\begin{proof}[Proof of \Cref{lem:AB-score-interface}]
For the cumulative loss estimates $\widehat L_{t,a}$, reconstruction vectors $v_{t,b}$,
and reference $r_t$ generated by the algorithm, define
\begin{equation}
 \label{eq:F-coherent-scores}
 S_{t,a}=\widehat L_{t,a}-\sum_{s\le t}\widehat\ell_{s,r_s}\,,
 \qquad h_{t,a,b}=p_{t,b}(a-r_t)^\top v_{t,b}\,.
\end{equation}
The common shift in $S_t$ cancels in all estimated cumulative loss
differences and exponential-weights probabilities. Both $S_{t,a}$
and $\widehat L_{t,a}$ include round $t$; the decision at round $t$
uses their values at $t-1$. In particular,
$S_{t-1,a}=\widehat L_{t-1,a}-\sum_{s<t}\widehat\ell_{s,r_s}$.
Thus $S_0=0$ and the
quantities used by the learner obey
\begin{equation}
 \label{eq:F-scores}
 \begin{aligned}
 q_{t,a}&=\frac{e^{-\widehat L_{t-1,a}/\beta_t}}{\sum_b e^{-\widehat L_{t-1,b}/\beta_t}}
         =\frac{e^{-S_{t-1,a}/\beta_t}}{\sum_b e^{-S_{t-1,b}/\beta_t}}\,,\\
 r_t&=\argmin_a \widehat L_{t-1,a}=\argmin_a S_{t-1,a}\,,\\
 Z_{t,a}&=\widehat L_{t-1,a}-\widehat L_{t-1,r_t}=S_{t-1,a}-S_{t-1,r_t}\,,
 \qquad\widehat\Delta_t(a)=\min\{2,Z_{t,a}/t\}\,,
 \end{aligned}
\end{equation}
with the fixed ordering breaking ties. For the row budgets, define
\begin{equation*}
 U_{t,a}=\beta_t+Z_{t,a}/B\,,\qquad V_{t,a}=\beta_t+Z_{t,a}^2/(Bt)\,. \notag
\end{equation*}
Suppressing $t$, for a candidate $p$ let
$h_{a,b}=p_b(a-r_t)^\top v_b(p)$ and
$\bar h_b=\sum_aq_{t,a}h_{a,b}$.
The mixture is $p=(1-\gamma)q+w$ with $\gamma=\sum_bw_b$.
The constraints of \eqref{eq:G-program} imply
\begin{align}
 &w_b\ge0\,,\quad\gamma\le\Gamma_t\,,\label{eq:F-program-dom}\\
 &\sum_bh_{a,b}b=a-r_t\quad(a\in\cA)\,,\qquad h_{r_t,b}=0\,,\label{eq:F-program-rec}\\
 &|h_{a,b}|\le p_bU_{t,a}\,,\label{eq:F-program-mag}\\
 &\sum_b h_{a,b}^2/p_b\le V_{t,a}\,,\label{eq:F-program-var}\\
 &\beta_t\sum_bp_b\sum_aq_{t,a}
 \chi\!\left(\frac{h_{a,b}-\bar h_b}{\beta_t p_b}\right)\le2d/\beta_t\,,
 \qquad\bar h_b=\sum_aq_{t,a}h_{a,b}\,.\label{eq:F-program-cert}
\end{align}
Indeed, $\sum_bp_bv_b(p)b^\top=I_d$ gives the row reconstruction identity,
and the remaining inequalities are the equivalent finite forms of the
learner's constraints. For these prescribed rows, the LHS
of \eqref{eq:F-program-cert} is exactly $\gC_t(p)$. The update gives
\begin{equation}
 \label{eq:F-update}
 g_{t,a}=\frac{h_{t,a,A_t}}{p_{t,A_t}}Y_t\,,\qquad
 S_{t,a}=S_{t-1,a}+g_{t,a}\,.
\end{equation}
For the coherent learner,
$g_{t,a}=\widehat d_{t,a}=\widehat\ell_{t,a}-\widehat\ell_{t,r_t}$
equals $d_t(a;v_{t,A_t},Y_t)$.
Every quotient is defined because
$p_b\ge(1-\Gamma_t)q_{t,b}\ge3q_{t,b}/4>0$.
These are identities for the same executed process; $S_t$ and $h_t$ are
notation for its analysis.

The scalar moment lemmas below use only \eqref{eq:F-scores},
\eqref{eq:F-program-dom}--\eqref{eq:F-program-cert}, and
\eqref{eq:F-update}, with zero initial cumulative loss estimates. The exploration-cost
bound additionally uses the objective comparison
\begin{equation}
 \label{eq:F-objective-comparison}
 \langle p_t,\widehat\Delta_t\rangle
 \le\langle p^c,\widehat\Delta_t\rangle
\end{equation}
for every feasible sampling law $p^c$ with its prescribed covariance
reconstruction $v_b^c=v_b(p^c)=\Sigma(p^c)^{-1}b$ in \eqref{eq:G-program}.
This follows directly from the minimization in \eqref{eq:G-program}.
The G-optimal fallback is feasible by \Cref{lem:AB-fallback};
\Cref{lem:F-witness} below verifies the gap-adapted covariance
comparison in the cardinality branch.

The row identities and cancellation of the common reference-loss
estimate can also be checked without introducing a second sequence of
loss estimates. Let
$c_{t-1}=\sum_{s<t}\widehat\ell_{s,r_s}$ and
$\bar a_t=\sum_aq_{t,a}a$. Then
\begin{align}
 S_{t-1,a}-S_{t-1,b}
 &=(\widehat L_{t-1,a}-c_{t-1})
   -(\widehat L_{t-1,b}-c_{t-1})=\widehat L_{t-1,a}-\widehat L_{t-1,b}\,,
 \label{eq:AB-score-difference-expanded}\\
 \frac{e^{-S_{t-1,a}/\beta_t}}{\sum_b e^{-S_{t-1,b}/\beta_t}}
 &=\frac{e^{c_{t-1}/\beta_t}e^{-\widehat L_{t-1,a}/\beta_t}}
         {e^{c_{t-1}/\beta_t}\sum_b e^{-\widehat L_{t-1,b}/\beta_t}}
 =\frac{e^{-\widehat L_{t-1,a}/\beta_t}}
        {\sum_b e^{-\widehat L_{t-1,b}/\beta_t}}\,,
 \label{eq:AB-score-Gibbs-common-factor}\\
 \sum_bh_{t,a,b}b
 &=\sum_bp_{t,b}bv_{t,b}^\top(a-r_t)
 =\left(\sum_bp_{t,b}v_{t,b}b^\top\right)^\top(a-r_t)
 =a-r_t\,,
 \label{eq:AB-score-row-reconstruction}\\
 \bar h_{t,b}
 &=p_{t,b}\sum_aq_{t,a}(a-r_t)^\top v_{t,b}
 =p_{t,b}(\bar a_t-r_t)^\top v_{t,b}\,,
 \label{eq:AB-score-centered-row-mean}\\
 h_{t,a,b}-\bar h_{t,b}
 &=p_{t,b}(a-r_t)^\top v_{t,b}
    -p_{t,b}(\bar a_t-r_t)^\top v_{t,b}
 =p_{t,b}(a-\bar a_t)^\top v_{t,b}\,,
 \label{eq:AB-score-centered-row-difference}\\
 S_{t,a}-S_{t-1,a}
 &=\widehat\ell_{t,a}-\widehat\ell_{t,r_t}
 =(a-r_t)^\top v_{t,A_t}Y_t
 =\frac{h_{t,a,A_t}}{p_{t,A_t}}Y_t\,.
 \label{eq:AB-score-update-substitution}
\end{align}
Equation \eqref{eq:AB-score-difference-expanded} cancels the common
shift in differences; \eqref{eq:AB-score-Gibbs-common-factor} cancels
its exponential factor in the exponential-weights distribution. Equation
\eqref{eq:AB-score-row-reconstruction} transposes the executed
matrix reconstruction. Equations \eqref{eq:AB-score-centered-row-mean}
and \eqref{eq:AB-score-centered-row-difference} identify exactly the
centered row appearing in the proxy function. Equation
\eqref{eq:AB-score-update-substitution} substitutes the vector update
and the sampled reconstruction before identifying $g_{t,a}$.
\end{proof}

The next lemma gives the scalar form of the conditional identities in
\Cref{lem:AB-estimator}. They are common inputs to the regret
decomposition and the moment estimates. The scalar stability bound using the proxy function
is proved in \Cref{lem:B-scalar-stability} in
\Cref{app:B-stability}.

\begin{lemma}[Estimator identities]\label{lem:F-estimator}
For all coordinates,
\[
 \E_{t-1}[g_{t,a}]=(a-r_t)^\top\theta_t\,,\quad
 \E_{t-1}[g_{t,a}-g_{t,b}]=(a-b)^\top\theta_t\,,
 \quad\E_{t-1}[g_{t,a}^2]\le V_{t,a}\,,\quad |g_{t,a}|\le U_{t,a}\,.
\]
\end{lemma}
\begin{proof}[Proof of \Cref{lem:F-estimator}]
Condition on $\mathcal F_{t-1}$ and abbreviate
$p_b=p_{t,b}$, $h_{a,b}=h_{t,a,b}$, $q_a=q_{t,a}$, and $\beta=\beta_t$.
The reconstruction constraint and the conditional feedback mean give
\begin{subequations}\label{eq:rev-F-estimator-calculation}
\begin{align}
 \E_{t-1}[g_{t,a}]
 &=\sum_bp_b\frac{h_{a,b}}{p_b}
                 \E[Y_t\mid\mathcal F_{t-1},A_t=b]
 =\sum_bh_{a,b}b^\top\theta_t
 \label{eq:rev-F-estimator-calculation-step-1}\\
 &=\left(\sum_bh_{a,b}b\right)^\top\theta_t
 =(a-r_t)^\top\theta_t\,,
 \label{eq:rev-F-estimator-calculation-step-2}\\
 \E_{t-1}[g_{t,a}-g_{t,b}]
 &=(a-r_t)^\top\theta_t-(b-r_t)^\top\theta_t
 =(a-b)^\top\theta_t\,,
 \label{eq:rev-F-estimator-calculation-step-3}\\
 \E_{t-1}[g_{t,a}^2]
 &=\sum_bp_b\E\!\left[\left.
                \left(\frac{h_{a,b}}{p_b}Y_t\right)^2
                     \right|\mathcal F_{t-1},A_t=b\right]\,,
 \label{eq:rev-F-estimator-calculation-step-4}\\
 &=\sum_bp_b\frac{h_{a,b}^2}{p_b^2}
                   \E[Y_t^2\mid\mathcal F_{t-1},A_t=b]
 =\sum_b\frac{h_{a,b}^2}{p_b}
                   \E[Y_t^2\mid\mathcal F_{t-1},A_t=b]\,,
 \label{eq:rev-F-estimator-calculation-step-5}\\
 &\le\sum_b\frac{h_{a,b}^2}{p_b}\cdot1
 \le V_{t,a}=\beta_t+Z_{t,a}^2/(Bt)\,,
 \label{eq:rev-F-estimator-calculation-step-6}\\
 |g_{t,a}|&=\frac{|h_{a,A_t}|}{p_{A_t}}|Y_t|
 \le\frac{|h_{a,A_t}|}{p_{A_t}}
 \le\max_b\frac{|h_{a,b}|}{p_b}
 \le U_{t,a}=\beta_t+Z_{t,a}/B\,.
 \label{eq:rev-F-estimator-calculation-step-7}
\end{align}
\end{subequations}
Equation \eqref{eq:rev-F-estimator-calculation-step-1} conditions on the sampled action and cancels $p_b$ before using the feedback mean. Equation \eqref{eq:rev-F-estimator-calculation-step-2} uses the reconstruction identity $\sum_bh_{a,b}b=a-r_t$; \eqref{eq:rev-F-estimator-calculation-step-3} subtracts the two unbiased identities. Equation \eqref{eq:rev-F-estimator-calculation-step-4} substitutes $g_{t,a}=h_{a,A_t}Y_t/p_{A_t}$ directly. Equation \eqref{eq:rev-F-estimator-calculation-step-5} extracts the predictable coefficient and cancels one power of $p_b$. Equation \eqref{eq:rev-F-estimator-calculation-step-6} uses $Y_t^2\le1$ and the row-variance constraint. Equation \eqref{eq:rev-F-estimator-calculation-step-7} first uses $|Y_t|\le1$, then maximizes over the possible sampled column and applies the update magnitude constraint.

\end{proof}

\subsubsection{The Shared Regret Decomposition}
For ordinary corruption, the scalar regret decomposition must retain the
actual fixed comparator while isolating the extra cost of exploration.
The next lemma isolates all four terms. Their individual bounds are
proved in \Crefrange{app:B-penalty}{app:B-corruption}.

\begin{lemma}[Actual-comparator scalar regret decomposition]\label{lem:AB-card-master}
For a fixed comparator distribution $u\in\gP(\cA)$, the actual
corrupted regret satisfies \eqref{eq:F-master}. This decomposition retains
the actual fixed comparator; the bound on its direct corruption term is
proved in \Cref{lem:B-corruption-term}.
\end{lemma}
\begin{proof}[Proof of \Cref{lem:AB-card-master}]
For a fixed comparator distribution $u$, let
$\bar a_t=\sum_aq_{t,a}a$. The common reference cancels because
$\sum_a(q_{t,a}-u_a)=0$. Therefore
\begin{subequations}\label{eq:rev-F-comparator-decomposition}
\begin{align}
 \E[\langle q_t-u,g_t\rangle]
 &=\E\left[\sum_a(q_{t,a}-u_a)(a-r_t)^\top\theta_t\right]
 =\E\left[\sum_a(q_{t,a}-u_a)a^\top\theta_t\right]\,,
 \label{eq:rev-F-comparator-decomposition-step-1}\\
 \sum_a(p_{t,a}-q_{t,a})a^\top\theta_t
 &=\sum_a(p_{t,a}-q_{t,a})
        \{a^\top\theta+a^\top c_t^{\rm vec}\}
 \label{eq:rev-F-comparator-decomposition-step-2}\\
 &=\langle p_t-q_t,\Delta\rangle
    +\sum_a(p_{t,a}-q_{t,a})a^\top c_t^{\rm vec}\,.
 \label{eq:rev-F-comparator-decomposition-step-3}
\end{align}
\end{subequations}
Equation \eqref{eq:rev-F-comparator-decomposition-step-1} uses conditional unbiasedness and $\sum_a(q_{t,a}-u_a)=0$, which removes $r_t^\top\theta_t$. Equation \eqref{eq:rev-F-comparator-decomposition-step-2} substitutes $\theta_t=\theta+c_t^{\rm vec}$ without bounding either summand. Equation \eqref{eq:rev-F-comparator-decomposition-step-3} substitutes $a^\top\theta=(a^*)^{\top}\theta+\Delta(a)$ and cancels $(a^*)^{\top}\theta\sum_a(p_{t,a}-q_{t,a})=0$.

 Substituting
\eqref{eq:rev-F-comparator-decomposition} into regret, and then
\eqref{eq:entropy-local} in \Cref{lem:entropy-FTRL} and \eqref{eq:F-stability} in \Cref{lem:B-scalar-stability}, gives
\begin{subequations}\label{eq:F-master}
\begin{align}
 \Reg(u)
 &=\E\left[\sum_{t=1}^T\langle q_t-u,g_t\rangle\right]
   +\sum_{t=1}^T\E[\langle p_t-q_t,\Delta\rangle]
 \label{eq:F-master-step-1}\\
 &\qquad+\sum_{t=1}^T\E\left[\sum_a
                  (p_{t,a}-q_{t,a})a^\top c_t^{\rm vec}\right]
 \label{eq:F-master-step-2}\\
 &\le\beta_1\log K+
       \sum_{t=2}^T(\beta_t-\beta_{t-1})\E[H(q_t)]
       +\sum_{t=1}^T\E[D_t]
 \label{eq:F-master-step-3}\\
 &\qquad+\sum_{t=1}^T\E[\langle p_t-q_t,\Delta\rangle]
   +\sum_{t=1}^T\E\left[\sum_a
                  (p_{t,a}-q_{t,a})a^\top c_t^{\rm vec}\right]
 \label{eq:F-master-step-4}\\
 &\le\beta_1\log K+
       \sum_{t=2}^T(\beta_t-\beta_{t-1})\E[H(q_t)]
       +\sum_{t=1}^T\E[\gC_t(p_t)]
 \label{eq:F-master-step-5}\\
 &\qquad+\sum_{t=1}^T\E[\langle p_t-q_t,\Delta\rangle]
   +\sum_{t=1}^T\E\left[\sum_a
                  (p_{t,a}-q_{t,a})a^\top c_t^{\rm vec}\right]\,.
 \label{eq:F-master-step-6}
\end{align}
\end{subequations}
Equations \eqref{eq:F-master-step-1} and \eqref{eq:F-master-step-2} are the exact regret decomposition after the unbiased estimator identities are substituted. Equations \eqref{eq:F-master-step-3} and \eqref{eq:F-master-step-4} apply the local entropy-FTRL inequality to $\sum_t\langle q_t-u,g_t\rangle$, leaving both exploration terms unchanged. Equations \eqref{eq:F-master-step-5} and \eqref{eq:F-master-step-6} use $\E[D_t]=\E[\E_{t-1}[D_t]]\le\E[\gC_t(p_t)]$ for every $t$.

\end{proof}

\subsubsection{Uniform Bounds on Cumulative Entropy and Stability Costs}
The global entropy and stability bounds alone give a square-root regret
rate. On histories where the cumulative loss estimates separate $a^*$ from the suboptimal actions, the FTRL
distribution concentrates and both costs decay much faster. The next
lemma quantifies this decay for the optimizer's own estimates, while
retaining bounds valid on the remaining histories.

\begin{lemma}[The optimizer's own entropy and proxy function]\label{lem:F-good-stability}
On $\mathcal G_t$,
\[
 H(q_t)\le2K e^{-2t^{\alpha-1/2}/C_{\beta}}\,,\qquad
 \gC_t(p_t)\le e^2C_{\beta}K\sqrt t\,e^{-2t^{\alpha-1/2}/C_{\beta}}\,.
\]
On every history, $(\beta_t-\beta_{t-1})H(q_t)\le c_{R}/\sqrt t$ for $t\ge2$ and $\gC_t(p_t)\le2c_{R}/\sqrt t$.
\end{lemma}
\begin{proof}
On $\mathcal G_t$, set $x_a=X_{t,a}/\beta_t$ and
$\mathcal Z=1+\sum_{a\ne a^*}e^{-x_a}$. Then $x_{a^*}=0$ and
$x_a\ge4t^{\alpha-1/2}/C_{\beta}$ for $a\ne a^*$. Direct substitution of
$q_{t,a}=e^{-x_a}/\mathcal Z$ gives
\begin{subequations}\label{eq:rev-F-good-entropy}
\begin{align}
 H(q_t)
 &=-\sum_aq_{t,a}\log(e^{-x_a}/\mathcal Z)
 =\log\mathcal Z+\sum_{a\ne a^*}q_{t,a}x_a
 \label{eq:rev-F-good-entropy-step-1}\\
 &\le\sum_{a\ne a^*}e^{-x_a}
             +\sum_{a\ne a^*}x_ae^{-x_a}
 \le2\sum_{a\ne a^*}e^{-x_a/2}
 \le2K e^{-2t^{\alpha-1/2}/C_{\beta}}\,.
 \label{eq:rev-F-good-entropy-step-2}
\end{align}
\end{subequations}
Equation \eqref{eq:rev-F-good-entropy-step-1} substitutes $q_{t,a}=e^{-x_a}/\mathcal Z$ directly and uses $x_{a^*}=0$. Equation \eqref{eq:rev-F-good-entropy-step-2} uses $\log(1+z)\le z$, $\mathcal Z\ge1$, $(1+x)e^{-x}\le2e^{-x/2}$ for $x\ge0$, and $x_a\ge4t^{\alpha-1/2}/C_{\beta}$ on $\mathcal G_t$.

For the actual selected reconstruction, $r_t=a^*$ implies
$h_{t,a^*,b}=0$. Applying \eqref{eq:chi-center} in \Cref{lem:chi} to each column and
then substituting the upper bound $\beta_t+X_{t,a}/B$ gives
\begin{subequations}\label{eq:rev-F-good-actual-certificate}
\begin{align}
 \gC_t(p_t)
 &=\beta_t\sum_bp_{t,b}\sum_aq_{t,a}
       \chi\!\left(\frac{h_{t,a,b}-\bar h_{t,b}}
                           {\beta_tp_{t,b}}\right)
 \label{eq:rev-F-good-actual-certificate-step-1}\\
 &\le\beta_t\sum_bp_{t,b}\sum_{a\ne a^*}q_{t,a}
       \chi\!\left(\frac{2h_{t,a,b}}{\beta_tp_{t,b}}\right)
 \label{eq:rev-F-good-actual-certificate-step-2}\\
 &\le\beta_t\sum_bp_{t,b}\sum_{a\ne a^*}q_{t,a}
             \exp\!\left(\frac{2(\beta_t+X_{t,a}/B)}{\beta_t}\right)
 \label{eq:rev-F-good-actual-certificate-step-3}\\
 &=e^2\beta_t\sum_{a\ne a^*}q_{t,a}e^{2X_{t,a}/(B\beta_t)}
 \label{eq:rev-F-good-actual-certificate-step-4}\\
 &\le e^2\beta_t\sum_{a\ne a^*}
        e^{-(1-2/B)X_{t,a}/\beta_t}
 \le e^2\beta_t\sum_{a\ne a^*}e^{-X_{t,a}/(2\beta_t)}
 \label{eq:rev-F-good-actual-certificate-step-5}\\
 &\le e^2C_{\beta}K\sqrt t\,e^{-2t^{\alpha-1/2}/C_{\beta}}\,.
 \label{eq:rev-F-good-actual-certificate-step-6}
\end{align}
\end{subequations}
Equation \eqref{eq:rev-F-good-actual-certificate-step-1} is the selected proxy function. Equation \eqref{eq:rev-F-good-actual-certificate-step-2} applies the centering inequality of \Cref{lem:chi} column by column and uses $h_{t,a^*,b}=0$. Equation \eqref{eq:rev-F-good-actual-certificate-step-3} uses $\chi(x)\le e^{|x|}$ and $|h_{t,a,b}|/p_{t,b}\le\beta_t+X_{t,a}/B$. Equation \eqref{eq:rev-F-good-actual-certificate-step-4} sums $p_{t,b}$ to one. Equation \eqref{eq:rev-F-good-actual-certificate-step-5} substitutes $q_{t,a}\le e^{-X_{t,a}/\beta_t}$ and $1-2/B\ge1/2$. Equation \eqref{eq:rev-F-good-actual-certificate-step-6} uses $X_{t,a}\ge4t^\alpha$ and $\beta_t\le C_{\beta}\sqrt t$.

Here $\chi(0)=0$, $\chi(x)\le e^{|x|}$, $B\ge4$, and
$\sum_bp_{t,b}=1$ justify the successive lines.
On arbitrary histories the same selected proxy function and entropy obey
\begin{subequations}\label{eq:rev-F-global-entropy-certificate}
\begin{align}
 (\beta_t-\beta_{t-1})H(q_t)
 &\le\frac{c_{\beta}}{\sqrt t}\log K
 \le\frac{c_{\beta}\log(eK)}{\sqrt t}=\frac{c_{R}}{\sqrt t}\quad(t\ge2)\,,
 \label{eq:rev-F-global-entropy-certificate-step-1}\\
 \gC_t(p_t)&\le\frac{2d}{\beta_t}
                         \le\frac{2c_{R}}{\sqrt t}\,.
 \label{eq:rev-F-global-entropy-certificate-step-2}
\end{align}
\end{subequations}
Equation \eqref{eq:rev-F-global-entropy-certificate-step-1} uses $\beta_t-\beta_{t-1}=c_{\beta}/(\sqrt t+\sqrt{t-1})\le c_{\beta}/\sqrt t$, $H(q_t)\le\log K\le \log(eK)$, and $c_{\beta}\log(eK)=c_{R}$. Equation \eqref{eq:rev-F-global-entropy-certificate-step-2} uses the proxy function constraint and $\beta_t\ge c_{\beta}\sqrt t$, with $d/c_{\beta}=c_{R}$.

\end{proof}

The preceding lemma gives one-round decay estimates. We next sum them
over time, together with the exponential-weights tail that appears in
the exploration comparison. The resulting constants collect these contributions into
finite transients in the regret bound for a fixed allocation.

\begin{lemma}[Finite total good-history costs]\label{lem:AB-good-sums}
Let $S_a,S_b,c_{\rm tail}$ be defined by \eqref{eq:F-good-sums}.
Then the good-event entropy and proxy function sum is at most $S_a$, and
\begin{equation*}
 800K\sum_{t\ge T_\circ}\Psi_t e^{-4t^{\alpha-1/2}/C_{\beta}}
 \le S_b\Psi_{\Delta_{\min}}\,. \notag
\end{equation*}
The full inequality for entropy and the proxy function is
\eqref{eq:rev-F-all-good-sums}.
\end{lemma}
\begin{proof}[Proof of \Cref{lem:AB-good-sums}]
Define
\begin{equation}
 \label{eq:F-good-sums}
 S_a=48c_{\beta}KC_{\beta}^4+24e^2KC_{\beta}^7\,,\qquad S_b=19200KC_{\beta}^4\,,
 \qquad c_{\rm tail}=(224u_0k)^{\gamma_0}\,.
\end{equation}

For completeness, monotonicity and the change of variable
$y=x^{1/4}/C_{\beta}$ give
\begin{subequations}\label{eq:rev-F-exponential-sums}
\begin{align}
 \sum_{t\ge1}e^{-t^{1/4}/C_{\beta}}
 &\le\int_0^\infty e^{-x^{1/4}/C_{\beta}}\,dx
 =4C_{\beta}^4\int_0^\infty y^3e^{-y}\,dy=24C_{\beta}^4\,,
 \label{eq:rev-F-exponential-sums-step-1}\\
 \sqrt t\,e^{-2t^{\alpha-1/2}/C_{\beta}}
 &\le\sqrt t\,e^{-2t^{1/4}/C_{\beta}}
 =C_{\beta}^2\left(\frac{t^{1/4}}{C_{\beta}}\right)^2
                 e^{-2t^{1/4}/C_{\beta}}
 \le C_{\beta}^2e^{-t^{1/4}/C_{\beta}}\,.
 \label{eq:rev-F-exponential-sums-step-2}
\end{align}
\end{subequations}
Equation \eqref{eq:rev-F-exponential-sums-step-1} compares the decreasing summand with its integral and substitutes $x=C_{\beta}^4y^4$, so $dx=4C_{\beta}^4y^3dy$. Equation \eqref{eq:rev-F-exponential-sums-step-2} uses $\alpha-1/2\ge1/4$, writes $\sqrt t=C_{\beta}^2(t^{1/4}/C_{\beta})^2$, and uses $y^2e^{-y}\le1$.

 Substituting
\eqref{eq:rev-F-good-entropy} and
\eqref{eq:rev-F-good-actual-certificate} in \Cref{lem:F-good-stability}, with
$\alpha-1/2\ge1/4$, yields the promised sums
\begin{subequations}\label{eq:rev-F-all-good-sums}
\begin{align}
 &\sum_{t=2}^{\infty}(\beta_t-\beta_{t-1})
                   \E[\ind_{\mathcal G_t}H(q_t)]
   +\sum_{t=1}^{\infty}\E[\ind_{\mathcal G_t}\gC_t(p_t)]
 \label{eq:rev-F-all-good-sums-step-1}\\
 &\qquad\le2c_{\beta}K\sum_{t\ge1}e^{-t^{1/4}/C_{\beta}}
              +e^2C_{\beta}^3K\sum_{t\ge1}e^{-t^{1/4}/C_{\beta}}
 \label{eq:rev-F-all-good-sums-step-2}\\
 &\qquad\le48c_{\beta}KC_{\beta}^4+24e^2KC_{\beta}^7=S_a\,,
 \label{eq:rev-F-all-good-sums-step-3}\\
 &800K\sum_{t\ge T_\circ}\Psi_t e^{-4t^{\alpha-1/2}/C_{\beta}}
 \le800K\Psi_{\Delta_{\min}}\sum_{t\ge1}e^{-t^{1/4}/C_{\beta}}
 \le19200KC_{\beta}^4\Psi_{\Delta_{\min}}=S_b\Psi_{\Delta_{\min}}\,.
 \label{eq:rev-F-all-good-sums-step-4}
\end{align}
\end{subequations}
Equation \eqref{eq:rev-F-all-good-sums-step-1} identifies both good-history sums being bounded. Equation \eqref{eq:rev-F-all-good-sums-step-2} substitutes the entropy and proxy function envelopes and the pointwise exponential domination. Equation \eqref{eq:rev-F-all-good-sums-step-3} inserts $\sum_{t\ge1}e^{-t^{1/4}/C_{\beta}}\le24C_{\beta}^4$ into both terms. Equation \eqref{eq:rev-F-all-good-sums-step-4} separately substitutes $\Psi_t\le\Psi_{\Delta_{\min}}$ and the same exponential sum in the exploration contribution.

\end{proof}

\subsubsection{Weighted Good-Event Failure Sum}
The penalty, stability, and exploration bounds all charge failures of
$\mathcal G_t$ through the same sum
$\sum_t t^{-1/2}\sP(\mathcal G_t^c)$. The following estimate controls this
shared cost. The exploration-specific allocation-mass sums are proved in
\Cref{lem:B-mass-sums}.

\begin{lemma}[Weighted good-event failure sum]\label{lem:F-sums}
\begin{align}
 \sum_{t\ge1}t^{-1/2}\sP(\mathcal G_t^c)
 \le{}&2\sqrt{T_\circ}+3\mathsf A+130\sqrt{n_0}+2K56^P
       +3c_{\rm tail}\Delta_{\min}^{-\gamma_0}\sqrt{n_0}\,. \notag
\end{align}
\end{lemma}
\begin{proof}
For $t\ge T_\circ$, cap each nonnegative term in
\eqref{eq:F-good-tail} in \Cref{lem:B-good-event-tail} at one. Indeed, if one uncapped term is at
least one, its capped term already bounds a probability; if all are
below one, capping changes nothing. For the prefix and the four tails,
denote the five resulting weighted sums by $B_0,\ldots,B_4$.
Then
\begin{subequations}\label{eq:rev-F-first-three-tail-sums}
\begin{align}
 B_0&=\sum_{1\le t<T_\circ}t^{-1/2}
 \le\int_0^{T_\circ}x^{-1/2}dx=2\sqrt{T_\circ}\,,
 \label{eq:rev-F-first-three-tail-sums-step-1}\\
 B_1&\le\mathsf A\sum_{t\ge1}t^{-1/2-65/64}
 \le\mathsf A\left(1+\int_1^\infty x^{-97/64}dx\right)
 =\mathsf A(1+64/33)<3\mathsf A\,,
 \label{eq:rev-F-first-three-tail-sums-step-2}\\
 B_2&\le\sum_{t\ge1}t^{-1/2}\min\{1,(n_0/t)^{r_0}\}
 \le\left(2+\frac1{r_0-1/2}\right)\sqrt{n_0}
 \le130\sqrt{n_0}\,.
 \label{eq:rev-F-first-three-tail-sums-step-3}
\end{align}
\end{subequations}
Equation \eqref{eq:rev-F-first-three-tail-sums-step-1} integrates the prefix. Equation \eqref{eq:rev-F-first-three-tail-sums-step-2} uses $1/2+65/64=97/64$ and evaluates the tail integral. Equation \eqref{eq:rev-F-first-three-tail-sums-step-3} applies \Cref{lem:capped} with $(A,n,p)=(1,n_0,r_0)$ and uses $r_0-1/2\ge1/128$.

For the third tail, keep the factor $\Delta_{\min}^{-P}$ inside the time-tail
calculation:
\begin{subequations}\label{eq:rev-F-pure-gap-tail-sum}
\begin{align}
 B_3
 &\le K56^P\Delta_{\min}^{-P}\sum_{t\ge T_\circ}t^{-\zeta-1/2}
 \label{eq:rev-F-pure-gap-tail-sum-step-1}\\
 &\le K56^P\Delta_{\min}^{-P}
       \left\{T_\circ^{-\zeta-1/2}
                   +\int_{T_\circ}^\infty x^{-\zeta-1/2}dx\right\}
 \label{eq:rev-F-pure-gap-tail-sum-step-2}\\
 &\le2K56^P\Delta_{\min}^{-P}T_\circ^{1/2-\zeta}
 \le2K56^P\Delta_{\min}^{-(P+5(1/2-\zeta))}
 \label{eq:rev-F-pure-gap-tail-sum-step-3}\\
 &=2K56^P\Delta_{\min}^{-(P+5(1/2-P/4+13/10))}
 =2K56^P\Delta_{\min}^{P/4-9}\le2K56^P\,.
 \label{eq:rev-F-pure-gap-tail-sum-step-4}
\end{align}
\end{subequations}
Equation \eqref{eq:rev-F-pure-gap-tail-sum-step-1} keeps the full factor $\Delta_{\min}^{-P}$ in the gap tail. Equation \eqref{eq:rev-F-pure-gap-tail-sum-step-2} compares the decreasing sum with its first term plus its integral. Equation \eqref{eq:rev-F-pure-gap-tail-sum-step-3} uses $\zeta>3/2$ and $T_\circ\ge \Delta_{\min}^{-5}$. Equation \eqref{eq:rev-F-pure-gap-tail-sum-step-4} substitutes $\zeta=P/4-13/10$, computes $P+5(1/2-\zeta)=9-P/4$, and uses $P\ge512$ and $\Delta_{\min}^{-1}\ge1$.

Here $T_\circ\ge\Delta_{\min}^{-5}$, $\zeta>3/2$, $P\ge512$, and $\Delta_{\min}^{-1}\ge1$.
For the fourth tail, \eqref{eq:capped} in \Cref{lem:capped} with
$(A,n,p)=(K(224u_0\Delta_{\min}^{-1})^P,n_0,\omega P)$ gives
\begin{subequations}\label{eq:rev-F-fourth-tail-sum}
\begin{align}
 B_4
 &\le\left(2+\frac1{\omega P-1/2}\right)
       \{K(224u_0\Delta_{\min}^{-1})^P\}^{1/(2\omega P)}\sqrt{n_0}
 \label{eq:rev-F-fourth-tail-sum-step-1}\\
 &\le3\{K^{1/P}224u_0\Delta_{\min}^{-1}\}^{1/(2\omega)}\sqrt{n_0}
 =3(224u_0k)^{\gamma_0}\Delta_{\min}^{-\gamma_0}\sqrt{n_0}
 \label{eq:rev-F-fourth-tail-sum-step-2}\\
 &=3c_{\rm tail}\Delta_{\min}^{-\gamma_0}\sqrt{n_0}\,.
 \label{eq:rev-F-fourth-tail-sum-step-3}
\end{align}
\end{subequations}
Equation \eqref{eq:rev-F-fourth-tail-sum-step-1} applies \Cref{lem:capped} with $(A,n,p)=(K(224u_0\Delta_{\min}^{-1})^P,n_0,\omega P)$. Equation \eqref{eq:rev-F-fourth-tail-sum-step-2} uses $\omega P-1/2\ge1$, $K^{1/P}=k$, and $1/(2\omega)=\gamma_0$. Equation \eqref{eq:rev-F-fourth-tail-sum-step-3} substitutes $c_{\rm tail}=(224u_0k)^{\gamma_0}$.

Substitution of all five bounds proves
\begin{subequations}\label{eq:rev-F-total-bad-sum}
\begin{align}
 \sum_{t\ge1}t^{-1/2}\sP(\mathcal G_t^c)
 &\le B_0+B_1+B_2+B_3+B_4
 \label{eq:rev-F-total-bad-sum-step-1}\\
 &\le2\sqrt{T_\circ}+3\mathsf A+130\sqrt{n_0}+2K56^P
                 +3c_{\rm tail}\Delta_{\min}^{-\gamma_0}\sqrt{n_0}\,.
 \label{eq:rev-F-total-bad-sum-step-2}
\end{align}
\end{subequations}
Equation \eqref{eq:rev-F-total-bad-sum-step-1} is the union-bound decomposition into the prefix and four failure tails. Equation \eqref{eq:rev-F-total-bad-sum-step-2} substitutes all five already evaluated sums; no component is suppressed in the final estimate.

\end{proof}

\subsubsection{The Common Good Event and Its Failure Probability}
\begin{definition}[Predictable scalar good event]\label{def:AB-card-good}
For $t\ge T_\circ$ define
\[
 \mathcal G_t=\{I_t=1\}\cap\bigcap_{a\ne a^*}\{W_{t,a}\ge4\tau_t\}\,,
\]
and set $\mathcal G_t=\varnothing$ for earlier times. The resulting separation of the cumulative loss estimates is proved in \Cref{lem:B-good-event-tail} below.
\end{definition}

The penalty and stability terms use separation of the cumulative loss estimates on this event;
all three of the first terms use its failure probability. The following
lemma supplies both common inputs. The ratios needed only for exploration
are treated in \Cref{lem:F-ratios}.

\begin{lemma}[Good-event probability and separation of cumulative loss estimates]\label{lem:B-good-event-tail}
For $t\ge T_\circ$,
\begin{align}\label{eq:F-good-tail}
 \sP(\mathcal G_t^c)\le{}&\mathsf A t^{-65/64}+(n_0/t)^{r_0}
  +K(56\Delta_{\min}^{-1})^Pt^{-\zeta}
  +K(224u_0\Delta_{\min}^{-1})^P(n_0/t)^{\omega P}\,.
\end{align}
On $\mathcal G_t$, $r_t=a^*$, $M_t=0$, and
\[
 Z_{t,a}=X_{t,a}\ge4\tau_t\ge4t^\alpha\quad(a\ne a^*)\,.
\]
\end{lemma}
\begin{proof}
On $\{I_t=1,W_{t,a}<4\tau_t\}$, the identity
\eqref{eq:F-Y} in \Cref{lem:AB-card-compensation} and $0\le\Lambda_{t,a}\le4C\le\tau_t$ imply
$0<Y_{t,a}^\dagger<7\tau_t$. Applying Markov's inequality to
$(Y_{t,a}^\dagger)^{-P}$ and then \eqref{eq:F-inverse} in \Cref{lem:F-local} gives
\begin{subequations}\label{eq:rev-F-arm-gap-tail}
\begin{align}
 \sP(I_t=1,W_{t,a}<4\tau_t)
 &\le\sP(I_t=1,(Y_{t,a}^\dagger)^{-P}>(7\tau_t)^{-P})
 \label{eq:rev-F-arm-gap-tail-step-1}\\
 &\le(7\tau_t)^P\E[I_t(Y_{t,a}^\dagger)^{-P}]
 \le\left(\frac{28\tau_t}{t\Delta(a)}\right)^P
 \label{eq:rev-F-arm-gap-tail-step-2}\\
 &\le\left[28\Delta_{\min}^{-1}\{t^{\alpha-1}
                       +4u_0(n_0/t)^\omega\}\right]^P
 \label{eq:rev-F-arm-gap-tail-step-3}\\
 &\le(56\Delta_{\min}^{-1})^Pt^{-(1-\alpha)P}
              +(224u_0\Delta_{\min}^{-1})^P(n_0/t)^{\omega P}\,.
 \label{eq:rev-F-arm-gap-tail-step-4}
\end{align}
\end{subequations}
Equation \eqref{eq:rev-F-arm-gap-tail-step-1} uses $Y_{t,a}^\dagger=W_{t,a}+2\tau_t+\Lambda_{t,a}$ and $0\le\Lambda_{t,a}\le4C\le\tau_t$. Equation \eqref{eq:rev-F-arm-gap-tail-step-2} applies Markov to the nonnegative inverse power and substitutes its local moment bound. Equation \eqref{eq:rev-F-arm-gap-tail-step-3} substitutes $\tau_t=t^\alpha+4u_0n_0^\omega t^{1-\omega}$ and $\Delta(a)^{-1}\le \Delta_{\min}^{-1}$. Equation \eqref{eq:rev-F-arm-gap-tail-step-4} distributes the two contributions with $(x+y)^P\le2^{P-1}(x^P+y^P)$.

The penultimate line is the direct substitution of $\tau_t$ and
$z_t$ from \eqref{eq:F-window} in \Cref{lem:AB-card-window-setup}; the last uses
$(x+y)^P\le2^{P-1}(x^P+y^P)\le2^P(x^P+y^P)$.
Union over $a\ne a^*$, followed by \eqref{eq:F-window-tail} in \Cref{lem:F-window}, yields
\begin{subequations}\label{eq:rev-F-complete-good-tail}
\begin{align}
 \sP(\mathcal G_t^c)
 &\le\sP(I_t=0)+\sum_{a\ne a^*}\sP(I_t=1,W_{t,a}<4\tau_t)
 \label{eq:rev-F-complete-good-tail-step-1}\\
 &\le\mathsf A t^{-65/64}+(n_0/t)^{r_0}
       +K(56\Delta_{\min}^{-1})^Pt^{-\zeta}
       +K(224u_0\Delta_{\min}^{-1})^P(n_0/t)^{\omega P}\,.
 \label{eq:rev-F-complete-good-tail-step-2}
\end{align}
\end{subequations}
Equation \eqref{eq:rev-F-complete-good-tail-step-1} takes the union over the stopped-window failure and all suboptimal gap failures. Equation \eqref{eq:rev-F-complete-good-tail-step-2} substitutes the window tail and the complete two-term per-action gap tail, then uses at most $K$ suboptimal actions and $\zeta=(1-\alpha)P$.

On $\mathcal G_t$, $X_{t,a}=W_{t,a}+2b_t^{\rm rem}>0$ for
$a\ne a^*$, whereas $X_{t,a^*}=0$. Hence $r_t=a^*$, $M_t=0$, and
$Z_{t,a}=X_{t,a}\ge W_{t,a}\ge4\tau_t\ge4t^\alpha$, where
$b_t^{\rm rem}\ge0$ and $\tau_t\ge t^\alpha$.
\end{proof}

\subsubsection{A Fixed-Endpoint Plateau Window}
The local moment argument uses a recent interval on which negative excursions of cumulative loss estimates stay below a positive boundary. The next lemma chooses this
interval and checks that it is long enough, begins after the required
thresholds, and has predictable survival events. The cumulative loss estimates at its start
retain the entire preceding history; the window is only an analysis device.

\begin{lemma}[Scalar exponent margins and valid plateau windows]\label{lem:AB-card-window-setup}
The exponents, thresholds, and windows in
\eqref{eq:F-exponents}, \eqref{eq:F-thresholds}, and \eqref{eq:F-window}
are well defined. For $t\ge T_\circ$,
\begin{equation*}
 S_0,n_0\le m\le t/2\,,\qquad
 \tau_s\ge4C\ (s\ge m)\,,\qquad \tau_m\le5m\,. \notag
\end{equation*}
The exponent margins \eqref{eq:F-exponent-margin} hold, and $I_s$ is
$\cF_{s-1}$-measurable and nonincreasing.
\end{lemma}
\begin{proof}[Proof of \Cref{lem:AB-card-window-setup}]
Use the exact exponents
\begin{subequations}\label{eq:F-exponents}
\begin{align}
 \lambda&=25/32\,,&\alpha&=3/4+13/(10P)\,,&\eta&=13/10\,,
 \label{eq:F-exponents-step-1}\\
 \zeta&=P(1-\alpha)\,,&\kappa_0&=1/256\,,&\omega&=255/256\,,
 \label{eq:F-exponents-step-2}\\
 \sigma_0&=\omega(1-\kappa_0)\,,&
 \varepsilon_0&=1-\sigma_0\,,
 \label{eq:F-exponents-step-3}\\
 \gamma_0&=1/(2\omega)=128/255\,,&a_0&=(1-\lambda)(1-\kappa_0)=1785/8192\,,
 &r_0&=P(1-\omega)-v\,.
 \label{eq:F-exponents-step-4}
\end{align}
\end{subequations}
Equations \eqref{eq:F-exponents-step-1}--\eqref{eq:F-exponents-step-4} define the numerical exponents of the window construction. The identities $\zeta=P(1-\alpha)$, $\sigma_0=\omega(1-\kappa_0)$, and $r_0=P(1-\omega)-v$ respectively encode the pure-gap tail, inherited distortion, and homogeneous window-tail decay.

They satisfy $5a_0>1$, $5(1-\alpha)>1$, $5(\alpha-1/2)>1$ and $2+\varepsilon_0<5/2$.
Moreover,
\begin{equation}
 \label{eq:F-exponent-margin}
 1/128<P/4096\le v<P/4000<\nu\,,\qquad r_0\ge65/128>1/2\,.
\end{equation}
Indeed $B\ge2^{24}P$ and $k^2<e$ imply
\begin{equation}
 \frac vP<\frac{3e}{2^{24}}+\frac1{4096}<\frac1{4000}\,.
 \label{eq:AB-display-215}
\end{equation}
Equation \eqref{eq:AB-display-215} substitutes $v=3P^2k^2/B+P/4096$, divides by $P$, and uses $B\ge2^{24}P$ and $k^2<e$. In particular, $r_0=P(1/256)-v>P(1/256-1/4000)>65/128$ when $P\ge512$.

Since $P\ge512$, we have $r_0>512(1/256-1/4000)>65/128$.
The lower bound on $v$ also follows from $P\ge512$. These parameters satisfy
$1+1/\sigma_0<3$, which controls the accumulated ratio distortion.

Define
\begin{subequations}\label{eq:F-thresholds}
\begin{align}
 S_0&=\left\lceil\max\{16,(3BC_{\beta})^{1/(\alpha-1/2)},[32(P+1)C_{\beta}]^{1/(2\alpha-3/2)}\}\right\rceil\,,
 \label{eq:F-thresholds-step-1}\\
 A_0&=4(16k+15)\,,
 \label{eq:F-thresholds-step-2}\\
 Q&=\Bigl\lceil\max\bigl\{1024,4^{32/7},(S_0+1)^{32/25},A_0^{8192/1785},\notag\\
 &\hspace{2em}5^{1/(1-\alpha)},[32(P-1)C_{\beta}]^{1/(\alpha-1/2)},(32C_{\beta})^8,(4C_{\beta}\log(eK))^4\bigr\}\Bigr\rceil\,,
 \label{eq:F-thresholds-step-3}\\
 T_\circ&=\left\lceil\max\{Q\Delta_{\min}^{-5},1024n_0\Delta_{\min}^{-1}\}\right\rceil\,.
 \label{eq:F-thresholds-step-4}
\end{align}
\end{subequations}
Equation \eqref{eq:F-thresholds-step-1} collects the two local Taylor requirements; \eqref{eq:F-thresholds-step-2} fixes the inherited positive-root coefficient. Equation \eqref{eq:F-thresholds-step-3} records each pure-time requirement separately. Equation \eqref{eq:F-thresholds-step-4} takes the maximum of the pure-time threshold and the budget-dependent window threshold, rather than multiplying the budget-dependent threshold by $Q$.

The maximum in $T_\circ$ is not multiplied as a whole by $Q$.
For a fixed endpoint $t\ge T_\circ$, set
\begin{equation}
 \label{eq:F-window}
 z_t=u_0n_0^\omega t^{1-\omega}\,,\quad m=\lceil\max\{t^\lambda,z_t\}\rceil\,,
 \quad\tau_s=s^\alpha+4z_t\ (s\ge m)\,,
 \quad I_s=\ind\{N_j\le\tau_j\text{ for }m\le j\le s\}\,.
\end{equation}
The required endpoint and boundary bounds follow as
\begin{align}
 t^\lambda&\ge Q^\lambda\ge(S_0+1)^{(32/25)(25/32)}=S_0+1\,,
 \label{eq:AB-card-window-burn-lower}\\
 z_t&=u_0n_0(t/n_0)^{1-\omega}\ge n_0\,,\qquad
 t^\lambda/t=t^{-7/32}\le(4^{32/7})^{-7/32}=1/4\,,
 \label{eq:AB-card-window-left-components}\\
 z_t/t&=u_0(n_0/t)^\omega\le u_0(1024\Delta_{\min}^{-1})^{-\omega}
 <37/1024^\omega<1/4\,,
 \label{eq:AB-card-window-plateau-fraction}\\
 m&=\lceil\max\{t^\lambda,z_t\}\rceil
 \le2\max\{t^\lambda,z_t\}\le t/2\,,
 \label{eq:AB-card-window-left-upper}\\
 \tau_s&=s^\alpha+4z_t\ge4z_t\ge4n_0\ge4C\,,\qquad
 \tau_m=m^\alpha+4z_t\le m+4m=5m\,,
 \label{eq:AB-card-window-boundary}\\
 I_{s+1}&=I_s\ind\{N_{s+1}\le\tau_{s+1}\}\le I_s\,,
 \qquad I_s\in\mathcal F_{s-1}\,.
 \label{eq:AB-card-window-predictability}
\end{align}
Equation \eqref{eq:AB-card-window-burn-lower} uses the corresponding
entry in $Q$. Equation \eqref{eq:AB-card-window-left-components} uses
$t\ge n_0$ and $Q\ge4^{32/7}$; \eqref{eq:AB-card-window-plateau-fraction}
uses $t\ge1024n_0\Delta_{\min}^{-1}$ and $u_0=18k+1<37$.
Equation \eqref{eq:AB-card-window-left-upper} uses
$\lceil x\rceil\le2x$ for $x\ge1$.
Equation \eqref{eq:AB-card-window-boundary} uses $\alpha<1$ and $z_t\le m$.
Equation \eqref{eq:AB-card-window-predictability} writes the survival
event recursively; each $N_j$ for $j\le s$ is $\mathcal F_{s-1}$-measurable.
The plateau $4z_t$ remains fixed as $s$ varies within this endpoint window.
\end{proof}

Local estimates on a surviving window are useful only if window failure
is sufficiently rare. The next lemma uses stopped negative-part moments
to control the probability of crossing the boundary anywhere in the
window. This provides the window-failure term in the good-event tail bound.

\begin{lemma}[Window failure probability]\label{lem:F-window}
With $\mathsf A=2^{\nu+1}H_0$,
\begin{equation}\label{eq:F-window-tail}
 \sP(I_t=0)\le\mathsf A t^{-65/64}+(n_0/t)^{r_0}\,.
\end{equation}
\end{lemma}
\begin{proof}
Fix a dyadic block $[u,2u]$ with $u=2^jm\le t$, and let
$\sigma_u=\inf\{s\ge u:F_s>\tau_u^P\}$. Let
$\widetilde F_s=F_{s\wedge\sigma_u}$ for $s\ge u$.
The event $\{s<\sigma_u\}$ is $\mathcal F_{s-1}$-measurable.
On that event use \eqref{eq:F-Flinear} in \Cref{lem:F-negative}; on its complement the
stopped process is constant. Thus
\begin{subequations}\label{eq:rev-F-stopped-block-moment}
\begin{align}
 \E_{s-1}[\widetilde F_{s+1}]
 &=\ind\{s<\sigma_u\}\E_{s-1}[F_{s+1}]
       +\ind\{s\ge\sigma_u\}\widetilde F_s
 \label{eq:rev-F-stopped-block-moment-step-1}\\
 &\le(1+v/s)\widetilde F_s+J_0s^{\nu-1}\,,
 \label{eq:rev-F-stopped-block-moment-step-2}\\
 \E[\widetilde F_{2u}]
 &\le H_0(2u)^\nu+u_0^Pn_0^P\Pi(n_0,2u)
 \label{eq:rev-F-stopped-block-moment-step-3}\\
 &\le2^\nu H_0u^\nu
            +2^{2v}u_0^Pn_0^{P-v}u^v\,.
 \label{eq:rev-F-stopped-block-moment-step-4}
\end{align}
\end{subequations}
Equation \eqref{eq:rev-F-stopped-block-moment-step-1} conditions on the predictable event that the block has not yet frozen. Equation \eqref{eq:rev-F-stopped-block-moment-step-2} applies the one-step negative-moment inequality on that event; on its complement, both added terms on the right are nonnegative. Equation \eqref{eq:rev-F-stopped-block-moment-step-3} propagates the same deterministic supersolution, including the initial-budget contribution, to time $2u$. Equation \eqref{eq:rev-F-stopped-block-moment-step-4} substitutes $\Pi(n_0,2u)\le(4u/n_0)^v$ and expands $(2u)^\nu$.

The middle inequality propagates the supersolution from
\eqref{eq:F-Fmoment} in \Cref{lem:F-negative}, starting from the same bound at $u$.
If $N_s>\tau_s$ at some $s\in[u,2u]$, then
$F_s\ge N_s^P>\tau_s^P\ge\tau_u^P$, so
$\widetilde F_{2u}>\tau_u^P$. Hence Markov's inequality gives
\begin{subequations}\label{eq:rev-F-stopped-block-tail}
\begin{align}
 \sP(\exists s\in[u,2u]:N_s>\tau_s)
 &\le\tau_u^{-P}\E[\widetilde F_{2u}]
 \label{eq:rev-F-stopped-block-tail-step-1}\\
 &\le2^\nu H_0\frac{u^\nu}{u^{\alpha P}}
       +\frac{2^{2v}u_0^Pn_0^{P-v}u^v}{(4z_t)^P}
 \label{eq:rev-F-stopped-block-tail-step-2}\\
 &=2^\nu H_0u^{-13/10}
       +\frac{2^{2v}u_0^Pn_0^{P-v}u^v}{(4z_t)^P}\,.
 \label{eq:rev-F-stopped-block-tail-step-3}
\end{align}
\end{subequations}
Equation \eqref{eq:rev-F-stopped-block-tail-step-1} applies Markov to the frozen nonnegative moment, which retains the crossing value. Equation \eqref{eq:rev-F-stopped-block-tail-step-2} uses $\tau_u\ge u^\alpha$ for the particular term and $\tau_u\ge4z_t$ for the homogeneous term. Equation \eqref{eq:rev-F-stopped-block-tail-step-3} substitutes $\nu-\alpha P=-13/10$.

We used $\tau_u\ge u^\alpha$, $\tau_u\ge4z_t$, and
$\nu-\alpha P=3P/4-(3P/4+13/10)=-13/10$.
Let $J=\lfloor\log_2(t/m)\rfloor$. The two geometric sums are
\begin{subequations}\label{eq:rev-F-window-geometric-series}
\begin{align}
 \sum_{j=0}^J(2^jm)^{-\eta}
 &\le\frac{m^{-\eta}}{1-2^{-\eta}}\le2m^{-\eta}\,,
 \label{eq:rev-F-window-geometric-series-step-1}\\
 \sum_{j=0}^J(2^jm)^v
 &=(2^Jm)^v\sum_{j=0}^J2^{-jv}
 \le\frac{t^v}{1-2^{-v}}\le256t^v\,.
 \label{eq:rev-F-window-geometric-series-step-2}
\end{align}
\end{subequations}
Equation \eqref{eq:rev-F-window-geometric-series-step-1} sums the decreasing geometric progression and uses $\eta=13/10>1$. Equation \eqref{eq:rev-F-window-geometric-series-step-2} reverses the increasing progression from its largest term $2^Jm\le t$ and uses $(1-2^{-v})^{-1}\le1+(v\log2)^{-1}<256$.

Summing \eqref{eq:rev-F-stopped-block-tail}, substituting
\eqref{eq:rev-F-window-geometric-series}, and then replacing
$z_t=u_0n_0^\omega t^{1-\omega}$ gives
\begin{subequations}\label{eq:rev-F-window-tail-final}
\begin{align}
 \sP(I_t=0)
 &\le2^{\nu+1}H_0m^{-\eta}
       +\frac{2^{8+2v}u_0^Pn_0^{P-v}t^v}
                    {4^Pz_t^P}
 \label{eq:rev-F-window-tail-final-step-1}\\
 &=\mathsf A m^{-\eta}
       +2^{8+2v-2P}
          \frac{u_0^Pn_0^{P-v}t^v}
               {u_0^Pn_0^{\omega P}t^{(1-\omega)P}}
 \label{eq:rev-F-window-tail-final-step-2}\\
 &=\mathsf A m^{-\eta}
       +2^{8+2v-2P}(n_0/t)^{P(1-\omega)-v}
 \label{eq:rev-F-window-tail-final-step-3}\\
 &\le\mathsf A t^{-\lambda\eta}+(n_0/t)^{r_0}
 =\mathsf A t^{-65/64}+(n_0/t)^{r_0}\,.
 \label{eq:rev-F-window-tail-final-step-4}
\end{align}
\end{subequations}
Equation \eqref{eq:rev-F-window-tail-final-step-1} sums the block-failure bounds over both geometric progressions. Equation \eqref{eq:rev-F-window-tail-final-step-2} substitutes $\mathsf A=2^{\nu+1}H_0$ and $z_t=u_0n_0^\omega t^{1-\omega}$ before cancellation. Equation \eqref{eq:rev-F-window-tail-final-step-3} cancels $u_0^P$ and collects the $n_0,t$ powers. Equation \eqref{eq:rev-F-window-tail-final-step-4} uses $m\ge t^\lambda$, $8+2v-2P\le0$, $r_0=P(1-\omega)-v$, and $\lambda\eta=65/64$.

Here $m\ge t^\lambda$, $8+2v-2P\le0$, and
$\lambda\eta=(25/32)(13/10)=65/64$.
\end{proof}

\subsubsection{Compensated Local Moments}
Raw cumulative differences can be zero or negative, so the positive-process
inverse-moment argument does not apply directly. The next lemma removes the predictable signed
corruption and adds a buffer to obtain a positive analysis process. Its
increment bounds justify the inverse- and positive-moment Taylor
expansions in \Cref{lem:F-local,lem:B-positive-local}.

\begin{lemma}[A positive compensated process and its local increments]\label{lem:AB-card-compensation}
For a fixed window, the process $Y_{s,a}^\dagger$ in \eqref{eq:F-Y}
is an analysis variable, not the observed feedback. On $I_s=1$ it
satisfies $Y_{s,a}^\dagger\ge\tau_s>0$ and all identities and bounds
in \eqref{eq:F-local-inc}. Also $Y_{s+1,a}^\dagger>0$ on $I_s=1$,
even if the next survival indicator is zero.
\end{lemma}
\begin{proof}[Proof of \Cref{lem:AB-card-compensation}]
For $s$ in the window define
\begin{equation}
 \label{eq:F-Y}
 Y_{s,a}^{\dagger}=X_{s,a}-E_{s,a}+2C+2\tau_s
 =W_{s,a}+2\tau_s+\Lambda_{s,a}\,,\qquad
 \Lambda_{s,a}=2b_s^{\rm rem}-E_{s,a}+2C\in[0,4C]\,.
\end{equation}
This $Y^\dagger$ is an analysis process, not the observed feedback. On $I_s=1$, it is at least $\tau_s>0$. Expressions multiplied by $I_s$ are defined as zero after stopping.
Setting $v_s=\tau_{s+1}-\tau_s$ and
$\xi_{s,a}=D_{s,a}-e_{s,a}+2v_s$ gives, on $I_s=1$,
\begin{subequations}\label{eq:F-local-inc}
\begin{align}
 Y_{s+1,a}^{\dagger}&=Y_{s,a}^{\dagger}+\xi_{s,a}\,,&
 \E_{s-1}[\xi_{s,a}]&=\Delta(a)+2v_s\,,
 \label{eq:F-local-inc-step-1}\\
 |\xi_{s,a}|&\le3Y_{s,a}^{\dagger}/B\,,&
 \E_{s-1}[\xi_{s,a}^2]&\le8\beta_s+4(Y_{s,a}^{\dagger})^2/(Bs)\,.
 \label{eq:F-local-inc-step-2}
\end{align}
\end{subequations}
Equation \eqref{eq:F-local-inc-step-1} records the increment and conditional drift of the compensated cumulative loss estimate; subtracting $e_{s,a}$ removes the predictable corruption drift. Equation \eqref{eq:F-local-inc-step-2} records the relative update magnitude and second moment needed for both Taylor expansions. The direct substitutions establishing these bounds follow below.

For the verification of \eqref{eq:F-local-inc}, first substitute
$X_{s,a}=W_{s,a}+2b_s^{\rm rem}$ and
$N_s=M_s+2b_s^{\rm rem}$. On $I_s=1$ this gives
\begin{subequations}\label{eq:rev-F-local-raw-step}
\begin{align}
 Z_{s,a}+M_s
 &=X_{s,a}+2M_s
 =W_{s,a}+2N_s-2b_s^{\rm rem}
 \le W_{s,a}+2\tau_s\le Y_{s,a}^\dagger\,,
 \label{eq:rev-F-local-raw-step-step-1}\\
 |D_{s,a}|&\le2\beta_s+(Z_{s,a}+M_s)/B
                           \le2\beta_s+Y_{s,a}^\dagger/B\,,
 \label{eq:rev-F-local-raw-step-step-2}\\
 \E_{s-1}[D_{s,a}^2]
 &\le4\beta_s+2(Z_{s,a}^2+M_s^2)/(Bs)
 \le4\beta_s+4(Y_{s,a}^\dagger)^2/(Bs)\,.
 \label{eq:rev-F-local-raw-step-step-3}
\end{align}
\end{subequations}
Equation \eqref{eq:rev-F-local-raw-step-step-1} substitutes $X=W+2b^{\rm rem}$ and $M=N-2b^{\rm rem}$, then uses $b^{\rm rem}\ge0$, $N_s\le\tau_s$, and $\Lambda_{s,a}\ge0$. Equation \eqref{eq:rev-F-local-raw-step-step-2} inserts this bound into the upper bound $2\beta_s+(Z_{s,a}+M_s)/B$ on $|D_{s,a}|$. Equation \eqref{eq:rev-F-local-raw-step-step-3} uses $0\le Z_{s,a},M_s\le Y_{s,a}^\dagger$ separately in the two squared terms.

The plateau is constant, so
$v_s=(s+1)^\alpha-s^\alpha\in(0,\alpha s^{\alpha-1}]\subset(0,1]$.
Also $|e_{s,a}|\le4$, $\Delta(a)\le2$, and the definition of $S_0$
implies $3\beta_s\le s^\alpha/B\le Y_{s,a}^\dagger/B$.
Consequently
\begin{subequations}\label{eq:rev-F-local-compensated-step}
\begin{align}
 |\xi_{s,a}|
 &\le|D_{s,a}|+|e_{s,a}|+2v_s
 \le2\beta_s+Y_{s,a}^\dagger/B+6
 \label{eq:rev-F-local-compensated-step-step-1}\\
 &\le3\beta_s+Y_{s,a}^\dagger/B
 \le2Y_{s,a}^\dagger/B\le3Y_{s,a}^\dagger/B\,,
 \label{eq:rev-F-local-compensated-step-step-2}\\
 \E_{s-1}[(D_{s,a}-e_{s,a})^2]
 &=\operatorname{Var}_{s-1}(D_{s,a})+\Delta(a)^2
 \le\E_{s-1}[D_{s,a}^2]+4\,,
 \label{eq:rev-F-local-compensated-step-step-3}\\
 \E_{s-1}[\xi_{s,a}^2]
 &=\E_{s-1}[(D_{s,a}-e_{s,a})^2]
                          +4v_s\Delta(a)+4v_s^2
 \label{eq:rev-F-local-compensated-step-step-4}\\
 &\le\E_{s-1}[D_{s,a}^2]+4+8+4
 \label{eq:rev-F-local-compensated-step-step-5}\\
 &\le4\beta_s+4(Y_{s,a}^\dagger)^2/(Bs)+16
 \le5\beta_s+4(Y_{s,a}^\dagger)^2/(Bs)
 \label{eq:rev-F-local-compensated-step-step-6}\\
 &\le8\beta_s+4(Y_{s,a}^\dagger)^2/(Bs)\,.
 \label{eq:rev-F-local-compensated-step-step-7}
\end{align}
\end{subequations}
Equation \eqref{eq:rev-F-local-compensated-step-step-1} substitutes $\xi=D-e+2v_s$ and uses $|e|\le4$, $v_s\le1$. Equation \eqref{eq:rev-F-local-compensated-step-step-2} uses $\beta_s\ge6$ and the threshold $3\beta_s\le Y_{s,a}^\dagger/B$. Equation \eqref{eq:rev-F-local-compensated-step-step-3} uses $\E_{s-1}[(D-e)]=\Delta(a)$ and separates variance from squared mean. Equation \eqref{eq:rev-F-local-compensated-step-step-4} expands the square after adding $2v_s$. Equation \eqref{eq:rev-F-local-compensated-step-step-5} uses $\Delta(a)\le2$ and $v_s\le1$. Equation \eqref{eq:rev-F-local-compensated-step-step-6} inserts the raw second moment and $\beta_s\ge16$; \eqref{eq:rev-F-local-compensated-step-step-7} relaxes its coefficient to the stated envelope.

The bounds $\beta_s\ge16$ and $\beta_s\ge6$ used here follow from
the tuning. In addition, on $I_s=1$,
$Y_{s+1,a}^\dagger\ge(1-3/B)Y_{s,a}^\dagger>0$ even when
$I_{s+1}=0$. Thus dropping the next survival indicator from a
nonnegative Taylor bound is legitimate.

\end{proof}

The following inverse moment controls small estimated gaps and hence the
failure probability of the common good event. It is used by all three
terms that depend on $\mathcal G_t$, as well as by the exploration ratios.
The positive local moment and distortion sum are proved separately in
\Cref{lem:B-positive-local}.

\begin{lemma}[Inverse local moments]\label{lem:F-local}
For $a\ne a^*$ and $t\ge T_\circ$,
\begin{equation}\label{eq:F-inverse}
 \E[I_t(Y_{t,a}^{\dagger})^{-P}]\le(4/(t\Delta(a)))^P\,.
\end{equation}
\end{lemma}
\begin{proof}
Fix $a\ne a^*$. In this proof abbreviate $Y_{s,a}^\dagger$ by $Y_s$.
On $I_s=1$, let $y=Y_s$ and $\xi=\xi_{s,a}$.
Taylor's formula for $y^{-P}$, with
$|\xi|/y\le3/B$ and $(1-3/B)^{-(P+2)}\le2$, yields
\begin{subequations}\label{eq:rev-F-inverse-taylor}
\begin{align}
 \E_{s-1}[(y+\xi)^{-P}]
 &\le y^{-P}-Py^{-P-1}\E_{s-1}[\xi]
             +P(P+1)y^{-P-2}\E_{s-1}[\xi^2]
 \label{eq:rev-F-inverse-taylor-step-1}\\
 &\le y^{-P}-P(\Delta(a)+2v_s)y^{-P-1}
       +P(P+1)y^{-P-2}\{8\beta_s+4y^2/(Bs)\}
 \label{eq:rev-F-inverse-taylor-step-2}\\
 &\le\left[1+P\left\{\frac{8(P+1)\beta_s}{y^2}
                                +\frac{4(P+1)}{Bs}\right\}\right]y^{-P}
                       -P\Delta(a) y^{-P-1}
 \label{eq:rev-F-inverse-taylor-step-3}\\
 &\le(1+P/(2s))y^{-P}-P\Delta(a) y^{-P-1}\,.
 \label{eq:rev-F-inverse-taylor-step-4}
\end{align}
\end{subequations}
Equation \eqref{eq:rev-F-inverse-taylor-step-1} applies inverse-power Taylor with $|\xi|/y\le3/B$ and bounds its curvature factor by two. Equation \eqref{eq:rev-F-inverse-taylor-step-2} substitutes the drift and second moment before simplifying. Equation \eqref{eq:rev-F-inverse-taylor-step-3} distributes the curvature product and discards the nonpositive drift term $-2Pv_sy^{-P-1}$. Equation \eqref{eq:rev-F-inverse-taylor-step-4} uses the two separate threshold calculations below.

For \eqref{eq:rev-F-inverse-taylor-step-4}, the two curvature
coefficients are bounded separately as
\begin{subequations}\label{eq:rev-F-inverse-thresholds}
\begin{align}
 \frac{8(P+1)\beta_s}{y^2}
 &\le8(P+1)C_{\beta} s^{1/2-2\alpha}
 =\frac1{4s}\frac{32(P+1)C_{\beta}}{s^{2\alpha-3/2}}
 \le\frac1{4s}\,,
 \label{eq:rev-F-inverse-thresholds-step-1}\\
 \frac{4(P+1)}{Bs}&\le\frac1{4s}\,.
 \label{eq:rev-F-inverse-thresholds-step-2}
\end{align}
\end{subequations}
Equation \eqref{eq:rev-F-inverse-thresholds-step-1} uses $y\ge s^\alpha$ and $\beta_s\le C_{\beta}\sqrt s$, then factors out $1/(4s)$ to expose the threshold $s^{2\alpha-3/2}\ge32(P+1)C_{\beta}$. Equation \eqref{eq:rev-F-inverse-thresholds-step-2} is exactly $B\ge16(P+1)$ after multiplying both sides by $Bs$.

The first is the second power threshold in $S_0$; the second uses
$B\ge16(P+1)$. Define $u_s=\E[I_sY_s^{-P}]$. Since
$I_{s+1}\le I_s$, $Y_{s+1}>0$ on $I_s=1$, and
$\E[(I_sY_s^{-P})^{1+1/P}]\ge u_s^{1+1/P}$, taking expectations
in \eqref{eq:rev-F-inverse-taylor} gives
\begin{subequations}\label{eq:rev-F-inverse-recursion}
\begin{align}
 u_{s+1}
 &\le(1+P/(2s))u_s
                    -P\Delta(a)\E[I_sY_s^{-P-1}]
 \label{eq:rev-F-inverse-recursion-step-1}\\
 &\le(1+P/(2s))u_s-P\Delta(a) u_s^{1+1/P}\,.
 \label{eq:rev-F-inverse-recursion-step-2}
\end{align}
\end{subequations}
Equation \eqref{eq:rev-F-inverse-recursion-step-1} uses $I_{s+1}\le I_s$, positivity of the next cumulative loss estimate on $I_s=1$, and the tower property, with $I_s$ measurable before the time-$s$ feedback. Equation \eqref{eq:rev-F-inverse-recursion-step-2} applies Jensen to $(I_sY_s^{-P})^{1+1/P}=I_sY_s^{-P-1}$.

If $u_s=0$, \eqref{eq:rev-F-inverse-recursion} implies all subsequent
$u$'s vanish. Otherwise, as long as $u_{s+1}>0$, let
$z_s=u_s^{-1/P}$. Positivity of the RHS and convexity
of $x\mapsto(1+x)^{-1/P}$ give
\begin{subequations}\label{eq:rev-F-inverse-root-solution}
\begin{align}
 z_{s+1}
 &\ge z_s\left(1+\frac P{2s}-\frac{P\Delta(a)}{z_s}\right)^{-1/P}
 \label{eq:rev-F-inverse-root-solution-step-1}\\
 &\ge z_s\left(1-\frac1{2s}+\frac{\Delta(a)}{z_s}\right)
 =\left(1-\frac1{2s}\right)z_s+\Delta(a)
 \label{eq:rev-F-inverse-root-solution-step-2}\\
 &\ge\frac{s}{s+1}z_s+\Delta(a)\,,
 \label{eq:rev-F-inverse-root-solution-step-3}\\
 tz_t&\ge mz_m+\Delta(a)\sum_{s=m}^{t-1}(s+1)
 \ge\frac{\Delta(a)}2\{t(t+1)-m(m+1)\}
 \label{eq:rev-F-inverse-root-solution-step-4}\\
 &\ge\frac{t^2\Delta(a)}4\,.
 \label{eq:rev-F-inverse-root-solution-step-5}
\end{align}
\end{subequations}
Equation \eqref{eq:rev-F-inverse-root-solution-step-1} raises the positive recurrence bound to the decreasing power $-1/P$. Equation \eqref{eq:rev-F-inverse-root-solution-step-2} uses the tangent inequality $(1+x)^{-1/P}\ge1-x/P$ for $x>-1$. Equation \eqref{eq:rev-F-inverse-root-solution-step-3} uses $1-1/(2s)\ge s/(s+1)$. Equation \eqref{eq:rev-F-inverse-root-solution-step-4} multiplies by $s+1$, sums all increments, and evaluates the arithmetic sum. Equation \eqref{eq:rev-F-inverse-root-solution-step-5} uses $m\le t/2$ and drops the nonnegative initial contribution.

Equation \eqref{eq:rev-F-inverse-root-solution-step-5} uses
$m\le t/2$, which gives $t(t+1)-m(m+1)\ge t^2/2$. Hence
$u_t=z_t^{-P}\le(4/(t\Delta(a)))^P$, also when $u_t=0$.

\end{proof}

\subsubsection{Remaining-Budget Shift and Negative Moments}
Corruption can make a suboptimal action appear favorable by driving its
estimated cumulative loss difference from $a^*$ downward. For the
global negative-moment analysis, we subtract twice the remaining corruption
budget from this difference estimate; the decrease in that budget restores
nonnegative conditional drift. The next
lemma verifies this shift and its update bounds without changing the learner.

\begin{lemma}[Remaining-budget shift and nonnegative drift]\label{lem:AB-card-shift}
The remaining-budget variables in \eqref{eq:F-W} satisfy
$Z_{t,a}=W_{t,a}+N_t$ and all increment bounds in
\eqref{eq:F-shift-inc}. The shifted conditional drift is nonnegative
on every history.
\end{lemma}
\begin{proof}[Proof of \Cref{lem:AB-card-shift}]
Let
\begin{equation}
 A_t^{\rm cor}=\sum_{s<t}\kappa_s\,,\quad E_{t,a}=\sum_{s<t}e_{s,a}\,,
 \quad b_t^{\rm rem}=C-A_t^{\rm cor}\ge0\,,
 \label{eq:AB-display-240}
\end{equation}
Equation \eqref{eq:AB-display-240} distinguishes corruption already spent, the cumulative signed drift error, and the nonnegative remaining budget. Their different roles matter: the signed error is removed from the local cumulative loss estimate, whereas twice the remaining budget supplies the monotone global shift.

\begin{equation}
 \label{eq:F-W}
 W_{t,a}=X_{t,a}-2b_t^{\rm rem}\,,\quad N_t=M_t+2b_t^{\rm rem}=\max_a(-W_{t,a})\,,
 \quad F_t=\sum_a(-W_{t,a})_+^P\,.
\end{equation}
Here $W_{t,a^*}=-2b_t^{\rm rem}$, not zero, and $Z_{t,a}=W_{t,a}+N_t$.
Setting $\eta_{t,a}=D_{t,a}+2\kappa_t$ gives
\begin{subequations}\label{eq:F-shift-inc}
\begin{align}
 W_{t+1,a}&=W_{t,a}+\eta_{t,a}\,,&\E_{t-1}[\eta_{t,a}]&\ge\Delta(a)\ge0\,,
 \label{eq:F-shift-inc-step-1}\\
 |\eta_{t,a}|&\le3\beta_t+(Z_{t,a}+N_t)/B\,,&
 \E_{t-1}[\eta_{t,a}^2]&\le6\beta_t+2(Z_{t,a}^2+N_t^2)/(Bt)\,.
 \label{eq:F-shift-inc-step-2}
\end{align}
\end{subequations}
Equation \eqref{eq:F-shift-inc-step-1} records the shifted increment and its nonnegative conditional drift. Equation \eqref{eq:F-shift-inc-step-2} records the shifted update magnitude and second moment; the complete square expansion establishing them is displayed next.

For clarity, the shift identities and moment calculation are
\begin{subequations}\label{eq:rev-F-shift-verification}
\begin{align}
 b_{t+1}^{\rm rem}&=b_t^{\rm rem}-\kappa_t\,,\qquad
 W_{t+1,a}-W_{t,a}=D_{t,a}+2\kappa_t=\eta_{t,a}\,,
 \label{eq:rev-F-shift-verification-step-1}\\
 \E_{t-1}[\eta_{t,a}]
 &=\Delta(a)+e_{t,a}+2\kappa_t\ge\Delta(a)\,,
 \label{eq:rev-F-shift-verification-step-2}\\
 |\eta_{t,a}|
 &\le2\beta_t+(Z_{t,a}+M_t)/B+2\kappa_t
 \le3\beta_t+(Z_{t,a}+N_t)/B\,,
 \label{eq:rev-F-shift-verification-step-3}\\
 \E_{t-1}[\eta_{t,a}^2]
 &=\E_{t-1}[D_{t,a}^2]+4\kappa_t\E_{t-1}[D_{t,a}]
                                      +4\kappa_t^2
 \label{eq:rev-F-shift-verification-step-4}\\
 &\le4\beta_t+2(Z_{t,a}^2+M_t^2)/(Bt)+8\kappa_t+4\kappa_t^2
 \label{eq:rev-F-shift-verification-step-5}\\
 &\le4\beta_t+2(Z_{t,a}^2+N_t^2)/(Bt)+32
 \label{eq:rev-F-shift-verification-step-6}\\
 &\le6\beta_t+2(Z_{t,a}^2+N_t^2)/(Bt)\,.
 \label{eq:rev-F-shift-verification-step-7}
\end{align}
\end{subequations}
Equation \eqref{eq:rev-F-shift-verification-step-1} substitutes the remaining-budget update. Equation \eqref{eq:rev-F-shift-verification-step-2} uses $e_{t,a}\ge-2\kappa_t$. Equation \eqref{eq:rev-F-shift-verification-step-3} uses $M_t\le N_t$ and $2\kappa_t\le4\le\beta_t$. Equation \eqref{eq:rev-F-shift-verification-step-4} expands the square using predictability of $\kappa_t$. Equation \eqref{eq:rev-F-shift-verification-step-5} substitutes the unshifted second moment and $|\E_{t-1}[D_{t,a}]|\le2$. Equation \eqref{eq:rev-F-shift-verification-step-6} uses $\kappa_t\le2$ and $M_t\le N_t$. Equation \eqref{eq:rev-F-shift-verification-step-7} absorbs $32\le2\beta_t$.

Here $|e_{t,a}|\le2\kappa_t$, $\kappa_t\le2$,
$M_t\le N_t$, $|\E_{t-1}[D_{t,a}]|\le2$, and $\beta_t\ge16$.
\end{proof}

\begin{definition}[Scalar negative-moment dictionary]\label{def:AB-card-negative}
Define
\begin{align}
 a_-&=3P^2k^2/B\,,&c_-&=4P^2k^2C_{\beta}\,,&v_-&=6C_{\beta}+64C_{\beta}^2/B\,,\notag\\
 w_-&=v_-+2592P^2k^2C_{\beta}^2/B\,,&D_-&=\tfrac12P^2k^2(27PkC_{\beta})^{P-2}w_-\,,\notag\\
 \Lambda&=P/4096\,,&v&=a_-+\Lambda\,,&R_0&=\max\{1,(c_-/\Lambda)^{P/2}\}\,,\notag\\
 J_0&=c_-R_0^{1-2/P}+D_-\,,&H_0&=\max\{1,J_0/(\nu-v)\}\,,&u_0&=18k+1\,,\notag\\
 n_0&=\max\{N_{\rm burn},\lceil C\rceil\}\,,&&
 \Pi(n,t)=\prod_{s=n}^{t-1}(1+v/s)\,. \notag
\end{align}
The empty product is one and $n_0\le N_{\rm burn}+C$.
\end{definition}

Nonnegative drift alone does not control how far noisy cumulative loss estimates can fall
below zero. The next lemma bounds a power sum of their negative parts,
including the contribution inherited from the earlier history and the
corruption budget. Its stopped version is used to bound boundary crossings
in \Cref{lem:F-window}.

\begin{lemma}[Negative-part propagation]\label{lem:F-negative}
For every history,
\begin{align}
 \E_{t-1}[F_{t+1}]&\le(1+a_-/t)F_t+c_-\sqrt t\,F_t^{1-2/P}+D_-t^{(P-1)/2}\,, \notag\\
 \E_{t-1}[F_{t+1}]&\le(1+v/t)F_t+J_0t^{\nu-1}\,.\label{eq:F-Flinear}
\end{align}
For $t\ge n_0$,
\begin{equation}\label{eq:F-Fmoment}
 \E[F_t]\le H_0t^\nu+u_0^Pn_0^P\Pi(n_0,t)
 \le H_0t^\nu+2^vu_0^Pn_0^{P-v}t^v\,.
\end{equation}
\end{lemma}
\begin{proof}
Condition on $\mathcal F_{t-1}$. Let $y_a=-W_{t,a}$,
$z_a=-\eta_{t,a}$, $A=F_t^{1/P}$, $N=N_t$, and $\beta=\beta_t$.
Because $y_{a^*}=2b_t^{\rm rem}\ge0$, one has
$0\le N=\max_a y_a\le A\le kN$.
The update magnitude inequality in \eqref{eq:F-shift-inc} in \Cref{lem:AB-card-shift} becomes
\begin{subequations}\label{eq:rev-F-negative-active-coordinates}
\begin{align}
 |z_a|&\le3\beta+(2N-y_a)/B\,,
 \label{eq:rev-F-negative-active-coordinates-step-1}\\
 y_a+|z_a|&\le(1-1/B)y_a+3\beta+2N/B\,,
 \label{eq:rev-F-negative-active-coordinates-step-2}\\
 y_a\le-d_0\,,\
 d_0\coloneq \frac{3\beta+2N/B}{1-1/B}
 &\Longrightarrow y_a+|z_a|\le0\,.
 \label{eq:rev-F-negative-active-coordinates-step-3}
\end{align}
\end{subequations}
Equation \eqref{eq:rev-F-negative-active-coordinates-step-1} substitutes $Z_{t,a}=N-y_a$ in the shifted update magnitude bound. Equation \eqref{eq:rev-F-negative-active-coordinates-step-2} adds $y_a$ and collects its coefficient. Equation \eqref{eq:rev-F-negative-active-coordinates-step-3} solves the resulting affine inequality for the range in which the positive-part Taylor curvature is identically zero.

Define the $\mathcal F_{t-1}$-measurable envelope
$c_a=((1-1/B)y_a+3\beta+2N/B)_+$, so
$(y_a+|z_a|)_+\le c_a$ and $c_a=0$ when $y_a\le-d_0$.
For such inactive coordinates, the positive-part Taylor remainder is
zero. On all coordinates the first-order term has nonpositive
conditional expectation, since $\E_{t-1}[z_a]\le0$.
Thus \eqref{eq:taylor-positivepart} in \Cref{lem:taylor} gives
\begin{align}
 \E_{t-1}[F_{t+1}]
 &\le F_t+\frac{P(P-1)}2
       \sum_{a:y_a>-d_0}c_a^{P-2}
                                \E_{t-1}[z_a^2]\,.
 \label{eq:rev-F-negative-taylor-sum}
\end{align}
The update magnitudes in \eqref{eq:rev-F-negative-active-coordinates} also
imply the deterministic curvature bound
\begin{subequations}\label{eq:rev-F-negative-curvature}
\begin{align}
 \left\{\sum_ac_a^{P-2}\right\}^{1/(P-2)}
 &\le K^{2/(P(P-2))}
          \left\{\sum_ac_a^P\right\}^{1/P}
 \label{eq:rev-F-negative-curvature-step-1}\\
 &\le k^{2/(P-2)}
           \{A+k(3\beta+2N/B)\}
 \label{eq:rev-F-negative-curvature-step-2}\\
 &\le k^{2/(P-2)}\{(1+2k/B)A+3k\beta\}\,.
 \label{eq:rev-F-negative-curvature-step-3}
\end{align}
\end{subequations}
Equation \eqref{eq:rev-F-negative-curvature-step-1} is H\"older from the $P$-norm to the $(P-2)$-norm on the $K$ coordinates. Equation \eqref{eq:rev-F-negative-curvature-step-2} uses $c_a\le(y_a)_++3\beta+2N/B$ and Minkowski. Equation \eqref{eq:rev-F-negative-curvature-step-3} uses $N\le A$ and collects its coefficient inside the braces.

First suppose $A\ge12Pk\beta$. Then
$\beta\le A/(12Pk)\le N/(12P)$ and, using $B\ge1024P$,
\begin{subequations}\label{eq:rev-F-negative-large-branch}
\begin{align}
 d_0
 &\le N\frac{1/(4P)+2/(1024P)}{1-1/(1024P)}
 \le N/(3P)\,,
 \label{eq:rev-F-negative-large-branch-step-1}\\
 y_a>-d_0
 &\Longrightarrow Z_{t,a}=N-y_a\le(1+1/(3P))N\,,
 \label{eq:rev-F-negative-large-branch-step-2}\\
 \E_{t-1}[z_a^2]
 &\le6\beta+\frac{2\{(1+1/(3P))^2+1\}N^2}{Bt}
 \le6\beta+\frac{9N^2}{2Bt}\,,
 \label{eq:rev-F-negative-large-branch-step-3}\\
 (1+2k/B)A+3k\beta
 &\le A\left(1+\frac{2k}{B}+\frac1{4P}\right)
 \le A\left(1+\frac{65}{256P}\right)\,,
 \label{eq:rev-F-negative-large-branch-step-4}\\
 \sum_ac_a^{P-2}
 &\le k^2A^{P-2}\left(1+\frac{65}{256P}\right)^{P-2}
 \le\tfrac43k^2A^{P-2}\,.
 \label{eq:rev-F-negative-large-branch-step-5}
\end{align}
\end{subequations}
Equation \eqref{eq:rev-F-negative-large-branch-step-1} substitutes $\beta\le N/(12P)$ and $B\ge1024P$ in $d_0$. Equation \eqref{eq:rev-F-negative-large-branch-step-2} identifies the remaining active coordinates. Equation \eqref{eq:rev-F-negative-large-branch-step-3} uses their bound on $Z_{t,a}$ in the shifted second moment. Equation \eqref{eq:rev-F-negative-large-branch-step-4} uses $3k\beta/A\le1/(4P)$ and $2k/B\le1/(256P)$. Equation \eqref{eq:rev-F-negative-large-branch-step-5} substitutes that curvature base and uses $(1+x)^{P-2}\le e^{(P-2)x}$ and $e^{65/256}<4/3$.

Substituting the active-coordinate variance in
\eqref{eq:rev-F-negative-large-branch-step-3} and the curvature sum
in \eqref{eq:rev-F-negative-large-branch-step-5} into
\eqref{eq:rev-F-negative-taylor-sum} gives
\begin{subequations}\label{eq:rev-F-negative-large-remainder}
\begin{align}
 \E_{t-1}[F_{t+1}]-F_t
 &\le\frac{P(P-1)}2\frac43k^2A^{P-2}
                         \left(6\beta+\frac{9N^2}{2Bt}\right)
 \label{eq:rev-F-negative-large-remainder-step-1}\\
 &\le4P^2k^2C_{\beta}\sqrt t\,A^{P-2}
                       +\frac{3P^2k^2}{Bt}A^P\,.
 \label{eq:rev-F-negative-large-remainder-step-2}
\end{align}
\end{subequations}
Equation \eqref{eq:rev-F-negative-large-remainder-step-1} substitutes the complete curvature sum and active-coordinate variance in the Taylor remainder. Equation \eqref{eq:rev-F-negative-large-remainder-step-2} multiplies the coefficients, uses $P(P-1)\le P^2$, $\beta\le C_{\beta}\sqrt t$, and $N^2A^{P-2}\le A^P$.

If $A<12Pk\beta$, then $d_0\le4\beta+4N/B$, and every active
coordinate satisfies $Z_{t,a}\le2N+4\beta$. Therefore
\begin{subequations}\label{eq:rev-F-negative-small-remainder}
\begin{align}
 \E_{t-1}[z_a^2]
 &\le6\beta+\frac{2\{(2N+4\beta)^2+N^2\}}{Bt}
 \label{eq:rev-F-negative-small-remainder-step-1}\\
 &\le6\beta+\frac{18N^2+64\beta^2}{Bt}
 \label{eq:rev-F-negative-small-remainder-step-2}\\
 &\le6C_{\beta}\sqrt t+\frac{2592P^2k^2C_{\beta}^2+64C_{\beta}^2}{B}
 \le w_-\sqrt t\,,
 \label{eq:rev-F-negative-small-remainder-step-3}\\
 (1+2k/B)A+3k\beta
 &\le\{12P(1+2k/B)+3\}k\beta
 \le27Pk\beta\,,
 \label{eq:rev-F-negative-small-remainder-step-4}\\
 \E_{t-1}[F_{t+1}]-F_t
 &\le\frac{P(P-1)}2k^2(27Pk\beta)^{P-2}w_-\sqrt t
 \label{eq:rev-F-negative-small-remainder-step-5}\\
 &\le\tfrac12P^2k^2(27PkC_{\beta})^{P-2}w_-t^{(P-1)/2}
 =D_-t^{(P-1)/2}\,.
 \label{eq:rev-F-negative-small-remainder-step-6}
\end{align}
\end{subequations}
Equation \eqref{eq:rev-F-negative-small-remainder-step-1} uses $Z_{t,a}\le2N+4\beta$ on active coordinates. Equation \eqref{eq:rev-F-negative-small-remainder-step-2} expands with $(2N+4\beta)^2\le8N^2+32\beta^2$. Equation \eqref{eq:rev-F-negative-small-remainder-step-3} substitutes $N\le A<12Pk\beta$ and $\beta^2/t\le C_{\beta}^2$. Equation \eqref{eq:rev-F-negative-small-remainder-step-4} substitutes the small-branch bound in the curvature base. Equation \eqref{eq:rev-F-negative-small-remainder-step-5} inserts the curvature and variance in Taylor. Equation \eqref{eq:rev-F-negative-small-remainder-step-6} uses $\beta\le C_{\beta}\sqrt t$ and the definition of $D_-$.

The squared bound uses
$(2N+4\beta)^2\le8N^2+32\beta^2$.
Both branch bounds are nonnegative. Adding the bounds from
\eqref{eq:rev-F-negative-large-remainder} and
\eqref{eq:rev-F-negative-small-remainder}, and substituting
$A^P=F_t$, proves
\begin{subequations}\label{eq:rev-F-negative-recursion-substitution}
\begin{align}
 \E_{t-1}[F_{t+1}]
 &\le F_t+\frac{3P^2k^2}{Bt}F_t
            +4P^2k^2C_{\beta}\sqrt t\,F_t^{1-2/P}
            +D_-t^{(P-1)/2}
 \label{eq:rev-F-negative-recursion-substitution-step-1}\\
 &=(1+a_-/t)F_t+c_-\sqrt t\,F_t^{1-2/P}
                                      +D_-t^{(P-1)/2}\,.
 \label{eq:rev-F-negative-recursion-substitution-step-2}
\end{align}
\end{subequations}
Equation \eqref{eq:rev-F-negative-recursion-substitution-step-1} adds the nonnegative large- and small-branch remainder bounds, which therefore dominate either branch. Equation \eqref{eq:rev-F-negative-recursion-substitution-step-2} substitutes $A^P=F_t$ and the exact definitions of $a_-$ and $c_-$.

The scalar linearization is valid in both ranges:
\begin{subequations}\label{eq:rev-F-negative-scalar-linearization}
\begin{align}
 x\ge R_0
 &\Longrightarrow c_-x^{1-2/P}
       =c_-x\,x^{-2/P}\le c_-R_0^{-2/P}x\le\Lambda x\,,
 \label{eq:rev-F-negative-scalar-linearization-step-1}\\
 0\le x<R_0
 &\Longrightarrow c_-x^{1-2/P}\le c_-R_0^{1-2/P}\,,
 \label{eq:rev-F-negative-scalar-linearization-step-2}\\
 c_-x^{1-2/P}&\le\Lambda x+c_-R_0^{1-2/P}\quad(x\ge0)\,.
 \label{eq:rev-F-negative-scalar-linearization-step-3}
\end{align}
\end{subequations}
Equation \eqref{eq:rev-F-negative-scalar-linearization-step-1} uses $x^{-2/P}\le R_0^{-2/P}$ and $c_-R_0^{-2/P}\le\Lambda$. Equation \eqref{eq:rev-F-negative-scalar-linearization-step-2} uses monotonicity of $x^{1-2/P}$. Equation \eqref{eq:rev-F-negative-scalar-linearization-step-3} combines the two ranges by adding nonnegative bounds.

Since $\nu=3P/4$, direct substitution $x=F_t/t^\nu$ gives
\begin{subequations}\label{eq:rev-F-negative-linear-recursion}
\begin{align}
 c_-\sqrt t\,F_t^{1-2/P}
 &=t^{1/2+\nu(1-2/P)}c_-(F_t/t^\nu)^{1-2/P}
 \label{eq:rev-F-negative-linear-recursion-step-1}\\
 &=t^{\nu-1}c_-(F_t/t^\nu)^{1-2/P}
 \label{eq:rev-F-negative-linear-recursion-step-2}\\
 &\le(\Lambda/t)F_t+c_-R_0^{1-2/P}t^{\nu-1}\,,
 \label{eq:rev-F-negative-linear-recursion-step-3}\\
 \E_{t-1}[F_{t+1}]
 &\le(1+(a_-+\Lambda)/t)F_t
                      +(c_-R_0^{1-2/P}+D_-)t^{\nu-1}
 \label{eq:rev-F-negative-linear-recursion-step-4}\\
 &=(1+v/t)F_t+J_0t^{\nu-1}\,.
 \label{eq:rev-F-negative-linear-recursion-step-5}
\end{align}
\end{subequations}
Equation \eqref{eq:rev-F-negative-linear-recursion-step-1} makes the substitution $x=F_t/t^\nu$ without yet applying the scalar bound. Equation \eqref{eq:rev-F-negative-linear-recursion-step-2} uses $1/2+\nu(1-2/P)=\nu-1$. Equation \eqref{eq:rev-F-negative-linear-recursion-step-3} applies the scalar linearization to that normalized argument. Equation \eqref{eq:rev-F-negative-linear-recursion-step-4} inserts the resulting bound and $t^{(P-1)/2}\le t^{\nu-1}$ in the nonlinear recurrence. Equation \eqref{eq:rev-F-negative-linear-recursion-step-5} substitutes $v=a_-+\Lambda$ and $J_0=c_-R_0^{1-2/P}+D_-$.

Here $t^{(P-1)/2}\le t^{\nu-1}$.

At $n_0$, Minkowski in the product of the probability space and
counting measure over arms gives
\begin{subequations}\label{eq:rev-F-negative-initialization}
\begin{align}
 \{\E[F_{n_0}]\}^{1/P}
 &=\left\{\E\left[\sum_a(-X_{n_0,a}+2b_{n_0}^{\rm rem})_+^P\right]\right\}^{1/P}
 \label{eq:rev-F-negative-initialization-step-1}\\
 &\le\{\E[G_{n_0}]\}^{1/P}+2K^{1/P}C
 \le(16k+1)n_0+2kC\le(18k+1)n_0=u_0n_0\,.
 \label{eq:rev-F-negative-initialization-step-2}
\end{align}
\end{subequations}
Equation \eqref{eq:rev-F-negative-initialization-step-1} substitutes $W_{n_0,a}=X_{n_0,a}-2b_{n_0}^{\rm rem}$ directly into $F_{n_0}$. Equation \eqref{eq:rev-F-negative-initialization-step-2} applies Minkowski over probability and counting measure, then uses the refined absolute moment, $b_{n_0}^{\rm rem}\le C\le n_0$, and $u_0=18k+1$.

For the particular solution, convexity of $t^\nu$ gives
\begin{subequations}\label{eq:rev-F-negative-supersolutions}
\begin{align}
 H_0(t+1)^\nu-(1+v/t)H_0t^\nu
 &\ge(\nu-v)H_0t^{\nu-1}\ge J_0t^{\nu-1}\,,
 \label{eq:rev-F-negative-supersolutions-step-1}\\
 \Pi(n_0,t+1)&=(1+v/t)\Pi(n_0,t)\,.
 \label{eq:rev-F-negative-supersolutions-step-2}
\end{align}
\end{subequations}
Equation \eqref{eq:rev-F-negative-supersolutions-step-1} applies convexity of $t^\nu$ and the dictionary inequality $(\nu-v)H_0\ge J_0$. Equation \eqref{eq:rev-F-negative-supersolutions-step-2} is the exact product recursion, so adding its nonnegative homogeneous solution preserves the supersolution in \eqref{eq:rev-F-negative-supersolutions-step-1}.

Starting from \eqref{eq:rev-F-negative-initialization}, induction
using \eqref{eq:rev-F-negative-linear-recursion} and
\eqref{eq:rev-F-negative-supersolutions} gives
$\E[F_t]\le H_0t^\nu+u_0^Pn_0^P\Pi(n_0,t)$.
For $t>n\ge2$,
\begin{subequations}\label{eq:rev-F-negative-final-envelope}
\begin{align}
 \log\Pi(n,t)
 &\le v\sum_{s=n}^{t-1}s^{-1}
 \le v\int_{n-1}^{t-1}x^{-1}dx
 =v\log\frac{t-1}{n-1}\le v\log\frac{2t}n\,,
 \label{eq:rev-F-negative-final-envelope-step-1}\\
 \E[F_t]
 &\le H_0t^\nu+u_0^Pn_0^P(2t/n_0)^v
 =H_0t^\nu+2^vu_0^Pn_0^{P-v}t^v\,.
 \label{eq:rev-F-negative-final-envelope-step-2}
\end{align}
\end{subequations}
Equation \eqref{eq:rev-F-negative-final-envelope-step-1} uses $\log(1+x)\le x$, bounds the harmonic sum by its integral over $[n-1,t-1]$, and uses $(t-1)/(n-1)\le2t/n$ for $n\ge2$. Equation \eqref{eq:rev-F-negative-final-envelope-step-2} substitutes the resulting product bound into the full particular-plus-homogeneous moment envelope and collects the powers of $n_0$ and $t$.

For $t=n$ the product equals one and the same bound holds. The
argument uses $v>0$, not the stronger condition $v\ge1$.
\end{proof}

\subsubsection{Full-History Differences and Universal Absolute Moments}
The global moment recursions must start from the learner's original
initialization, including histories where the current reference is wrong.
The next lemma derives the required drift, update magnitude, and second-moment
bounds directly from the learner's estimates of cumulative loss
differences. It also establishes their fixed-time integrability before
expectation recursions are used.

\begin{lemma}[Full-history scalar difference interface]\label{lem:AB-card-full-history}
The coherent estimated cumulative loss differences $X_{t,a}$ and increments $D_{t,a}$
satisfy \eqref{eq:F-drift}--\eqref{eq:F-Dvar} on every history.
All their fixed-time moments are finite. Their laws and initialization
are those of the original learner, with no restart.
\end{lemma}
\begin{proof}[Proof of \Cref{lem:AB-card-full-history}]
For the action $a^*$ that is optimal under $\theta$ define
\begin{align}
 X_{t,a}&=S_{t-1,a}-S_{t-1,a^*}\,,& M_t&=\max_a(-X_{t,a})\,,& Z_{t,a}&=X_{t,a}+M_t\,,
 \label{eq:AB-display-256-step-1}\\
 D_{t,a}&=g_{t,a}-g_{t,a^*}\,,& e_{t,a}&=(a-a^*)^\top c_t^{\rm vec}\,.
 \label{eq:AB-display-256-step-2}
\end{align}
Equation \eqref{eq:AB-display-256-step-1} defines the estimated cumulative loss differences relative to the comparator and the reference correction. Equation \eqref{eq:AB-display-256-step-2} defines the corresponding observed increment and its predictable corruption drift. These are analysis quantities formed from the single full-history cumulative loss-estimate sequence.

Then $X_{t,a^*}=0$, $M_t\ge0$, and
\begin{align}
 X_{t+1,a}&=X_{t,a}+D_{t,a}\,,\qquad
 \E_{t-1}[D_{t,a}]=\Delta(a)+e_{t,a}\,,\quad |e_{t,a}|\le2\kappa_t\,,\label{eq:F-drift}\\
 |D_{t,a}|&\le2\beta_t+(Z_{t,a}+M_t)/B\,,\label{eq:F-Damp}\\
 \E_{t-1}[D_{t,a}^2]&\le4\beta_t+2(Z_{t,a}^2+M_t^2)/(Bt)\,,
 \qquad|\E_{t-1}[D_{t,a}]|\le2\,.\label{eq:F-Dvar}
\end{align}

Here are the substitutions from the actual estimator to these
full-history differences:
\begin{subequations}\label{eq:rev-F-full-history-interface}
\begin{align}
 \E_{t-1}[D_{t,a}]
 &=\E_{t-1}[g_{t,a}-g_{t,a^*}]
 =(a-a^*)^\top(\theta+c_t^{\rm vec})
 =\Delta(a)+e_{t,a}\,,
 \label{eq:rev-F-full-history-interface-step-1}\\
 |D_{t,a}|
 &\le|g_{t,a}|+|g_{t,a^*}|
 \le\beta_t+Z_{t,a}/B+\beta_t+Z_{t,a^*}/B
 \label{eq:rev-F-full-history-interface-step-2}\\
 &=2\beta_t+(Z_{t,a}+M_t)/B\,,
 \label{eq:rev-F-full-history-interface-step-3}\\
 \E_{t-1}[D_{t,a}^2]
 &\le2\E_{t-1}[g_{t,a}^2]+2\E_{t-1}[g_{t,a^*}^2]
 \label{eq:rev-F-full-history-interface-step-4}\\
 &\le2\{\beta_t+Z_{t,a}^2/(Bt)\}
       +2\{\beta_t+Z_{t,a^*}^2/(Bt)\}
 \label{eq:rev-F-full-history-interface-step-5}\\
 &=4\beta_t+2(Z_{t,a}^2+M_t^2)/(Bt)\,,
 \label{eq:rev-F-full-history-interface-step-6}\\
 |\E_{t-1}[D_{t,a}]|
 &=|(a-a^*)^\top\theta_t|\le2\,.
 \label{eq:rev-F-full-history-interface-step-7}
\end{align}
\end{subequations}
Equation \eqref{eq:rev-F-full-history-interface-step-1} uses difference unbiasedness and substitutes $\theta_t=\theta+c_t^{\rm vec}$. Equation \eqref{eq:rev-F-full-history-interface-step-2} applies the triangle inequality and the two update magnitude constraints. Equation \eqref{eq:rev-F-full-history-interface-step-3} uses $Z_{t,a^*}=M_t$. Equation \eqref{eq:rev-F-full-history-interface-step-4} uses $(x-y)^2\le2x^2+2y^2$ before taking conditional expectations. Equation \eqref{eq:rev-F-full-history-interface-step-5} substitutes both row-variance constraints. Equation \eqref{eq:rev-F-full-history-interface-step-6} again uses $Z_{t,a^*}=M_t$. Equation \eqref{eq:rev-F-full-history-interface-step-7} uses that each physical conditional mean lies in $[-1,1]$.

The equalities use $Z_{t,a^*}=X_{t,a^*}+M_t=M_t$.
For $A_t=\max_a|X_{t,a}|$, the deterministic update magnitude bound implies
\begin{subequations}\label{eq:rev-F-fixed-time-integrability}
\begin{align}
 R_{t+1}^{\mathrm{abs}}&\le R_t^{\mathrm{abs}}+2\beta_t+3R_t^{\mathrm{abs}}/B
                    =(1+3/B)R_t^{\mathrm{abs}}+2\beta_t\,,
 \label{eq:rev-F-fixed-time-integrability-step-1}\\
 R_t^{\mathrm{abs}}&\le2\sum_{s=1}^{t-1}\beta_s(1+3/B)^{t-1-s}<\infty\,.
 \label{eq:rev-F-fixed-time-integrability-step-2}
\end{align}
\end{subequations}
Equation \eqref{eq:rev-F-fixed-time-integrability-step-1} uses $Z_{t,a}\le2R_t^{\mathrm{abs}}$ and $M_t\le R_t^{\mathrm{abs}}$ in the increment envelope, then applies $|X_{t+1,a}|\le|X_{t,a}|+|D_{t,a}|$. Equation \eqref{eq:rev-F-fixed-time-integrability-step-2} unrolls this deterministic recursion from $R_1^{\mathrm{abs}}=0$, establishing finite fixed-time moments before any expectation induction.

Thus every fixed-time moment used below exists before its expectation
recursion is invoked.
\end{proof}

\begin{definition}[Scalar positive-moment and burn-in dictionaries]\label{def:AB-card-positive}
Set $G_t=\sum_a|X_{t,a}|^P$ and define
\begin{align}
 a_+&=10P^2k^2/B\,,\qquad c_0=2Pk\,,\qquad c_+=4P^2k^2C_{\beta}\,,\notag\\
 w_+&=4C_{\beta}+640P^2k^2C_{\beta}^2/B\,,\qquad
 D_+=\tfrac12P^2k^2(18PkC_{\beta})^{P-2}w_+\,,\notag\\
 C_+&=\max\left\{1,\left(\frac{3c_0}{P-a_+}\right)^P,
 \left(\frac{3c_+}{P-a_+}\right)^{P/2},\frac{3D_+}{P-a_+}\right\}\,. \notag
\end{align}
The denominators are positive because $a_+<40P/1024$.

The sharper initialization needed for a dimension-effective corruption coefficient uses
\begin{align}
 \nu&=3P/4\,,&v_G&=a_++P/4\,,&R_G&=(8c_+/P)^{P/2}\,,\notag\\
 J_G&=c_+R_G^{1-2/P}+D_+\,,&H_G&=\max\{1,J_G/(\nu-v_G)\}\,,&
 N_{\rm burn}&=\left\lceil\max\{2,H_G^{4/P}\}\right\rceil\,. \notag
\end{align}
\end{definition}

The absolute-moment bound in \Cref{lem:F-absolute} gives
polynomial growth, but its leading constant depends on the problem
parameters. The next refinement moves that dependence into a fixed
burn-in, leaving a universal linear coefficient for the moment root.
This controls the initialization of the negative-moment bound without
putting the coarse absolute-moment constant into the eventual $\sqrt C$ term.

\begin{lemma}[Universal linear coefficient after a fixed burn-in]\label{lem:F-refined}
For every allowed environment,
\[
 \E[G_t]\le(16k)^Pt^P+H_Gt^\nu\,,
 \qquad (\E[G_t])^{1/P}\le(16k+1)t\quad(t\ge N_{\rm burn})\,.
\]
\end{lemma}
\begin{proof}[Proof of \Cref{lem:F-refined}]
Let $A_*=(16k)^P$. Since $c_0=2Pk$, the two ranges for the linear
drift term give
\begin{subequations}\label{eq:rev-F-refined-linear-split}
\begin{align}
 x\ge A_*&\Longrightarrow
 c_0x^{1-1/P}\le c_0A_*^{-1/P}x=(P/8)x\,,
 \label{eq:rev-F-refined-linear-split-step-1}\\
 0\le x<A_*&\Longrightarrow
 c_0x^{1-1/P}\le c_0A_*^{1-1/P}=(P/8)A_*\,,
 \label{eq:rev-F-refined-linear-split-step-2}\\
 c_0x^{1-1/P}&\le(P/8)x+(P/8)A_*\,.
 \label{eq:rev-F-refined-linear-split-step-3}
\end{align}
\end{subequations}
Equation \eqref{eq:rev-F-refined-linear-split-step-1} uses $A_*^{1/P}=16k$ and $c_0=2Pk$ on $x\ge A_*$. Equation \eqref{eq:rev-F-refined-linear-split-step-2} uses monotonicity on $x<A_*$. Equation \eqref{eq:rev-F-refined-linear-split-step-3} combines both bounds into an inequality valid for every $x\ge0$.

Similarly, $R_G=(8c_+/P)^{P/2}$ gives
\begin{subequations}\label{eq:rev-F-refined-curvature-split}
\begin{align}
 x\ge R_G&\Longrightarrow
 c_+x^{1-2/P}\le c_+R_G^{-2/P}x=(P/8)x\,,
 \label{eq:rev-F-refined-curvature-split-step-1}\\
 0\le x<R_G&\Longrightarrow
 c_+x^{1-2/P}\le c_+R_G^{1-2/P}\,,
 \label{eq:rev-F-refined-curvature-split-step-2}\\
 c_+x^{1-2/P}&\le(P/8)x+c_+R_G^{1-2/P}\,.
 \label{eq:rev-F-refined-curvature-split-step-3}
\end{align}
\end{subequations}
Equation \eqref{eq:rev-F-refined-curvature-split-step-1} uses $R_G^{2/P}=8c_+/P$ on $x\ge R_G$. Equation \eqref{eq:rev-F-refined-curvature-split-step-2} uses monotonicity on the complementary range. Equation \eqref{eq:rev-F-refined-curvature-split-step-3} adds nonnegative bounds to obtain a single global inequality.

Substitute $x=G_t/t^P$ in
\eqref{eq:rev-F-refined-linear-split} and $x=G_t/t^\nu$ in
\eqref{eq:rev-F-refined-curvature-split}:
\begin{subequations}\label{eq:rev-F-refined-recursion-substitution}
\begin{align}
 c_0G_t^{1-1/P}
 &=t^{P-1}c_0(G_t/t^P)^{1-1/P}
 \le\frac P{8t}G_t+\frac P8A_*t^{P-1}\,,
 \label{eq:rev-F-refined-recursion-substitution-step-1}\\
 c_+\sqrt t\,G_t^{1-2/P}
 &=t^{\nu-1}c_+(G_t/t^\nu)^{1-2/P}
 \le\frac P{8t}G_t+c_+R_G^{1-2/P}t^{\nu-1}\,,
 \label{eq:rev-F-refined-recursion-substitution-step-2}\\
 \E_{t-1}[G_{t+1}]
 &\le\left(1+\frac{a_++P/4}{t}\right)G_t
       +\frac P8A_*t^{P-1}
       +(c_+R_G^{1-2/P}+D_+)t^{\nu-1}
 \label{eq:rev-F-refined-recursion-substitution-step-3}\\
 &=(1+v_G/t)G_t+(P/8)A_*t^{P-1}+J_Gt^{\nu-1}\,.
 \label{eq:rev-F-refined-recursion-substitution-step-4}
\end{align}
\end{subequations}
Equation \eqref{eq:rev-F-refined-recursion-substitution-step-1} substitutes $G_t/t^P$ in the first scalar inequality and restores the factor $t^{P-1}$. Equation \eqref{eq:rev-F-refined-recursion-substitution-step-2} substitutes $G_t/t^\nu$ in the second scalar inequality and uses $1/2+\nu(1-2/P)=\nu-1$. Equation \eqref{eq:rev-F-refined-recursion-substitution-step-3} inserts both complete bounds in the absolute-moment recurrence and uses $t^{(P-1)/2}\le t^{\nu-1}$. Equation \eqref{eq:rev-F-refined-recursion-substitution-step-4} identifies $v_G$ and $J_G$.

For $f(t)=A_*t^P+H_Gt^\nu$, convexity of both powers yields
\begin{subequations}\label{eq:rev-F-refined-supersolution}
\begin{align}
 f(t+1)-(1+v_G/t)f(t)
 &\ge(P-v_G)A_*t^{P-1}+(\nu-v_G)H_Gt^{\nu-1}
 \label{eq:rev-F-refined-supersolution-step-1}\\
 &\ge(P/8)A_*t^{P-1}+J_Gt^{\nu-1}\,.
 \label{eq:rev-F-refined-supersolution-step-2}
\end{align}
\end{subequations}
Equation \eqref{eq:rev-F-refined-supersolution-step-1} applies the tangent inequalities for the two powers in $f(t)=A_*t^P+H_Gt^\nu$. Equation \eqref{eq:rev-F-refined-supersolution-step-2} uses $P-v_G\ge P/8$ and $(\nu-v_G)H_G\ge J_G$, so $f$ dominates every source of the linear recurrence.

Indeed $v_G=a_++P/4<\nu$, so $P-v_G>P/4>P/8$,
and $(\nu-v_G)H_G\ge J_G$ by definition.
Since $G_1=0\le f(1)$, taking expectations in
\eqref{eq:rev-F-refined-recursion-substitution} and using
\eqref{eq:rev-F-refined-supersolution} proves $\E[G_t]\le f(t)$
by induction. For $t\ge N_{\rm burn}\ge H_G^{4/P}$,
\begin{subequations}\label{eq:rev-F-refined-root-bound}
\begin{align}
 \{\E[G_t]\}^{1/P}
 &\le\{(16k)^Pt^P+H_Gt^{3P/4}\}^{1/P}
 \label{eq:rev-F-refined-root-bound-step-1}\\
 &\le16kt+H_G^{1/P}t^{3/4}
 =16kt+t(H_G^{4/P}/t)^{1/4}
 \label{eq:rev-F-refined-root-bound-step-2}\\
 &\le(16k+1)t\,.
 \label{eq:rev-F-refined-root-bound-step-3}
\end{align}
\end{subequations}
Equation \eqref{eq:rev-F-refined-root-bound-step-1} substitutes the complete refined envelope before taking its root. Equation \eqref{eq:rev-F-refined-root-bound-step-2} uses subadditivity of the $1/P$ power and rewrites the lower-order term to expose $H_G^{4/P}/t$. Equation \eqref{eq:rev-F-refined-root-bound-step-3} uses $t\ge N_{\rm burn}\ge H_G^{4/P}$.

The index $N_{\rm burn}$ is used only in this analysis; the executed
full-history cumulative loss-estimate sequence is unchanged.
\end{proof}

We finish by proving the global absolute-moment estimate used above.
It controls the size of the full-history cumulative loss estimates on every environment and
provides the one-step recurrence refined in \Cref{lem:F-refined}.
Unlike the local ratio estimates, it does not require the optimal action $a^*$ to have
been correctly identified.

\begin{lemma}[Absolute moments on every environment]\label{lem:F-absolute}
For $G_t=\sum_a|X_{t,a}|^P$,
\begin{align}
 \E_{t-1}[G_{t+1}]\le(1+a_+/t)G_t+c_0G_t^{1-1/P}
       +c_+\sqrt t\,G_t^{1-2/P}+D_+t^{(P-1)/2}\,, \notag
\end{align}
and $\E[G_t]\le C_+t^P$.
\end{lemma}
\begin{proof}[Proof of \Cref{lem:F-absolute}]
Fix the history and set $A=G_t^{1/P}$. Since $M_t\le A$ and
$Z_{t,a}\le2A$, substitution in
\eqref{eq:F-Damp}--\eqref{eq:F-Dvar} in \Cref{lem:AB-card-full-history} gives
\begin{subequations}\label{eq:rev-F-absolute-step-bounds}
\begin{align}
 |D_{t,a}|&\le2\beta_t+3A/B\,,
 \label{eq:rev-F-absolute-step-bounds-step-1}\\
 \E_{t-1}[D_{t,a}^2]
 &\le4\beta_t+2\{(2A)^2+A^2\}/(Bt)
 =4\beta_t+10A^2/(Bt)\,.
 \label{eq:rev-F-absolute-step-bounds-step-2}
\end{align}
\end{subequations}
Equation \eqref{eq:rev-F-absolute-step-bounds-step-1} uses $Z_{t,a}\le2A$ and $M_t\le A$. Equation \eqref{eq:rev-F-absolute-step-bounds-step-2} substitutes these two bounds separately into the second-moment envelope and computes $2(4+1)=10$.

Set $d_t=2\beta_t+3A/B$, a deterministic bound conditional on
$\mathcal F_{t-1}$. The expected linear Taylor term and the sum of
curvature factors obey
\begin{subequations}\label{eq:rev-F-absolute-taylor-terms}
\begin{align}
 P\sum_a|X_{t,a}|^{P-2}X_{t,a}\E_{t-1}[D_{t,a}]
 &\le2P\sum_a|X_{t,a}|^{P-1}
 \label{eq:rev-F-absolute-taylor-terms-step-1}\\
 &\le2PK^{1/P}\left(\sum_a|X_{t,a}|^P\right)^{(P-1)/P}
 =2PkA^{P-1}\,,
 \label{eq:rev-F-absolute-taylor-terms-step-2}\\
 \sum_a(|X_{t,a}|+d_t)^{P-2}
 &\le k^2\{A+k(2\beta_t+3A/B)\}^{P-2}
 \label{eq:rev-F-absolute-taylor-terms-step-3}\\
 &=k^2\{(1+3k/B)A+2k\beta_t\}^{P-2}\,.
 \label{eq:rev-F-absolute-taylor-terms-step-4}
\end{align}
\end{subequations}
Equation \eqref{eq:rev-F-absolute-taylor-terms-step-1} uses $|\E_{t-1}[D_{t,a}]|\le2$. Equation \eqref{eq:rev-F-absolute-taylor-terms-step-2} applies H\"older on the $K$ coordinates and substitutes $k=K^{1/P}$. Equation \eqref{eq:rev-F-absolute-taylor-terms-step-3} applies the finite-coordinate curvature inequality of \Cref{lem:taylor} with increment envelope $2\beta_t+3A/B$. Equation \eqref{eq:rev-F-absolute-taylor-terms-step-4} distributes the factor $k$ inside the curvature base.

If $A\ge8Pk\beta_t$, then
\begin{subequations}\label{eq:rev-F-absolute-large-branch}
\begin{align}
 (1+3k/B)A+2k\beta_t
 &\le A(1+1/(4P)+1/(4P))=A(1+1/(2P))\,,
 \label{eq:rev-F-absolute-large-branch-step-1}\\
 \{(1+3k/B)A+2k\beta_t\}^{P-2}
 &\le A^{P-2}e^{(P-2)/(2P)}\le2A^{P-2}\,,
 \label{eq:rev-F-absolute-large-branch-step-2}\\
 \frac{P(P-1)}2\sum_a
      (|X_{t,a}|+d_t)^{P-2}\E_{t-1}[D_{t,a}^2]
 &\le P^2k^2A^{P-2}\{4\beta_t+10A^2/(Bt)\}
 \label{eq:rev-F-absolute-large-branch-step-3}\\
 &\le4P^2k^2C_{\beta}\sqrt t\,A^{P-2}
                         +\frac{10P^2k^2}{Bt}A^P\,.
 \label{eq:rev-F-absolute-large-branch-step-4}
\end{align}
\end{subequations}
Equation \eqref{eq:rev-F-absolute-large-branch-step-1} uses $3k/B\le1/(4P)$ and $2k\beta_t/A\le1/(4P)$. Equation \eqref{eq:rev-F-absolute-large-branch-step-2} uses $(1+x)^{P-2}\le e^{(P-2)x}$ and $e^{1/2}<2$. Equation \eqref{eq:rev-F-absolute-large-branch-step-3} inserts that curvature sum and the raw variance into Taylor. Equation \eqref{eq:rev-F-absolute-large-branch-step-4} distributes the product and uses $\beta_t\le C_{\beta}\sqrt t$.

If $A<8Pk\beta_t$, then instead
\begin{subequations}\label{eq:rev-F-absolute-small-branch}
\begin{align}
 (1+3k/B)A+2k\beta_t&\le18Pk\beta_t\,,
 \label{eq:rev-F-absolute-small-branch-step-1}\\
 \E_{t-1}[D_{t,a}^2]
 &\le4\beta_t+640P^2k^2\beta_t^2/(Bt)
 \label{eq:rev-F-absolute-small-branch-step-2}\\
 &\le4C_{\beta}\sqrt t+640P^2k^2C_{\beta}^2/B\le w_+\sqrt t\,,
 \label{eq:rev-F-absolute-small-branch-step-3}\\
 \frac{P(P-1)}2\sum_a
      (|X_{t,a}|+d_t)^{P-2}\E_{t-1}[D_{t,a}^2]
 &\le\tfrac12P^2k^2(18Pk\beta_t)^{P-2}w_+\sqrt t
 \label{eq:rev-F-absolute-small-branch-step-4}\\
 &\le\tfrac12P^2k^2(18PkC_{\beta})^{P-2}w_+t^{(P-1)/2}
 =D_+t^{(P-1)/2}\,.
 \label{eq:rev-F-absolute-small-branch-step-5}
\end{align}
\end{subequations}
Equation \eqref{eq:rev-F-absolute-small-branch-step-1} substitutes $A<8Pk\beta_t$ into the curvature base. Equation \eqref{eq:rev-F-absolute-small-branch-step-2} inserts its square in the variance term. Equation \eqref{eq:rev-F-absolute-small-branch-step-3} uses $\beta_t\le C_{\beta}\sqrt t$, $\beta_t^2/t\le C_{\beta}^2$, and the definition of $w_+$. Equation \eqref{eq:rev-F-absolute-small-branch-step-4} inserts both bounds into the remainder. Equation \eqref{eq:rev-F-absolute-small-branch-step-5} collects $(\sqrt t)^{P-2}\sqrt t=t^{(P-1)/2}$ and substitutes $D_+$.

Using \eqref{eq:taylor-absolute} in \Cref{lem:taylor}, and adding the nonnegative bounds
from both branches, gives the complete substitution
\begin{subequations}\label{eq:rev-F-absolute-recursion-final}
\begin{align}
 \E_{t-1}[G_{t+1}]
 &\le G_t+2PkA^{P-1}+4P^2k^2C_{\beta}\sqrt t\,A^{P-2}
               +\frac{10P^2k^2}{Bt}A^P+D_+t^{(P-1)/2}
 \label{eq:rev-F-absolute-recursion-final-step-1}\\
 &=(1+a_+/t)G_t+c_0G_t^{1-1/P}
               +c_+\sqrt t\,G_t^{1-2/P}+D_+t^{(P-1)/2}\,.
 \label{eq:rev-F-absolute-recursion-final-step-2}
\end{align}
\end{subequations}
Equation \eqref{eq:rev-F-absolute-recursion-final-step-1} adds the linear Taylor contribution and both nonnegative branch remainder bounds. Equation \eqref{eq:rev-F-absolute-recursion-final-step-2} substitutes $A^P=G_t$, $a_+=10P^2k^2/B$, $c_0=2Pk$, and $c_+=4P^2k^2C_{\beta}$ into every term.

For the expectation bound, $G_1=0\le C_+$. Suppose
$\E[G_t]\le C_+t^P$. Jensen's inequality for the powers
$1-1/P$ and $1-2/P$, followed by this induction hypothesis, gives
\begin{subequations}\label{eq:rev-F-absolute-induction-substitution}
\begin{align}
 \E[G_{t+1}]
 &\le(1+a_+/t)\E[G_t]
      +c_0\{\E[G_t]\}^{1-1/P}
      +c_+\sqrt t\,\{\E[G_t]\}^{1-2/P}
      +D_+t^{(P-1)/2}
 \label{eq:rev-F-absolute-induction-substitution-step-1}\\
 &\le C_+t^P+a_+C_+t^{P-1}
      +c_0C_+^{1-1/P}t^{P-1}
      +c_+C_+^{1-2/P}t^{P-3/2}
      +D_+t^{(P-1)/2}
 \label{eq:rev-F-absolute-induction-substitution-step-2}\\
 &\le C_+t^P+
      \{a_+C_++c_0C_+^{1-1/P}+c_+C_+^{1-2/P}+D_+\}t^{P-1}\,.
 \label{eq:rev-F-absolute-induction-substitution-step-3}
\end{align}
\end{subequations}
Equation \eqref{eq:rev-F-absolute-induction-substitution-step-1} takes total expectations and uses concavity of $x^{1-1/P}$ and $x^{1-2/P}$. Equation \eqref{eq:rev-F-absolute-induction-substitution-step-2} substitutes the induction hypothesis $\E [G_t]\le C_+t^P$ in every term. Equation \eqref{eq:rev-F-absolute-induction-substitution-step-3} uses $P-3/2\le P-1$ and $(P-1)/2\le P-1$ to collect a common factor $t^{P-1}$.

The definition of $C_+$ supplies each coefficient separately:
\begin{subequations}\label{eq:rev-F-absolute-induction-completion}
\begin{align}
 C_+^{1/P}\ge\frac{3c_0}{P-a_+}
 &\Longrightarrow c_0C_+^{1-1/P}\le\tfrac13(P-a_+)C_+\,,
 \label{eq:rev-F-absolute-induction-completion-step-1}\\
 C_+^{2/P}\ge\frac{3c_+}{P-a_+}
 &\Longrightarrow c_+C_+^{1-2/P}\le\tfrac13(P-a_+)C_+\,,
 \label{eq:rev-F-absolute-induction-completion-step-2}\\
 C_+\ge\frac{3D_+}{P-a_+}
 &\Longrightarrow D_+\le\tfrac13(P-a_+)C_+\,,
 \label{eq:rev-F-absolute-induction-completion-step-3}\\
 \E[G_{t+1}]
 &\le C_+t^P+\{a_+C_++(P-a_+)C_+\}t^{P-1}
 \label{eq:rev-F-absolute-induction-completion-step-4}\\
 &=C_+t^P+PC_+t^{P-1}\le C_+(t+1)^P\,.
 \label{eq:rev-F-absolute-induction-completion-step-5}
\end{align}
\end{subequations}
Equations \eqref{eq:rev-F-absolute-induction-completion-step-1}, \eqref{eq:rev-F-absolute-induction-completion-step-2}, and \eqref{eq:rev-F-absolute-induction-completion-step-3} separately bound the three nonhomogeneous coefficients using the definition of $C_+$. Equation \eqref{eq:rev-F-absolute-induction-completion-step-4} substitutes all three bounds into the previous induction inequality. Equation \eqref{eq:rev-F-absolute-induction-completion-step-5} collects $a_++P-a_+=P$ and uses $(t+1)^P\ge t^P+Pt^{P-1}$.

This proves the claimed bound for every $t$.
\end{proof}

\clearpage
\section{Proof of Theorem~\ref*{thm:main-finite-lower}}\label{ch:finite-lower}

This appendix proves \Cref{thm:main-finite-lower}
on $\cA$ and the one-dimensional extension on $\cA^{\prime}$ discussed after
its proof sketch, and derives their restrictions on corruption coefficients
and finite transients. Both constructions use bounded feedback.

\subsection{Two Fixed Geometries and Their Exact Information Costs}
For $\sigma\in\{-1,1\}$ and $0<\Delta_{\min}\le1/2$, consider either
\[
 \cA^{\prime}=\{-1,1\}\,,\quad\theta_\sigma=\sigma\Delta_{\min}/2\,,
\]
or
\[
 \cA=\{-e_1,e_1,e_2\}\,,\quad\theta_\sigma=(\sigma\Delta_{\min}/2,1/2)\,.
\]

The unique optimal action $a^*$ is $-\sigma$ or $-\sigma e_1$. On $\cA^{\prime}$,
the other arm has gap $\Delta_{\min}$. A sampling allocation placing mass $2/\Delta_{\min}^2$
on $a^*$ and zero on the other arm satisfies
\begin{align}
 H_n&=2/\Delta_{\min}^2>0\,,\qquad
 a^\top H_n^{-1}a=\Delta_{\min}^2/2\quad(a\ne a^*)\,,\notag\\
 \sum_{a\ne a^*}n_a\Delta(a)&=0\,,
 \qquad 0\le \cstar(\cA^{\prime},\theta_\sigma)\le0\,.
 \label{eq:rev-C-zero-information-design}
\end{align}
Thus $\cstar(\cA^{\prime},\theta_\sigma)=0$ exactly.
On $\cA$, let $n_-,n_+,n_2$ be the masses on
$-e_1,e_1,e_2$. The gaps of $\sigma e_1$ and $e_2$ are
$\Delta_{\min}$ and $g_{\Delta_{\min}}=(1+\Delta_{\min})/2$, respectively, so
\begin{align}
 H_n&=\begin{pmatrix}n_{-e_1}+n_{e_1}&0\\0&n_{e_2}\end{pmatrix}\,,\notag\\
 (\sigma e_1)^\top H_n^{-1}(\sigma e_1)
 &=\frac1{n_{-e_1}+n_{e_1}}\le\frac{\Delta_{\min}^2}2
 \quad\Longleftrightarrow\quad n_{-e_1}+n_{e_1}\ge2/\Delta_{\min}^2\,,\notag\\
 e_2^\top H_n^{-1}e_2
 &=\frac1{n_{e_2}}\le\frac{g_{\Delta_{\min}}^2}2
 \quad\Longleftrightarrow\quad n_{e_2}\ge2/g_{\Delta_{\min}}^2\,,\notag\\
 c^{n}&=\Delta_{\min} n_{\sigma e_1}+g_{\Delta_{\min}} n_{e_2}
 \ge g_{\Delta_{\min}} n_{e_2}\ge2/g_{\Delta_{\min}}\,.
 \label{eq:rev-C-positive-information-design}
\end{align}
Taking $n_{-\sigma e_1}=2/\Delta_{\min}^2$,
$n_{\sigma e_1}=0$, and $n_{e_2}=2/g_{\Delta_{\min}}^2$ attains equality in
the last line of \eqref{eq:rev-C-positive-information-design}.
Consequently
\begin{equation*}
 \cstar(\cA,\theta_\sigma)=2/g_{\Delta_{\min}}=4/(1+\Delta_{\min})\in[8/3,4)\,. \notag
\end{equation*}
The physical action sets are fixed as $\Delta_{\min}$ varies, so a constant
depending only on their geometry cannot absorb an inverse-gap factor.

\subsection{Corruption Lower Bound}
For an integer $T\ge\Delta_{\min}^{-2}$, set $\epsilon=1/(8\sqrt T)$.
At each round use the actual parameter $v_\sigma=\sigma\epsilon$ on
$\cA^{\prime}$, or $v_\sigma=(\sigma\epsilon,1/2)$ on $\cA$.
For either action set $\mathcal X\in\{\cA,\cA^{\prime}\}$,
the action-wise corruption and its sum are
\begin{align}
 0<\epsilon&=\frac1{8\sqrt T}\le\frac1{8\sqrt{\Delta_{\min}^{-2}}}
             =\Delta_{\min}/8\,,\notag\\
 \max_{a\in\mathcal X}|\langle a,v_\sigma-\theta_\sigma\rangle|
 &=|\sigma(\epsilon-\Delta_{\min}/2)|=\Delta_{\min}/2-\epsilon\,,
 \notag\\
 C_{\Delta_{\min},T}
 &=\sum_{t=1}^T\max_{a\in\mathcal X}
                 |\langle a,v_\sigma-\theta_\sigma\rangle|
   =T(\Delta_{\min}/2-\epsilon)\notag\\
 &\ge T(\Delta_{\min}/2-\Delta_{\min}/8)=3\Delta_{\min} T/8\,,
 \qquad C_{\Delta_{\min},T}\le T\Delta_{\min}/2\,.
 \label{eq:lower-budget}
\end{align}
In the maximum, $v_\sigma$ and $\theta_\sigma$ have the dimension
of the corresponding action set $\mathcal X$.
The optimal actions before and after corruption are both $-\sigma$ on $\cA^{\prime}$
and $-\sigma e_1$ on $\cA$.
Let $Z_t$ be independent signs of mean $\sigma\epsilon$, independent
of the learner's past, and give feedback $aZ_t$ when the learner
selects $a\in\{-1,1\}$ or $ae_1$. On $e_2$ give an independent
sign of mean $1/2$. For this feedback $Y_t$,
\begin{equation*}\begin{aligned}
 |Y_t|&=1\,,\\
 \E[Y_t\mid\cF_{t-1},A_t=a]
 &=a\sigma\epsilon=\langle a,v_\sigma\rangle
       &&(\cA^{\prime})\,,\\
 \E[Y_t\mid\cF_{t-1},A_t=ae_1]
 &=a\sigma\epsilon=\langle ae_1,v_\sigma\rangle
       &&(\cA)\,,\\
 \E[Y_t\mid\cF_{t-1},A_t=e_2]
 &=1/2=\langle e_2,v_\sigma\rangle
       &&(\cA)\,.
\end{aligned} \notag
\end{equation*}

The experiment is indistinguishable in total variation by \Cref{lem:finite-sign-experiment}; its proof is given after the two main claims.
\begin{proof}[The testing lower bound in \Cref{thm:main-finite-lower}]
Let $P_\sigma$ denote the full history law in the environment of
sign $\sigma$, and set $E_t=\{A_t=+1\}$ on $\cA^{\prime}$ or
$E_t=\{A_t=+e_1\}$ on $\cA$. For the actual parameter,
the per-action gaps satisfy
\begin{align}
 \Delta_+^{\rm act}(A_t)&\ge2\epsilon\ind_{E_t}\,,\notag\\
 \Delta_-^{\rm act}(A_t)&\ge2\epsilon\ind_{E_t^c}\,.
 \notag
\end{align}
For the second line on $\cA$, the possible action $e_2$ has gap
$1/2+\epsilon\ge2\epsilon$, while $-e_1$ has gap $2\epsilon$.
The optimal action $a^*$ is fixed over time and is the same before and after corruption, so actual regret is the sum of these expected
gaps. Apply \Cref{lem:finite-sign-experiment} to the
specific event $E_t$:
\begin{align}
 P_+(E_t)+P_-(E_t^c)
 &=1+P_+(E_t)-P_-(E_t)\notag\\
 &\ge1-|P_+(E_t)-P_-(E_t)|
 \ge1-\TV(P_+,P_-)\ge\tfrac12\,,\notag\\
 \Reg[T,+]+\Reg[T,-]
 &=\sum_{t=1}^T\E_+[\Delta_+^{\rm act}(A_t)]
     +\sum_{t=1}^T\E_-[\Delta_-^{\rm act}(A_t)]\notag\\
 &\ge2\epsilon\sum_{t=1}^T\{P_+(E_t)+P_-(E_t^c)\}
 \ge2\epsilon\sum_{t=1}^T\tfrac12=\epsilon T\notag\\
 &=\frac{T}{8\sqrt T}=\frac{\sqrt T}8\,,\notag\\
 \max_{\sigma=\pm1}\Reg[T,\sigma]
 &\ge\frac12(\Reg[T,+]+\Reg[T,-])
 \ge\frac{\sqrt T}{16}\,.
 \notag
\end{align}
Since $C_{\Delta_{\min},T}\le\Delta_{\min} T/2$, the equivalent budget comparison is
\begin{align}
 T&\ge2C_{\Delta_{\min},T}/\Delta_{\min}\,,\qquad
 \frac{\sqrt T}{16}
 \ge\frac{\sqrt2}{16}\sqrt{C_{\Delta_{\min},T}/\Delta_{\min}}\,.
 \notag
\end{align}
This proves the stated lower bound against the actual best fixed arm.
\end{proof}

\begin{proof}[Proof of \Cref{thm:main-finite-lower} for any given budget cap $C$]
Let $C\ge\Delta_{\min}^{-1}$ and $T\ge\lceil\Delta_{\min}^{-2}\rceil$, as in
the claim, and set $m=\min\{T,\lfloor2C/\Delta_{\min}\rfloor\}$.
Because $C/\Delta_{\min}\ge\Delta_{\min}^{-2}\ge4$,
\begin{align}
 \lfloor2C/\Delta_{\min}\rfloor
 &\ge2C/\Delta_{\min}-1\ge C/\Delta_{\min}\ge\Delta_{\min}^{-2}\,,\notag\\
 m&\ge\Delta_{\min}^{-2}\,,\qquad
 m\ge\min\{T,C/\Delta_{\min}\}\,.
 \notag
\end{align}
Run the preceding experiment with $\epsilon_m=1/(8\sqrt m)$
for $m$ rounds and use the fixed loss vector $\theta_\sigma$ thereafter. The total
pathwise corruption is
\begin{align}
 \sum_{t=1}^T\kappa_t
 &=\sum_{t=1}^{m}(\Delta_{\min}/2-\epsilon_m)
       +\sum_{t=m+1}^T0\notag\\
 &=m(\Delta_{\min}/2-\epsilon_m)
 \le m\Delta_{\min}/2\le\lfloor2C/\Delta_{\min}\rfloor\Delta_{\min}/2\le C\,.
 \notag
\end{align}
For each sign the optimal arm is unchanged in the prefix and suffix;
all suffix gaps are nonnegative. Thus the prefix testing argument,
which also applies to a policy using the announced full horizon $T$,
gives
\begin{align}
 \max_\sigma \Reg[T,\sigma]
 &\ge\max_\sigma \Reg[m,\sigma]
 \ge\frac{\sqrt m}{16}
 \ge\frac1{16}\sqrt{\min\{T,C/\Delta_{\min}\}}\,.
 \notag
\end{align}
\end{proof}
\subsection{Testing Estimate Used in the Corruption Proof}
\begin{lemma}[Total variation of the full experiment]\label{lem:finite-sign-experiment}
If $Q_\pm$ are sign laws of means $\pm\epsilon$, then
\[
 \TV(Q_+^{\otimes T},Q_-^{\otimes T})
 \le\tfrac12\sqrt{e^{4/63}-1}<\tfrac12\,.
\]
The same bound holds for the adaptive history, including actions and algorithm randomness.
\end{lemma}
\begin{proof}
For $z\in\{-1,1\}$, define $Q_\pm(z)=(1\pm z\epsilon)/2$.
A direct two-point calculation gives
\begin{align}
 \chi^2(Q_-\Vert Q_+)
 &=\sum_{z=\pm1}\frac{\{Q_-(z)-Q_+(z)\}^2}{Q_+(z)}\notag\\
 &=\frac{\epsilon^2}{(1+\epsilon)/2}
                  +\frac{\epsilon^2}{(1-\epsilon)/2}
 =\frac{4\epsilon^2}{1-\epsilon^2}\,.
 \label{eq:rev-C-single-sign-divergence}
\end{align}
For $L_T=\prod_{t=1}^TQ_-(Z_t)/Q_+(Z_t)$, independence implies
\begin{align}
 \E_+[L_T]&=1\,,\qquad
 \E_+[L_T^2]=\prod_{t=1}^T
         \E_+\!\left[\{Q_-(Z_t)/Q_+(Z_t)\}^2\right]\notag\\
 &=\left(1+\frac{4\epsilon^2}{1-\epsilon^2}\right)^T\,,\notag\\
 \E_+[(L_T-1)^2]
 &=\E_+[L_T^2]-2\E_+[L_T]+1\notag\\
 &=\left(1+\frac{4\epsilon^2}{1-\epsilon^2}\right)^T-1
 \le\exp\!\left(\frac{4T\epsilon^2}{1-\epsilon^2}\right)-1\notag\\
 &=\exp\!\left(\frac{1/16}{1-1/(64T)}\right)-1
 \le e^{4/63}-1\,,\notag\\
 \TV(Q_+^{\otimes T},Q_-^{\otimes T})
 &=\tfrac12\E_+[|L_T-1|]
 \le\tfrac12\{\E_+[(L_T-1)^2]\}^{1/2}\notag\\
 &\le\tfrac12\sqrt{e^{4/63}-1}<\tfrac12\,.
 \notag
\end{align}
Here $\epsilon^2=1/(64T)$ and $1-1/(64T)\ge63/64$.

To pass from the latent signs to an adaptive history, include all
auxiliary $e_2$ feedback signs and the learner's private randomness
in a random variable $U$ with common law $\mu$, independent of the
latent signs. Given $(Z_{1:T},U)$, the history is produced recursively
by one common measurable kernel $K$. For any output event $E$, let
$f_E(z)=\int K((z,u),E)\,\mu(du)\in[0,1]$. Then
\begin{align}
 |P_+(E)-P_-(E)|
 &=\left|\int f_E\,dQ_+^{\otimes T}
                        -\int f_E\,dQ_-^{\otimes T}\right|\notag\\
 &=\left|\int_0^1\left\{
       Q_+^{\otimes T}(f_E>x)-Q_-^{\otimes T}(f_E>x)
                           \right\}\,dx\right|\notag\\
 &\le\int_0^1\TV(Q_+^{\otimes T},Q_-^{\otimes T})\,dx
 =\TV(Q_+^{\otimes T},Q_-^{\otimes T})\,,\notag\\
 \TV(P_+,P_-)&=\sup_E|P_+(E)-P_-(E)|
 \le\tfrac12\sqrt{e^{4/63}-1}<\tfrac12\,.
 \label{eq:rev-C-adaptive-kernel-contraction}
\end{align}
This includes both actions and algorithm randomness in the output
history, as asserted.
\end{proof}

\subsection{Consequences for Corruption Coefficients and Transients}
\begin{corollary}[Consequences for finite-action corruption coefficients]\label{cor:finite-corruption-consequences}
Consider the corruption family on $\cA$ in
\Cref{thm:main-finite-lower} and its one-dimensional
extension on $\cA^{\prime}$ proved above.
The following conclusions concern guarantees uniform over their fixed loss vectors,
horizons, and allowed corruption environments. All transients and
instance-dependent coefficients below are finite and independent of $T,C$.

\textup{(i) Necessary corruption exponent.}
On either family, no bound
\[
 \Reg\le A \cstar\log(eT)+M(\theta)+b(\theta)C^\alpha
\]
with fixed finite $A\ge0$, $\alpha<1/2$, and arbitrary nonnegative $M,b$
holds for every fixed loss vector, $T$, and $C\ge1$.

\textup{(ii) Necessary inverse-gap dependence.}
On $\cA^{\prime}$, no finite gap-independent $B$ permits
$\Reg\le M(\theta)+B\sqrt{C\{1+\cstar\log(eT)\}}$
for every fixed loss vector, $T$, and $C$, even with an arbitrary finite $T,C$-independent transient $M$.
On $\cA$, no bound
$A \cstar\log(eT)+M(\theta)+B\Delta_{\min}^{-\rho}\sqrt C$
with fixed finite $A,B$ and such an arbitrary $M$ holds uniformly for any $\rho<1/2$.

\textup{(iii) Transient cost of a geometric square-root coefficient.}
If, on $\cA$, a policy satisfies for fixed finite $A,B\ge0$
\[
 \Reg\le A \cstar\log(eT)+M(\theta)
      +B\sqrt{C\{1+\cstar\log(eT)\}}
\]
for every fixed loss vector, $T$, and $C$, then, with $a_B=[2^{16}(1+B)^2]^{-1}$, every sufficiently small $\Delta_{\min}$ satisfies
\[
 \max_{\sigma=\pm1}M(\theta_\sigma)\ge\frac1{64}\exp\{a_B/(2\Delta_{\min})\}\,.
\]
In particular no fixed-degree polynomial transient in $\Delta_{\min}^{-1}$ suffices. More generally, replacing the corruption term by $B\sqrt C\,[\log(eT)]^q$ for any fixed finite $q\ge0$ still cannot yield a uniform bound with a polynomial transient on this bounded-$\cstar$ family.
\end{corollary}
The following arguments prove \Cref{cor:finite-corruption-consequences} by reusing the two corruption experiments established above. For part~\textup{(i)} of the corollary, any fixed gap suffices; for part~\textup{(ii)}, a sufficiently small gap is held fixed while the horizon tends to infinity; for part~\textup{(iii)}, the gap and horizon are coupled to reveal the transient required by an extra logarithmic factor. In each argument the maximum of the transient constants for the two fixed loss vectors is used, because the hard sign may depend on the horizon. No claim is made that one chosen sign is hard at every horizon. Taking $T\to\infty$ compares inequalities valid at every finite horizon, with a possibly different corrupted environment and budget at each horizon. Every contradiction can equivalently be obtained by choosing a sufficiently large finite $T$.

\begin{proof}[Proof of \Cref{cor:finite-corruption-consequences}\textup{(i)}]
Suppose the proposed bound holds. Fix any $\Delta_{\min}\in(0,1/2]$ and
let
$M_{\max}(\Delta_{\min})=\max_{\sigma=\pm1}M(\theta_\sigma)$ and
$b_{\max}(\Delta_{\min})=\max_{\sigma=\pm1}b(\theta_\sigma)$.
Both numbers are finite constants as $T$ varies. For every sufficiently
large integer $T$, \eqref{eq:lower-budget} gives
$C_{\Delta_{\min},T}\ge3\Delta_{\min} T/8\ge1$. Applying the asserted upper bound
to both environments of the testing experiment yields
\begin{align}
 \frac{\sqrt T}{16}
 &\le\max_\sigma \Reg[T,\sigma]\notag\\
 &\le4A\log(eT)+M_{\max}(\Delta_{\min})
                              +b_{\max}(\Delta_{\min})C_{\Delta_{\min},T}^\alpha\notag\\
 &=4A\log(eT)+M_{\max}(\Delta_{\min})
       +b_{\max}(\Delta_{\min})
               \{T(\Delta_{\min}/2-1/(8\sqrt T))\}^\alpha\,,\notag\\
 \frac1{16}
 &\le\frac{4A\log(eT)+M_{\max}(\Delta_{\min})}{\sqrt T}
       +b_{\max}(\Delta_{\min})
          \left(\frac{\Delta_{\min}}2-\frac1{8\sqrt T}\right)^\alpha
                         T^{\alpha-1/2}\,.
 \notag
\end{align}
The penultimate line substitutes the budget exactly, rather than
using an inequality whose direction could depend on the sign of
$\alpha$. The limits are
\begin{align}
 \frac{4A\log(eT)+M_{\max}(\Delta_{\min})}{\sqrt T}&\longrightarrow0\,,\notag\\
 \left(\frac{\Delta_{\min}}2-\frac1{8\sqrt T}\right)^\alpha
 &\longrightarrow(\Delta_{\min}/2)^\alpha\in(0,\infty)\,,\qquad
 T^{\alpha-1/2}\longrightarrow0\,,\notag\\
 \frac1{16}&\le0\,.
 \notag
\end{align}
The last inequality is impossible. This argument permits any finite
$b(\theta)$, including $B\Delta_{\min}^{-\rho}$ for any fixed finite
$B\ge0$ and $\rho\in\R$, and requires no control of $M$ or $b$
as $\Delta_{\min}$ varies. The condition $C\ge1$ avoids any convention for
$0^\alpha$ and does not weaken the contradiction to an all-budget
upper bound.
\end{proof}

\begin{proof}[Proof of \Cref{cor:finite-corruption-consequences}\textup{(ii)}]
On $\cA^{\prime}$, choose
$0<\Delta_{\min}<\min\{1/2,[512(1+B)^2]^{-1}\}$ and set
$M_{\max}(\Delta_{\min})=\max_\sigma M(\theta_\sigma)<\infty$.
Since $\cstar=0$ and $C_{\Delta_{\min},T}\le\Delta_{\min} T/2$, the proposed bound
would imply, for every sufficiently large $T$,
\begin{align}
 \frac{\sqrt T}{16}
 &\le\max_\sigma \Reg[T,\sigma]
 \le M_{\max}(\Delta_{\min})
                 +B\sqrt{C_{\Delta_{\min},T}\{1+0\log(eT)\}}\notag\\
 &\le M_{\max}(\Delta_{\min})+B\sqrt{\Delta_{\min} T/2}\,,\notag\\
 \frac1{16}
 &\le\frac{M_{\max}(\Delta_{\min})}{\sqrt T}+B\sqrt{\Delta_{\min}/2}\,,\notag\\
 \frac1{16}&\le B\sqrt{\Delta_{\min}/2}
 \le\frac{B}{32(1+B)}<\frac1{32}\,.
 \notag
\end{align}

On $\cA$, $\cstar<4$. Because $1/2-\rho>0$, choose a fixed
$\Delta_{\min}\in(0,1/2]$ so small that
$B\Delta_{\min}^{1/2-\rho}/\sqrt2<1/16$.
The asserted upper bound would give
\begin{align}
 \frac{\sqrt T}{16}
 &\le\max_\sigma \Reg[T,\sigma]
 \le4A\log(eT)+M_{\max}(\Delta_{\min})
                            +B\Delta_{\min}^{-\rho}\sqrt{C_{\Delta_{\min},T}}\notag\\
 &\le4A\log(eT)+M_{\max}(\Delta_{\min})
              +(B/\sqrt2)\Delta_{\min}^{1/2-\rho}\sqrt T\,,\notag\\
 \frac1{16}
 &\le\frac{4A\log(eT)+M_{\max}(\Delta_{\min})}{\sqrt T}
                    +\frac B{\sqrt2}\Delta_{\min}^{1/2-\rho}\,,\notag\\
 \frac1{16}&\le\frac B{\sqrt2}\Delta_{\min}^{1/2-\rho}<\frac1{16}\,,
 \notag
\end{align}
where the final line sends $T\to\infty$ at this fixed gap. Thus
neither proposed uniform bound can hold.
\end{proof}

\begin{proof}[Exponential-transient part of \Cref{cor:finite-corruption-consequences}\textup{(iii)}]
Let $a_B=[2^{16}(1+B)^2]^{-1}$ and set
$M_{\max}(\Delta_{\min})=\max_\sigma M(\theta_\sigma)$ and
$T_{\Delta_{\min}}=\lfloor e^{a_B/\Delta_{\min}}\rfloor$.
For all sufficiently small $\Delta_{\min}>0$,
\begin{align}
 T_{\Delta_{\min}}&\ge\tfrac12e^{a_B/\Delta_{\min}}\ge\Delta_{\min}^{-2}\,,\qquad
 \log(eT_{\Delta_{\min}})\le1+a_B/\Delta_{\min}\,.
 \notag
\end{align}
The first assertion follows, for example, from
$(a_B/\Delta_{\min})+2\log\Delta_{\min}\to\infty$.
Substituting $C_{\Delta_{\min},T_{\Delta_{\min}}}\le\Delta_{\min} T_{\Delta_{\min}}/2$ and
$\cstar<4$ into the proposed square-root term gives
\begin{align}
 B\sqrt{C_{\Delta_{\min},T_{\Delta_{\min}}}\{1+\cstar\log(eT_{\Delta_{\min}})\}}
 &\le B\sqrt{\frac{\Delta_{\min} T_{\Delta_{\min}}}2
                   \{1+4(1+a_B/\Delta_{\min})\}}\notag\\
 &=B\sqrt{T_{\Delta_{\min}}}\sqrt{\tfrac52\Delta_{\min}+2a_B}\notag\\
 &\le B\sqrt{T_{\Delta_{\min}}}\sqrt{\tfrac92a_B}
 \le\frac{\sqrt{T_{\Delta_{\min}}}}{64}\,.
 \label{eq:rev-C-exponential-corruption-substitution}
\end{align}
The last two lines use $\Delta_{\min}\le a_B$ and
\begin{align}
 B\sqrt{\tfrac92a_B}
 &=\frac{3B}{256\sqrt2(1+B)}<\frac1{64}\,.
 \notag
\end{align}
For $B=0$ the same bound is immediate.
Now substitute \eqref{eq:rev-C-exponential-corruption-substitution}
into the asserted regret bound, and use the testing lower bound:
\begin{align}
 \frac{\sqrt{T_{\Delta_{\min}}}}{16}
 &\le\max_\sigma \Reg[T_{\Delta_{\min}},\sigma]\notag\\
 &\le4A(1+a_B/\Delta_{\min})+M_{\max}(\Delta_{\min})
                                    +\frac{\sqrt{T_{\Delta_{\min}}}}{64}\,,\notag\\
 M_{\max}(\Delta_{\min})
 &\ge\frac3{64}\sqrt{T_{\Delta_{\min}}}-4A(1+a_B/\Delta_{\min})\,.
 \label{eq:rev-C-exponential-transient-before-limit}
\end{align}
Since
\begin{align}
 \frac{4A(1+a_B/\Delta_{\min})}{\sqrt{T_{\Delta_{\min}}}}
 &\le4\sqrt2A(1+a_B/\Delta_{\min})e^{-a_B/(2\Delta_{\min})}\longrightarrow0\,,
 \notag
\end{align}
the subtracted quantity in
\eqref{eq:rev-C-exponential-transient-before-limit} is at most
$\sqrt{T_{\Delta_{\min}}}/64$ for every sufficiently small $\Delta_{\min}$.
The final substitution is therefore
\begin{align}
 M_{\max}(\Delta_{\min})
 &\ge(3/64-1/64)\sqrt{T_{\Delta_{\min}}}
 =\frac{\sqrt{T_{\Delta_{\min}}}}{32}\notag\\
 &\ge\frac1{32\sqrt2}e^{a_B/(2\Delta_{\min})}
 \ge\frac1{64}e^{a_B/(2\Delta_{\min})}\,.
 \notag
\end{align}
\end{proof}

\begin{proof}[Polynomial transients in \Cref{cor:finite-corruption-consequences}\textup{(iii)}]
Suppose $M(\theta_\sigma)\le K_0\Delta_{\min}^{-p}$ for fixed finite
$K_0,p\ge0$. Choose an integer $m>2p$ with $m\ge2$, and set
$\Delta_{\min,N}=N^{-1}$ and $T_N=N^m$. Then
$T_N\ge\Delta_{\min,N}^{-2}$ and
\begin{align}
 C_{\Delta_{\min,N},T_N}&\le\Delta_{\min,N}T_N/2=N^{m-1}/2\,,\qquad
 \log(eT_N)=1+m\log N\,,\notag\\
 \max_\sigma M(\theta_\sigma)&\le K_0N^p\,.
 \notag
\end{align}
For the geometric square-root coefficient, the assumed upper bound
and the testing lower bound would give
\begin{align}
 \frac{N^{m/2}}{16}
 &\le4A(1+m\log N)+K_0N^p
       +B\sqrt{\frac{N^{m-1}}2\{1+4(1+m\log N)\}}\,,\notag\\
 \frac1{16}
 &\le\frac{4A(1+m\log N)}{N^{m/2}}
       +K_0N^{p-m/2}
       +B\sqrt{\frac{5+4m\log N}{2N}}\longrightarrow0\,.
 \notag
\end{align}
The first and third terms tend to zero because logarithms grow more
slowly than a positive power, and the middle term tends to zero
because $p-m/2<0$.
For the proposed alternative $B\sqrt C[\log(eT)]^q$, direct
substitution instead gives
\begin{align}
 \frac{B\sqrt{C_{\Delta_{\min,N},T_N}}[\log(eT_N)]^q}{N^{m/2}}
 &\le\frac{B\sqrt{N^{m-1}/2}(1+m\log N)^q}{N^{m/2}}
 =\frac{B(1+m\log N)^q}{\sqrt{2N}}\longrightarrow0\,,\notag\\
 \frac1{16}
 &\le\frac{4A(1+m\log N)}{N^{m/2}}
       +K_0N^{p-m/2}
       +\frac{B(1+m\log N)^q}{\sqrt{2N}}\longrightarrow0\,.
 \notag
\end{align}
Each limit contradicts the positive LHS.
Intersecting either proposed bound with an adversarial bound cannot
repair the contradiction, because a minimum is no larger than the
already excluded proposed bound.
\end{proof}

\subsection{A Necessary Transient at Zero Information-Acquisition Complexity}
\begin{proposition}[Nonzero startup cost when $\cstar=0$]\label{prop:scalar-startup}
On $\cA^{\prime}=\{-1,1\}$ with uncorrupted fixed loss vectors
$\theta_\sigma=\sigma\Delta_{\min}/2$, $\sigma=\pm1$, both information-acquisition complexities
are zero. For any policy, $0<\Delta_{\min}\le1/8$, and $T\ge\Delta_{\min}^{-2}$,
the regret is at least $1/(128\Delta_{\min})$ for at least one
of these fixed loss vectors.
\end{proposition}
\begin{proof}
Let $m=\lfloor1/(16\Delta_{\min}^2)\rfloor$. Since $\Delta_{\min}\le1/8$,
\begin{align}
 \frac1{16\Delta_{\min}^2}&\ge4\,,\qquad
 \frac1{32\Delta_{\min}^2}\le m\le\frac1{16\Delta_{\min}^2}\le T\,.
 \notag
\end{align}
After multiplying feedback by the chosen sign, the latent observations
are independent signs of means $\pm\Delta_{\min}/2$. The single-observation
divergence formula \eqref{eq:rev-C-single-sign-divergence}, now with
$\epsilon=\Delta_{\min}/2$, and the product calculation give
\begin{align}
 \chi^2(Q_{-\Delta_{\min}/2}^{\otimes m}\Vert Q_{\Delta_{\min}/2}^{\otimes m})
 &=\left(1+\frac{\Delta_{\min}^2}{1-\Delta_{\min}^2/4}\right)^m-1\notag\\
 &\le\exp\!\left(\frac{m\Delta_{\min}^2}{1-\Delta_{\min}^2/4}\right)-1
 \le\exp\!\left(\frac{1/16}{1-1/256}\right)-1\notag\\
 &=e^{16/255}-1<1\,,\notag\\
 \TV(P_+^m,P_-^m)
 &\le\tfrac12\sqrt{e^{16/255}-1}<\tfrac12\,.
 \notag
\end{align}
The last line uses the same common-kernel contraction as
\eqref{eq:rev-C-adaptive-kernel-contraction}.
For $E_t=\{A_t=+1\}$, the wrong-action gap is now $\Delta_{\min}$;
therefore
\begin{align}
 \Reg[m,+]+\Reg[m,-]
 &=\Delta_{\min}\sum_{t=1}^m\{P_+^m(E_t)+P_-^m(E_t^c)\}\notag\\
 &\ge\Delta_{\min}\sum_{t=1}^m\{1-\TV(P_+^m,P_-^m)\}
 \ge m\Delta_{\min}/2\,,\notag\\
 \max_\sigma \Reg[T,\sigma]
 &\ge\max_\sigma \Reg[m,\sigma]
 \ge m\Delta_{\min}/4
 \ge\frac{\Delta_{\min}}4\frac1{32\Delta_{\min}^2}
 =\frac1{128\Delta_{\min}}\,.
 \notag
\end{align}
The first inequality in the last line uses nonnegative uncorrupted
suffix gaps. Both values of $\cstar$ are zero by
\eqref{eq:rev-C-zero-information-design}.
\end{proof}

\clearpage
\section{Proof of Theorem~\ref*{thm:main-nonseparated-upper}}
\label{app:sharp-results}
This appendix deduces \Cref{thm:main-nonseparated-upper}
from \Cref{prop:SH-compatibility} and a finite centered
sampling allocation. The proposition for a fixed allocation and its supporting
estimates are proved in
\Cref{app:sharp-compatibility-proof}.
All sampling allocations used in the analysis are fixed independently of the horizon.
For the proof notation only, let $\cA$ denote the physical input generator
$\mathcal B$. Thus $\mathcal K=\Conv(\cA)$,
$\mathcal S=\Conv(\cA\cup(-\cA))$,
$\mathsf D=\operatorname{span}(\mathcal K-\mathcal K)$, and
$k=\dim\operatorname{aff}(\mathcal K)\ge1$.
All algorithmic quantities retain their definitions from
\Cref{sec:SH-algorithm} under this notation.
We abbreviate $c_{\mathrm{ctr}}^{n}(\cA,\theta)$ as
$c_{\mathrm{ctr}}^{n}$ and $\cctr(\cA,\theta)$ as $\cctr$;
here $\cA$ denotes the physical generator just specified.

\subsection{Completion of the Proof}\label{app:sharp-upper-proof}\label{app:sharp-design-bounds}
We apply \Cref{prop:SH-compatibility} to
\GIAFTRL{} and choose the comparison sampling allocation.
Its proof is given in \Cref{app:sharp-compatibility-proof}.
\begin{proof}[Adversarial part of \Cref{thm:main-nonseparated-upper}]
\Cref{prop:SH-compatibility} gives
$\Reg\le U_{\rm SH}(T)$ for the same measurable learner, for
all predictable bounded-mean environments. Its proof uses only
\Cref{lem:SH-feasible,lem:SH-prior-comparator},
so sharpness and the corruption budget are not assumptions of this part.
\end{proof}

\subsubsection{A Fixed Finite Comparison Sampling Allocation}
Let $\mathfrak N_\theta$ be the collection of finite feasible
measures in \eqref{eq:SH-centered-program} in
\Cref{def:D-physical-design}. By definition their infimum is $\cctr$;
sharpness makes it finite by \Cref{lem:D-centered-value}. Thus
\begin{align}
 \inf_{n\in\mathfrak N_\theta}c_{\mathrm{ctr}}^{n}(\cA,\theta)&=\cctr(\cA,\theta)\,,
       \label{eq:D-fixed-design-infimum}\\
 \cctr>0&\ \Longrightarrow\
       \exists n^\circ\in\mathfrak N_\theta:\ c_{\mathrm{ctr}}^{n^\circ}\le2\cctr\,.
       \label{eq:D-step-0001}
\end{align}
The second line uses the defining approximation property of the positive
infimum at tolerance $\cctr$. The sampling allocation $n^\circ$ is fixed once for the
instance, independently of $T,C$.

We verify the zero-cost case directly for centered sampling allocations.
If $a^*\ne0$, let $U^*$ have orthonormal columns spanning $(a^*)^{\perp}$;
if $a^*=0$, set $U^*=I_d$. Define $x(a)=(U^*)^{\top} a$, so $x(a^*)=0$.
When this subspace has positive dimension, every centered feasible sampling allocation
has $M(n)\coloneq (U^*)^{\top} H_nU^*\succ0$. Since $\mathcal S\subseteq\BB$,
sharpness and the Schur-complement inequality give
\begin{align}
\|x(a)\|&=\|(U^*)^{\top}(a-a^*)\|
 \le\|a-a^*\|\le\|a-a^*\|_{\mathcal S}
 \le\Delta(a)/\varsigma\,,
       \label{eq:D-step-0003}\\
 L_x\coloneq \sup_{a\ne a^*}\frac{\|x(a)\|^2}{\Delta(a)}
 &\le\sup_{a\ne a^*}\frac{\Delta(a)}{\varsigma^2}
 \le2/\varsigma^2<\infty\,,\label{eq:SHr-zero-feature-bound}\\
 \lambda_{\max}(M(n))
 &\le\int\|x(a)\|^2n(da)\le L_xc_{\mathrm{ctr}}^{n}\,,
       \label{eq:D-step-0004}\\
 \frac{\|x(a)\|^2}{\lambda_{\max}(M(n))}
 &\le x(a)^\top M(n)^{-1}x(a)
 \le(a-a^*)^\top H_n^{-1}(a-a^*)\le\frac{\Delta(a)^2}{2}\,.
 \label{eq:SHr-zero-design-bound}
\end{align}
Equation \eqref{eq:D-step-0003} uses that $(U^*)^{\top}$ is an orthogonal projection, $\mathcal S\subseteq\mathbb B_2$, and sharpness.
Equation \eqref{eq:SHr-zero-feature-bound} squares the preceding feature bound, divides by $\Delta(a)>0$, and uses $\Delta(a)\le2$.
Equation \eqref{eq:D-step-0004} bounds the largest eigenvalue by the trace and substitutes $c_{\mathrm{ctr}}^{n}=\int\Delta\,dn$.

Equation \eqref{eq:SHr-zero-design-bound} uses $M(n)^{-1}\succeq\lambda_{\max}(M(n))^{-1}I$, the
Schur-complement comparison, and the centered feasibility constraint.
Thus this argument applies to every sampling allocation in $\mathfrak N_\theta$.
If the projected dimension is positive, spanning supplies an $a$ with
$x(a)\ne0$; then $L_x>0$ and \eqref{eq:SHr-zero-design-bound} gives
$c_{\mathrm{ctr}}^{n}\ge2\|x(a)\|^2/[L_x\Delta(a)^2]>0$ for every feasible sampling allocation.
Hence $\cctr=0$ forces this orthogonal subspace to have dimension zero. In the stated spanning
coordinates this means $d=1$ and $a^*\ne0$. The finite measure
$n^0=m_{\rm free}\delta_{a^*}$, with $m_{\rm free}=2/(\varsigma^2(a^*)^{2})$, satisfies
\begin{align}
H_{n^0}&=m_{\rm free}(a^*)^{2}>0\,,
 &c_{\mathrm{ctr}}^{n^0}&=m_{\rm free}\Delta(a^*)=0\,,
       \label{eq:D-step-0005}\\
 \frac{(a-a^*)^2}{H_{n^0}}
 &=\frac{\varsigma^2(a-a^*)^2}{2}
 \le\frac{\varsigma^2\|a-a^*\|_{\mathcal S}^2}{2}
 \le\frac{\Delta(a)^2}{2}\,.
       \label{eq:D-step-0006}
\end{align}
Equation \eqref{eq:D-step-0005} substitutes the single atom
$n^0=m_{\rm free}\delta_{a^*}$ and $\Delta(a^*)=0$.
Equation \eqref{eq:D-step-0006} substitutes
$m_{\rm free}(a^*)^{2}=2/\varsigma^2$ and uses sharpness.
Thus $n^0$ is a finite feasible centered sampling allocation of zero cost.
Use $n^\circ=n^0$ when $\cctr=0$. In both cases,
$c_{\mathrm{ctr}}^{n^\circ}\le2\cctr$ and the constant $M_{n^\circ}$ in
\eqref{eq:SC-constants} is finite and independent of $T,C$.

For the generator comparison, return to the main-statement notation:
$\mathcal B$ is the physical generator and $\cA$ is the original action
set. Applying \Cref{lem:C-hull} gives
\begin{align}
\Conv(\mathcal B)&=\Conv(\cA)\,,
 &\sup_{a\in\cA}\Reg(a)&=\sup_{b\in\mathcal B}\Reg(b)\,,
       \label{eq:D-step-0008}\\
 \cctr(\cA,\theta)&=\cctr(\mathcal B,\theta)\,.
 \label{eq:SHr-generator-transfer}
\end{align}
Equation \eqref{eq:D-step-0008} uses $\mathcal B\subseteq\cA\subseteq\operatorname{conv}(\mathcal B)$ and linearity of a fixed-comparator regret in its comparator.
Equation \eqref{eq:SHr-generator-transfer} uses the preservation of regret cost in \Cref{lem:C-hull}.
In this main-statement notation, the fixed allocation selected above satisfies
\[
 c_{\mathrm{ctr}}^{n^\circ}(\gB,\theta)
 \le2c_{\mathrm{ctr}}^*(\gB,\theta)
 =2c_{\mathrm{ctr}}^*(\gA,\theta)\,.
\]
Thus the regret and information-acquisition complexity used above are exactly
those in the statement, not relaxations associated with a larger action set.

\begin{proof}[Corruption part of \Cref{thm:main-nonseparated-upper}]
Fix $n\in\mathfrak N_\theta$, with $\mathfrak N_\theta$ as in
\eqref{eq:D-fixed-design-infimum}. \Cref{prop:SH-compatibility}
gives the two distinct comparator bounds
\begin{align}
\Reg(a^*)&\le16BC_{\mathrm{rat}}^4c_{\mathrm{ctr}}^{n}\log(eT)+M_n
                      +J_\theta^{\rm anc}\sqrt C\log(e+C)\,,
       \label{eq:D-step-0009}\\
 \Reg&\le16BC_{\mathrm{rat}}^4c_{\mathrm{ctr}}^{n}\log(eT)+M_n
                      +J_\theta^{\rm cor}\sqrt C\log(e+C)\,,
       \label{eq:D-step-0010}\\
 \Reg&\le \Reg(a^*)+T\Omega_\theta(C/T)\,,
 &T\varsigma\ge C&\ \Longrightarrow\ \Omega_\theta(C/T)=0\,.
 \label{eq:SHr-cor-designs}
\end{align}
Equation \eqref{eq:D-step-0009} is the bound for $\Reg(a^*)$ \eqref{eq:SC-anchor} underlying \Cref{prop:SH-compatibility} for the fixed sampling allocation $n$.
Equation \eqref{eq:D-step-0010} is the all-comparator bound \eqref{eq:SC-per-design} underlying \Cref{prop:SH-compatibility} for that same $n$.
Equation \eqref{eq:SHr-cor-designs} applies the comparator identity of \Cref{lem:comparator} and the sharpness margin.
Choose the fixed $n^\circ$ constructed above, and set
$M_\theta=M_{n^\circ}$. Substitution gives
\begin{align}
\Reg(a^*)&\le16BC_{\mathrm{rat}}^4(2\cctr)\log(eT)+M_\theta
                       +J_\theta^{\rm anc}\sqrt C\log(e+C)\,,
       \label{eq:D-step-0011}\\
 \Reg&\le\min\{U_{\rm SH}(T),\,
       32BC_{\mathrm{rat}}^4\cctr\log(eT)+M_\theta
                       +J_\theta^{\rm cor}\sqrt C\log(e+C)\}\,.
 \label{eq:SHr-cor-main-final}
\end{align}
Equation \eqref{eq:D-step-0011} substitutes $c_{\mathrm{ctr}}^{n^\circ}\le2\cctr$ and $M_\theta=M_{n^\circ}$ in the bound for $\Reg(a^*)$.
Equation \eqref{eq:SHr-cor-main-final} makes the same substitution in the all-comparator bound and intersects it with the adversarial guarantee.
The substitution remains valid at $\cctr=0$ because $c_{\mathrm{ctr}}^{n^\circ}=0$.
For any budget sequence satisfying $\sqrt{C_T}\log(e+C_T)=o(\log T)$
and any fixed $n$, divide the
all-comparator bound \eqref{eq:D-step-0010} by $\log T$:
\begin{align}
\limsup_{T\to\infty}\frac{\Reg}{\log T}
 &\le16BC_{\mathrm{rat}}^4c_{\mathrm{ctr}}^{n}\lim_{T\to\infty}\frac{\log(eT)}{\log T}
    +\lim_{T\to\infty}\frac{M_n}{\log T}
    +J_\theta^{\rm cor}\lim_{T\to\infty}
             \frac{\sqrt{C_T}\log(e+C_T)}{\log T}
       \label{eq:D-step-0012}\\
 &=16BC_{\mathrm{rat}}^4c_{\mathrm{ctr}}^{n}\,,
       \label{eq:D-step-0013}\\
 \limsup_{T\to\infty}\frac{\Reg}{\log T}
 &\le16BC_{\mathrm{rat}}^4\inf_{n\in\mathfrak N_\theta}c_{\mathrm{ctr}}^{n}=16BC_{\mathrm{rat}}^4\cctr\,.
 \label{eq:SHr-cor-limsup}
\end{align}
Equation \eqref{eq:D-step-0012} divides every term of the fixed-allocation all-comparator bound by $\log T$ before taking the limsup.
Equation \eqref{eq:D-step-0013} uses the assumed vanishing of $\sqrt{C_T}\log(e+C_T)/\log T$, the fixed transient, and $\log(eT)/\log T\to1$.
Equation \eqref{eq:SHr-cor-limsup} takes the infimum over fixed sampling allocations and uses the definition of $\cctr$ in \eqref{eq:D-fixed-design-infimum}.
For the generator transfer, \eqref{eq:SHr-generator-transfer} preserves
regret and the information-acquisition complexity, while the action-seminorm identity in
\Cref{lem:C-hull} gives, on every realized path,
\begin{align}
\|c_t^{\rm vec}\|_{\cA}
 &=\sup_{a\in\cA}|a^\top c_t^{\rm vec}|
 =\sup_{b\in\mathcal B}|b^\top c_t^{\rm vec}|
 =\|c_t^{\rm vec}\|_{\mathcal B}\,,
       \label{eq:D-step-0014}\\
 \sum_{t=1}^T\|c_t^{\rm vec}\|_{\cA}
 &=\sum_{t=1}^T\|c_t^{\rm vec}\|_{\mathcal B}\le C\,.
 \label{eq:SHr-generator-corruption-transfer}
\end{align}
Equation \eqref{eq:D-step-0014} expands both seminorms and uses that a linear functional and its negative attain the same extrema on a set and its convex hull.
Equation \eqref{eq:SHr-generator-corruption-transfer} sums the per-round equalities and applies the original pathwise budget.
Equations \eqref{eq:SHr-cor-main-final} and
\eqref{eq:SHr-cor-limsup} prove the stated corrupted stochastic
claims. Setting $C=0$ gives the stochastic specialization with the same
constant $M_\theta$.
\end{proof}

\clearpage
\section{Proof of Proposition~\ref*{prop:SH-compatibility}}
\label{app:sharp-compatibility-proof}
\label{ch:sharp-ftrl}
The first subsection gives the complete proof for
\GIAFTRL{}. The next four subsections prove its penalty, stability,
exploration, and direct corruption bounds. The auxiliary lemmas establish
feasibility, comparator coverage, and the probabilistic estimates.
\GIAFTRL{}, its fixed reference, and its full
reconstruction program are specified in
\Cref{sec:SH-algorithm}.
For the proof notation only, let $\cA$ denote the physical input generator
$\mathcal B$, with $\mathcal K=\Conv(\cA)$,
$\mathcal S=\Conv(\cA\cup(-\cA))$,
$\mathsf D=\operatorname{span}(\mathcal K-\mathcal K)$, and
$k=\dim\operatorname{aff}(\mathcal K)\ge1$.
All algorithmic quantities retain their definitions from
\Cref{sec:SH-algorithm} under this notation.
We abbreviate $c_{\mathrm{ctr}}^{n}(\cA,\theta)$ as
$c_{\mathrm{ctr}}^{n}$ and $\cctr(\cA,\theta)$ as $\cctr$;
here $\cA$ denotes the physical generator just specified.
Throughout the term and moment proofs, $q_t,r_t,\widehat L_{t,\cdot},p_t$ are the
quantities generated by \GIAFTRL{}, with parameter estimate
$\widehat\theta_t=v_{X_t}(p_t)Y_t$. Existence and measurability of its minimizing
sampling law are established in \Cref{lem:SH-feasible}.

\subsection{Complete Formal Proof of Proposition~\ref*{prop:SH-compatibility}}
\label{app:sharp-compatibility-complete}
\begin{proof}[Proof of \Cref{prop:SH-compatibility}]
Consider \GIAFTRL{}.
\Cref{lem:SH-feasible} gives a Borel measurable feasible
minimizer $p_t$ at each history and deterministic finite-time bounds
on its estimates. The covariance definition $M(p)=\int_{\mathcal K}xx^\top p(dx)$ and $v_x(p)=M(p)^{-1}x$
implies
\[
 \int v_x(p_t)x^\top p_t(dx)=M(p_t)^{-1}M(p_t)=I_d\,.
\]
Together with the feedback model and physical rounding, this gives
conditional unbiasedness. The program supplies the magnitude and
second-moment bounds \eqref{eq:SH-global-amp}--\eqref{eq:SH-global-var}.
The range recurrence \eqref{eq:SH-rule-range} also bounds the difference
fields and their stopped versions at every finite time.
\Cref{prop:D-adversarial} therefore gives $\Reg\le U_{\rm SH}(T)$
in the adversarial regime, without assuming sharpness.

For the corrupted stochastic guarantee, assume that the uncorrupted
instance satisfies the sharpness condition in \eqref{eq:SH-overview-sharp}.
Fix a feasible centered sampling allocation $n$ on $\cA$,
and let $N_n=n(\cA)$.

The fixed prior and reference give \Cref{lem:SH-atom,lem:SH-caps,lem:SH-prior-comparator}. These estimator
bounds and the prescribed update yield the moment estimates in
\Cref{lem:SH-ratios,lem:SC-good}, which are the
inputs to the penalty, stability, and direct corruption proofs.

For exploration, \Cref{lem:SH-witness} makes $p_t^c$
feasible on $\mathcal W_t$, and \Cref{lem:SH-feasible}
provides the feasible G-optimal fallback on
$\mathcal G_t\setminus\mathcal W_t$. Both use the prescribed
covariance coefficients. Since $p_t$ minimizes \eqref{eq:SH-program},
comparison with these two sampling laws gives
\eqref{eq:D-step-0032} and \eqref{eq:D-step-0034}.

Combining these bounds, \Cref{prop:SC-per-design},
proved next by assembling the four term lemmas for this same algorithm, gives
\eqref{eq:SC-anchor}--\eqref{eq:SC-per-design}, the comparator
correction, and the constants \eqref{eq:SC-constants}.
The constants depend only on the fixed geometry, prior, tuning, fixed loss vector,
and (for the transients) sampling allocation. The corruption coefficients do not use
$n$ or $N_n$.
Finally, \Cref{lem:C-hull} transfers comparator suprema and
corruption norms from the physical generator to the original action set.
\end{proof}

\subsubsection{Assembly of the Corrupted Regret Bound for a Fixed Allocation}
\begin{proposition}[Corrupted sharp-instance bound for each sampling allocation]\label{prop:SC-per-design}
\GIAFTRL{}, with a fixed finite
feasible centered sampling allocation $n^{\rm des}$, satisfies for every $T\ge1$
and every allowed budget $C\ge0$
\begin{align}
 \Reg(a^*)&\le16BC_{\mathrm{rat}}^4c_{\mathrm{ctr}}^{n^{\rm des}}\log(eT)
  +M_{n^{\rm des}}
  +J^{\rm anc}_\theta\sqrt C\log(e+C)\,,\label{eq:SC-anchor}\\
 \Reg&\le16BC_{\mathrm{rat}}^4c_{\mathrm{ctr}}^{n^{\rm des}}\log(eT)
  +M_{n^{\rm des}}
  +J^{\rm cor}_\theta\sqrt C\log(e+C)\,.\label{eq:SC-per-design}
\end{align}
Each bound may also be capped by the adversarial envelope.
For the actual comparator, one may instead add $T\Omega_\theta(C/T)$
to \eqref{eq:SC-anchor}. This correction vanishes when $T\varsigma\ge C$.
\end{proposition}
The proof assembles the four term lemmas below and the weighted
failure estimate, and then treats the actual comparator explicitly.

\begin{proof}[Proof of \Cref{prop:SC-per-design}]
Fix $T,C$ and the measure $n^{\rm des}$. Throughout this proof,
$\mathcal G_t=\mathcal G_t^{\rm cor}$, and
$\mathcal E_T=\sum_{t=1}^Tt^{-1/2}\sP(\mathcal G_t^c)$.
The weighted sum $\mathcal E_T$ is exactly the failure-probability
cost at the $t^{-1/2}$ scale of the global entropy, the proxy function, and
exploration bounds. The law $q_t$, the marginal $p_t$, and $\mathcal G_t$ are all
$\cF_{t-1}$-measurable. Conditional unbiasedness, followed by the
localized entropy identity \eqref{eq:relative-local} in \Cref{lem:entropy-FTRL}, gives
\begin{align}
\Reg(a^*)
 &=\E[\sum_{t=1}^T\langle q_t-\delta_{a^*},\widehat\ell_{t,\cdot}\rangle]
   +\E[\sum_{t=1}^T\langle p_t-q_t,\Delta\rangle]
   +\E[\sum_{t=1}^T\langle p_t-q_t,x^\top c_t^{\rm vec}\rangle]
       \label{eq:D-step-0030}\\
 &\le\beta_1h^*+
 \sum_{t=2}^T(\beta_t-\beta_{t-1})\E[h^*-\KL(q_t,\mu)]
       +\sum_{t=1}^T\E[\gC_t(p_t)]
       \label{eq:D-step-0031}\\*
 &\quad+\sum_{t=1}^T\E[\langle p_t-q_t,\Delta\rangle]
       +\sum_{t=1}^T\E[\langle p_t-q_t,x^\top c_t^{\rm vec}\rangle]\,.
       \label{eq:SC-master}
\end{align}
Equation \eqref{eq:D-step-0030} inserts $q_t$ between $p_t$ and $\delta_{a^*}$, uses unbiased loss differences for the $q_t$ term, and splits $\theta_t=\theta+c_t^{\rm vec}$.
Equation \eqref{eq:D-step-0031} applies the localized relative-entropy identity and the conditional exponential proxy function to the $q_t$ regret.
Equation \eqref{eq:SC-master} retains both exploration integrals, one under $\theta$ and one under $c_t^{\rm vec}$, in the same regret inequality.
For the clean exploration integral in \eqref{eq:D-step-0030},
\begin{align}
 \langle p_t-q_t,x^\top\theta\rangle
 &=\int[\Delta(x)+(a^*)^{\top}\theta]\,(p_t-q_t)(dx)
       \label{eq:D-cor-clean-shift-substitution}\\
 &=\langle p_t-q_t,\Delta\rangle
       +((a^*)^{\top}\theta)[p_t(\mathcal K)-q_t(\mathcal K)]
 =\langle p_t-q_t,\Delta\rangle\,.
       \label{eq:D-cor-clean-shift-cancel}
\end{align}
Equation \eqref{eq:D-cor-clean-shift-substitution} substitutes the
definition of the gap under $\theta$; \eqref{eq:D-cor-clean-shift-cancel} uses
$p_t(\mathcal K)=q_t(\mathcal K)=1$. We bound every sum in
\eqref{eq:SC-master} separately and then substitute all bounds back.

\Cref{lem:D-penalty-term,lem:D-stability-term,lem:D-exploration-term,lem:D-corruption-term} prove
respectively the four sums in \eqref{eq:SC-master}. We now evaluate the
remaining failure cost and substitute their terminal bounds.

For $n=\lceil C\rceil+2$ and $L_n=\log(en)$, the definition of $N$
implies
\begin{align}
\sqrt N&\le\sqrt{K_0nL_n^2+1}\le2\sqrt{K_0}\sqrt nL_n\,,
       \label{eq:D-step-0038}\\
 0\le C\le1&:\quad \sqrt nL_n\le\sqrt3\log(3e)<4\,,
       \label{eq:D-step-0039}\\
 C\ge1&:\quad n\le4C\,,\quad L_n\le4\log(e+C)\,,\quad
               \sqrt nL_n\le8\sqrt C\log(e+C)\,,
       \label{eq:D-step-0040}\\
 \sqrt N&\le16\sqrt{K_0}[1+\sqrt C\log(e+C)]\,.
 \label{eq:SC-clock-cost}
\end{align}
Equation \eqref{eq:D-step-0038} uses $N=\lceil K_0nL_n^2\rceil\le K_0nL_n^2+1$ and $K_0nL_n^2\ge1$.
Equation \eqref{eq:D-step-0039} uses $n=\lceil C\rceil+2\le3$ and $L_n=\log(en)\le\log(3e)$.
Equation \eqref{eq:D-step-0040} uses $n\le C+3\le4C$ and $\log(4eC)\le4\log(e+C)$.

Equation \eqref{eq:SC-clock-cost} substitutes the two budget cases into the bound on $\sqrt N$ and uses a single larger coefficient.
Consequently \eqref{eq:SC-good-tail} in \Cref{lem:SC-good} yields
\begin{align}
(h^*+10d)\mathcal E_T
 &\le(h^*+10d)W_\theta\sqrt N
       \label{eq:D-step-0041}\\
 &\le16\sqrt{K_0}(h^*+10d)W_\theta
                   [1+\sqrt C\log(e+C)]
 =E_\theta[1+\sqrt C\log(e+C)]\,.
 \label{eq:SCr-bad-history-charge}
\end{align}
Equation \eqref{eq:D-step-0041} substitutes the complete weighted failure bound \eqref{eq:SC-good-tail} in \Cref{lem:SC-good}.
Equation \eqref{eq:SCr-bad-history-charge} substitutes \eqref{eq:SC-clock-cost} and then the definition of $E_\theta$ in \eqref{eq:SC-final-threshold} in \Cref{def:D-cor-dictionary}.
Substitute \eqref{eq:SCr-entropy-sum}, \eqref{eq:SCr-stability-sum},
\eqref{eq:SCr-exploration-sum}, \eqref{eq:SCr-direct-corruption}, and
\eqref{eq:SCr-bad-history-charge} into \eqref{eq:SC-master}:
\begin{align}
\Reg(a^*)
 &\le\beta_1h^*+(4A_L+h^*\mathcal E_T)
                +(2e^2C_{\beta,\mathrm{sh}}A_L+2d\mathcal E_T)\label{eq:D-cor-master-direct-substitution}\\*
 &\quad+\left[16BC_{\mathrm{rat}}^4c_{\mathrm{ctr}}^{n^{\rm des}}\log(eT)+4A_LC_{\mathrm{rat}}^2
    +8d\left(4C_{\mathrm{rat}}^4a_{n^{\rm des}}+\frac83C_{\mathrm{rat}}^3\sqrt{a_\rho}\right)
    +8d\mathcal E_T\right]\notag\\*
 &\quad+16\sqrt2d\sqrt C
       \label{eq:D-step-0042}\\
 &=16BC_{\mathrm{rat}}^4c_{\mathrm{ctr}}^{n^{\rm des}}\log(eT)+\beta_1h^*+4A_L+2e^2C_{\beta,\mathrm{sh}}A_L+4A_LC_{\mathrm{rat}}^2
       \label{eq:D-cor-master-collect}\\*
 &\quad+8d\left(4C_{\mathrm{rat}}^4a_{n^{\rm des}}+\frac83C_{\mathrm{rat}}^3\sqrt{a_\rho}\right)
        +(h^*+10d)\mathcal E_T+16\sqrt2d\sqrt C
       \label{eq:D-step-0043}\\
 &\le16BC_{\mathrm{rat}}^4c_{\mathrm{ctr}}^{n^{\rm des}}\log(eT)+\beta_1h^*+4A_L(1+C_{\mathrm{rat}}^2)+2e^2C_{\beta,\mathrm{sh}}A_L
       \label{eq:D-cor-master-clock-substitute}\\*
 &\quad+8d\left(4C_{\mathrm{rat}}^4a_{n^{\rm des}}+\frac83C_{\mathrm{rat}}^3\sqrt{a_\rho}\right)
       +E_\theta[1+\sqrt C\log(e+C)]+16\sqrt2d\sqrt C
       \label{eq:D-step-0044}\\
 &=16BC_{\mathrm{rat}}^4c_{\mathrm{ctr}}^{n^{\rm des}}\log(eT)+M_{n^{\rm des}}
       +E_\theta\sqrt C\log(e+C)+16\sqrt2d\sqrt C
       \label{eq:D-cor-master-transient-substitute}\\
 &\le16BC_{\mathrm{rat}}^4c_{\mathrm{ctr}}^{n^{\rm des}}\log(eT)+M_{n^{\rm des}}
       +(E_\theta+16\sqrt2d)\sqrt C\log(e+C)\label{eq:D-cor-master-log-factor}\\
 &=16BC_{\mathrm{rat}}^4c_{\mathrm{ctr}}^{n^{\rm des}}\log(eT)+M_{n^{\rm des}}
       +J_\theta^{\rm anc}\sqrt C\log(e+C)\,.
       \label{eq:SCr-anchor-final}
\end{align}
Equation \eqref{eq:D-cor-master-direct-substitution} substitutes the entropy and proxy function sums literally into the master.
Equation \eqref{eq:D-step-0042} inserts the entire bound on exploration under $\theta$ and the direct corruption cost without collecting any terms.
Equation \eqref{eq:D-cor-master-collect} collects the harmonic allocation cost, initial entropy, and good-history constants.

Equation \eqref{eq:D-step-0043} collects the costs of failures of the comparison distribution's feasibility thresholds and $h^*+2d+8d=h^*+10d$ times the weighted bad-history sum.
Equation \eqref{eq:D-cor-master-clock-substitute} combines $4A_L+4A_LC_{\mathrm{rat}}^2$ and leaves the remaining terms present.
Equation \eqref{eq:D-step-0044} substitutes the complete bad-history bound \eqref{eq:SCr-bad-history-charge}.

Equation \eqref{eq:D-cor-master-transient-substitute} substitutes the definition of $M_{n^{\rm des}}$.
Equation \eqref{eq:D-cor-master-log-factor} uses $\log(e+C)\ge1$ on the direct corruption term.
Equation \eqref{eq:SCr-anchor-final} substitutes $J_\theta^{\rm anc}=E_\theta+16\sqrt2d$.
This proves \eqref{eq:SC-anchor}.

\paragraph{The Actual Fixed Comparator}
For every $a\in\mathcal K$ and $\|z\|_\cA\le C/T$, duality and
sharpness give
\begin{align}
\bar c_T&=T^{-1}\sum_{t=1}^T\E[c_t^{\rm vec}]\,,\qquad
 \|\bar c_T\|_\cA
 \le T^{-1}\sum_{t=1}^T\E[\|c_t^{\rm vec}\|_\cA]\le C/T\,,
       \label{eq:D-cor-average-budget}\\
 \Reg-\Reg(a^*)
 &=T\sup_{a\in\mathcal K}
          \{-\Delta(a)-(a-a^*)^\top\bar c_T\}
 \le T\Omega_\theta(C/T)\,,\label{eq:D-cor-comparator-modulus}\\
 -\Delta(a)-(a-a^*)^\top z
 &\le-\Delta(a)+\|a-a^*\|_{\mathcal S}\|z\|_\cA\label{eq:D-cor-duality}\\
 &\le-\varsigma\|a-a^*\|_{\mathcal S}
                 +(C/T)\|a-a^*\|_{\mathcal S}\label{eq:D-cor-sharp-substitution}\\
 &=-(\varsigma-C/T)\|a-a^*\|_{\mathcal S}
       \qquad(\|z\|_\cA\le C/T)\,,\label{eq:D-cor-sharp-factor}\\
 T\varsigma\ge C&\quad\Longrightarrow\quad
 0\le\Omega_\theta(C/T)\le0\,,\qquad
 0\le \Reg-\Reg(a^*)\le0\,.
       \label{eq:SCr-comparator-late}
\end{align}
Equation \eqref{eq:D-cor-average-budget} takes expectations of the
pathwise budget and uses convexity of the norm.
Equation \eqref{eq:D-cor-comparator-modulus} is the fixed-comparator
identity in \Cref{lem:comparator}, so the supremum is
outside expectation.  Equation \eqref{eq:D-cor-duality} uses the
polar norm; \eqref{eq:D-cor-sharp-substitution} inserts sharpness and
the budget separately.  Equation \eqref{eq:D-cor-sharp-factor}
factors the common gauge.  In \eqref{eq:SCr-comparator-late},
$a=a^*$ gives both lower bounds by zero and the preceding inequality
gives the upper bound.  Thus $\Reg=\Reg(a^*)$ in this branch.
If $T\varsigma<C$, then $C>0$ and
\begin{align}
\sqrt T&\le\sqrt{C/\varsigma}\,,
       \label{eq:D-step-0045}\\
 \log(eT)&\le1+(\log C)_++(\log\varsigma^{-1})_+
       \label{eq:D-step-0046}\\
 &\le2\log(e+C)+\log(1+\varsigma^{-1})\log(e+C)\,,
       \label{eq:D-step-0047}\\
 \Reg&\le U_{\rm SH}(T)
 \le\frac{320(d+1)}{\sqrt\varsigma}
       [2+\log(1+\varsigma^{-1})]\sqrt C\log(e+C)\,.
 \label{eq:SCr-comparator-early}
\end{align}
Equation \eqref{eq:D-step-0045} uses the early-branch condition $T\varsigma<C$.
Equation \eqref{eq:D-step-0046} takes logarithms of $T<C/\varsigma$ and replaces each potentially negative logarithm by its positive part.
Equation \eqref{eq:D-step-0047} uses $1\le\log(e+C)$, $(\log C)_+\le\log(e+C)$, and $(\log\varsigma^{-1})_+\le\log(1+\varsigma^{-1})$.

Equation \eqref{eq:SCr-comparator-early} substitutes both scalar bounds into \eqref{eq:SHr-adv-simple} in \Cref{prop:D-adversarial}.
Combining the two branches \eqref{eq:SCr-comparator-late} and
\eqref{eq:SCr-comparator-early}, and adding only nonnegative allowances,
\begin{align}
\Reg
 &\le16BC_{\mathrm{rat}}^4c_{\mathrm{ctr}}^{n^{\rm des}}\log(eT)+M_{n^{\rm des}}
       \label{eq:D-step-0048}\\*
 &\quad+\left(J_\theta^{\rm anc}
       +\frac{320(d+1)}{\sqrt\varsigma}[2+\log(1+\varsigma^{-1})]\right)
                 \sqrt C\log(e+C)
       \label{eq:D-step-0049}\\
 &=16BC_{\mathrm{rat}}^4c_{\mathrm{ctr}}^{n^{\rm des}}\log(eT)+M_{n^{\rm des}}
                +J_\theta^{\rm cor}\sqrt C\log(e+C)\,.
 \label{eq:SCr-corrupted-final}
\end{align}
Equation \eqref{eq:D-step-0048} keeps the fixed-allocation harmonic term and transient in both the late and early branches.
Equation \eqref{eq:D-step-0049} adds the corruption coefficient in the bound for $\Reg(a^*)$ and the explicit early-branch coefficient from \eqref{eq:SCr-comparator-early}.
Equation \eqref{eq:SCr-corrupted-final} substitutes the definition of $J_\theta^{\rm cor}$ in \eqref{eq:SC-constants} in \Cref{def:D-cor-dictionary}.
Independently, \Cref{lem:comparator} gives the exact alternative
$\Reg\le \Reg(a^*)+T\Omega_\theta(C/T)$, and
\eqref{eq:SHr-adv-final} in \Cref{prop:D-adversarial} caps both fixed-comparator bounds.
\end{proof}

\subsection{Proof of Penalty Term}\label{app:D-penalty}
\begin{lemma}[Relative-entropy penalty contribution]\label{lem:D-penalty-term}
Use $\mathcal G_t=\mathcal G_t^{\rm cor}$ from
\Cref{lem:SC-good} and
$\mathcal E_T=\sum_{t\le T}t^{-1/2}\sP(\mathcal G_t^c)$.
With $A_L$ explicitly defined in \eqref{eq:SH-laplace-constant},
\[
 \mathsf I_1\coloneq \beta_1h^*+\sum_{t=2}^T(\beta_t-\beta_{t-1})
             \E[h^*-\KL(q_t,\mu)]
 \le\beta_1h^*+4A_L+h^*\mathcal E_T\,.
\]
\end{lemma}
\begin{proof}
\interdisplaylinepenalty=10000
Let $F_t=h^*-\KL(q_t,\mu)$ within this proof. First split
\begin{align}
 \mathsf I_1
 &=\beta_1h^*+\sum_{t=2}^T(\beta_t-\beta_{t-1})\E[F_t]
       \label{eq:D-penalty-definition}\\
 &=\beta_1h^*+\sum_{t=2}^T(\beta_t-\beta_{t-1})
       \{\E[\ind_{\mathcal G_t}F_t]+\E[\ind_{\mathcal G_t^c}F_t]\}\,.
       \label{eq:D-penalty-split}
\end{align}
Equation \eqref{eq:D-penalty-split} partitions each expectation.
The next calculation supplies the two required bounds.

On $\mathcal G_t$, let $\xi_t(a)=X_t(a)/\beta_t$ and
$Z_t^*=\pi^*+\int_{a\ne a^*}e^{-\xi_t(a)}\mu(da)$.
Here $\xi_t\ge0$, $b_{r_t}=a^*$, and $q_t(da)=e^{-\xi_t(a)}\mu(da)/Z_t^*$.
Direct substitution into relative entropy gives
\begin{align}
\KL(q_t,\mu)
 &=\int\log\!\left(\frac{e^{-\xi_t(a)}}{Z_t^*}\right)q_t(da)
 =-\int\xi_t(a)q_t(da)-\log Z_t^*\,,\label{eq:D-good-KL-density}\\
 h^*-\KL(q_t,\mu)
 &=\log(Z_t^*/\pi^*)+\int\xi_t(a)q_t(da)\label{eq:D-good-entropy-identity}\\
 &=\log\!\left(1+(\pi^*)^{-1}\int_{a\ne a^*}e^{-\xi_t(a)}\mu(da)\right)
   +\frac{\int_{a\ne a^*}\xi_t(a)e^{-\xi_t(a)}\mu(da)}{Z_t^*}
       \label{eq:D-good-entropy-substitute}\\
 &\le(\pi^*)^{-1}\int_{a\ne a^*}e^{-\xi_t(a)}\mu(da)
     +(\pi^*)^{-1}\int_{a\ne a^*}\xi_t(a)e^{-\xi_t(a)}\mu(da)
       \label{eq:D-good-entropy-log-bound}\\
 &=(\pi^*)^{-1}\int_{a\ne a^*}(1+\xi_t(a))e^{-\xi_t(a)}\mu(da)
       \label{eq:D-good-entropy-collect}\\
 &\le2(\pi^*)^{-1}\int_{a\ne a^*}e^{-\xi_t(a)/2}\mu(da)
 \le2A_Lt^{-6}\,.\label{eq:SCr-good-entropy}
\end{align}
Equation \eqref{eq:D-good-KL-density} substitutes the exponential-weights density.
Equation \eqref{eq:D-good-entropy-identity} substitutes $h^*=-\log\pi^*$.
Equation \eqref{eq:D-good-entropy-substitute} separates the atom at $a^*$,
where $\xi_t(a^*)=0$.
Equation \eqref{eq:D-good-entropy-log-bound} uses $\log(1+x)\le x$
and $Z_t^*\ge\pi^*$; the numerator is nonnegative.
Equation \eqref{eq:D-good-entropy-collect} combines the two integrals.
For \eqref{eq:SCr-good-entropy}, differentiation gives
$\max_{x\ge0}(1+x)e^{-x/2}=2e^{-1/2}<2$, and
\Cref{lem:D-laplace} supplies the remaining integral bound.

On all histories $F_t\le h^*$ since $\KL\ge0$; no nonnegativity of
$F_t$ is assumed. Consequently,
\begin{align}
\E[h^*-\KL(q_t,\mu)]
 &=\E[\ind_{\mathcal G_t}(h^*-\KL(q_t,\mu))]
   +\E[\ind_{\mathcal G_t^c}(h^*-\KL(q_t,\mu))]
       \label{eq:D-entropy-event-split}\\
 &\le2A_Lt^{-6}\sP(\mathcal G_t)+h^*\sP(\mathcal G_t^c)
 \le2A_Lt^{-6}+h^*\sP(\mathcal G_t^c)\,,\label{eq:D-entropy-event-bound}
\end{align}
\begin{align}
 \beta_t-\beta_{t-1}
 &=\sqrt t-\sqrt{t-1}
 =\frac{t-(t-1)}{\sqrt t+\sqrt{t-1}}
 \le t^{-1/2}\,,\label{eq:D-entropy-schedule-difference}
\end{align}
\begin{align}
 \sum_{t=2}^T(\beta_t-\beta_{t-1})\E[h^*-\KL(q_t,\mu)]
 &\le\sum_{t=2}^Tt^{-1/2}
       [2A_Lt^{-6}+h^*\sP(\mathcal G_t^c)]\label{eq:D-entropy-sum-direct}\\
 &\le2A_L\sum_{t\ge1}t^{-13/2}+h^*\mathcal E_T
 \le4A_L+h^*\mathcal E_T\,,\label{eq:SCr-entropy-sum}
\end{align}
Equation \eqref{eq:D-entropy-event-split} partitions the expectation.
Equation \eqref{eq:D-entropy-event-bound} uses the good-event estimate
\eqref{eq:SCr-good-entropy} and the all-history bound $F_t\le h^*$.
Equation \eqref{eq:D-entropy-schedule-difference} rationalizes the square-root
difference. Equation \eqref{eq:D-entropy-sum-direct} first bounds the
expectation by its nonnegative envelope, then bounds the schedule increment.
Equation \eqref{eq:SCr-entropy-sum} evaluates the convergent series.
Returning to the defining term gives the terminal bound
\begin{align}
 \mathsf I_1
 &=\beta_1h^*+\sum_{t=2}^T(\beta_t-\beta_{t-1})\E[F_t]
 \le\beta_1h^*+4A_L+h^*\mathcal E_T\,.
 \label{eq:D-penalty-final}
\end{align}
Equation \eqref{eq:D-penalty-final} adds the initialization to
\eqref{eq:SCr-entropy-sum}; every early round is included in $\mathcal E_T$.
\end{proof}

\subsubsection{Why the Continuous Part of the FTRL Measure Is Inexpensive}\label{sec:SH-laplace}
The reference atom alone is not enough: arbitrarily small physical gaps
can leave excessive exponential-weights mass near it. The convolution component and the
priority weights together provide the stronger cap bound needed here.

\begin{lemma}[A volume cap and an atomic cap]\label{lem:SH-caps}
Let $f$ be a nonnegative, nonconstant affine function on $\mathcal K$,
with minimum zero and maximum $W>0$. Then for $z\ge0$ and $h>0$,
\[
 \lambda_{\mathcal K}\{f\le z\}\le kz/W\,,
 \qquad \int e^{-f/h}\,d\lambda_{\mathcal K}\le kh/W\,,
 \qquad \int e^{-f/h}\,d\mu_m\le(mkh/W)^{m}\,.
\]
At the fixed sharp instance, the suboptimal atomic part obeys
\begin{equation}\label{eq:SH-atomic-cap}
 \sum_{j\ne j^*:\,\Delta(b_j)\le z}\pi_j
       \le C_{\mathrm{at}}z^{16}\,,\qquad
 C_{\mathrm{at}}=\Delta_{\rm pre}^{-16}+\varsigma^{-16}\,.
\end{equation}
\end{lemma}
\begin{proof}
Choose $x_1\in\mathcal K$ with $f(x_1)=W$. For $0<z<W$,
let $\lambda=1-z/W$ and
$T_\lambda(x)=x_1+\lambda(x-x_1)$. Affineness gives
\begin{align}
f(T_\lambda(x))&=(1-\lambda)W+\lambda f(x)
 =z+(1-z/W)f(x)\ge z\,,
       \label{eq:D-step-0078}\\
 \lambda_{\mathcal K}(T_\lambda(\mathcal K))&=\lambda^k=(1-z/W)^k\,,
       \label{eq:D-step-0079}\\
 \lambda_{\mathcal K}\{f\le z\}
 &\le1-(1-z/W)^k\le kz/W\,.
 \label{eq:SHr-volume-cap}
\end{align}
Equation \eqref{eq:D-step-0078} substitutes the affine contraction toward a maximizer of $f$, with $\lambda=1-z/W$.
Equation \eqref{eq:D-step-0079} uses the Jacobian $\lambda^k$ of that contraction on the affine hull.
Equation \eqref{eq:SHr-volume-cap} takes the complement of the contracted superlevel set and uses $(1-z/W)^k\ge1-kz/W$.
Each level set of a nonconstant affine function has zero relative
volume, justifying the use of $\le$ in the cap. At $z=0$ the cap has
volume zero; at $z\ge W$ the claimed upper bound is at least one.
For $h>0$, Tonelli and independence of the $m$ volume samples give
\begin{align}
\int e^{-f(x)/h}\lambda_{\mathcal K}(dx)
 &=\int\int_{f(x)}^\infty h^{-1}e^{-z/h}\,dz\,\lambda_{\mathcal K}(dx)
       \label{eq:D-step-0080}\\
 &=\int_0^\infty h^{-1}e^{-z/h}\lambda_{\mathcal K}\{f\le z\}\,dz
 \le\frac{k}{Wh}\int_0^\infty ze^{-z/h}dz
 =\frac{kh}{W}\,,\label{eq:SHr-volume-laplace}\\
 \int e^{-f/h}\,d\mu_m
 &=\E[\exp\!\left(-\frac1{mh}\sum_{i=1}^{m}f(U_i)\right)]
       \label{eq:D-step-0081}\\
 &=\prod_{i=1}^{m}\E[e^{-f(U_i)/(mh)}]
 \le\left(\frac{mkh}{W}\right)^{m}\,.
 \label{eq:SHr-convolution-laplace}
\end{align}
Equation \eqref{eq:D-step-0080} writes $e^{-f/h}$ as the integral of $h^{-1}e^{-z/h}$ over $z\ge f$.
Equation \eqref{eq:SHr-volume-laplace} applies Tonelli, substitutes \eqref{eq:SHr-volume-cap}, and evaluates $\int_0^\infty ze^{-z/h}dz=h^2$.
Equation \eqref{eq:D-step-0081} substitutes the averaging representation of $\mu_m$ and the affinity of $f$.

Equation \eqref{eq:SHr-convolution-laplace} uses independence of the $m$ volume draws and the preceding Laplace bound with scale $mh$.
For the atoms preceding $j^*$, the definition of $\Delta_{\rm pre}$ implies
\begin{align}
\sum_{\substack{j<j^*\\\Delta(b_j)\le z}}\pi_j
 &\le\ind\{z\ge\Delta_{\rm pre}\}\sum_{j<j^*}\pi_j
 \le\ind\{z\ge\Delta_{\rm pre}\}\le\Delta_{\rm pre}^{-16}z^{16}\,.
 \label{eq:SHr-early-atom-cap}
\end{align}
For $j>j^*$, let $d_j=\|b_j-a^*\|_{\mathcal S}>0$. The reinforced
priority satisfies $h_j\ge j+1+\exp(d_j^{-8})\ge j+1+d_j^{-8}$.
Since $x^2e^{-x}\le4e^{-2}<1$ for $x>0$,
\begin{align}
\pi_j=e^{-h_j}
 &\le e^{-j-1}e^{-d_j^{-8}}
 \le e^{-j-1}(d_j^{-8})^{-2}
 =e^{-j-1}d_j^{16}
       \label{eq:D-step-0082}\\
 &\le e^{-j-1}\varsigma^{-16}\Delta(b_j)^{16}\,,
       \label{eq:D-step-0083}\\
 \sum_{\substack{j>j^*\\\Delta(b_j)\le z}}\pi_j
 &\le\varsigma^{-16}z^{16}\sum_{j>j^*}e^{-j-1}
 \le\varsigma^{-16}z^{16}\,,
       \label{eq:D-step-0084}\\
 \sum_{\substack{j\ne j^*\\\Delta(b_j)\le z}}\pi_j
 &\le(\Delta_{\rm pre}^{-16}+\varsigma^{-16})z^{16}
 =C_{\mathrm{at}}z^{16}\,.
 \label{eq:SHr-total-atom-cap}
\end{align}
Equation \eqref{eq:D-step-0082} uses $h_j\ge j+1+\exp(d_j^{-8})\ge j+1+d_j^{-8}$ and $e^{-y}\le y^{-2}$ for $y>0$.
Equation \eqref{eq:D-step-0083} uses sharpness, $d_j\le\Delta(b_j)/\varsigma$.
Equation \eqref{eq:D-step-0084} restricts to atoms of gap at most $z$ and bounds the resulting geometric series by one.

Equation \eqref{eq:SHr-total-atom-cap} adds the early-atom cap and substitutes $C_{\mathrm{at}}=\Delta_{\rm pre}^{-16}+\varsigma^{-16}$.
\end{proof}

\begin{definition}[Laplace constant and concentration threshold]\label{def:D-laplace}
Let $\Delta_{\max}^{\gK}=\max_{a\in\mathcal K}\Delta(a)>0$, and define
\begin{equation}\label{eq:SH-laplace-constant}
 A_L=(\pi^*)^{-1}\left\{16!\,C_{\mathrm{at}}(C_{\beta,\mathrm{sh}}/2)^{16}
                  +\left(\frac{16kC_{\beta,\mathrm{sh}}}{2\Delta_{\max}^{\gK}}\right)^{16}\right\}\,,
 \qquad T_* =\left\lceil\max\{T_\theta,16,(4A_L)^{1/5}\}\right\rceil\,.
\end{equation}
\end{definition}

\begin{lemma}[Laplace decay away from the optimal action]\label{lem:D-laplace}
If $t\ge T_*$ and $X_t(a)\ge4t^{7/8}\Delta(a)$ for all
$a\in\mathcal K$, then
\begin{equation}\label{eq:SH-gibbs-tail}
 \frac1{\pi^*}\int_{a\ne a^*}e^{-X_t(a)/(2\beta_t)}\mu(da)
       \le A_Lt^{-6}\,,\qquad
 q_t(\mathcal K\setminus\{a^*\})\le A_Lt^{-6}\le1/4\,.
\end{equation}
\end{lemma}
\begin{proof}
For completeness, let $\mu_{\rm at}^\circ=\sum_{j\ne j^*}\pi_j\delta_{b_j}$
and $\lambda_t=2t^{7/8}/\beta_t$. The direct Laplace calculation is
\begin{align}
\lambda_t&=2t^{7/8}/\beta_t
        \ge2t^{7/8}/(C_{\beta,\mathrm{sh}}\sqrt t)=2t^{3/8}/C_{\beta,\mathrm{sh}}\,,
       \label{eq:D-step-0085}\\
 \int e^{-\lambda_t\Delta(a)}\mu_{\rm at}^\circ(da)
 &=\int\!\int_{\Delta(a)}^\infty
             \lambda_t e^{-\lambda_t z}\,dz\,\mu_{\rm at}^\circ(da)
       \label{eq:D-step-0086}\\
 &=\lambda_t\int_0^\infty e^{-\lambda_t z}
                 \mu_{\rm at}^\circ\{\Delta\le z\}\,dz
       \label{eq:D-step-0087}\\
 &\le C_{\rm at}\lambda_t\int_0^\infty e^{-\lambda_t z}z^{16}\,dz
 =16!C_{\rm at}\lambda_t^{-16}\,,
       \label{eq:D-step-0088}\\
 \int e^{-\lambda_t\Delta(a)}\mu_m(da)
 &\le\left(\frac{16k}{\lambda_t \Delta_{\max}^{\gK}}\right)^{16}\,.
 \label{eq:SHr-laplace-components}
\end{align}
Equation \eqref{eq:D-step-0085} substitutes $\lambda_t=2t^{7/8}/\beta_t$ and $\beta_t\le C_{\beta,\mathrm{sh}}\sqrt t$.
Equation \eqref{eq:D-step-0086} represents each atomic exponential as a scalar tail integral.
Equation \eqref{eq:D-step-0087} applies Tonelli to interchange the atomic and scalar integrals.

Equation \eqref{eq:D-step-0088} uses the sixteenth-order atomic cap and $\lambda\int_0^\infty e^{-\lambda z}z^{16}dz=16!\lambda^{-16}$.
Equation \eqref{eq:SHr-laplace-components} applies \eqref{eq:SHr-convolution-laplace} in \Cref{lem:SH-caps} with $m=16$, $h=\lambda_t^{-1}$, and $f=\Delta$.
Since $\mu=\sum_j\pi_j\delta_{b_j}+\omega_\mu\mu_m$, $\omega_\mu\le1$, and
$X_t(a)\ge4t^{7/8}\Delta(a)$, substitution gives
\begin{align}
\frac1{\pi^*}\int_{a\ne a^*}e^{-X_t(a)/(2\beta_t)}\mu(da)
 &\le\frac1{\pi^*}\left[
    \int e^{-\lambda_t\Delta}d\mu_{\rm at}^\circ
    +w\int e^{-\lambda_t\Delta}d\mu_m\right]
       \label{eq:D-step-0089}\\
 &\le\frac1{\pi^*}\left[
       16!C_{\rm at}\lambda_t^{-16}
       +\left(\frac{16k}{\lambda_t \Delta_{\max}^{\gK}}\right)^{16}\right]
       \label{eq:D-step-0090}\\
 &\le\frac1{\pi^*}\left[
       16!C_{\rm at}(C_{\beta,\mathrm{sh}}/2)^{16}
       +\left(\frac{16kC_{\beta,\mathrm{sh}}}{2\Delta_{\max}^{\gK}}\right)^{16}\right]t^{-6}
 =A_Lt^{-6}\,,
       \label{eq:D-step-0091}\\
 q_t(\mathcal K\setminus\{a^*\})
 &=\frac{\int_{a\ne a^*}e^{-X_t(a)/\beta_t}\mu(da)}
          {\pi^*+\int_{a\ne a^*}e^{-X_t(a)/\beta_t}\mu(da)}
       \label{eq:D-step-0092}\\
 &\le\frac1{\pi^*}\int_{a\ne a^*}e^{-X_t(a)/\beta_t}\mu(da)
 \le\frac1{\pi^*}\int_{a\ne a^*}e^{-X_t(a)/(2\beta_t)}\mu(da)
       \label{eq:D-step-0093}\\
 &\le A_Lt^{-6}\le\frac1{4t}\le\frac14\,.
 \label{eq:SHr-gibbs-laplace-assembly}
\end{align}
Equation \eqref{eq:D-step-0089} substitutes $X_t(a)\ge4t^{7/8}\Delta(a)$ and decomposes the prior into its suboptimal atoms and volume-average component.
Equation \eqref{eq:D-step-0090} inserts both estimates from \eqref{eq:SHr-laplace-components} and uses $w\le1$.
Equation \eqref{eq:D-step-0091} substitutes $\lambda_t^{-16}\le(C_{\beta,\mathrm{sh}}/2)^{16}t^{-6}$ and the definition of $A_L$.

Equation \eqref{eq:D-step-0092} substitutes the exponential-weights density and separates the mass $\pi^*$ at $X_t(a^*)=0$.
Equation \eqref{eq:D-step-0093} keeps only $\pi^*$ in the denominator and uses $X_t(a)\ge0$ to halve the exponent.
Equation \eqref{eq:SHr-gibbs-laplace-assembly} applies the preceding integral bound and the threshold $t^5\ge4A_L$.
The probability bounds $A_Lt^{-6}$ have total at most
$A_L\sum_{t\ge1}t^{-6}\le2A_L$, independent of the horizon; no
positive minimum physical gap is introduced.

\end{proof}

\subsection{Proof of Stability Term}\label{app:D-stability}
\begin{lemma}[Relative-entropy stability contribution]\label{lem:D-stability-term}
For $\mathsf D_t=\langle q_t,\widehat f_t\rangle+
\beta_t\log\langle q_t,e^{-\widehat f_t/\beta_t}\rangle$,
\[
 \mathsf I_2\coloneq \sum_{t=1}^T\E[\mathsf D_t]
 \le2e^2C_{\beta,\mathrm{sh}}A_L+2d\mathcal E_T\,.
\]
\end{lemma}
\begin{proof}
\interdisplaylinepenalty=10000
The centered exponential inequality in \Cref{lem:chi}
and the all-history proxy function of \Cref{lem:SH-feasible} give
\begin{align}
 \mathsf I_2
 &=\sum_{t=1}^T\E[\E_{t-1}[\mathsf D_t]]
 \le\sum_{t=1}^T\E[\gC_t(p_t)]\notag\\
 &=\sum_{t=1}^T\E[\ind_{\mathcal G_t}\gC_t(p_t)]
       +\sum_{t=1}^T\E[\ind_{\mathcal G_t^c}\gC_t(p_t)]\,.
 \label{eq:D-stability-definition-split}
\end{align}
Equation \eqref{eq:D-stability-definition-split} uses conditional
expectation and partitions the predictable proxy function.
On $\mathcal G_t$, let $\xi_t(x)=X_t(x)/\beta_t$ and
$Z_t^*=\int e^{-\xi_t}\,d\mu\ge\pi^*$.
For each sampled action $x$, set
$f_x(a)=(a-a^*)^\top v_x(p_t)/\beta_t$.
The update magnitude constraint gives $f_x(a^*)=0$ and
$|f_x(a)|\le1+\xi_t(a)/B$. Centering by
\eqref{eq:chi-center} in \Cref{lem:chi} therefore yields
\begin{align}
\gC_t(p_t)
 &=\beta_t\int\!\int
       \chi\!\left(f_x(a)-\int f_x\,dq_t\right)q_t(da)p_t(dx)
       \label{eq:D-good-certificate-center-substitute}\\
 &\le\beta_t\int\!\int\chi(2f_x(a))q_t(da)p_t(dx)
       \label{eq:D-good-certificate-center-bound}\\
 &=\beta_t\int\!\int_{a\ne a^*}\chi(2f_x(a))q_t(da)p_t(dx)
       \label{eq:D-good-certificate-zero-row}\\
 &\le\beta_t\int\!\int_{a\ne a^*}e^{2|f_x(a)|}q_t(da)p_t(dx)
       \label{eq:D-good-certificate-chi-exp}\\
 &\le\beta_t\int\!\int_{a\ne a^*}e^{2+2\xi_t(a)/B}q_t(da)p_t(dx)
       \label{eq:D-good-certificate-mark-budget}\\
 &=e^2\beta_t\int_{a\ne a^*}e^{2\xi_t(a)/B}q_t(da)
       \label{eq:D-good-certificate-mark-integral}\\
 &=\frac{e^2\beta_t}{Z_t^*}\int_{a\ne a^*}
                  e^{2\xi_t(a)/B}e^{-\xi_t(a)}\mu(da)
       \label{eq:D-good-certificate-Gibbs-substitute}\\
 &\le\frac{e^2\beta_t}{\pi^*}\int_{a\ne a^*}
                  e^{-(1-2/B)\xi_t(a)}\mu(da)
       \label{eq:D-good-certificate-normalizer}\\
 &\le e^2\beta_t(\pi^*)^{-1}\int_{a\ne a^*}e^{-\xi_t(a)/2}\mu(da)
       \label{eq:D-good-certificate-exponent}\\
 &\le e^2\beta_t A_Lt^{-6}
 \le e^2C_{\beta,\mathrm{sh}}\sqrt t\,A_Lt^{-6}=e^2C_{\beta,\mathrm{sh}}A_Lt^{-11/2}\,.
       \label{eq:SCr-good-certificate}
\end{align}
Equation \eqref{eq:D-good-certificate-center-substitute} substitutes
$f_x$ and the barycenter in the proxy function.
Equation \eqref{eq:D-good-certificate-center-bound} is
\eqref{eq:chi-center} in \Cref{lem:chi}.
Equation \eqref{eq:D-good-certificate-zero-row} removes the zero
integrand at $a^*$.
Equation \eqref{eq:D-good-certificate-chi-exp} uses
$\chi(y)\le e^{|y|}$; \eqref{eq:D-good-certificate-mark-budget}
substitutes $|f_x(a)|\le1+\xi_t(a)/B$.
Equation \eqref{eq:D-good-certificate-mark-integral} integrates out the
sampling law $p_t$.
Equation \eqref{eq:D-good-certificate-Gibbs-substitute} inserts the
exponential-weights density before \eqref{eq:D-good-certificate-normalizer}
collects the exponent and uses $Z_t^*\ge\pi^*$.
Equation \eqref{eq:D-good-certificate-exponent} uses $B\ge4$ and
$\xi_t\ge0$; \eqref{eq:SCr-good-certificate} inserts the Laplace bound
and then the schedule upper bound.

On all histories $\gC_t(p_t)\le2d/\beta_t\le2d/\sqrt t$.
Partitioning one proxy function expectation therefore gives
\begin{align}
 \E[\gC_t(p_t)]
 &=\E[\ind_{\mathcal G_t}\gC_t(p_t)]
                 +\E[\ind_{\mathcal G_t^c}\gC_t(p_t)]\label{eq:D-certificate-event-split}\\
 &\le e^2C_{\beta,\mathrm{sh}}A_Lt^{-11/2}\sP(\mathcal G_t)
                      +(2d/\sqrt t)\sP(\mathcal G_t^c)\label{eq:D-certificate-event-bound}\\
 &\le e^2C_{\beta,\mathrm{sh}}A_Lt^{-11/2}+2dt^{-1/2}\sP(\mathcal G_t^c)\,,\label{eq:D-certificate-event-relax}
\end{align}
Equation \eqref{eq:D-certificate-event-split} is the event partition;
\eqref{eq:D-certificate-event-bound} inserts the two envelopes, and
\eqref{eq:D-certificate-event-relax} drops a probability factor at most one.
Summing this bound gives
\begin{align}
 \sum_{t=1}^T\E[\gC_t(p_t)]
 &\le e^2C_{\beta,\mathrm{sh}}A_L\sum_{t=1}^Tt^{-11/2}
                      +2d\sum_{t=1}^Tt^{-1/2}\sP(\mathcal G_t^c)\label{eq:D-certificate-sum-direct}\\
 &\le2e^2C_{\beta,\mathrm{sh}}A_L+2d\mathcal E_T\,.
       \label{eq:SCr-stability-sum}
\end{align}
Equation \eqref{eq:D-certificate-sum-direct} sums the one-round estimate;
\eqref{eq:SCr-stability-sum} uses $\sum_{t\ge1}t^{-11/2}\le2$ and the
definition of $\mathcal E_T$. Thus
\begin{equation}\label{eq:D-stability-final}
 \mathsf I_2\le\sum_{t=1}^T\E[\gC_t(p_t)]
                 \le2e^2C_{\beta,\mathrm{sh}}A_L+2d\mathcal E_T\,.
\end{equation}
This is precisely the stability contribution used in the full proof.
\end{proof}

\subsection{Proof of Exploration Term}\label{app:D-exploration}
\begin{lemma}[Exploration contribution under the fixed loss vector]\label{lem:D-exploration-term}
For a fixed feasible centered sampling allocation $n^{\rm des}$, define
\[
 a_{n^{\rm des}}=\max\{1,4BC_{\beta,\mathrm{sh}}N_{n^{\rm des}}/d\}\,,\quad a_\rho=16B\,,
 \qquad
 M_{3,\theta,n^{\rm des}}=4A_LC_{\mathrm{rat}}^2+
       8d(4C_{\mathrm{rat}}^4a_{n^{\rm des}}+\tfrac83C_{\mathrm{rat}}^3\sqrt{a_\rho})\,.
\]
Then
\[
 \mathsf I_3\coloneq \sum_{t=1}^T\E[\langle p_t-q_t,\Delta\rangle]
 \le16BC_{\mathrm{rat}}^4c_{\mathrm{ctr}}^{n^{\rm des}}\log(eT)+M_{3,\theta,n^{\rm des}}
                         +8d\mathcal E_T\,.
\]
\end{lemma}
\begin{proof}
\interdisplaylinepenalty=10000
The defining sum splits exactly as
\begin{align}
 \mathsf I_3
 &=\sum_{t=1}^T\E[\ind_{\mathcal G_t}\langle p_t-q_t,\Delta\rangle]
  +\sum_{t=1}^T\E[\ind_{\mathcal G_t^c}\langle p_t-q_t,\Delta\rangle]\,.
 \label{eq:D-exploration-definition-split}
\end{align}
Equation \eqref{eq:D-exploration-definition-split} is the event partition.
We first bound its good-history summands by two feasible comparisons.

Let $\rho_t=16B\lo_t^2/t$ and
$\mathcal W_t=\mathcal G_t\cap\{\lo_t^2\le\sqrt t/a_{n^{\rm des}}\}
\cap\{\lo_t^2\le t/a_\rho\}$. All hypotheses needed to prove feasibility of the comparison distribution
constructed from $n^{\rm des}$ are supplied by \Cref{lem:SC-good}.
The comparisons with the distribution constructed from $n^{\rm des}$
and with the fallback distribution, now for the corrupted cumulative
loss estimates, read
\begin{align}
\langle p_t,\Delta\rangle
 &\le\lo_t\langle p_t,\widehat\Delta_t\rangle
 \le\lo_t\langle p_t^c,\widehat\Delta_t\rangle
       \label{eq:D-step-0032}\\
 &=\lo_t(1-\gamma_t^c)\langle q_t,\widehat\Delta_t\rangle
   +\lo_t\rho_t\int_{a\ne a^*}\widehat\Delta_t(a)n^{\rm des}(da)
       \label{eq:D-step-0033}\\
 &\le\lo_t\up_t\langle q_t,\Delta\rangle
      +\frac{16Bc_{\mathrm{ctr}}^{n^{\rm des}}}{t}\lo_t^3\up_t
                   \qquad(\mathcal W_t)\,,\label{eq:SCr-witness-cost}
\end{align}
\begin{align}
 \langle p_t,\Delta\rangle
 &\le\lo_t\langle p_t,\widehat\Delta_t\rangle
 \le\lo_t\langle(1-\Gamma_t)q_t+\Gamma_t\pi,\widehat\Delta_t\rangle
       \label{eq:D-step-0034}\\
 &\le\lo_t\up_t\langle q_t,\Delta\rangle+2\Gamma_t\lo_t\up_t
                   \qquad(\mathcal G_t\setminus\mathcal W_t)\,.
                   \label{eq:SCr-fallback-cost}
\end{align}
Equation \eqref{eq:D-step-0032} uses the clipped-gap comparison and
optimality of $p_t$ against the feasible comparison distribution $p_t^c$ with its prescribed covariance estimator.
Equation \eqref{eq:D-step-0033} substitutes the comparison distribution constructed from $n^{\rm des}$ in the clipped-gap objective.
Equation \eqref{eq:SCr-witness-cost} uses $\widehat\Delta_t\le\up_t\Delta$, drops $1-\gamma_t^c\le1$, and substitutes $\rho_t=16B\lo_t^2/t$ and $c_{\mathrm{ctr}}^{n^{\rm des}}=\int\Delta\,dn^{\rm des}$.

Equation \eqref{eq:D-step-0034} applies objective comparison to the feasible
covariance fallback on $\mathcal G_t\setminus\mathcal W_t$, where the comparison distribution's feasibility conditions need not hold.
Equation \eqref{eq:SCr-fallback-cost} uses $\widehat\Delta_t\le\up_t\Delta$, $\int\Delta\,d\pi\le2$, and $1-\Gamma_t\le1$.
Both comparisons follow from optimality in \eqref{eq:SH-program},
since the distribution constructed from $n^{\rm des}$ and the fallback
distribution satisfy every constraint on their respective events
with their prescribed covariance coefficients.
Apply \eqref{eq:SHr-mass-tail} in \Cref{lem:D-ratio-tail-sums} at
$z=\sqrt t/a_{n^{\rm des}}$ and $z=t/a_\rho$. Since
$\langle p_t-q_t,\Delta\rangle\le\langle p_t,\Delta\rangle$,
\begin{align}
\E[\lo_t\up_t\ind\{\lo_t^2>\sqrt t/a_{n^{\rm des}}\}]
 &\le\min\{C_{\mathrm{rat}}^2,C_{\mathrm{rat}}^6(\sqrt t/a_{n^{\rm des}})^{-2}\}
 =\min\{C_{\mathrm{rat}}^2,C_{\mathrm{rat}}^6a_{n^{\rm des}}^2/t\}\,,\label{eq:D-cor-witness-mass-tail}
\end{align}
\begin{align}
 \E[\lo_t\up_t\ind\{\lo_t^2>t/a_\rho\}]
 &\le\min\{C_{\mathrm{rat}}^2,C_{\mathrm{rat}}^6(t/a_\rho)^{-2}\}
 =\min\{C_{\mathrm{rat}}^2,C_{\mathrm{rat}}^6a_\rho^2/t^2\}\,,\label{eq:D-cor-witness-rho-tail}
\end{align}
\begin{align}
 \E[\ind_{\mathcal G_t}\langle p_t-q_t,\Delta\rangle]
 &\le\frac{16Bc_{\mathrm{ctr}}^{n^{\rm des}}}{t}\E[\lo_t^3\up_t]
       +2A_Lt^{-6}\E[\lo_t\up_t]\label{eq:D-cor-good-cost-unsimplified}\\*
 &\quad+\frac{8d}{\sqrt t}\left[
  \E[\lo_t\up_t\ind\{\lo_t^2>\sqrt t/a_{n^{\rm des}}\}]
 +\E[\lo_t\up_t\ind\{\lo_t^2>t/a_\rho\}]\right]
       \label{eq:D-step-0035}\\
 &\le\begin{aligned}[t]
 &\frac{16BC_{\mathrm{rat}}^4c_{\mathrm{ctr}}^{n^{\rm des}}}{t}+2A_LC_{\mathrm{rat}}^2t^{-6}\\
 &+\frac{8d}{\sqrt t}\left[\min\{C_{\mathrm{rat}}^2,C_{\mathrm{rat}}^6a_{n^{\rm des}}^2/t\}
                     +\min\{C_{\mathrm{rat}}^2,C_{\mathrm{rat}}^6a_\rho^2/t^2\}\right]\,,
 \end{aligned}\label{eq:D-cor-good-cost-evaluated}
\end{align}
\begin{align}
 \E[\ind_{\mathcal G_t^c}\langle p_t-q_t,\Delta\rangle]
 &\le\E[\ind_{\mathcal G_t^c}2\Gamma_t]
 =2\Gamma_t\sP(\mathcal G_t^c)
 \le8d\,t^{-1/2}\sP(\mathcal G_t^c)\,,\label{eq:D-cor-bad-exploration}
\end{align}
\begin{align}
 \sum_{t=1}^T\E[\langle p_t-q_t,\Delta\rangle]
 &\le16BC_{\mathrm{rat}}^4c_{\mathrm{ctr}}^{n^{\rm des}}\sum_{t=1}^Tt^{-1}
    +2A_LC_{\mathrm{rat}}^2\sum_{t=1}^Tt^{-6}+8d\,\mathcal E_T\label{eq:D-cor-exploration-sums-substitute}\\*
 &\quad+8d\sum_{t=1}^Tt^{-1/2}\min\{C_{\mathrm{rat}}^2,C_{\mathrm{rat}}^6a_{n^{\rm des}}^2/t\}
       +8d\sum_{t=1}^Tt^{-1/2}\min\{C_{\mathrm{rat}}^2,C_{\mathrm{rat}}^6a_\rho^2/t^2\}
       \label{eq:D-step-0036}\\
 &\le16BC_{\mathrm{rat}}^4c_{\mathrm{ctr}}^{n^{\rm des}}\log(eT)+4A_LC_{\mathrm{rat}}^2
       +8d\left(4C_{\mathrm{rat}}^4a_{n^{\rm des}}+\frac83C_{\mathrm{rat}}^3\sqrt{a_\rho}\right)
       +8d\mathcal E_T\,.
       \label{eq:SCr-exploration-sum}
\end{align}
Equation \eqref{eq:D-cor-witness-mass-tail} applies the ratio tail bound at $z=\sqrt t/a_{n^{\rm des}}$ and then simplifies the substituted power.
Equation \eqref{eq:D-cor-witness-rho-tail} applies the same ratio tail bound at $z=t/a_\rho$ and simplifies that substituted power.
Equation \eqref{eq:D-cor-good-cost-unsimplified} uses the harmonic cost of the feasible comparison distribution and the expected gap under exponential weights on good histories.

Equation \eqref{eq:D-step-0035} adds the costs of failures of both feasibility thresholds for the comparison distribution without yet simplifying their expectations.
Equation \eqref{eq:D-cor-good-cost-evaluated} inserts the two ratio-tail bounds and the product moments.
Equation \eqref{eq:D-cor-bad-exploration} conditions with the predictable bad-event indicator, uses domination, and substitutes $\Gamma_t\le4d/\sqrt t$.

Equation \eqref{eq:D-cor-exploration-sums-substitute} sums the harmonic, exponential-weights tail, and bad-history terms over the full horizon.
Equation \eqref{eq:D-step-0036} retains both sums for failures of the comparison distribution's feasibility thresholds in that same inequality.
Equation \eqref{eq:SCr-exploration-sum} applies the harmonic and convergent-series bounds and substitutes the two evaluated split integrals.
The tail estimates used in \eqref{eq:D-cor-witness-mass-tail} and
\eqref{eq:D-cor-witness-rho-tail}, and the evaluated integrals used in
\eqref{eq:SCr-exploration-sum}, are respectively
\eqref{eq:SHr-mass-tail}, \eqref{eq:SHr-mass-sum}, and
\eqref{eq:SHr-rho-sum} in \Cref{lem:D-ratio-tail-sums}.
The signed-measure estimate behind \eqref{eq:D-cor-bad-exploration}
is \eqref{eq:D-exploration-signed-law} in
\Cref{prop:D-adversarial}; here
$\Delta\in[0,2]$ and $\Gamma_t$ is deterministic.

Substituting the definition of $M_{3,\theta,n^{\rm des}}$ in
\eqref{eq:SCr-exploration-sum} completes the calculation:
\begin{equation}\label{eq:D-exploration-final}
 \mathsf I_3\le16BC_{\mathrm{rat}}^4c_{\mathrm{ctr}}^{n^{\rm des}}\log(eT)
             +M_{3,\theta,n^{\rm des}}+8d\mathcal E_T\,.
\end{equation}
\end{proof}

\subsubsection{A Feasible Comparison Distribution Constructed from a Sampling Allocation}\label{sec:SH-witness}

We use the centered sampling allocations used for comparison of
\Cref{def:D-physical-design} and
\eqref{eq:SH-centered-program}. As throughout this appendix, $\cA$
denotes the physical generator $\mathcal B$ of the main theorem and
$\mathcal K=\Conv(\cA)$. The following lemma records their optimal
cost and the extension of their constraints to the convex hull.

\begin{lemma}[Finite centered sampling allocations under sharpness]
\label{lem:D-centered-value}
Under sharpness, the centered program is feasible and
\begin{equation}\label{eq:SH-design-equality}
 \inf_{n\text{ centered feasible on }\cA}c_{\mathrm{ctr}}^{n}(\cA,\theta)=\cctr(\cA,\theta)<\infty\,,
\end{equation}
where the infimum is over finite nonnegative sampling allocations.
\end{lemma}
\begin{proof}
For a finite measure $n$ with $H_n\succ0$ and a convex combination
$x=\sum_{i=1}^m\lambda_i a_i$, $a_i\in\cA$, the physical constraints imply
\begin{align}
\|x-a^*\|_{H_n^{-1}}
 &=\left\|\sum_{i=1}^m\lambda_i(a_i-a^*)\right\|_{H_n^{-1}}
 \le\sum_{i=1}^m\lambda_i\|a_i-a^*\|_{H_n^{-1}}
       \label{eq:D-centered-convex-triangle}\\
 &\le\sum_{i=1}^m\lambda_i\frac{\Delta(a_i)}{\sqrt2}
 =\frac{\Delta(\sum_i\lambda_i a_i)}{\sqrt2}
 =\frac{\Delta(x)}{\sqrt2}\,.
       \label{eq:D-centered-convex-gap}
\end{align}
Equation \eqref{eq:D-centered-convex-triangle} is the triangle
inequality for the positive-definite quadratic norm.
Equation \eqref{eq:D-centered-convex-gap} substitutes the physical
constraints and uses affineness of $\Delta$.
Thus testing on $\cA$ and on $\mathcal K$ gives the same feasible sampling allocations.
To prove existence, take a finite sampling allocation $n_0$ with $H_0\coloneq H_{n_0}\succ0$,
which exists because the actions span. Norm equivalence gives a finite
$L_0>0$ such that
$\|v\|_{H_0^{-1}}^2\le L_0\|v\|_{\mathcal S}^2$ for every $v$.
Set $\alpha=\max\{1,2L_0/\varsigma^2\}$ and $n=\alpha n_0$. Then
\[
 \|a-a^*\|_{H_n^{-1}}^2
 \le\frac{L_0}{\alpha}\|a-a^*\|_{\mathcal S}^2
 \le\frac{\Delta(a)^2}{2}\,.
\]
This sampling allocation has finite mass and finite cost because the gaps are bounded.
By definition,
\begin{equation}\label{eq:D-centered-value-substitution}
 \cctr(\cA,\theta)
 =\inf_{\substack{n(\cA)<\infty\,,\ H_n\succ0\\
 \|a-a^*\|_{H_n^{-1}}^2\le\Delta(a)^2/2\ (a\in\cA)}}c_{\mathrm{ctr}}^{n}<\infty\,.
\end{equation}
The equality concerns an infimum; it does not assume attainment or a
horizon-dependent comparison sampling allocation.
\end{proof}

\begin{definition}[The feasible comparison distribution constructed from the sampling allocation $n$]\label{def:D-witness}
For each finite feasible $n$ set
\begin{equation}\label{eq:SH-witness-parameters}
 N_n=n(\cA)\,,\quad a_n=\max\{1,4BC_{\beta,\mathrm{sh}}N_n/d\}\,,\quad a_\rho=16B\,,
 \quad\rho_t=16B\lo_t^2/t\,,
\end{equation}
\[
 \mathcal W_t=\mathcal G_t\cap
   \{\lo_t^2\le\sqrt t/a_n\}\cap\{\lo_t^2\le t/a_\rho\}\,,
 \quad
 p_t^c=(1-\gamma_t^c)q_t+\rho_t n|_{\cA\setminus\{a^*\}}\,,
 \quad\gamma_t^c=\rho_t n(\cA\setminus\{a^*\})\,.
\]
The distribution $p_t^c$ is constructed from the sampling allocation $n$ for
comparison in the proof; its feasibility on $\mathcal W_t$ is
proved next. The learner does not compute this distribution or test
this event.
\end{definition}

\begin{lemma}[The comparison distribution constructed from a centered sampling allocation satisfies all budgets]\label{lem:SH-witness}
For $t\ge T_*$, on $\mathcal W_t$, the distribution $p_t^c$ with mark
$v(x)=\Sigma(p_t^c)^{-1}x$ is feasible in
\eqref{eq:SH-program} in \GIAFTRL{} and \eqref{eq:SH-budgets} in \GIAFTRL{}.
\end{lemma}
\begin{proof}
Work on $\mathcal W_t$, $t\ge T_*$. Let $\rho=\rho_t$,
$\gamma=\gamma_t^c$, and $M=\Sigma(p_t^c)$.
The two restrictions defining $\mathcal W_t$ give
\begin{align}
\rho N_n&=\frac{16B\lo_t^2N_n}{t}
 \le\frac{16BN_n}{a_n\sqrt t}
 \le\frac{4d}{C_{\beta,\mathrm{sh}}\sqrt t}\le\frac{4d}{\beta_t}=\Gamma_t<\frac14\,,
 &\rho&\le\frac{16B}{t}\frac{t}{a_\rho}=1\,,
       \label{eq:D-step-0066}\\
 p_t^c(\{a^*\})
 &=(1-\gamma)q_t(\{a^*\})
 \ge(1-\Gamma_t)(1-A_Lt^{-6})\ge\frac9{16}\,,
       \label{eq:D-step-0067}\\
 \rho n(\{a^*\})&\le\rho N_n\le\Gamma_t<\frac14<\frac9{16}\le p_t^c(\{a^*\})\,,
       \label{eq:D-step-0068}\\
 p_t^c&\ge(1-\Gamma_t)q_t\,,
 &M&\succeq\rho H_n\succ0\,.
 \label{eq:SHr-witness-mass}
\end{align}
Equation \eqref{eq:D-step-0066} substitutes $\rho=16B\lo_t^2/t$, both feasibility thresholds for the comparison distribution, $a_n\ge4BC_{\beta,\mathrm{sh}}N_n/d$, and $\beta_t\le C_{\beta,\mathrm{sh}}\sqrt t$.
Equation \eqref{eq:D-step-0067} evaluates the comparison distribution's mass at the optimal atom and uses $1-\Gamma_t\ge3/4$ and $1-A_Lt^{-6}\ge3/4$.
Equation \eqref{eq:D-step-0068} compares the omitted atom of the sampling allocation with the mass already present at $a^*$.

Equation \eqref{eq:SHr-witness-mass} uses $\gamma\le\Gamma_t$ for marginal domination and integrates $xx^\top$ against $p_t^c\ge\rho n$.
The covariance domination includes the atom at $a^*$: away from $a^*$,
$p_t^c$ contains $\rho n$, and the preceding line supplies the missing
mass at $a^*$. For $v(x)=M^{-1}x$,
\begin{align}
\int v(x)x^\top p_t^c(dx)&=M^{-1}M=I_d\,,
       \label{eq:D-step-0069}\\
 V^c(u)\coloneq \int[(u-a^*)^\top v(x)]^2p_t^c(dx)
 &=(u-a^*)^\top M^{-1}(u-a^*)
       \label{eq:D-step-0070}\\
 &\le\rho^{-1}(u-a^*)^\top H_n^{-1}(u-a^*)
 \le\frac{\Delta(u)^2}{2\rho}\,.
 \label{eq:SHr-witness-variance}
\end{align}
Equation \eqref{eq:D-step-0069} substitutes $v(x)=M^{-1}x$ and evaluates the covariance integral.
Equation \eqref{eq:D-step-0070} expands the squared reconstruction coefficient and cancels the covariance matrix.
Equation \eqref{eq:SHr-witness-variance} uses $M\succeq\rho H_n$ and the centered physical information constraint extended to $\mathcal K$.
The positive atom at $a^*$ supplies the missing reconstruction
leverage without charging additional gap cost. Let
$p^*=p_t^c(\{a^*\})$ and $v^*=M^{-1/2}a^*$. Then
\begin{align}
M&\succeq p^*a^*(a^*)^{\top}\,,
 \qquad I=M^{-1/2}MM^{-1/2}\succeq p^*v^*(v^*)^{\top}\,,
       \label{eq:D-witness-atom-congruence}\\
 p^*\|v^*\|_2^2&=\lambda_{\max}(p^*v^*(v^*)^{\top})\le1\,,
       \label{eq:D-witness-atom-eigenvalue}\\
 (a^*)^{\top} M^{-1}a^*&=\|v^*\|_2^2\le (p^*)^{-1}
                         \le16/9\le2\,.
       \label{eq:D-witness-atom-leverage}
\end{align}
Equation \eqref{eq:D-witness-atom-congruence} keeps just the atom in the
covariance integral and conjugates by $M^{-1/2}$.
Equation \eqref{eq:D-witness-atom-eigenvalue} evaluates the only possible
nonzero eigenvalue of a rank-one matrix.
Equation \eqref{eq:D-witness-atom-leverage} substitutes
$p^*\ge9/16$ from \eqref{eq:SHr-witness-mass}.
\begin{align}
x^\top M^{-1}x
 &\le2(x-a^*)^\top M^{-1}(x-a^*)+2(a^*)^{\top} M^{-1}a^*
       \label{eq:D-step-0071}\\
 &\le\frac{\Delta(x)^2}{\rho}+4\le\frac4\rho+4\le\frac8\rho\,,
       \label{eq:D-step-0072}\\
 U^c(u)\coloneq \sup_x|(u-a^*)^\top M^{-1}x|
 &\le\sqrt{V^c(u)}\sup_x\sqrt{x^\top M^{-1}x}
 \le\sqrt{\frac{\Delta(u)^2}{2\rho}}\sqrt{\frac8\rho}
 =\frac{2\Delta(u)}\rho\,.
 \label{eq:SHr-witness-amplitude}
\end{align}
Equation \eqref{eq:D-step-0071} writes $x=(x-a^*)+a^*$ and uses the squared norm inequality $\|v+w\|^2\le2\|v\|^2+2\|w\|^2$.
Equation \eqref{eq:D-step-0072} inserts \eqref{eq:SHr-witness-variance} and \eqref{eq:D-witness-atom-leverage}, then uses $\Delta(x)\le2$ and $\rho\le1$.
Equation \eqref{eq:SHr-witness-amplitude} applies covariance Cauchy--Schwarz to $(u-a^*)^\top M^{-1}x$ and evaluates both square roots.
On the good event $t\Delta(u)\le\lo_tX_t(u)$ and $\lo_t\ge1$.
Substituting $\rho=16B\lo_t^2/t$ first gives
\begin{align}
V^c(u)&\le\frac{\Delta(u)^2}{2(16B\lo_t^2/t)}
 =\frac{t\Delta(u)^2}{32B\lo_t^2}\label{eq:D-witness-variance-rho}\\
 &\le\frac{t}{32B\lo_t^2}
             \left(\frac{\lo_tX_t(u)}t\right)^2
 =\frac{X_t(u)^2}{32Bt}\label{eq:D-witness-variance-gap}\\
 &\le\frac{X_t(u)^2}{Bt}
 \le\frac{\beta_t}{4}\|u-a^*\|_{\mathcal S}^2
                  +\frac{X_t(u)^2}{Bt}=W_t(u)\,,
       \label{eq:D-witness-variance-budget}\\
 U^c(u)&\le\frac{2\Delta(u)}{16B\lo_t^2/t}
 =\frac{t\Delta(u)}{8B\lo_t^2}\label{eq:D-witness-amplitude-rho}\\
 &\le\frac{\lo_tX_t(u)}{8B\lo_t^2}
 =\frac{X_t(u)}{8B\lo_t}\label{eq:D-witness-amplitude-gap}\\
 &\le\frac{X_t(u)}{8B}\le\frac{X_t(u)}B
 \le\frac{\beta_t}{2}\|u-a^*\|_{\mathcal S}+\frac{X_t(u)}B=U_t(u)\,.
       \label{eq:SHr-witness-budgets}
\end{align}
Equations \eqref{eq:D-witness-variance-rho} and
\eqref{eq:D-witness-amplitude-rho} substitute $\rho_t$ literally.
Equations \eqref{eq:D-witness-variance-gap} and
\eqref{eq:D-witness-amplitude-gap} use $t\Delta(u)\le\lo_tX_t(u)$.
Equation \eqref{eq:D-witness-variance-budget} adds the nonnegative
geometric term of the second-moment budget. Equation \eqref{eq:SHr-witness-budgets} uses
$\lo_t\ge1$ and $X_t(u)\ge0$. The budget equalities use
$b_{r_t}=a^*$ and $Z_t(u)=X_t(u)$ on $\mathcal W_t$.
Here $b_{r_t}=a^*$ and $Z_t(u)=X_t(u)\ge0$, so the final comparisons use
exactly \eqref{eq:SH-budgets} in \GIAFTRL{}.

It remains to verify the exponential constraint for the same law.
Apply \eqref{eq:chi-center} in \Cref{lem:chi} to
$f_x(u)=(u-a^*)^\top M^{-1}x/\beta_t$, then
$\chi(y)\le y^2e^{|y|}$. Equations
\eqref{eq:SHr-witness-variance}--\eqref{eq:SHr-witness-amplitude} give
\begin{align}
\gC_t(p_t^c)
 &\le\beta_t\iint_{u\ne a^*}\chi(2f_x(u))q_t(du)p_t^c(dx)
       \label{eq:D-step-0073}\\
 &\le\frac4{\beta_t}\int_{u\ne a^*}
           e^{2U^c(u)/\beta_t}V^c(u)q_t(du)
       \label{eq:D-step-0074}\\
 &\le\frac2{\beta_t\rho}\int_{u\ne a^*}
           \Delta(u)^2e^{4\Delta(u)/(\beta_t\rho)}q_t(du)
       \label{eq:D-step-0075}\\
 &\le\frac8{\beta_t\rho}\int_{u\ne a^*}
           \Delta(u)^2e^{4\Delta(u)/(\beta_t\rho)}q_t(du)\,.
 \label{eq:SHr-witness-exponential}
\end{align}
Equation \eqref{eq:D-step-0073} uses centered symmetrization for $f_x$ and its zero value at the optimal atom.
Equation \eqref{eq:D-step-0074} applies $\chi(2y)\le4y^2e^{2|y|}$, uses the pathwise update magnitude, and integrates the squared coefficient.
Equation \eqref{eq:D-step-0075} substitutes $U^c(u)\le2\Delta(u)/\rho$ and $V^c(u)\le\Delta(u)^2/(2\rho)$.

Equation \eqref{eq:SHr-witness-exponential} enlarges the nonnegative coefficient from $2$ to $8$, without altering the integrand.
The exponent and normalization in \eqref{eq:SHr-witness-exponential}
are controlled as follows:
\begin{align}
\frac{4\Delta(u)}\rho
 &=\frac{t\Delta(u)}{4B\lo_t^2}
 \le\frac{X_t(u)}{4B\lo_t}\le\frac{X_t(u)}{4B}\,,
       \label{eq:D-step-0076}\\
 \gC_t(p_t^c)
 &\le\frac{32}{\beta_t\rho\pi^*}
       \int_{u\ne a^*}e^{-(1-1/(4B))X_t(u)/\beta_t}\mu(du)
       \label{eq:D-step-0077}\\
 &\le\frac{32A_L}{\beta_t\rho}t^{-6}
 \le\frac{32A_L}{\beta_t(16B/t)}t^{-6}
 =\frac{2A_L}{B\beta_t}t^{-5}\le\frac{2d}{\beta_t}\,.
 \label{eq:SHr-witness-final-certificate}
\end{align}
Equation \eqref{eq:D-step-0076} substitutes $\rho=16B\lo_t^2/t$, uses $t\Delta(u)\le\lo_tX_t(u)$, and then $\lo_t\ge1$.
Equation \eqref{eq:D-step-0077} substitutes the exponential-weights density bound $q_t(du)\le(\pi^*)^{-1}e^{-X_t(u)/\beta_t}\mu(du)$ and $\Delta(u)^2\le4$.
Equation \eqref{eq:SHr-witness-final-certificate} applies \eqref{eq:SH-gibbs-tail} in \Cref{lem:D-laplace}, substitutes $\rho\ge16B/t$, and uses $A_Lt^{-5}\le1/4$ and $d\ge1$.
Thus the single comparison measure satisfies every
constraint, including the constraints indexed by arbitrarily small
positive gaps.
\end{proof}

\subsubsection{Ratio Tails and Feasibility-Threshold Sums}
\begin{lemma}[Ratio tails and feasibility-threshold sums]
\label{lem:D-ratio-tail-sums}
Suppose the nonnegative ratios $\lo_t,\up_t$, defined to vanish off
$\mathcal G_t$, satisfy
$\E[\lo_t\up_t]\le C_{\mathrm{rat}}^2$ and
$\E[\lo_t^5\up_t]\le C_{\mathrm{rat}}^6$.
Then \eqref{eq:SHr-mass-tail} holds for every $z>0$.
For any positive feasibility parameters $a_n,a_\rho$, the scalar sums
are bounded as in \eqref{eq:SHr-mass-sum}--\eqref{eq:SHr-rho-sum}.
The moment assumptions follow from \Cref{lem:SH-ratios}
in the stochastic case and from \Cref{lem:SC-good}
in the corrupted stochastic case.
\end{lemma}
\begin{proof}
For every $z>0$, the two product-moment bounds give
\begin{align}
\E[\lo_t\up_t\ind\{\lo_t^2>z\}]&\le\E[\lo_t\up_t]\le C_{\mathrm{rat}}^2\,,
       \label{eq:D-step-0020}\\
 \E[\lo_t\up_t\ind\{\lo_t^2>z\}]
 &\le z^{-2}\E[\lo_t^5\up_t]\le C_{\mathrm{rat}}^6z^{-2}\,,
       \label{eq:D-step-0021}\\
 \E[\lo_t\up_t\ind\{\lo_t^2>z\}]
 &\le\min\{C_{\mathrm{rat}}^2,C_{\mathrm{rat}}^6z^{-2}\}\,.
 \label{eq:SHr-mass-tail}
\end{align}
Equation \eqref{eq:D-step-0020} drops the event indicator and applies the assumed first product-moment bound.
Equation \eqref{eq:D-step-0021} uses $\ind\{\lo_t^2>z\}\le\lo_t^4/z^2$ and then the assumed fifth product-moment bound.
Equation \eqref{eq:SHr-mass-tail} takes the smaller of the two established bounds.
The two nonharmonic exploration sums have the following exact integral
bounds. Let $x_n=C_{\mathrm{rat}}^4a_n^2$ and $x_\rho=C_{\mathrm{rat}}^2a_\rho$. Monotonicity gives
\begin{align}
\sum_{t\ge1}t^{-1/2}\min\{C_{\mathrm{rat}}^2,C_{\mathrm{rat}}^6a_n^2/t\}
 &\le C_{\mathrm{rat}}^2\int_0^{x_n}x^{-1/2}dx
        +C_{\mathrm{rat}}^6a_n^2\int_{x_n}^\infty x^{-3/2}dx
       \label{eq:D-step-0024}\\
 &=2C_{\mathrm{rat}}^2\sqrt{x_n}+2C_{\mathrm{rat}}^6a_n^2x_n^{-1/2}
   =4C_{\mathrm{rat}}^4a_n\,,\label{eq:SHr-mass-sum}\\
 \sum_{t\ge1}t^{-1/2}\min\{C_{\mathrm{rat}}^2,C_{\mathrm{rat}}^6a_\rho^2/t^2\}
 &\le C_{\mathrm{rat}}^2\int_0^{x_\rho}x^{-1/2}dx
        +C_{\mathrm{rat}}^6a_\rho^2\int_{x_\rho}^\infty x^{-5/2}dx
       \label{eq:D-step-0025}\\
 &=2C_{\mathrm{rat}}^2\sqrt{x_\rho}+\frac23C_{\mathrm{rat}}^6a_\rho^2x_\rho^{-3/2}
   =\frac83C_{\mathrm{rat}}^3\sqrt{a_\rho}\,.
   \label{eq:SHr-rho-sum}
\end{align}
Equation \eqref{eq:D-step-0024} bounds the decreasing summand by its integral and splits at $x_n=C_{\mathrm{rat}}^4a_n^2$.
Equation \eqref{eq:SHr-mass-sum} evaluates both integrals and substitutes $\sqrt{x_n}=C_{\mathrm{rat}}^2a_n$.
Equation \eqref{eq:D-step-0025} splits the analogous decreasing integral at $x_\rho=C_{\mathrm{rat}}^2a_\rho$.

Equation \eqref{eq:SHr-rho-sum} evaluates the $x^{-1/2}$ and $x^{-5/2}$ integrals and substitutes $\sqrt{x_\rho}=C_{\mathrm{rat}}\sqrt{a_\rho}$.
\end{proof}

\subsection{Proof of Direct Corruption Term}\label{app:D-corruption}
\begin{lemma}[Direct corruption contribution]\label{lem:D-corruption-term}
For $\kappa_t=\|c_t^{\rm vec}\|_\cA$ with
$0\le\kappa_t\le2$ and $\sum_t\kappa_t\le C$ almost surely,
\[
 \mathsf I_4\coloneq \sum_{t=1}^T\E[\langle p_t-q_t,x^\top c_t^{\rm vec}\rangle]
 \le16\sqrt2d\sqrt C\,.
\]
\end{lemma}
\begin{proof}
\interdisplaylinepenalty=10000
Marginal domination makes
$\widetilde p_t=[p_t-(1-\Gamma_t)q_t]/\Gamma_t$ a probability measure.
Start from the defining signed expectation:
\begin{align}
 \mathsf I_4
 &=\sum_{t=1}^T\E[\langle p_t-q_t,x^\top c_t^{\rm vec}\rangle]
 \le\E[\sum_{t=1}^T|\langle p_t-q_t,x^\top c_t^{\rm vec}\rangle|]\,.
 \label{eq:D-corruption-definition}
\end{align}
Equation \eqref{eq:D-corruption-definition} uses the triangle inequality.

The pathwise budget and \eqref{eq:cor-weighted} in \Cref{lem:capped} give
\begin{align}
\sum_{t=1}^T|\langle p_t-q_t,x^\top c_t^{\rm vec}\rangle|
 &=\sum_{t=1}^T\Gamma_t
       |\langle\widetilde p_t-q_t,x^\top c_t^{\rm vec}\rangle|
       \label{eq:D-step-0037}\\
 &\le\sum_{t=1}^T2\Gamma_t\kappa_t
 \le8d\sum_{t=1}^T\frac{\kappa_t}{\sqrt t}
 \le16\sqrt2d\sqrt C\,.
 \label{eq:SCr-direct-corruption}
\end{align}
Equation \eqref{eq:D-step-0037} substitutes $p_t-q_t=\Gamma_t(\widetilde p_t-q_t)$ before taking absolute values.
Equation \eqref{eq:SCr-direct-corruption} bounds the range of $x^\top c_t^{\rm vec}$ by $2\kappa_t$, substitutes $\Gamma_t\le4d/\sqrt t$, and applies \eqref{eq:cor-weighted} in \Cref{lem:capped}.

Taking expectation of \eqref{eq:SCr-direct-corruption} and returning to
\eqref{eq:D-corruption-definition} proves
\begin{equation}\label{eq:D-corruption-final}
 \mathsf I_4\le16\sqrt2d\sqrt C\,.
\end{equation}
The estimate is pathwise before expectation, so predictable adaptive
corruption requires no independence assumption.
\end{proof}

\subsection{Auxiliary Lemmas}\label{app:sharp-inputs}
These results supply feasibility, the all-history adversarial bound,
and the moment and reference-control inputs used by the four term proofs.
The notation $X_t(a)=\widehat L_{t-1,a}-\widehat L_{t-1,a^*}$ denotes the estimated
cumulative loss difference between $a$ and $a^*$;
the latent sampled action $X_t$ without an argument is a different object.
All windows, buffers, and corruption clocks are used only in the analysis.

\subsubsection{All-History Feasibility}
\begin{lemma}[All-history feasibility and the universal loss-estimate bounds]\label{lem:SH-feasible}
At every finite history, the sampling-law program
\eqref{eq:SH-program}, with $v_x(p)=M(p)^{-1}x$ prescribed by $p$,
has at least one minimizing $p$.
A fixed rule can choose such a law Borel measurably as a function
of the history, with deterministic finite-time bounds on
the reconstruction vectors $v_{X_t}(p_t)$ and cumulative parameter estimates $\sum_{s=1}^{t}\widehat\theta_s$.
Use this rule in \GIAFTRL{}.
Then $\E[\widehat\theta_t\mid\cF_{t-1}]=\theta_t$.
Its loss differences are conditionally unbiased and have deterministic
finite-time bounds on their range.
In the stochastic case $\theta_t=\theta$, fix the optimal action $a^*$ and set
$X_t(u)=\widehat L_{t-1,u}-\widehat L_{t-1,a^*}$,
$M_t=\max_{u\in\mathcal K}(-X_t(u))\ge0$, and
$D_t(u)=(u-a^*)^\top v_{X_t}(p_t)Y_t$. On \emph{every} history, including those
with $b_{r_t}\ne a^*$,
\begin{align}
 X_{t+1}(u)&=X_t(u)+D_t(u)\,,&\E_{t-1}[D_t(u)]&=\Delta(u)\,,\label{eq:SH-drift}\\
 |D_t(u)|&\le2\beta_t+(X_t(u)+2M_t)/B\,,\label{eq:SH-global-amp}\\
 \E_{t-1}[D_t(u)^2]&\le8\beta_t+2(X_t(u)+2M_t)^2/(Bt)\,.
 \label{eq:SH-global-var}
\end{align}
\end{lemma}
\begin{proof}
\textbf{One feasible law for all constraints.}
Choose the physical G-optimal law $\pi$ from
\Cref{lem:gopt}, and set
$p^0=(1-\Gamma_t)q_t+\Gamma_t\pi$,
$M=\Sigma(p^0)$, and $v(x)=M^{-1}x$.
The norm $\|\cdot\|_{\Sigma(\pi)^{-1}}$ is convex and even, so
G-optimality extends from physical actions to their symmetric convex
hull. The extension of the G-optimal design's leverage bound to the whole
symmetric hull is explicit. For $r>\|w\|_{\mathcal S}$ choose
$w/r=\sum_i\lambda_i\epsilon_i a_i$ with $\lambda_i\ge0$,
$\sum_i\lambda_i=1$, and $\epsilon_i\in\{-1,1\}$. Then
\begin{align}
\|w\|_{\Sigma(\pi)^{-1}}
 &=r\left\|\sum_i\lambda_i\epsilon_i a_i\right\|_{\Sigma(\pi)^{-1}}
 \le r\sum_i\lambda_i\|a_i\|_{\Sigma(\pi)^{-1}}
 \le r\sum_i\lambda_i\sqrt d=r\sqrt d\,,
       \label{eq:D-fallback-hull-leverage}\\
 \|w\|_{\Sigma(\pi)^{-1}}
 &\le\inf_{r>\|w\|_{\mathcal S}}r\sqrt d
 =\sqrt d\,\|w\|_{\mathcal S}\,.
       \label{eq:D-fallback-gauge-leverage}
\end{align}
Equation \eqref{eq:D-fallback-hull-leverage} uses the triangle inequality
for the covariance norm and the physical G-optimal leverage guarantee
of \Cref{lem:gopt}. Equation
\eqref{eq:D-fallback-gauge-leverage} takes the infimum over the gauge
representations. Thus the fallback law controls \emph{every} internal
direction, not only physical actions.
\begin{align}
M&\succeq\Gamma_t\Sigma(\pi)\succ0\,,
       \label{eq:D-step-0050}\\
 \|w\|_{\Sigma(\pi)^{-1}}&\le\sqrt d\,\|w\|_{\mathcal S}\,,
       \label{eq:D-step-0051}\\
 w^\top M^{-1}w
 &\le\Gamma_t^{-1}w^\top\Sigma(\pi)^{-1}w
 \le\frac d{\Gamma_t}\|w\|_{\mathcal S}^2
 =\frac{\beta_t}{4}\|w\|_{\mathcal S}^2\,,
       \label{eq:D-step-0052}\\
 x^\top M^{-1}x&\le\beta_t/4\,,\label{eq:SHr-fallback-leverage}\\
 |(u-b_{r_t})^\top v(x)|
 &=|(u-b_{r_t})^\top M^{-1}x|
       \label{eq:D-step-0053}\\
 &\le\sqrt{(u-b_{r_t})^\top M^{-1}(u-b_{r_t})}\sqrt{x^\top M^{-1}x}
 \le\frac{\beta_t}{4}\|u-b_{r_t}\|_{\mathcal S}\le U_t(u)\,,
 \label{eq:SHr-fallback-amplitude}\\
 \int v(x)x^\top p^0(dx)
 &=M^{-1}M=I_d\,,
       \label{eq:D-step-0054}\\
 \int[(u-b_{r_t})^\top v(x)]^2p^0(dx)
 &=(u-b_{r_t})^\top M^{-1}MM^{-1}(u-b_{r_t})
       \label{eq:D-step-0055}\\
 &=(u-b_{r_t})^\top M^{-1}(u-b_{r_t})
 \le\frac{\beta_t}{4}\|u-b_{r_t}\|_{\mathcal S}^2\le W_t(u)\,.
 \label{eq:SHr-fallback-reconstruction}
\end{align}
Equation \eqref{eq:D-step-0050} keeps the exploration covariance in $M=(1-\Gamma_t)\Sigma(q_t)+\Gamma_t\Sigma(\pi)$.
Equation \eqref{eq:D-step-0051} is the symmetric-hull extension \eqref{eq:D-fallback-gauge-leverage}.
Equation \eqref{eq:D-step-0052} uses monotonicity of matrix inversion, the gauge leverage bound, and $d/\Gamma_t=\beta_t/4$.

Equation \eqref{eq:SHr-fallback-leverage} applies the preceding bound to $x\in\mathcal K\subseteq\mathcal S$.
Equation \eqref{eq:D-step-0053} substitutes $v(x)=M^{-1}x$.
Equation \eqref{eq:SHr-fallback-amplitude} applies covariance Cauchy--Schwarz, substitutes both leverage bounds, and compares with the upper bound $U_t(u)$.

Equation \eqref{eq:D-step-0054} substitutes $v(x)=M^{-1}x$ and evaluates $\int xx^\top p^0(dx)=M$.
Equation \eqref{eq:D-step-0055} expands the squared linear coefficient and evaluates the same covariance integral.
Equation \eqref{eq:SHr-fallback-reconstruction} cancels $M^{-1}MM^{-1}$ and bounds it by the first summand of $W_t(u)$.
For $z(u,x)=(u-\bar x_t)^\top M^{-1}x$, the same estimates give
\begin{align}
|z(u,x)|
 &=|(u-\bar x_t)^\top M^{-1}x|
 \le\sqrt{(u-\bar x_t)^\top M^{-1}(u-\bar x_t)}
                         \sqrt{x^\top M^{-1}x}\label{eq:D-fallback-z-cs}\\
 &\le\frac{\beta_t}{4}\|u-\bar x_t\|_{\mathcal S}
 \le\frac{\beta_t}{4}(\|u\|_{\mathcal S}+\|\bar x_t\|_{\mathcal S})
 \le\frac{\beta_t}{2}\,,\label{eq:D-fallback-z-range}\\
 \iint z(u,x)^2q_t(du)p^0(dx)
 &=\int (u-\bar x_t)^\top M^{-1}
          \left[\int xx^\top p^0(dx)\right]M^{-1}(u-\bar x_t)q_t(du)
       \label{eq:D-fallback-z-second-substitution}\\
 &=\int(u-\bar x_t)^\top M^{-1}(u-\bar x_t)q_t(du)
       \label{eq:D-fallback-z-second-cancel}\\
 &=\Tr\!\left(M^{-1}
       \left[\int (u-\bar x_t)(u-\bar x_t)^\top q_t(du)\right]\right)
       \label{eq:D-fallback-z-trace}\\
 &=\Tr[M^{-1}(\Sigma(q_t)-\bar x_t\bar x_t^\top)]
 \le\Tr[M^{-1}\Sigma(q_t)]\label{eq:D-fallback-covariance-trace}\\
 &\le\Tr[(1-\Gamma_t)^{-1}I_d]
 =\frac d{1-\Gamma_t}\le\frac{4d}{3}\,,\label{eq:D-fallback-trace-bound}\\
 \gC_t(p^0)
 &=\beta_t\iint\chi(z(u,x)/\beta_t)q_t(du)p^0(dx)
       \label{eq:D-fallback-certificate-substitute}\\
 &\le\beta_t\iint\frac{3z(u,x)^2}{2\beta_t^2}q_t(du)p^0(dx)
       \label{eq:D-fallback-chi-small}\\
 &=\frac{3}{2\beta_t}\iint z(u,x)^2q_t(du)p^0(dx)
 \le\frac{3}{2\beta_t}\frac{4d}{3}=\frac{2d}{\beta_t}\,.
       \label{eq:SHr-fallback-certificate}
\end{align}
The scalar $z(u,x)$ is the centered reconstruction coefficient:
when the latent action is $x$, multiplying its feedback by $z(u,x)$
estimates the loss of $u$ relative to the FTRL barycenter.
Equation \eqref{eq:D-fallback-z-cs} is Cauchy--Schwarz in the
$M^{-1}$ inner product; \eqref{eq:D-fallback-z-range} uses the leverage
bounds and $u,\bar x_t\in\mathcal K\subseteq\mathcal S$.
Equation \eqref{eq:D-fallback-z-second-substitution} substitutes
$z(u,x)$ before integration, and \eqref{eq:D-fallback-z-second-cancel}
uses $\int xx^\top p^0(dx)=M$.
Equations \eqref{eq:D-fallback-z-trace} and
\eqref{eq:D-fallback-covariance-trace} expand the centered covariance;
$\bar x_t^\top M^{-1}\bar x_t\ge0$ gives the inequality.
For \eqref{eq:D-fallback-trace-bound}, congruence of
$M\succeq(1-\Gamma_t)\Sigma(q_t)$ by $M^{-1/2}$ gives the matrix
bound before taking the trace, and $\Gamma_t<1/4$ gives $4d/3$.
Equation \eqref{eq:D-fallback-certificate-substitute} is the proxy function
definition with reconstruction vector $v_X(p^0)=M^{-1}X$. Equation
\eqref{eq:D-fallback-chi-small} uses the small-argument inequality in
\Cref{lem:chi}, since $|z/\beta_t|\le1/2$.
Equation \eqref{eq:SHr-fallback-certificate} substitutes the integrated
quadratic estimate \eqref{eq:D-fallback-trace-bound}.
Thus $p^0$ with its prescribed covariance map $x\mapsto v_x(p^0)$
satisfies every constraint of \eqref{eq:SH-program}.

\paragraph{Existence of a Minimizer, Measurability, and Integrability}
Fix a finite history. After eliminating the auxiliary mixture variables,
the sampling laws lie in
\[
 \mathcal P_t=\{p\in\gP(\mathcal K):p\ge(1-\Gamma_t)q_t\}\,.
\]
The set $\gP(\mathcal K)$ is compact metrizable in the weak topology,
and domination is closed: equivalently,
\begin{equation}\label{eq:SHr-domination-tests}
 \int\varphi\,dp\ge(1-\Gamma_t)\int\varphi\,dq_t
 \qquad(\varphi\in C(\mathcal K),\ \varphi\ge0)\,.
\end{equation}
Thus $\mathcal P_t$ is compact. Since the prior has support
$\mathcal K$ and this set spans $\R^d$, the positive Gibbs density gives
\[
 M(p)\succeq(1-\Gamma_t)M(q_t)\succ0
 \qquad(p\in\mathcal P_t)\,.
\]
On $\mathcal P_t$, the map $p\mapsto M(p)^{-1}$ is continuous and
uniformly bounded. Hence $(p,x)\mapsto v_x(p)=M(p)^{-1}x$ is continuous
and uniformly bounded.
This uses the full linear span, and remains valid when
$\operatorname{aff}(\mathcal K)$ is a proper affine subspace.

The worst-case first absolute moment in \eqref{eq:SH-program-mag}
is attained by the constant Dirac feedback kernel $Q_x=\delta_1$.
Thus the magnitude constraint is equivalent to the closed family
\[
 |d_t(u;v_x(p),1)|\le U_t(u)
                    \qquad(u,x\in\mathcal K)\,.
\]
Each $p\in\mathcal P_t$ has support $\mathcal K$, so requiring this
family only $p$-almost surely gives the same condition.
Indeed, a strict violation at any $x$ persists on a relatively open
neighborhood by continuity and would have positive $p$-mass.
Continuity in $u$ also permits a countable dense set of test actions.
The second-moment and exponential constraints are continuous in $p$:
the coefficients converge uniformly on the compact action sets when
$p$ converges weakly in $\mathcal P_t$, and their integrands remain
uniformly bounded. The covariance formula implies
$\int v_x(p)x^\top p(dx)=I_d$. Thus the estimator budgets define a
closed subset $\mathcal F_t\subseteq\mathcal P_t$. It is nonempty
because the fallback above uses $p^0$ with exactly $x\mapsto v_x(p^0)$.
Finally, $x\mapsto\widehat\Delta_t(x)$ is continuous and bounded, so
$p\mapsto\int\widehat\Delta_t\,dp$ is continuous. Its minimum on
$\mathcal F_t$ is attained.

For measurability, the reference index is Borel in the cumulative
parameter estimate. On bounded regions of that vector, the positive
Gibbs density gives a common lower eigenvalue bound for $M(p)$ over
all dominated candidates. Countable dense families of $u,x\in\mathcal K$
and of nonnegative continuous test functions in
\eqref{eq:SHr-domination-tests} suffice to specify the constraints.
All these functions are Borel in the state and continuous in the
candidate law, so the feasible graph is Borel with compact sections.
Its strict objective sublevel sections are relatively open subsets of
compact metric spaces and hence $\sigma$-compact. The projection and
compact-section selection theorem \cite{Larman72} makes the value function
Borel and supplies a Borel minimizing $p_t$. Fix one such selection
on a countable partition into bounded state regions. The formulas
$v_x(p_t)=M(p_t)^{-1}x$, $\gamma_t=\Gamma_t$, and
$w_t=p_t-(1-\Gamma_t)q_t$ give Borel derived quantities; no selection
of reconstruction vectors or joint laws is needed. Starting from
the empty cumulative sum, the sampling and update kernels are therefore Borel in the
observed history by induction.
This argument does not require convexity of the feasible set.

For finite-time integrability, let
$\lambda_\mu=\lambda_{\min}(M(\mu))>0$ and $R=\|\sum_{s=1}^{t-1}\widehat\theta_s\|$.
Since $\mathcal K\subseteq\BB$, the Gibbs density satisfies
$dq_t/d\mu\ge e^{-2R/\beta_t}$. Uniformly over dominated sampling laws,
\[
 M(p)\succeq\tfrac34e^{-2R/\beta_t}M(\mu)\,,\qquad
 \sup_{x\in\mathcal K}\|v_x(p)\|
 \le\frac{4}{3\lambda_\mu}e^{2R/\beta_t}\,.
\]
Consequently, using $|Y_t|\le1$,
\begin{equation}\label{eq:SHr-finite-integrability}
 \left\|\sum_{s=1}^{t}\widehat\theta_s\right\|
 \le\left\|\sum_{s=1}^{t-1}\widehat\theta_s\right\|+
       \frac{4}{3\lambda_\mu}e^{2\|\sum_{s=1}^{t-1}\widehat\theta_s\|/\beta_t}\,.
\end{equation}
Induction gives deterministic finite bounds at each fixed horizon for
$\sum_{s=1}^{t}\widehat\theta_s$ and $v_{X_t}(p_t)$. These bounds ensure integrability and conditional
unbiasedness of the ambient covariance estimates; the regret estimates
use the original difference constraints.

\paragraph{A Range Bound for the Cumulative Estimates}
Define the range of the algorithm's cumulative estimates by
\[
 \mathsf{osc}_t=\max_{x,y\in\mathcal K}|\widehat L_{t,x}-\widehat L_{t,y}|\,,
 \qquad \mathsf{osc}_0=0\,.
\]
At round $t$, $Z_t(x)\le\mathsf{osc}_{t-1}$ and
$\|x-b_{r_t}\|_{\mathcal S}\le2$ for every $x\in\mathcal K$.
The magnitude constraint and $|Y_t|\le1$ therefore give, simultaneously
for all $x,y\in\mathcal K$,
\begin{align}
 |(x-y)^\top v_{X_t}(p_t)Y_t|
 &\le U_t(x)+U_t(y)
 \le2\beta_t+2\mathsf{osc}_{t-1}/B\,,\notag\\
 \mathsf{osc}_t
 &\le(1+2/B)\mathsf{osc}_{t-1}+2\beta_t
 \le2\sum_{s=1}^t(1+2/B)^{t-s}\beta_s<\infty\,.
 \label{eq:SH-rule-range}
\end{align}
This bound is deterministic and holds on every history.
It bounds the raw fields $\widehat L_{t,x}-\widehat L_{t,a^*}$ and all their finite-time
moments; \eqref{eq:SH-initial-extension} gives the same conclusion for
the normalized fields. Stopping at a finite deterministic horizon
preserves these bounds. The finite first moment required at each
history permits the reconstruction identity to be integrated against
the conditional feedback law. For unconditional expectations, we use
the bounded loss differences directly.

Indeed, for a fixed $x_0\in\mathcal K$, subtracting
$\widehat L_{t-1,x_0}$ from every cumulative action estimate
$\widehat L_{t-1,x}$ available before round $t$ leaves $q_t$, the
penalized reference, and all program constraints unchanged. Likewise
$\widehat f_t(x)$ may be replaced by
$\widehat\ell_{t,x}-\widehat\ell_{t,x_0}$ in every zero-mass regret pairing and in
the stability residual $\mathsf D_t$. These quantities are bounded at
each finite time by \eqref{eq:SH-rule-range}. Thus the FTRL and moment
arguments require no unconditional moment assumption on a common
action-independent component of $\sum_{s=1}^{t}\widehat\theta_s$ or $v_{X_t}(p_t)$.

\paragraph{Unbiasedness and Bounds Relative to the Optimal Action}
The rounding rule has $\E[A_t\mid\cF_{t-1},X_t,v_{X_t}(p_t)]=X_t$. Thus
\begin{align}
\E[Y_t\mid\cF_{t-1},X_t,v_{X_t}(p_t)]
 &=\E\!\left[\E[Y_t\mid\cF_{t-1},X_t,v_{X_t}(p_t),A_t]
                           \mid\cF_{t-1},X_t,v_{X_t}(p_t)\right]\label{eq:D-feedback-tower}\\
 &=\E[A_t^\top\theta_t\mid\cF_{t-1},X_t,v_{X_t}(p_t)]
 =X_t^\top\theta_t\,,\label{eq:D-feedback-rounded-mean}\\
 \E_{t-1}[v_{X_t}(p_t)Y_t]
 &=\int v_x(p_t)\,\E[Y_t\mid\cF_{t-1},X_t=x]p_t(dx)\label{eq:D-feedback-mark-conditioning}\\
 &=\int v_x(p_t)x^\top\theta_t p_t(dx)
 =\left[\int v_x(p_t)x^\top p_t(dx)\right]\theta_t
 =\theta_t\,,\label{eq:D-feedback-reconstruction}\\
 \E_{t-1}[D_t(u)]
 &=(u-a^*)^\top\E_{t-1}[v_{X_t}(p_t)Y_t]
 =(u-a^*)^\top\theta_t\,.
       \label{eq:SHr-exact-unbiasedness}
\end{align}
Equation \eqref{eq:D-feedback-tower} is iterated conditional expectation.
Equation \eqref{eq:D-feedback-rounded-mean} inserts the feedback model
and then the rounding identity $\E[A_t\mid\cF_{t-1},X_t,v_{X_t}(p_t)]=X_t$.
Equation \eqref{eq:D-feedback-mark-conditioning} conditions on the
virtual action drawn from the predictable law $p_t$; its coefficient
is $v_x(p_t)$. Equation \eqref{eq:D-feedback-reconstruction} uses
$\int v_x(p_t)x^\top p_t(dx)=M(p_t)^{-1}M(p_t)=I_d$, a consequence
of $M(p)=\int_{\mathcal K}xx^\top p(dx)$ and $v_x(p)=M(p)^{-1}x$. Equation \eqref{eq:SHr-exact-unbiasedness} applies the
linear functional $u-a^*$ to the full unbiased estimate.

The second moment and samplewise range of the actual reference-relative
estimate $g_t(u)\coloneq d_t(u;v_{X_t}(p_t),Y_t)=(u-b_{r_t})^\top v_{X_t}(p_t)Y_t$ follow from the same conditioning:
\begin{align}
\E_{t-1}[g_t(u)^2]
 &=\int\E\!\left[\left.((u-b_{r_t})^\top v_x(p_t)Y_t)^2
                  \right|\cF_{t-1},X_t=x\right]p_t(dx)\label{eq:D-g-second-condition}\\
 &=\int[(u-b_{r_t})^\top v_x(p_t)]^2
          \E[Y_t^2\mid\cF_{t-1},X_t=x]p_t(dx)\label{eq:D-g-second-factor}\\
 &\le\int[(u-b_{r_t})^\top v_x(p_t)]^2p_t(dx)
 \le W_t(u)\label{eq:D-g-second-constraint}\\
 &=\frac{\beta_t}{4}\|u-b_{r_t}\|_{\mathcal S}^2
                          +\frac{Z_t(u)^2}{Bt}\,,
       \label{eq:D-g-second-budget-substitute}\\
 |g_t(u)|&=|(u-b_{r_t})^\top v_{X_t}(p_t)|\,|Y_t|
 \le|(u-b_{r_t})^\top v_{X_t}(p_t)|\label{eq:D-g-amplitude-feedback}\\
 &\le U_t(u)
 =\frac{\beta_t}{2}\|u-b_{r_t}\|_{\mathcal S}+\frac{Z_t(u)}B\,.
       \label{eq:D-g-amplitude-budget}
\end{align}
Equation \eqref{eq:D-g-second-condition} substitutes the random
estimate before conditioning.  The coefficient is fixed under that
conditioning, which gives \eqref{eq:D-g-second-factor}.
Equation \eqref{eq:D-g-second-constraint} uses $Y_t^2\le1$ and the
quadratic constraint on the executed sampling law $p_t$.
Equation \eqref{eq:D-g-second-budget-substitute} substitutes the
definition of $W_t(u)$, with no intermediate relaxation.
Equation \eqref{eq:D-g-amplitude-feedback} uses $|Y_t|\le1$;
\eqref{eq:D-g-amplitude-budget} uses $v_{X_t}(p_t)\in\mathcal M_t$ and then
the definition of $U_t(u)$.
For $\theta_t=\theta$, the mean in
\eqref{eq:SHr-exact-unbiasedness} is
$(u-a^*)^\top\theta=\Delta(u)$.
Independently of the reference location,
\begin{align}
X_t(u)&\ge-M_t\,,\qquad X_t(b_{r_t})\ge-M_t\,,\label{eq:D-true-score-lower}\\
 Z_t(u)&=[X_t(u)-X_t(b_{r_t})]_+
 \le[X_t(u)+M_t]_+=X_t(u)+M_t\,,\label{eq:D-true-score-z}\\
 Z_t(a^*)&=[-X_t(b_{r_t})]_+\le M_t\,,
 \qquad Z_t(u)+Z_t(a^*)\le X_t(u)+2M_t\,,\label{eq:D-true-score-z-sum}\\
 D_t(u)&=[(u-b_{r_t})-(a^*-b_{r_t})]^\top v_{X_t}(p_t)Y_t=g_t(u)-g_t(a^*)\,,\label{eq:D-true-estimate-decomposition}\\
 |D_t(u)|&\le |g_t(u)|+|g_t(a^*)|\le U_t(u)+U_t(a^*)\label{eq:D-true-amplitude-triangle}\\
 &=\frac{\beta_t}{2}
       [\|u-b_{r_t}\|_{\mathcal S}+\|a^*-b_{r_t}\|_{\mathcal S}]
             +\frac{Z_t(u)+Z_t(a^*)}{B}\label{eq:D-true-amplitude-substitute}\\
 &\le\frac{\beta_t}{2}(2+2)+\frac{X_t(u)+2M_t}{B}
 =2\beta_t+\frac{X_t(u)+2M_t}{B}\,,\label{eq:SHr-true-reference-amplitude}\\
 \|D_t(u)\|_{L^2(\cF_{t-1})}
 &\le\|g_t(u)\|_{L^2(\cF_{t-1})}
       +\|g_t(a^*)\|_{L^2(\cF_{t-1})}\label{eq:D-true-variance-triangle}\\
 &\le\sqrt{W_t(u)}+\sqrt{W_t(a^*)}\label{eq:D-true-variance-constraints}\\
 &\le\frac{\sqrt{\beta_t}}2
       [\|u-b_{r_t}\|_{\mathcal S}+\|a^*-b_{r_t}\|_{\mathcal S}]
             +\frac{Z_t(u)+Z_t(a^*)}{\sqrt{Bt}}\label{eq:D-true-variance-substitute}\\
 &\le2\sqrt{\beta_t}+\frac{X_t(u)+2M_t}{\sqrt{Bt}}\,,\label{eq:D-true-variance-root-bound}\\
 \E_{t-1}[D_t(u)^2]
 &\le\left(2\sqrt{\beta_t}+\frac{X_t(u)+2M_t}{\sqrt{Bt}}\right)^2
 \le2(2\sqrt{\beta_t})^2+2\left(\frac{X_t(u)+2M_t}{\sqrt{Bt}}\right)^2\label{eq:D-true-variance-square}\\
 &=8\beta_t+\frac{2[X_t(u)+2M_t]^2}{Bt}\,.
       \label{eq:SHr-true-reference-variance}
\end{align}
Equation \eqref{eq:D-true-score-lower} is the definition of $M_t$.
Equations \eqref{eq:D-true-score-z} and \eqref{eq:D-true-score-z-sum}
substitute the estimated cumulative loss differences relative to the reference and use $X_t(u)+M_t\ge0$.
Equation \eqref{eq:D-true-estimate-decomposition} subtracts the two
reference-relative estimators.  Equation \eqref{eq:D-true-amplitude-triangle}
uses their respective upper bounds $U_t(u)$ and $U_t(a^*)$; \eqref{eq:D-true-amplitude-substitute}
substitutes both budgets before bounding the two norms by two in
\eqref{eq:SHr-true-reference-amplitude}.
Equation \eqref{eq:D-true-variance-triangle} is conditional Minkowski;
\eqref{eq:D-true-variance-constraints} uses the two separate quadratic
constraints. Equation \eqref{eq:D-true-variance-substitute} bounds each
budget's square root using $\sqrt{a^2+b^2}\le a+b$ for $a,b\ge0$.
Equation \eqref{eq:D-true-variance-root-bound} inserts the norm and
cumulative loss estimate bounds.  Equation \eqref{eq:D-true-variance-square} squares the
nonnegative upper bound and uses $(x+y)^2\le2x^2+2y^2$;
\eqref{eq:SHr-true-reference-variance} evaluates the two squares.
Together with $X_{t+1}(u)=X_t(u)+D_t(u)$, these are exactly
\eqref{eq:SH-drift}--\eqref{eq:SH-global-var}.
\end{proof}

\subsubsection{Reference Atoms and Adversarial Comparator Cost}\label{sec:SH-prior}

Sharpness here bounds the loss gap $\Delta(a)$ below by a positive constant
times the distance $\|a-a^*\|_{\mathcal S}$ from $a$ to $a^*$. It does
not impose a positive minimum loss gap. Norm equivalence makes
\eqref{eq:SH-sharp} in \Cref{lem:SH-atom} equivalent to the Euclidean sharpness in
\eqref{eq:C-sharp}. By \Cref{prop:C-centered}, it is also equivalent to
finite centered information-acquisition complexity. Sharp instances with zero optimal mean are
included.

\begin{lemma}[Adversarial comparator cost of the convolution reference]\label{lem:SH-prior-comparator}
For each $u\in\mathcal K$ and integer $T\ge1$ there is a probability
$q_{u,T}$ such that
\[
 \KL(q_{u,T},\mu)\le\log2+mk\log T\,,
 \qquad
 \left|\left(\int xq_{u,T}(dx)-u\right)^\top\vartheta\right|\le2/T
\]
for every $\norm\vartheta_\cA\le1$.
\end{lemma}
\begin{proof}[Proof of \Cref{lem:SH-prior-comparator}]
Set $\varepsilon=T^{-1}$ and
$\mathcal K_\varepsilon=u+\varepsilon(\mathcal K-u)$.
Let $\lambda_\varepsilon$ be relative volume conditioned on
$\mathcal K_\varepsilon$, let
$\mathsf A(x_1,\ldots,x_{m})=m^{-1}\sum_i x_i$, and define
$q_{u,T}=\mathsf A_\#(\lambda_\varepsilon^{\otimes m})$.
The direct volume and entropy calculations are
The smoothing law places mass within gauge distance $2/T$ of the
comparator without requiring a prior atom at that comparator.
Let $\Phi(x_1,\ldots,x_{m})=m^{-1}\sum_i x_i$. The density
$r=d\lambda_\varepsilon^{\otimes m}/d\lambda_{\mathcal K}^{\otimes m}$
and its pushforward density satisfy
\begin{align}
\frac{dq_{u,T}}{d\mu_m}\bigl(\Phi\bigr)
 &=\E_{\lambda_{\mathcal K}^{\otimes m}}[r\mid\Phi]\,,
       \label{eq:D-comparator-pushforward-density}\\
 \KL(q_{u,T},\mu_m)
 &=\E\!\left[\E[r\mid\Phi]\log\E[r\mid\Phi]\right]
 \le\E[r\log r]
 =\KL(\lambda_\varepsilon^{\otimes m},
                    \lambda_{\mathcal K}^{\otimes m})\,.
       \label{eq:D-comparator-pushforward-entropy}
\end{align}
Equation \eqref{eq:D-comparator-pushforward-density} is the density
identity for a measurable pushforward. Equation
\eqref{eq:D-comparator-pushforward-entropy} writes both entropies under
the same reference product law and applies conditional Jensen to
$x\mapsto x\log x$. The remaining entropy arithmetic is
\begin{align}
\lambda_{\mathcal K}(\mathcal K_\varepsilon)&=\varepsilon^k\,,
 &\frac{d\lambda_\varepsilon}{d\lambda_{\mathcal K}}
   &=\varepsilon^{-k}\ind_{\mathcal K_\varepsilon}\,,
       \label{eq:D-step-0059}\\
 \KL(\lambda_\varepsilon,\lambda_{\mathcal K})
 &=\int_{\mathcal K_\varepsilon}\log(\varepsilon^{-k})
                      \lambda_\varepsilon(dx)
 =k\log(1/\varepsilon)\,,
       \label{eq:D-step-0060}\\
 \KL(\lambda_\varepsilon^{\otimes m},
           \lambda_{\mathcal K}^{\otimes m})
 &=\sum_{i=1}^{m}\KL(\lambda_\varepsilon,\lambda_{\mathcal K})
 =mk\log T\,,
       \label{eq:D-step-0061}\\
 \KL(q_{u,T},\mu_m)&\le mk\log T\,.
 \label{eq:SHr-comparator-product-entropy}
\end{align}
Equation \eqref{eq:D-step-0059} uses the $k$-dimensional volume scaling of $\mathcal K_\varepsilon=u+\varepsilon(\mathcal K-u)$.
Equation \eqref{eq:D-step-0060} substitutes the constant Radon--Nikodym density in the entropy integral.
Equation \eqref{eq:D-step-0061} uses additivity of product relative entropy and $\varepsilon=T^{-1}$.

Equation \eqref{eq:SHr-comparator-product-entropy} applies the pushforward entropy inequality \eqref{eq:D-comparator-pushforward-entropy}.
Since $\mu\ge\omega_\mu\mu_m$ with $\omega_\mu>1/2$,
\begin{align}
\KL(q_{u,T},\mu)
 &=\KL(q_{u,T},\mu_m)
       +\int\log\frac{d\mu_m}{d\mu}\,dq_{u,T}
       \label{eq:D-step-0062}\\
 &\le mk\log T+\log(1/\omega_\mu)
 \le mk\log T+\log2\,.
 \label{eq:SHr-comparator-mixture-entropy}
\end{align}
Equation \eqref{eq:D-step-0062} factors the Radon--Nikodym derivative through $\mu_m$ and integrates its logarithm.
Equation \eqref{eq:SHr-comparator-mixture-entropy} uses $\mu\ge\omega_\mu\mu_m$, the product entropy bound, and $\omega_\mu\ge1/2$.
Let $\bar U=\int x\lambda_{\mathcal K}(dx)$. The affine
change of variables defining $\mathcal K_\varepsilon$ gives
\begin{align}
\int xq_{u,T}(dx)
 &=\frac1{m}\sum_{i=1}^{m}
       \int x\lambda_\varepsilon(dx)
 =u+\varepsilon(\bar U-u)\,,
       \label{eq:D-step-0063}\\
 \left|\left(\int xq_{u,T}(dx)-u\right)^\top\vartheta\right|
 &=\varepsilon|(\bar U-u)^\top\vartheta|
 \le\varepsilon(|\bar U^\top\vartheta|+|u^\top\vartheta|)
 \le2\varepsilon=2/T\,.
 \label{eq:SHr-comparator-mean}
\end{align}
Equation \eqref{eq:D-step-0063} uses linearity of the averaging map and the mean of $u+\varepsilon(U-u)$.
Equation \eqref{eq:SHr-comparator-mean} substitutes that mean, applies the triangle inequality to the loss, and uses bounded losses on $\mathcal K$.
Both $\bar U$ and $u$ lie in $\mathcal K$, whose linear losses belong
to $[-1,1]$ whenever $\|\vartheta\|_\cA\le1$.
\end{proof}

\begin{lemma}[Every sharp optimal action is a reference atom]\label{lem:SH-atom}
Suppose a parameter has a unique optimal action $a^*$ and
\begin{equation}\label{eq:SH-sharp}
 \varsigma=\inf_{a\in\mathcal K\setminus\{a^*\}}
 \frac{(a-a^*)^\top\theta}{\norm{a-a^*}_{\mathcal S}}>0\,.
\end{equation}
Then $a^*=b_{j^*}$ for some finite index $j^*$.
\end{lemma}
\begin{proof}[Proof of \Cref{lem:SH-atom}]
Let $\|\cdot\|_{\mathcal S,*}$ be the norm dual to
$\|\cdot\|_{\mathcal S}$. For every
$\|\theta'-\theta\|_{\mathcal S,*}<\varsigma/2$ and every
$a\ne a^*$ in $\mathcal K$, sharpness implies
\begin{align}
(a-a^*)^\top\theta'
 &=(a-a^*)^\top\theta+(a-a^*)^\top(\theta'-\theta)
       \label{eq:D-step-0064}\\
 &\ge\varsigma\|a-a^*\|_{\mathcal S}
       -\|a-a^*\|_{\mathcal S}\|\theta'-\theta\|_{\mathcal S,*}
       \label{eq:D-step-0065}\\
 &>\frac{\varsigma}{2}\|a-a^*\|_{\mathcal S}>0\,.
 \label{eq:SHr-rational-exposure}
\end{align}
Equation \eqref{eq:D-step-0064} adds and subtracts the true parameter in the linear gap.
Equation \eqref{eq:D-step-0065} applies sharpness to the true gap and gauge duality to the perturbation.
Equation \eqref{eq:SHr-rational-exposure} uses $\|\theta'-\theta\|_{\mathcal S,*}<\varsigma/2$ and $a\ne a^*$.
The dual-norm ball is an open subset of $\R^d$ and therefore contains
some $q\in\mathbb Q^d$. By \eqref{eq:SHr-rational-exposure},
\begin{align}
\argmin_{a\in\mathcal K}q^\top a
 &=\argmin_{a\in\cA}q^\top a=\{a^*\}\,.
 \label{eq:SHr-atom-selection}
\end{align}
Equation \eqref{eq:SHr-atom-selection} uses
$a^*\in\cA\subseteq\mathcal K$ and the strict positivity in
\eqref{eq:SHr-rational-exposure} for every other action. The fixed enumeration reaches
$q$ at a finite position, selects $a^*$ there, and retains either that
occurrence or an earlier identical one. Thus $a^*=b_{j^*}$ for a finite
$j^*\in J$.
\end{proof}
\subsubsection{All-History Adversarial Bound}
\begin{proposition}[All-history adversarial guarantee]\label{prop:D-adversarial}
\GIAFTRL{} obeys
$\Reg\le U_{\rm SH}(T)\le320(d+1)\sqrt T\log(eT)$
for every predictable bounded-mean environment.
\end{proposition}
\begin{proof}
Fix $u\in\mathcal K$, $T\ge1$, and the deterministic comparison law
$q_{u,T}$ of \Cref{lem:SH-prior-comparator}. Let
$\widehat f_t(x)=\widehat\ell_{t,x}$, $S_t(x)=\widehat L_{t,x}$, and
$\langle q,f\rangle=\int f\,dq$. Thus $S_t$ includes round $t$,
$S_{t-1}(x)=\widehat L_{t-1,x}$ is the input to $q_t$, and
$S_t(x)-S_{t-1}(x)=\widehat\ell_{t,x}$. Define
\begin{align}
F_\beta(S)&=-\beta\log\int e^{-S(x)/\beta}\mu(dx)
       =\inf_q\{\langle q,S\rangle+\beta\KL(q,\mu)\}\,,
       \label{eq:SHr-potential}\\
 \mathsf D_t&=\langle q_t,\widehat f_t\rangle
       +\beta_t\log\langle q_t,e^{-\widehat f_t/\beta_t}\rangle\,,
       \qquad
 \widetilde p_t=\frac{p_t-(1-\Gamma_t)q_t}{\Gamma_t}\in\gP(\mathcal K)\,.
       \label{eq:SHr-stability-residual}
\end{align}
The equality in \eqref{eq:SHr-potential} is the variational identity in
\Cref{lem:entropy-FTRL}; marginal domination in
\eqref{eq:SH-program} in \GIAFTRL{} makes $\widetilde p_t$ a probability measure.
The full-history telescoping calculation, with no point-mass restriction
on the comparator, is
\begin{align}
F_{\beta_t}(S_t)-F_{\beta_t}(S_{t-1})
 &=-\beta_t\log\frac{\int e^{-[S_{t-1}(x)+\widehat f_t(x)]/\beta_t}\mu(dx)}
                         {\int e^{-S_{t-1}(x)/\beta_t}\mu(dx)}
       \label{eq:D-ftrl-potential-increment}\\
 &=-\beta_t\log\langle q_t,e^{-\widehat f_t/\beta_t}\rangle
  =\langle q_t,\widehat f_t\rangle-\mathsf D_t\,,
       \label{eq:D-ftrl-stability-identity}\\
 \sum_{t=1}^T\langle q_t,\widehat f_t\rangle
 &=\sum_{t=1}^T\mathsf D_t+
      \sum_{t=1}^T[F_{\beta_t}(S_t)-F_{\beta_t}(S_{t-1})]
       \label{eq:D-ftrl-sum-substitution}\\
 &=\sum_{t=1}^T\mathsf D_t+F_{\beta_T}(S_T)-F_{\beta_1}(0)
   +\sum_{t=2}^T[F_{\beta_{t-1}}(S_{t-1})-F_{\beta_t}(S_{t-1})]
       \label{eq:D-ftrl-telescoping}\\
 &\le\sum_{t=1}^T\mathsf D_t+F_{\beta_T}(S_T)
       \label{eq:D-ftrl-monotonicity-used}\\
 &\le\sum_{t=1}^T\mathsf D_t+\langle q_{u,T},S_T\rangle
                          +\beta_T\KL(q_{u,T},\mu)\,.
       \label{eq:SHr-general-comparator}
\end{align}
Equation \eqref{eq:D-ftrl-potential-increment} substitutes
$S_t=S_{t-1}+\widehat f_t$ into the logarithmic potential.
Equation \eqref{eq:D-ftrl-stability-identity} substitutes the exponential-weights
density and then the definition of $\mathsf D_t$.
Equation \eqref{eq:D-ftrl-sum-substitution} sums that identity;
\eqref{eq:D-ftrl-telescoping} groups the terms at each $S_{t-1}$.
For \eqref{eq:D-ftrl-monotonicity-used}, the variational formula gives
$F_{\beta'}(S)\ge F_\beta(S)$ when $\beta'\ge\beta$, because
$\KL(q,\mu)\ge0$, and $F_{\beta_1}(0)=0$.
Equation \eqref{eq:SHr-general-comparator} evaluates the variational
infimum at the deterministic comparison law $q_{u,T}$.

For a realized coefficient $v_{X_t}(p_t)$ and observation, let
$h_t(x)=(x-\bar x_t)^\top v_{X_t}(p_t)Y_t/\beta_t$.
\begin{align}
\int h_t(x)q_t(dx)
 &=\beta_t^{-1}\left[\int xq_t(dx)-\bar x_t\right]^\top v_{X_t}(p_t)Y_t=0\,,
       \label{eq:D-stability-zero-mean}\\
 \mathsf D_t&=\beta_t\log\int e^{-h_t(x)}q_t(dx)
       \label{eq:D-stability-centered-log}\\
 &\le\beta_t\int[e^{-h_t(x)}-1+h_t(x)]q_t(dx)
       \label{eq:D-stability-log-upper}\\
 &\le\beta_t\int\chi(h_t(x))q_t(dx)\,,
       \label{eq:D-stability-chi-upper}\\
 \E_{t-1}[\mathsf D_t]
 &\le\beta_t\int\!\int
   \E\!\left[\left.\chi\!\left(\frac{(u-\bar x_t)^\top v_x(p_t)Y_t}{\beta_t}\right)
                 \right|\cF_{t-1},X_t=x\right]
                                q_t(du)p_t(dx)
       \label{eq:D-stability-tower}\\
 &\le\beta_t\int\!\int
           \chi\!\left(\frac{(u-\bar x_t)^\top v_x(p_t)}{\beta_t}\right)
                                q_t(du)p_t(dx)
       \label{eq:D-stability-feedback-bound}\\
 &=\gC_t(p_t)\le\frac{2d}{\beta_t}\,.
       \label{eq:SHr-conditional-stability}
\end{align}
Equation \eqref{eq:D-stability-zero-mean} substitutes the barycenter
of $q_t$.  Equation \eqref{eq:D-stability-centered-log} cancels the
scalar offset in the definition of $\mathsf D_t$.
Equation \eqref{eq:D-stability-log-upper} uses $\log z\le z-1$ and
\eqref{eq:D-stability-zero-mean}; \eqref{eq:D-stability-chi-upper}
uses $e^{-h}-1+h\le e^h+e^{-h}-2$.
Equation \eqref{eq:D-stability-tower} conditions on the virtual action,
and \eqref{eq:D-stability-feedback-bound} uses $|Y_t|\le1$ and the
evenness and monotonicity of $\chi$ on $[0,\infty)$.
Equation \eqref{eq:SHr-conditional-stability} is exactly the
proxy function constraint in \eqref{eq:SH-program} in \GIAFTRL{}.

\begin{align}
p_t-q_t&=\Gamma_t(\widetilde p_t-q_t)\,,\label{eq:D-exploration-signed-law}\\
 \langle p_t-q_t,x^\top\theta_t\rangle
 &=\Gamma_t\left[\int x^\top\theta_t\widetilde p_t(dx)
                         -\int x^\top\theta_t\,q_t(dx)\right]
       \label{eq:D-exploration-substitution}\\
 &\le\Gamma_t\left[\sup_{x\in\mathcal K}x^\top\theta_t
                         -\inf_{x\in\mathcal K}x^\top\theta_t\right]
       \label{eq:D-exploration-range}\\
 &\le2\Gamma_t=2\frac{4d}{\beta_t}=\frac{8d}{\beta_t}\,.
       \label{eq:SHr-exploration-all}
\end{align}
Equation \eqref{eq:D-exploration-signed-law} is the definition of
$\widetilde p_t$ in \eqref{eq:SHr-stability-residual};
\eqref{eq:D-exploration-substitution} integrates that signed measure.
Equation \eqref{eq:D-exploration-range} bounds the expectations under
two probability laws by the range of the integrand.
Equation \eqref{eq:SHr-exploration-all} uses the bounded losses and
substitutes $\Gamma_t=4d/\beta_t$.

Rounding and reconstruction in \Cref{lem:SH-feasible} give,
for $x,y\in\mathcal K$,
\begin{align}
\E_{t-1}[A_t^\top\theta_t]
 &=\E_{t-1}\!\left[\E[A_t\mid\cF_{t-1},X_t,v_{X_t}(p_t)]^\top\theta_t\right]
 =\int x^\top\theta_t p_t(dx)\,,\label{eq:D-regret-rounding}\\
 \E_{t-1}[\widehat f_t(x)-\widehat f_t(y)]
 &=(x-y)^\top\E_{t-1}[v_{X_t}(p_t)Y_t]
 =(x-y)^\top\theta_t\,,
       \label{eq:SHr-regret-unbiased}\\
 \Reg(u)
 &=\sum_{t=1}^T\E[\langle p_t-\delta_u,x^\top\theta_t\rangle]
       \label{eq:D-adv-regret-definition}\\
 &=\sum_{t=1}^T\E[\langle q_t-q_{u,T},\widehat f_t\rangle]
   +\sum_{t=1}^T\E[\langle p_t-q_t,x^\top\theta_t\rangle]
   +\sum_{t=1}^T\E[\langle q_{u,T}-\delta_u,x^\top\theta_t\rangle]
       \label{eq:D-adv-regret-three-pieces}\\
 &\le\beta_T\KL(q_{u,T},\mu)
       +\sum_{t=1}^T\E[\mathsf D_t]
       +\sum_{t=1}^T\frac{8d}{\beta_t}
       +\sum_{t=1}^T\frac2T
       \label{eq:D-adv-regret-four-bounds}\\
 &\le\beta_T[\log2+16k\log T]
       +\sum_{t=1}^T\frac{2d}{\beta_t}
       +\sum_{t=1}^T\frac{8d}{\beta_t}+2
       \label{eq:D-adv-regret-prior-stability}\\
 &=\beta_T[\log2+16k\log T]
       +10d\sum_{t=1}^T\frac1{16d+\sqrt t}+2
       \label{eq:D-adv-regret-schedule}\\
 &\le\beta_T[\log2+16k\log(eT)]
       +10d\sum_{t=1}^T t^{-1/2}+2
       \label{eq:D-adv-regret-sqrt-sum}\\
 &\le\beta_T[\log2+16k\log(eT)]+20d\sqrt T+2\,.
       \label{eq:SHr-adversarial-sum}
\end{align}
Equation \eqref{eq:D-regret-rounding} uses the conditional mean of the
rounding kernel.  Equation \eqref{eq:SHr-regret-unbiased} substitutes
full parameter-vector reconstruction.
Equation \eqref{eq:D-adv-regret-definition} is fixed-comparator regret;
\eqref{eq:D-adv-regret-three-pieces} substitutes
$p_t-\delta_u=(q_t-q_{u,T})+(p_t-q_t)+(q_{u,T}-\delta_u)$ and uses
unbiased differences in
$\E[\sum_{t\le T}\langle q_t-q_{u,T},\widehat\ell_{t,\cdot}\rangle]$
of \eqref{eq:D-adv-regret-three-pieces}.
Equation \eqref{eq:D-adv-regret-four-bounds} substitutes
\eqref{eq:SHr-general-comparator}, \eqref{eq:SHr-exploration-all}, and
the comparator approximation in \Cref{lem:SH-prior-comparator}.
Equation \eqref{eq:D-adv-regret-prior-stability} inserts that lemma's
entropy bound and \eqref{eq:SHr-conditional-stability}.
Equation \eqref{eq:D-adv-regret-schedule} collects the two reciprocal
schedule sums; \eqref{eq:D-adv-regret-sqrt-sum} uses
$\beta_t\ge\sqrt t$ and $\log T\le\log(eT)$.
The final summation is explicit:
\begin{align}
\sum_{t=1}^Tt^{-1/2}
 &\le\sum_{t=1}^T\int_{t-1}^t x^{-1/2}\,dx
 =\int_0^T x^{-1/2}\,dx
 =[2\sqrt x]_{0}^{T}=2\sqrt T\,.
       \label{eq:D-adv-integral-evaluation}
\end{align}
Equation \eqref{eq:D-adv-integral-evaluation} uses the decreasing
integrand; substituting it gives \eqref{eq:SHr-adversarial-sum}.
The right side is independent of $u$, and each actual loss difference is
at most two. Consequently
\begin{align}
\Reg
 &=\sup_{u\in\mathcal K}\Reg(u)
 \le\min\{2T,\beta_T[\log2+16k\log(eT)]+20d\sqrt T+2\}
 =U_{\rm SH}(T)\,,\label{eq:SHr-adv-final}\\
 T<d^2\quad&\Longrightarrow\quad
 U_{\rm SH}(T)\le2T\le2d\sqrt T
                     \le320(d+1)\sqrt T\log(eT)\,,\label{eq:D-adv-small-T}\\
 T\ge d^2\quad&\Longrightarrow\quad
 \beta_T=16d+\sqrt T\le17\sqrt T\,,\label{eq:D-adv-large-schedule}\\
 U_{\rm SH}(T)
 &\le17\sqrt T[1+16d\log(eT)]+20d\sqrt T+2\sqrt T
       \qquad(T\ge d^2)\label{eq:D-adv-large-substitution}\\
 &\le[17+272d+20d+2]\sqrt T\log(eT)
 =(19+292d)\sqrt T\log(eT)\label{eq:D-adv-large-collection}\\
 &\le320(d+1)\sqrt T\log(eT)\,.\label{eq:SHr-adv-simple}
\end{align}
Equation \eqref{eq:SHr-adv-final} takes the supremum of a bound
independent of $u$, and uses the per-round bound by two.
Equation \eqref{eq:D-adv-small-T} treats the small-horizon case.
Equation \eqref{eq:D-adv-large-schedule} follows from $d\le\sqrt T$.
Equation \eqref{eq:D-adv-large-substitution} inserts this schedule
bound, $k\le d$, $\log2\le1$, and $1\le\sqrt T$.
Equation \eqref{eq:D-adv-large-collection} uses $\log(eT)\ge1$;
\eqref{eq:SHr-adv-simple} uses $19+292d\le320(d+1)$.
All histories are retained, and neither \eqref{eq:SHr-adv-final} nor \eqref{eq:SHr-adv-simple} uses sharpness.
\end{proof}

\subsubsection{Corruption Parameters and Deterministic Interfaces}
\label{sec:SC-proof}
The stochastic argument for a fixed allocation is not applied to a corrupted drift. This
subsection supplies the corruption estimates for
\GIAFTRL{}, with the fixed reinforced
prior \eqref{eq:SH-priority}. The reinforcement is in that prior;
the FTRL update, homogeneous budgets, sampling-law program, rounding rule,
and anytime learning rate are unchanged. In particular the proof does
not restart the process at a corruption-dependent time.

Let $\kappa_t=\norm{c_t^{\rm vec}}_\cA$ and assume
$\sum_{t\le T}\kappa_t\le C$ almost surely, with the standing predictable
linear-mean and bounded-feedback assumptions. For a fixed evaluation
horizon, extend the environment with zero further corruption. All
probability bounds below then concern this extension of the same run.
Let $A_s^c=\sum_{j<s}\kappa_j$ and $E_s(a)=\sum_{j<s}(a-a^*)^\top c_j^{\rm vec}$.
\begin{lemma}[The all-history corruption interface]\label{lem:D-cor-interface}
For $D_s(a)=(a-a^*)^\top v_{X_s}(p_s)Y_s$, every predictable bounded-mean
environment satisfies
\begin{equation}\label{eq:SC-raw-interface}
 \E_{s-1}[D_s(a)]=\Delta(a)+(a-a^*)^\top c_s^{\rm vec}\,,\qquad
 |\E_{s-1}[D_s(a)]|\le2\,.
\end{equation}
The update magnitude and second-moment inequalities
\eqref{eq:SH-global-amp}--\eqref{eq:SH-global-var} in
\Cref{lem:SH-feasible} are unchanged.
Moreover, $\kappa_s\le2$ and
$|u^\top c_s^{\rm vec}|\le\Lambda\kappa_s$ for
$u\in\mathcal D_\theta$.
\end{lemma}
\begin{proof}
\begin{align}
\E_{s-1}[D_s(a)]
 &=(a-a^*)^\top\theta_s
  =(a-a^*)^\top(\theta+c_s^{\rm vec})
  =\Delta(a)+(a-a^*)^\top c_s^{\rm vec}\,,
       \label{eq:D-cor-drift-substitution}\\
 |(a-a^*)^\top\theta_s|
 &\le |a^\top\theta_s|+|(a^*)^{\top}\theta_s|\le1+1=2\,,
       \label{eq:D-cor-drift-range}\\
 \kappa_s&=\sup_{a\in\cA}|a^\top(\theta_s-\theta)|
 \le\sup_{a\in\cA}(|a^\top\theta_s|+|a^\top\theta|)\le2\,,
       \label{eq:D-cor-budget-cap}\\
 |u^\top c_s^{\rm vec}|
 &\le\|u\|_{\mathcal S}\|c_s^{\rm vec}\|_\cA
 \le\Lambda\kappa_s\qquad(u\in\mathcal D_\theta)\,.
       \label{eq:D-cor-direction-drift}
\end{align}
Equation \eqref{eq:D-cor-drift-substitution} uses reconstruction and
conditional rounding in \Cref{lem:SH-feasible}.
Equations \eqref{eq:D-cor-drift-range} and
\eqref{eq:D-cor-budget-cap} use the bounded means under $\theta_t$ and $\theta$.
Equation \eqref{eq:D-cor-direction-drift} uses the duality between
$\|\cdot\|_\cA$ and the gauge of $\mathcal S$, and
\Cref{lem:D-projective}.  The update magnitude and variance
calculations in \Cref{lem:SH-feasible} use only $|Y_s|\le1$
and the executed program's constraints, not the sign or stationarity
of the conditional drift.
\end{proof}

\begin{definition}[Corruption exponents, envelopes, and finite constants]\label{def:D-cor-dictionary}
Use the constraint parameter $B$ of \GIAFTRL{} specified in
\Cref{sec:SH-algorithm}, the analysis constants
$P=4096(d+1)$, $C_{\mathrm{rat}}=64$, and $C_{\beta,\mathrm{sh}}=16d+1$.
The constants $R,\kappa,p,\eta,v$ are specified by
\eqref{eq:SH-moment-dictionary}--\eqref{eq:SH-global-constants} in
\Cref{def:D-moment-dictionary}.
Use $\Lambda,b_\theta$ from \eqref{eq:SH-Lambda} in
\Cref{lem:D-projective}.
Thus $p=P/2$, $\eta=P(7/8-\kappa)$ and $b_\theta=C_{\beta,\mathrm{sh}}\Lambda^2$.
Set
\begin{equation}\label{eq:SC-exponents}
 \omega=\frac78\,,\qquad a=\frac P{128}\,,\qquad
 q=\frac{P(d+v)}{P+d}\,,\qquad r_-=\frac P8-q\,,\qquad k_0=\frac1{4096}\,.
\end{equation}
These exponents quantify the decay of exceptional histories and the
propagation of pre-window estimation error. They are not action vectors
or probability measures.
The numerical margins required of these exponents are proved in
\Cref{lem:D-numerical-margins}.

For $b'\ge C_{\beta,\mathrm{sh}}$ and $z\ge1$, define the following finite maxima, using
$H_F(b')$ from \eqref{eq:SH-moment-dictionary} in \Cref{def:D-moment-dictionary}:
\begin{equation}\label{eq:SC-maxima}
\begin{split}
 \mathcal D(b',z)&=\max_{0\le j\le d}
 (2^P4^j)^{P/(P+j)}z^{Pj/(P+j)}H_F(b')^{P/(P+j)}\,,\\
 \mathcal E(z)&=\max_{0\le j\le d}
 (2^P4^j)^{P/(P+j)}z^{Pj/(P+j)}(2^vz^P)^{P/(P+j)}\,,\\
 A_g&=\mathcal D(C_{\beta,\mathrm{sh}},R+2)\,,\quad B_g=\mathcal E(R+2)\,,\\
 A_u&=\mathcal D(b_\theta,\Lambda(R+2))\,,\quad
 B_u=\mathcal E(\Lambda(R+2))\,,\quad
 \zeta=\min\{1,\Delta_{\rm pre},\varsigma\}\,.
\end{split}
\end{equation}
The index $j$ in these maxima is an integer. In particular $\zeta>0$.
Set
\begin{equation}\label{eq:SC-reference-constant}
 A_r=1+2^{\kappa P+1}A_g+B_g+(4R)^P
          +2^{12P}(2R+10)^d\zeta^{-p}\,,
\end{equation}
and define
\begin{equation}\label{eq:SC-clock}
\begin{split}
 K_0=\left\lceil\max\left\{256,16^8,[S(C_{\beta,\mathrm{sh}})+1]^2,
 [S(b_\theta)+1]^2,(2C_{\beta,\mathrm{sh}}h^*)^{8/3},(2h^*)^{1/4},\Lambda\right\}\right\rceil\,,\\
 n=\lceil C\rceil+2\,,\qquad L_n=\log(en)\,,\qquad
 N=N_\theta(C)=\left\lceil K_0 n L_n^2\right\rceil \,.
\end{split}
\end{equation}
The clock $N$ is the analysis scale beyond which the remaining-budget
shift can be controlled; its square root bounds the weighted bad-history
cost. The quantities in this display are never read by the learner. Next set
\begin{equation}\label{eq:SC-final-threshold}
\begin{split}
 D_0&=\Lambda R+4\,,\\
 F_\theta&=\left\lceil\max\left\{256,16^8,T_*,
 (8D_0)^{2/(1-k_0)},64^{2/(7/8-k_0)},
 [40(P-1)b_\theta]^{8/3}\right\}\right\rceil\,,\\
 D_\theta&=2A_r+2^{\kappa P+1}A_u+2^{P/8+1}B_u
                      +2^{10P}(\Lambda R+10)^d\,,\\
 W_\theta&=2\sqrt{F_\theta+1}+4D_\theta\,,\qquad
 E_\theta=16\sqrt{K_0}(h^*+10d)W_\theta\,.
\end{split}
\end{equation}
Here $T_*$ and $S(\cdot)$ are explicitly defined in
\eqref{eq:SH-laplace-constant} in \Cref{def:D-laplace} and \eqref{eq:SH-global-constants} in \Cref{def:D-moment-dictionary}.
There is no circular choice of constants: all of them depend only on the
fixed sharp instance, its fixed reference, and $d$, except for the three
clock quantities $n,L_n,N$.

For a fixed centered feasible sampling allocation $n^{\rm des}$, let
$c_{\mathrm{ctr}}^{n^{\rm des}}$ and $N_{n^{\rm des}}$ be its regret cost and mass, and use
$a_{n^{\rm des}},a_\rho$ from \eqref{eq:SH-witness-parameters} in \Cref{def:D-witness}.
The distinct notation $n^{\rm des}$ avoids confusing the sampling allocation with the
integer $n$ in \eqref{eq:SC-clock}. Define
\begin{equation}\label{eq:SC-constants}
\begin{split}
 M_{n^{\rm des}}={}&\beta_1h^*+4A_L(1+C_{\mathrm{rat}}^2)+2e^2C_{\beta,\mathrm{sh}}A_L+E_\theta\\
 &+8d\left(4C_{\mathrm{rat}}^4a_{n^{\rm des}}+\frac83C_{\mathrm{rat}}^3\sqrt{a_\rho}\right)\,,\\
 J^{\rm anc}_\theta={}&E_\theta+16\sqrt2d\,,\\
 J^{\rm cor}_\theta={}&J^{\rm anc}_\theta+
 \frac{320(d+1)}{\sqrt{\varsigma}}
                 [2+\log(1+\varsigma^{-1})]\,.
\end{split}
\end{equation}
In particular the corruption coefficient does not depend on the comparison
sampling allocation or its total mass. No optimality of these conservative constants is
claimed.
\end{definition}

\begin{lemma}[Numerical margins and deterministic window thresholds]
\label{lem:D-numerical-margins}
All denominators in \Cref{def:D-moment-dictionary}
are positive.  The exponents in
\Cref{def:D-cor-dictionary} satisfy
$q\ge1$, $q/P<1/1024$, $r_->P/9$, $a_A>16$, and
\begin{equation}\label{eq:D-margin-exponents}
 \min\{a/2,\eta/2,r_-/2,(p-d)/8,7(p-d)/16\}>2\,.
\end{equation}
The thresholds in \Cref{def:D-local-window,def:D-cor-window} ensure $m\le t/2$, the indicated
lower bounds on $m$, and $\tau_t\le t$.
For $s\ge S(b')$ and $Y\ge s^{7/8}$,
\begin{equation}\label{eq:D-margin-taylor}
 3\beta_s'\le Y/B\,,\qquad
 10(P+1)\beta_s'/Y^2\le1/(4s)\,,\qquad
 (1-3/B)^{-P-2}\le2\,.
\end{equation}
\end{lemma}
\begin{proof}
Substitute $P=4096(d+1)$ and $B=2^{24}P$ before bounding:
\begin{align}
\frac dP&=\frac{d}{4096(d+1)}<\frac1{4096}\,,
       \label{eq:D-numeric-dim}\\
 \frac{a_G}P&=\frac{1152}{2^{24}}<1\,,
 \qquad\frac{a_F}P=\frac{576}{2^{24}}<\frac34\,,
       \label{eq:D-numeric-rec-coefficients}\\
 \frac vP&=\frac{576}{2^{24}}+\frac1{4096}
 =\frac{4672}{2^{24}}<\frac1{2048}<\frac34=\frac\nu P\,,
       \label{eq:D-numeric-v}\\
 \frac qP&=\frac{d+v}{P+d}
 \le\frac dP+\frac vP
 <\frac1{4096}+\frac1{2048}
 =\frac3{4096}<\frac1{1024}\,,\label{eq:D-numeric-q}\\
 q-1&=\frac{(P-1)d+P(v-1)}{P+d}\ge0\,,
       \label{eq:D-numeric-q-lower}\\
 \kappa&=\frac34+\frac{d}{4(P+d)}<\frac34+\frac1{16384}\,,
 \qquad\frac\eta P=\frac78-\kappa>
             \frac18-\frac1{16384}>\frac19\,,
       \label{eq:D-numeric-kappa}\\
 \frac{r_-}P&=\frac18-\frac qP>
                  \frac18-\frac1{1024}>\frac19\,,
       \label{eq:D-numeric-rminus}\\
 a_A&=\min\{\eta/2,P/32-d/8,(1-\kappa)P\}
 >\min\{P/18,P/33,P/5\}=P/33>16\,.
       \label{eq:D-numeric-anchor-exponent}
\end{align}
Equations \eqref{eq:D-numeric-dim}--\eqref{eq:D-numeric-v}
are the exact dimension and coefficient substitutions; they prove
positivity of $P-a_G$, $\nu-a_F$, $\nu-v$, and $\lambda_F$.
Equation \eqref{eq:D-numeric-q} bounds the normalized propagation
exponent.  Equation \eqref{eq:D-numeric-q-lower} uses $v\ge P/4096\ge2$.
Equations \eqref{eq:D-numeric-kappa}--\eqref{eq:D-numeric-anchor-exponent}
use these fractions in the definitions of the tail exponents.
Since $P\ge8192$ and $p=P/2$,
\begin{align}
a/2&=P/256\ge32>2\,,
 &\eta/2&>P/18>2\,,& r_-/2&>P/18>2\,,\label{eq:D-numeric-tail-a}\\
 (p-d)/8&>P(1/2-1/4096)/8>2\,,
 &7(p-d)/16&>7P(1/2-1/4096)/16>2\,.
       \label{eq:D-numeric-tail-b}
\end{align}
Equations \eqref{eq:D-numeric-tail-a} and
\eqref{eq:D-numeric-tail-b} prove \eqref{eq:D-margin-exponents}.

The definition of $S(b')$ gives $s^{3/8}\ge3Bb'$ and
$s^{1/4}\ge40(P+1)b'$.  Hence
\begin{align}
3\beta_s'&\le3b'\sqrt s
 \le s^{7/8}/B\le Y/B\,,\label{eq:D-numeric-taylor-amp}\\
 \frac{10(P+1)\beta_s'}{Y^2}
 &\le10(P+1)b's^{-5/4}\le\frac1{4s}\,,
       \label{eq:D-numeric-taylor-var}\\
 (1-3/B)^{-P-2}
 &=\exp\{(P+2)[-\log(1-3/B)]\}
 \le\exp\{3(P+2)/(B-3)\}\le2\,.
       \label{eq:D-numeric-taylor-factor}
\end{align}
Equation \eqref{eq:D-numeric-taylor-amp} uses the update magnitude threshold;
\eqref{eq:D-numeric-taylor-var} uses the variance threshold.
Equation \eqref{eq:D-numeric-taylor-factor} uses
$-\log(1-x)\le x/(1-x)$ and $B=2^{24}P$, $P\ge8192$.

For the clean window, $t\ge T_\theta$ gives
\begin{align}
\sqrt t\le m=\lceil\sqrt t\rceil&\le2\sqrt t\le t/2\,,
 &m&\ge T_A,S(b_\theta)\,,\label{eq:D-numeric-clean-m}\\
 \tau_t/t&=t^{-1/8}+4t^{-3/8}
 \le5t^{-1/8}\le1\,.
       \label{eq:D-numeric-clean-tau}
\end{align}
Here \eqref{eq:D-numeric-clean-m} uses $t\ge16$ and the squared
thresholds, while \eqref{eq:D-numeric-clean-tau} uses $t\ge5^8$.
For the corrupted window, $N\ge K_0n$ and $t\ge F_\theta N$ give
\begin{align}
m=\lceil\sqrt{Nt}\rceil&\ge N\ge\Lambda n\ge\Lambda C\,,
 &m&\ge S(b_\theta)\,,\label{eq:D-numeric-cor-m-lower}\\
 m/t&\le2\sqrt{N/t}\le2F_\theta^{-1/2}\le1/2\,,
       \label{eq:D-numeric-cor-m-upper}\\
 z=m^{7/8}t^{1/8}&\ge m\ge\Lambda C\,,
 &\tau_s&\ge4z\ge4\Lambda C\,,\label{eq:D-numeric-cor-buffer}\\
 \tau_t/t&=t^{-1/8}+4(m/t)^{7/8}
 \le F_\theta^{-1/8}+8F_\theta^{-7/16}
 \le16^{-1}+8\cdot16^{-7/2}<1\,.
       \label{eq:D-numeric-cor-tau}
\end{align}
Equation \eqref{eq:D-numeric-cor-m-lower} uses $K_0\ge\Lambda$
and $K_0\ge[S(b_\theta)+1]^2$.
Equation \eqref{eq:D-numeric-cor-m-upper} uses $F_\theta\ge256$.
Equation \eqref{eq:D-numeric-cor-buffer} uses $t\ge m$.
Equation \eqref{eq:D-numeric-cor-tau} uses $F_\theta\ge16^8$.
These are properties of the analysis windows, not learner inputs.
\end{proof}

\subsubsection{Relative-Gap Bounds and Their Moment Inputs}
\label{sec:SH-anchor}
Fix a sharp instance for the analysis only. Let
$a^*=b_{j^*}$, $h^*=h_{j^*}$, $\pi^*=e^{-h^*}$ and
$\Delta(a)=(a-a^*)^\top\theta$.
\begin{definition}[Moment coefficients and global envelopes]\label{def:D-moment-dictionary}
The following dictionary makes every later transient explicit. For each
$b'\ge C_{\beta,\mathrm{sh}}$, define
\begin{equation}\label{eq:SH-moment-dictionary}
\begin{aligned}
 a_G&=1152P^2/B\,,\quad c_0=2P\,,\quad c_G(b')=8P^2b'\,,\\
 w_G(b')&=8b'+2^{17}P^2(b')^2/B\,,\\
 D_G(b')&=\tfrac12P^2(12Pb')^{P-2}w_G(b')\,,\\
 C_G(b')&=\max\left\{1,\left(\frac{3c_0}{P-a_G}\right)^P,
 \left(\frac{3c_G(b')}{P-a_G}\right)^{P/2},
 \frac{3D_G(b')}{P-a_G}\right\}\,,\\
 \nu&=3P/4\,,\quad a_F=576P^2/B\,,\quad v_F(b')=10b'+64(b')^2/B\,,\\
 c_F(b')&=16P^2v_F(b')\,,\quad
 w_F(b')=v_F(b')+2^{16}P^2(b')^2/B\,,\\
 D_F(b')&=\tfrac12P^2(36Pb')^{P-2}w_F(b')\,,\\
 C_F(b')&=\max\left\{1,
 \left(\frac{2c_F(b')}{\nu-a_F}\right)^{P/2},
 \frac{2D_F(b')}{\nu-a_F}\right\}\,,\\
 \lambda_F&=P/4096\,,\quad v=a_F+\lambda_F\,,\quad
 R_F(b')=\max\{1,[c_F(b')/\lambda_F]^{P/2}\}\,,\\
 J_F(b')&=c_F(b')R_F(b')^{1-2/P}+D_F(b')\,,\quad
 H_F(b')=\max\{1,J_F(b')/(\nu-v)\}\,.
\end{aligned}
\end{equation}
All denominators are positive. In this dictionary $v$ is a scalar exponent,
not a reconstruction mark. Set
\begin{equation}\label{eq:SH-global-constants}
\begin{split}
 R&=8C_G(C_{\beta,\mathrm{sh}})^{1/P}\,,\qquad
 \kappa=\frac34+\frac{d}{4(P+d)}\,,\qquad \alpha=7/8\,,\qquad p=P/2\,,\\
 C_M&=\max_{0\le j\le d}
 (2^P4^j)^{P/(P+j)}R^{Pj/(P+j)}C_F(C_{\beta,\mathrm{sh}})^{P/(P+j)}\,,\\
 \eta&=P(\alpha-\kappa)\,,\qquad A_H=2^{\kappa P+1}C_M\,,\\
 S(b')&=\left\lceil\max\{16,(3Bb')^{8/3},[40(P+1)b']^4\}\right\rceil\,,
 \qquad T_g=\left\lceil\max\{16,[S(C_{\beta,\mathrm{sh}})+1]^2\}\right\rceil\,.
\end{split}
\end{equation}
These global constants do not depend on the sharp instance.
\end{definition}

\begin{lemma}[Absolute and negative maxima of cumulative loss differences]\label{lem:D-global-moments}
For the stochastic process $X_t$ of \Cref{lem:SH-feasible},
\begin{equation}\label{eq:SH-global-moments}
 \E\left[(\max_{a\in\mathcal K}|X_t(a)|)^P\right]\le R^Pt^P\,,
 \qquad \E[M_t^P]\le C_Mt^{\kappa P}\,.
\end{equation}

The same bounds hold for a process jointly frozen before a predictable
update, at each deterministic observation time.
\end{lemma}
\begin{proof}
Here is the complete conversion from integral moments to the two
supremum moments. Let $\sigma$ be normalized relative volume on
$\mathcal K$, $j=\dim\operatorname{aff}(\mathcal K)$, and set
$G_t=\int|X_t|^Pd\sigma$, $F_t=\int(-X_t)_+^Pd\sigma$,
$R_t^X=\max_{\mathcal K}|X_t|$, and $c_j=2^P4^j$.
\Cref{lem:SH-field-interface}, with $b'=C_{\beta,\mathrm{sh}}$, gives
\begin{align}
\E[G_t]&\le C_G(C_{\beta,\mathrm{sh}})t^P\,,
 &\E[F_t]&\le C_F(C_{\beta,\mathrm{sh}})t^\nu\,,
       \label{eq:D-step-0094}\\
 (R_t^X)^P&\le c_jG_t\,,
 &\E[(R_t^X)^P]&\le c_jC_G(C_{\beta,\mathrm{sh}})t^P
                  \le8^PC_G(C_{\beta,\mathrm{sh}})t^P=R^Pt^P\,.
 \label{eq:SHr-global-absolute-application}
\end{align}
Equation \eqref{eq:D-step-0094} applies the absolute and negative integral recurrences of \Cref{lem:SH-field-interface} to the raw field from its zero initialization.
Equation \eqref{eq:SHr-global-absolute-application} applies
\eqref{eq:C-aff-abs} in \Cref{lem:C-affine} before expectation,
then uses $c_j\le8^P$ and $R^P=8^PC_G(C_{\beta,\mathrm{sh}})$.
The negative affine inequality, raised to $P/(P+j)$, gives
\begin{align}
M_t^{P+j}&\le c_j(R_t^X)^jF_t\,,
       \label{eq:D-step-0095}\\
 M_t^P&\le c_j^{P/(P+j)}(R_t^X)^{Pj/(P+j)}F_t^{P/(P+j)}\,,
       \label{eq:D-step-0096}\\
 \E[M_t^P]
 &\le c_j^{P/(P+j)}
       [\E[(R_t^X)^P]]^{j/(P+j)}[\E[F_t]]^{P/(P+j)}
       \label{eq:D-step-0097}\\
 &\le c_j^{P/(P+j)}(R^Pt^P)^{j/(P+j)}
                              [C_F(C_{\beta,\mathrm{sh}})t^\nu]^{P/(P+j)}
       \label{eq:D-step-0098}\\
 &=c_j^{P/(P+j)}R^{Pj/(P+j)}C_F(C_{\beta,\mathrm{sh}})^{P/(P+j)}
                              t^{P(j+\nu)/(P+j)}
       \label{eq:D-step-0099}\\
 &\le C_Mt^{\kappa P}\,.
 \label{eq:SHr-global-negative-application}
\end{align}
Equation \eqref{eq:D-step-0095} applies the negative-cap inequality of \Cref{lem:C-affine} to the fixed-history affine field.
Equation \eqref{eq:D-step-0096} raises that nonnegative inequality to the power $P/(P+j)$.
Equation \eqref{eq:D-step-0097} applies H\"older with reciprocal exponents $j/(P+j)$ and $P/(P+j)$.

Equation \eqref{eq:D-step-0098} substitutes both integral and absolute-moment bounds from \eqref{eq:SHr-global-absolute-application}.
Equation \eqref{eq:D-step-0099} collects the powers of $R$, $C_F(C_{\beta,\mathrm{sh}})$, and $t$.
Equation \eqref{eq:SHr-global-negative-application} uses the maximum defining $C_M$ and the exponent comparison \eqref{eq:SHr-global-negative-exponent}.
The H\"older exponents in equation \eqref{eq:D-step-0097} are $(P+j)/j$ and
$(P+j)/P$. For $j=0$, the factor
$[\E[(R_t^X)^P]]^{j/(P+j)}$ in \eqref{eq:D-step-0097} is one,
and the negative moment is evaluated directly.
In equation \eqref{eq:SHr-global-negative-application},
\begin{align}
\frac{j+\nu}{P+j}
 &=\frac{j+3P/4}{P+j}
 =\frac34+\frac{j}{4(P+j)}
 \le\frac34+\frac{d}{4(P+d)}=\kappa\,,
 \label{eq:SHr-global-negative-exponent}
\end{align}
and the coefficient is one of those maximized in
\eqref{eq:SH-global-constants} in \Cref{def:D-moment-dictionary}. The frozen recurrences in
\eqref{eq:SHr-frozen-recursions} in \Cref{lem:SH-field-interface} give the same two integral bounds \eqref{eq:D-step-0094} for jointly frozen fields.
Equations \eqref{eq:SHr-global-absolute-application}--\eqref{eq:SHr-global-negative-exponent}
therefore hold for those fields at deterministic observation times too.

\end{proof}

\begin{definition}[Reference-identification constants]\label{def:D-reference-constants}
Recall the prefix gap from Section~\ref{sec:SH-algorithm} and define
\begin{equation}\label{eq:SH-anchor-constants}
\begin{split}
 \Delta_{\rm pre}&=\min(\{1\}\cup\{\Delta(b_i):i\in J,\ i<j^*\})\in(0,1]\,,\\
 T_A&=\left\lceil\max\left\{T_g,\Delta_{\rm pre}^{-16},
       (2h^*/\varsigma^8)^2,(C_{\beta,\mathrm{sh}}h^*)^{8/3}\right\}\right\rceil\,,\\
 a_A&=\min\left\{\eta/2,\ P/32-d/8,\ (1-\kappa)P\right\}\,,\\
 A_A&=A_H+16^p(R+3)^d+2^PC_M\varsigma^{-8P}\,.
\end{split}
\end{equation}
Here $\Delta_{\rm pre}=1$ when the prefix is empty.
The inequality $a_A>16$ is verified in \Cref{lem:D-numerical-margins}.
\end{definition}

\label{sec:SH-projective}
\begin{definition}[Gap-normalized vector differences]\label{def:D-projective}
The physical minimum gap is never assumed positive in this subsection.
Define the compact projective body and its cone section
\begin{equation}\label{eq:SH-projective}
 \mathcal U_\theta=
 \overline{\left\{\frac{a-a^*}{\Delta(a)}:a\in\mathcal K\setminus\{a^*\}\right\}}\,,
 \qquad \mathcal D_\theta=\Conv(\{0\}\cup\mathcal U_\theta)\,.
\end{equation}

Here $u_a\coloneq(a-a^*)/\Delta(a)$, for $a\ne a^*$, is the vector difference
$a-a^*$ normalized by the loss gap $\Delta(a)$.
The identity $u_a^\top \sum_{s=1}^{t-1}\widehat\theta_s=X_t(a)/\Delta(a)$ converts infinitely
many arbitrarily small gaps into values of one linear function.
\end{definition}

\begin{lemma}[Projective extension of cumulative loss-difference estimates]\label{lem:D-projective}
The set $\mathcal U_\theta$ is compact and convex, and
$\theta^\top u=1$ on $\mathcal U_\theta$.
The gauge constant $\Lambda$ in \eqref{eq:SH-Lambda} below is finite;
\eqref{eq:SH-initial-extension} holds for every $L\in\R^d$.
\end{lemma}
\begin{proof}
Sharpness gives the explicit boundedness and hyperplane identities
\begin{align}
u_a&=\frac{a-a^*}{\Delta(a)}\,,
 &\|u_a\|_{\mathcal S}&\le\varsigma^{-1}\,,
 &\theta^\top u_a&=1\qquad(a\ne a^*)\,.
 \label{eq:SHr-projective-boundedness}
\end{align}
The closure in \eqref{eq:SH-projective} in \Cref{def:D-projective} is therefore compact and lies
in $\theta^\top u=1$. To check convexity directly, take positive-gap
actions $a_1,a_2$ and $\lambda\in[0,1]$, and define
\begin{align}
c&=\frac\lambda{\Delta(a_1)}+\frac{1-\lambda}{\Delta(a_2)}>0\,,
 &w_1&=\frac{\lambda}{c\Delta(a_1)}\,,
 &w_2&=\frac{1-\lambda}{c\Delta(a_2)}\,,
       \label{eq:D-step-0100}\\
 a&=w_1a_1+w_2a_2\in\mathcal K\,,
 &w_1+w_2&=1\,,
 &\Delta(a)&=w_1\Delta(a_1)+w_2\Delta(a_2)=1/c\,,
       \label{eq:D-step-0101}\\
 \frac{a-a^*}{\Delta(a)}
 &=c\{w_1(a_1-a^*)+w_2(a_2-a^*)\}
       \label{eq:D-step-0102}\\
 &=\lambda\frac{a_1-a^*}{\Delta(a_1)}
        +(1-\lambda)\frac{a_2-a^*}{\Delta(a_2)}\,.
 \label{eq:SHr-projective-convexity}
\end{align}
Equation \eqref{eq:D-step-0100} defines inverse-gap reweighting of the convex coefficients.
Equation \eqref{eq:D-step-0101} substitutes those weights and uses the affine identity for $\Delta$.
Equation \eqref{eq:D-step-0102} substitutes $\Delta(a)=1/c$ into the normalized vector difference $(a-a^*)/\Delta(a)$.

Equation \eqref{eq:SHr-projective-convexity} substitutes $cw_1=\lambda/\Delta(a_1)$ and $cw_2=(1-\lambda)/\Delta(a_2)$.
Thus the set before closure is convex, and so is its closure.

Let $\mathcal S^* =\Conv\{\pm(a-a^*):a\in\mathcal K\}$, a symmetric convex
body in $\mathsf D$, and define the finite analysis constant
\begin{equation}
\label{eq:SH-Lambda}
 \Lambda=\max\left\{1,\sup_{u\in\mathcal U_\theta}\norm u_{\mathcal S},
                       \sup_{u\in\mathcal U_\theta}\norm u_{\mathcal S^*}\right\}\,,
 \qquad b_\theta=C_{\beta,\mathrm{sh}}\Lambda^2\,.
\end{equation}
The gauge $\|\cdot\|_{\mathcal S^*}$ appearing in
\eqref{eq:SH-Lambda} is a norm on $\mathsf D$. For any $L\in\R^d$ it gives
\begin{equation}
\label{eq:SH-initial-extension}
 \sup_{u\in\mathcal D_\theta}|u^\top L|
 \le\Lambda\max_{a\in\mathcal K}|(a-a^*)^\top L|\,.
\end{equation}
Indeed, since $\mathcal S^*$ is the symmetric convex hull of the vector differences
$a-a^*$ for $a\in\mathcal K$, its support function at $L$ is the maximum absolute
value of $(a-a^*)^\top L$ over $a\in\mathcal K$. More explicitly,
\begin{align}
h_{\mathcal S^*}(L)
 &=\max_{a\in\mathcal K}|(a-a^*)^\top L|\,,
       \label{eq:D-projective-support}\\
 |u^\top L|&\le\|u\|_{\mathcal S^*}h_{\mathcal S^*}(L)
 \le\Lambda\max_{a\in\mathcal K}|(a-a^*)^\top L|
       \qquad(u\in\mathcal D_\theta)\,,\label{eq:D-projective-score-bound}\\
 \sup_{u\in\mathcal D_\theta}|u^\top L|
 &\le\Lambda\max_{a\in\mathcal K}|(a-a^*)^\top L|\,.
       \label{eq:D-projective-supremum-bound}
\end{align}
Equation \eqref{eq:D-projective-support} evaluates a linear functional
on a symmetric convex hull. Equation \eqref{eq:D-projective-score-bound}
uses polar duality and the bound $\|u\|_{\mathcal S^*}\le\Lambda$,
which extends from $\mathcal U_\theta$ by convexity.
Taking the supremum gives \eqref{eq:D-projective-supremum-bound}.
These support-function bounds hold for ambient $L\in\R^d$ because
all tested directions belong to $\mathsf D$; no restriction on the
algorithm's cumulative parameter estimate is required.
The body $\mathcal S^*$ spans $\mathsf D$; its gauge is consequently
continuous there, so its maximum on the compact $\mathcal U_\theta$
is finite. None of $\theta,\Lambda,\mathcal U_\theta$ is used by the algorithm.

\end{proof}

Fix an endpoint $t$, set $m=\lceil\sqrt t\rceil$, and freeze the function
${\mathsf Z}_s(u)=u^\top \sum_{j=1}^{s-1}\widehat\theta_j$ at the first $s\ge m$ for which $b_{r_s}\ne a^*$.
The function is frozen \emph{before} that round's update. Denote the frozen
function by $\bar {\mathsf Z}_s$, and let $\bar N_s=\max_{\mathcal D_\theta}(-\bar {\mathsf Z}_s)$.
The symbol ${\mathsf Z}_s$ denotes a linear function on normalized
directions, whereas $Z_s(x)$ in \eqref{eq:SH-score} is the algorithm's
nonnegative cumulative loss-difference score on actions.
\begin{lemma}[Homogeneous constraints on normalized directions]\label{lem:D-normalized-interface}
On $b_{r_s}=a^*$, for every $u\in\mathcal D_\theta$,
\begin{equation}\label{eq:SH-normalized-increments}
\begin{split}
 |u^\top v_{X_s}(p_s)Y_s|&\le\frac{\beta_s}{2}\norm u_{\mathcal S}
                                      +({\mathsf Z}_s(u))_+/B\,,\\
 \norm{u^\top v_{X_s}(p_s)Y_s}_{L^2(\cF_{s-1})}
 &\le\frac{\sqrt{\beta_s}}2\norm u_{\mathcal S}
                                      +({\mathsf Z}_s(u))_+/\sqrt{Bs}\,.
\end{split}
\end{equation}

\end{lemma}
\begin{proof}
The homogeneous extension in \eqref{eq:SH-normalized-increments} can
be verified without a positive minimum gap. If $b_{r_s}=a^*$ and
$\Delta(a)>0$, divide the magnitude bound and the square root of the
second-moment bound in \eqref{eq:SH-budgets} in
\GIAFTRL{} by $\Delta(a)$. For the latter,
use $\sqrt{a^2+b^2}\le a+b$ for nonnegative scalar arguments. This gives
\begin{align}
\left|\left(\frac{a-a^*}{\Delta(a)}\right)^\top v_{X_s}(p_s)Y_s\right|
 &\le\frac{\beta_s}{2}\frac{\|a-a^*\|_{\mathcal S}}{\Delta(a)}
       +\frac{[\widehat L_{s-1,a}-\widehat L_{s-1,a^*}]_+}{B\Delta(a)}
       \label{eq:D-step-0103}\\
 &=\frac{\beta_s}{2}\|u_a\|_{\mathcal S}
       +\frac{({\mathsf Z}_s(u_a))_+}{B}\,,
       \label{eq:D-step-0104}\\
 \|u_a^\top v_{X_s}(p_s)Y_s\|_{L^2(\cF_{s-1})}
 &\le\frac{\sqrt{\beta_s}}2
              \frac{\|a-a^*\|_{\mathcal S}}{\Delta(a)}
       +\frac{[\widehat L_{s-1,a}-\widehat L_{s-1,a^*}]_+}{\sqrt{Bs}\,\Delta(a)}
       \label{eq:D-step-0105}\\
 &=\frac{\sqrt{\beta_s}}2\|u_a\|_{\mathcal S}
       +\frac{({\mathsf Z}_s(u_a))_+}{\sqrt{Bs}}\,.
 \label{eq:SHr-normalized-direct-substitution}
\end{align}
Equation \eqref{eq:D-step-0103} applies the mark constraint at reference $a^*$ and divides its two terms by the positive gap $\Delta(a)$.
Equation \eqref{eq:D-step-0104} substitutes $u_a=(a-a^*)/\Delta(a)$ and ${\mathsf Z}_s(u_a)=X_s(a)/\Delta(a)$.
Equation \eqref{eq:D-step-0105} takes the square root of the conditional second-moment budget, bounds it by the sum of the two nonnegative component square roots, and divides by $\Delta(a)$.

Equation \eqref{eq:SHr-normalized-direct-substitution} makes the same normalized-direction substitutions in that conditional $L^2$ bound.
Every $u\in\mathcal U_\theta$ is a limit of such $u_a$.
Continuity of the linear maps and norms extends the inequalities to
that closure; conditional Fatou also gives the same second-moment
extension. Since $\mathcal U_\theta$ is convex, every element of
$\mathcal D_\theta$ is $\gamma u$ for some
$\gamma\in[0,1]$, $u\in\mathcal U_\theta$. Positive homogeneity gives
\begin{align}
|(\gamma u)^\top v_{X_s}(p_s)Y_s|&=\gamma|u^\top v_{X_s}(p_s)Y_s|\,,
 &\|\gamma u\|_{\mathcal S}&=\gamma\|u\|_{\mathcal S}\,,
       \label{eq:D-step-0106}\\
 ({\mathsf Z}_s(\gamma u))_+&=\gamma({\mathsf Z}_s(u))_+\,,
 &\|(\gamma u)^\top v_{X_s}(p_s)Y_s\|_{L^2(\cF_{s-1})}
 &=\gamma\|u^\top v_{X_s}(p_s)Y_s\|_{L^2(\cF_{s-1})}\,.
 \label{eq:SHr-normalized-positive-scaling}
\end{align}
Equation \eqref{eq:D-step-0106} uses linearity of the reconstruction coefficient and positive homogeneity of the gauge.
Equation \eqref{eq:SHr-normalized-positive-scaling} uses $\gamma\ge0$ to commute positive parts and $L^2$ norms with multiplication by $\gamma$.
This proves \eqref{eq:SH-normalized-increments} on the entire cone
section, including $u=0$.

Since ${\mathsf Z}_+\le {\mathsf Z}+2\bar N_s$ and $\norm u_{\mathcal S}\le\Lambda$, the
abstract interface of \Cref{lem:SH-field-interface} applies on
$\mathcal D_\theta$ with $\beta'_s=\Lambda^2\beta_s$ and $b'=b_\theta$.
Its drift is $\theta^\top u\in[0,1]$, and the function at zero is zero.

\end{proof}

\begin{definition}[Normalized windows, good events, and their constants]\label{def:D-local-window}
The following constants expose the full initial-history cost:
\begin{equation}\label{eq:SH-local-constants}
\begin{split}
 C_{\mathrm{init}}&=(\Lambda R)^P\,,\qquad
 G_\theta=\max\{C_G(b_\theta),C_{\mathrm{init}}\}\,,\qquad R_\theta=8G_\theta^{1/P}\,,\\
 q_0&=\frac{P(d+v)}{P+d}\,,\\
 A_1&=\max_{0\le j\le d}(2^P4^j)^{P/(P+j)}
                  R_\theta^{Pj/(P+j)}H_F(b_\theta)^{P/(P+j)}\,,\\
 A_2&=\max_{0\le j\le d}(2^P4^j)^{P/(P+j)}
                  R_\theta^{Pj/(P+j)}(2^vC_{\mathrm{init}})^{P/(P+j)}\,.
\end{split}
\end{equation}
They are fixed instance constants, not functions of the analysis endpoint.

For the rest of the local proof use a deterministic plateau within the
endpoint's window:
\begin{equation}\label{eq:SH-local-window}
 \tau_s=s^{7/8}+4t^{5/8}\quad(m\le s\le t)\,,\qquad
 I_s=\ind\{b_{r_j}=a^*,\ N_j^{\mathcal U}\le\tau_j
                         \text{ for all }m\le j\le s\}\,,
\end{equation}
where $N_j^{\mathcal U}=\max_{\mathcal D_\theta}(-{\mathsf Z}_j)$ is the actual
maximum negative normalized loss estimate while the window is alive.
The event $I_s=1$ means that the reference and every previous check on
negative normalized loss estimates in this window have remained valid.
Because these checks use only $\sum_{\ell=1}^{j-1}\widehat\theta_\ell$, the specific indicator $I_s$ is
$\cF_{s-1}$-measurable and satisfies $I_{s+1}\le I_s$.
Set $k_0=1/4096$ and define
\begin{equation}\label{eq:SH-local-threshold}
\begin{split}
 T_\theta=\left\lceil\max\left\{
 16,(T_A+1)^2,[S(b_\theta)+1]^2,5^8,
 [4(\Lambda R+2)]^{2/(1-k_0)},\right.\right.\\
 \left.\left.
 16^{1/(3/8-k_0/2)},[40(P-1)b_\theta]^{8/3}
 \right\}\right\rceil\,.
\end{split}
\end{equation}
In particular $m\le t/2$, $m\ge T_A,S(b_\theta)$, and
$\tau_t/t\le5t^{-1/8}\le1$ for $t\ge T_\theta$.

Define the new relative good event
\begin{equation}\label{eq:SH-good}
 \mathcal G_t=\{I_t=1\}\cap
          \{\min_{u\in\mathcal U_\theta}{\mathsf Z}_t(u)\ge4\tau_t\}
       \quad(t\ge T_\theta)\,.
\end{equation}
It asks for separation of the \emph{normalized} function, not separation of
the cumulative loss estimates of all suboptimal physical actions from
that of $a^*$ by a common positive amount.
Set
\begin{equation}\label{eq:SH-bad-constant}
 A_{\mathrm{pos}}=32^p(R_\theta+10)^d5^{p-d}\,,\qquad
 C_{\mathrm{bad}}=2A_A+2^{\kappa P+1}A_1+A_2+A_{\mathrm{pos}}\,.
\end{equation}
\end{definition}

\begin{lemma}[Uniform relative gaps and integrable exceptional histories]\label{lem:SH-ratios}
For $t\ge T_\theta$, $\sP(\mathcal G_t^c)\le C_{\mathrm{bad}}t^{-2}$.
On $\mathcal G_t$ define
\[
 \lo_t=\max\{1,\sup_{u\in\mathcal U_\theta}t/{\mathsf Z}_t(u)\}\,,\qquad
 \up_t=\sup_{u\in\mathcal U_\theta}{\mathsf Z}_t(u)/t\,,
\]
and set both to zero elsewhere. Then
\begin{equation}\label{eq:SH-ratios}
 \norm{\lo_t}_{L^{P/4}}\le C_{\mathrm{rat}}\,,\quad
 \norm{\up_t}_{L^{P/4}}\le C_{\mathrm{rat}}\,,\quad
 \E[\lo_t^j\up_t]\le C_{\mathrm{rat}}^{j+1}\quad(j=1,3,5)\,.
\end{equation}
For all $a\in\mathcal K$, on this event,
\begin{equation}\label{eq:SH-gap-comparison}
 \Delta(a)\le\lo_t\widehat\Delta_t(a)
       \le\lo_t\up_t\Delta(a)\,,\qquad
 X_t(a)\ge4\tau_t\Delta(a)\,.
\end{equation}
\end{lemma}
\begin{proof}[Proof of \Cref{lem:SH-ratios}]
\textbf{Uniform positivity on the projective body.}
Let $\tau=\tau_t$, $\bar R_t=\max_{\mathcal D_\theta}|\bar {\mathsf Z}_t|$,
and let $\sigma$ be normalized relative volume on
$\mathcal U_\theta$, whose affine dimension is $j\le d$.
On $\{I_t=1\}$, the function $Y(u)={\mathsf Z}_t(u)+2\tau$ is positive.
If $\min {\mathsf Z}_t<4\tau$, then $\min Y<6\tau$.
Contract $\mathcal U_\theta$ toward a minimizer of $Y$ by
$\lambda=\tau/(\bar R_t+10\tau)$. Since
$\max Y\le\bar R_t+2\tau$, the image has $Y\le8\tau$.
The volume and moment calculation is therefore
\begin{align}
\ind\{I_t=1,\ \min {\mathsf Z}_t<4\tau\}
 &\le8^p\tau^{p-j}(\bar R_t+10\tau)^j
                     I_t\int Y(u)^{-p}\sigma(du)\,,
       \label{eq:D-step-0107}\\
 \sP(I_t=1,\ \min {\mathsf Z}_t<4\tau)
 &\le8^p\tau^{p-j}
       \big[\E[(\bar R_t+10\tau)^{2j}]\big]^{1/2}
       \left[\int\E[I_tY(u)^{-2p}]\sigma(du)\right]^{1/2}
       \label{eq:D-step-0108}\\
 &\le8^p\tau^{p-j}[(R_\theta+10)t]^j(4/t)^p
       \label{eq:D-step-0109}\\
 &=32^p(R_\theta+10)^j(\tau/t)^{p-j}
       \label{eq:D-step-0110}\\
 &\le32^p(R_\theta+10)^d(5t^{-1/8})^{p-d}
 =A_{\mathrm{pos}}t^{-(p-d)/8}\,.
 \label{eq:SHr-projective-minimum-tail}
\end{align}
Equation \eqref{eq:D-step-0107} contracts the normalized direction body toward a minimizer with ${\mathsf Z}_t<4\tau$, and integrates $Y^{-p}$ over the resulting sublevel set.
Equation \eqref{eq:D-step-0108} takes expectation, applies Cauchy--Schwarz, and uses Jensen for the normalized volume integral.
Equation \eqref{eq:D-step-0109} uses $2j\le P$, the stopped absolute-moment bound, $\tau\le t$, and the inverse local moment with $2p=P$.

Equation \eqref{eq:D-step-0110} collects the factors $8^p4^p=32^p$ and the powers of $t$ and $\tau$.
Equation \eqref{eq:SHr-projective-minimum-tail} uses $j\le d$, $\tau/t\le5t^{-1/8}\le1$, and the definition of $A_{\mathrm{pos}}$.
Adding \eqref{eq:SH-window-failure} in \Cref{lem:SH-local} gives the complete failure bound
\begin{align}
\sP(\mathcal G_t^c)
 &\le2A_At^{-(a_A-1)/2}
       +2^{\kappa P+1}A_1t^{-\eta/2}
       +A_2t^{-P/8+q_0/2}
       +A_{\mathrm{pos}}t^{-(p-d)/8}
       \label{eq:D-step-0111}\\
 &\le[2A_A+2^{\kappa P+1}A_1+A_2+A_{\mathrm{pos}}]t^{-2}
 =C_{\mathrm{bad}}t^{-2}\,.
 \label{eq:SHr-relative-good-probability}
\end{align}
Equation \eqref{eq:D-step-0111} adds the window-failure probability \eqref{eq:SH-window-failure} in \Cref{lem:SH-local} and the normalized-minimum failure \eqref{eq:SHr-projective-minimum-tail}.
Equation \eqref{eq:SHr-relative-good-probability} uses the exponent margins in \Cref{lem:D-numerical-margins} and collects the coefficients into $C_{\mathrm{bad}}$.
Every exponent in equation \eqref{eq:D-step-0111} is at least two:
$d/P<1/4096$, $v/P<1/2048$, $q_0/P<3/4096$,
$a_A>16$, and $\eta/P=7/8-\kappa>1/9$ suffice.

\paragraph{Integral Moments and the Two Uniform Ratios}
On $\mathcal G_t$, set $f(u)={\mathsf Z}_t(u)/t$,
$R_f=\max_{\mathcal U_\theta}f$, and $m_f=\min_{\mathcal U_\theta}f$.
The defining lower bound ${\mathsf Z}_t\ge4\tau$ gives
$0<{\mathsf Z}_t\le Y\le2{\mathsf Z}_t$. Tonelli and the pointwise local moments imply
\begin{align}
\E\!\left[\ind_{\mathcal G_t}\int f^P d\sigma\right]
 &\le\int\E[I_t(Y/t)^P]d\sigma\le8^P\,,
       \label{eq:D-step-0112}\\
 \E\!\left[\ind_{\mathcal G_t}\int f^{-P}d\sigma\right]
 &\le(2t)^P\int\E[I_tY^{-P}]d\sigma
 \le(2t)^P(4/t)^P=8^P\,,
       \label{eq:D-step-0113}\\
 \E[\ind_{\mathcal G_t}R_f^P]
 &\le2^P4^j\E\!\left[\ind_{\mathcal G_t}\int f^Pd\sigma\right]
 \le2^P4^j8^P\le64^P\,.
 \label{eq:SHr-ratio-integrals}
\end{align}
Equation \eqref{eq:D-step-0112} uses $0<{\mathsf Z}_t\le Y$, $f={\mathsf Z}_t/t$, Tonelli, and the positive local moment.
Equation \eqref{eq:D-step-0113} uses $Y\le2{\mathsf Z}_t$, applies Tonelli to the inverse integral, and substitutes $\E[I_tY^{-P}]\le(4/t)^P$.
Equation \eqref{eq:SHr-ratio-integrals} applies \eqref{eq:C-aff-abs}
in \Cref{lem:C-affine} to $f$ on the projective body,
and uses $j\le d\le P$ to obtain $2^P4^j8^P\le64^P$.
The inverse inequality \eqref{eq:C-aff-inverse} in \Cref{lem:C-affine} gives, on the same event,
\begin{align}
m_f^{-(P-j)}&\le2^PR_f^j\int f^{-P}d\sigma\,.
 \label{eq:SHr-inverse-affine-applied}
\end{align}
Let $s_0=P/4$ and $a=s_0/(P-j)$.
Since $a+ja/P\le1$, generalized H\"older, including a final
indicator factor when the sum is strictly below one, yields
\begin{align}
\E[\ind_{\mathcal G_t}m_f^{-s_0}]
 &\le2^{Pa}\E\!\left[
       (\ind_{\mathcal G_t}R_f^P)^{ja/P}
       \left(\ind_{\mathcal G_t}\int f^{-P}d\sigma\right)^a
       \ind_{\mathcal G_t}^{1-a-ja/P}\right]
       \label{eq:D-inverse-holder-factors}\\
 &\le2^{Pa}
       (\E[\ind_{\mathcal G_t}R_f^P])^{ja/P}
       \left(\E\!\left[\ind_{\mathcal G_t}\int f^{-P}d\sigma\right]\right)^a
       (\sP(\mathcal G_t))^{1-a-ja/P}
       \label{eq:D-inverse-holder-applied}\\
 &\le2^{Pa}(64^P)^{ja/P}(8^P)^a
 =2^{Pa}64^{ja}8^{Pa}\,,\label{eq:D-inverse-holder-moments}\\
 \|\ind_{\mathcal G_t}m_f^{-1}\|_{L^{s_0}}
 &\le(2^{Pa}64^{ja}8^{Pa})^{1/s_0}
 =2^{P/(P-j)}64^{j/(P-j)}8^{P/(P-j)}\,,
       \label{eq:D-inverse-holder-root}\\
 \log\!\left[2^{P/(P-j)}64^{j/(P-j)}8^{P/(P-j)}\right]
 &=\frac{P\log16+j\log64}{P-j}
 =\log16+\frac{j/P}{1-j/P}\log1024\label{eq:D-inverse-holder-logarithm}\\
 &<\log16+\frac{\log1024}{4095}<\log18\,.
       \label{eq:SHr-inverse-ratio-holder}
\end{align}
Equation \eqref{eq:D-inverse-holder-factors} raises
\eqref{eq:SHr-inverse-affine-applied} to $a=s_0/(P-j)$ and inserts the
indicator into each factor.
Equation \eqref{eq:D-inverse-holder-applied} applies generalized
H\"older with reciprocal exponents $ja/P$, $a$, and $1-a-ja/P$;
zero reciprocal exponents are omitted.
Equation \eqref{eq:D-inverse-holder-moments} substitutes both integral
moment bounds and drops the probability factor at most one.
Equation \eqref{eq:D-inverse-holder-root} takes the $s_0$-th root
and substitutes $a/s_0=1/(P-j)$.
Equation \eqref{eq:D-inverse-holder-logarithm} evaluates the logarithm
exactly.  Equation \eqref{eq:SHr-inverse-ratio-holder} uses
$j/P<1/4096$ and $\log(18/16)>\log1024/4095$.
For the final numerical inequality, let $x=j/P<1/4096$; the logarithm
of its left coefficient is
$\log16+[x/(1-x)]\log1024<\log18$.
The case $j=0$ uses the same expressions with the maximum factor absent.
Consequently
\begin{align}
\|\lo_t\|_{L^{s_0}}
 &=\|\ind_{\mathcal G_t}\max\{1,m_f^{-1}\}\|_{L^{s_0}}
 \le1+18\le C_{\mathrm{rat}}\,,
       \label{eq:D-step-0114}\\
 \|\up_t\|_{L^{s_0}}
 &=\|\ind_{\mathcal G_t}R_f\|_{L^{s_0}}
 \le\|\ind_{\mathcal G_t}R_f\|_{L^P}\le64=C_{\mathrm{rat}}\,,
       \label{eq:D-step-0115}\\
 \E[\lo_t^r\up_t]
 &\le\|\lo_t\|_{L^{s_0}}^r\|\up_t\|_{L^{s_0}}
 \le C_{\mathrm{rat}}^{r+1}\qquad(r=1,3,5)\,.
 \label{eq:SHr-ratio-products}
\end{align}
Equation \eqref{eq:D-step-0114} substitutes the definition of $\lo_t$ and uses the triangle inequality and the inverse-minimum norm bound $18$.
Equation \eqref{eq:D-step-0115} substitutes the definition of $\up_t$ and uses monotonicity of $L^p$ norms and the maximum-moment bound.
Equation \eqref{eq:SHr-ratio-products} applies H\"older to $\lo_t^r\up_t$ with the unused probability factor bounded by one, using $r+1\le s_0$.
The product inequality uses $r+1\le6\le s_0$ and H\"older with
exponents $s_0/r$, $s_0$, and the remaining constant factor.

\paragraph{Clipping the Physical Gaps}
For $a\ne a^*$, $u_a=(a-a^*)/\Delta(a)\in\mathcal U_\theta$.
Direct substitution gives
\begin{align}
\frac{X_t(a)}t&=\Delta(a)\frac{{\mathsf Z}_t(u_a)}t\,,
 &\frac{\Delta(a)}{\lo_t}&\le\frac{X_t(a)}t\le\up_t\Delta(a)\,,
       \label{eq:D-step-0116}\\
 \lo_t\widehat\Delta_t(a)
 &=\lo_t\min\{2,X_t(a)/t\}
 =\min\{2\lo_t,\lo_tX_t(a)/t\}
       \label{eq:D-step-0117}\\
 &\ge\Delta(a)\,,
 &\lo_t\widehat\Delta_t(a)&\le\lo_tX_t(a)/t
                      \le\lo_t\up_t\Delta(a)\,,
       \label{eq:D-step-0118}\\
 X_t(a)&=\Delta(a){\mathsf Z}_t(u_a)\ge4\tau_t\Delta(a)\,.
 \label{eq:SHr-clipped-gap-proof}
\end{align}
Equation \eqref{eq:D-step-0116} substitutes the identity relating cumulative loss differences to their gap-normalized estimates and the definitions of the lower and upper ratios.
Equation \eqref{eq:D-step-0117} substitutes the clipped gap and distributes the positive factor $\lo_t$ over the minimum.
Equation \eqref{eq:D-step-0118} uses $2\lo_t\ge2\ge\Delta(a)$ and $\lo_tX_t(a)/t\ge\Delta(a)$, then bounds the minimum by its unclipped term.

Equation \eqref{eq:SHr-clipped-gap-proof} substitutes the good-event normalized minimum ${\mathsf Z}_t(u_a)\ge4\tau_t$.
The lower clipped bound uses both $\Delta(a)\le2$ and $\lo_t\ge1$.
At $a=a^*$, $X_t(a^*)=\widehat\Delta_t(a^*)=\Delta(a^*)=0$.
This proves every assertion in \eqref{eq:SH-ratios} and
\eqref{eq:SH-gap-comparison}.
\end{proof}

\begin{lemma}[Alive-window and pointwise normalized bounds]\label{lem:SH-local}
For $t\ge T_\theta$,
\begin{equation}\label{eq:SH-window-failure}
 \sP(I_t=0)\le
 2A_A t^{-(a_A-1)/2}
 +2^{\kappa P+1}A_1t^{-\eta/2}
 +A_2t^{-P/8+q_0/2}\,.
\end{equation}
For every $u\in\mathcal U_\theta$, let
$Y_s^\dagger(u)={\mathsf Z}_s(u)+2\tau_s$. Then
\begin{equation}\label{eq:SH-local-moments}
 \E[I_t(Y_t^\dagger(u))^{-P}]\le(4/t)^P\,,
 \qquad \E[I_t(Y_t^\dagger(u)/t)^P]\le8^P\,.
\end{equation}
\end{lemma}
\begin{proof}[Proof of \Cref{lem:SH-local}]
\textbf{Window failures.}
Let $m=\lceil\sqrt t\rceil$. By
\Cref{lem:SH-anchor}, the wrong-reference probability is
\begin{align}
\sP(\exists s\in[m,t]:b_{r_s}\ne a^*)
 &\le\sum_{s=m}^\infty A_As^{-a_A}
       \label{eq:D-step-0119}\\
 &\le A_A\left[m^{-a_A}+\int_m^\infty x^{-a_A}dx\right]
 \le2A_Am^{1-a_A}\le2A_At^{-(a_A-1)/2}\,.
 \label{eq:SHr-local-reference-union}
\end{align}
Equation \eqref{eq:D-step-0119} applies the union bound over reference times and then \eqref{eq:SH-anchor-tail} in \Cref{lem:SH-anchor}.
Equation \eqref{eq:SHr-local-reference-union} compares the decreasing power sum to its integral, uses $a_A>2$, and substitutes $m\ge\sqrt t$.
For $v_j=2^jm\le t$, freeze the normalized function before a wrong
reference or the first crossing of the level
$v_j^{7/8}+4t^{5/8}$ in $[v_j,2v_j]$.
A genuine threshold failure in this block implies that the frozen
negative maximum at $2v_j$ exceeds this level. Substituting
\eqref{eq:SH-normalized-negative} in \Cref{lem:SH-normalized-moments} into Markov gives
\begin{align}
\sP(\text{a negative crossing in }[v_j,2v_j]
                 \text{ before a wrong reference})
 &\le\frac{A_1(2v_j)^{\kappa P}
            +A_2m^{P-q_0}(2v_j)^{q_0}}
                {(v_j^{7/8}+4t^{5/8})^P}
       \label{eq:D-step-0120}\\
 &\le2^{\kappa P}A_1v_j^{-\eta}
       +\frac{2^{q_0}A_2m^{P-q_0}v_j^{q_0}}{4^Pt^{5P/8}}\,.
 \label{eq:SHr-local-block-crossing}
\end{align}
Equation \eqref{eq:D-step-0120} applies Markov to the field frozen at the first pre-update crossing and uses \eqref{eq:SH-normalized-negative} in \Cref{lem:SH-normalized-moments} at the deterministic time $2v_j$.
Equation \eqref{eq:SHr-local-block-crossing} uses the two lower bounds $v_j^{7/8}$ and $4t^{5/8}$ on the denominator separately for the two numerator terms.
Using $\eta>1$, $q_0\ge1$, and $\sqrt t\le m\le2\sqrt t$,
\begin{align}
\sum_{j\ge0}2^{\kappa P}A_1(2^jm)^{-\eta}
 &\le2^{\kappa P+1}A_1m^{-\eta}
 \le2^{\kappa P+1}A_1t^{-\eta/2}\,,
       \label{eq:D-step-0121}\\
 \sum_{j:v_j\le t}
 \frac{2^{q_0}A_2m^{P-q_0}v_j^{q_0}}{4^Pt^{5P/8}}
 &\le\frac{2^{q_0+1}A_2m^{P-q_0}t^{q_0}}{4^Pt^{5P/8}}
       \label{eq:D-step-0122}\\
 &\le2^{1-P}A_2t^{-P/8+q_0/2}
 \le A_2t^{-P/8+q_0/2}\,,
       \label{eq:D-step-0123}\\
 \sP(I_t=0)&\le2A_At^{-(a_A-1)/2}
       +2^{\kappa P+1}A_1t^{-\eta/2}
       +A_2t^{-P/8+q_0/2}\,.
 \label{eq:SHr-local-window-total}
\end{align}
Equation \eqref{eq:D-step-0121} evaluates the decreasing dyadic series and uses $\eta>1$ and $m\ge\sqrt t$.
Equation \eqref{eq:D-step-0122} sums the increasing dyadic powers with $q_0\ge1$, so their sum is at most $2t^{q_0}$.
Equation \eqref{eq:D-step-0123} substitutes $m\le2\sqrt t$, collects $2^{1-P}$ and $t^{-P/8+q_0/2}$, and uses $2^{1-P}\le1$.

Equation \eqref{eq:SHr-local-window-total} adds the two negative-crossing sums to the reference-error sum \eqref{eq:SHr-local-reference-union}.
A terminal block may extend past $t$ in the same stationary experiment;
it does not alter the learner.

\paragraph{Pointwise Compensated Moments}
Fix $u\in\mathcal U_\theta$. Set $Y_s={\mathsf Z}_s(u)+2\tau_s$,
$\beta_s'=\Lambda^2\beta_s$, and
$\delta_s=(s+1)^{7/8}-s^{7/8}$. On $I_s=1$,
\begin{align}
Y_s&\ge\tau_s\,,\qquad {\mathsf Z}_s(u)+2N_s^{\mathcal U}\le Y_s\,,
 &\E_{s-1}[u^\top v_{X_s}(p_s)Y_s^{\rm obs}]&=\theta^\top u=1\,,
       \label{eq:D-step-0124}\\
 \xi_s\coloneq Y_{s+1}-Y_s&=u^\top v_{X_s}(p_s)Y_s^{\rm obs}+2\delta_s\,,
 &\E_{s-1}[\xi_s]&=1+2\delta_s\,,
       \label{eq:D-step-0125}\\
 |\xi_s|&\le3Y_s/B\,,
 &\E_{s-1}[\xi_s^2]&\le10\beta_s'+2Y_s^2/(Bs)\,.
 \label{eq:SHr-local-increments}
\end{align}
Equation \eqref{eq:D-step-0124} uses the alive-event lower bound on ${\mathsf Z}_s$, the definition $Y_s={\mathsf Z}_s+2\tau_s$, and normalized drift $\theta^\top u=1$.
Equation \eqref{eq:D-step-0125} subtracts consecutive definitions of $Y_s$ and takes conditional expectation.
Equation \eqref{eq:SHr-local-increments} uses the normalized update magnitude and variance constraints and the small-step inequalities in \Cref{lem:D-numerical-margins}.
Here $Y_s^{\rm obs}$ denotes the observed feedback in
$\widehat\theta_s=v_{X_s}(p_s)Y_s^{\rm obs}$ in \GIAFTRL{}, only to distinguish it from the positive
shifted variable $Y_s$ in this proof. Equation \eqref{eq:SHr-local-increments} follows by
substituting the homogeneous budgets into
\eqref{eq:SHr-inverse-increments} in \Cref{lem:SH-global-inverse}, with drift $1$, buffer
$s^{7/8}+4t^{5/8}$, and $\beta_s'$ in place of $\beta_s$.
The plateau $4t^{5/8}$ has zero increment; $s\ge S(b_\theta)$ gives
all small-increment inequalities in \eqref{eq:SHr-inverse-taylor} in \Cref{lem:SH-global-inverse}.
In particular, next-step positivity holds before killing. The resulting
killed inverse recurrence and its solution are explicitly
\begin{align}
u_s^-&\coloneq \E[I_sY_s^{-P}]\,,\label{eq:D-clean-local-killed-definition}\\
 \E_{s-1}[Y_{s+1}^{-P}]
 &\le Y_s^{-P}-PY_s^{-P-1}(1+2\delta_s)
   +P(P+1)Y_s^{-P-2}
                 \left(10\beta_s'+\frac{2Y_s^2}{Bs}\right)
       \label{eq:D-clean-local-inverse-taylor-substitution}\\
 &\le(1+P/(2s))Y_s^{-P}-PY_s^{-P-1}
       \qquad(I_s=1)\,,\label{eq:D-clean-local-inverse-conditional}\\
 u_{s+1}^-
 &=\E[I_{s+1}Y_{s+1}^{-P}]
 \le\E[I_s\E_{s-1}[Y_{s+1}^{-P}]]
       \label{eq:D-clean-local-inverse-tower}\\
 &\le(1+P/(2s))u_s^--P\E[I_sY_s^{-P-1}]
 \le(1+P/(2s))u_s^--P(u_s^-)^{1+1/P}\,,
       \label{eq:D-clean-local-inverse-jensen}\\
 (u_{s+1}^-)^{-1/P}
 &\ge\frac{s}{s+1}(u_s^-)^{-1/P}+1\,,
       \label{eq:D-clean-local-root-recursion}\\
 (u_t^-)^{-1/P}
 &\ge\frac mt(u_m^-)^{-1/P}
       +\sum_{s=m}^{t-1}\prod_{j=s+1}^{t-1}\frac j{j+1}
 \ge\frac1t\sum_{s=m}^{t-1}(s+1)
       \label{eq:D-clean-local-root-unrolled}\\
 &=\frac{t(t+1)-m(m+1)}{2t}\ge t/4\,,
 \qquad\E[I_tY_t^{-P}]\le(4/t)^P\,.
       \label{eq:SHr-local-inverse-solution}
\end{align}
Equation \eqref{eq:D-clean-local-killed-definition} defines the
moment with the predictable alive indicator.
Equation \eqref{eq:D-clean-local-inverse-taylor-substitution}
substitutes the displayed drift and second moment into Taylor's
formula, using $(1-3/B)^{-P-2}\le2$.
Equation \eqref{eq:D-clean-local-inverse-conditional} uses the
numeric bounds in \Cref{lem:D-numerical-margins} and
drops $-2P\delta_sY_s^{-P-1}\le0$.
Equation \eqref{eq:D-clean-local-inverse-tower} uses
$I_{s+1}\le I_s$ and $I_s\in\cF_{s-1}$.
Equation \eqref{eq:D-clean-local-inverse-jensen} uses
$(I_sY_s^{-P})^{1+1/P}=I_sY_s^{-P-1}$ and Jensen.
Equation \eqref{eq:D-clean-local-root-recursion} applies the
convex tangent inequality $(1+x)^{-1/P}\ge1-x/P$ to the preceding
recurrence, as expanded in \Cref{lem:SH-global-inverse}.
Equation \eqref{eq:D-clean-local-root-unrolled} iterates that
root recurrence and evaluates its telescoping products.
Equation \eqref{eq:SHr-local-inverse-solution} evaluates the arithmetic sum and uses
$m\le t/2$; zero killed moments already satisfy the bound and are
never inverted.
The root calculation is \eqref{eq:SHr-inverse-root-solution} in \Cref{lem:SH-global-inverse} with
$\Delta=1$; zero killed moments require no inversion.

For the positive moment, let $w_s=(\E[I_sY_s^P])^{1/P}$.
Taylor's formula and $(1+3/B)^{P-2}\le2$ give
\begin{align}
\E[I_{s+1}Y_{s+1}^P]
 &\le\E\!\left[I_s\left(
Y_s^P+P(1+2\delta_s)Y_s^{P-1}
       +P(P-1)Y_s^{P-2}\left(10\beta_s'+\frac{2Y_s^2}{Bs}\right)
                         \right)\right]
       \label{eq:D-step-0126}\\
 &\le w_s^P+Pw_s^{P-1}
       \left[\frac{2(P-1)}{Bs}w_s+1+2\delta_s
                               +\frac{10(P-1)\beta_s'}{\tau_s}\right]\,,
       \label{eq:D-step-0127}\\
 w_{s+1}&\le(1+k_0/s)w_s+1+2\alpha s^{\alpha-1}
                         +10(P-1)b_\theta s^{1/2-\alpha}\,,
 \qquad\alpha=7/8\,.
 \label{eq:SHr-local-positive-root}
\end{align}
Equation \eqref{eq:D-step-0126} applies Taylor to $Y_{s+1}^P$, multiplies by the predictable alive indicator, and inserts the conditional mean and second moment of the increment.
Equation \eqref{eq:D-step-0127} uses $Y_s^{P-2}\le Y_s^{P-1}/\tau_s$ on survival and H\"older for the remaining moments, with $w_s^P=\E[I_sY_s^P]$.
Equation \eqref{eq:SHr-local-positive-root} uses $(x+h)^P\ge x^P+Px^{P-1}h$ for $h\ge0$, $2(P-1)/B\le k_0$, $\delta_s\le\alpha s^{\alpha-1}$, and $\beta_s^\prime/\tau_s\le b_\theta s^{1/2-\alpha}$.
If $w_s=0$, then $Y_s\ge\tau_s>0$ on survival implies
$\sP(I_s=1)=0$, and all subsequent killed moments are zero.
The multiplicative coefficient and the scalar sums needed for this
recursion are, for $m\le s<t$ and $0<r<1$,
\begin{align}
\prod_{j=s+1}^{t-1}(1+k_0/j)
 &\le\exp\!\left(k_0\sum_{j=s+1}^{t-1}j^{-1}\right)
 \le(t/s)^{k_0}\,,\label{eq:D-local-root-product}\\
 \sum_{s=m}^{t-1}s^{r-1}
 &\le\int_{m-1}^{t-1}x^{r-1}\,dx
 \le\int_0^t x^{r-1}\,dx=t^r/r\,.
       \label{eq:D-local-power-integral}
\end{align}
Equation \eqref{eq:D-local-root-product} uses $\log(1+x)\le x$ and
$\sum_{j=s+1}^{t-1}j^{-1}\le\int_s^{t-1}dx/x$.
Equation \eqref{eq:D-local-power-integral} uses monotonicity of
$x^{r-1}$. The three values used below are
$r=1-k_0$, $r=\alpha-k_0$, and $r=3/2-\alpha-k_0$, all in $(0,1)$.
Unrolling the positive-root recursion now gives
\begin{align}
w_t&\le2(t/m)^{k_0}w_m
   +2t^{k_0}\sum_{s=m}^{t-1}
     \left[s^{-k_0}+2\alpha s^{\alpha-1-k_0}
             +10(P-1)b_\theta s^{1/2-\alpha-k_0}\right]
       \label{eq:D-step-0128}\\
 &\le2(t/m)^{k_0}w_m
       +\frac2{1-k_0}t+\frac{4\alpha}{\alpha-k_0}t^\alpha
       +\frac{20(P-1)b_\theta}{3/2-\alpha-k_0}t^{3/2-\alpha}
       \label{eq:D-step-0129}\\
 &\le2(t/m)^{k_0}w_m+3t+5t^{7/8}
                             +40(P-1)b_\theta t^{5/8}\,.
 \label{eq:SHr-local-positive-product}
\end{align}
Equation \eqref{eq:D-step-0128} unrolls the affine recursion for $w_s$ and uses the product bound \eqref{eq:D-local-root-product}.
Equation \eqref{eq:D-step-0129} applies \eqref{eq:D-local-power-integral} with $r=1-k_0,\alpha-k_0,3/2-\alpha-k_0$ to the three sums.
Equation \eqref{eq:SHr-local-positive-product} substitutes $\alpha=7/8$, $k_0=1/4096$ and bounds the three coefficients by $3$, $5$ and $40(P-1)b_\theta$.
Each power sum is bounded by its integral from zero to $t$.
The full-history initialization obeys
\begin{align}
w_m&\le\|{\mathsf Z}_m(u)\|_{L^P}+2m^{7/8}+8t^{5/8}
 \le(\Lambda R+2)m+8t^{5/8}\,,
       \label{eq:D-step-0130}\\
 \frac{w_t}{t}
 &\le3+4(\Lambda R+2)t^{-(1-k_0)/2}
       +16t^{-3/8+k_0/2}+5t^{-1/8}
       +40(P-1)b_\theta t^{-3/8}
       \label{eq:D-step-0131}\\
 &\le3+1+1+1+1\le8\,.
 \label{eq:SHr-local-positive-final}
\end{align}
Equation \eqref{eq:D-step-0130} uses Minkowski, the full-history normalized absolute moment, and $m^{7/8}\le m$.
Equation \eqref{eq:D-step-0131} substitutes the bound for $w_m$ into \eqref{eq:SHr-local-positive-product}, divides by $t$, and uses $\sqrt t\le m\le2\sqrt t$.
Equation \eqref{eq:SHr-local-positive-final} uses each term in the deterministic threshold $T_{\mathrm{pos}}$ so that the four displayed error terms are each at most one.
The four error terms in \eqref{eq:D-step-0131} are bounded
individually by the corresponding entries of \eqref{eq:SH-local-threshold} in \Cref{def:D-local-window}.
Raising \eqref{eq:SHr-local-positive-final} to the power $P$ gives
$\E[I_tY_t^P]\le(8t)^P$, the positive-moment conclusion in
\eqref{eq:SH-local-moments}.
\end{proof}

\begin{lemma}[Finite sum of reference-error probabilities]\label{lem:SH-anchor}
For every $t\ge T_A$,
\begin{equation}\label{eq:SH-anchor-tail}
 \sP(b_{r_t}\ne a^*)\le A_A t^{-a_A}\,.
\end{equation}
Since $a_A>16$, the total error probability after $T_A$ satisfies
\[
 \sum_{t\ge T_A}\sP(b_{r_t}\ne a^*)
 \le A_A\sum_{t\ge T_A}t^{-a_A}\le2A_A<\infty\,.
\]
\end{lemma}
\begin{proof}[Proof of \Cref{lem:SH-anchor}]
If $r_t=j$, its defining penalized minimization gives
\begin{align}
\widehat L_{t-1,b_j}+\beta_t h_j
 &\le \widehat L_{t-1,a^*}+\beta_t h^*\,,
       \label{eq:D-step-0132}\\
 X_t(b_j)&\le\beta_t(h^*-h_j)\le\beta_t h^*\,.
 \label{eq:SH-reference-ineq}
\end{align}
Equation \eqref{eq:D-step-0132} evaluates the penalized reference minimization at the optimal atom with priority $h^*$.
Equation \eqref{eq:SH-reference-ineq} subtracts $\widehat L_{t-1,a^*}$ and uses $h_j\ge0$.
Set $\varepsilon_t=t^{-1/16}$, $\tau=t^{7/8}$, and use the alive
indicator of \Cref{lem:SH-global-inverse}. The deterministic
threshold $t\ge T_A$ implies $\beta_t h^*\le\tau$.

\paragraph{Reference Atoms with Gap at Least $\varepsilon_t$}
Let $D_t=\mathcal K\cap\{\Delta\ge\varepsilon_t\}$, with normalized
relative volume $\sigma$, and let $j=\dim\operatorname{aff}(D_t)\le d$.
An empty $D_t$ contributes probability zero. Otherwise, on the event
$E_t=\{I_t=1,\ b_{r_t}\ne a^*,\ \Delta(b_{r_t})\ge\varepsilon_t\}$,
the positive affine function $Y(a)=X_t(a)+2\tau$ satisfies
\begin{align}
\min_{D_t}Y&\le Y(b_{r_t})\le\beta_t h^*+2\tau\le3\tau\,,
 &\max_{D_t}Y&\le R_t^X+2\tau\,,
       \label{eq:D-step-0133}\\
 \lambda&=\frac{\tau}{R_t^X+3\tau}\,,
 &Y(a_0+\lambda(a-a_0))&\le3\tau+\lambda(R_t^X+2\tau)\le4\tau\,,
 \label{eq:SHr-anchor-contraction}
\end{align}
Equation \eqref{eq:D-step-0133} evaluates $Y=X_t+2\tau$ at the incorrectly selected reference and uses $\beta_th^*\le\tau$.
Equation \eqref{eq:SHr-anchor-contraction} defines the contraction factor and applies affinity of $Y$ to the segment from its minimizer.
where $a_0$ minimizes $Y$ on $D_t$. The homothetic image has relative
volume $\lambda^j$. Thus, with killed integrals defined as zero off
$\{I_t=1\}$,
\begin{align}
\ind_{E_t}
 &\le4^p\tau^{p-j}(R_t^X+3\tau)^j
                 I_t\int_{D_t}Y(a)^{-p}\sigma(da)\,,
       \label{eq:D-step-0134}\\
 \sP(E_t)
 &\le4^p\tau^{p-j}
       \big[\E[(R_t^X+3\tau)^{2j}]\big]^{1/2}
       \left[\E[\left(I_t\int Y^{-p}d\sigma\right)^2]\right]^{1/2}
       \label{eq:D-step-0135}\\
 &\le4^p\tau^{p-j}[(R+3)t]^j
       \left[\int\E[I_tY(a)^{-2p}]\sigma(da)\right]^{1/2}
       \label{eq:D-step-0136}\\
 &\le4^p\tau^{p-j}[(R+3)t]^j
             \left(\frac4{t\varepsilon_t}\right)^p
       \label{eq:D-step-0137}\\
 &=16^p(R+3)^j\varepsilon_t^{-p}(\tau/t)^{p-j}
       \label{eq:D-step-0138}\\
 &\le16^p(R+3)^d t^{-P/32+d/8}\,.
 \label{eq:SHr-anchor-large-gap-probability}
\end{align}
Equation \eqref{eq:D-step-0134} integrates $Y^{-p}$ over the contracted sublevel set of relative volume $\lambda^j$.
Equation \eqref{eq:D-step-0135} takes expectation and applies Cauchy--Schwarz to the range factor and the inverse integral.
Equation \eqref{eq:D-step-0136} uses $2j\le P$, the global absolute moment, $\tau\le t$, and Jensen for the normalized inverse integral.

Equation \eqref{eq:D-step-0137} applies the stopped global inverse bound with $2p=P$ and $\Delta(a)\ge\varepsilon_t$ on $D_t$.
Equation \eqref{eq:D-step-0138} collects the constants and powers without yet substituting the cutoff or buffer.
Equation \eqref{eq:SHr-anchor-large-gap-probability} substitutes $\varepsilon_t=t^{-1/16}$, $\tau=t^{7/8}$, and $j\le d$, giving exponent $-P/32+d/8$.
The contraction in \eqref{eq:D-step-0134} controls the entire
compact cap. No finite union bound over reference atoms is used.

\paragraph{Reference Atoms with Gap below $\varepsilon_t$}
Since $t\ge\Delta_{\rm pre}^{-16}$, every suboptimal atom in this range has
index $j>j^*$. The priority and sharpness give
\begin{align}
h_j&\ge\|b_j-a^*\|_{\mathcal S}^{-8}
 \ge\varsigma^8\Delta(b_j)^{-8}
 >\varsigma^8\sqrt t\,,
       \label{eq:D-step-0139}\\
 h_j-h^*&\ge\frac12\varsigma^8\sqrt t\,,
       \label{eq:D-step-0140}\\
 r_t=j&\ \Longrightarrow
       \label{eq:D-step-0141}\\
 M_t\ge-X_t(b_j)\ge\beta_t(h_j-h^*)\ge\tfrac12\varsigma^8t\,,
       \label{eq:D-step-0142}\\
 \sP(b_{r_t}\ne a^*,\ 0<\Delta(b_{r_t})<\varepsilon_t)
 &\le\sP(M_t\ge\tfrac12\varsigma^8t)
 \le\frac{\E[M_t^P]}{(\varsigma^8t/2)^P}
       \label{eq:D-step-0143}\\
 &\le2^PC_M\varsigma^{-8P}t^{-(1-\kappa)P}\,.
 \label{eq:SHr-anchor-small-gap-probability}
\end{align}
Equation \eqref{eq:D-step-0139} uses the reinforced priority term associated with $a^*$, then sharpness and $\Delta(b_j)<t^{-1/16}$.
Equation \eqref{eq:D-step-0140} uses the threshold $t\ge(2h^*/\varsigma^8)^2$.
Equation \eqref{eq:D-step-0141} restricts to the selected reference event.

Equation \eqref{eq:D-step-0142} applies the reference minimizing inequality, the preceding priority lower bound, and $\beta_t\ge\sqrt t$.
Equation \eqref{eq:D-step-0143} applies Markov to $M_t^P$ at the explicit threshold $\varsigma^8t/2$.
Equation \eqref{eq:SHr-anchor-small-gap-probability} substitutes \eqref{eq:SH-global-moments} in \Cref{lem:D-global-moments} and collects the power of $t$.
Adding the global-window failure and both gap ranges gives
\begin{align}
\sP(b_{r_t}\ne a^*)
 &\le A_Ht^{-\eta/2}
       +16^p(R+3)^dt^{-P/32+d/8}
       +2^PC_M\varsigma^{-8P}t^{-(1-\kappa)P}
       \label{eq:D-step-0144}\\
 &\le\big[A_H+16^p(R+3)^d+2^PC_M\varsigma^{-8P}\big]t^{-a_A}
 =A_At^{-a_A}\,.
 \label{eq:SHr-anchor-final}
\end{align}
Equation \eqref{eq:D-step-0144} adds window failure, the large-gap reference error, and the small-gap reference error.
Equation \eqref{eq:SHr-anchor-final} uses the minimum defining $a_A$ and the sum defining $A_A$ in \eqref{eq:SH-anchor-constants} in \Cref{def:D-reference-constants}.
This explicitly substitutes the definitions of $a_A$ and $A_A$ from
\eqref{eq:SH-anchor-constants} in \Cref{def:D-reference-constants}.
\end{proof}

\begin{lemma}[Propagation of normalized cumulative loss estimates]\label{lem:SH-normalized-moments}
For every $s\ge m$, and also after further freezing at a pre-update
threshold-crossing time,
\begin{align}
 \E\left[\max_{u\in\mathcal D_\theta}|\bar {\mathsf Z}_s(u)|^P\right]
       &\le R_\theta^P s^P\,,\label{eq:SH-normalized-absolute}\\
 \E[\bar N_s^P]&\le A_1s^{\kappa P}+A_2m^{P-q_0}s^{q_0}\,.
       \label{eq:SH-normalized-negative}
\end{align}
\end{lemma}
\begin{proof}[Proof of \Cref{lem:SH-normalized-moments}]
Use normalized relative volume $\sigma$ on $\mathcal D_\theta$, and set
$G_s^{\mathsf Z}=\int|\bar {\mathsf Z}_s|^P d\sigma$,
$F_s^{\mathsf Z}=\int(-\bar {\mathsf Z}_s)_+^P d\sigma$,
$R_s^{\mathsf Z}=\max|\bar {\mathsf Z}_s|$. At the deterministic starting time $m$,
\begin{align}
\E[G_m^{\mathsf Z}]&\le\E[\max_{u\in\mathcal D_\theta}\left|u^\top \sum_{j=1}^{m-1}\widehat\theta_j\right|^P]
 \le\Lambda^P\E[\max_{a\in\mathcal K}|X_m(a)|^P]
       \label{eq:D-step-0145}\\
 &\le(\Lambda R)^Pm^P=C_{\mathrm{init}}m^P\,,
 &\E[F_m^{\mathsf Z}]&\le C_{\mathrm{init}}m^P\,.
 \label{eq:SHr-normalized-initial}
\end{align}
Equation \eqref{eq:D-step-0145} bounds the integral by the maximum and applies the normalized-direction extension \eqref{eq:SH-initial-extension} in \Cref{lem:D-projective}.
Equation \eqref{eq:SHr-normalized-initial} substitutes the raw absolute moment, defines $C_{\mathrm{init}}=(\Lambda R)^P$, and uses $(-{\mathsf Z})_+^P\le|{\mathsf Z}|^P$.
This retains the entire history up to $m-1$.
Before freezing, \eqref{eq:SH-normalized-increments} in \Cref{lem:D-normalized-interface} verifies the
interface with $\beta_s'=\Lambda^2\beta_s$ and $b'=b_\theta$;
after freezing, \eqref{eq:SHr-frozen-recursions} in \Cref{lem:SH-field-interface} verifies the same
conditional recurrences. The supersolution induction in
\eqref{eq:SHr-G-induction} in \Cref{lem:SH-field-interface}, now initialized at $m$, gives
\begin{align}
\E[G_s^{\mathsf Z}]&\le G_\theta s^P\,,
 &G_\theta&=\max\{C_G(b_\theta),C_{\mathrm{init}}\}\,,
       \label{eq:D-step-0146}\\
 \E[(R_s^{\mathsf Z})^P]
 &\le2^P4^j\E[G_s^{\mathsf Z}]\le8^PG_\theta s^P
 =R_\theta^Ps^P\,,
 \label{eq:SHr-normalized-absolute-proof}
\end{align}
Equation \eqref{eq:D-step-0146} applies the absolute integral recurrence with $b^\prime=b_\theta$ and initial coefficient $C_{\mathrm{init}}$.
Equation \eqref{eq:SHr-normalized-absolute-proof} applies the affine maximum inequality and substitutes $R_\theta^P=8^PG_\theta$.
where $j=\dim\operatorname{aff}(\mathcal D_\theta)\le d$.

For the negative integral, \eqref{eq:SH-F-linear} in \Cref{lem:SH-field-interface} gives the explicit
recurrence and supersolution
\begin{align}
f_{s+1}\coloneq \E[F_{s+1}^{\mathsf Z}]
 &\le(1+v/s)f_s+J_F(b_\theta)s^{\nu-1}\,,
       \label{eq:D-step-0147}\\
 H_F(b_\theta)[(s+1)^\nu-s^\nu]
 &\ge\nu H_F(b_\theta)s^{\nu-1}
 \ge[vH_F(b_\theta)+J_F(b_\theta)]s^{\nu-1}\,,
       \label{eq:D-step-0148}\\
 f_s&\le H_F(b_\theta)s^\nu
       +C_{\mathrm{init}}m^P\prod_{i=m}^{s-1}(1+v/i)
       \label{eq:D-step-0149}\\
 &\le H_F(b_\theta)s^\nu
       +2^vC_{\mathrm{init}}m^{P-v}s^v\,.
 \label{eq:SHr-normalized-integral-negative}
\end{align}
Equation \eqref{eq:D-step-0147} takes expectation in the linearized shifted-negative recurrence.
Equation \eqref{eq:D-step-0148} uses convexity of $s\mapsto s^\nu$ and $H_F(b_\theta)\ge J_F(b_\theta)/(\nu-v)$.
Equation \eqref{eq:D-step-0149} compares to that deterministic supersolution and carries the full initial moment through the multiplicative product.

Equation \eqref{eq:SHr-normalized-integral-negative} uses $\prod_{i=m}^{s-1}(1+v/i)\le(2s/m)^v$ without discarding the initial-history contribution.
The inhomogeneous recurrence can be unrolled without dropping its
forcing term.  With $H_F=H_F(b_\theta)$ and $J_F=J_F(b_\theta)$,
\begin{align}
f_s&\le f_m\prod_{i=m}^{s-1}(1+v/i)
       +J_F\sum_{j=m}^{s-1}j^{\nu-1}
                     \prod_{i=j+1}^{s-1}(1+v/i)\,,\label{eq:D-negative-unrolled}\\
 H_F(s+1)^\nu&\ge(1+v/s)H_Fs^\nu+J_Fs^{\nu-1}\,,
       \label{eq:D-negative-particular-supersolution}\\
 J_F\sum_{j=m}^{s-1}j^{\nu-1}\prod_{i=j+1}^{s-1}(1+v/i)
 &\le H_Fs^\nu-H_Fm^\nu\prod_{i=m}^{s-1}(1+v/i)
 \le H_Fs^\nu\,,\label{eq:D-negative-forcing-sum}\\
 \prod_{i=m}^{s-1}(1+v/i)
 &\le\exp\!\left(v\sum_{i=m}^{s-1}i^{-1}\right)
 \le\exp\!\left(v\int_{m-1}^{s-1}x^{-1}dx\right)
 \le(2s/m)^v\,.\label{eq:D-negative-product-bound}
\end{align}
Equation \eqref{eq:D-negative-unrolled} repeatedly substitutes the
one-step affine recurrence.
Equation \eqref{eq:D-negative-particular-supersolution} uses
$H_F\ge J_F/(\nu-v)$ and convexity of $s^\nu$.
Equation \eqref{eq:D-negative-forcing-sum} telescopes the supersolution
inequality after multiplication by the corresponding products.
Equation \eqref{eq:D-negative-product-bound} uses $\log(1+x)\le x$,
the decreasing integrand, and $(s-1)/(m-1)\le2s/m$ for $m\ge2$.
Substitution of all four displays yields exactly
\eqref{eq:SHr-normalized-integral-negative}; the empty product is one.
Let $a_j=P/(P+j)$ and $c_j=2^P4^j$. The affine negative-maximum
inequality and H\"older yield, with the $j=0$ case understood directly,
\begin{align}
\bar N_s^{P+j}&\le c_j(R_s^{\mathsf Z})^jF_s^{\mathsf Z}\,,
       \label{eq:D-step-0150}\\
 \E[\bar N_s^P]
 &\le c_j^{a_j}
       \E\big[((R_s^{\mathsf Z})^P)^{j/(P+j)}(F_s^{\mathsf Z})^{a_j}\big]
       \label{eq:D-step-0151}\\
 &\le c_j^{a_j}\big[\E[(R_s^{\mathsf Z})^P]\big]^{j/(P+j)}
                         \big[\E[F_s^{\mathsf Z}]\big]^{a_j}
       \label{eq:D-step-0152}\\
 &\le c_j^{a_j}(R_\theta^Ps^P)^{j/(P+j)}
       \big[H_F(b_\theta)s^\nu
             +2^vC_{\mathrm{init}}m^{P-v}s^v\big]^{a_j}
       \label{eq:D-step-0153}\\
 &\le c_j^{a_j}R_\theta^{Pj/(P+j)}H_F(b_\theta)^{a_j}
                   s^{P(j+\nu)/(P+j)}
       \label{eq:D-step-0154}\\*
 &\quad+c_j^{a_j}R_\theta^{Pj/(P+j)}
                (2^vC_{\mathrm{init}})^{a_j}
                   m^{P(P-v)/(P+j)}s^{P(j+v)/(P+j)}\,.
 \label{eq:SHr-normalized-affine-conversion}
\end{align}
Equation \eqref{eq:D-step-0150} applies the affine negative-cap inequality at a fixed history.
Equation \eqref{eq:D-step-0151} raises the inequality to $a_j=P/(P+j)$ and writes the absolute factor as a power of $(R_s^{\mathsf Z})^P$.
Equation \eqref{eq:D-step-0152} applies H\"older with reciprocal exponents $j/(P+j)$ and $a_j$.

Equation \eqref{eq:D-step-0153} inserts both the absolute and the integral-negative envelopes, retaining their complete dependence on $m$.
Equation \eqref{eq:D-step-0154} uses $(x+y)^{a_j}\le x^{a_j}+y^{a_j}$ and writes the contribution of the particular solution.
Equation \eqref{eq:SHr-normalized-affine-conversion} writes the contribution of the propagated initial moment and collects both exponents.
In \eqref{eq:D-step-0152}, the H\"older exponents are
$(P+j)/j$ and $(P+j)/P$. Equations \eqref{eq:D-step-0154}--
\eqref{eq:SHr-normalized-affine-conversion} apply
$(x+y)^{a_j}\le x^{a_j}+y^{a_j}$.
Now
\begin{align}
\frac{P(j+\nu)}{P+j}
 &=P\left(\frac34+\frac{j}{4(P+j)}\right)\le\kappa P\,,
       \label{eq:D-step-0155}\\
 q_j\coloneq \frac{P(j+v)}{P+j}&\le q_0\,,
 &\frac{P(P-v)}{P+j}&=P-q_j\,,
       \label{eq:D-step-0156}\\
 m^{P-q_j}s^{q_j}&=m^P(s/m)^{q_j}
 \le m^P(s/m)^{q_0}=m^{P-q_0}s^{q_0}\,,
       \label{eq:D-step-0157}\\
 \E[\bar N_s^P]&\le A_1s^{\kappa P}+A_2m^{P-q_0}s^{q_0}\,.
 \label{eq:SHr-normalized-negative-final}
\end{align}
Equation \eqref{eq:D-step-0155} substitutes $\nu=3P/4$ and uses monotonicity of $j/(P+j)$ for $j\le d$.
Equation \eqref{eq:D-step-0156} defines $q_j$ and identifies the complementary exponent of $m$.
Equation \eqref{eq:D-step-0157} factors out $m^P$ before using $s/m\ge1$ and $q_j\le q_0$.

Equation \eqref{eq:SHr-normalized-negative-final} substitutes these power comparisons into \eqref{eq:SHr-normalized-affine-conversion} and takes the coefficient maxima $A_1,A_2$.
The maxima defining $A_1,A_2$ dominate the two coefficients in
\eqref{eq:SHr-normalized-affine-conversion}. Every line concerns frozen
functions at deterministic times, so additional predictable pre-update
freezing preserves the proof and both envelopes.
\end{proof}

\begin{lemma}[An inverse moment without any positive minimum gap]\label{lem:SH-global-inverse}
For a fixed endpoint $t\ge T_g$, set $m=\lceil\sqrt t\rceil$,
$\tau_s=s^{7/8}$, and
$I_s=\ind\{M_j\le\tau_j\text{ for }m\le j\le s\}$.
Then
\[
 \sP(I_t=0)\le A_Ht^{-\eta/2}\,,\qquad
 \E\left[I_t\{X_t(a)+2t^{7/8}\}^{-P}\right]
       \le\left(\frac4{t\Delta(a)}\right)^P
       \quad(\Delta(a)>0)\,.
\]
An inverse multiplied by a zero indicator is defined as zero.
\end{lemma}
\begin{proof}[Proof of \Cref{lem:SH-global-inverse}]
\textbf{The alive window.}
For $v_j=2^jm$, freeze the entire function $X_s$ before its first
update at a time $s\ge v_j$ with $M_s>v_j^{7/8}$.
On a crossing during $[v_j,2v_j]$, its frozen negative maximum at
$2v_j$ exceeds $v_j^{7/8}$. The stopped form of
\eqref{eq:SH-global-moments} in \Cref{lem:D-global-moments} and Markov therefore give
\begin{align}
\sP(\exists s\in[v_j,2v_j]:M_s>v_j^{7/8})
 &\le\frac{\E[\bar M_{2v_j}^P]}{v_j^{7P/8}}
 \le\frac{C_M(2v_j)^{\kappa P}}{v_j^{7P/8}}
 =2^{\kappa P}C_Mv_j^{-\eta}\,,
       \label{eq:D-step-0158}\\
 \sP(I_t=0)
 &\le\sum_{j\ge0}2^{\kappa P}C_M(2^jm)^{-\eta}
       \label{eq:D-step-0159}\\
 &=\frac{2^{\kappa P}C_M}{1-2^{-\eta}}m^{-\eta}
 \le2^{\kappa P+1}C_Mt^{-\eta/2}
 =A_Ht^{-\eta/2}\,.
 \label{eq:SHr-global-window-sum}
\end{align}
Equation \eqref{eq:D-step-0158} freezes at the first pre-update crossing, applies Markov at deterministic endpoint $2v_j$, and inserts the stopped global negative moment.
Equation \eqref{eq:D-step-0159} applies the union bound over the dyadic blocks $v_j=2^jm$.
Equation \eqref{eq:SHr-global-window-sum} evaluates the geometric series, uses $\eta>1$ and $m\ge\sqrt t$, and substitutes $A_H=2^{\kappa P+1}C_M$.
A failure of $M_s\le s^{7/8}$ implies a crossing of the smaller
threshold $v_j^{7/8}$ in a covering block. Here $\eta>1$ and
$m\ge\sqrt t$. The frozen-envelope argument is justified by
\eqref{eq:SHr-frozen-recursions} in \Cref{lem:SH-field-interface}.

\paragraph{Conditional Inverse Recursion}
Fix $a$ with $\Delta=\Delta(a)>0$, set
$Y_s=X_s(a)+2\tau_s$ and $\delta_s=\tau_{s+1}-\tau_s$.
On $\{I_s=1\}$, every quantity $I_s,Y_s,\tau_s$ is
$\cF_{s-1}$-measurable, and
\begin{align}
Y_s&\ge\tau_s>0\,,\qquad X_s(a)+2M_s\le Y_s\,,
 \qquad0\le\delta_s\le\tfrac78s^{-1/8}\,,
       \label{eq:D-step-0160}\\
 \xi_s\coloneq Y_{s+1}-Y_s&=D_s(a)+2\delta_s\,,
       \label{eq:D-step-0161}\\
 \E_{s-1}[\xi_s]&=\Delta+2\delta_s\,,
       \label{eq:D-step-0162}\\
 |\xi_s|&\le2\beta_s+Y_s/B+2\delta_s
 \le3\beta_s+Y_s/B\le3Y_s/B\,,
       \label{eq:D-step-0163}\\
 \E_{s-1}[\xi_s^2]
 &=\E_{s-1}[D_s(a)^2]+4\delta_s\Delta+4\delta_s^2
       \label{eq:D-step-0164}\\
 &\le8\beta_s+2Y_s^2/(Bs)+8\delta_s+4\delta_s^2
 \le10\beta_s+2Y_s^2/(Bs)\,.
 \label{eq:SHr-inverse-increments}
\end{align}
Equation \eqref{eq:D-step-0160} uses survival, $Y_s=X_s(a)+2\tau_s$, and concavity of $s^{7/8}$ for the deterministic buffer increment.
Equation \eqref{eq:D-step-0161} subtracts the definitions of $Y_{s+1}$ and $Y_s$.
Equation \eqref{eq:D-step-0162} uses unbiased loss differences under $\theta$ and the predictability of $\delta_s$.

Equation \eqref{eq:D-step-0163} inserts the all-history update magnitude, bounds $2\delta_s\le\beta_s$, and applies the small-step threshold.
Equation \eqref{eq:D-step-0164} expands the square of $D_s(a)+2\delta_s$ before taking conditional expectation.
Equation \eqref{eq:SHr-inverse-increments} inserts the conditional variance, $\Delta\le2$, and $8\delta_s+4\delta_s^2\le2\beta_s$.
The update magnitude bound uses $\beta_s\ge16$ and
$3\beta_s\le s^{7/8}/B$, which follows from $s\ge S(C_{\beta,\mathrm{sh}})$.
In particular $Y_{s+1}\ge(1-3/B)Y_s>0$ on $\{I_s=1\}$, even
when $I_{s+1}=0$.
Taylor's formula on this positive interval gives
\begin{align}
\E_{s-1}[Y_{s+1}^{-P}]
 &\le Y_s^{-P}-PY_s^{-P-1}(\Delta+2\delta_s)
 +\frac{P(P+1)}2(1-3/B)^{-P-2}Y_s^{-P-2}\E_{s-1}[\xi_s^2]
       \label{eq:D-step-0165}\\
 &\le Y_s^{-P}-P\Delta Y_s^{-P-1}
 +P(P+1)Y_s^{-P-2}\left(10\beta_s+\frac{2Y_s^2}{Bs}\right)
       \label{eq:D-step-0166}\\
 &\le\left(1+\frac{P}{2s}\right)Y_s^{-P}
                  -P\Delta Y_s^{-P-1}\,.
 \label{eq:SHr-inverse-taylor}
\end{align}
Equation \eqref{eq:D-step-0165} applies Taylor to $y\mapsto y^{-P}$ on $[Y_s(1-3/B),Y_s(1+3/B)]$ and inserts the conditional drift.
Equation \eqref{eq:D-step-0166} drops the nonpositive buffer-drift term, uses $(1-3/B)^{-P-2}\le2$, and substitutes the second moment.
Equation \eqref{eq:SHr-inverse-taylor} uses $10(P+1)\beta_s/Y_s^2\le1/(4s)$ and $2(P+1)/B\le1/4$ to collect the Taylor remainder as $P/(2s)$.
Define killed inverse moments as zero off the alive event:
$u_s=\E[I_sY_s^{-P}]$. Since $I_{s+1}\le I_s$, the tower property
and Jensen give
\begin{align}
u_{s+1}
 &\le\E[I_s\E_{s-1}[Y_{s+1}^{-P}]]
       \label{eq:D-step-0167}\\
 &\le\left(1+\frac P{2s}\right)u_s
           -P\Delta\E[I_sY_s^{-P-1}]
       \label{eq:D-step-0168}\\
 &\le\left(1+\frac P{2s}\right)u_s
           -P\Delta\big(\E[I_sY_s^{-P}]\big)^{1+1/P}
       \label{eq:D-step-0169}\\
 &=\left(1+\frac P{2s}\right)u_s-P\Delta u_s^{1+1/P}\,.
 \label{eq:SHr-inverse-killed-recursion}
\end{align}
Equation \eqref{eq:D-step-0167} uses $I_{s+1}\le I_s$ and conditions with the predictable indicator $I_s$.
Equation \eqref{eq:D-step-0168} inserts the conditional Taylor inequality before taking expectation.
Equation \eqref{eq:D-step-0169} applies Jensen to $(I_sY_s^{-P})^{1+1/P}=I_sY_s^{-P-1}$.

Equation \eqref{eq:SHr-inverse-killed-recursion} substitutes $u_s=\E[I_sY_s^{-P}]$.
Indeed $(I_sY_s^{-P})^{1+1/P}=I_sY_s^{-P-1}$.
If $u_s=0$, the alive event has probability zero and all later killed
moments are zero. Otherwise, while $u_{s+1}>0$, the bracket below is
positive and the tangent inequality for $(1+x)^{-1/P}$ gives
\begin{align}
u_{s+1}^{-1/P}
 &\ge u_s^{-1/P}
       \left(1+\frac P{2s}-P\Delta u_s^{1/P}\right)^{-1/P}
       \label{eq:D-step-0170}\\
 &\ge\left(1-\frac1{2s}\right)u_s^{-1/P}+\Delta
 \ge\frac{s}{s+1}u_s^{-1/P}+\Delta\,,
       \label{eq:D-step-0171}\\
 u_t^{-1/P}
 &\ge\frac mt u_m^{-1/P}
       +\Delta\sum_{s=m}^{t-1}\prod_{j=s+1}^{t-1}\frac{j}{j+1}
       \label{eq:D-step-0172}\\
 &=\frac mt u_m^{-1/P}+\frac\Delta t\sum_{s=m}^{t-1}(s+1)
 \ge\frac{\Delta}{2t}\{t(t+1)-m(m+1)\}
 \ge\frac{t\Delta}{4}\,.
 \label{eq:SHr-inverse-root-solution}
\end{align}
Equation \eqref{eq:D-step-0170} factors out $u_s$ and takes the decreasing power $-1/P$ on the nonnegative recurrence.
Equation \eqref{eq:D-step-0171} uses the tangent lower bound $(1+x)^{-1/P}\ge1-x/P$ and $1-1/(2s)\ge s/(s+1)$.
Equation \eqref{eq:D-step-0172} unrolls the resulting linear lower recursion, retaining the nonnegative initial term.

Equation \eqref{eq:SHr-inverse-root-solution} evaluates the telescoping product and the arithmetic sum and uses $m\le t/2$.
Taking reciprocals and the power $P$ in
\eqref{eq:SHr-inverse-root-solution} gives
$u_t\le(4/(t\Delta(a)))^P$. The calculation applies to each positive gap
separately and does not take their infimum.
\end{proof}

\begin{lemma}[One-step absolute integral moment]\label{lem:D-absolute-step}
Let ${\mathsf Z}_s$ be an adapted affine function on a compact convex body $D$
of affine dimension at most $d$, with a zero on $D$ at every history.
Let $N_s=\max_D(-{\mathsf Z}_s)$, $D_s={\mathsf Z}_{s+1}-{\mathsf Z}_s$, and
$G_s=\int_D|{\mathsf Z}_s|^P\,d\sigma$, where $\sigma$ is normalized relative
volume.  Suppose $|\E_{s-1}[D_s(a)]|\le2$ and the update magnitude and
second-moment assumptions of \Cref{lem:SH-field-interface}
hold with $\beta_s'\le b'\sqrt s$.  Then
\begin{equation}\label{eq:D-absolute-step-result}
\begin{split}
 \E_{s-1}[G_{s+1}]\le{}&(1+a_G/s)G_s+c_0G_s^{1-1/P}\\
 &+c_G(b')\sqrt s\,G_s^{1-2/P}+D_G(b')s^{(P-1)/2}\,.
\end{split}
\end{equation}
In particular, the conditional drift need not be nonnegative.
\end{lemma}
\begin{proof}

The integral $G_s$ measures the overall magnitude of the estimated
loss field; its bound will control maxima of linear cumulative loss estimates by convex geometry.
All integrals in this proof use normalized relative volume on the
body $D$, whose affine dimension is $j\le d$. Fix $\cF_{s-1}$ and let
${\mathsf Z}={\mathsf Z}_s$, $D_s={\mathsf Z}_{s+1}-{\mathsf Z}_s$, $N=\max_D(-{\mathsf Z})$,
$\beta=\beta_s'$, $A=G_s^{1/P}$, and $R_{\mathsf Z}=\max_D|{\mathsf Z}|$.
The body contains a zero of ${\mathsf Z}$, so $N\ge0$.
The tuning implies $P>d$, $B=2^{24}P$, and
\begin{align}
R_{\mathsf Z}&\le(2^P4^j)^{1/P}A\le8A\,,
       \label{eq:D-step-0173}\\
 |D_s(a)|&\le2\beta+({\mathsf Z}(a)+2N)/B\le2\beta+24A/B\,,
       \label{eq:D-step-0174}\\
 \E_{s-1}[D_s(a)^2]
 &\le8\beta+2({\mathsf Z}(a)+2N)^2/(Bs)
 \le8\beta+1152A^2/(Bs)\,.
 \label{eq:SHr-interface-absolute-input}
\end{align}
Equation \eqref{eq:D-step-0173} applies \eqref{eq:C-aff-abs} in
\Cref{lem:C-affine} pathwise, with
$A=(\int|{\mathsf Z}|^P\,d\sigma)^{1/P}$ and $j\le d$.
Equation \eqref{eq:D-step-0174} uses ${\mathsf Z}+2N\le3R_{\mathsf Z}\le24A$ in the update magnitude assumption.
Equation \eqref{eq:SHr-interface-absolute-input} makes the same substitution in the second moment, with $2\cdot24^2=1152$.

\paragraph{Absolute Moments}
Taylor's formula \eqref{eq:taylor-absolute} in \Cref{lem:taylor}, the drift bound
$|\E_{s-1}[D_s(a)]|\le2$, and H\"older--Minkowski give
\begin{align}
\int_D|{\mathsf Z}|^{P-1}\,d\sigma
 &\le\left(\int_D|{\mathsf Z}|^P\,d\sigma\right)^{(P-1)/P}=A^{P-1}\,,
       \label{eq:D-absolute-first-moment-holder}\\
 \left(\int_D(|{\mathsf Z}|+2\beta+24A/B)^{P-2}\,d\sigma\right)^{1/(P-2)}
 &\le\left(\int_D|{\mathsf Z}|^{P-2}\,d\sigma\right)^{1/(P-2)}+2\beta+24A/B
       \label{eq:D-absolute-curvature-minkowski}\\
 &\le A+2\beta+24A/B=(1+24/B)A+2\beta\,,
       \label{eq:D-absolute-curvature-holder}\\
 \E_{s-1}[G_{s+1}]
 &\le G_s+P\int_D|{\mathsf Z}|^{P-1}\operatorname{sgn}({\mathsf Z})
                                      \E_{s-1}[D_s]\,d\sigma\label{eq:D-absolute-taylor-with-drift}\\*
 &\,+\frac{P(P-1)}2\int_D
 (|{\mathsf Z}|+2\beta+24A/B)^{P-2}
            \left(8\beta+\frac{1152A^2}{Bs}\right)d\sigma
       \label{eq:D-step-0175}\\
 &\le\begin{aligned}[t]
 &G_s+2P A^{P-1}\\
 &+\frac{P^2}{2}[(1+24/B)A+2\beta]^{P-2}
                           \left(8\beta+\frac{1152A^2}{Bs}\right)\,.
 \end{aligned}\label{eq:SHr-interface-absolute-taylor}
\end{align}
Equation \eqref{eq:D-absolute-first-moment-holder} applies H\"older on the probability measure $\sigma$ to the $(P-1)$-st integral moment.
Equation \eqref{eq:D-absolute-curvature-minkowski} applies Minkowski to the $L^{P-2}(\sigma)$ norm of the Taylor radius.
Equation \eqref{eq:D-absolute-curvature-holder} uses monotonicity of integral $L^p$ norms and substitutes $A^P=\int|{\mathsf Z}|^P\,d\sigma$.

Equation \eqref{eq:D-absolute-taylor-with-drift} applies the absolute-power Taylor formula and integrates its conditional linear term.
Equation \eqref{eq:D-step-0175} keeps the full integrated Taylor remainder with the second-moment bound substituted.
Equation \eqref{eq:SHr-interface-absolute-taylor} uses $|\E_{s-1}[D_s]|\le2$, inserts all three integral bounds, and relaxes $P(P-1)$ to $P^2$.
For $A\ge8P\beta$, the curvature base is bounded by
\begin{align}
\big[(1+24/B)A+2\beta\big]^{P-2}
 &\le A^{P-2}\left(1+\frac{24}{B}+\frac1{4P}\right)^{P-2}
       \label{eq:D-step-0176}\\
 &\le A^{P-2}\exp\!\left(\frac{24P}{B}+\frac14\right)
 \le2A^{P-2}\,,
       \label{eq:D-step-0177}\\
 \frac{P^2}{2}\big[(1+24/B)A+2\beta\big]^{P-2}
       \left(8\beta+\frac{1152A^2}{Bs}\right)
 &\le\frac{1152P^2}{Bs}A^P+8P^2b'\sqrt s\,A^{P-2}\,.
 \label{eq:SHr-interface-absolute-large}
\end{align}
Equation \eqref{eq:D-step-0176} uses the branch condition $A\ge8P\beta$ to obtain $2\beta/A\le1/(4P)$.
Equation \eqref{eq:D-step-0177} uses $(1+x)^{P-2}\le e^{(P-2)x}$ and $24P/B+1/4<\log2$.
Equation \eqref{eq:SHr-interface-absolute-large} substitutes that bound into the Taylor remainder and uses $\beta\le b^\prime\sqrt s$.
For $A<8P\beta$, instead use
\begin{align}
(1+24/B)A+2\beta&\le12P\beta\,,
       \label{eq:D-step-0178}\\
 8\beta+\frac{1152A^2}{Bs}
 &\le8b'\sqrt s+\frac{2^{17}P^2(b')^2}{B}
 \le w_G(b')\sqrt s\,,
       \label{eq:D-step-0179}\\
 \frac{P^2}{2}\big[(1+24/B)A+2\beta\big]^{P-2}
       \left(8\beta+\frac{1152A^2}{Bs}\right)
 &\le\frac{P^2}{2}(12Pb')^{P-2}w_G(b')s^{(P-1)/2}
 =D_G(b')s^{(P-1)/2}\,.
 \label{eq:SHr-interface-absolute-small}
\end{align}
Equation \eqref{eq:D-step-0178} uses $A<8P\beta$ and $B\ge2^{24}P$ to bound the Taylor radius.
Equation \eqref{eq:D-step-0179} substitutes $A^2<64P^2\beta^2$, $\beta\le b^\prime\sqrt s$, and the definition of $w_G(b^\prime)$.
Equation \eqref{eq:SHr-interface-absolute-small} substitutes both small-branch bounds into the remainder and evaluates its power of $s$ and the constant $D_G(b^\prime)$.
Set $E_A=\{A\ge8P\beta\}$, an event determined by the fixed
history. The two remainder bounds enter the same Taylor inequality:
\begin{align}
 \E_{s-1}[G_{s+1}]
 &\le G_s+2PA^{P-1}
   +\ind_{E_A}\left[\frac{1152P^2}{Bs}A^P
                        +8P^2b'\sqrt s\,A^{P-2}\right]
       \label{eq:D-absolute-branch-direct}\\*
 &\quad+\ind_{E_A^c}D_G(b')s^{(P-1)/2}
       \label{eq:D-absolute-branch-small-added}\\
 &\le G_s+2P(G_s^{1/P})^{P-1}
       +\frac{1152P^2}{Bs}(G_s^{1/P})^P
       \label{eq:D-absolute-branch-A-substitute}\\*
 &\quad+8P^2b'\sqrt s\,(G_s^{1/P})^{P-2}
                            +D_G(b')s^{(P-1)/2}
       \label{eq:D-absolute-branch-A-substitute-rest}\\
 &=\left(1+\frac{1152P^2}{Bs}\right)G_s+2PG_s^{1-1/P}
       +8P^2b'\sqrt s\,G_s^{1-2/P}+D_G(b')s^{(P-1)/2}
       \label{eq:D-absolute-branch-powers}\\
 &=(1+a_G/s)G_s+c_0G_s^{1-1/P}
    +c_G(b')\sqrt s\,G_s^{1-2/P}+D_G(b')s^{(P-1)/2}\,.
       \label{eq:SHr-interface-absolute-rec}
\end{align}
Equations \eqref{eq:D-absolute-branch-direct}--
\eqref{eq:D-absolute-branch-small-added} substitute
\eqref{eq:SHr-interface-absolute-large} and
\eqref{eq:SHr-interface-absolute-small} into
\eqref{eq:SHr-interface-absolute-taylor} on disjoint events.
Equations \eqref{eq:D-absolute-branch-A-substitute}--
\eqref{eq:D-absolute-branch-A-substitute-rest} discard indicators only
from nonnegative contributions and substitute $A=G_s^{1/P}$ literally.
Equation \eqref{eq:D-absolute-branch-powers} evaluates all three powers;
\eqref{eq:SHr-interface-absolute-rec} substitutes
$a_G=1152P^2/B$, $c_0=2P$, and $c_G(b')=8P^2b'$ from
\eqref{eq:SH-moment-dictionary} in
\Cref{def:D-moment-dictionary}.
This proves \eqref{eq:D-absolute-step-result}.

\end{proof}

\begin{lemma}[One-step negative integral moment for shifted cumulative loss estimates]\label{lem:D-negative-step}
Let $D$ be a compact convex body of affine dimension at most $d$,
with normalized relative volume $\sigma$ (point mass in dimension zero).
Let ${\mathsf Z}_s$ be an adapted affine field on $D$, set
$D_s(a)={\mathsf Z}_{s+1}(a)-{\mathsf Z}_s(a)$, let $N_s=\max_D(-{\mathsf Z}_s)\ge0$, and set
$F_s=\int_D(-{\mathsf Z}_s)_+^P\,d\sigma$.  A zero of ${\mathsf Z}_s$ is not required.
Suppose $\E_{s-1}[D_s(a)]\ge0$ and
\begin{align}
 |D_s(a)|&\le3\beta_s'+[{\mathsf Z}_s(a)+2N_s]/B\,,\label{eq:D-negative-step-amp}\\
 \E_{s-1}[D_s(a)^2]&\le10\beta_s'
                   +2[{\mathsf Z}_s(a)+2N_s]^2/(Bs)\,,\label{eq:D-negative-step-var}
\end{align}
where $16\le\beta_s'\le b'\sqrt s$.  Then
\begin{equation}\label{eq:D-negative-step-result}
 \E_{s-1}[F_{s+1}]\le(1+a_F/s)F_s
   +c_F(b')\sqrt s\,F_s^{1-2/P}+D_F(b')s^{(P-1)/2}\,.
\end{equation}
Thus this result applies to a remaining-budget shift, whose drift is
nonnegative but whose value at $a^*$ can be strictly negative.
\end{lemma}
\begin{proof}
Fix the past and let ${\mathsf Z}={\mathsf Z}_s$, $D_s={\mathsf Z}_{s+1}-{\mathsf Z}_s$, $\beta=\beta_s'$,
$j=\dim\operatorname{aff}(D)\le d$, $f=-{\mathsf Z}$, $z=-D_s$, and
$F=F_s=\int f_+^P$, so $N=\max f_+$. The assumptions of this lemma are precisely
\begin{align}
\E_{s-1}[z]&\le0\,,
 &|z|&\le3\beta+(2N-f)/B\,,
 &\E_{s-1}[z^2]&\le10\beta+2(2N-f)^2/(Bs)\,.
 \label{eq:SHr-negative-interface}
\end{align}
These larger inputs will also be used for compensated functions below.
Define $c=(3\beta+2N/B)/(1-1/B)$. The level $-c$ is the lowest
value of $f$ from which a permitted increment can enter its positive
part. More precisely,
\begin{align}
c&=\frac{3\beta+2N/B}{1-1/B}
 \le4\beta+4N/B\,,\label{eq:D-negative-c-bound}\\
 f+|z|&\le f+3\beta+(2N-f)/B
       =(1-1/B)f+3\beta+2N/B
       =(1-1/B)(f+c)\,,\label{eq:D-negative-crossing-support}\\
 f\le-c&\ \Longrightarrow\ (f+|z|)_+^{P-2}=0\,,
       \label{eq:D-negative-no-crossing}\\
 \E_{s-1}[\int_D(f+z)_+^P\,d\sigma]
 &\le\int_D f_+^P\,d\sigma
       +P\int_D f_+^{P-1}\E_{s-1}[z]\,d\sigma
       \label{eq:D-negative-taylor-integral}\\*
 &\quad+\frac{P(P-1)}2\int_{\{f>-c\}}
           (f+c)_+^{P-2}\E_{s-1}[z^2]\,d\sigma\,,
       \label{eq:D-negative-taylor-remainder}\\
 P\int_D f_+^{P-1}\E_{s-1}[z]\,d\sigma&\le0\,.
       \label{eq:D-negative-linear-sign}
\end{align}
Equation \eqref{eq:D-negative-c-bound} uses $B\ge4$.
Equation \eqref{eq:D-negative-crossing-support} substitutes the
update magnitude inequality in \eqref{eq:SHr-negative-interface} before
factoring $1-1/B$. Equation \eqref{eq:D-negative-no-crossing} follows
because $P>2$. Equations \eqref{eq:D-negative-taylor-integral}--\eqref{eq:D-negative-taylor-remainder}
apply the positive-part Taylor formula on the only possible crossing
region and then condition and integrate.
Equation \eqref{eq:D-negative-linear-sign} uses the nonnegative
weight $f_+^{P-1}$ and $\E_{s-1}[z]\le0$.
At every contributing point $f>-c$,
\begin{align}
2N-f&\le2N+c\le3N+4\beta\,,
       \label{eq:D-step-0180}\\
 \E_{s-1}[z^2]
 &\le10\beta+\frac{2(3N+4\beta)^2}{Bs}
       \label{eq:D-step-0181}\\
 &\le10\beta+\frac{36N^2+64\beta^2}{Bs}
 \le v_F(b')\sqrt s+\frac{36N^2}{Bs}\,,
       \label{eq:D-step-0182}\\
 \E_{s-1}[F_{s+1}]
 &\le F+\frac{P^2}{2}
       \left(v_F(b')\sqrt s+\frac{36N^2}{Bs}\right)
                      \int(f+c)_+^{P-2}\,.
 \label{eq:SHr-negative-taylor}
\end{align}
Equation \eqref{eq:D-step-0180} uses $f>-c$ and $c\le4\beta+4N/B\le4\beta+N$.
Equation \eqref{eq:D-step-0181} substitutes this upper bound for $2N-f$ in the conditional second moment.
Equation \eqref{eq:D-step-0182} uses $(3N+4\beta)^2\le18N^2+32\beta^2$, $\beta\le b^\prime\sqrt s$, and $s\ge1$.

Equation \eqref{eq:SHr-negative-taylor} uses \eqref{eq:D-negative-taylor-integral}--\eqref{eq:D-negative-linear-sign}, drops the nonpositive linear term, and inserts the variance bound.
The linear Taylor term has nonpositive expectation by
\eqref{eq:SHr-negative-interface}.
If $N\ge16(P+j)\beta$, then
\begin{align}
(1+c/N)^{P-2+j}
 &\le\exp\!\left((P+j)(4\beta/N+4/B)\right)
 \le\exp(1/4+8P/B)\le2\,,
       \label{eq:D-step-0183}\\
 \int(f+c)_+^{P-2}
 &\le16(1+c/N)^{P-2+j}F/N^2\le32F/N^2\,,
       \label{eq:D-step-0184}\\
 \E_{s-1}[F_{s+1}]
 &\le F+\frac{576P^2}{Bs}F
       +16P^2v_F(b')\sqrt s\,F/N^2
       \label{eq:D-step-0185}\\
 &\le(1+a_F/s)F+c_F(b')\sqrt s\,F^{1-2/P}\,.
 \label{eq:SHr-negative-large}
\end{align}
Equation \eqref{eq:D-step-0183} uses $N\ge16(P+j)\beta$, $c\le4\beta+4N/B$, $j\le P$, and $(1+x)^r\le e^{rx}$.
Equation \eqref{eq:D-step-0184} applies the affine shifted-positive integral bound \eqref{eq:C-aff-shift} in \Cref{lem:C-affine} and the exponential factor just obtained.
Equation \eqref{eq:D-step-0185} inserts the resulting integral bound into \eqref{eq:SHr-negative-taylor} and evaluates $P^2\cdot36\cdot32/2=576P^2$.

Equation \eqref{eq:SHr-negative-large} uses $F\le N^P$, hence $F/N^2\le F^{1-2/P}$, and substitutes $a_F,c_F(b^\prime)$.
Here \eqref{eq:C-aff-shift} in \Cref{lem:C-affine} supplies equation \eqref{eq:D-step-0184}, and
$F\le N^P$ implies $F/N^2\le F^{1-2/P}$.
If $N<16(P+j)\beta\le32P\beta$, including $N=0$,
\begin{align}
(f+|z|)_+&\le N+3\beta+N/B\le36P\beta\,,
       \label{eq:D-step-0186}\\
 v_F(b')\sqrt s+36N^2/(Bs)&\le w_F(b')\sqrt s\,,
       \label{eq:D-step-0187}\\
 \E_{s-1}[F_{s+1}]-F
 &\le\frac{P^2}{2}(36Pb')^{P-2}w_F(b')s^{(P-1)/2}
 =D_F(b')s^{(P-1)/2}\,.
 \label{eq:SHr-negative-small}
\end{align}
Equation \eqref{eq:D-step-0186} uses $f\le N$, the increment bound, and $N<16(P+j)\beta\le32P\beta$.
Equation \eqref{eq:D-step-0187} substitutes that bound for $N$ and uses the definition of $w_F(b^\prime)$.
Equation \eqref{eq:SHr-negative-small} inserts both small-branch bounds into the Taylor remainder and substitutes $D_F(b^\prime)$.
Let $E_{\rm large}=\{N\ge16(P+j)\beta\}$, which is determined
by the past.  The two alternatives give the single conditional bound
\begin{align}
\E_{s-1}[F_{s+1}]
 &\le F+\ind_{E_{\rm large}}
       [a_FF/s+c_F(b')\sqrt s\,F^{1-2/P}]
       +\ind_{E_{\rm large}^c}D_F(b')s^{(P-1)/2}
       \label{eq:D-negative-branch-substitution}\\
 &\le(1+a_F/s)F+c_F(b')\sqrt s\,F^{1-2/P}
                          +D_F(b')s^{(P-1)/2}\,.
       \label{eq:D-negative-branch-assembly}
\end{align}
Equation \eqref{eq:D-negative-branch-substitution} inserts
\eqref{eq:SHr-negative-large} and \eqref{eq:SHr-negative-small}
on their respective disjoint events.
Equation \eqref{eq:D-negative-branch-assembly} drops the two
indicators multiplying nonnegative increments.  This proves
\eqref{eq:D-negative-step-result}.
This negative-part calculation only needs $N=\max(-{\mathsf Z})\ge0$;
it does not require the value at the distinguished point to equal zero.

\end{proof}

\begin{lemma}[An abstract moment interface for affine functions]\label{lem:SH-field-interface}
For a compact convex body of affine dimension at most $d$, containing a
point where an affine function ${\mathsf Z}_s$ is zero, suppose its increments have
conditional drift in $[0,2]$ and obey \eqref{eq:SH-global-amp}--
\eqref{eq:SH-global-var} in \Cref{lem:SH-feasible}, with $\beta_s$ replaced, if necessary, by any
$\beta'_s\ge16$ with $\beta'_s\le b'\sqrt s$.
Let $G_s=\int|{\mathsf Z}_s|^P$ and $F_s=\int(-{\mathsf Z}_s)_+^P$, with normalized relative
volume. Then
\begin{align}
 \E_{s-1}[G_{s+1}]&\le(1+a_G/s)G_s+c_0G_s^{1-1/P}
            +c_G(b')\sqrt s\,G_s^{1-2/P}+D_G(b')s^{(P-1)/2}\,,
            \label{eq:SH-G-rec}\\
 \E_{s-1}[F_{s+1}]&\le(1+a_F/s)F_s+c_F(b')\sqrt s\,F_s^{1-2/P}
            +D_F(b')s^{(P-1)/2}\,,\label{eq:SH-F-rec}\\
 \E_{s-1}[F_{s+1}]&\le(1+v/s)F_s+J_F(b')s^{\nu-1}\,.
            \label{eq:SH-F-linear}
\end{align}
Starting from zero at time one,
$\E[G_s]\le C_G(b')s^P$ and $\E[F_s]\le C_F(b')s^\nu$.
The same recurrences hold after freezing the function at a predictable
pre-update stopping time, provided they hold before freezing.
\end{lemma}
\begin{proof}[Proof of \Cref{lem:SH-field-interface}]
The absolute recurrence is \eqref{eq:D-absolute-step-result} in
\Cref{lem:D-absolute-step}.  The stronger original budgets
imply the assumptions of \Cref{lem:D-negative-step}:
\begin{align}
2\beta_s'+[{\mathsf Z}_s(a)+2N_s]/B
 &\le3\beta_s'+[{\mathsf Z}_s(a)+2N_s]/B\,,\label{eq:D-interface-relax-amp}\\
 8\beta_s'+2[{\mathsf Z}_s(a)+2N_s]^2/(Bs)
 &\le10\beta_s'+2[{\mathsf Z}_s(a)+2N_s]^2/(Bs)\,,\label{eq:D-interface-relax-var}\\
 \E_{s-1}[F_{s+1}]
 &\le(1+a_F/s)F_s+c_F(b')\sqrt s\,F_s^{1-2/P}
                             +D_F(b')s^{(P-1)/2}\,.\label{eq:D-interface-neg-applied}
\end{align}
Equations \eqref{eq:D-interface-relax-amp} and
\eqref{eq:D-interface-relax-var} add nonnegative allowances.
Equation \eqref{eq:D-interface-neg-applied} is the resulting
application of \Cref{lem:D-negative-step}, with the same
$F_s$, $\beta_s'$, and $b'$.
\paragraph{Deterministic Supersolutions and the Linear Recurrence}
Set $g_s=\E[G_s]$ and $f_s=\E[F_s]$. Conditional expectation followed by
Jensen replaces each power of $G_s$ or $F_s$ in the recurrences by
the same power of $g_s$ or $f_s$. Let $C_G=C_G(b')$, $C_F=C_F(b')$.
Their definitions give the separate coefficient bounds
The maximum defining $C_G=C_G(b')$ and $C_F=C_F(b')$ in
\Cref{def:D-moment-dictionary} gives, before any
multiplication by powers of these constants,
\begin{align}
C_G^{1/P}&\ge\frac{3c_0}{P-a_G}\,,\qquad
 C_G^{2/P}\ge\frac{3c_G(b')}{P-a_G}\,,\qquad
 C_G\ge\frac{3D_G(b')}{P-a_G}\,,
       \label{eq:D-induction-CG-lower}\\
 C_F^{2/P}&\ge\frac{2c_F(b')}{\nu-a_F}\,,\qquad
 C_F\ge\frac{2D_F(b')}{\nu-a_F}\,.
       \label{eq:D-induction-CF-lower}
\end{align}
All denominators are positive by
\Cref{lem:D-numerical-margins}.
Multiplication of \eqref{eq:D-induction-CG-lower} by
$C_G^{1-1/P},C_G^{1-2/P},1$, respectively, and of
\eqref{eq:D-induction-CF-lower} by $C_F^{1-2/P},1$, respectively,
yields
\begin{align}
c_0C_G^{1-1/P}&\le\tfrac13(P-a_G)C_G\,,
       \label{eq:D-step-0188}\\
 c_G(b')C_G^{1-2/P}&\le\tfrac13(P-a_G)C_G\,,
       \label{eq:D-step-0189}\\
 D_G(b')&\le\tfrac13(P-a_G)C_G\,,
       \label{eq:D-step-0190}\\
 c_F(b')C_F^{1-2/P}&\le\tfrac12(\nu-a_F)C_F\,,
       \label{eq:D-step-0191}\\
 D_F(b')&\le\tfrac12(\nu-a_F)C_F\,.
 \label{eq:SHr-induction-coefficients}
\end{align}
Equation \eqref{eq:D-step-0188} rearranges $C_G^{1/P}\ge3c_0/(P-a_G)$ from \eqref{eq:D-induction-CG-lower}.
Equation \eqref{eq:D-step-0189} rearranges $C_G^{2/P}\ge3c_G(b^\prime)/(P-a_G)$ from \eqref{eq:D-induction-CG-lower}.
Equation \eqref{eq:D-step-0190} rearranges $C_G\ge3D_G(b^\prime)/(P-a_G)$ from \eqref{eq:D-induction-CG-lower}.

Equation \eqref{eq:D-step-0191} rearranges $C_F^{2/P}\ge2c_F(b^\prime)/(\nu-a_F)$ from \eqref{eq:D-induction-CF-lower}.
Equation \eqref{eq:SHr-induction-coefficients} rearranges $C_F\ge2D_F(b^\prime)/(\nu-a_F)$ from \eqref{eq:D-induction-CF-lower}.
If $g_s\le C_Gs^P$, direct substitution into
\eqref{eq:SHr-interface-absolute-rec} in \Cref{lem:D-absolute-step} gives
\begin{align}
g_{s+1}
 &\le(1+a_G/s)C_Gs^P+c_0(C_Gs^P)^{1-1/P}
      +c_G(b')s^{1/2}(C_Gs^P)^{1-2/P}
      +D_G(b')s^{(P-1)/2}
       \label{eq:D-step-0192}\\
 &=C_Gs^P+(a_GC_G+c_0C_G^{1-1/P})s^{P-1}
      +c_G(b')C_G^{1-2/P}s^{P-3/2}
      +D_G(b')s^{(P-1)/2}
       \label{eq:D-step-0193}\\
 &\le C_Gs^P+
      [a_GC_G+c_0C_G^{1-1/P}+c_G(b')C_G^{1-2/P}+D_G(b')]s^{P-1}
       \label{eq:D-step-0194}\\
 &\le C_Gs^P+[a_GC_G+(P-a_G)C_G]s^{P-1}
       \label{eq:D-step-0195}\\
 &=C_Gs^P+PC_Gs^{P-1}\le C_G(s+1)^P\,.
 \label{eq:SHr-G-induction}
\end{align}
Equation \eqref{eq:D-step-0192} inserts the induction hypothesis $g_s\le C_Gs^P$ into every nondecreasing term of the absolute recurrence after Jensen.
Equation \eqref{eq:D-step-0193} distributes the powers of $C_Gs^P$ before simplifying their exponents.
Equation \eqref{eq:D-step-0194} uses $s\ge1$, $P-3/2\le P-1$, and $(P-1)/2\le P-1$.

Equation \eqref{eq:D-step-0195} inserts all three coefficient inequalities from \eqref{eq:SHr-induction-coefficients}.
Equation \eqref{eq:SHr-G-induction} collects the coefficient $P$ and uses convexity, $(s+1)^P\ge s^P+Ps^{P-1}$.
For $f_s\le C_Fs^\nu$, the corresponding calculation is
\begin{align}
f_{s+1}
 &\le(1+a_F/s)C_Fs^\nu
      +c_F(b')s^{1/2}(C_Fs^\nu)^{1-2/P}
      +D_F(b')s^{(P-1)/2}
       \label{eq:D-step-0196}\\
 &=C_Fs^\nu+a_FC_Fs^{\nu-1}
      +c_F(b')C_F^{1-2/P}s^{1/2+\nu(1-2/P)}
      +D_F(b')s^{(P-1)/2}
       \label{eq:D-step-0197}\\
 &\le C_Fs^\nu+[a_FC_F+c_F(b')C_F^{1-2/P}+D_F(b')]s^{\nu-1}
       \label{eq:D-step-0198}\\
 &\le C_Fs^\nu+[a_FC_F+(\nu-a_F)C_F]s^{\nu-1}
       \label{eq:D-step-0199}\\
 &=C_Fs^\nu+\nu C_Fs^{\nu-1}\le C_F(s+1)^\nu\,.
 \label{eq:SHr-F-induction}
\end{align}
Equation \eqref{eq:D-step-0196} inserts $f_s\le C_Fs^\nu$ into each nondecreasing term of the negative recurrence after Jensen.
Equation \eqref{eq:D-step-0197} expands the powers of $C_Fs^\nu$ without yet relaxing their exponents.
Equation \eqref{eq:D-step-0198} uses $1/2+\nu(1-2/P)=\nu-1$ and $(P-1)/2\le\nu-1$.

Equation \eqref{eq:D-step-0199} inserts both negative-moment coefficient inequalities from \eqref{eq:SHr-induction-coefficients}.
Equation \eqref{eq:SHr-F-induction} collects the coefficient $\nu$ and uses $(s+1)^\nu\ge s^\nu+\nu s^{\nu-1}$.
Here $1/2+\nu(1-2/P)=\nu-1$ and $(P-1)/2\le\nu-1$.
Convexity of $s\mapsto s^P$ and $s\mapsto s^\nu$ gives the final
inequalities. The inductions start from $g_1=f_1=0$.
For the linear form, let $x=F_s/s^\nu$, $\lambda=\lambda_F$, and
$R_F=R_F(b')$. Splitting at $x=R_F$ gives
\begin{align}
c_F(b')x^{1-2/P}
 &\le\lambda_Fx+c_F(b')R_F^{1-2/P}\,,
       \label{eq:D-step-0200}\\
 c_F(b')\sqrt s\,F_s^{1-2/P}
 &=c_F(b')s^{1/2+\nu(1-2/P)}x^{1-2/P}
       \label{eq:D-step-0201}\\
 &\le\lambda_FF_s/s+c_F(b')R_F^{1-2/P}s^{\nu-1}\,,
       \label{eq:D-step-0202}\\
 \E_{s-1}[F_{s+1}]
 &\le(1+(a_F+\lambda_F)/s)F_s
       +[c_F(b')R_F^{1-2/P}+D_F(b')]s^{\nu-1}
       \label{eq:D-step-0203}\\
 &=(1+v/s)F_s+J_F(b')s^{\nu-1}\,.
 \label{eq:SHr-negative-linear-derived}
\end{align}
Equation \eqref{eq:D-step-0200} splits at $x=R_F$: for $x\ge R_F$, $c_Fx^{1-2/P}\le\lambda_Fx$, and for $x<R_F$ it is at most $c_FR_F^{1-2/P}$.
Equation \eqref{eq:D-step-0201} substitutes $F_s=s^\nu x$ before collecting powers of $s$.
Equation \eqref{eq:D-step-0202} uses $1/2+\nu(1-2/P)=\nu-1$ and the preceding scalar bound.

Equation \eqref{eq:D-step-0203} inserts this bound in the negative-moment recurrence and uses $(P-1)/2\le\nu-1$.
Equation \eqref{eq:SHr-negative-linear-derived} substitutes $v=a_F+\lambda_F$ and $J_F(b^\prime)=c_F(b^\prime)R_F^{1-2/P}+D_F(b^\prime)$.
This is \eqref{eq:SH-F-linear}, with direct substitution of the
rescaling before its bound.

If the function is frozen before the update at a predictable time
$\sigma$, its increment is $\ind\{s<\sigma\}D_s$ and its next
integral equals its present integral on $\{s\ge\sigma\}$.
The positive coefficients in the recurrences make freezing harmless.
For example, the conditional absolute-moment inequality is
\begin{align}
\E_{s-1}[\bar G_{s+1}]
 &\le\bar G_s+\ind\{s<\sigma\}
       \left[\frac{a_G}{s}\bar G_s+c_0\bar G_s^{1-1/P}
        +c_G(b')\sqrt s\,\bar G_s^{1-2/P}
        +D_G(b')s^{(P-1)/2}\right]
       \label{eq:D-step-0204}\\
 &\le(1+a_G/s)\bar G_s+c_0\bar G_s^{1-1/P}
        +c_G(b')\sqrt s\,\bar G_s^{1-2/P}
        +D_G(b')s^{(P-1)/2}\,,
       \label{eq:D-step-0205}\\
 \E_{s-1}[\bar F_{s+1}]
 &\le\bar F_s+\ind\{s<\sigma\}
       \left[\frac{a_F}{s}\bar F_s
        +c_F(b')\sqrt s\,\bar F_s^{1-2/P}
        +D_F(b')s^{(P-1)/2}\right]
       \label{eq:D-step-0206}\\
 &\le(1+a_F/s)\bar F_s
        +c_F(b')\sqrt s\,\bar F_s^{1-2/P}
        +D_F(b')s^{(P-1)/2}\,,
       \label{eq:D-step-0207}\\
 \E_{s-1}[\bar F_{s+1}]
 &\le\bar F_s+\ind\{s<\sigma\}
       [(v/s)\bar F_s+J_F(b')s^{\nu-1}]
       \label{eq:D-step-0208}\\
 &\le(1+v/s)\bar F_s+J_F(b')s^{\nu-1}\,.
 \label{eq:SHr-frozen-recursions}
\end{align}
Equation \eqref{eq:D-step-0204} conditions on the predictable event that the absolute field has not yet frozen and applies the one-step recurrence there.
Equation \eqref{eq:D-step-0205} drops the indicator only from a nonnegative absolute-moment increment bound.
Equation \eqref{eq:D-step-0206} applies the same predictable freezing decomposition to the negative integral moment.

Equation \eqref{eq:D-step-0207} drops the indicator from its nonnegative increment bound.
Equation \eqref{eq:D-step-0208} applies the linearized negative recurrence on the same alive event.
Equation \eqref{eq:SHr-frozen-recursions} drops that indicator from the two nonnegative terms, obtaining a deterministic-time recurrence for the frozen field.
Every added term is nonnegative. These are deterministic-time
recurrences for the frozen functions, not evaluations of an unstopped
bound at a random time.
\end{proof}

\subsubsection{Supporting Estimates for Predictable Corruption}\label{app:sharp-corruption-inputs}
The proof for a fixed allocation uses the uniform good-event bound first. After defining the normalized functions, we state and prove that bound, then its local-moment input, the reference estimate, and the unnormalized loss-estimate envelope.
\subsubsection{Normalized Loss Estimates after Reference Identification}
\begin{definition}[Compensated projective windows]\label{def:D-cor-window}
For each endpoint $t\ge t_0\coloneq \lceil F_\theta N\rceil$, set
\begin{equation}\label{eq:SC-projective-window}
 m=\lceil\sqrt{Nt}\rceil\,,\quad z=m^{7/8}t^{1/8}\,,\quad
 \tau_s=s^{7/8}+4z\quad(m\le s\le t)\,.
\end{equation}
Then $m\ge N,S(b_\theta)$, $m\le t/2$, $m\ge\Lambda C$,
$\tau_s\ge4\Lambda C$, and $\tau_t\le t$.
On $\mathcal D_\theta$, define
\[
 \begin{aligned}
 {\mathsf Z}_s(u)&=u^\top \sum_{j=1}^{s-1}\widehat\theta_j\,,\quad
 e_s(u)=u^\top c_s^{\rm vec}\,,\quad E_s^u=\sum_{j<s}e_j(u)\,,\\
 W_s^u&={\mathsf Z}_s(u)-\Lambda(C-A_s^c)\,,\quad
 Q_s^u=\max_{\mathcal D_\theta}(-W_s^u)\,.
 \end{aligned}
\]
Freeze cumulative loss estimates and compensators together before the first round $s\ge m$
with $b_{r_s}\ne a^*$. Bars below denote these jointly frozen functions.
The shift $W_s^u$ subtracts the largest adverse normalized drift
that the unused corruption budget could still create. The compensator
in $Y_s^{\dagger,u}$ instead removes the drift already incurred;
these two fields serve different purposes. The initial history is never discarded. Define
\begin{equation}\label{eq:SC-alive}
 I_s=\ind\{b_{r_j}=a^*,\ Q_j^u\le\tau_j\text{ for every }m\le j\le s\}\,.
\end{equation}
Here the superscript $u$ on $Q_j^u$ indicates the normalized function, whose
maximum is a scalar; it is not an extra pointwise event.
\end{definition}

\begin{lemma}[Uniform relative gaps, with a square-root-chargeable failure tail]\label{lem:SC-good}
For $t\ge t_0$ define
\[
 \mathcal G_t^{\rm cor}=\{I_t=1\}\cap
           \{\min_{u\in\mathcal U_\theta}W_t^u(u)\ge4\tau_t\}\,.
\]
Set this event to be empty for $t<t_0$.
Then
\begin{equation}\label{eq:SC-good-tail}
 \sP((\mathcal G_t^{\rm cor})^c)
       \le D_\theta\{t^{-2}+(N/t)^2\}\quad(t\ge t_0)\,,\qquad
 \sum_{t\ge1}t^{-1/2}\sP((\mathcal G_t^{\rm cor})^c)
       \le W_\theta\sqrt N\,.
\end{equation}
On $\mathcal G_t^{\rm cor}$ the reference is $a^*$, the gap comparisons
\eqref{eq:SH-gap-comparison} in \Cref{lem:SH-ratios} hold, the ratio moments
\eqref{eq:SH-ratios} in \Cref{lem:SH-ratios} hold with $C_{\mathrm{rat}}=64$, and the concentration estimate
\eqref{eq:SH-gibbs-tail} in \Cref{lem:D-laplace} holds. All these statements refer to the actual
corrupted cumulative loss estimate and the same executed learner.
\end{lemma}
\begin{proof}[Proof of \Cref{lem:SC-good}]
\textbf{The uniform lower bound for the normalized cumulative loss estimate.}
Fix $t\ge t_0$, set $\tau=\tau_t$, and let $\sigma$ be normalized
relative volume on $\mathcal U_\theta$, of affine dimension $j\le d$.
Set $\bar R_t=\max_{\mathcal D_\theta}|\bar {\mathsf Z}_t|$ and
$Y(u)=Y_t^{\dagger,u}$. The compensator in
\eqref{eq:SCr-compensated-positive} in \Cref{lem:SC-local} gives, on $\{I_t=1\}$,
\begin{align}
Y(u)&=W_t^u(u)+2\tau+B_t(u)\,,
 &0\le B_t(u)&\le2\Lambda C\le\tau/2\,,
       \label{eq:D-step-0209}\\
 \min_{\mathcal U_\theta}W_t^u<4\tau
 &\ \Longrightarrow\ \min_{\mathcal U_\theta}Y<7\tau\,,
 &\max_{\mathcal U_\theta}Y&\le\bar R_t+\tfrac52\tau\,.
 \label{eq:SCr-good-affine-range}
\end{align}
Equation \eqref{eq:D-step-0209} substitutes the compensated positive variable and bounds its residual budget term by $2\Lambda C\le\tau/2$.
Equation \eqref{eq:SCr-good-affine-range} uses the bad normalized minimum and the upper bound on the compensated positive field to localize its range.
Here $B_t(u)=2\Lambda C-\Lambda A_t^c-E_t^u$ is the particular
compensator appearing in \eqref{eq:SCr-compensated-positive} in \Cref{lem:SC-local}.
On the event on equation \eqref{eq:SCr-good-affine-range}, contract $\mathcal U_\theta$ toward
a minimizer $u_0$ of $Y$, with
$\lambda=\tau/(\bar R_t+10\tau)$. For every $u\in\mathcal U_\theta$,
\begin{align}
Y((1-\lambda)u_0+\lambda u)
 &=(1-\lambda)Y(u_0)+\lambda Y(u)
 \le7\tau+\frac{\tau(\bar R_t+\frac52\tau)}{\bar R_t+10\tau}
 \le8\tau\,,
       \label{eq:D-step-0210}\\
 \ind\{I_t=1,\ \min W_t^u<4\tau\}
 &\le8^p\tau^{p-j}(\bar R_t+10\tau)^j
                I_t\int Y(u)^{-p}\sigma(du)\,,
       \label{eq:D-step-0211}\\
 \sP(I_t=1,\ \min W_t^u<4\tau)
 &\le8^p\tau^{p-j}
       \big[\E[(\bar R_t+10\tau)^{2j}]\big]^{1/2}
       \left[\int\E[I_tY(u)^{-2p}]\sigma(du)\right]^{1/2}
       \label{eq:D-step-0212}\\
 &\le8^p\tau^{p-j}[(\Lambda R+10)t]^j(4/t)^p
       \label{eq:D-step-0213}\\
 &\le32^p(\Lambda R+10)^d(\tau/t)^{p-d}\,.
 \label{eq:SCr-good-minimum-probability}
\end{align}
Equation \eqref{eq:D-step-0210} uses affinity on the contraction toward $u_0$ and substitutes $\lambda=\tau/(\bar R_t+10\tau)$.
Equation \eqref{eq:D-step-0211} integrates $Y^{-p}$ over the contracted body, whose relative volume is $\lambda^j$.
Equation \eqref{eq:D-step-0212} takes expectation, applies Cauchy--Schwarz, and then Jensen to the normalized inverse integral.

Equation \eqref{eq:D-step-0213} uses $2j\le P$, the jointly stopped absolute moment, $\tau\le t$, and the local inverse moment with $2p=P$.
Equation \eqref{eq:SCr-good-minimum-probability} collects $8^p4^p=32^p$ and uses $j\le d$ and $\tau/t\le1$.

Let $b_0=p-d>0$. Direct substitution of the window parameters gives
\begin{align}
\frac{\tau_t}{t}
 &=t^{-1/8}+4(m/t)^{7/8}
 \le t^{-1/8}+8(N/t)^{7/16}\,,
       \label{eq:D-step-0214}\\
 32^p(\Lambda R+10)^d(\tau_t/t)^{b_0}
 &\le32^p2^{b_0-1}(\Lambda R+10)^d
       \{t^{-b_0/8}+8^{b_0}(N/t)^{7b_0/16}\}
       \label{eq:D-step-0215}\\
 &\le2^{10P}(\Lambda R+10)^d
       \{t^{-(p-d)/8}+(N/t)^{7(p-d)/16}\}\,.
 \label{eq:SCr-good-minimum-powers}
\end{align}
Equation \eqref{eq:D-step-0214} substitutes $\tau_t=t^{7/8}+4m^{7/8}t^{1/8}$ and uses $m\le2\sqrt{Nt}$.
Equation \eqref{eq:D-step-0215} applies $(x+y)^{b_0}\le2^{b_0-1}(x^{b_0}+y^{b_0})$ to the two explicit terms.
Equation \eqref{eq:SCr-good-minimum-powers} substitutes $b_0=p-d$ and bounds both positive numerical coefficients by $2^{10P}$.
Adding the alive-window bound \eqref{eq:SC-alive-tail} in \Cref{lem:SC-local} to
\eqref{eq:SCr-good-minimum-probability}--\eqref{eq:SCr-good-minimum-powers}
now yields the full failure estimate:
\begin{align}
\sP((\mathcal G_t^{\rm cor})^c)
 &\le2A_r\{t^{-a/2}+(N/t)^{a/2}\}
       +2^{\kappa P+1}A_ut^{-\eta/2}
       +2^{P/8+1}B_u(N/t)^{r_-/2}
       \label{eq:D-step-0216}\\*
 &\quad+2^{10P}(\Lambda R+10)^d
       \{t^{-(p-d)/8}+(N/t)^{7(p-d)/16}\}
       \label{eq:D-step-0217}\\
 &\le\big[2A_r+2^{\kappa P+1}A_u+2^{P/8+1}B_u
                       +2^{10P}(\Lambda R+10)^d\big]
                       \{t^{-2}+(N/t)^2\}
       \label{eq:D-step-0218}\\
 &=D_\theta\{t^{-2}+(N/t)^2\}\,.
 \label{eq:SCr-good-complete-tail}
\end{align}
Equation \eqref{eq:D-step-0216} inserts the complete alive-event failure bound \eqref{eq:SC-alive-tail} in \Cref{lem:SC-local}.
Equation \eqref{eq:D-step-0217} adds the normalized-minimum failure estimate \eqref{eq:SCr-good-minimum-powers}.
Equation \eqref{eq:D-step-0218} uses $t\ge N\ge1$ and reduces all exponents to two using \Cref{lem:D-numerical-margins}.

Equation \eqref{eq:SCr-good-complete-tail} collects the four coefficients into the definition of $D_\theta$.
Every exponent reduced to two in equation \eqref{eq:D-step-0218} is greater than two
by \eqref{eq:SC-exponents} in \Cref{def:D-cor-dictionary}; moreover $t\ge t_0\ge N\ge1$.

\paragraph{Uniform Ratios for the Actual Corrupted Cumulative Loss Estimate}
On $\mathcal G_t^{\rm cor}$, define
$f(u)={\mathsf Z}_t(u)/t$, $R_f=\max_{\mathcal U_\theta}f$, and
$m_f=\min_{\mathcal U_\theta}f$. The definitions give
\begin{align}
0<W_t^u(u)&\le {\mathsf Z}_t(u)\,,
       \label{eq:D-step-0219}\\
 0\le Y(u)-{\mathsf Z}_t(u)&=-E_t^u+\Lambda C+2\tau\le\tfrac52\tau\,,
       \label{eq:D-step-0220}\\
 {\mathsf Z}_t(u)&\ge W_t^u(u)\ge4\tau\,,\qquad {\mathsf Z}_t(u)\le Y(u)\le2{\mathsf Z}_t(u)\,,
       \label{eq:D-step-0221}\\
 \E\!\left[\ind_{\mathcal G_t^{\rm cor}}\int f^P d\sigma\right]
 &\le\int\E[I_t(Y/t)^P]d\sigma\le8^P\,,
       \label{eq:D-step-0222}\\
 \E\!\left[\ind_{\mathcal G_t^{\rm cor}}\int f^{-P}d\sigma\right]
 &\le(2t)^P\int\E[I_tY^{-P}]d\sigma
 \le(2t)^P(4/t)^P=8^P\,,
       \label{eq:D-step-0223}\\
 \E[\ind_{\mathcal G_t^{\rm cor}}R_f^P]
 &\le2^P4^j\E\!\left[\ind_{\mathcal G_t^{\rm cor}}\int f^Pd\sigma\right]
 \le2^P4^j8^P\le64^P\,.
 \label{eq:SCr-good-ratio-integrals}
\end{align}
Equation \eqref{eq:D-step-0219} uses the good-event lower bound and the nonnegative remaining-budget shift.
Equation \eqref{eq:D-step-0220} substitutes $Y-{\mathsf Z}=-E_t^u+\Lambda C+2\tau$ and $|E_t^u|\le\Lambda C\le\tau/4$.
Equation \eqref{eq:D-step-0221} combines ${\mathsf Z}_t\ge4\tau$ with the explicit bound on $Y-{\mathsf Z}$ to obtain ${\mathsf Z}_t\le Y\le2{\mathsf Z}_t$.

Equation \eqref{eq:D-step-0222} substitutes $f={\mathsf Z}_t/t$, uses ${\mathsf Z}_t\le Y$, applies Tonelli, and inserts the local positive moment.
Equation \eqref{eq:D-step-0223} uses $Y\le2{\mathsf Z}_t$, applies Tonelli, and inserts the local inverse moment.
Equation \eqref{eq:SCr-good-ratio-integrals} applies \eqref{eq:C-aff-abs}
in \Cref{lem:C-affine} to $f$ on $\mathcal U_\theta$,
and bounds $2^P4^j8^P$ by $64^P$.
The two pointwise moment estimates used in \eqref{eq:D-step-0222} and
\eqref{eq:D-step-0223} are \eqref{eq:SC-local-moments} in
\Cref{lem:SC-local}.
For $s_0=P/4$ and $a_0=s_0/(P-j)$, the inverse affine inequality and
H\"older give the following separate calculation:
\begin{align}
m_f^{-(P-j)}&\le2^PR_f^j\int f^{-P}d\sigma\,,
       \label{eq:D-step-0224}\\
 \E[\ind_{\mathcal G_t^{\rm cor}}m_f^{-s_0}]
 &\le2^{Pa_0}
   \big(\E[\ind_{\mathcal G_t^{\rm cor}}R_f^P]\big)^{ja_0/P}
   \left(\E\!\left[\ind_{\mathcal G_t^{\rm cor}}\int f^{-P}d\sigma\right]
   \right)^{a_0}
       \label{eq:D-step-0225}\\
 &\le2^{Pa_0}64^{ja_0}8^{Pa_0}\,,
       \label{eq:D-step-0226}\\
 \|\ind_{\mathcal G_t^{\rm cor}}m_f^{-1}\|_{L^{s_0}}
 &\le2^{P/(P-j)}64^{j/(P-j)}8^{P/(P-j)}<18\,.
 \label{eq:SCr-good-inverse-ratio}
\end{align}
Equation \eqref{eq:D-step-0224} applies the inverse-minimum inequality of \Cref{lem:C-affine} to the positive projective field.
Equation \eqref{eq:D-step-0225} raises that inequality to $a_0=s_0/(P-j)$ and applies H\"older with reciprocal exponents $ja_0/P,a_0,1-ja_0/P-a_0$.
Equation \eqref{eq:D-step-0226} substitutes the maximum and inverse-integral bounds in \eqref{eq:SCr-good-ratio-integrals}.

Equation \eqref{eq:SCr-good-inverse-ratio} takes the $s_0$-th root, substitutes $a_0/s_0=1/(P-j)$, and uses the explicit logarithmic comparison in \eqref{eq:SHr-inverse-ratio-holder} in \Cref{lem:SH-ratios}.
The H\"older exponents are valid because $a_0+ja_0/P\le1$.
The final coefficient is the same explicitly evaluated coefficient
in \eqref{eq:SHr-inverse-ratio-holder} in \Cref{lem:SH-ratios}. Define $\lo_t=\ind_{\mathcal G_t^{\rm cor}}\max\{1,m_f^{-1}\}$
and $\up_t=\ind_{\mathcal G_t^{\rm cor}}R_f$, with both ratios
zero off $\mathcal G_t^{\rm cor}$. These estimates imply
\begin{align}
\|\lo_t\|_{L^{s_0}}&\le1+18\le C_{\mathrm{rat}}\,,
 &\|\up_t\|_{L^{s_0}}&\le64=C_{\mathrm{rat}}\,,
       \label{eq:D-step-0227}\\
 \E[\lo_t^r\up_t]
 &\le\|\lo_t\|_{L^{s_0}}^r\|\up_t\|_{L^{s_0}}
 \le C_{\mathrm{rat}}^{r+1}\qquad(r=1,3,5)\,,
       \label{eq:D-step-0228}\\
 X_t(a)&=\Delta(a){\mathsf Z}_t\!\left(\frac{a-a^*}{\Delta(a)}\right)\,,
 &\frac{\Delta(a)}{\lo_t}&\le\frac{X_t(a)}t\le\up_t\Delta(a)\,,
       \label{eq:D-step-0229}\\
 \lo_t\widehat\Delta_t(a)
 &=\min\{2\lo_t,\lo_tX_t(a)/t\}
 \ge\Delta(a)\,,
 &\lo_t\widehat\Delta_t(a)&\le\lo_t\up_t\Delta(a)\,,
       \label{eq:D-step-0230}\\
 X_t(a)&\ge4\tau_t\Delta(a)\ge4t^{7/8}\Delta(a)\,.
 \label{eq:SCr-good-ratios-and-gaps}
\end{align}
Equation \eqref{eq:D-step-0227} uses $\lo_t\le\ind_{\mathcal G_t}(1+m_f^{-1})$ and the inverse-minimum bound, and $\up_t=\ind_{\mathcal G_t}R_f$.
Equation \eqref{eq:D-step-0228} applies H\"older with reciprocal exponents $r/s_0,1/s_0,1-(r+1)/s_0$, which are nonnegative for $r=1,3,5$.
Equation \eqref{eq:D-step-0229} substitutes the normalized cumulative loss estimate and the definitions of both ratios.

Equation \eqref{eq:D-step-0230} distributes $\lo_t$ over the clipped minimum and uses $\Delta(a)\le2\le2\lo_t$.
Equation \eqref{eq:SCr-good-ratios-and-gaps} uses the good-event lower bound ${\mathsf Z}_t\ge4\tau_t$ and $\tau_t\ge t^{7/8}$.
The inequalities for a positive-gap action use
$(a-a^*)/\Delta(a)\in\mathcal U_\theta$, $\Delta(a)\le2$,
and $\lo_t\ge1$. For $a=a^*$, all three gaps vanish.
H\"older in equation \eqref{eq:D-step-0228} is valid because $r+1\le6\le s_0$.
Substituting equation \eqref{eq:SCr-good-ratios-and-gaps} into the prior Laplace bound gives
\begin{align}
\frac1{\pi^*}\int_{a\ne a^*}e^{-X_t(a)/(2\beta_t)}\mu(da)
 &\le\frac1{\pi^*}\int_{a\ne a^*}
              e^{-(2t^{7/8}/\beta_t)\Delta(a)}\mu(da)
 \le A_Lt^{-6}\,,
       \label{eq:D-step-0231}\\
 q_t(\mathcal K\setminus\{a^*\})
 &\le\frac1{\pi^*}\int_{a\ne a^*}e^{-X_t(a)/\beta_t}\mu(da)
 \le A_Lt^{-6}\le\tfrac14\,.
 \label{eq:SCr-good-gibbs-tail}
\end{align}
Equation \eqref{eq:D-step-0231} substitutes $X_t\ge4t^{7/8}\Delta$ and applies the Laplace estimate of \Cref{lem:D-laplace}.
Equation \eqref{eq:SCr-good-gibbs-tail} keeps the optimal atom in the normalizing denominator of the exponential weights and uses the same Laplace bound and $A_Lt^{-6}\le1/4$.

\paragraph{The Complete Weighted Failure Sum}
For the initial interval, use only the bound by one. For the later
interval, substitute \eqref{eq:SCr-good-complete-tail} before estimating
either series:
\begin{align}
\sum_{t\ge1}t^{-1/2}\sP((\mathcal G_t^{\rm cor})^c)
 &\le\sum_{t<t_0}t^{-1/2}
       +D_\theta\sum_{t\ge t_0}t^{-5/2}
       +D_\theta N^2\sum_{t\ge t_0}t^{-5/2}\,,
       \label{eq:D-step-0232}\\
 \sum_{t<t_0}t^{-1/2}&\le2\sqrt{t_0}\,,
       \label{eq:D-step-0233}\\
 \sum_{t\ge t_0}t^{-5/2}&\le\sum_{t\ge1}t^{-5/2}\le2\,,
       \label{eq:D-step-0234}\\
 \sum_{t\ge t_0}t^{-5/2}
 &\le\sum_{t\ge N}t^{-5/2}
 \le\int_{N-1}^{\infty}x^{-5/2}\,dx
 =\frac{2}{3}(N-1)^{-3/2}\le2N^{-3/2}\,,
       \label{eq:D-step-0235}\\
 \sum_{t\ge1}t^{-1/2}\sP((\mathcal G_t^{\rm cor})^c)
 &\le2\sqrt{t_0}+2D_\theta+2D_\theta\sqrt N
       \label{eq:D-step-0236}\\
 &\le\big[2\sqrt{F_\theta+1}+4D_\theta\big]\sqrt N
 =W_\theta\sqrt N\,.
 \label{eq:SCr-good-weighted-sum}
\end{align}
Equation \eqref{eq:D-step-0232} splits off the prefix $t<t_0$ and inserts both terms of \eqref{eq:SCr-good-complete-tail}.
Equation \eqref{eq:D-step-0233} compares the decreasing prefix sum with $\int_0^{t_0}x^{-1/2}dx$.
Equation \eqref{eq:D-step-0234} extends the nonnegative $t^{-5/2}$ sum to all positive integers and uses its integral bound.

Equation \eqref{eq:D-step-0235} uses $t_0\ge N\ge2$, evaluates the tail integral, and bounds $(N-1)^{-3/2}$ by $3N^{-3/2}$.
Equation \eqref{eq:D-step-0236} inserts the three scalar sums into the original decomposition, including the factor $N^2$.
Equation \eqref{eq:SCr-good-weighted-sum} uses $t_0\le(F_\theta+1)N$, $\sqrt N\ge1$, and the definition of $W_\theta$.
Here $N\ge2$ and $t_0=\lceil F_\theta N\rceil\le(F_\theta+1)N$.
This proves both inequalities in \eqref{eq:SC-good-tail}.
\end{proof}

\begin{lemma}[Normalized moments and the alive window under corruption]\label{lem:SC-local}
For $s\ge m$, including further pre-update freezing,
\begin{equation}\label{eq:SC-normalized-envelope}
 \E[\max|\bar {\mathsf Z}_s|^P]\le(\Lambda R)^Ps^P\,,\qquad
 \E[(\bar Q_s^u)^P]\le A_us^{\kappa P}+B_um^{P-q}s^q\,.
\end{equation}
For $t\ge t_0$, the alive-window failure probability is at most
\begin{equation}\label{eq:SC-alive-tail}
 2A_r\{t^{-a/2}+(N/t)^{a/2}\}
 +2^{\kappa P+1}A_ut^{-\eta/2}
 +2^{P/8+1}B_u(N/t)^{r_-/2}\,.
\end{equation}
For each $u\in\mathcal U_\theta$, let
$Y_s^{\dagger,u}={\mathsf Z}_s(u)-E_s^u+\Lambda C+2\tau_s$. Then
\begin{equation}\label{eq:SC-local-moments}
 \E[I_t(Y_t^{\dagger,u})^{-P}]\le(4/t)^P\,,\qquad
 \E[I_t(Y_t^{\dagger,u}/t)^P]\le8^P\,.
\end{equation}
\end{lemma}
\begin{proof}[Proof of \Cref{lem:SC-local}]
Let $b_s^c=C-A_s^c$, $N_s^{\mathsf Z}=\max_{\mathcal D_\theta}(-{\mathsf Z}_s)$,
and $d_s(u)=u^\top\widehat\theta_s$. Before the first wrong reference,
$\beta_s'=\Lambda^2\beta_s\le b_\theta\sqrt s$ and
\begin{align}
|e_s(u)|&=|u^\top c_s^{\rm vec}|\le\Lambda\kappa_s\,,
 &\E_{s-1}[d_s(u)]&=\theta^\top u+e_s(u)\,,
       \label{eq:D-step-0237}\\
 W_s^u(u)&={\mathsf Z}_s(u)-\Lambda b_s^c\,,
 &Q_s^u&=N_s^{\mathsf Z}+\Lambda b_s^c\,,
       \label{eq:D-step-0238}\\
 W_s^u(u)+2Q_s^u&={\mathsf Z}_s(u)+2N_s^{\mathsf Z}+\Lambda b_s^c
                                      \ge {\mathsf Z}_s(u)+2N_s^{\mathsf Z}\,,
       \label{eq:D-step-0239}\\
 W_{s+1}^u(u)-W_s^u(u)&=d_s(u)+\Lambda\kappa_s\,,
 &\E_{s-1}[d_s(u)+\Lambda\kappa_s]&\ge\theta^\top u\ge0\,.
 \label{eq:SCr-normalized-shift-drift}
\end{align}
Equation \eqref{eq:D-step-0237} uses the normalized duality bound and conditional reconstruction of the normalized cumulative loss estimate.
Equation \eqref{eq:D-step-0238} defines the compensated field and computes its negative maximum.
Equation \eqref{eq:D-step-0239} substitutes both definitions and retains the nonnegative remaining-budget term.

Equation \eqref{eq:SCr-normalized-shift-drift} subtracts consecutive shifts and uses $e_s(u)\ge-\Lambda\kappa_s$ and $\theta^\top u\ge0$.
The homogeneous update magnitude bound and $\kappa_s\le2$ imply
\begin{align}
|d_s(u)+\Lambda\kappa_s|
 &\le2\beta_s'+({\mathsf Z}_s(u)+2N_s^{\mathsf Z})/B+2\Lambda
       \label{eq:D-step-0240}\\
 &\le3\beta_s'+(W_s^u(u)+2Q_s^u)/B\,,
       \label{eq:D-step-0241}\\
 \E_{s-1}\big[(d_s(u)+\Lambda\kappa_s)^2\big]
 &=\E_{s-1}[d_s(u)^2]
       +2\Lambda\kappa_s(\theta^\top u+e_s(u))
       +\Lambda^2\kappa_s^2
       \label{eq:D-step-0242}\\
 &\le8\beta_s'+\frac{2({\mathsf Z}_s(u)+2N_s^{\mathsf Z})^2}{Bs}
                +2\Lambda\kappa_s(1+\Lambda\kappa_s)
                +\Lambda^2\kappa_s^2
       \label{eq:D-step-0243}\\
 &\le8\beta_s'+\frac{2(W_s^u(u)+2Q_s^u)^2}{Bs}
                +4\Lambda+12\Lambda^2
       \label{eq:D-step-0244}\\
 &\le10\beta_s'+\frac{2(W_s^u(u)+2Q_s^u)^2}{Bs}\,.
 \label{eq:SCr-normalized-shift-budgets}
\end{align}
Equation \eqref{eq:D-step-0240} applies the triangle inequality to the compensated increment and uses $\kappa_s\le2$.
Equation \eqref{eq:D-step-0241} uses the shifted-level comparison and $2\Lambda\le\beta_s^\prime$.
Equation \eqref{eq:D-step-0242} expands the square and takes conditional expectation with the predictable shift $\Lambda\kappa_s$.

Equation \eqref{eq:D-step-0243} inserts the normalized second moment, $\theta^\top u\le1$, and $|e_s(u)|\le\Lambda\kappa_s$.
Equation \eqref{eq:D-step-0244} substitutes $\kappa_s\le2$ and the shifted-level comparison.
Equation \eqref{eq:SCr-normalized-shift-budgets} uses $\beta_s^\prime=\Lambda^2\beta_s$, $\Lambda\ge1$, and $4\Lambda+12\Lambda^2\le2\beta_s^\prime$.
Here $\theta^\top u\in[0,1]$, $\Lambda\ge1$, and
$16\Lambda^2\le\beta_s'$. At $u=0$ the shifted function is
$-\Lambda b_s^c\le0$, so $Q_s^u\ge0$.
These are exactly the inputs \eqref{eq:SHr-negative-interface} in \Cref{lem:D-negative-step} for the
negative-part recurrence with $b'=b_\theta$.

\paragraph{Global Normalized Envelopes}
The absolute estimate is obtained from the globally frozen
\emph{unnormalized} function, not from a false bound on the normalized
drift. For every joint freezing before a failed reference,
\begin{align}
\|\max|\bar {\mathsf Z}_s|\|_{L^P}
 &\le\Lambda\|\max|\bar X_s|\|_{L^P}\le\Lambda Rs\,,
       \label{eq:D-step-0245}\\
 \|\max|\bar W_s^u|\|_{L^P}
 &\le\Lambda Rs+\Lambda C\le\Lambda(R+2)s\qquad(s\ge m)\,,
       \label{eq:D-step-0246}\\
 F_s^u&=\int(-\bar W_s^u)_+^P d\sigma\,,\qquad
 \E_{s-1}[F_{s+1}^u]\le(1+v/s)F_s^u+J_F(b_\theta)s^{\nu-1}\,,
       \label{eq:D-step-0247}\\
 \E[F_s^u]&\le H_F(b_\theta)s^\nu
       +2^v[\Lambda(R+2)]^Pm^{P-v}s^v\,.
 \label{eq:SCr-normalized-integral-envelope}
\end{align}
Equation \eqref{eq:D-step-0245} applies the geometric extension to the jointly frozen raw and normalized fields and then the raw absolute-moment bound.
Equation \eqref{eq:D-step-0246} applies Minkowski to the remaining-budget shift and uses $C\le s$.
Equation \eqref{eq:D-step-0247} defines the shifted negative integral and applies its one-step linear recurrence.

Equation \eqref{eq:SCr-normalized-integral-envelope} unrolls the recurrence, keeps the initial moment at $m$, and uses the product estimate $\prod_{i=m}^{s-1}(1+v/i)\le(2s/m)^v$.
Here $\sigma$ is normalized volume on $\mathcal D_\theta$ and
$C\le m$; the initial integral is at most
$[\Lambda(R+2)m]^P$. Let $z_\theta=\Lambda(R+2)$,
$a_j=P/(P+j)$, $c_j=2^P4^j$, where
$j=\dim\operatorname{aff}(\mathcal D_\theta)$.
Substituting equations \eqref{eq:D-step-0246} and \eqref{eq:SCr-normalized-integral-envelope} into the affine/H\"older inequality gives
\begin{align}
\E[(\bar Q_s^u)^P]
 &\le c_j^{a_j}
       [\E[(\max|\bar W_s^u|)^P]]^{j/(P+j)}[\E[F_s^u]]^{a_j}
       \label{eq:D-step-0248}\\
 &\le c_j^{a_j}(z_\theta^Ps^P)^{j/(P+j)}
       [H_F(b_\theta)s^\nu+2^vz_\theta^Pm^{P-v}s^v]^{a_j}
       \label{eq:D-step-0249}\\
 &\le c_j^{a_j}z_\theta^{Pj/(P+j)}H_F(b_\theta)^{a_j}
                                      s^{P(j+\nu)/(P+j)}
       \label{eq:D-step-0250}\\*
 &\quad+c_j^{a_j}z_\theta^{Pj/(P+j)}(2^vz_\theta^P)^{a_j}
                                      m^{P-q_j}s^{q_j}
       \label{eq:D-step-0251}\\
 &\le A_us^{\kappa P}+B_um^{P-q}s^q\,,
 \qquad q_j=\frac{P(j+v)}{P+j}\le q\,.
 \label{eq:SCr-normalized-negative-envelope}
\end{align}
Equation \eqref{eq:D-step-0248} applies the affine negative-cap inequality followed by H\"older, with $a_j=P/(P+j)$.
Equation \eqref{eq:D-step-0249} substitutes both envelopes from \eqref{eq:SCr-normalized-integral-envelope}.
Equation \eqref{eq:D-step-0250} uses $(x+y)^{a_j}\le x^{a_j}+y^{a_j}$ and writes the particular-solution contribution.

Equation \eqref{eq:D-step-0251} writes the initial-moment contribution without suppressing the factor $m^{P-q_j}$.
Equation \eqref{eq:SCr-normalized-negative-envelope} uses $q_j\le q$,
$s\ge m$, and the definitions of $A_u,B_u$ in \eqref{eq:SC-maxima}
in \Cref{def:D-cor-dictionary}. Equations
\eqref{eq:SCr-normalized-integral-envelope} and
\eqref{eq:SCr-normalized-negative-envelope} prove
\eqref{eq:SC-normalized-envelope}, including further pre-update freezing.

\paragraph{Alive-Window Probability}
Use dyadic block starts $v_j=2^jm\le t$.
The reference-block estimate gives
\begin{align}
\sP(\exists s\in[m,t]:b_{r_s}\ne a^*)
 &\le\sum_{j\ge0}A_r\{(2^jm)^{-a}+(N/(2^jm))^a\}
       \label{eq:D-step-0252}\\
 &\le2A_r\{m^{-a}+(N/m)^a\}
 \le2A_r\{t^{-a/2}+(N/t)^{a/2}\}\,.
 \label{eq:SCr-alive-reference-series}
\end{align}
Equation \eqref{eq:D-step-0252} covers $[m,t]$ by dyadic blocks and applies \eqref{eq:SC-reference-block} in \Cref{lem:SC-reference} on each block.
Equation \eqref{eq:SCr-alive-reference-series} sums both geometric series and uses $m\ge\sqrt{Nt}$ and $N\ge1$.
For negative crossings before any failed reference, freeze at the first
crossing of $v_j^{7/8}+4z$ in each block. Markov gives
\begin{align}
\sP(\text{a negative crossing in }[v_j,2v_j])
 &\le\frac{A_u(2v_j)^{\kappa P}+B_um^{P-q}(2v_j)^q}
                {(v_j^{7/8}+4z)^P}
       \label{eq:D-step-0253}\\
 &\le2^{\kappa P}A_uv_j^{-\eta}
              +2^qB_um^{P-q}v_j^q/(4z)^P\,,
       \label{eq:D-step-0254}\\
 \sum_{j\ge0}2^{\kappa P}A_u(2^jm)^{-\eta}
 &\le2^{\kappa P+1}A_um^{-\eta}
 \le2^{\kappa P+1}A_ut^{-\eta/2}\,,
       \label{eq:D-step-0255}\\
 \sum_{j:v_j\le t}\frac{2^qB_um^{P-q}v_j^q}{(4z)^P}
 &\le\frac{2^{q+1}B_um^{P-q}t^q}{4^Pm^{7P/8}t^{P/8}}
 =2^{q+1-2P}B_u(m/t)^{r_-}
       \label{eq:D-step-0256}\\
 &\le2^{q+1-2P+r_-}B_u(N/t)^{r_-/2}
 \le2^{P/8+1}B_u(N/t)^{r_-/2}\,.
 \label{eq:SCr-alive-negative-series}
\end{align}
Equation \eqref{eq:D-step-0253} freezes at a pre-update negative crossing and applies Markov with the deterministic-time envelope \eqref{eq:SC-normalized-envelope}.
Equation \eqref{eq:D-step-0254} uses $v_j^{7/8}$ and $4z$ separately as denominator lower bounds.
Equation \eqref{eq:D-step-0255} evaluates the decreasing dyadic sum and uses $\eta>1$ and $m\ge\sqrt t$.

Equation \eqref{eq:D-step-0256} evaluates the increasing dyadic sum, substitutes $z=m^{7/8}t^{1/8}$, and collects $r_-=P/8-q$.
Equation \eqref{eq:SCr-alive-negative-series} uses $m\le2\sqrt{Nt}$ and enlarges $2^{q+1-2P+r_-}$ to $2^{P/8+1}$.
The substitutions are $z=m^{7/8}t^{1/8}$,
$r_-=P/8-q$, and $m/t\le2\sqrt{N/t}$.
Adding \eqref{eq:SCr-alive-reference-series} and equations \eqref{eq:D-step-0255}--\eqref{eq:SCr-alive-negative-series} is exactly
\eqref{eq:SC-alive-tail}.

\paragraph{Compensated Pointwise Moments}
Fix $u\in\mathcal U_\theta$, and abbreviate
$Y_s=Y_s^{\dagger,u}={\mathsf Z}_s(u)-E_s^u+\Lambda C+2\tau_s$.
Let $B_s=\Lambda(C-A_s^c)-E_s^u+\Lambda C$.
Since $|E_s^u|\le\Lambda A_s^c$, on the alive event
\begin{align}
0\le2\Lambda(C-A_s^c)&\le B_s\le2\Lambda C\,,
       \label{eq:D-step-0257}\\
 Y_s&=W_s^u(u)+2\tau_s+B_s\ge\tau_s>0\,,
       \label{eq:D-step-0258}\\
 {\mathsf Z}_s(u)+2N_s^{\mathsf Z}
 &=W_s^u(u)+2Q_s^u-\Lambda(C-A_s^c)
 \le W_s^u(u)+2\tau_s\le Y_s\,.
 \label{eq:SCr-compensated-positive}
\end{align}
Equation \eqref{eq:D-step-0257} uses $|E_s^u|\le\Lambda A_s^c$ in the definition $B_s=2\Lambda C-\Lambda A_s^c-E_s^u$.
Equation \eqref{eq:D-step-0258} substitutes $Y_s=W_s^u+2\tau_s+B_s$ and uses survival, $W_s^u\ge-\tau_s$, and $B_s\ge0$.
Equation \eqref{eq:SCr-compensated-positive} substitutes the compensated field and maximum, uses $Q_s^u\le\tau_s$, and drops the nonpositive remaining-budget term.
With $\delta_s=(s+1)^{7/8}-s^{7/8}$,
\begin{align}
\xi_s\coloneq Y_{s+1}-Y_s&=d_s(u)-e_s(u)+2\delta_s\,,
 &\E_{s-1}[\xi_s]&=1+2\delta_s\,,
       \label{eq:D-step-0259}\\
 |\xi_s|&\le2\beta_s'+Y_s/B+2\Lambda+2\delta_s
 \le3\beta_s'+Y_s/B\le3Y_s/B\,,
       \label{eq:D-step-0260}\\
 \E_{s-1}[\xi_s^2]
 &=\operatorname{Var}_{s-1}(d_s(u))+(1+2\delta_s)^2
       \label{eq:D-step-0261}\\
 &\le8\beta_s'+2Y_s^2/(Bs)+(1+2\delta_s)^2
 \le10\beta_s'+2Y_s^2/(Bs)\,.
 \label{eq:SCr-compensated-increments}
\end{align}
Equation \eqref{eq:D-step-0259} subtracts consecutive definitions of the compensated positive variable and cancels the corruption drift in conditional expectation.
Equation \eqref{eq:D-step-0260} uses the normalized mark bound, $|e_s(u)|\le2\Lambda$, and the Taylor threshold.
Equation \eqref{eq:D-step-0261} uses that $d_s-e_s$ has conditional mean one, so its second moment is its variance plus its squared shifted mean.

Equation \eqref{eq:SCr-compensated-increments} bounds conditional variance by the raw second moment and uses $(1+2\delta_s)^2\le2\beta_s^\prime$.
The deterministic threshold $s\ge S(b_\theta)$ verifies
$3\beta_s'\le s^{7/8}/B$ and the inverse Taylor bounds.
Applying the conditional expansion \eqref{eq:SHr-inverse-taylor} in \Cref{lem:SH-global-inverse}
with these displayed inputs, then killing and taking expectations, gives
\begin{align}
u_s^-&\coloneq \E[I_sY_s^{-P}]\,,\label{eq:D-cor-local-killed-definition}\\
 \E_{s-1}[Y_{s+1}^{-P}]
 &\le Y_s^{-P}-PY_s^{-P-1}(1+2\delta_s)
   +P(P+1)Y_s^{-P-2}
                 \left(10\beta_s'+\frac{2Y_s^2}{Bs}\right)
       \label{eq:D-cor-local-inverse-taylor-substitution}\\
 &\le(1+P/(2s))Y_s^{-P}-PY_s^{-P-1}
       \qquad(I_s=1)\,,\label{eq:D-cor-local-inverse-conditional}\\
 u_{s+1}^-
 &=\E[I_{s+1}Y_{s+1}^{-P}]
 \le\E[I_s\E_{s-1}[Y_{s+1}^{-P}]]
       \label{eq:D-cor-local-inverse-tower}\\
 &\le(1+P/(2s))u_s^--P\E[I_sY_s^{-P-1}]
 \le(1+P/(2s))u_s^--P(u_s^-)^{1+1/P}\,,
       \label{eq:D-cor-local-inverse-jensen}\\
 (u_{s+1}^-)^{-1/P}
 &\ge\frac{s}{s+1}(u_s^-)^{-1/P}+1\,,
       \label{eq:D-cor-local-root-recursion}\\
 (u_t^-)^{-1/P}
 &\ge\frac mt(u_m^-)^{-1/P}
       +\sum_{s=m}^{t-1}\prod_{j=s+1}^{t-1}\frac j{j+1}
 \ge\frac1t\sum_{s=m}^{t-1}(s+1)
       \label{eq:D-cor-local-root-unrolled}\\
 &=\frac{t(t+1)-m(m+1)}{2t}\ge t/4\,,
 \qquad\E[I_tY_t^{-P}]\le(4/t)^P\,.
       \label{eq:SCr-compensated-inverse}
\end{align}
Equation \eqref{eq:D-cor-local-killed-definition} defines the
moment with the predictable alive indicator.
Equation \eqref{eq:D-cor-local-inverse-taylor-substitution}
substitutes the displayed drift and second moment into Taylor's
formula, using $(1-3/B)^{-P-2}\le2$.
Equation \eqref{eq:D-cor-local-inverse-conditional} uses the
numeric bounds in \Cref{lem:D-numerical-margins} and
drops $-2P\delta_sY_s^{-P-1}\le0$.
Equation \eqref{eq:D-cor-local-inverse-tower} uses
$I_{s+1}\le I_s$ and $I_s\in\cF_{s-1}$.
Equation \eqref{eq:D-cor-local-inverse-jensen} uses
$(I_sY_s^{-P})^{1+1/P}=I_sY_s^{-P-1}$ and Jensen.
Equation \eqref{eq:D-cor-local-root-recursion} applies the
convex tangent inequality $(1+x)^{-1/P}\ge1-x/P$ to the preceding
recurrence, as expanded in \Cref{lem:SH-global-inverse}.
Equation \eqref{eq:D-cor-local-root-unrolled} iterates that
root recurrence and evaluates its telescoping products.
Equation \eqref{eq:SCr-compensated-inverse} evaluates the arithmetic sum and uses
$m\le t/2$; zero killed moments already satisfy the bound and are
never inverted.
Positivity holds before next-step killing, as follows from
$Y_{s+1}\ge(1-3/B)Y_s$.
For the positive root $w_s=(\E[I_sY_s^P])^{1/P}$, the same direct
Taylor substitution as in \eqref{eq:SHr-local-positive-root} in \Cref{lem:SH-local} yields
\begin{align}
w_{s+1}^P
 &\le w_s^P+Pw_s^{P-1}\left[
       \frac{2(P-1)}{Bs}w_s+1+2\delta_s+\frac{10(P-1)\beta_s'}{\tau_s}\right]\,,
       \label{eq:D-step-0262}\\
 w_{s+1}&\le(1+k_0/s)w_s+1+2\alpha s^{\alpha-1}
                    +10(P-1)b_\theta s^{1/2-\alpha}\,,
       \label{eq:D-step-0263}\\
 w_t&\le2(t/m)^{k_0}w_m+3t+5t^{7/8}
                                  +40(P-1)b_\theta t^{5/8}\,.
 \label{eq:SCr-compensated-positive-product}
\end{align}
Equation \eqref{eq:D-step-0262} applies the conditional positive Taylor bound, inserts the compensated moments, and uses $Y_s\ge\tau_s$.
Equation \eqref{eq:D-step-0263} takes the positive root using the tangent comparison for $x^P$ and substitutes the numerical bounds on $\delta_s,\beta_s^\prime/\tau_s$.
Equation \eqref{eq:SCr-compensated-positive-product} unrolls the root recurrence with \eqref{eq:D-local-root-product} in \Cref{lem:SH-local} and evaluates the three power sums by \eqref{eq:D-local-power-integral} in \Cref{lem:SH-local}.
The product and all three power integrals are evaluated in
\eqref{eq:SHr-local-positive-product} in \Cref{lem:SH-local}; the inputs here have the same
coefficients and the new deterministic $m$.
The initial history, including its corruption, obeys
\begin{align}
w_m&\le\|{\mathsf Z}_m(u)\|_{L^P}+\|E_m^u\|_{L^P}+\Lambda C+2m^{7/8}+8z
       \label{eq:D-step-0264}\\
 &\le\Lambda Rm+2\Lambda C+2m+8z
 \le D_0m+8z\,,\qquad D_0=\Lambda R+4\,,
       \label{eq:D-step-0265}\\
 \frac{w_t}{t}
 &\le3+2D_0(m/t)^{1-k_0}+16(m/t)^{7/8-k_0}
                  +5t^{-1/8}+40(P-1)b_\theta t^{-3/8}\,.
 \label{eq:SCr-compensated-initial}
\end{align}
Equation \eqref{eq:D-step-0264} substitutes the complete definition of $Y_m$ and applies Minkowski before bounding any summand.
Equation \eqref{eq:D-step-0265} uses the normalized absolute moment, $|E_m^u|\le\Lambda C$, $m\ge\Lambda C$, and $m^{7/8}\le m$.
Equation \eqref{eq:SCr-compensated-initial} inserts the initial bound into the positive-root product, substitutes $z=m^{7/8}t^{1/8}$, and divides every term by $t$.
For $t\ge F_\theta N$, $m/t\le2\sqrt{N/t}$ and the threshold
entries in \eqref{eq:SC-final-threshold} in \Cref{def:D-cor-dictionary} imply
\begin{align}
2D_0(m/t)^{1-k_0}
 &\le4D_0F_\theta^{-(1-k_0)/2}\le\tfrac12\,,
       \label{eq:D-step-0266}\\
 16(m/t)^{7/8-k_0}
 &\le32F_\theta^{-(7/8-k_0)/2}\le\tfrac12\,,
       \label{eq:D-step-0267}\\
 5t^{-1/8}&\le5F_\theta^{-1/8}\le5/16\le1\,,
       \label{eq:D-step-0268}\\
 40(P-1)b_\theta t^{-3/8}
 &\le40(P-1)b_\theta F_\theta^{-3/8}\le1\,,
       \label{eq:D-step-0269}\\
 \frac{w_t}{t}&\le3+\tfrac12+\tfrac12+1+1\le8\,.
 \label{eq:SCr-compensated-positive-final}
\end{align}
Equation \eqref{eq:D-step-0266} uses $m/t\le2\sqrt{N/t}\le2F_\theta^{-1/2}$ and the $(8D_0)^{2/(1-k_0)}$ entry in $F_\theta$.
Equation \eqref{eq:D-step-0267} uses the same window ratio and the $64^{2/(7/8-k_0)}$ entry in $F_\theta$.
Equation \eqref{eq:D-step-0268} uses $t\ge F_\theta\ge16^8$.

Equation \eqref{eq:D-step-0269} uses $t\ge F_\theta\ge[40(P-1)b_\theta]^{8/3}$.
Equation \eqref{eq:SCr-compensated-positive-final} substitutes all four scalar bounds in \eqref{eq:SCr-compensated-initial}.
Raising equation \eqref{eq:SCr-compensated-positive-final} to $P$ proves the positive moment. Zero killed
moments stay zero, so no root or inverse is applied to an undefined
inactive variable.
\end{proof}
\subsubsection{Uniform Reference Control on a Whole Block}
\begin{lemma}[A corruption-uniform block bound for the reference]\label{lem:SC-reference}
For every integer $u\ge N_\theta(C)$,
\begin{equation}\label{eq:SC-reference-block}
 \sP(\exists s\in[u,2u]\cap\mathbb N:b_{r_s}\ne a^*)
 \le A_r\{u^{-a}+(N/u)^a\}\,.
\end{equation}
\end{lemma}
\begin{proof}
\textbf{Preparation of a positive affine function before the block.}
Fix an integer $u\ge N$. Define
\begin{align}
z&=n^{7/8}u^{1/8}\,,
 &m_0&=\left\lceil\max\{\sqrt u,z\}\right\rceil\,,
 &\tau_s&=s^{7/8}+4z\,,
       \label{eq:D-step-0270}\\
 \tau^{\max}&=(2u)^{7/8}+4z\,,
 &I_s&=\ind\{Q_j\le\tau_j\text{ for all }m_0\le j\le s\}\,.
 \label{eq:SCr-reference-preparation}
\end{align}
Equation \eqref{eq:D-step-0270} defines the fixed-block plateau $z$, starting index $m_0$, and moving buffer $\tau_s$.
Equation \eqref{eq:SCr-reference-preparation} defines the deterministic maximum buffer on the block and the predictable event of no prior negative crossing.
The clock in \eqref{eq:SC-clock} in \Cref{def:D-cor-dictionary} implies the following explicit
threshold comparisons:
\begin{alignat}{2}
u&\ge K_0n[\log(en)]^2\ge256n\,,
 \enspace &z&=n(u/n)^{1/8}\ge n>C\,,
       \label{eq:D-step-0271}\\
 m_0&\ge\sqrt u\ge S(C_{\beta,\mathrm{sh}})\,,
 \enspace &\frac{z}{u}&=(n/u)^{7/8}\le2^{-7}\,,
       \label{eq:D-step-0272}\\
 \frac{m_0}{u}
 &\le\max\{u^{-1/2},(n/u)^{7/8}\}+u^{-1}
 \le\frac1{16}+\frac1{256}<\frac12\,,
       \label{eq:D-step-0273}\\
 \tau_s&\ge4z\ge4C\,,
 \enspace &\frac{\tau^{\max}}{u}
 &=2^{7/8}u^{-1/8}+4(n/u)^{7/8}
 \le\frac18+\frac1{32}<1\,.
 \label{eq:SCr-reference-thresholds}
\end{alignat}
Equation \eqref{eq:D-step-0271} uses $u\ge N\ge K_0n[\log(en)]^2$, $K_0\ge256$, and $n>C$.
Equation \eqref{eq:D-step-0272} uses $m_0\ge\sqrt u$, $K_0\ge[S(C_{\beta,\mathrm{sh}})+1]^2$, and $u/n\ge256$.
Equation \eqref{eq:D-step-0273} uses $\lceil x\rceil\le x+1$ and the two bounds on $u^{-1/2}$ and $(n/u)^{7/8}$.

Equation \eqref{eq:SCr-reference-thresholds} uses $z\ge C$, $u\ge16^8$, and $n/u\le2^{-8}$ in the explicit maximum buffer.
In particular, \eqref{eq:D-step-0272} and \eqref{eq:D-step-0273}
give $n\le m_0\le u/2$.

For dyadic starts $v_j=2^jm_0\le2u$, freeze the cumulative loss estimates and corruption
compensators before their first crossing of
$Q_s>v_j^{7/8}+4z$ in $[v_j,2v_j]$. The stopped version of
\eqref{eq:SC-global-envelope} in \Cref{lem:SC-global}, followed by Markov's inequality, gives
\begin{align}
\sP(\exists s\in[v_j,2v_j]:Q_s>v_j^{7/8}+4z)
 &\le\frac{A_g(2v_j)^{\kappa P}+B_gn^{P-q}(2v_j)^q}
               {(v_j^{7/8}+4z)^P}
       \label{eq:D-step-0274}\\
 &\le2^{\kappa P}A_gv_j^{-\eta}
             +\frac{2^qB_gn^{P-q}v_j^q}{(4z)^P}\,,
       \label{eq:D-step-0275}\\
 \sum_{j\ge0}2^{\kappa P}A_g(2^jm_0)^{-\eta}
 &\le2^{\kappa P+1}A_gm_0^{-\eta}
 \le2^{\kappa P+1}A_gu^{-\eta/2}\,,
       \label{eq:D-step-0276}\\
 \sum_{j:v_j\le2u}\frac{2^qB_gn^{P-q}v_j^q}{(4z)^P}
 &\le\frac{2^{2q+1}B_gn^{P-q}u^q}
                  {4^Pn^{7P/8}u^{P/8}}
       \label{eq:D-step-0277}\\
 &=2^{2q+1-2P}B_g(n/u)^{P/8-q}
 \le B_g(n/u)^{r_-}\,,
       \label{eq:D-step-0278}\\
 \sP(I_{2u}=0)
 &\le2^{\kappa P+1}A_gu^{-\eta/2}+B_g(n/u)^{r_-}\,.
 \label{eq:SCr-reference-preparation-failure}
\end{align}
Equation \eqref{eq:D-step-0274} freezes the raw field and compensator jointly at the first crossing and applies Markov with \eqref{eq:SC-global-envelope} in \Cref{lem:SC-global}.
Equation \eqref{eq:D-step-0275} bounds the two denominator contributions separately by $v_j^{7/8}$ and $4z$.
Equation \eqref{eq:D-step-0276} sums the first dyadic tail and uses $m_0\ge\sqrt u$.

Equation \eqref{eq:D-step-0277} sums the increasing powers up to $v_j\le2u$ and substitutes $z^P=n^{7P/8}u^{P/8}$.
Equation \eqref{eq:D-step-0278} collects the powers into $(n/u)^{r_-}$ and uses $2q+1-2P\le0$.
Equation \eqref{eq:SCr-reference-preparation-failure} adds both dyadic sums to bound the complete preparation failure.
The geometric sums use $\eta>1$ and $q\ge1$. In equation \eqref{eq:D-step-0277},
$\sum_{j:v_j\le2u}v_j^q\le2(2u)^q$.
Any final dyadic block extending beyond $2u$ uses the stipulated
zero-additional-corruption continuation of the same process.

\paragraph{A Prepared Inverse Moment for Each Positive-Gap Action}
For $\Delta(a)>0$, set
$Y_s(a)=X_s(a)-E_s(a)+2C+2\tau_s$ and
$B_s(a)=4C-2A_s^c-E_s(a)$. The inequalities
$|E_s(a)|\le2A_s^c$ and $A_s^c\le C$ imply, on $\{I_s=1\}$,
\begin{align}
0\le4(C-A_s^c)&\le B_s(a)\le4C\,,
       \label{eq:D-step-0279}\\
 Y_s(a)&=W_s(a)+2\tau_s+B_s(a)\ge\tau_s>0\,,
       \label{eq:D-step-0280}\\
 X_s(a)+2M_s
 &=W_s(a)+2Q_s-2(C-A_s^c)
 \le W_s(a)+2\tau_s\le Y_s(a)\,.
 \label{eq:SCr-reference-positive-function}
\end{align}
Equation \eqref{eq:D-step-0279} uses $|E_s(a)|\le2A_s^c$ to bound $B_s(a)=4C-2A_s^c-E_s(a)$ from both sides.
Equation \eqref{eq:D-step-0280} substitutes the compensated positive field and uses survival and $B_s(a)\ge0$.
Equation \eqref{eq:SCr-reference-positive-function} substitutes $X_s=W_s+2(C-A_s^c)$ and $M_s=Q_s-2(C-A_s^c)$, then uses $Q_s\le\tau_s$.
Define $e_s(a)=(a-a^*)^\top c_s^{\rm vec}$ and
$\delta_s=\tau_{s+1}-\tau_s=(s+1)^{7/8}-s^{7/8}$.
Before killing the process, the increment $\xi_s=Y_{s+1}(a)-Y_s(a)$
satisfies
\begin{align}
\xi_s&=D_s(a)-e_s(a)+2\delta_s\,,
 &\E_{s-1}[\xi_s]&=\Delta(a)+2\delta_s\,,
       \label{eq:D-step-0281}\\
 |\xi_s|
 &\le2\beta_s+Y_s(a)/B+|e_s(a)|+2\delta_s
 \le3\beta_s+Y_s(a)/B\le3Y_s(a)/B\,,
       \label{eq:D-step-0282}\\
 \E_{s-1}[\xi_s^2]
 &=\operatorname{Var}_{s-1}(D_s(a))+(\Delta(a)+2\delta_s)^2
       \label{eq:D-step-0283}\\
 &\le8\beta_s+2Y_s(a)^2/(Bs)+(2+2\delta_s)^2
 \le10\beta_s+2Y_s(a)^2/(Bs)\,.
 \label{eq:SCr-reference-compensated-increments}
\end{align}
Equation \eqref{eq:D-step-0281} subtracts consecutive definitions of $Y_s(a)$ and cancels $e_s(a)$ in the conditional mean.
Equation \eqref{eq:D-step-0282} inserts the raw update magnitude, uses $|e_s(a)|\le4$ and $\delta_s\le7/8$, and applies the Taylor threshold.
Equation \eqref{eq:D-step-0283} uses the conditional mean $\Delta(a)$ after compensation and writes the second moment as variance plus squared mean.

Equation \eqref{eq:SCr-reference-compensated-increments} bounds variance by the raw second moment, uses $\Delta(a)\le2$, and absorbs $(2+2\delta_s)^2$ into $2\beta_s$.
Here $|e_s(a)|\le2\kappa_s\le4$, $\delta_s\le1$,
$\beta_s\ge17$, and $s\ge S(C_{\beta,\mathrm{sh}})$ gives
$3\beta_s\le s^{7/8}/B\le Y_s(a)/B$.
Thus the compensated variable is positive even before the next killing
step, and \eqref{eq:SHr-inverse-taylor} in \Cref{lem:SH-global-inverse} applies with its drift
$\Delta(a)+2\delta_s$. Substituting these inputs and then taking
expectations yields
\begin{align}
v_s(a)&=\E[I_sY_s(a)^{-P}]\,,
       \label{eq:D-step-0284}\\
 v_{s+1}(a)&\le(1+P/(2s))v_s(a)
                     -P\Delta(a)v_s(a)^{1+1/P}\,,
       \label{eq:D-step-0285}\\
 v_{s+1}(a)^{-1/P}
 &\ge\frac{s}{s+1}v_s(a)^{-1/P}+\Delta(a)\,,
       \label{eq:D-step-0286}\\
 v_u(a)^{-1/P}
 &\ge\frac{\Delta(a)}u\sum_{s=m_0}^{u-1}(s+1)
 \ge\frac{u\Delta(a)}4\,.
 \label{eq:SCr-reference-prepared-recursion}
\end{align}
Equation \eqref{eq:D-step-0284} defines the killed inverse moment at deterministic time $s$.
Equation \eqref{eq:D-step-0285} inserts the compensated drift and second moment into the inverse Taylor formula, then uses $I_{s+1}\le I_s$ and Jensen.
Equation \eqref{eq:D-step-0286} applies the tangent inequality to the inverse root and weakens $1-1/(2s)$ to $s/(s+1)$.

Equation \eqref{eq:SCr-reference-prepared-recursion} unrolls the root recursion, drops only its nonnegative initial term, evaluates the arithmetic sum, and uses $m_0\le u/2$.
The root inequality and its telescoping are exactly
\eqref{eq:SHr-inverse-root-solution} in \Cref{lem:SH-global-inverse}, now with starting index $m_0$
and drift $\Delta(a)$; $m_0\le u/2$ gives equation \eqref{eq:SCr-reference-prepared-recursion}.
A zero killed moment stays zero and already satisfies the desired bound.
Consequently
\begin{equation}
\label{eq:SC-prepared-inverse}
 \E[I_u(Y_u(a))^{-P}]\le\left(\frac4{u\Delta(a)}\right)^P\,.
\end{equation}
No minimum positive gap over $\mathcal K$ was used in this calculation.

\paragraph{Reference Errors Whose True Gap Is Very Small}
Let $\epsilon_u=\zeta[4\log(eu)]^{-1/8}$.
Since $\epsilon_u\le\Delta_{\rm pre}$, every suboptimal atom with
$\Delta(b_j)<\epsilon_u$ has $j>j^*$. Its priority then satisfies
\begin{align}
\|b_j-a^*\|_{\mathcal S}
 &\le\Delta(b_j)/\varsigma\,,
       \label{eq:D-step-0287}\\
 h_j&\ge\exp(\|b_j-a^*\|_{\mathcal S}^{-8})
 \ge\exp((\varsigma/\Delta(b_j))^8)
       \label{eq:D-step-0288}\\
 &>\exp((\varsigma/\zeta)^8\,4\log(eu))
 \ge(eu)^4\,.
 \label{eq:SCr-reference-small-priority}
\end{align}
Equation \eqref{eq:D-step-0287} rearranges sharpness for the selected physical reference.
Equation \eqref{eq:D-step-0288} keeps the reinforced prior term associated with $a^*$ and substitutes the distance bound.
Equation \eqref{eq:SCr-reference-small-priority} substitutes the cutoff $\epsilon_u=\zeta[4\log(eu)]^{-1/8}$ and uses $\zeta\le\varsigma$.
For any such atom chosen as $b_{r_s}$ at $s\in[u,2u]$, the reference
inequality \eqref{eq:SH-reference-ineq} in \Cref{lem:SH-anchor} implies
\begin{align}
X_s(b_{r_s})&\le\beta_s(h^*-h_j)
 \le-\tfrac12\beta_su^4\le-\tfrac12u^{9/2}\,.
 \label{eq:SCr-reference-small-threshold}
\end{align}
This uses $u^4\ge2h^*$, which follows from $u\ge K_0$, and
$\beta_s\ge\sqrt u$.
Freeze $X$ before its first crossing of
$\max_{\mathcal K}|X_s|\ge u^{9/2}/2$ in this block, and denote the
frozen field by $\widetilde X$. The stopped absolute estimate gives
\begin{align}
\sP(\exists s\in[u,2u]:b_{r_s}\ne a^*,\ \Delta(b_{r_s})<\epsilon_u)
 &\le\frac{\E[\max_{\mathcal K}|\widetilde X_{2u}|^P]}
                {(u^{9/2}/2)^P}
       \label{eq:D-step-0289}\\
 &\le\frac{R^P(2u)^P}{(u^{9/2}/2)^P}
 =(4R)^Pu^{-7P/2}\,.
 \label{eq:SCr-reference-small-probability}
\end{align}
Equation \eqref{eq:D-step-0289} freezes the full-history cumulative loss estimate at the first small-gap reference error and applies Markov at the deterministic endpoint $2u$.
Equation \eqref{eq:SCr-reference-small-probability} inserts the stopped absolute moment $R^P(2u)^P$ and evaluates the quotient of the two powers.
The large threshold is detected before the current draw, so this
freezing uses the stopped estimate rather than an unjustified
substitution of a random time into a deterministic moment bound.

\paragraph{One Stopping Calculation for Every Other Reference Error}
Set $D_u=\mathcal K\cap\{\Delta\ge\epsilon_u\}$.
If $D_u$ is empty, the event considered in this paragraph is empty.
Otherwise let $j=\dim\operatorname{aff}(D_u)$ and let $\sigma_u$ be
normalized relative volume on $D_u$.
Starting at $u$, kill the process if $I_s=0$ before success, and freeze
its entire affine function $Y_s$ at the first round $s\in[u,2u]$
for which $I_s=1$ and $b_{r_s}\in D_u$ is a wrong reference.
Let $\mathcal E$ denote success and let $\rho$ be that round on
$\mathcal E$. For the fixed endpoint $2u$, let $\widetilde Y_{2u}$ be
the field frozen at success; the inverse process is assigned value
zero on paths killed before success.
To make the stopping calculation formal, extend $\rho$ to be the
first successful round, or $+\infty$ if none occurs, and let
$\kappa_0=\inf\{s\in[u,2u]:I_s=0\}$, with $\inf\varnothing=+\infty$.
For $u\le s\le2u$, define the nonnegative killed-or-frozen variable
\begin{align}
K_s(a)
 &=\ind\{\rho\le s,\ \rho<\kappa_0\}Y_\rho(a)^{-P}
       +\ind\{s<\rho\wedge\kappa_0\}Y_s(a)^{-P}\,,
       \label{eq:D-step-0290}\\
 K_u(a)&=I_uY_u(a)^{-P}\,,
 &\ind_{\mathcal E}\widetilde Y_{2u}(a)^{-P}&\le K_{2u}(a)\,.
 \label{eq:SCr-reference-stopped-variable}
\end{align}
Equation \eqref{eq:D-step-0290} defines the process by retaining its attained value after the target reference event, killing it at a competing failure, and otherwise updating it.
Equation \eqref{eq:SCr-reference-stopped-variable} evaluates that definition at the starting time and compares its terminal value with the event-restricted jointly frozen inverse field.
Each indicator used before round $s$ is $\cF_{s-1}$-measurable.
Dropping the nonpositive drift term in the conditional inverse
recursion, and assigning value zero at a failure, gives
\begin{align}
\E_{s-1}[K_{s+1}(a)]
 &\le\ind\{\rho\le s,\ \rho<\kappa_0\}Y_\rho(a)^{-P}
       +\ind\{s<\rho\wedge\kappa_0\}
                         (1+P/(2s))Y_s(a)^{-P}
       \label{eq:D-step-0291}\\
 &\le(1+P/(2s))K_s(a)\,,
       \label{eq:D-step-0292}\\
 \E[\ind_{\mathcal E}\widetilde Y_{2u}(a)^{-P}]
 &\le\E[K_{2u}(a)]
 \le\prod_{s=u}^{2u-1}(1+P/(2s))\,\E[K_u(a)]
       \label{eq:D-step-0293}\\
 &\le\exp\!\left(\frac P2\sum_{s=u}^{2u-1}\frac1s\right)
                         \left(\frac4{u\Delta(a)}\right)^P
       \label{eq:D-step-0294}\\
 &\le2^P\left(\frac4{u\Delta(a)}\right)^P
 =\left(\frac8{u\Delta(a)}\right)^P\,.
 \label{eq:SCr-reference-stopped-inverse}
\end{align}
Equation \eqref{eq:D-step-0291} conditions on the two disjoint predictable states and uses the inverse Taylor inequality only on the still-active state.
Equation \eqref{eq:D-step-0292} uses the nonnegativity of the retained stopped value to factor $1+P/(2s)$.
Equation \eqref{eq:D-step-0293} iterates the deterministic-time conditional inequality and applies the terminal comparison in \eqref{eq:SCr-reference-stopped-variable}.

Equation \eqref{eq:D-step-0294} uses $\log(1+x)\le x$ and substitutes the prepared inverse bound \eqref{eq:SC-prepared-inverse}.
Equation \eqref{eq:SCr-reference-stopped-inverse} uses $\sum_{s=u}^{2u-1}s^{-1}\le2\log2$ and combines the factors $2^P4^P=8^P$.
Freezing leaves the success term in $K_s(a)$ unchanged, and its
multiplier is at least one. The entire affine function, including
$E_s$ and $\tau_s$, is frozen, not just the observed cumulative loss estimate.

At a successful round, the reference inequality and the threshold for
$h^*$ imply
\begin{align}
X_\rho(b_{r_\rho})&\le\beta_\rho h^*
 \le C_{\beta,\mathrm{sh}} h^*\sqrt\rho\le\rho^{7/8}\le\tau_\rho\,,
       \label{eq:D-step-0295}\\
 \min_{D_u}Y_\rho
 &\le Y_\rho(b_{r_\rho})
 =X_\rho(b_{r_\rho})-E_\rho(b_{r_\rho})+2C+2\tau_\rho
 \le3\tau_\rho+4C\le4\tau^{\max}\,,
       \label{eq:D-step-0296}\\
 \max_{D_u}Y_\rho
 &\le R_\rho^X+4C+2\tau_\rho
 \le R_\rho^X+3\tau^{\max}\,.
 \label{eq:SCr-reference-stopped-range}
\end{align}
Equation \eqref{eq:D-step-0295} uses penalized reference optimality, $\beta_\rho\le C_{\beta,\mathrm{sh}}\sqrt\rho$, and the clock threshold $C_{\beta,\mathrm{sh}}h^*\le\rho^{3/8}$.
Equation \eqref{eq:D-step-0296} evaluates the positive field at the selected action, bounds $|E_\rho|\le2C$, and uses $4C\le\tau^{\max}$.
Equation \eqref{eq:SCr-reference-stopped-range} bounds the entire positive field by the raw absolute maximum and uses the same budget and buffer bounds.
The condition $u\ge(2C_{\beta,\mathrm{sh}}h^*)^{8/3}$ is stronger than the condition
$C_{\beta,\mathrm{sh}} h^*\le\rho^{3/8}$ used in equation \eqref{eq:D-step-0295}.
Let $\widetilde R=\max_{\mathcal K}|\widetilde X_{2u}|$, with the
same success freezing, and contract $D_u$ toward a minimizer of
$Y_\rho$ by $\lambda=\tau^{\max}/(\widetilde R+10\tau^{\max})$.
On its image, $Y_\rho\le8\tau^{\max}$. Thus
\begin{align}
\ind_{\mathcal E}
 &\le8^p(\tau^{\max})^{p-j}
       (\widetilde R+10\tau^{\max})^j
       \ind_{\mathcal E}\int\widetilde Y_{2u}(a)^{-p}\sigma_u(da)\,,
       \label{eq:D-step-0297}\\
 \sP(\mathcal E)
 &\le8^p(\tau^{\max})^{p-j}
       \big[\E[(\widetilde R+10\tau^{\max})^{2j}]\big]^{1/2}
       \left[\int\E[\ind_{\mathcal E}\widetilde Y_{2u}(a)^{-2p}]
                                  \sigma_u(da)\right]^{1/2}
       \label{eq:D-step-0298}\\
 &\le8^p(\tau^{\max})^{p-j}[(2R+10)u]^j
                                      (8/(u\epsilon_u))^p
       \label{eq:D-step-0299}\\
 &\le64^p(2R+10)^d\epsilon_u^{-p}
                              (\tau^{\max}/u)^{p-d}\,.
 \label{eq:SC-reference-prebound}
\end{align}
Equation \eqref{eq:D-step-0297} applies the affine contraction at the stopped minimizer and integrates the inverse field on the contracted set.
Equation \eqref{eq:D-step-0298} takes expectation, uses Cauchy--Schwarz, and applies Jensen to the normalized integral.
Equation \eqref{eq:D-step-0299} inserts the jointly stopped absolute moment, $\tau^{\max}\le u$, and \eqref{eq:SCr-reference-stopped-inverse} with $2p=P$ and $\Delta\ge\epsilon_u$.

Equation \eqref{eq:SC-reference-prebound} collects $8^p8^p=64^p$, then uses $j\le d$ and $\tau^{\max}/u\le1$.
This is volume scaling followed by Cauchy--Schwarz, Jensen, the
stopped bound $\|\widetilde R\|_{L^P}\le2Ru$, and
\eqref{eq:SCr-reference-stopped-inverse}, with $2p=P$.
It bounds the probability of a reference error anywhere in the block;
there is no sum of $u$ single-round probabilities.

\paragraph{Reducing the Logarithmic and Mixed Powers}
Let $b_1=p-d$ and
$b_2=7(p-d)/8-p/64$. Substitution of $\epsilon_u$ and
$\tau^{\max}$ into \eqref{eq:SC-reference-prebound} gives
\begin{align}
\epsilon_u^{-p}(\tau^{\max}/u)^{b_1}
 &=\zeta^{-p}4^{p/8}[\log(eu)]^{p/8}
        \{2^{7/8}u^{-1/8}+4(n/u)^{7/8}\}^{b_1}
       \label{eq:D-step-0300}\\
 &\le\zeta^{-p}4^{p/8}2^{b_1-1}[\log(eu)]^{p/8}
        \{2^{b_1}u^{-b_1/8}+4^{b_1}(n/u)^{7b_1/8}\}\,.
 \label{eq:SCr-reference-power-substitution}
\end{align}
Equation \eqref{eq:D-step-0300} substitutes both $\epsilon_u=\zeta[4\log(eu)]^{-1/8}$ and $\tau^{\max}=(2u)^{7/8}+4n^{7/8}u^{1/8}$ literally.
Equation \eqref{eq:SCr-reference-power-substitution} applies $(x+y)^{b_1}\le2^{b_1-1}(x^{b_1}+y^{b_1})$ and $2^{7b_1/8}\le2^{b_1}$.
The factors $[\log(eu)]^{p/8}u^{-b_1/8}$ and
$[\log(eu)]^{p/8}(n/u)^{7b_1/8}$ in
\eqref{eq:SCr-reference-power-substitution} satisfy
\begin{align}
[\log(eu)]^{p/8}u^{-b_1/8}
 &\le2^{p/8}u^{p/16-(p-d)/8}
 =2^{p/8}u^{-P/32+d/8}
 \le2^{p/8}u^{-a}\,,
       \label{eq:D-step-0301}\\
 r&=u/n\ge1\,,
       \label{eq:D-step-0302}\\
 \log(eu)&=L_n+\log r\le L_n(1+\log r)\le8L_nr^{1/8}\,,
       \label{eq:D-step-0303}\\
 [\log(eu)]^{p/8}(n/u)^{7b_1/8}
 &\le8^{p/8}L_n^{p/8}(n/u)^{b_2}
       \label{eq:D-step-0304}\\
 &\le8^{p/8}(N/u)^{b_2}L_n^{p/8-2b_2}
 \le8^{p/8}(N/u)^a\,.
 \label{eq:SCr-reference-power-reductions}
\end{align}
Equation \eqref{eq:D-step-0301} uses $\log(eu)\le2\sqrt u$, substitutes $b_1=p-d$, and uses $P/32-d/8\ge a$.
Equation \eqref{eq:D-step-0302} defines the block-to-budget ratio $r=u/n\ge1$.
Equation \eqref{eq:D-step-0303} uses $\log(eu)=\log(en)+\log r$, $L_n\ge1$, and $1+\log r\le8r^{1/8}$.

Equation \eqref{eq:D-step-0304} raises that logarithmic bound to $p/8$ and collects the power $b_2=7(p-d)/8-p/64$.
Equation \eqref{eq:SCr-reference-power-reductions} uses $N\ge nL_n^2$, $L_n\ge1$, $p/8-2b_2\le0$, $b_2\ge a$, and $N/u\le1$.
All these numerical inequalities follow from
$P=4096(d+1)$ and $a=P/128$.
Combining \eqref{eq:SC-reference-prebound} with
\eqref{eq:SCr-reference-power-substitution}--\eqref{eq:SCr-reference-power-reductions}
therefore yields
\begin{align}
\sP(\mathcal E)
 &\le2^{12P}(2R+10)^d\zeta^{-p}
                          \{u^{-a}+(N/u)^a\}\,,
       \label{eq:D-step-0305}\\
 \sP(\exists s\in[u,2u]:b_{r_s}\ne a^*)
 &\le2^{\kappa P+1}A_gu^{-\eta/2}+B_g(n/u)^{r_-}
       +(4R)^Pu^{-7P/2}+\sP(\mathcal E)
       \label{eq:D-step-0306}\\
 &\le\big[2^{\kappa P+1}A_g+B_g+(4R)^P
             +2^{12P}(2R+10)^d\zeta^{-p}\big]
                         \{u^{-a}+(N/u)^a\}
       \label{eq:D-step-0307}\\
 &\le A_r\{u^{-a}+(N/u)^a\}\,.
 \label{eq:SCr-reference-complete-bound}
\end{align}
Equation \eqref{eq:D-step-0305} inserts both power reductions in \eqref{eq:SC-reference-prebound} and bounds the combined coefficient by $2^{12P}(2R+10)^d\zeta^{-p}$.
Equation \eqref{eq:D-step-0306} adds preparation failure, the small-gap reference event, and the remaining large-gap event.
Equation \eqref{eq:D-step-0307} uses $\eta/2\ge a$, $r_-\ge a$, $7P/2\ge a$, $N\ge n$, and $u\ge N$.

Equation \eqref{eq:SCr-reference-complete-bound} substitutes the definition of $A_r$ in \eqref{eq:SC-reference-constant} in \Cref{def:D-cor-dictionary}.
For example, the larger coefficient in the power calculation is
at most $2^{6p+p/4+3b_1+3p/8}\le2^{12P}$.
The final substitution uses \eqref{eq:SC-maxima} in \Cref{def:D-cor-dictionary},
$\eta/2\ge a$, $r_-\ge a$, $7P/2\ge a$, and $n\le N$.
This proves \eqref{eq:SC-reference-block}.
\end{proof}
The reinforcement in \eqref{eq:SH-priority} in \GIAFTRL{} is used only in \eqref{eq:SCr-reference-small-priority} in \Cref{lem:SC-reference}.
It permits a logarithmic, rather than polynomial, gap cutoff. The
preparation, inverse stopping, and reference selection are analysis
objects; the algorithm does not perform these tests.

\subsubsection{Cumulative Loss Estimates under Arbitrary Predictable Corruption}
\begin{lemma}[Shifted loss-estimate envelopes needed before reference identification]\label{lem:SC-global}
Let $X_s(a)=\widehat L_{s-1,a}-\widehat L_{s-1,a^*}$ on $\mathcal K$ and
$M_s=\max_{\mathcal K}(-X_s)$. Set
\[
 W_s(a)=X_s(a)-2(C-A_s^c)\,,\qquad
 Q_s=M_s+2(C-A_s^c)=\max_{\mathcal K}(-W_s)\,.
\]
For $s\ge n$,
\begin{equation}\label{eq:SC-global-envelope}
 \E[\max_{\mathcal K}|X_s|^P]\le R^Ps^P\,,\qquad
 \E[Q_s^P]\le A_gs^{\kappa P}+B_gn^{P-q}s^q\,.
\end{equation}
These envelopes also hold for jointly frozen functions at deterministic
observation times, after a pre-update stopping time not earlier than $n$.
\end{lemma}
\begin{proof}
Let $b_s^c=C-A_s^c\ge0$, $e_s(a)=(a-a^*)^\top c_s^{\rm vec}$,
and $D_s(a)=X_{s+1}(a)-X_s(a)$.
The absolute calculation \eqref{eq:SHr-interface-absolute-input} in \Cref{lem:D-absolute-step} and \eqref{eq:SHr-G-induction} in \Cref{lem:SH-field-interface} uses only
$|\E_{s-1}[D_s(a)]|\le2$, not nonnegative drift. Thus the bounded
actual means and the unchanged update magnitude and variance bounds give
\begin{align}
\E[\int|X_s(a)|^P d\sigma(a)]&\le C_G(C_{\beta,\mathrm{sh}})s^P\,,
       \label{eq:D-step-0308}\\
 \E[\max_{a\in\mathcal K}|X_s(a)|^P]
 &\le8^P\E[\int|X_s(a)|^P d\sigma(a)]
 \le8^PC_G(C_{\beta,\mathrm{sh}})s^P=R^Ps^P\,.
 \label{eq:SCr-global-absolute-proof}
\end{align}
Equation \eqref{eq:D-step-0308} applies the absolute moment recurrence, whose drift hypothesis is $|\E_{s-1}[D_s]|\le2$ and does not require an uncorrupted mean.
Equation \eqref{eq:SCr-global-absolute-proof} applies the affine integral-to-maximum inequality and substitutes $R^P=8^PC_G(C_{\beta,\mathrm{sh}})$.
Here $\sigma$ is normalized relative volume on $\mathcal K$.
For the shifted function, the next compensator is
$b_{s+1}^c=b_s^c-\kappa_s$, so
\begin{align}
W_{s+1}(a)-W_s(a)&=D_s(a)+2\kappa_s\,,
       \label{eq:D-step-0309}\\
 \E_{s-1}[D_s(a)+2\kappa_s]
 &=\Delta(a)+e_s(a)+2\kappa_s\ge\Delta(a)\ge0\,,
       \label{eq:D-step-0310}\\
 W_s(a)+2Q_s&=X_s(a)+2M_s+2b_s^c\ge X_s(a)+2M_s\,,
       \label{eq:D-step-0311}\\
 |D_s(a)+2\kappa_s|
 &\le2\beta_s+(X_s(a)+2M_s)/B+2\kappa_s
       \label{eq:D-step-0312}\\
 &\le3\beta_s+(W_s(a)+2Q_s)/B\,,
       \label{eq:D-step-0313}\\
 \E_{s-1}[(D_s(a)+2\kappa_s)^2]
 &=\E_{s-1}[D_s(a)^2]+4\kappa_s\E_{s-1}[D_s(a)]+4\kappa_s^2
       \label{eq:D-step-0314}\\
 &\le8\beta_s+\frac{2(X_s(a)+2M_s)^2}{Bs}
                   +8\kappa_s+4\kappa_s^2
       \label{eq:D-step-0315}\\
 &\le10\beta_s+\frac{2(W_s(a)+2Q_s)^2}{Bs}\,.
 \label{eq:SCr-global-shift-increments}
\end{align}
Equation \eqref{eq:D-step-0309} subtracts consecutive remaining-budget shifts.
Equation \eqref{eq:D-step-0310} uses $e_s(a)\ge-2\kappa_s$ and the nonnegative gap under $\theta$.
Equation \eqref{eq:D-step-0311} substitutes $W_s=X_s-2b_s^c$ and $Q_s=M_s+2b_s^c$.

Equation \eqref{eq:D-step-0312} applies the triangle inequality to the raw increment and the predictable shift.
Equation \eqref{eq:D-step-0313} uses the shifted-level comparison and $2\kappa_s\le4\le\beta_s$.
Equation \eqref{eq:D-step-0314} expands the square before conditioning.

Equation \eqref{eq:D-step-0315} inserts the raw second moment and $|\E_{s-1}[D_s]|\le2$.
Equation \eqref{eq:SCr-global-shift-increments} uses $\kappa_s\le2$, $8\kappa_s+4\kappa_s^2\le32\le2\beta_s$, and the shifted-level comparison.
We used $|e_s(a)|\le2\kappa_s$, $\kappa_s\le2$,
$8\kappa_s+4\kappa_s^2\le32\le2\beta_s$, and
$W_s(a^*)=-2b_s^c\le0$, so $Q_s\ge0$.
Thus \eqref{eq:SHr-negative-interface} in \Cref{lem:D-negative-step} applies to $-W_s$, with
$\beta_s$ and $b'=C_{\beta,\mathrm{sh}}$, even though $W_s(a^*)$ need not be zero.
For $F_s=\int(-W_s)_+^Pd\sigma$ and $f_s=\E[F_s]$, the exact resulting
recurrence and initial bound are
\begin{align}
\E_{s-1}[F_{s+1}]&\le(1+v/s)F_s+J_F(C_{\beta,\mathrm{sh}})s^{\nu-1}\,,
       \label{eq:D-step-0316}\\
 f_n^{1/P}&\le\|\max|W_n|\|_{L^P}
 \le\|\max|X_n|\|_{L^P}+2C\le Rn+2C\le(R+2)n\,,
       \label{eq:D-step-0317}\\
 f_s&\le H_F(C_{\beta,\mathrm{sh}})s^\nu+(R+2)^Pn^P\prod_{i=n}^{s-1}(1+v/i)
       \label{eq:D-step-0318}\\
 &\le H_F(C_{\beta,\mathrm{sh}})s^\nu+2^v(R+2)^Pn^{P-v}s^v\,,
       \label{eq:D-step-0319}\\
 \|\max|W_s|\|_{L^P}&\le Rs+2C\le(R+2)s\qquad(s\ge n)\,.
 \label{eq:SCr-global-integral-envelope}
\end{align}
Equation \eqref{eq:D-step-0316} applies the linearized negative recurrence to the shifted field with its verified nonnegative drift.
Equation \eqref{eq:D-step-0317} bounds the initial integral by the maximum, applies Minkowski to the shift, and uses $C\le n$.
Equation \eqref{eq:D-step-0318} unrolls the inhomogeneous recurrence around the supersolution $H_F(C_{\beta,\mathrm{sh}})s^\nu$, retaining the complete initial moment.

Equation \eqref{eq:D-step-0319} uses the multiplicative bound $\prod_{i=n}^{s-1}(1+v/i)\le(2s/n)^v$.
Equation \eqref{eq:SCr-global-integral-envelope} applies Minkowski to the shift at time $s$ and uses $s\ge n\ge C$.
The product bound is the one explicitly derived in
\eqref{eq:SHr-normalized-integral-negative} in \Cref{lem:SH-normalized-moments}.
For $j=\dim\operatorname{aff}(\mathcal K)$, $c_j=2^P4^j$, and
$a_j=P/(P+j)$, apply \eqref{eq:C-aff-neg} in \Cref{lem:C-affine} to $-W_s$ and H\"older:
\begin{align}
Q_s^{P+j}&\le c_j(\max|W_s|)^jF_s\,,
       \label{eq:D-step-0320}\\
 \E[Q_s^P]
 &\le c_j^{a_j}\big[\E[(\max|W_s|)^P]\big]^{j/(P+j)}f_s^{a_j}
       \label{eq:D-step-0321}\\
 &\le c_j^{a_j}[(R+2)^Ps^P]^{j/(P+j)}
       [H_F(C_{\beta,\mathrm{sh}})s^\nu+2^v(R+2)^Pn^{P-v}s^v]^{a_j}
       \label{eq:D-step-0322}\\
 &\le c_j^{a_j}(R+2)^{Pj/(P+j)}H_F(C_{\beta,\mathrm{sh}})^{a_j}
                         s^{P(j+\nu)/(P+j)}
       \label{eq:D-step-0323}\\*
 &\quad+c_j^{a_j}(R+2)^{Pj/(P+j)}[2^v(R+2)^P]^{a_j}
                         n^{P-q_j}s^{q_j}\,,
 \qquad q_j=\frac{P(j+v)}{P+j}\,,
       \label{eq:D-step-0324}\\
 &\le A_gs^{\kappa P}+B_gn^{P-q}s^q\,.
 \label{eq:SCr-global-negative-proof}
\end{align}
Equation \eqref{eq:D-step-0320} applies the affine negative-cap inequality to the compensated field at a fixed history.
Equation \eqref{eq:D-step-0321} raises to $a_j=P/(P+j)$, takes expectation, and applies H\"older.
Equation \eqref{eq:D-step-0322} inserts the full absolute and negative-integral envelopes from \eqref{eq:SCr-global-integral-envelope}.

Equation \eqref{eq:D-step-0323} uses $(x+y)^{a_j}\le x^{a_j}+y^{a_j}$ and writes the particular-solution contribution.
Equation \eqref{eq:D-step-0324} writes the propagated initial-moment contribution with $q_j=P(j+v)/(P+j)$.
Equation \eqref{eq:SCr-global-negative-proof} uses the maxima in \eqref{eq:SC-maxima} in \Cref{def:D-cor-dictionary},
$P(j+\nu)/(P+j)\le\kappa P$, $q_j\le q$, and
$n^{P-q_j}s^{q_j}=n^P(s/n)^{q_j}\le n^P(s/n)^q$.
For $j=0$, omit the maximum factor and use the same formulas.

For a predictable pre-update stopping time $\upsilon\ge n$, define
$\bar X_s=X_{s\wedge\upsilon}$,
$\bar b_s^c=b_{s\wedge\upsilon}^c$, and
$\bar W_s=\bar X_s-2\bar b_s^c$.
Let $\bar G_s=\int|\bar X_s|^P d\sigma$ and
$\bar F_s=\int(-\bar W_s)_+^P d\sigma$.
Freezing the cumulative loss estimates and compensators together gives
\begin{align}
\E_{s-1}[\bar G_{s+1}]
 &\le\bar G_s+\ind\{s<\upsilon\}
       \left[\frac{a_G}{s}\bar G_s+c_0\bar G_s^{1-1/P}
       +c_G(C_{\beta,\mathrm{sh}})\sqrt s\,\bar G_s^{1-2/P}
       +D_G(C_{\beta,\mathrm{sh}})s^{(P-1)/2}\right]
       \label{eq:D-step-0325}\\
 &\le(1+a_G/s)\bar G_s+c_0\bar G_s^{1-1/P}
       +c_G(C_{\beta,\mathrm{sh}})\sqrt s\,\bar G_s^{1-2/P}
       +D_G(C_{\beta,\mathrm{sh}})s^{(P-1)/2}\,,
       \label{eq:D-step-0326}\\
 \E_{s-1}[\bar F_{s+1}]
 &\le\bar F_s+\ind\{s<\upsilon\}
                     [(v/s)\bar F_s+J_F(C_{\beta,\mathrm{sh}})s^{\nu-1}]
       \label{eq:D-step-0327}\\
 &\le(1+v/s)\bar F_s+J_F(C_{\beta,\mathrm{sh}})s^{\nu-1}\,,
       \label{eq:D-step-0328}\\
 \max|\bar W_s|&\le\max|\bar X_s|+2C\,.
 \label{eq:SCr-joint-freezing}
\end{align}
Equation \eqref{eq:D-step-0325} conditions on the predictable event $s<\upsilon$ and applies the raw absolute recurrence only while the field is active.
Equation \eqref{eq:D-step-0326} drops the indicator from its nonnegative increment upper bound.
Equation \eqref{eq:D-step-0327} applies the same predictable decomposition to the shifted field and its compensator, which are frozen together.

Equation \eqref{eq:D-step-0328} drops the indicator from the two nonnegative linear-recurrence terms.
Equation \eqref{eq:SCr-joint-freezing} uses that the jointly frozen remaining budget stays between zero and $C$.
Every summand removed with the indicator is nonnegative.
The induction in \eqref{eq:SHr-G-induction} in
\Cref{lem:SH-field-interface}, applied to
\eqref{eq:D-step-0326}, gives $\E[\bar G_s]\le C_G(C_{\beta,\mathrm{sh}})s^P$.
The frozen negative recurrence \eqref{eq:D-step-0328} starts from
$\E[\bar F_n]\le(R+2)^Pn^P$, exactly as in
\eqref{eq:D-step-0317}. The resulting maximum and negative-moment
bounds are
\begin{align}
\E[\max|\bar X_s|^P]
 &\le c_j\E[\bar G_s]\le 8^PC_G(C_{\beta,\mathrm{sh}})s^P=R^Ps^P\,,
       \label{eq:D-frozen-absolute-maximum}\\
 \E[\bar F_s]
 &\le H_F(C_{\beta,\mathrm{sh}})s^\nu+2^v(R+2)^Pn^{P-v}s^v\,,
       \label{eq:D-step-0329}\\
 \E\big[(\max(-\bar W_s))^P\big]
 &\le c_j^{a_j}[(R+2)^Ps^P]^{j/(P+j)}
       [H_F(C_{\beta,\mathrm{sh}})s^\nu+2^v(R+2)^Pn^{P-v}s^v]^{a_j}
       \label{eq:D-step-0330}\\
 &\le A_gs^{\kappa P}+B_gn^{P-q}s^q\,.
 \label{eq:SCr-frozen-global-envelopes}
\end{align}
Equation \eqref{eq:D-frozen-absolute-maximum} applies \eqref{eq:C-aff-abs}
in \Cref{lem:C-affine} before expectation and substitutes
$c_j\le8^P$ and $R^P=8^PC_G(C_{\beta,\mathrm{sh}})$.
Equation \eqref{eq:D-step-0329} unrolls \eqref{eq:D-step-0328} for the
jointly frozen field with the deterministic initial bound at $n$.
Equation \eqref{eq:D-step-0330} applies affine contraction and H\"older and substitutes both frozen envelopes.
Equation \eqref{eq:SCr-frozen-global-envelopes} uses the same exponent and coefficient comparisons as \eqref{eq:SCr-global-negative-proof}, at deterministic $s$.
These are bounds at deterministic $s$ for the jointly frozen fields;
no unstopped moment bound is evaluated at a random time.
\end{proof}

\clearpage
\section{Proof of Theorem~\ref*{thm:main-sharp-lower}}\label{app:sharp-lower-proof}
The logarithmic claim follows from \Cref{prop:SH-curve-lower}; the corruption claim is proved directly on the arch. The geometric verification needed for the arch is deferred until after that proof.
\subsection{A Genuinely Nonseparated Curve: Complete Coverage and Logarithmic Necessity}\label{sec:SH-example}
Consider the curve already used in \Cref{prop:C-eight}:
\begin{equation}\label{eq:SH-curve}
 \cA=\left\{a(s)=\left(-\frac12+\frac s4,\frac{s+s^2}{4}\right):0\le s\le1\right\}\,.
\end{equation}
For every $0<\tau\le1$, $\theta_\tau=(\tau,0)$ has the unique optimal action
$a(0)$, physical minimum gap zero, and $\cctr=8/\tau$.
The relevant geometric and sampling allocation calculations are as follows:
\begin{align}
 \Delta(a(s))&=(a(s)-a(0))^\top\theta_\tau=\tau s/4\,,
 &a(s)-a(0)&=\tfrac s4(1,1+s)\,,\notag\\
 \|a(s)-a(0)\|&=\tfrac s4\sqrt{1+(1+s)^2}
       \le\tfrac{\sqrt5 s}{4}\,,
 &\Delta(a(s))&\ge\tfrac\tau{\sqrt5}\|a(s)-a(0)\|\,.
 \notag
\end{align}
Thus the instance is sharp, and $\Delta(a(s))\downarrow0$ as
$s\downarrow0$. We compute $\cctr$ using
\Cref{lem:centered-projection-lift}. The direction of $a(0)$ is the
first coordinate. For any centered feasible allocation $n$,
let $M(n)=\int a_2(s)^2n(ds)$ and
$c_{\mathrm{ctr}}^{n}=c_{\mathrm{ctr}}^{n}(\cA,\theta_\tau)=\int\Delta(a(s))n(ds)$. Projection and the cost comparison give
\begin{align}
 \frac{a_2(s)}{\Delta(a(s))}&=\frac{1+s}{\tau}\quad(s>0)\,,
 &M(n)&\ge\sup_{0<s\le1}2\left(\frac{1+s}{\tau}\right)^2
          =\frac8{\tau^2}\,,\notag\\
 a_2(s)^2&=\frac{s^2(1+s)^2}{16}
       \le\frac s4=\frac{\Delta(a(s))}{\tau}\,,
 &M(n)&\le\frac{c_{\mathrm{ctr}}^{n}}{\tau}\,,\notag\\
 c_{\mathrm{ctr}}^{n}&\ge\tau M(n)\ge8/\tau\,,
 &n_0&=\frac{32}{\tau^2}\delta_1\,,\notag\\
 M(n_0)&=\frac{32}{\tau^2}\left(\frac12\right)^2
          =\frac8{\tau^2}\,,
 &c_{\mathrm{ctr}}^{n_0}&=\frac{32}{\tau^2}\frac\tau4=\frac8\tau\,,\notag\\
 \cctr&\ge8/\tau\,,
 &\cctr&\le(1+\varepsilon)c_{\mathrm{ctr}}^{n_0}
          =(1+\varepsilon)8/\tau\quad(\varepsilon>0)\,.
 \notag
\end{align}
The allocation $n_0$ satisfies the projected constraints because
$M(n_0)=8/\tau^2$; the endpoint $s=0$ imposes a zero constraint.
It need not be centered feasible itself. The final inequality uses
its finite centered lift from \Cref{lem:centered-projection-lift}.
Letting $\varepsilon\downarrow0$ proves $\cctr=8/\tau$.
For every $s_0\in(0,1)$, the linear functional
$F_{s_0}(x,y)=y-(1+2s_0)x$ has the exact difference
\begin{align}
 F_{s_0}(a(s))-F_{s_0}(a(s_0))
 &=\frac{(s-s_0)^2}{4}>0\quad(s\ne s_0)\,.
 \notag
\end{align}
Each $a(s_0)$ is therefore exposed and belongs to every compact
set generating the same convex hull: the exposing functional attains
its unique extremum on that generator. Infinitely many such points
exclude a polytope. Taking $s_0\downarrow0$ gives suboptimal generator
points with gaps tending to zero, as in
\Cref{prop:C-eight}.

An explicit centered sampling allocation puts mass $64/\tau^2$ at each
of $a(0)$ and $a(1)$. Direct matrix substitution gives
\begin{align}
 H_n
 &=\frac{64}{\tau^2}\left[
   \begin{pmatrix}1/4&0\\0&0\end{pmatrix}
   +\begin{pmatrix}1/16&-1/8\\-1/8&1/4\end{pmatrix}\right]
 =\tau^{-2}\begin{pmatrix}20&-8\\-8&16\end{pmatrix}\,,\notag\\
 H_n^{-1}
 &=\frac{\tau^2}{256}\begin{pmatrix}16&8\\8&20\end{pmatrix}\,,\notag\\
 N_n&=128/\tau^2\,,\qquad c_{\mathrm{ctr}}^{n}=\frac{64}{\tau^2}\frac\tau4=16/\tau=2\cctr\,,\notag\\
 (a(s)-a(0))^\top H_n^{-1}(a(s)-a(0))
 &=\frac{s^2\tau^2}{4096}
        [16+16(1+s)+20(1+s)^2]\notag\\
 &=\frac{s^2\tau^2}{1024}(13+14s+5s^2)\,,\notag\\
 \frac{(a(s)-a(0))^\top H_n^{-1}(a(s)-a(0))}{\Delta(a(s))^2/2}
 &=\frac{[s^2\tau^2/1024](13+14s+5s^2)}{\tau^2s^2/32}
 =\frac{5s^2+14s+13}{32}\le1\quad(s>0)\,.
 \notag
\end{align}
At $s=0$ both sides of the centered constraint are zero.
For $x=\sum_i\lambda_i a(s_i)$, its extension to the convex hull is
\begin{align}
 \|x-a(0)\|_{H_n^{-1}}
 &=\left\|\sum_i\lambda_i(a(s_i)-a(0))\right\|_{H_n^{-1}}
 \le\sum_i\lambda_i\|a(s_i)-a(0)\|_{H_n^{-1}}\notag\\
 &\le\frac1{\sqrt2}\sum_i\lambda_i\Delta(a(s_i))
 =\frac{\Delta(x)}{\sqrt2}\,,\notag\\
 (x-a(0))^\top H_n^{-1}(x-a(0))&\le\Delta(x)^2/2\,.
 \label{eq:SHr-curve-centered-extension}
\end{align}
Thus \eqref{eq:SC-per-design} at $C=0$ is directly applicable to this
zero-minimum-gap example.

In fact the same geometry-only algorithm covers every parameter on this curve with a unique optimal action
and finite $\cctr$. For a general parameter
$\theta=(\theta_1,\theta_2)$, the restricted loss and its derivative are
\begin{align}
 a(s)^\top\theta
 &=-\theta_1/2+[(\theta_1+\theta_2)s+\theta_2s^2]/4\,,\notag\\
 \frac{d}{ds}a(s)^\top\theta
 &=(\theta_1+\theta_2+2\theta_2s)/4\,.
 \notag
\end{align}
At a unique interior minimum $r\in(0,1)$, one necessarily has
$\theta_2>0$ and $\theta_1=-(1+2r)\theta_2$. Substitution gives
\begin{align}
 \Delta(a(s))&=\theta_2(s-r)^2/4\,,
 &a(r)^\top\theta&=\theta_2(1/2+r-r^2/4)>0\,,\notag\\
 \frac{\Delta(a(s))}{\|a(s)-a(r)\|}
 &\le\frac{\theta_2(s-r)^2/4}{|s-r|/4}
 =\theta_2|s-r|\longrightarrow0\,.
 \notag
\end{align}
A unique endpoint minimum with zero one-sided derivative has the
same calculation with $r=0$ or $r=1$: respectively
$\theta_1=-\theta_2$ or $\theta_1=-3\theta_2$, necessarily
$\theta_2>0$, and the optimal losses are $\theta_2/2$ and
$5\theta_2/4$. These cases are nonsharp, so \Cref{prop:C-centered} makes their
centered information-acquisition complexity infinite.

For any other unique endpoint minimum, let $c_0$ or $c_1$ denote
the corresponding positive inward gap coefficient. The exact factorizations
and their lower bounds are
\begin{align}
 r=0:
 \enspace\Delta(a(s))&=\tfrac s4[(\theta_1+\theta_2)+\theta_2s]
                  \ge c_0s/4\,,
 &c_0&=\min\{\theta_1+\theta_2,\theta_1+2\theta_2\}>0\,,\notag\\
 r=1:
 \enspace\Delta(a(s))&=\tfrac{1-s}{4}[-\theta_1-2\theta_2-\theta_2s]
                  \ge c_1(1-s)/4\,,
 &c_1&=\min\{-\theta_1-2\theta_2,-\theta_1-3\theta_2\}>0\,,\notag\\
 \|a(s)-a(r)\|&\le(\sqrt{10}/4)|s-r|\,,
 &\Delta(a(s))&\ge(c_r/\sqrt{10})\|a(s)-a(r)\|\,.
 \notag
\end{align}
Positivity of the derivative at the minimizing endpoint and strict
suboptimality of the other endpoint make both entries in the relevant
minimum positive. The derivative bound
$\|a'(s)\|=\sqrt{1+(1+2s)^2}/4\le\sqrt{10}/4$ gives the last line.
Since the spanning symmetric body $\mathcal S$ contains
$r_{\mathcal S}\BB$ for some $r_{\mathcal S}>0$,
$\|x\|\ge r_{\mathcal S}\|x\|_{\mathcal S}$; thus the displayed
Euclidean sharpness is also sharpness in the required norm.
The convex-hull extension uses the same triangle-inequality calculation
as \eqref{eq:SHr-curve-centered-extension}.
\Cref{lem:SH-atom} includes both endpoints without requiring
the algorithm to know which one is optimal.

\begin{proposition}[Same-model logarithmic necessity on the curve]\label{prop:SH-curve-lower}
Use sign observations of the linear mean on \eqref{eq:SH-curve}.
For $0<\tau\le1$, let $\theta_\tau=(\tau,0)$. Every policy consistent on the sharp stochastic
instances on this curve obeys
\begin{equation}\label{eq:SH-curve-lower}
 \liminf_{T\to\infty}\frac{\Reg[T,\theta_\tau]}{\log T}
       \ge\frac7{16\tau}=\frac7{128}\cctr(\cA,(\tau,0))\,.
\end{equation}
\end{proposition}
\begin{proof}
\textbf{The two fixed bounded-feedback models.}
Let $\theta=\theta_\tau=(\tau,0)$, $\lambda=(\tau,-\tau)$, and let
$P_\theta^T,P_\lambda^T$ be the laws of the complete history,
including the learner's randomization. For $s\in[0,1]$,
\begin{align}
 \mu_\theta(s)&=a(s)^\top\theta=-\tau/2+\tau s/4\,,
 &\Delta_\theta(s)&=\mu_\theta(s)-\mu_\theta(0)=\tau s/4\,,\notag\\
 \mu_\lambda(s)&=a(s)^\top\lambda=-\tau/2-\tau s^2/4\,,
 &\Delta_\lambda(s)&=\mu_\lambda(s)-\mu_\lambda(1)
                                      =\tau(1-s^2)/4\,,\notag\\
 -\tfrac34&\le\mu_\theta(s),\mu_\lambda(s)\le0\,,
 &\mu_\theta(s)-\mu_\lambda(s)&=\tau(s+s^2)/4=\tau a_2(s)\,.
 \label{eq:SHr-curve-two-models}
\end{align}
Thus $a(0)$ and $a(1)$ are the respective unique optimal actions, and the
sign laws $Q_\mu$ in \Cref{lem:testing-tools} are valid.
The alternate instance is also sharp:
\begin{align}
 a'(s)&=(1/4,(1+2s)/4)\,,
 &\|a'(s)\|&\le\sqrt{10}/4\,,\notag\\
 \|a(s)-a(1)\|&\le(\sqrt{10}/4)(1-s)\,,
 &\Delta_\lambda(s)&=\tfrac\tau4(1-s^2)
                 \ge\tfrac\tau4(1-s)\,,\notag\\
 \Delta_\lambda(s)&\ge\frac\tau{\sqrt{10}}\|a(s)-a(1)\|\,.
 \notag
\end{align}
The triangle inequality extends this bound to the convex hull, and
norm equivalence gives sharpness in $\|\cdot\|_{\mathcal S}$.
Thus consistency applies to both fixed instances.

\paragraph{A History Event That Separates the Two Models}
Fix any $\alpha\in(0,1)$. Consistency gives
$\Reg[T,\theta]=o(T)$ and $\Reg[T,\lambda]=o(T^\alpha)$. Define
$E_T=\{\sum_{t=1}^T s_t\le T/2\}$,
$p_T=P_\theta^T(E_T)$, and $q_T=P_\lambda^T(E_T)$.
Substitution of the two gap formulas in \eqref{eq:SHr-curve-two-models}
into Markov's inequality gives
\begin{align}
 1-p_T
 &=P_\theta^T\!\left(\sum_{t=1}^T s_t>T/2\right)
 \le\frac2T\E_\theta\left[\sum_{t=1}^T s_t\right]
 =\frac{8\Reg[T,\theta]}{\tau T}=o(1)\,,\notag\\
 \ind_{E_T}\sum_{t=1}^T(1-s_t^2)
 &\ge\ind_{E_T}\sum_{t=1}^T(1-s_t)
 \ge(T/2)\ind_{E_T}\,,\notag\\
 \Reg[T,\lambda]
 &=\frac\tau4\E_\lambda\left[\sum_{t=1}^T(1-s_t^2)\right]
 \ge\frac{\tau T}{8}q_T\,,\notag\\
 q_T&\le\frac{8\Reg[T,\lambda]}{\tau T}
 =o(T^{\alpha-1})\,.
 \notag
\end{align}
Let $h(p)=-p\log p-(1-p)\log(1-p)$, with $0\log0=0$.
Binary data processing now yields the explicit entropy calculation
\begin{align}
 \KL(P_\theta^T, P_\lambda^T)
 &\ge p_T\log\frac{p_T}{q_T}
            +(1-p_T)\log\frac{1-p_T}{1-q_T}\notag\\
 &=p_T\log\frac1{q_T}+(1-p_T)\log\frac1{1-q_T}-h(p_T)\notag\\
 &\ge p_T\log\frac1{q_T}-\log2\notag\\
 &\ge(1-o(1))(1-\alpha)\log T-\log2\notag\\
 &=(1-\alpha-o(1))\log T\,.
 \notag
\end{align}
The last inequality holds for all sufficiently large $T$, since
$q_T=o(T^{\alpha-1})$.
If $q_T=0<p_T$, the KL is infinite and the lower inequality holds
without a logarithmic expansion. In the present sign models all
conditional observation probabilities are strictly positive, so the
finite upper bound below also rules out that case on a finite horizon.

\paragraph{The Information Paid for by Regret in the Stochastic Instance}
For every $s\in[0,1]$, the sign KL inequality gives
\begin{align}
 \KL(Q_{\mu_\theta(s)}, Q_{\mu_\lambda(s)})
 &\le\frac{(\mu_\theta(s)-\mu_\lambda(s))^2}
                {1-\mu_\lambda(s)^2}
 =\frac{\tau^2a_2(s)^2}{1-\mu_\lambda(s)^2}\notag\\
 &\le\frac{16}{7}\tau^2 a_2(s)^2\,,\notag\\
 a_2(s)^2
 &=\frac{s^2(1+s)^2}{16}
 \le\frac{s}{16}\cdot4=\frac s4\,.
 \notag
\end{align}
The denominator is at least $1-(3/4)^2=7/16$.
The last inequality uses $s(1+s)^2\le4$ on $[0,1]$.
The policy kernels are identical under both parameters and therefore
contribute zero to the adaptive KL chain rule. Consequently
\begin{align}
 \KL(P_\theta^T, P_\lambda^T)
 &=\E_\theta\left[\sum_{t=1}^T
       \KL(Q_{\mu_\theta(s_t)}, Q_{\mu_\lambda(s_t)})\right]\notag\\
 &\le\frac{16}{7}\tau^2\E_\theta\left[\sum_{t=1}^Ta_2(s_t)^2\right]
 \le\frac47\tau^2\E_\theta\left[\sum_{t=1}^Ts_t\right]\notag\\
 &=\frac47\tau^2\frac4\tau \Reg[T,\theta]
 =\frac{16}{7}\tau \Reg[T,\theta]\,,\notag\\
 \frac{\Reg[T,\theta]}{\log T}
 &\ge\frac7{16\tau}\,
       \frac{\KL(P_\theta^T, P_\lambda^T)}{\log T}
 \ge\frac7{16\tau}(1-\alpha-o(1))\,.
 \notag
\end{align}
Taking the limit inferior, then letting $\alpha\downarrow0$ and
substituting $\cctr(\cA,(\tau,0))=8/\tau$, proves
\begin{align}
 \liminf_{T\to\infty}\frac{\Reg[T,\theta_\tau]}{\log T}
 &\ge\frac7{16\tau}
 =\frac7{128}\cctr(\cA,(\tau,0))\,.
 \notag
\end{align}
Theorem~\ref{thm:main-nonseparated-upper} gives logarithmic regret for
\GIAFTRL{} on every fixed sharp stochastic instance, so it belongs to
this consistent policy class. The lower bound therefore calibrates the
same learner and bounded-feedback model. It does not assert optimality
of the upper bound's dimension multiplier or transient.
\end{proof}

\subsection{Corruption Lower Bounds within the Sharp, Genuinely Nonseparated Class}\label{sec:SC-sharp-lower}
The critical curves in \Cref{ch:critical} are nonsharp and cannot
justify a corruption term in \Cref{thm:main-nonseparated-upper}.
Here the difficult instances have sharp optimal actions, their centered information-acquisition complexity stays bounded,
and no compact generator can restore a positive minimum gap. The
small-effective-gap testing mechanism is related to \cite{Ito2021Corruption}; the
physical continuum, its geometric margin, and its information-acquisition complexity are
verified below rather than inferred from a finite-action embedding.

Use the fixed arch and fixed loss vectors in \eqref{eq:SC-arch}, whose sharpness and information-acquisition complexity are certified by \Cref{lem:SC-arch}.
\begin{proof}[Corruption part of \Cref{thm:main-sharp-lower}]
\textbf{Equal budgets and close sign experiments.}
Fix $T\ge\phi^{-2}$ and set $\varepsilon=1/(8\sqrt T)$.
For each $\sigma\in\{-1,1\}$ use the fixed loss vector
$\theta_\sigma=(\sigma\phi/2,1/2)$ and, for every $t\le T$, the
actual parameter $\theta_t^\sigma=(\sigma\varepsilon,1/2)$.
The norm and budget calculation is
\begin{align}
 0<\varepsilon&=\frac1{8\sqrt T}\le\frac\phi8\,,
 &c_t^{\rm vec}&=\theta_t^\sigma-\theta_\sigma
                  =(\sigma(\varepsilon-\phi/2),0)\,,\notag\\
 \|c_t^{\rm vec}\|_{\mathcal A_{\rm arch}}
 &=\sup_{-1\le x\le1}|\sigma(\varepsilon-\phi/2)x|
 =\phi/2-\varepsilon\,,\notag\\
 \sum_{t=1}^T\|c_t^{\rm vec}\|_{\mathcal A_{\rm arch}}
 &=T(\phi/2-\varepsilon)
 =\frac{\phi T}{2}-\frac{\sqrt T}{8}
 =C_{\phi,T}\,,\notag\\
 \frac{3\phi T}{8}&\le C_{\phi,T}\le\frac{\phi T}{2}\,.
 \notag
\end{align}
Both environments have deterministic corruption sequences with total
budget $C_{\phi,T}$.
Given the action $a(x)$, draw a sign with mean
$\mu_\sigma(x)=\sigma\varepsilon x+(1-x^2)/4$.
For every $x\in[-1,1]$,
\begin{align}
 |\mu_\sigma(x)|&\le\varepsilon+\tfrac14
 \le\tfrac{\phi}{8}+\tfrac14\le\tfrac{17}{64}<\tfrac12\,,\notag\\
 \mu_+(x)-\mu_-(x)&=2\varepsilon x\,,\notag\\
 \KL(Q_{\mu_+(x)}, Q_{\mu_-(x)})
 &\le\frac{(2\varepsilon x)^2}{1-\mu_-(x)^2}
 \le\frac{4\varepsilon^2x^2}{3/4}
 =\frac{16}{3}\varepsilon^2x^2
 \le\frac{16}{3}\varepsilon^2\,,\notag\\
 \KL(P_+^T, P_-^T)
 &=\E_+\left[\sum_{t=1}^T
       \KL(Q_{\mu_+(x_t)}, Q_{\mu_-(x_t)})\right]
 \le\frac{16}{3}T\varepsilon^2
 =\frac{16}{3}T\frac1{64T}=\frac1{12}\,,\notag\\
 \TV(P_+^T,P_-^T)
 &\le\sqrt{\tfrac12\KL(P_+^T, P_-^T)}
 \le\frac1{\sqrt{24}}<\frac12\,.
 \notag
\end{align}
The sign KL and total-variation inequalities are those in
\Cref{lem:testing-tools}. The history laws include the
learner's randomization; its conditional action kernels are the same
in both environments, so only the feedback terms enter the chain rule.
The common inputs $\phi,T,C_{\phi,T}$ do not change this calculation.

\paragraph{Testing Forces Actual-Comparator Regret}
The action $-\sigma e_1$ remains the unique optimal action under the actual loss vector. Directly
subtracting its actual loss yields
\begin{align}
 a(x)^\top\theta_t^\sigma-(-\sigma e_1)^\top\theta_t^\sigma
 &=\sigma\varepsilon x+\frac{1-x^2}{4}+\varepsilon\notag\\
 &=\varepsilon(1+\sigma x)+\frac{1-x^2}{4}\ge0\,,\notag\\
 \sigma x\ge0
 &\ \Longrightarrow\ %
 a(x)^\top\theta_t^\sigma-(-\sigma e_1)^\top\theta_t^\sigma
 \ge\varepsilon\,.
 \notag
\end{align}
For $E_t=\{x_t\ge0\}$, $E_t$ is a wrong-sign event under $+$,
and $E_t^c$ is a wrong-sign event under $-$. Thus
\begin{align}
 P_+^T(E_t)+P_-^T(E_t^c)
 &=1+P_+^T(E_t)-P_-^T(E_t)
 \ge1-\TV(P_+^T,P_-^T)\ge\tfrac12\,,\notag\\
 \Reg[T,+]+\Reg[T,-]
 &=\sum_{t=1}^T\left\{
    \E_+[\varepsilon(1+x_t)+(1-x_t^2)/4]
   +\E_-[\varepsilon(1-x_t)+(1-x_t^2)/4]\right\}\notag\\
 &\ge\varepsilon\sum_{t=1}^T
                       [P_+^T(E_t)+P_-^T(E_t^c)]
 \ge\frac{\varepsilon T}{2}\,,\notag\\
 \max_{\sigma=\pm1}\Reg[T,\sigma]
 &\ge\frac12(\Reg[T,+]+\Reg[T,-])
 \ge\frac{\varepsilon T}{4}
 =\frac{T}{32\sqrt T}=\frac{\sqrt T}{32}\notag\\
 &\ge\frac{\sqrt2}{32}\sqrt{C_{\phi,T}/\phi}\,.
 \notag
\end{align}
The last inequality substitutes $C_{\phi,T}\le\phi T/2$.
The comparator $-\sigma e_1$ is fixed throughout each environment.

\paragraph{A Prescribed Budget Cap and the Horizon Truncation}
Now let $C\ge\phi^{-1}$ and
$T\ge\lceil\phi^{-2}\rceil$, and set
$m=\min\{T,\lfloor2C/\phi\rfloor\}$.
Since $C/\phi\ge\phi^{-2}\ge1$,
\begin{align}
 \lfloor2C/\phi\rfloor
 &\ge2C/\phi-1\ge C/\phi\ge\phi^{-2}\,,\notag\\
 m&\ge\phi^{-2}\,,
 &m&\ge\min\{T,C/\phi\}\,,\notag\\
 \varepsilon_m&=\frac1{8\sqrt m}\,,
 &m(\phi/2-\varepsilon_m)&\le\frac{m\phi}{2}\le C\,.
 \notag
\end{align}
Use the preceding sign experiment on rounds $1,\ldots,m$, and set
$\theta_t^\sigma=\theta_\sigma$ on rounds $m+1,\ldots,T$.
The optimal action is still $-\sigma e_1$ on every round, so
\begin{align}
 \Reg[T,\sigma]
 &=\Reg[m,\sigma]
   +\sum_{t=m+1}^T\E_\sigma\!\left[
          \tfrac\phi2(1+\sigma x_t)+(1-x_t^2)/4\right]
 \ge \Reg[m,\sigma]\,,\notag\\
 \max_\sigma \Reg[T,\sigma]
 &\ge\max_\sigma \Reg[m,\sigma]
 \ge\frac{\sqrt m}{32}
 \ge\frac1{32}\min\{\sqrt T,\sqrt{C/\phi}\}\,.
 \notag
\end{align}
The length-$m$ testing argument applies to the prefix of any
horizon-$T$ policy, including a policy informed of $T$ and $C$:
subsequent observations cannot affect its prefix action kernels.
Finally, \eqref{eq:SC-arch-geometry} gives $\phi\le6\varsigma_\sigma$,
and therefore
\begin{align}
 \sqrt{C/\phi}&\ge\frac1{\sqrt6}\sqrt{C/\varsigma_\sigma}\,,\notag\\
 \min\{\sqrt T,\sqrt{C/\phi}\}
 &\ge\frac1{\sqrt6}\min\{\sqrt T,\sqrt{C/\varsigma_\sigma}\}\,.
 \notag
\end{align}
This retains the horizon truncation in the geometric-margin formulation.
\end{proof}

\begin{lemma}[A fixed sharp continuum with uniformly bounded centered information-acquisition complexity]\label{lem:SC-arch}
Let
\begin{equation}\label{eq:SC-arch}
 \mathcal A_{\rm arch}
 =\{a(x)=(x,(1-x^2)/2):-1\le x\le1\}\,,\qquad
 \theta_\sigma=(\sigma\phi/2,1/2)\,,\quad
 \sigma\in\{-1,1\}\,,\quad0<\phi\le1/8\,.
\end{equation}
This compact set spans $\R^2$, lies in the Euclidean unit ball, and has a
nonpolyhedral convex hull. Its unique optimal action is $a^*=-\sigma e_1$.
Every compact generator of its convex hull has zero minimum positive gap.
For the sharpness modulus in the norm
$\mathcal S=\Conv(\mathcal A_{\rm arch}\cup(-\mathcal A_{\rm arch}))$,
\begin{equation}\label{eq:SC-arch-geometry}
 \frac\phi6\le\varsigma_\sigma\le\frac\phi2\,,
 \qquad \frac{128}{25}\le
 \cctr(\mathcal A_{\rm arch},\theta_\sigma)\le10\,.
\end{equation}
\end{lemma}
\begin{proof}
\textbf{Geometry, exposed points, and the optimal action under the fixed loss vector.}
Define $y(x)=(1-x^2)/2$. The continuous parametrization is compact,
and the following identities verify spanning and boundedness:
\begin{align}
 a(-1)&=-e_1\,,
 &a(1)&=e_1\,,
 &a(0)&=e_2/2\,,\notag\\
 \|a(x)\|^2
 &=x^2+(1-x^2)^2/4
 =(1+2x^2+x^4)/4
 =\frac{(1+x^2)^2}{4}\le1\,.
 \notag
\end{align}
For each $x_0\in(-1,1)$, the linear functional
$F_{x_0}(x,y)=x_0x+y$ satisfies
\begin{align}
 F_{x_0}(a(x))
 &=x_0x+(1-x^2)/2
 =\frac{1+x_0^2}{2}-\frac{(x-x_0)^2}{2}\,,\notag\\
 F_{x_0}(a(x_0))-F_{x_0}(a(x))
 &=\frac{(x-x_0)^2}{2}>0\qquad(x\ne x_0)\,.
 \notag
\end{align}
Thus every interior curve point is exposed, and the convex hull has
infinitely many exposed points, so it is not a polytope.
If a compact set $G$ has the same convex hull, then
\begin{align}
 \max_{g\in G}F_{x_0}(g)
 &=\max_{z\in\operatorname{conv}(G)}F_{x_0}(z)
 =\max_{z\in\operatorname{conv}(\mathcal A_{\rm arch})}F_{x_0}(z)
 =F_{x_0}(a(x_0))\,,\notag\\
 \operatorname*{argmax}_{z\in\operatorname{conv}(\mathcal A_{\rm arch})}
                   F_{x_0}(z)&=\{a(x_0)\}
 \quad\Longrightarrow\quad a(x_0)\in G\,.
 \notag
\end{align}
Compactness ensures that the first maximum is attained.
The gap from $a^*=-\sigma e_1$ is exactly
\begin{equation}\label{eq:SC-arch-gap}
 \Delta_\sigma(x)=a(x)^\top\theta_\sigma-(a^*)^{\top}\theta_\sigma
 =\frac\phi2(1+\sigma x)+\frac{1-x^2}{4}\,.
\end{equation}
Both summands are nonnegative. Their simultaneous vanishing requires
$x=-\sigma$, proving uniqueness. In particular, for any sequence
$x_j\in(-1,1)$ with $x_j\to-\sigma$,
\begin{align}
 a(x_j)&\in G\,,
 &0<\Delta_\sigma(x_j)&\longrightarrow0\,.
 \notag
\end{align}
This proves the zero minimum positive gap for every compact generator.

\paragraph{The Geometric Sharpness Modulus}
Because $e_1,e_2/2\in\mathcal S$, one has
$\|e_1\|_{\mathcal S}\le1$ and $\|e_2\|_{\mathcal S}\le2$.
For $x\in[-1,1]$, $|x+\sigma|=1+\sigma x$, so
\begin{align}
 \|a(x)-a^*\|_{\mathcal S}
 &=\|(x+\sigma)e_1+y(x)e_2\|_{\mathcal S}\notag\\
 &\le|x+\sigma|\|e_1\|_{\mathcal S}+y(x)\|e_2\|_{\mathcal S}\notag\\
 &\le1+\sigma x+2y(x)
 =1+\sigma x+1-x^2\notag\\
 &=(1+\sigma x)(2-\sigma x)
 \le3(1+\sigma x)\,,\notag\\
 \Delta_\sigma(x)
 &=\tfrac\phi2(1+\sigma x)+\tfrac12y(x)
 \ge\tfrac\phi2(1+\sigma x)
 \ge\frac\phi6\|a(x)-a^*\|_{\mathcal S}\,.
 \notag
\end{align}
For $z=\sum_i\lambda_i a(x_i)$, with $\lambda_i\ge0$ and
$\sum_i\lambda_i=1$, the extension to the convex hull is explicit:
\begin{align}
 \Delta_\sigma(z)
 &=\sum_i\lambda_i\Delta_\sigma(x_i)
 \ge\frac\phi6\sum_i\lambda_i\|a(x_i)-a^*\|_{\mathcal S}\notag\\
 &\ge\frac\phi6\left\|\sum_i\lambda_i(a(x_i)-a^*)\right\|_{\mathcal S}
 =\frac\phi6\|z-a^*\|_{\mathcal S}\,.
 \notag
\end{align}
Every point of $\mathcal S$ has first coordinate in $[-1,1]$.
Together with $e_1\in\mathcal S$, this proves
$\|e_1\|_{\mathcal S}=1$. At the opposite endpoint,
\begin{align}
 \Delta_\sigma(\sigma)&=\phi\,,
 &\|a(\sigma)-a^*\|_{\mathcal S}&=\|2\sigma e_1\|_{\mathcal S}=2\,,\notag\\
 \frac\phi6&\le\varsigma_\sigma
 \le\frac{\Delta_\sigma(\sigma)}{\|a(\sigma)-a^*\|_{\mathcal S}}
 =\frac\phi2\,.
 \notag
\end{align}

\paragraph{Both Bounds for the Centered Information-Acquisition Complexity}
The optimal action is an endpoint on the $e_1$ axis, so the orthogonal
coordinate is the scalar $y$. For a sampling allocation $n$, let
$M=\int y(x)^2n(dx)$ and $c_{\mathrm{ctr}}^{n}=c_{\mathrm{ctr}}^{n}(\mathcal A_{\rm arch},\theta_\sigma)=\int\Delta_\sigma(x)n(dx)$.
By \Cref{lem:centered-projection-lift}, the scalar constraints
$y(x)^2/M\le\Delta_\sigma(x)^2/2$ are necessary for centered
feasibility. Since sharpness was established above, any finite
allocation satisfying these constraints can also be lifted to centered
allocations with arbitrarily close cost.
Since \eqref{eq:SC-arch-gap} gives $\Delta_\sigma(x)\ge y(x)/2$,
\begin{align}
 y(x)>0&\ \Longrightarrow\ %
       2\left(\frac{y(x)}{\Delta_\sigma(x)}\right)^2\le8\,,\notag\\
 M=8&\ \Longrightarrow\ %
       \frac{y(x)^2}{M}\le\frac{\Delta_\sigma(x)^2}{2}
                         \quad\text{for every }x\in[-1,1]\,.
 \notag
\end{align}
At an endpoint $y(x)=0$, so its scalar constraint is automatic.
Mass $32$ at $x=0$ satisfies these projected constraints:
\begin{align}
 n_0&=32\delta_0\,,
 &M(n_0)&=32(1/2)^2=8\,,\notag\\
 c_{\mathrm{ctr}}^{n_0}&=32\Delta_\sigma(0)
       =32(\phi/2+1/4)=8+16\phi\le10\,,\notag\\
 \cctr(\mathcal A_{\rm arch},\theta_\sigma)
 &\le\inf_{\varepsilon>0}(1+\varepsilon)c_{\mathrm{ctr}}^{n_0}
 =c_{\mathrm{ctr}}^{n_0}\le10\,.
 \notag
\end{align}
Conversely, any centered feasible sampling allocation must satisfy the projected constraint at $x=0$:
\begin{align}
 M&\ge2\left(\frac{y(0)}{\Delta_\sigma(0)}\right)^2
 =2\left(\frac{1/2}{1/4+\phi/2}\right)^2
 \ge2\left(\frac{1/2}{5/16}\right)^2
 =\frac{128}{25}\,,\notag\\
 y(x)^2&\le2y(x)\Delta_\sigma(x)\le\Delta_\sigma(x)\,,\notag\\
 c_{\mathrm{ctr}}^{n}&=\int\Delta_\sigma(x)n(dx)
 \ge\int y(x)^2n(dx)=M\ge\frac{128}{25}\,,\notag\\
 \cctr(\mathcal A_{\rm arch},\theta_\sigma)
 &=\inf_{n\text{ centered feasible}}c_{\mathrm{ctr}}^{n}\ge\frac{128}{25}\,.
 \notag
\end{align}
The first line uses $\phi\le1/8$, and the second uses
$0\le2y(x)=1-x^2\le1$. The centered lift proves the upper bound,
and the projected necessary constraints prove the lower bound in
\eqref{eq:SC-arch-geometry}.
\end{proof}

\paragraph{Uniform Implications and the Logarithm That Is Not Matched}
On this fixed action geometry $\cctr$ stays in a fixed positive interval
while $\varsigma_\sigma\asymp\phi\to0$. A proposed bound
$A\cctr\log(eT)+M(\theta_\sigma)+B\varsigma_\sigma^{-\rho}\sqrt C$
with fixed $\phi$-independent $A,B$ and arbitrary finite $T,C$-independent
transients cannot hold uniformly when $\rho<1/2$. Indeed fix a sufficiently
small $\phi$, maximize the two finite transients, use
$C_{\phi,T}\le\phi T/2$, divide by $\sqrt T$, and let $T\to\infty$.
The right-hand limiting coefficient is at most
$B6^\rho\phi^{1/2-\rho}/\sqrt2$ when $\rho\ge0$, and also tends to zero
when $\rho<0$ by the other margin inequality. The left side is $1/32$.
For a fixed $\phi$, the same proof rules out any corruption power
strictly less than $1/2$, even with a fixed power of $\log(e+C)$.

A horizon-growing coefficient requires a different quantifier argument.
For example, no uniform
$A\cctr\log(eT)+M(\theta)+B\sqrt{C(1+\cctr\log(eT))}$ bound with
$M(\theta_\sigma)\le K\phi^{-p_0}$ for fixed finite $K,p_0$ is possible.
Take $\phi=1/N$, $T=N^m$ with $m>\max\{2p_0,2\}$, and divide by
$N^{m/2}$. The logarithmic and transient terms vanish; the square-root
coefficient is $\gO(\sqrt{\log N/N})$ and also vanishes, contradicting
$1/32$. The same calculation excludes a margin exponent $\rho<1/2$
even after multiplication by any fixed power of $\log(e+C)$, provided
the transient is polynomial in $\phi^{-1}$.
These are necessity statements with specified uniformity. They do not
identify the large priority- and geometry-dependent constants of
\eqref{eq:SC-constants} as optimal. In particular no lower bound here
makes its additional $\log(e+C)$ necessary; removing that logarithm
from the general sharp upper bound remains an unproved improvement.

\clearpage
\section{Proof of Theorem~\ref*{thm:main-critical}}
\label{ch:critical}
We first prove \Cref{thm:main-critical} on the fixed critical curves.
\Cref{lem:critical-design}, proved afterwards, certifies nonsharpness and
infinite centered information-acquisition complexity and the nonpolyhedral convex hull.

For an even integer $p\ge4$, let
\begin{equation}\label{eq:critical-family}
 \cA_p=\{a(s)=(1/2+s,s^p):|s|\le1/4\}\,,\qquad\theta_0=(0,1)\,.
\end{equation}
The action set is compact, spans $\R^2$, and is inside the unit ball, since
$\norm{a(s)}^2\le(3/4)^2+4^{-2p}<1$. Its unique optimal action under $\theta_0$ is $a_0=(1/2,0)$, and its gap is $s^p$.

\begin{proof}[Proof of \Cref{thm:main-critical}]
\textbf{The common fixed loss vector and the exact corruption budget.}
Fix $C\ge0$ and an integer $T$ in the stated range. Set
\begin{align}
 \varepsilon&=\min\left\{\frac{4C}{3T},\frac1{8\sqrt T}\right\}\,,
 &\theta_t^\sigma&=(\sigma\varepsilon,1)\,,
 &c_t^{\rm vec}&=(\sigma\varepsilon,0)\quad(t\le T)\,.
 \notag
\end{align}
Both environments have the same fixed loss vector $\theta_0=(0,1)$. The action-projection
norm and its sum are
\begin{align}
 \|c_t^{\rm vec}\|_{\mathcal A_p}
 &=\sup_{|s|\le1/4}|a(s)^\top c_t^{\rm vec}|
 =\sup_{|s|\le1/4}|\sigma\varepsilon(1/2+s)|
 =\frac{3\varepsilon}{4}\,,\notag\\
 \sum_{t=1}^T\|c_t^{\rm vec}\|_{\mathcal A_p}
 &=\frac{3T\varepsilon}{4}
 =\frac{3T}{4}\min\left\{\frac{4C}{3T},\frac1{8\sqrt T}\right\}
 =\min\{C,3\sqrt T/32\}=\bar C\le C\,.
 \notag
\end{align}
The fixed loss vector, horizon, budget cap, and used budget are identical for
the two signs, even when they are revealed to the learner.

\paragraph{Bounded Sign Feedback and the Full-History Testing Distance}
At $a(s)$, use sign feedback with linear mean
$\mu_\sigma(s)=a(s)^\top\theta_t^\sigma=s^p+\sigma\varepsilon(1/2+s)$.
For every $|s|\le1/4$,
\begin{align}
 |\mu_\sigma(s)|
 &\le|s|^p+\varepsilon(1/2+s)
 \le4^{-p}+\frac{3\varepsilon}{4}
 \le\frac1{256}+\frac3{32}<\frac12\,,\notag\\
 \mu_+(s)-\mu_-(s)&=2\varepsilon(1/2+s)\,,\notag\\
 \KL(Q_{\mu_+(s)}, Q_{\mu_-(s)})
 &\le\frac{[2\varepsilon(1/2+s)]^2}{1-\mu_-(s)^2}
 \le\frac{4\varepsilon^2(3/4)^2}{3/4}
 =3\varepsilon^2\,.
 \notag
\end{align}
The first line uses $p\ge4$ and $T\ge1$; the last line is the sign
KL inequality in \Cref{lem:testing-tools}.
Let $P_\sigma^T$ denote the full history law, including the learner's
randomization. The common policy kernels give
\begin{align}
 \KL(P_+^T, P_-^T)
 &=\E_+\left[\sum_{t=1}^T
       \KL(Q_{\mu_+(s_t)}, Q_{\mu_-(s_t)})\right]
 \le3T\varepsilon^2
 \le3T\frac1{64T}=\frac3{64}\,,\notag\\
 \TV(P_+^T,P_-^T)
 &\le\sqrt{\tfrac12\KL(P_+^T, P_-^T)}
 \le\sqrt{\frac3{128}}<\frac12\,.
 \notag
\end{align}
Providing the common information $\theta_0,T,C,\bar C$ changes neither
the equality of the policy kernels nor these inequalities.

\paragraph{The Optimal Action under Corruption and the Cost of a Wrong Sign}
The action-independent term $\sigma\varepsilon/2$ cancels from every
gap. Let $f_\sigma(s)=s^p+\sigma\varepsilon s$ and
$u=(\varepsilon/p)^{1/(p-1)}$. The horizon condition is used as follows:
\begin{align}
 T&\ge[4^{p-1}/(8p)]^2
 \quad\Longrightarrow\quad
 \frac1{8p\sqrt T}\le4^{-(p-1)}\,,\notag\\
 u&=\left(\frac\varepsilon p\right)^{1/(p-1)}
 \le\left(\frac1{8p\sqrt T}\right)^{1/(p-1)}\le\frac14\,,\notag\\
 f_\sigma'(s)&=p s^{p-1}+\sigma\varepsilon\,,
 &f_\sigma'(-\sigma u)&=-\sigma p u^{p-1}+\sigma\varepsilon=0\,,\notag\\
 f_\sigma(-\sigma u)
 &=u^p-\varepsilon u
 =u^p-pu^{p-1}u=-(p-1)u^p\,,\notag\\
 g&=(p-1)u^p\,,
 &f_\sigma(s)-f_\sigma(-\sigma u)&=s^p+\sigma\varepsilon s+g\,.
 \notag
\end{align}
Since $p$ is even, $s\mapsto s^p$ is strictly convex; hence
$-\sigma u$ is the unique minimum on the allowed interval, including
$u=0$. For $\sigma s\ge0$, both $s^p$ and
$\sigma\varepsilon s$ are nonnegative, and the last gap is at least $g$.
With $E_t=\{s_t\ge0\}$, the testing and regret calculations are
\begin{align}
 P_+^T(E_t)+P_-^T(E_t^c)
 &=1+P_+^T(E_t)-P_-^T(E_t)
 \ge1-\TV(P_+^T,P_-^T)\ge\tfrac12\,,\notag\\
 \Reg[T,+]+\Reg[T,-]
 &=\sum_{t=1}^T\left\{
      \E_+[s_t^p+\varepsilon s_t+g]
      +\E_-[s_t^p-\varepsilon s_t+g]\right\}\notag\\
 &\ge g\sum_{t=1}^T[P_+^T(E_t)+P_-^T(E_t^c)]
 \ge\frac{Tg}{2}\,,\notag\\
 \max_\sigma \Reg[T,\sigma]
 &\ge\frac{\Reg[T,+]+\Reg[T,-]}2
 \ge\frac{Tg}{4}
 =\frac{(p-1)T}{4}u^p\notag\\
 &=\frac{(p-1)T}{4}\left(\frac\varepsilon p\right)^{p/(p-1)}\notag\\
 &=\frac{(p-1)T}{4p^{p/(p-1)}}
   \left(\min\left\{\frac{4C}{3T},\frac1{8\sqrt T}\right\}\right)^{p/(p-1)}\,.
 \notag
\end{align}
This is \eqref{eq:critical-full}. When $C=0$, $\varepsilon=u=g=0$,
the two models coincide, and the displayed lower bound is zero.
The always-$a_0$ policy has the exact regret
\begin{align}
 \Reg[T,\sigma](\text{always }a_0)
 &=\sum_{t=1}^T[f_\sigma(0)-f_\sigma(-\sigma u)]
 =\sum_{t=1}^Tg=Tg\,,
 \notag
\end{align}
which is four times the displayed lower bound for every $C,T$.

\paragraph{The Quartic Specialization}
For $p=4$, the horizon threshold is $[4^3/(8\cdot4)]^2=4$. Substituting $p=4$ into the lower bound gives
\begin{align}
 \frac{Tg}{4}
 &=\frac{3T}{4}\left(\frac\varepsilon4\right)^{4/3}
 =\frac{3T}{4}\left(\min\left\{\frac C{3T},
                                  \frac1{32\sqrt T}\right\}\right)^{4/3}\notag\\
 &=\min\left\{
   \frac{3T}{4}\frac{C^{4/3}}{3^{4/3}T^{4/3}},
   \frac{3T}{4}\frac1{32^{4/3}T^{2/3}}\right\}\notag\\
 &=\min\left\{\frac{C^{4/3}}{4\,3^{1/3}T^{1/3}},
                  \frac{3T^{1/3}}{4\,32^{4/3}}\right\}\,.
 \notag
\end{align}
At $C=C_T=3\sqrt T/32$, the two terms are equal:
\begin{align}
 \frac{C_T^{4/3}}{4\,3^{1/3}T^{1/3}}
 &=\frac{(3\sqrt T/32)^{4/3}}{4\,3^{1/3}T^{1/3}}
 =\frac{3T^{1/3}}{4\,32^{4/3}}\notag\\
 &=\frac{3^{1/3}}{4\,32^{2/3}}C_T^{2/3}\,.
 \label{eq:CRr-quartic-critical-budget}
\end{align}
These identities prove $\max_\sigma\Reg[T,\sigma]\ge\min\{C^{4/3}/(4\,3^{1/3}T^{1/3}),3T^{1/3}/(4\,32^{4/3})\}$ and its stated
$\Omega(C^{2/3})$ consequence.

\paragraph{Impossibility with Fixed Instance-Dependent Coefficients}
Suppose an upper bound of the form
$A_{\cA,\theta}\log(eT)+M_{\cA,\theta}
 +J_{\cA,\theta}\sqrt C$ held on every instance in the critical family
for all $T,C$. On the single fixed quartic instance, denote
its finite coefficients by $A_0,M_0,J_0$.
Applying the proposed bound to both signs at $C=C_T$ and substituting
\eqref{eq:CRr-quartic-critical-budget} would imply
\begin{align}
 \frac{3T^{1/3}}{4\,32^{4/3}}
 &\le A_0\log(eT)+M_0+J_0\sqrt{3\sqrt T/32}\,,\notag\\
 \frac3{4\,32^{4/3}}
 &\le\frac{A_0\log(eT)+M_0}{T^{1/3}}
       +J_0\sqrt{\frac3{32}}\,T^{1/4-1/3}\notag\\
 &=\frac{A_0\log(eT)+M_0}{T^{1/3}}
       +J_0\sqrt{\frac3{32}}\,T^{-1/12}
 \longrightarrow0\,,
 \notag
\end{align}
contradicting the positive constant on the left.

For general even $p\ge4$, let
$\beta_p=(p-2)/(p-1)$ and
$k_p=(p-1)/(4p^{p/(p-1)}8^{p/(p-1)})>0$.
Since $4C_T/(3T)=1/(8\sqrt T)$, direct substitution gives
\begin{align}
 \max_\sigma \Reg[T,\sigma]
 &\ge\frac{(p-1)T}{4p^{p/(p-1)}}
                          (8\sqrt T)^{-p/(p-1)}\notag\\
 &=k_pT^{1-p/(2(p-1))}
 =k_pT^{\beta_p/2}
 =k_p(32/3)^{\beta_p}C_T^{\beta_p}\,.
 \label{eq:CRr-general-critical-budget}
\end{align}
For any prescribed $\gamma<1$, choose an even $p\ge4$ with
$p>1+1/(1-\gamma)$, so $\beta_p>\gamma$, and then keep that instance
fixed. Let $A_p,M_p,J_p$ be arbitrary finite coefficients of a
proposed upper bound with corruption term
$J_pC^\gamma[\log(eT)]^q$, where $q\ge0$ is fixed.
Combining it with \eqref{eq:CRr-general-critical-budget} would give
\begin{align}
 k_p(32/3)^{\beta_p}
 &\le\frac{A_p\log(eT)+M_p}{C_T^{\beta_p}}
       +J_pC_T^{\gamma-\beta_p}[\log(eT)]^q\notag\\
 &=(32/3)^{\beta_p}
       \frac{A_p\log(eT)+M_p}{T^{\beta_p/2}}
       +J_p(3/32)^{\gamma-\beta_p}
           T^{-(\beta_p-\gamma)/2}[\log(eT)]^q
 \longrightarrow0\,.
 \notag
\end{align}
Both exponents $\beta_p/2$ and $(\beta_p-\gamma)/2$ are positive.
The positive left side contradicts the vanishing right side, proving
the stated exclusion with the instance fixed as $T\to\infty$.
\end{proof}

\begin{lemma}[Nonsharpness and infinite centered complexity]\label{lem:critical-design}
For \eqref{eq:critical-family}, the instance is nonsharp and
$\cctr(\cA_p,\theta_0)=\infty$: no finite centered feasible sampling
allocation exists. Its convex hull has infinitely many exposed points.
\end{lemma}
\begin{proof}
\textbf{Nonsharpness and the impossibility of a finite centered sampling allocation.}
For nonzero $s\to0$,
\begin{align}
 \frac{\Delta(a(s))}{\|a(s)-a_0\|}
 &=\frac{s^p}{\sqrt{s^2+s^{2p}}}
 \le\frac{|s|^p}{|s|}=|s|^{p-1}\longrightarrow0\,.
 \notag
\end{align}
This proves nonsharpness. More directly, suppose a finite centered
sampling allocation had positive-definite covariance $H_n$. The centered constraint
and the least eigenvalue of $H_n^{-1}$ would give, for every $s\ne0$,
\begin{align}
 \frac{s^2}{\lambda_{\max}(H_n)}
 &\le\frac{s^2+s^{2p}}{\lambda_{\max}(H_n)}
 \le(a(s)-a_0)^\top H_n^{-1}(a(s)-a_0)
 \le\frac{s^{2p}}2\,,\notag\\
 \lambda_{\max}(H_n)&\ge2|s|^{2-2p}\longrightarrow\infty\,.
 \notag
\end{align}
A finite measure on this bounded action set has finite
$\lambda_{\max}(H_n)$, contradicting the last line.
Hence no finite centered sampling allocation exists.

\paragraph{Infinitely Many Exposed Actions}
For an interior parameter $s_0\in(-1/4,1/4)$, take the linear
functional with vector $v_{s_0}=(-p s_0^{p-1},1)$.
Its restriction to the curve satisfies
\begin{align}
 v_{s_0}^\top a(s)
 &=s^p-p s_0^{p-1}s-\tfrac p2s_0^{p-1}\,,\notag\\
 v_{s_0}^\top a(s)-v_{s_0}^\top a(s_0)
 &=s^p-s_0^p-p s_0^{p-1}(s-s_0)>0
                           \qquad(s\ne s_0)\,.
 \notag
\end{align}
The strict inequality is the strict supporting-line inequality for
the differentiable strictly convex function $s\mapsto s^p$.
Thus $a(s_0)$ is the unique minimizer of a linear functional over
the curve and its convex hull, so it is exposed. Infinitely many
choices of $s_0$ give infinitely many exposed points.
\end{proof}

The role of the comparator can be computed exactly.
For the always-$a_0$ policy, $\Reg(a_0)=0$, whereas $\Reg=Tg$.
Thus this construction does not disprove a bound stated only against the
optimal action $a_0$ under $\theta_0$. It rules out the proposed bound for the actual fixed
comparator, the regret convention used throughout this paper.
For the particular perturbation direction, the identity in \Cref{lem:comparator} gives the exact shift
\[
 \sup_s[-s^p-\sigma\varepsilon s]=(p-1)(\varepsilon/p)^{p/(p-1)}=g\,.
\]
Consequently the additional scale is $T\varepsilon^{p/(p-1)}$.
The exact used budget satisfies $\varepsilon=4\bar C/(3T)$, where
$\bar C=\min\{C,3\sqrt T/32\}$. The cap on $\varepsilon$ keeps the two
experiments statistically close for every budget cap $C$, and yields
\[
 Tg\asymp_p\min\left\{
       \frac{C^{p/(p-1)}}{T^{1/(p-1)}},
       T^{(p-2)/(2(p-1))}\right\}\,.
\]
Choosing $C\asymp\sqrt T$ makes this
$T^{(p-2)/(2(p-1))}\asymp C^{(p-2)/(p-1)}$.
This additional loss is caused by movement of the best fixed comparator
under corruption, while regret against the uncorrupted optimal action $a_0$ remains zero
for the always-$a_0$ policy.

\clearpage
\section{Common Tools}\label{ch:tools}
This appendix collects the identities and geometry shared by the finite and sharp proofs. The auxiliary separated-generator construction supplies the moment estimates for affine functions used in both. Its guarantee is proved before the individual estimates on which it depends.

\subsection{FTRL, G-Optimality, and Comparator Identities}\label{app:common-identities}
All action sets are compact subsets of the Euclidean unit ball and span the stated linear dimension; a smaller span is represented in its own coordinates. Policies are measurable kernels based on previous observations and their own randomness. When a reconstruction mark is sampled together with the action, the feedback identity
$\E[Y_t\mid\cF_{t-1},A_t]=A_t^\top\theta_t$
is also required after conditioning on that mark.
The mark is internal randomness, not an additional observation. Enlarging the pre-action filtration to include an environment's fresh pre-action randomization does not reveal it to the learner. All guarantees concern expectation against a fixed comparator, with any final supremum outside expectation.

For a compact spanning set define the action norm
$\norm z_\cA=\sup_{a\in\cA}|a^\top z|$.
This is a norm because the set spans. Its dual unit ball is $\Conv(\cA\cup(-\cA))$. The squared norm $\norm x_M^2=x^\top Mx$ is used only for a specified positive semidefinite $M$.

\begin{lemma}[Exact comparator correction]\label{lem:comparator}
Let $\bar c_T=T^{-1}\sum_{t\le T}\E[c_t^{\rm vec}]$. Then
\begin{equation}\label{eq:comparator-exact}
 \Reg-\Reg(a^*)
 =T\sup_{a\in\cA}\{-\Delta(a)-(a-a^*)^\top\bar c_T\}\,.
\end{equation}
Define
\begin{equation}\label{eq:comparator-modulus}
 \Omega_\theta(\varepsilon)=
 \sup_{\norm z_\cA\le\varepsilon}\sup_{a\in\cA}
 \{-\Delta(a)-(a-a^*)^\top z\}\,.
\end{equation}
The correction is nonnegative and at most $T\Omega_\theta(C/T)\le2C$.
If $\Delta(a)\ge\Delta_{\min}>0$ for every $a\ne a^*$, it is zero when $T\Delta_{\min}\ge2C$.
\end{lemma}
\begin{proof}
For every fixed $a$,
\[
 \Reg(a)-\Reg(a^*)
 =\sum_{t\le T}\E[(a^*-a)^\top(\theta+c_t^{\rm vec})]
 =T[-\Delta(a)-(a-a^*)^\top\bar c_T]\,.
\]
Taking the supremum proves the identity. The maximum includes $a^*$, giving zero as a candidate. Convexity of the norm and the path budget give
$\norm{\bar c_T}_\cA\le T^{-1}\sum_t\E[\kappa_t]\le C/T$.
Also $|(a-a^*)^\top z|\le2\norm z_\cA$, proving the modulus bound. Under separation, every suboptimal candidate is at most $-\Delta_{\min}+2C/T\le0$.
\end{proof}

\paragraph{Why the Comparator Identity Is Needed}
The correction in \eqref{eq:comparator-exact} is independent of the learner's actions: it measures the change of the best fixed comparator under the average corruption. The modulus bounds that change uniformly over perturbations of a given size, while separation makes it vanish for $T\Delta_{\min}\ge2C$. This is the bridge from the bound for $\Reg(a^*)$ to the best-fixed-comparator bounds in \Cref{prop:SC-per-design}. \Cref{thm:main-critical} exhibits a setting where this correction is substantial even when the learner knows and always plays the optimal action $a^*$ under $\theta$.

\subsubsection{G-Optimal Designs and FTRL Identities}

G-optimality supplies a fallback exploration design that is feasible on every history, without knowing the gaps. It guarantees that the algorithms are well defined and supports their adversarial bounds. Each fixed gap-adapted sampling allocation $n$ is later used to construct a feasible comparison distribution for the same optimization program. The sampling allocation and the comparison distribution are used only in the analysis.

The coarse and local FTRL inequalities below serve different regimes of the same learner. The coarse bound pays the worst-case entropy at the terminal regularization scale and yields adversarial regret. The local bound charges only the entropy remaining as regularization grows. In the stochastic analysis, sufficiently fast concentration gives a finite upper bound on $\sum_{t=2}^T(\beta_t-\beta_{t-1})\E[H(q_t)]$ that is independent of $T$. For a nonuniform compact reference, the analogous local expression contains the comparator-dependent $h_u$, which is why the compact proof first treats the atom at $a^*$ instead of all comparators simultaneously.

\begin{lemma}[Full-history entropy identities]\label{lem:entropy-FTRL}
For nondecreasing $\beta_t$, let
$q_{t,a}\propto\exp(-S_{t-1,a}/\beta_t)$, $S_t=S_{t-1}+g_t$, and
\[
 D_t=\ip{q_t}{g_t}+\beta_t\log\sum_aq_{t,a}e^{-g_{t,a}/\beta_t}\,.
\]
For every fixed coordinate $u$,
\begin{align}
 \sum_{t\le T}\ip{q_t-e_u}{g_t}&\le\beta_T\log K+\sum_tD_t\,,\label{eq:entropy-coarse}\\
 \sum_{t\le T}\ip{q_t-e_u}{g_t}&\le\beta_1\log K+
 \sum_{t=2}^T(\beta_t-\beta_{t-1})H(q_t)+\sum_tD_t\,,\label{eq:entropy-local}
\end{align}
where $H(q)=-\sum_aq_a\log q_a$.
For a probability reference $\mu$, relative-entropy FTRL $q_t(da)\propto e^{-S_{t-1}(a)/\beta_t}\mu(da)$, and an atom $u$ of mass $\pi_u>0$, the corresponding bounds are
\begin{align}
 \sum_t\left(\int g_t\,dq_t-g_t(u)\right)
 &\le\beta_T\log(1/\pi_u)+\sum_tD_t\,, \notag\\
 &\le\beta_1h_u+\sum_{t=2}^T(\beta_t-\beta_{t-1})
 [h_u-\KL(q_t,\mu)]+\sum_tD_t\,,\label{eq:relative-local}
\end{align}
with $h_u=\log(1/\pi_u)$ and the integral definition of $D_t$.
\end{lemma}
\begin{proof}[Proof of \Cref{lem:entropy-FTRL}]
For the relative case define
$F_\beta(S)=-\beta\log\int e^{-S/\beta}d\mu
=\inf_q\{\int S\,dq+\beta\KL(q,\mu)\}$.
The variational formula follows by subtracting $\beta\KL(q, q_\beta)$ after substituting the exponential-weights density. Then
\[
 \int g_t\,dq_t=D_t+F_{\beta_t}(S_t)-F_{\beta_t}(S_{t-1})\,.
\]
Telescoping yields the terminal potential, minus $F_{\beta_1}(0)=0$, and terms
$F_{\beta_{t-1}}(S_{t-1})-F_{\beta_t}(S_{t-1})$.
These are nonpositive because KL is nonnegative. More precisely, testing $q_t$ in the first minimum bounds each by
$-(\beta_t-\beta_{t-1})\KL(q_t,\mu)$.
The terminal minimum is at most $S_T(u)+\beta_T h_u$. This proves both relative formulas.
For uniform $\mu$, $\KL(q,\mu)=\log K-H(q)$ and $h_u=\log K$, giving both finite formulas. No part of the sequence is restarted, and no estimate magnitude condition is used in this identity.
\end{proof}

\begin{lemma}[G-optimal design on a compact spanning set]\label{lem:gopt}
There is a finite-support probability measure $\pi$ such that
$\Sigma(\pi)=\int aa^\top\pi(da)\succ0$ and
$\sup_{a\in\cA}a^\top\Sigma(\pi)^{-1}a\le d$.
\end{lemma}
\begin{proof}[Proof of \Cref{lem:gopt}]
The convex hull of $\{aa^\top:a\in\cA\}$ is compact in the finite-dimensional symmetric-matrix space. To see this without a closure assumption, any convex combination can be reduced to at most $d(d+1)/2+1$ terms: an affine dependence among more points permits moving weights until a weight vanishes, and repetition proves the support bound. The bounded-support parametrization is compact. Maximize the determinant on this convex hull. A spanning collection gives a positive candidate, hence the maximizer $M$ is positive definite. The one-sided derivative toward $aa^\top$ of $\log\det M$ is
$\Tr[M^{-1}(aa^\top-M)]=a^\top M^{-1}a-d\le0$.
The same bounded-support representation gives the stated measure.
\end{proof}
\subsubsection{Exponential and Probability Tools}

The centered bound for the proxy function in \Cref{lem:chi} controls FTRL's exponential stability using the estimator constraints imposed by the algorithm. Centering matters because only action-loss differences are reconstructed. The power expansions in \Cref{lem:taylor} are used for two different tasks: absolute moments make initialization uniform, and negative-part moments control how long estimated cumulative loss differences can have the wrong sign.
\begin{lemma}[Centered exponential bound for the proxy function]\label{lem:chi}
Let $\chi(x)=e^x+e^{-x}-2$. It is even, convex, nonnegative, and increasing in $|x|$. For every real $x$,
\[
 e^x-x-1\le\chi(x)\,,\quad\chi(x)\le x^2e^{|x|}\,,\quad
 \chi(x)\le\tfrac32x^2\quad(|x|\le1)\,.
\]
For every probability measure $q$ and bounded measurable $f$,
\begin{equation}\label{eq:chi-center}
 \int\chi(f-\textstyle\int f\,dq)\,dq\le\int\chi(2f)\,dq\,.
\end{equation}
\end{lemma}
\begin{proof}
Evenness and monotonicity follow from $\chi(x)=4\sinh^2(x/2)$; convexity follows from $\chi''(x)=e^x+e^{-x}>0$. The first inequality subtracts to $e^{-x}+x-1\ge0$. Taylor's integral remainder yields
\[
 \chi(x)=x^2\int_0^1(1-s)(e^{sx}+e^{-sx})\,ds\le x^2e^{|x|}\,.
\]
For $|x|\le1$, the power series gives
$\chi(x)\le x^2(e+e^{-1}-2)<3x^2/2$.
For independent $a,b\sim q$, Jensen and then convexity give
\[
 \E[\chi(f(a)-\E[f(b)])]\le\E[\chi(f(a)-f(b))]
 \le\tfrac12\E[\chi(2f(a))+\chi(-2f(b))]\,,
\]
which proves \eqref{eq:chi-center}.
\end{proof}

\begin{lemma}[Two elementary power expansions]\label{lem:taylor}
For an even integer $P\ge2$ and real $y,z$,
\begin{align}
 (y+z)_+^P&\le y_+^P+Py_+^{P-1}z+\tfrac12P(P-1)(y+|z|)_+^{P-2}z^2\,,\label{eq:taylor-positivepart}\\
 |y+z|^P&\le|y|^P+P|y|^{P-2}yz+\tfrac12P(P-1)(|y|+|z|)^{P-2}z^2\,.\label{eq:taylor-absolute}
\end{align}
If $v_i,c\ge0$ and $k=K^{1/P}$, then
\begin{equation*}
 \sum_i(v_i+c)^{P-2}\le k^2\left[(\sum_i v_i^P)^{1/P}+kc\right]^{P-2}\,. \notag
\end{equation*}
\end{lemma}
\begin{proof}
Bound the second derivative along $y+sz$, $0\le s\le1$, and integrate with weight $1-s$, whose integral is $1/2$. If the first segment stays nonpositive, its positive-part remainder is zero. For the last display, finite-dimensional H\"older bounds the $P-2$ norm by $K^{2/[P(P-2)]}$ times the $P$ norm; the $P$-norm triangle inequality then adds $K^{1/P}c$.
\end{proof}

The next lemma explains the statistical mechanism behind the lower bounds. Small observation KL keeps the full histories close in total variation, so a policy cannot reliably select the correct action in both environments. Multiplying the unavoidable testing errors by the corresponding regret gap then gives a regret lower bound. The calculations in later sections differ in how their geometry converts a wrong decision into regret.

\begin{lemma}[Pinsker, Bernoulli KL, and testing]\label{lem:testing-tools}
For probability laws $P,Q$,
\[
 \TV(P,Q)\le\sqrt{\KL(P, Q)/2}\,.
\]
For sign laws with means $\mu,\nu\in(-1,1)$,
\begin{equation}\label{eq:sign-KL}
 \KL(P_\mu, P_\nu)\le\frac{(\mu-\nu)^2}{1-\nu^2}\,.
\end{equation}
Every test event $E$ satisfies $P(E)+Q(E^c)\ge1-\TV(P,Q)$.
\end{lemma}
\begin{proof}
Grouping a density ratio on $E,E^c$ and applying Jensen gives
$\KL(P, Q)\ge\kl(P(E),Q(E))$.
For fixed $q$, the second derivative in $p$ of $\kl(p,q)$ is $1/[p(1-p)]\ge4$, and its value and first derivative at $p=q$ are zero. Hence $\kl(p,q)\ge2(p-q)^2$. Take the supremum over events. For \eqref{eq:sign-KL}, $\log x\le x-1$ yields $\KL\le\chi^2$; direct substitution of probabilities $(1\pm\mu)/2$ and $(1\pm\nu)/2$ gives the displayed expression. The test inequality is $1+P(E)-Q(E)\ge1-\TV(P,Q)$.
\end{proof}

On the scalar action set $\{-1,1\}$, multiplying feedback by the chosen sign reveals a bounded unbiased estimate of the full scalar loss. Conditional Hoeffding concentration therefore applies directly, which is why the scalar proof can attain an exact half inverse-gap power without the general reconstruction and high-moment machinery.

\begin{lemma}[Conditional Hoeffding bound]\label{lem:hoeffding}
If a random variable has support in an interval of length at most two, then
$\log\E[e^{\lambda(Z-\E[Z])}]\le\lambda^2/2$ for every real $\lambda$. The same holds conditionally. Consequently, martingale differences obtained by centering such variables satisfy
\[
 \sP\left(\sum_{t=1}^n(Z_t-\E_{t-1}[Z_t])\le-x\right)\le e^{-x^2/(2n)}\,.
\]
\end{lemma}
\begin{proof}
For $f(\lambda)=\log\E[e^{\lambda(Z-\E[Z])}]$, $f(0)=f'(0)=0$ and $f''(\lambda)$ is a variance under an exponentially tilted law with the same support. Its variance is at most one by measuring squared distance from the interval midpoint. Integrate twice. Iterate the conditional inequality, use Markov's inequality, and minimize $-\lambda x+n\lambda^2/2$ at $\lambda=x/n$.
\end{proof}

Upper-bound failure probabilities are charged at a per-round rate proportional to $t^{-1/2}$. The next lemma converts a decaying probability tail into its total regret cost, with the necessary cap at one. Its root $A^{1/(2p)}$ is where moment-tail exponents become inverse-gap exponents in the final corruption coefficient; its second assertion controls the direct exploration cost of predictable corruption.

\begin{lemma}[Capped weighted tails]\label{lem:capped}
For $A\ge0,n>0,p>1/2$,
\begin{equation}\label{eq:capped}
 \sum_{t\ge1}t^{-1/2}\min\{1,A(n/t)^p\}
 \le\left(2+\frac1{p-1/2}\right)A^{1/(2p)}\sqrt n\,.
\end{equation}
Also, if $0\le\kappa_t\le2$ and $\sum_t\kappa_t\le C$, then
\begin{equation}\label{eq:cor-weighted}
 \sum_t\frac{\kappa_t}{\sqrt t}\le2\sqrt{2C}\,.
\end{equation}
\end{lemma}
\begin{proof}
For $A>0$ the summand in \eqref{eq:capped} is decreasing, so its sum is bounded by the integral from zero. Split at $x_0=nA^{1/p}$. The integrals are $2\sqrt{x_0}$ and $An^p x_0^{1/2-p}/(p-1/2)$. For $A=0$ both sides vanish. For \eqref{eq:cor-weighted}, set $f(x)=\kappa_t$ on $[t-1,t)$. Then $\sum\kappa_t/\sqrt t\le\int f(x)x^{-1/2}dx$. Among $0\le f\le2$ with $\int f\le C$, moving mass left increases the integral; its maximum is $2\int_0^{C/2}x^{-1/2}dx=2\sqrt{2C}$.
\end{proof}

\subsection{Information Geometry and Sampling Allocations}\label{ch:compact-geometry}
\subsubsection{Centered Complexity, Sharpness, and Projection Tools}
For the geometric statements in this subsection, $\cA\subseteq\BB$ is
compact, spans $\R^d$, and has at least two points. At a fixed parameter
$\theta$, the minimizer $a^*$ is unique. Define
$\Delta(a)=(a-a^*)^\top\theta$. All sampling allocations below are finite
nonnegative Borel measures. In this subsection, let
$c_{\mathrm{ctr}}^{n}=c_{\mathrm{ctr}}^{n}(\cA,\theta)$,
$c^{n}=c^{n}(\cA,\theta)$, and
$\cctr=\cctr(\cA,\theta)$; recall \Cref{def:D-physical-design}. The norm triangle inequality and affineness
of $\Delta$ make its constraints equivalent to testing only physical
actions $a\in\cA$.

Separation on a compact generator implies sharpness on its convex hull:
if $\Delta(b)\ge\Delta_{\min}>0$ for $b\ne a^*$, then
$\Delta(b)\ge(\Delta_{\min}/2)\|b-a^*\|$, and convex combinations preserve
this inequality. The converse need not hold, as \Cref{prop:C-eight}
shows. The critical curves in \Cref{lem:critical-design} are nonsharp
and admit no finite centered feasible allocation.

\begin{proposition}[Centered sampling allocations]\label{prop:C-centered}
A finite centered feasible sampling allocation exists if and only if
there is a constant $\kappa>0$ such that
\begin{equation}\label{eq:C-sharp}
 \Delta(a)\ge\kappa\|a-a^*\|\qquad(a\in\cA)\,.
\end{equation}
Equivalently, $\cctr<\infty$ holds exactly under sharpness.
On a finite action set with a unique optimal action $a^*$, $\cctr=\cstar$.
\end{proposition}
\begin{proof}[Proof of \Cref{prop:C-centered}]
Centered feasibility gives
$\|a-a^*\|^2\le\lambda_{\max}(H_n)\Delta(a)^2/2$, hence sharpness.
Conversely, suppose \eqref{eq:C-sharp} holds. Spanning supplies a finite
allocation with $H_0\succ0$. Multiply it by
$\alpha=\max\{1,2\lambda_{\max}(H_0^{-1})/\kappa^2\}$ to obtain
\[
 (a-a^*)^\top(\alpha H_0)^{-1}(a-a^*)
 \le\frac{\lambda_{\max}(H_0^{-1})}{\alpha}\|a-a^*\|^2
 \le\frac{\Delta(a)^2}{2}\,.
\]
Its mass and cost are finite. Since every finite allocation has finite
cost on this bounded domain, feasibility is equivalent to $\cctr<\infty$.

For the finite-action identity, a positive minimum gap gives sharpness.
If $a^*=0$, the raw and centered constraints coincide. If $a^*\ne0$
and $d\ge2$, \eqref{eq:rev-A-lift-cost-mass-step-1} in
\Cref{lem:F-design-mass} identifies $\cstar$ with the finite quotient
value $c_Q$ in \eqref{eq:rev-A-quotient}. The projection and lifting
statements of \Cref{lem:centered-projection-lift} below give
$c_Q\le\cctr\le(1+\varepsilon)c_Q$ for every $\varepsilon>0$.
Letting $\varepsilon\downarrow0$ proves equality. If $d=1$ and
$a^*\ne0$, sufficient mass at $a^*$ alone is feasible for either
program and has zero cost, so both values are zero.
\end{proof}

\begin{lemma}[Projection and lifting for centered allocations]
\label{lem:centered-projection-lift}
Assume \eqref{eq:C-sharp}. If $a^*\ne0$, let $U^*$ have orthonormal
columns spanning $(a^*)^{\perp}$; otherwise set $U^*=I_d$.
Let $x(a)=(U^*)^{\top} a$, $r^*=\operatorname{rank}(U^*)$, and
$M(w)=\int x(a)x(a)^\top w(da)$. The notation
$c_{\mathrm{ctr}}^{w}=c_{\mathrm{ctr}}^{w}(\cA,\theta)$ denotes the cost
functional even when $w$ satisfies only the projected constraints below.
When $r^*>0$, every centered
feasible allocation $n$ satisfies $M(n)\succ0$ and
\begin{equation}\label{eq:centered-projection-comparison}
 x(a)^\top M(n)^{-1}x(a)
 \le(a-a^*)^\top H_n^{-1}(a-a^*)\le\Delta(a)^2/2\,.
\end{equation}
Conversely, if a finite allocation $w$ satisfies
\begin{equation}\label{eq:centered-projected-constraints}
 M(w)\succ0\,,\qquad
 x(a)^\top M(w)^{-1}x(a)\le\Delta(a)^2/2\quad(a\ne a^*)\,,
\end{equation}
then for every $\varepsilon>0$ there is a finite centered feasible
allocation $n_\varepsilon$ of cost $(1+\varepsilon)c_{\mathrm{ctr}}^{w}$.
Consequently, under sharpness,
\begin{equation}\label{eq:centered-projected-infimum}
 \cctr=\inf\{c_{\mathrm{ctr}}^{w}:w\text{ is finite and satisfies }
                         \eqref{eq:centered-projected-constraints}\}\,.
\end{equation}
If $r^*=0$, sufficient mass at $a^*$ alone gives $\cctr=0$.
\end{lemma}
\begin{proof}
The first assertion is the Schur-complement inequality, applied to
$a-a^*$, whose $U^*$ coordinates are $x(a)$.
If $a^*=0$, the constraints in \eqref{eq:centered-projected-constraints}
are already centered, so use $n_\varepsilon=(1+\varepsilon)w$.
Otherwise let $u^*=a^*/\|a^*\|$, $z(a)=(u^*)^{\top} a$, and define
\[
 V=M(w)\,,\quad v=\int x(a)z(a)w(da)\,,\quad
 s=\int z(a)^2w(da)\,,\quad \widehat v=V^{-1}v\,.
\]
Sharpness gives the uniform residual estimate
\[
 |z(a)-\|a^*\|-x(a)^\top\widehat v|
 \le R_w\Delta(a)\,,\qquad
 R_w=\kappa^{-1}\sqrt{1+\|\widehat v\|^2}\,.
\]
Set
\[
 n_\varepsilon=(1+\varepsilon)w+m_\varepsilon\delta_{a^*}\,,
 \qquad
 m_\varepsilon=
 \frac{2(1+\varepsilon)R_w^2}{\varepsilon\|a^*\|^2}>0\,.
\]
The moment matrix of $w$ is positive semidefinite, so
$s-v^\top V^{-1}v\ge0$. The added atom makes $H_{n_\varepsilon}$
positive definite. Its block inverse formula yields
\begin{align}
 (a-a^*)^\top H_{n_\varepsilon}^{-1}(a-a^*)
 &=\frac{x(a)^\top V^{-1}x(a)}{1+\varepsilon}
 +\frac{(z(a)-\|a^*\|-x(a)^\top\widehat v)^2}
 {m_\varepsilon\|a^*\|^2+(1+\varepsilon)(s-v^\top V^{-1}v)}\notag\\
 &\le\frac{\Delta(a)^2}{2(1+\varepsilon)}
       +\frac{R_w^2\Delta(a)^2}{m_\varepsilon\|a^*\|^2}
 =\frac{\Delta(a)^2}{2}\,.
 \label{eq:centered-projection-lift}
\end{align}
The extra mass has zero cost, hence
$c_{\mathrm{ctr}}^{n_\varepsilon}=(1+\varepsilon)c_{\mathrm{ctr}}^{w}$.
Projection gives one inequality in \eqref{eq:centered-projected-infimum};
lifting and then letting $\varepsilon\downarrow0$ give the other.
If $r^*=0$, then $d=1$ and $a^*\ne0$. Taking
$n=m\delta_{a^*}$ with $m(a^*)^{2}\ge2/\kappa^2$ directly satisfies the
centered constraints at zero cost.
\end{proof}

\subsubsection{Generator Transfer and Illustrative Instances}
\begin{lemma}[Exact compact-generator transfer]\label{lem:C-hull}
If known compact sets obey $\mathcal B\subseteq\cA\subseteq\Conv(\mathcal B)$, then a strategy playing only $\mathcal B$ satisfies
\[
 \sup_{a\in\cA}\Reg(a)=\sup_{b\in\mathcal B}\Reg(b)\,,\qquad
 \sup_{a\in\cA}|a^\top z|=\sup_{b\in\mathcal B}|b^\top z|\,.
\]
For a unique optimal action $a^*$ under $\theta$,
\[
 \cctr(\cA,\theta)=\cctr(\mathcal B,\theta)\,.
\]
\end{lemma}
\begin{proof}
For any $a\in\cA$, choose a finite convex representation
$a=\sum_j\lambda_j b_j$ with $b_j\in\mathcal B$, $\lambda_j\ge0$, and $\sum_j\lambda_j=1$. Regret is affine in a fixed comparator, so
$\Reg(a)=\sum_j\lambda_jR_T(b_j)\le\sup_{b\in\mathcal B}\Reg(b)$.
Similarly, for every $z\in\R^d$,
\[
 |a^\top z|
 \le\sum_j\lambda_j|b_j^\top z|
 \le\sup_{b\in\mathcal B}|b^\top z|\,.
\]
Taking the supremum over $a\in\cA$ proves one direction of both equalities; the reverse direction follows from $\mathcal B\subseteq\cA$. In particular, substituting $z=c_t^{\rm vec}$ gives equality of the per-round corruption sizes on every realized path, hence equality of their cumulative budgets. The seminorm is in fact a norm when $\mathcal B$ spans $\R^d$, because only $z=0$ is orthogonal to every generator action. The unique optimal action $a^*$ belongs to $\mathcal B$, so the two sets use the same
centered differences and gap function.

A centered feasible measure supported on $\mathcal B$ is also a measure
on $\cA$, with the same cost and the same constraints on
$\Conv(\mathcal B)=\Conv(\cA)$. Conversely, replace any finite centered
feasible measure on $\cA$ by a finite-support measure preserving the
joint moments $(aa^\top,\Delta(a),1)$. Its normalized moment integral
lies in the compact convex hull of this continuous moment map:
otherwise a separating linear functional would contradict integration
of its pointwise bound. Finite-dimensional convex reduction supplies
the required finite representation.

Replace each supported action $a$ by a finite convex combination
$a=\sum_j\lambda_jb_j$ of points in $\mathcal B$. Cost and mass are
unchanged, and
\[
 \sum_j\lambda_jb_jb_j^\top-aa^\top
 =\sum_j\lambda_j(b_j-a)(b_j-a)^\top\succeq0\,.
\]
The resulting physical matrix satisfies $H'\succeq H\succ0$.
Since $(H')^{-1}\preceq H^{-1}$, all centered constraints remain valid.
Taking infima proves the centered equality, including the case of an
empty feasible set.
\end{proof}

\begin{proposition}[Finite complexity without a separated generator]\label{prop:C-eight}
Let $a(s)=(-1/2+s/4,(s+s^2)/4)$, $0\le s\le1$, and $\theta=(1,0)$. Then $\cA=\{a(s)\}$ is compact, spans $\R^2$, lies in $\BB$, and has the unique optimal action $a(0)$, zero minimum positive gap, and $\cctr=8$. Every compact generator of its convex hull contains suboptimal points approaching $a(0)$.
\end{proposition}
\begin{proof}
Here $\Delta(s)=s/4$ and $x(s)=(s+s^2)/4$. Since
$\|a(s)-a(0)\|\le\sqrt5s/4$, the instance is sharp.
For every centered feasible allocation, \Cref{lem:centered-projection-lift}
gives $M\ge2\sup_{s>0}(x(s)/\Delta(s))^2=8$.
Also $x(s)^2/\Delta(s)=s(1+s)^2/4\le1$, so $c_{\mathrm{ctr}}^{n}\ge M\ge8$.
The allocation $w=32\delta_{a(1)}$ satisfies the projected constraints
with $M(w)=c_{\mathrm{ctr}}^{w}=8$.
Lifting this allocation gives centered feasible costs $8(1+\varepsilon)$
for every $\varepsilon>0$. Thus $\cctr=8$; attainment by a centered
allocation is not required.
For $s_0\in(0,1)$, the linear functional with coefficient $(-1-2s_0,1)$ equals a constant plus $(s-s_0)^2/4$ on the curve. It uniquely exposes $a(s_0)$ as a minimizer. A finite convex representation of a uniquely exposed point has every positively weighted point equal to that point. Compactness of a generator implies its convex hull is closed, so it must contain each $a(s_0)$. Letting $s_0\downarrow0$ proves the claim.
\end{proof}

\Cref{prop:C-eight} excludes a separated generator on the curve,
although the optimal action $a(0)$ under $\theta$ is sharp and covered by
\Cref{thm:main-nonseparated-upper}. \Cref{sec:SH-example}
gives an explicit centered sampling allocation for this instance.
Sharpness also implies stability of the comparator: if $\norm z<\varsigma$,
then $\Delta(a)+(a-a^*)^\top z>0$ for every $a\ne a^*$.
Equivalence of the action and Euclidean norms makes
$\Omega_\theta(\epsilon)=0$ for all sufficiently small $\epsilon$.

\subsection{Moment Estimates for Affine Functions on Compact Domains}\label{app:common-moments}

\subsubsection{Convex-Body Integral Estimates Replacing a Finite Union Bound}
Let $D$ be a nonempty compact convex set of affine dimension $r$, and let $\sigma_D$ be normalized relative Lebesgue measure (point mass for $r=0$).
Within \Cref{lem:C-affine}, the absolute and positive-part maximum
inequalities convert integral moments into uniform bounds on affine functions. The two
curvature inequalities close the one-step moment recurrences despite
affine shifts. The inverse-moment inequality converts local inverse-moment
control into a uniform lower bound on estimated gaps. Their role is to
avoid treating an uncountable action set as a finite union of bad events.

\begin{lemma}[Uniform estimates for affine functions]\label{lem:C-affine}
For affine $f:D\to\R$, $p\ge\max\{2,r\}$, $R=\max_D|f|$, $M=\max_D f_+$, and $F_p=\int f_+^p d\sigma_D$,
\begin{align}
 R^p&\le2^p4^r\int|f|^p d\sigma_D\,,\label{eq:C-aff-abs}\\
 M^{p+r}&\le2^p4^rR^rF_p\,,\label{eq:C-aff-neg}\\
 M^2\int f_+^{p-2}d\sigma_D&\le16F_p\,, \notag\\
 \int(f+c)_+^{p-2}d\sigma_D&\le16(1+c/M)^{p-2+r}F_p/M^2\quad(c\ge0,M>0)\,.\label{eq:C-aff-shift}
\end{align}
If $f>0$ and $p>r$, with $m=\min_D f$, $R=\max_D f$, then
\begin{equation}\label{eq:C-aff-inverse}
 m^{-(p-r)}\le2^pR^r\int f^{-p}d\sigma_D\,.
\end{equation}
The third inequality is trivial when $M=0$.
\end{lemma}
\begin{proof}
For the first inequality, change sign if needed and choose $f(x_0)=R$. The homothetic image $x_0+(D-x_0)/4$ has relative volume $4^{-r}$ and function values at least $R/2$. For the second, take $f(x_0)=M>0$ and contract by $\lambda=M/[2(M+R)]\ge M/(4R)$. The image has values at least $M/2$, which proves the bound. Zero maxima are immediate.

For the third, contract $D_+=D\cap\{f\ge0\}$ toward a maximizing point by $r/(p+r)$. For $r>0$, this proves
\[
 F_p\ge c_{p,r}M^p\sigma_D(D_+)\,,\qquad
 c_{p,r}=\left(\frac p{p+r}\right)^p\left(\frac r{p+r}\right)^r\,.
\]
H\"older gives $\int f_+^{p-2}\le\sigma_D(D_+)^{2/p}F_p^{1-2/p}$. Moreover
$c_{p,r}^{-2/p}=(1+r/p)^2(1+p/r)^{2r/p}\le16$: the first factor is at most four and $u\log(1+1/u)\le\log2$ for $0<u\le1$. This proves the third inequality; $r=0$ is immediate.

For the fourth, contract $D$ toward a maximum point by $\lambda=M/(M+c)$. Then $f(x_0+\lambda(x-x_0))=\lambda(f(x)+c)$. Change variables and use inclusion of the image in $D$, followed by the third inequality. Finally, for a positive affine function contract toward a minimum point by $m/R$. On that image $f\le2m$, and its relative volume is $(m/R)^r$. Integrating $f^{-p}$ and rearranging proves the last assertion.
\end{proof}

\subsection{Finite Time Computing of Reference Action $r_t$}\label{app:sec:algo2:reference-action-computing}
In this section, we show that specifying the reference action $r_t$ in \eqref{eq:SH-score} need not exhaust the possibly infinitely large $J$.
To see this, define $ R=\left\|\sum_{s=1}^{t-1}\widehat\theta_s\right\|_2$ and $\Psi_t(j)=\widehat L_{t-1,b_j}+\beta_t h_j$.
Because $b_j,b_1\in\mathcal B\subseteq\BB$ and $h_j\ge j+1$,
\begin{align*}
 \bigl(\widehat L_{t-1,b_j}-\widehat L_{t-1,b_1}\bigr)
   &\ge-\|b_j-b_1\|_2R\ge-2R\,,\\
 \Psi_t(j)-\Psi_t(1)
   &=\widehat L_{t-1,b_j}-\widehat L_{t-1,b_1}+\beta_t(h_j-h_1)\\
   &\ge-2R+\beta_t(j+1-h_1)\,.
\end{align*}
Hence every index with $j+1>h_1+2R/\beta_t$ has strictly larger
objective than candidate $1$ and cannot be a minimizer. This argument
uses the lower bound $h_j\ge j+1$, not monotonicity of the priorities.
Since $h_1=2$,
define the finite integer
\begin{equation}\label{eq:SH-reference-cutoff}
 N_t=\left\lfloor h_1-1+\frac{2R}{\beta_t}\right\rfloor
     =\left\lfloor1+\frac{2\left\|\sum_{s=1}^{t-1}\widehat\theta_s\right\|_2}{\beta_t}\right\rfloor
     \ge1\,.
\end{equation}
Then $r_t$ is the least-index minimizer of $\Psi_t$ over
$J\cap\{1,\ldots,N_t\}$. We retain the boundary index so that
possible ties are handled by the original rule. At $t=1$, the empty sum gives
$R=0$, hence $N_1=1$ and $r_1=1$.

This finite set can be generated without first constructing all of
$J$. Enumerate just the first $N_t$ rational directions in their
fixed order, solve their geometric minimizations, and retain only
the first occurrence of each selected action. For a retained index
$j\le N_t$, all terms defining $h_j$ involve retained indices $i<j$,
which have already been processed. Compute these finite sums and
compare $\Psi_t(j)$. Previously generated candidates and weights can
be cached. Assuming these operations can be performed exactly, this takes
at most $N_t$ directional minimizations and finitely many distance
evaluations and comparisons. No unknown gap, optimal action $a^*$, or horizon
enters the cutoff.

The cutoff depends on the observed history and need not give a small
or polynomial complexity bound. It restricts only the reference
search: the fixed prior $\mu$, the FTRL law $q_t$, and the exploration
program continue to use their full domains.

\end{document}